\documentclass[10pt,twocolumn,letterpaper]{article}

\usepackage{cvpr}              % To produce the CAMERA-READY version
\usepackage{microtype}

\renewcommand{\paragraph}[1]{\vspace{.0em}\noindent\textbf{#1}}

\usepackage{soul}
\setuldepth{foobar}

\definecolor{cvprblue}{rgb}{0.21,0.49,0.74}
\usepackage[pagebackref,breaklinks,colorlinks,allcolors=cvprblue]{hyperref}

\usepackage{multirow}

\usepackage[accsupp]{axessibility}

\usepackage{amssymb}
\usepackage{pifont}
\newcommand{\cmark}{\ding{51}}
\newcommand{\xmark}{\ding{55}}

\newcommand{\algoname}{\textsc{ModMap}}
\usepackage{amsmath}
\DeclareMathOperator{\mean}{mean}
\DeclareMathOperator{\product}{prod}

\usepackage{algorithm}
\usepackage{algpseudocode}
\algrenewcommand\algorithmicrequire{\textbf{Input:}}
\algrenewcommand\algorithmicensure{\textbf{Output:}}

\definecolor{custom_green}{RGB}{130,179,102}
\definecolor{custom_red}{RGB}{184,84,80}
\definecolor{custom_blue}{RGB}{108,142,191}

\def\confName{CVPR}
\def\confYear{2026}

\title{Modulate-and-Map: Crossmodal Feature Mapping with Cross-View Modulation for 3D Anomaly Detection}

\author{ 
    Alex Costanzino\textsuperscript{1} \hspace{0.5cm} 
    Pierluigi Zama Ramirez\textsuperscript{2} \hspace{0.5cm} 
    Giuseppe Lisanti\textsuperscript{1} \hspace{0.5cm} 
    Luigi Di Stefano\textsuperscript{1} \\
    \small \textsuperscript{1}CVLab, University of Bologna \hspace{0.5cm} \textsuperscript{2}Ca' Foscari University of Venice \\
    \normalsize \url{https://alex-costanzino.github.io/modmap/}
}

\begin{document}
\maketitle

\begin{abstract}
    We present \algoname{}, a natively multiview and multimodal framework for 3D anomaly detection and segmentation. 
    Unlike existing methods that process views independently, our method draws inspiration from the crossmodal feature mapping paradigm to learn to map features across both modalities and views, while explicitly modelling view-dependent relationships through feature-wise modulation.
    We introduce a cross-view training strategy that leverages all possible view combinations, enabling effective anomaly scoring through multiview ensembling and aggregation.
    To process high-resolution 3D data, we train and publicly release a foundational depth encoder tailored to industrial datasets.  
    Experiments on SiM3D, a recent benchmark that introduces the first multiview and multimodal setup for 3D anomaly detection and segmentation, demonstrate that \algoname{} attains state-of-the-art performance by surpassing previous methods by wide margins.
\end{abstract}    
\section{Introduction}
\label{sec:introduction}
    \begin{figure}[t]
    \centering
    \resizebox{\linewidth}{!}{
        \setlength{\tabcolsep}{2px}
        \begin{tabular}{ccc}
            %& $v_{i}$ & $v_{j}$ \\

            %\rotatebox{90}{\hspace{1.25em} Image} &
            \includegraphics[width=0.33\linewidth]{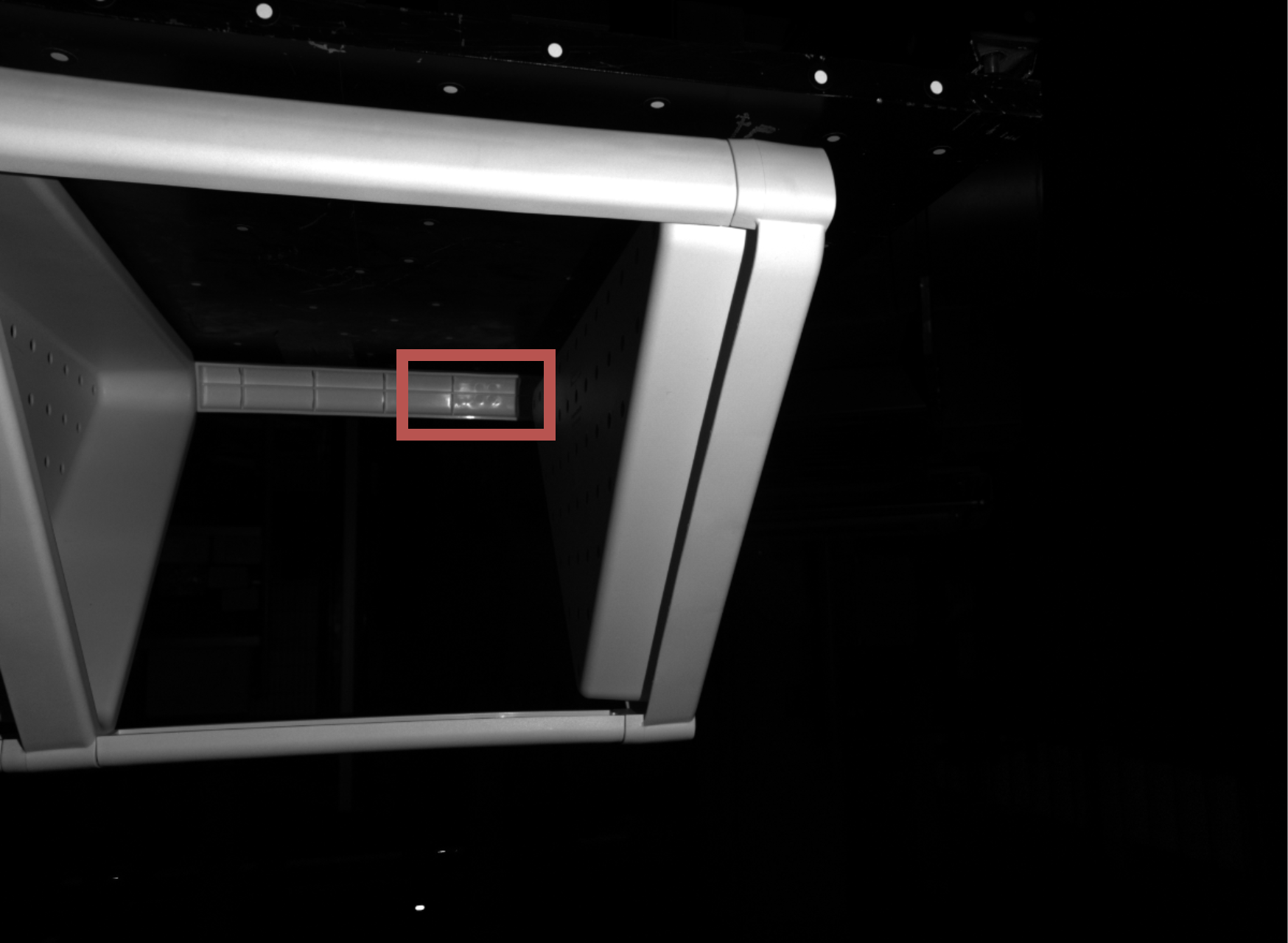} &
            \includegraphics[width=0.33\linewidth]{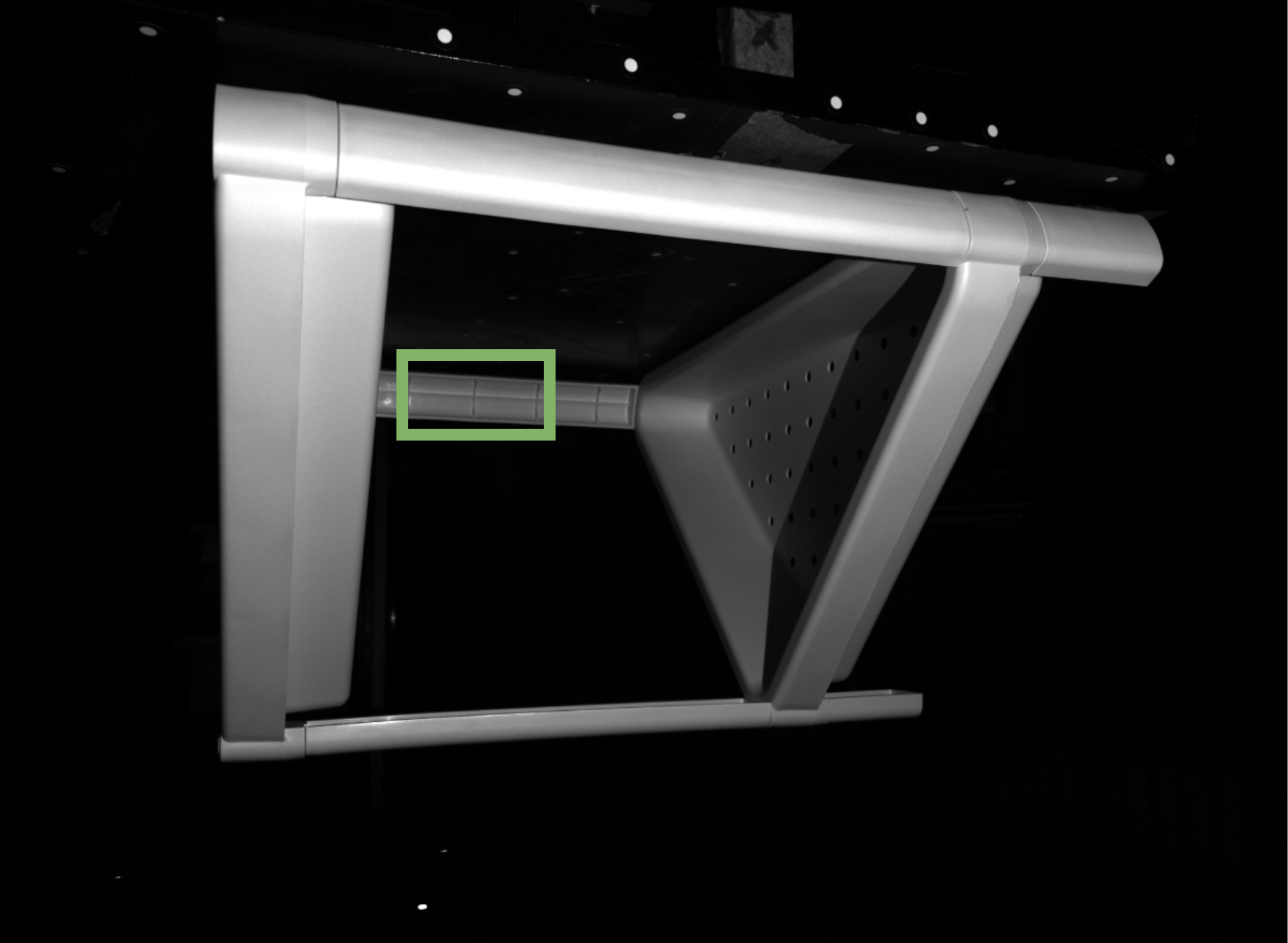} \\

            %\rotatebox{90}{\hspace{1.25em} Depth} &
            \includegraphics[width=0.33\linewidth]{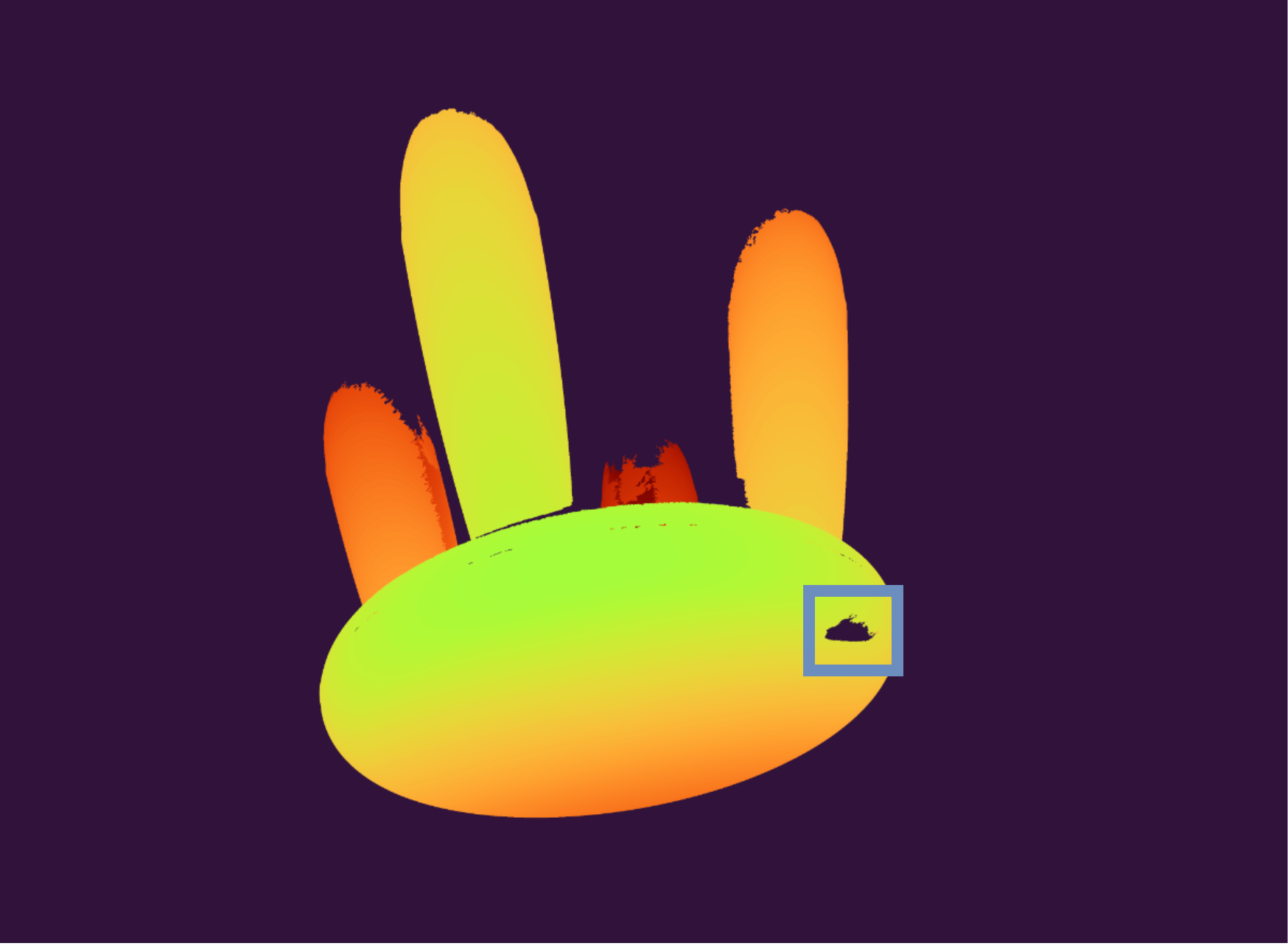} &
            \includegraphics[width=0.33\linewidth]{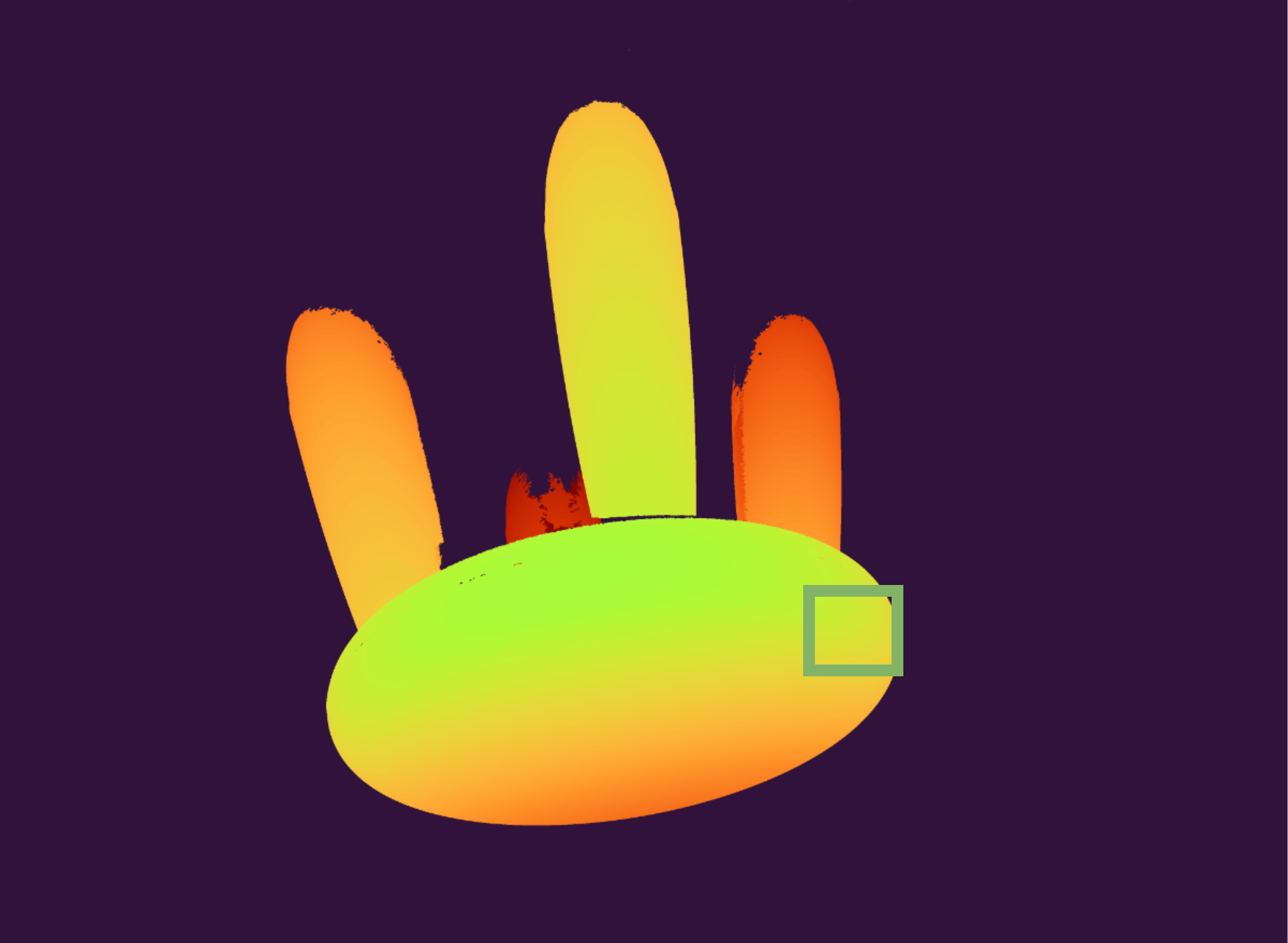} \\
        \end{tabular}
    }
    \caption{
        \textbf{View-dependent Artefacts.}
        The first column shows acquisition artefacts observed in an image (top row: specular highlights, \textcolor{custom_red}{\textbf{red box}}) and a depth map (bottom row: missing depths, \textcolor{custom_blue}{\textbf{blue box}}), dealing with two objects of SiM3D. 
        As shown in the right column, other views of the same object are not affected by artefacts at the positions (\textcolor{custom_green}{\textbf{green boxes}}) corresponding to those highlighted in the left column. 
        }
    \label{fig:teaser}
\end{figure}

    The introduction of benchmark datasets such as MVTec AD~\cite{bergmann2019mvtec} has significantly accelerated research on \emph{unsupervised} anomaly detection and segmentation (ADS), where the goal is to detect and localise defects by training solely on nominal samples. 
    Most approaches~\cite{roth2022patchcore,Cohen2020SubImageAD,defard2021padim,bergmann2018improving,wyatt2022anoddpm,yang2020dfr,bergmann2020uninformed,wang2021student_teacher,batzner2024efficientad,rudolph2021same,gudovskiy2022cflow,sohn2021learning,yi2020patch,li2021cutpaste} focus on 2D anomaly detection, which consists of processing a single image of the inspected object to produce a 2D anomaly map.
    While effective for many applications, this paradigm has several limitations. 
    First, geometric defects -- such as dents, deformations, or missing parts -- may not be clearly visible in RGB images alone, especially under diffuse lighting.
    Second, 2D anomaly maps do not allow precise localisation of defects in 3D, which is necessary for efficient and possibly automated, additional assessment and rework of high-value products.
    To address these limitations, more recent benchmarks~\cite{bergmann20223dad, bonfiglioli2022eyecandies} provide multimodal data, namely RGB images along with pixel-registered 3D information, with methods leveraging both modalities yielding improved performance~\cite{horwitz2023btf,wang2023m3dm,costanzino2024cfm} due to the ability to perceive both colour and geometric anomalies.
    However, these benchmarks consider a setup where an object is captured from a single viewpoint, which prevents comprehensive inspection when defects may occur over the entire surface. 
    Besides, even partial surface scans may require multiple high-resolution views to detect subtle anomalies. 
    Eventually, in~\cite{bergmann20223dad, bonfiglioli2022eyecandies} the task is still cast as 2D anomaly detection, and, hence, it does not assess the ability of methods to localise defects in the 3D space precisely.

    The recently introduced SiM3D benchmark~\cite{costanzino2025sim3d} features the first dataset for multiview multimodal 3D anomaly detection, thereby addressing the limitations highlighted above.
    Each object is scanned through multiple (12 to 36) multimodal views captured from vantage points designed to ensure comprehensive surface coverage, each view including an image along with a pixel-aligned depth map. 
    Unlike previous benchmarks, the task consists of producing a 3D anomaly volume: a voxel grid where each voxel carries an anomaly score.
    As reported in~\cite{costanzino2025sim3d}, existing multimodal ADS methods~\cite{rudolph2023ast, horwitz2023btf, wang2023m3dm, costanzino2024cfm} face significant challenges when naively adapted to the multiview scenario due to the lack of mechanisms designed to deploy synergistically multimodal cues gathered from different viewpoints. 
    Instead, one may argue that the co-occurrence of multimodal cues across the views observed in the training data may help establish a more robust, holistic model of the nominal samples compared to learning from individual views in isolation.

    In this work, we show how the Crossmodal Feature Mapping (CFM) paradigm, introduced in~\cite{costanzino2024cfm} to tackle single-view anomaly detection, can be extended to obtain an inherently multiview approach and face the challenges set forth by the 3D Anomaly Detection task proposed by SiM3D. 
    CFM relies on the intuition that, given a multimodal view of an object, such as, e.g., an image and depth map, an anomaly manifests itself as an unlikely co-occurrence of the cues -- namely the features -- observed in the two modalities at corresponding positions. 
    Hence, CFM learns the co-occurrence of features in nominal data by training two neural networks to predict image features from depth features and vice versa. 
    Then, at inference time, it computes per-pixel anomaly scores based on the discrepancy between predicted and observed features. 
    In this paper, we reckon that learning the crossmodal mappings not only within a view but also across views may help robustly handle view-dependent acquisition artefacts, such as specular highlights in images or missing measurements in depth maps (\cref{fig:teaser}, left column).
    More in detail, we extend CFM's crossmodal mapping model so that the two neural networks are trained to predict not only a given feature (either an image feature or a depth feature) based on that observed in the other modality in the same view, but also based on those observed in the other modality in all other views.  
    In this way, if at inference time a feature in a view is corrupted by an acquisition artefact unseen at training time, the correct corresponding feature in the other modality may be predicted from another view not affected by the artefact (\cref{fig:teaser}, right column).
    Based on this observation, at inference time we compare the observed feature in a view to those predicted from the other modality from all views, and compute the anomaly score based on the closest prediction. 
    Therefore, a feature in a modality can be deemed as nominal if at least one crossmodal mapping from all views can predict it as such. 
    This cross-view and crossmodal feature mapping mechanism is designed to synergistically leverage multimodal cues obtained from all viewpoints, with the goal of maximising robustness to acquisition artefacts. 
    As such, it is a mechanism that inherently trades recall for precision: to counterbalance the potential loss of sensitivity, when computing the final 3D anomaly segmentation and detection outputs, we adopt an aggregation strategy that sifts out the strongest anomaly score, increasing the final recall of our approach.

    Thus, the main contribution of this paper is the introduction of the first \emph{natively multiview} multimodal algorithm for 3D anomaly detection, specifically designed to synergistically exploit crossmodal relationships across multiple viewpoints rather than processing views in isolation. 
    Several key technical novelties are instrumental in realising this contribution.
    We propose a cross-view training strategy that enables effective multiview learning. 
    During training, we consider all pairs of source and target views and, for both modalities, learn to predict target features from source ones at corresponding positions.
    Crucially, to prevent one-to-many mapping ambiguities, we introduce conditioning on view pairs: an additional network \emph{modulates} the source features, conditioned on both the source and target views, before passing them to the main \emph{mapping} network.
    Accordingly, we dub our method Modulate-and-Map (\algoname{}).
    Previous multimodal methods~\cite{rudolph2023ast, horwitz2023btf, wang2023m3dm, costanzino2024cfm} rely on point cloud encoders designed for low-resolution inputs (8K points). 
    Conversely, to enable processing of high-resolution depth maps, we deploy multiple industrial datasets and train a \emph{depth foundational model} using self-supervised learning. 
    Our depth encoder can handle high-resolution 3D data (5-7M points) without aggressive downsampling, and provides depth features that are pixel-aligned to image features. 
    Extensive experiments demonstrate that \algoname{} achieves state-of-the-art performance on SiM3D, the recent benchmark designed to address multiview and multimodal 3D anomaly detection.
\section{Related Work}
\label{sec:related_work}

    \paragraph{Anomaly Detection Benchmarks.}
        The standardisation of evaluation protocols through MVTec AD~\cite{bergmann2019mvtec} catalysed rapid progress in anomaly detection, fostering the development of numerous methods and subsequent specialised benchmarks.
        Subsequent benchmarks expanded the scope of ADS to address various challenges: MVTec LOCO~\cite{mvtec_loco} introduced logical anomalies, VisA~\cite{visa} provided high-resolution images of complex scenes, and PAD~\cite{zhou2023pad} addressed pose-agnostic detection. 
        Real-IAD~\cite{Wang_2024_CVPR_real_iad} recently introduced a large-scale multiview dataset with RGB images, though it still focuses on 2D anomaly maps.
        To address the limitations of image-only approaches, several benchmarks have incorporated 3D information. 
        MVTec 3D-AD~\cite{bergmann20223dad} provides RGB images with pixel-aligned XYZ coordinates captured by structured light sensors, while Eyecandies~\cite{bonfiglioli2022eyecandies} offers synthetic data with pixel-aligned depth and normal maps. 
        Real3D-AD~\cite{liu2023real3d} introduced the first point cloud anomaly detection benchmark, using front-and-back scans for training and single-view point clouds for testing. 
        However, the above-described benchmarks still evaluate methods on single-view inputs and yield 2D anomaly maps.
        SiM3D~\cite{costanzino2025sim3d} addresses these limitations by introducing the first benchmark for multiview 3D anomaly detection, where the task is to produce a 3D anomaly volume by integrating information from multiple views. 
        Moreover, it focuses on the challenging single-instance scenario, where only one nominal object -- either real or synthetic -- is available for training, closely resembling real industrial scenarios.

    \paragraph{Multimodal Anomaly Detection.}
        Multimodal single-view benchmarks have fostered the introduction of approaches that effectively integrate different modalities (e.g., RGB images and 3D data) to perform 2D anomaly detection.
        BTF~\cite{horwitz2023btf} extended PatchCore~\cite{roth2022patchcore}, a solution based on memory banks, to multimodal input by combining 2D features from pre-trained CNNs with hand-crafted 3D features (FPFH~\cite{fpfh}). 
        M3DM~\cite{wang2023m3dm} improved upon BTF by using Transformer-based backbones (DINO-v1~\cite{caron2021emerging} for RGB and Point-MAE~\cite{pang2022masked} for point clouds) and learning a fusion function to combine modality-specific features. 
        AST~\cite{rudolph2023ast} adopted a teacher-student paradigm using normalising flows, processing depth information as additional input channels rather than extracting explicit 3D features. 
        EasyNet~\cite{chen2023easynet} introduced a lightweight architecture combining multiple feature extraction strategies.
        More recently, CFM~\cite{costanzino2024cfm} proposed learning crossmodal feature mappings between 2D and 3D features on nominal samples, detecting anomalies by identifying inconsistencies between predicted and observed features. 
        While these methods achieve strong performance on single-view multimodal benchmarks, they face significant challenges when extended to multiview scenarios. 
        Memory bank methods suffer from computational overhead that increases with the number of views, while reconstruction-based approaches process views independently without leveraging geometric consistency. 
        In addition, solutions based on pre-trained point cloud feature extractors, such as~\cite{wang2023m3dm,costanzino2024cfm}, are unable to take full advantage of high-resolution 3D data (e.g., SiM3D provides point clouds of 5–7 million points) due to the severe downsampling performed by~\cite{pang2022masked} (i.e., 2048 points), which limits their ability to detect small geometric defects.
        Our work extends the crossmodal feature mapping paradigm to the multiview settings by introducing view-conditioned modulation and cross-view training, along with a pre-trained depth encoder, enabling efficient and robust multiview anomaly detection.
\section{Method}
\label{sec:method}
    \begin{figure*}[t]
    \centering
        \includegraphics[width=0.94\linewidth]{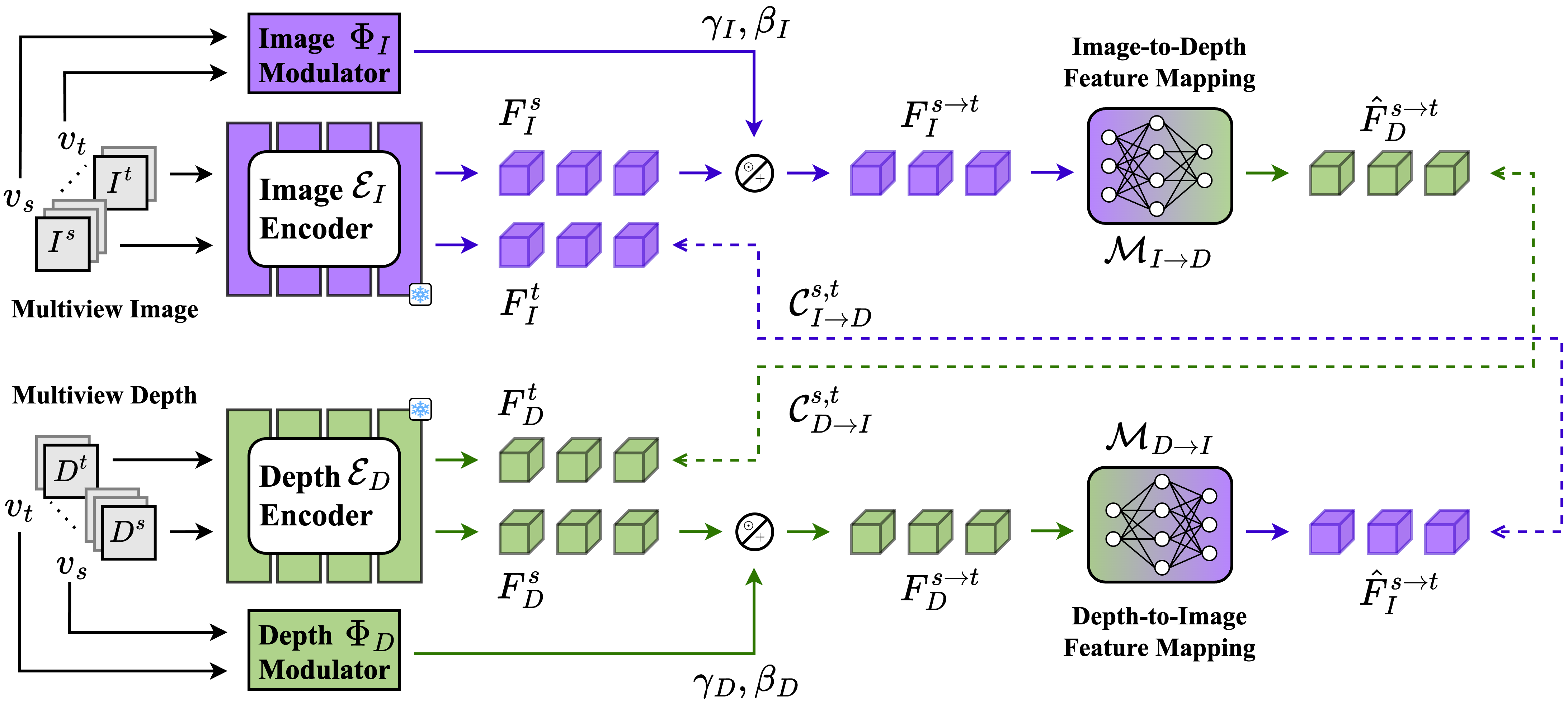}
    \caption{\textbf{\algoname{} Training.}
    Starting from the set of images $I$ and depths $D$ from a training sample, we select a source view $s$ and target view $t$, and forward their images $I^s, I^t$ and depths $D^s, D^t$ to the image, $\mathcal{E}_I$, and depth, $\mathcal{E}_D$, encoders, respectively, so as to compute modality-specific features, $F^s_I, F^t_I$ and $F^s_D, F^t_D$. 
    Moreover, the one-hot encodings of the view indexes, $v_s$ and $v_t$, are fed into the feature modulators, $\Phi_I$ and $\Phi_D$, to generate modality-specific scale-and-shift parameters, $\gamma_I, \beta_I$ and $\gamma_D, \beta_D$.
    Then, for both modalities, the source features are scale-and-shifted to obtain modulated source features that incorporate view conditioning: $F^{s \rightarrow t}_I, F^{s \rightarrow t}_D$.  
    The modulated features are passed as inputs to the mapping networks $\mathcal{M}_{I \rightarrow D}, \mathcal{M}_{D \rightarrow I}$ that predict the corresponding features from the other modality $\hat{F}^{s \rightarrow t}_{D}, \hat{F}^{s \rightarrow t}_{I}$. 
    The predicted features are then compared to the actual target features $F^t_D, F^t_I$ to optimise both the modulators and the mapping networks.
    }
    \label{fig:pipeline}
\end{figure*}

    \subsection{Task Definition}
        Our method addresses the multiview multimodal 3D anomaly detection task introduced by the recent SiM3D benchmark~\cite{costanzino2025sim3d}. 
        Specifically, the inputs consist of a set of images $\{I^i\}_{i=1}^n$ and corresponding 3D data $\{D^i\}_{i=1}^n$, captured from $n$ viewpoints of an object instance. 
        The viewpoints remain consistent across different instances of the same object category. 
        In our approach, the 3D data are represented as depth maps, enabling efficient high-resolution processing. 
        For each object instance, the detection task involves predicting a global anomaly score. 
        In contrast, the segmentation task aims to estimate an anomaly volume $\Omega \in \mathbb{R}^{X \times Y \times Z}$, where $X$, $Y$, and $Z$ denote the grid dimensions, and each voxel encodes an anomaly score.

    \subsection{Cross-View Crossmodal Feature Mapping}
        \label{subsec:modmap}
        Our approach draws inspiration from CFM~\cite{costanzino2024cfm} by learning mappings that predict features from one modality to another on nominal samples. 
        At inference time, inconsistencies between the predicted and actual features highlight anomalies. 
        We extend such a mechanism by leveraging multiview inputs to generate more accurate anomaly maps and reduce false positives caused by acquisition artefacts.
        In particular, we propose a Modulate-and-Map framework, \algoname{}, capable of mapping features not only across modalities but also across views. 

    \subsubsection{\algoname{} Architecture}
        \algoname{} comprises six main components described in the following sections: an image feature extractor, a depth feature extractor, two modulators and two mapping networks (\cref{fig:pipeline}).

        \paragraph{Image Feature Extractor.}
            We employ a frozen DINO-v2~\cite{oquab2023dinov2} encoder, denoted as $\mathcal{E}_{I}$, as our image feature extractor.
            Given an input image $I^i \in \mathbb{R}^{H \times W}$ from view $i$, $\mathcal{E}_{I}$ produces a feature map:
            \begin{equation}
                F_{I}^i = \mathcal{E}_{I}(I^i) \in \mathbb{R}^{h \times w \times c_{I}}
            \end{equation}
            where $h = H/p$ and $w = W/p$ correspond to the spatial resolution of the feature map given the patch size $p$, and $c_{I}$ denotes the number of feature channels.
            Finally, the feature map $F_{I}^i$ is unrolled into $h \cdot w$ elements of dimension $c_{I}$ for subsequent processing.
    
        \paragraph{Depth Feature Extractor.}
            Existing pre-trained point-cloud feature extractors (e.g.,~Point-MAE~\cite{pang2022masked}) struggle to handle the high-resolution 3D data provided in SiM3D~\cite{costanzino2025sim3d, guo2020deep}, which contain 5-7M points per sample.
            This limitation mainly arises from the unstructured nature of point clouds, which makes it computationally expensive to apply local operations and efficiently extract fine-grained geometric features.
            To overcome this issue, we operate on depth maps, which allow efficient high-resolution processing thanks to their structured, image-like representation~\cite{costanzino2025sim3d}.
            Since no pre-trained depth feature extractors do exist, we train a dedicated encoder $\mathcal{E}_{D}$ based on a Vision Transformer, adapted to single-channel depth inputs.
            The encoder is trained from scratch following a self-supervised protocol inspired by DINO-v2, with one key difference: only spatial augmentations are applied during training, including random rotations, horizontal and vertical flips, and random crops with scaling.
            Photometric augmentations (e.g., brightness, contrast, or colour jittering) are deliberately excluded, as they would corrupt the information encoded in depth maps, where pixel intensities correspond to distance values.
            Beyond efficiency, depth processing provides several advantages over point-cloud processing.
            First, it produces features that are pixel-aligned with images and architecturally coherent. 
            Indeed, by employing the same Vision Transformer design with shared normalisation layers, positional encodings, and self-attention mechanisms, the depth encoder learns representations that naturally align with image features in both spatial structure and semantic space.
            Second, it allows leveraging multiple industrial datasets during training, yielding representations specifically tailored for anomaly detection tasks.
            In particular, $\mathcal{E}_{D}$ is trained on depth maps from various multimodal industrial anomaly detection datasets, including MVTec 3D-AD~\cite{bergmann20223dad}, Eyecandies~\cite{bonfiglioli2022eyecandies}, and Real-IAD D\textsuperscript{3}~\cite{wang2024real}.
            We further augment the training set with depth maps obtained from a monocular depth foundation model, DepthAnything-v2~\cite{depth_anything_v2}, on 2D anomaly detection datasets such as MVTec AD~\cite{bergmann2019mvtec}, MVTec LOCO~\cite{mvtec_loco}, MVTec AD 2~\cite{heckler2025mvtecad2}, and ViSA~\cite{visa}.
            After this pre-training, the encoder is frozen and integrated into our pipeline. 
            Since SiM3D -- the dataset used in our experimental evaluation -- is not included in the training data, our depth encoder is regarded as an off-the-shelf foundation model, similar to DINO-v2, within our framework.

            Given a depth map $D^i \in \mathbb{R}^{H \times W}$ from view $i$, $\mathcal{E}_{D}$ produces a depth feature map:
            \begin{equation}
                F_{D}^i = \mathcal{E}_{D}(D^i) \in \mathbb{R}^{h \times w \times c_{D}}
            \end{equation}
            where $h = H/p$ and $w = W/p$ correspond to the spatial resolution of the feature map given the patch size $p$, and $c_{D}$ denotes the number of feature channels.
            Finally, the feature map $F_{D}^i$ is unrolled into $h \cdot w$ elements of dimension $c_{D}$ for subsequent processing.
    
        \paragraph{Modulate-and-Map.}
            The proposed Modulate-and-Map strategy, depicted in~\cref{fig:pipeline}, transforms features, $F^s_m$, of one modality, $m \in \{I, D\}$, obtained from a source viewpoint, $s \in \{1, .., N\}$, to features, $F^t_n$, of the other modality, $n \in \{I, D\}$, with $m \neq n$, of a target viewpoint, $t \in \{1, .., N\}$.
    
            First, for each view $i$, we encode its identity as a one-hot vector $v_i \in \{0,1\}^N$, where $N$ is the total number of views.
            Given a source view code, $v_s$, and a target view code, $v_t$, we use a modulator $\Phi$ to condition $F^s$ based on the desired source to target mapping:
            \begin{equation}
                F^{s \rightarrow t} = \Phi(F^s, v_s, v_t)
            \end{equation}
            The modulator, $\Phi$, inspired by FiLM~\cite{perez2018film}, is implemented as a lightweight MLP that takes as input the concatenation of the source and target view codes and produces scale-and-shift parameters:
            \begin{align}
                [\gamma, \beta] &= \text{MLP}([v_s; v_t])
                \\
                \Phi(F^s, v_s, v_t) &= \gamma \odot F^s + \beta
                \end{align}
            where $\gamma, \beta \in \mathbb{R}^{d}$ are the modulation parameters initialised at $\gamma=\bf{1}$ and $\beta=\bf{0}$, $\odot$ denotes element-wise multiplication, and $d \in \{c_I, c_D\}$ is the corresponding feature dimension for each modality.
            We employ feature-wise linear modulations, which enable view-dependent feature adaptation via learnable affine transformations while preserving the geometric structure of pre-trained feature spaces. 
            This preservation is crucial as the mapping networks establish correspondences between image and depth features based on semantic similarity (e.g., edge of hole, flat surface), and these relationships are encoded in the spatial organisation of the feature space. 
            Indeed, disrupting such a structure would prevent learning consistent crossmodal mappings. 
            Moreover, identity initialisation provides a natural starting point, allowing the network to smoothly learn view-specific adaptations without disrupting the rich semantic content encoded by the frozen pre-trained encoders.
    
            Then, we employ two lightweight MLP-based mapping networks that operate feature-wise: 
            $\mathcal{M}_{I \rightarrow D}: \mathbb{R}^{c_{I}} \rightarrow \mathbb{R}^{c_{D}}$ maps image features to depth features, and $\mathcal{M}_{D \rightarrow I}: \mathbb{R}^{c_{D}} \rightarrow \mathbb{R}^{c_{I}}$ performs the opposite mapping.
            Given modulated features from source view $s$ to target view $t$, the mapped features are:
            \begin{align}
                \hat{F}_{D}^{s \rightarrow t} &= \mathcal{M}_{I \rightarrow D} \left( \Phi_{I}(F_{I}^s, v_s, v_t) \right)
                \\
                \hat{F}_{I}^{s \rightarrow t} &= \mathcal{M}_{D \rightarrow I} \left( \Phi_{D}(F_{D}^s, v_s, v_t) \right)
            \end{align}
    
    \subsubsection{\algoname{} Training}
        As introduced in \cref{sec:introduction}, we realise a natively multiview approach robust to acquisition artefacts by training the mapping networks to predict crossmodal mappings for all possible source-target view pairs. 
        Moreover, this enables us to vastly augment the training samples, which may help prevent overfitting in data scarcity regimes, as is indeed the case of SiM3D, which ships only one training instance, either real or synthetic, per object category.    
        Thus, as shown in~\cref{fig:pipeline}, for a training sample with $N$ views, we process all $N \times N$ source-target view pairs.
        For each pair $(s,t)$, we optimise cosine distances between predicted and actual features:
        \begin{align}
            \mathcal{C}_{I \rightarrow D}^{s,t} &= 1 - \frac{\hat{F}_{D}^{s \rightarrow t} \cdot F_{D}^t}{\|\hat{F}_{D}^{s \rightarrow t}\| \|F_{D}^t\|}
            \\
            \mathcal{C}_{D \rightarrow I}^{s,t} &= 1 - \frac{\hat{F}_{I}^{s \rightarrow t} \cdot F_{I}^t}{\|\hat{F}_{I}^{s \rightarrow t}\| \|F_{I}^t\|}
        \end{align}    
        The overall loss is defined as the sum of the two terms:
        \begin{equation}
            \mathcal{L} = \mathcal{C}_{I \rightarrow D}^{s,t} + \mathcal{C}_{D \rightarrow I}^{s,t}
        \end{equation}

    \begin{figure}[t]
    \centering
        \includegraphics[width=\linewidth]{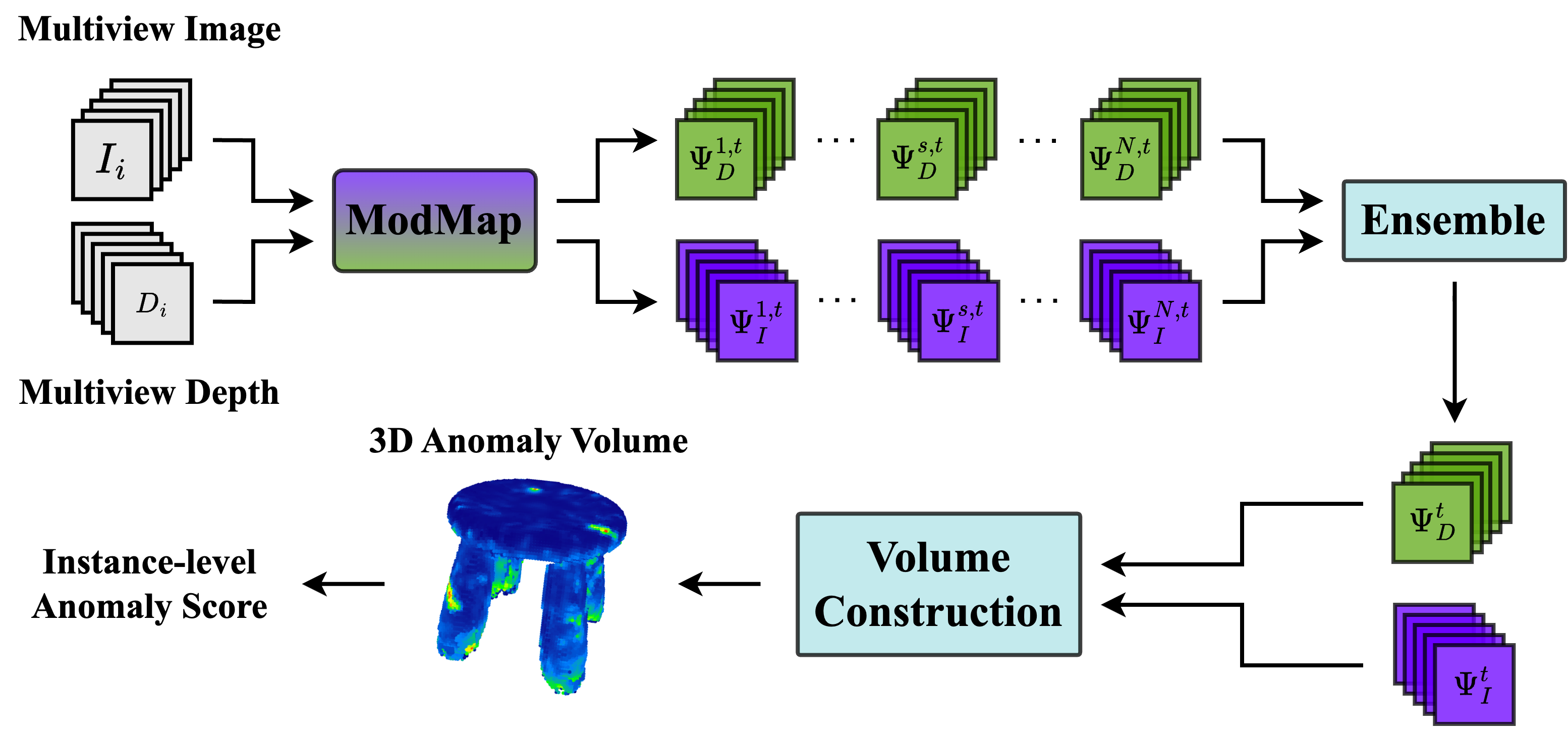}
    \caption{
        \textbf{\algoname{} Inference.}
        We process the set of $N$ images $I^i$ and $N$ depths $D^i$, obtaining $N \times N$ anomaly maps for each of the two modalities. 
        We ensemble the anomaly scores into $N$ refined 2D anomaly maps for each modality.
        Finally, we aggregate the 2D anomaly maps to obtain a 3D Anomaly Volume and an Instance-level Anomaly Score. 
    }
    \label{fig:inference}
\end{figure}

    \subsubsection{\algoname{} Inference}
        \label{subsubsec:inference}
        As pointed out in~\cref{sec:introduction}, our framework achieves robustness to acquisition artefacts -- and thereby maximises precision -- by selecting, for each spatial position and modality, the closest prediction across all views. 
        Hence, as depicted in~\cref{fig:inference}, we predict multiple anomaly maps for each viewpoint and ensemble them so as to maximise precision. 
        In particular, during inference, given a sample with $N$ views, for each of the $N \times N$ combinations of source and target views $(s, t)$, we compute anomaly scores associated with both modalities, $\Psi_{I}^{s,t}$ and $\Psi_{D}^{s,t}$, via the cosine distance between predicted and actual features:
        \begin{equation}
            \Psi_{I}^{s,t} = \mathcal{C}_{D \rightarrow I}^{s,t}
            \quad
            \Psi_{D}^{s,t} = \mathcal{C}_{I \rightarrow D}^{s,t}
        \end{equation}
        with $\Psi_{I}^{s,t}$ and $\Psi_{D}^{s,t}$ reshaped as a 2D anomaly map of resolution $h \times w$.
        Afterwards, for each target view $t$ and for both modalities, we ensemble the anomaly scores computed at each spatial location based on the crossmodal predictions from all source views: % (the same as the target and all other ones): 
        \begin{equation}
            \Psi^t_I = \min_{s \in \{1, \ldots, N\}}  \Psi^{s,t}_I \quad \Psi^t_D = \min_{s \in \{1, \ldots, N\}} \Psi^{s,t}_D
        \end{equation}
        Following common practice in multimodal anomaly detection \cite{wang2023m3dm,costanzino2024cfm,rudolph2023ast}, we filter out background regions based on depth information.
        
        \paragraph{Rationale.}
            \begin{figure}[t]
\centering
    \includegraphics[width=\linewidth]{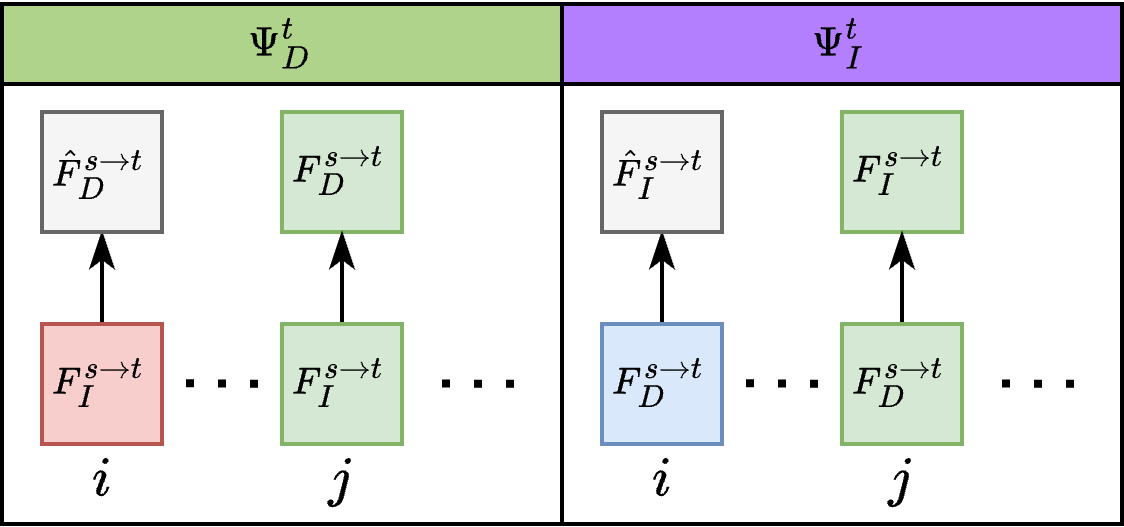}
        \caption{
        \textbf{Rationale for ensembling.}
            Squares represent features (\textcolor{custom_green}{\textbf{green}}: uncorrupted, \textcolor{custom_red}{\textbf{red}}: corrupted by image artefacts, \textcolor{custom_blue}{\textbf{blue}}: corrupted by depth artefacts, \textcolor{gray}{\textbf{grey}}: incorrect prediction), while arrows show mapping predictions.
        }
    \label{fig:rationale}
\end{figure}
            We discuss here in more detail the rationale behind our minimum-based cross-view ensembling strategy, aimed at maximising the robustness of $\Psi^t_D$ and $\Psi^t_I$ with respect to view-dependent image and depth artefacts, respectively.  
            As illustrated in~\cref{fig:rationale} (first column), if an image artefact corrupts a feature (\textcolor{custom_red}{\textbf{red box}}) at a certain position in view $i$, the crossmodal mapping will likely predict an incorrect depth feature (\textcolor{gray}{\textbf{grey box}}), which, therefore, will not match the actual one observed at the considered position in the same view (\textcolor{custom_green}{\textbf{green box}}, top row).
            Yet, the image artefact may not corrupt the feature at the same position in view $j$ (\textcolor{custom_green}{\textbf{green box}}, bottom row), which, therefore, may predict a feature close to the actual one, yielding a low anomaly score in $\Psi^{i,j}_D$.
            Hence, minimum-based ensembling allows for sifting out the lowest anomaly score at every position, thereby alleviating the impact of the image artefacts in $\Psi^t_D$. 
            As shown in~\cref{fig:rationale} (second column), similar considerations may be drawn for a depth artefact and $\Psi^t_I$. 

    \subsubsection{3D Anomaly Detection and Segmentation}
        \label{subsec:volume_aggregation}
        The anomaly maps $\Psi^t_I$,  $\Psi^t_D$ produced for each viewpoint by our minimum-based ensembling strategy are highly precise but may exhibit insufficient sensitivity. 
        To counterbalance this effect, we adopt a recall-oriented, maximum-based strategy when aggregating the per-view anomaly maps across all viewpoints in order to obtain, for each test sample, the 3D anomaly volume.
        In particular, for each view $t$, we project the 2D anomaly maps $\Psi^t_I$, $\Psi^t_D$ into the 3D space using the known camera intrinsics and extrinsics. 
        Each pixel's anomaly scores are assigned to the corresponding 3D voxel. 
        After processing all views, each voxel accumulates multiple scores from different viewpoints and both modalities.
        To obtain the final 3D anomaly volume, we take, for each voxel, the \emph{maximum} score. % across all projections.
        This \emph{maximum}-based aggregation strategy across views ensures that a voxel may be segmented out as anomalous if it contains a defect visible from at least a single viewpoint.
        At last, the global, instance-level anomaly score is taken as the maximum value across all the scores within the anomaly volume.

\section{Experiments}
\label{sec:experiments}

    \subsection{Multiview ADS with \algoname{}}
        We evaluate our method against state-of-the-art anomaly detection approaches adapted to the multiview and multimodal 3D anomaly detection scenario set forth by SiM3D~\cite{costanzino2025sim3d}.
        The considered competitors include memory bank methods (PatchCore~\cite{roth2022patchcore}, BTF~\cite{horwitz2023btf}, M3DM~\cite{wang2023m3dm}), teacher-student approaches (EfficientAD~\cite{batzner2024efficientad}, AST~\cite{rudolph2023ast}), and the original, single-view Crossmodal Feature Mapping (CFM~\cite{costanzino2024cfm}). 
        Hence, as proposed in \cite{costanzino2025sim3d}, all the competitors are adapted to produce 3D anomaly volumes by processing each view independently and aggregating the resulting 2D anomaly maps into the 3D space using the projection strategy described in the SiM3D paper.
        \cref{tab:competitors_real} and~\cref{tab:competitors_synth} report results on the real-to-real and synthetic-to-real setups of SiM3D, respectively, with qualitative results shown in~\cref{fig:qualitatives}.
        More experiments are reported in the Supplementary Material.
        \begin{table*}[ht]
    \centering
    \resizebox{\linewidth}{!}{
        \begin{tabular}{l cc ccccccccc ccccccccc}

            \toprule
        
            \multirow{2}{*}{\textbf{{Algorithm}}} & \multicolumn{2}{c}{\textbf{Features}} & \multicolumn{9}{c}{\textbf{Detection}} & \multicolumn{9}{c}{\textbf{Segmentation}} \\
            
            \cmidrule(lr){4-12} \cmidrule(lr){13-21}
            
            & Image & Depth / Point Cloud & Pl. Stool & Rub. Bin & W. Vase & B. Furn. & Cont. & Pl. Vase & W. Stool & Sink Cab. & Mean & Pl. Stool & Rub. Bin & W. Vase & B. Furn. & Cont. & Pl. Vase & W. Stool & Sink Cab. & Mean \\ 

            \cmidrule(lr){1-1} \cmidrule(lr){2-3} \cmidrule(lr){4-12} \cmidrule(lr){13-21}
            
            \multirow{5}{*}{PatchCore} & WRN-101 & \xmark{} & 0.740 & 0.987 & 0.636 & 0.777 & 0.774 & 0.551 & 1.000 & 0.566 & \underline{0.754} & 0.710 & 0.461 & 0.761 & 0.675 & 0.694 & 0.747 & 0.380 & 0.609 & 0.630 \\ 
                                       & DINO-v2 & \xmark{} & 0.500 & 0.958 & 0.636 & 0.622 & 0.578 & 0.563 & 1.000 & 0.563 & 0.678 & 0.745 & 0.469 & 0.775 & 0.792 & 0.709 & 0.753 & 0.435 & 0.690 & 0.671 \\ 
                                       & \xmark{} & WRN-101 & -- & -- & -- & -- & -- & -- & -- & -- & 0.582 & -- & -- & -- & -- & -- & -- & -- & -- & 0.486 \\
                                       & \xmark{} & DINO-v2 & -- & -- & -- & -- & -- & -- & -- & -- & 0.600 & -- & -- & -- & -- & -- & -- & -- & -- & 0.469 \\
                                       & \xmark{} & FPFH & -- & -- & -- & -- & -- & -- & -- & -- & 0.415 & -- & -- & -- & -- & -- & -- & -- & -- & 0.464 \\
            
            \cmidrule(lr){1-1} \cmidrule(lr){2-3} \cmidrule(lr){4-12} \cmidrule(lr){13-21}
            
            EfficientAD                  & PDN & \xmark{} & 0.280  & 0.732 & 0.000 & 0.878 & 0.424 & 0.730 & 0.928 & 0.712 & 0.586 & 0.682 & 0.462 & 0.763 & 0.534 & 0.680 & 0.743 & 0.407 & 0.488 & 0.595 \\ 
            
            \cmidrule(lr){1-1} \cmidrule(lr){2-3} \cmidrule(lr){4-12} \cmidrule(lr){13-21}
            
            \multirow{2}{*}{AST} & EffNet-B5 & \xmark{} & 0.537 & 0.566 & 0.611 & 0.515 & 0.573 & 0.496 & 0.839 & 0.537 & 0.584 & 0.720 & 0.502 & 0.789 & 0.803 & 0.716 & 0.764 & 0.446 & 0.758 & 0.687 \\
                                 & EffNet-B5 & EffNet-B5 & 0.950 & 0.927 & 0.785 & 0.474 & 0.542 & 0.470 & 0.428 & 0.925 & 0.688 & 0.750 & 0.503 & 0.792 & 0.807 & 0.716 & 0.764 & 0.467 & 0.798 & \underline{0.700} \\ 
            
            \cmidrule(lr){1-1} \cmidrule(lr){2-3} \cmidrule(lr){4-12} \cmidrule(lr){13-21}
            
            \multirow{2}{*}{BTF} & DINO-v2 & FPFH & 0.421 & 0.217 & 0.504 & 0.565 & 0.545 & 0.471 & 0.678 & 0.424 & 0.478 & 0.551 & 0.402 & 0.750 & 0.377 & 0.614 & 0.741 & 0.092 & 0.030 & 0.445 \\
                                 & WRN-101 & WRN-101 & -- & -- & -- & -- & -- & -- & -- & -- & 0.707 & -- & -- & -- & -- & -- & -- & -- & -- & 0.609 \\
            
            \cmidrule(lr){1-1} \cmidrule(lr){2-3} \cmidrule(lr){4-12} \cmidrule(lr){13-21}
            
            \multirow{2}{*}{CFM} & DINO-v2 & FPFH & 0.198 & 0.301 & 0.074 & 0.515 & 0.483 & 0.456 & 0.732 & 0.825 & 0.448 & 0.597 & 0.415 & 0.768 & 0.505 & 0.640 & 0.743 & 0.315 & 0.321 & 0.538 \\
                                 & DINO-v2 & DINO-v2 & -- & -- & -- & -- & -- & -- & -- & -- & 0.548 & -- & -- & -- & -- & -- & -- & -- & -- & 0.663 \\
            
            \cmidrule(lr){1-1} \cmidrule(lr){2-3} \cmidrule(lr){4-12} \cmidrule(lr){13-21}
            
            \multirow{2}{*}{M3DM} & DINO-v2 & FPFH & 0.702 & 0.988 & 0.661 & 0.545 & 0.556 & 0.649 & 0.392 & 0.475 & 0.621 & 0.733 & 0.452 & 0.767 & 0.702 & 0.702 & 0.752 & 0.288 & 0.126 & 0.565 \\ 
                                  & DINO-v2 & DINO-v2 & -- & -- & -- & -- & -- & -- & -- & -- & 0.659 & -- & -- & -- & -- & -- & -- & -- & -- & 0.503 \\
            
            \cmidrule(lr){1-1} \cmidrule(lr){2-3} \cmidrule(lr){4-12} \cmidrule(lr){13-21}
            
            \algoname{} & DINO-v2  & DINO-Depth    & 0.909 & 0.990 & 0.945 & 0.647 & 0.740 & 0.607 & 0.916 & 1.000 & \textbf{0.844} & 0.855 & 0.707 & 0.863 & 0.791 & 0.831 & 0.885 & 0.711 & 0.785 & \textbf{0.804} \\ 
            
            \bottomrule
            
        \end{tabular}}
    \caption{
        \textbf{Results on real-to-real setup of SiM3D.} 
        Best results in \textbf{bold}, runner-ups \underline{underlined}.
        Missing entries (--) not reported in~\cite{costanzino2025sim3d}.
        }
    \label{tab:competitors_real}

\end{table*}
        \begin{table*}[ht]
    \centering
    \resizebox{\linewidth}{!}{
        \begin{tabular}{l cc ccccccccc ccccccccc}

            \toprule
        
            \multirow{2}{*}{\textbf{{Algorithm}}} & \multicolumn{2}{c}{\textbf{Features}} & \multicolumn{9}{c}{\textbf{Detection}} & \multicolumn{9}{c}{\textbf{Segmentation}} \\
            
            \cmidrule(lr){4-12} \cmidrule(lr){13-21}
            
            & Image & Depth / Point Cloud & Pl. Stool & Rub. Bin & W. Vase & B. Furn. & Cont. & Pl. Vase & W. Stool & Sink Cab. & Mean & Pl. Stool & Rub. Bin & W. Vase & B. Furn. & Cont. & Pl. Vase & W. Stool & Sink Cab. & Mean \\ 

            \cmidrule(lr){1-1} \cmidrule(lr){2-3} \cmidrule(lr){4-12} \cmidrule(lr){13-21}
            
            \multirow{5}{*}{PatchCore} & WRN-101 & \xmark{} & 0.462 & 0.235 & 0.537 & 0.353 & 0.678 & 0.576 & 0.517 & 0.250 & 0.451 & 0.734 & 0.437 & 0.751 & 0.454 & 0.678 & 0.741 & 0.386 & 0.618 & 0.600 \\
                                       & DINO-v2 & \xmark{} & 0.958 & 0.897 & 0.413 & 0.464 & 0.207 & 0.684 & 0.607 & 0.087 & 0.540 & 0.701 & 0.457 & 0.772 & 0.563 & 0.648 & 0.734 & 0.321 & 0.575 & 0.596\\
                                       & \xmark{} & WRN-101 & -- & -- & -- & -- & -- & -- & -- & -- & 0.507 & -- & -- & -- & -- & -- & -- & -- & -- & 0.484 \\
                                       & \xmark{} & DINO-v2 & -- & -- & -- & -- & -- & -- & -- & -- & 0.246 & -- & -- & -- & -- & -- & -- & -- & -- & 0.471 \\
                                       & \xmark{} & FPFH    & -- & -- & -- & -- & -- & -- & -- & -- & 0.313 & -- & -- & -- & -- & -- & -- & -- & -- & 0.460 \\
            
            \cmidrule(lr){1-1} \cmidrule(lr){2-3} \cmidrule(lr){4-12} \cmidrule(lr){13-21}
            
            EfficientAD                  & PDN & \xmark{} & 0.123 & 0.673 & 0.000 & 0.848 & 0.008 & 0.487 & 0.750 & 0.662 & 0.444 & 0.680 & 0.449 & 0.729 & 0.523 & 0.669 & 0.747 & 0.431 & 0.361 & 0.574 \\
            
            \cmidrule(lr){1-1} \cmidrule(lr){2-3} \cmidrule(lr){4-12} \cmidrule(lr){13-21}
            
            \multirow{2}{*}{AST} & EffNet-B5 & \xmark{}        & 1.000 & 0.344 & 1.000 & 0.525 & 0.482 & 0.388 & 0.428 & 0.187 & \underline{0.544} & 0.726 & 0.488 & 0.791 & 0.806 & 0.705 & 0.764 & 0.461 & 0.695 & 0.680 \\
                                 & EffNet-B5 & EffNet-B5 & 0.636 & 0.002 & 0.504 & 0.474 & 0.462 & 0.569 & 0.428 & 0.887 & 0.495 & 0.751 & 0.502 & 0.793 & 0.805 & 0.717 & 0.764 & 0.450 & 0.785 & \underline{0.696} \\
            
            \cmidrule(lr){1-1} \cmidrule(lr){2-3} \cmidrule(lr){4-12} \cmidrule(lr){13-21}
            
            \multirow{2}{*}{BTF} & DINO-v2 & FPFH    & 0.462 & 0.294 & 0.520 & 0.444 & 0.464 & 0.424 & 0.482 & 0.287 & 0.422 & 0.547 & 0.402 & 0.756 & 0.369 & 0.610 & 0.741 & 0.103 & 0.040 & 0.446 \\
                                 & WRN-101 & WRN-101 & -- & -- & -- & -- & -- & -- & -- & -- & 0.448 & -- & -- & -- & -- & -- & -- & -- & -- & 0.543 \\
            
            \cmidrule(lr){1-1} \cmidrule(lr){2-3} \cmidrule(lr){4-12} \cmidrule(lr){13-21}
            
            \multirow{2}{*}{CFM} & DINO-v2 & FPFH    & 0.008 & 0.000 & 0.190 & 0.424 & 0.313 & 0.459 & 0.428 & 0.362 & 0.273 & 0.523 & 0.413 & 0.753 & 0.542 & 0.621 & 0.741 & 0.222 & 0.314 & 0.516 \\
                                 & DINO-v2 & DINO-v2 & -- & -- & -- & -- & -- & -- & -- & -- & 0.311 & -- & -- & -- & -- & -- & -- & -- & -- & 0.620 \\
            
            \cmidrule(lr){1-1} \cmidrule(lr){2-3} \cmidrule(lr){4-12} \cmidrule(lr){13-21}
            
            \multirow{2}{*}{M3DM} & DINO-v2 & FPFH    & 0.107 & 0.117 & 0.735 & 0.565 & 0.381 & 0.586 & 0.214 & 0.512 & 0.402 & 0.690 & 0.454 & 0.770 & 0.461 & 0.645 & 0.734 & 0.318 & 0.132 & 0.526 \\
                                  & DINO-v2 & DINO-v2 & -- & -- & -- & -- & -- & -- & -- & -- & 0.254 & -- & -- & -- & -- & -- & -- & -- & -- & 0.476 \\
            
            \cmidrule(lr){1-1} \cmidrule(lr){2-3} \cmidrule(lr){4-12} \cmidrule(lr){13-21}
            
            \algoname{} & DINO-v2 & DINO-Depth     & 0.681 & 0.707 & 0.954 & 0.472 & 0.805 & 0.594 & 0.312 & 0.458 & \textbf{0.623} & 0.778 & 0.663 & 0.861 & 0.752 & 0.850 & 0.864 & 0.550 & 0.725 & \textbf{0.755} \\

            \bottomrule
            
        \end{tabular}}
    \caption{
        \textbf{Results on synthetic-to-real setup of SiM3D.} 
        Best results in \textbf{bold}, runner-ups \underline{underlined}.
        Missing entries (--) not reported in~\cite{costanzino2025sim3d}.
        }
    \label{tab:competitors_synth}

\end{table*}
        \begin{figure}[t]
    \centering
        \includegraphics[width=\linewidth]{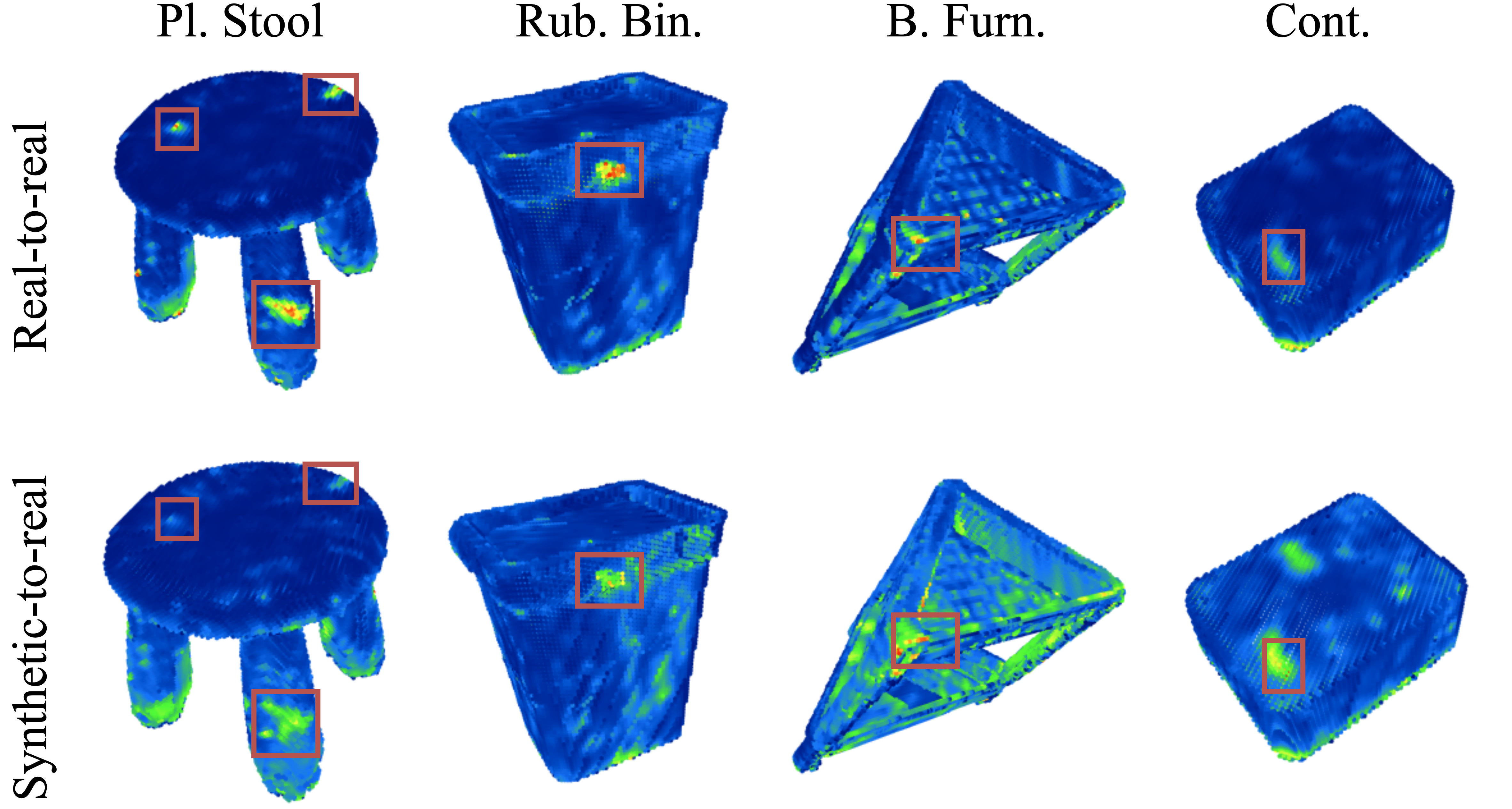}
    \caption{\textbf{Qualitative results.}
    Real-to-real (top) vs. Synthetic-to-real (bottom). 
    Anomalies are highlighted by \textcolor{custom_red}{{\textbf{red boxes}}}.
    }
    \label{fig:qualitatives}
\end{figure}

        \paragraph{Real-to-real Setup.}
            In the real-to-real setup (\cref{tab:competitors_real}), our method (\algoname{}) achieves the best performance in both the detection and segmentation tasks, with a mean I-AUROC of $0.844$ and a mean V-AUPRO@1\% of $0.804$. 
            This represents a substantial improvement -- $\sim$14\% in detection and $\sim$10\% in segmentation -- over previous multimodal methods, i.e., BTF (WRN-101) and AST, which achieve $0.707$ I-AUROC and $0.700$ V-AUPRO@1\%, respectively. 
            Notably, \algoname{} significantly outperforms the original CFM ($0.448$ I-AUROC, $0.538$ V-AUPRO@1\%), demonstrating the effectiveness of our \emph{natively multiview} framework.
            Compared to the runner-up in detection, i.e., PatchCore with WRN-101, \algoname{} yields $\sim$9\% improvement (I-AUROC\%: $0.844$ vs $0.754$), with the gap in segmentation being $\sim$17\% (V-AUPRO@1\%: $0.804$ vs $0.630$).

        \paragraph{Synthetic-to-real Setup.}
            The synthetic-to-real setup (\cref{tab:competitors_synth}) presents a more challenging scenario due to the domain shift between synthetic training data and real test samples. Indeed, all methods show a significant performance drop, with \algoname{} achieving $0.623$ I-AUROC and $0.755$ V-AUPRO@1\%, outperforming the runner-ups of $\sim$8\% in detection and $\sim$6\% in segmentation.
            Interestingly, some competitors exhibit erratic behaviour in this setup, i.e., AST achieves $1.000$ I-AUROC on Plastic Stool but drops to $0.002$ on Rubbish Bin, suggesting instability when faced with domain shift. 
            In contrast, \algoname{} demonstrates more consistent performance across categories, which vouches for the robustness of our paradigm. 
            
    \subsection{Ablation Studies}
        We conduct comprehensive ablation studies to analyse the contribution of the key components of our framework. 
        All ablations are performed on the real-to-real setup of SiM3D.
        More ablations are included in the Supplementary Material.
        
        \paragraph{View Conditioning.}
            \begin{table}[t]
    \centering
    \resizebox{\linewidth}{!}{
        \begin{tabular}{c cc cc}

            \toprule
        
            \multirow{2}{*}{\textbf{{View Conditioning}}} & \multicolumn{2}{c}{\textbf{Features}} & \multirow{2}{*}{\textbf{Detection}} & \multirow{2}{*}{\textbf{Segmentation}} \\
            
            \cmidrule(lr){2-3}
            
            & Image & Point Cloud & & \\ 

            \cmidrule(lr){1-1} \cmidrule(lr){2-3} \cmidrule(lr){4-5}
                        
            \xmark{} & DINO-v2 & FPFH & 0.448 & 0.538 \\
            \cmark{} & DINO-v2 & FPFH & \textbf{0.548} & \textbf{0.663} \\

            \bottomrule
            
        \end{tabular}}
    \caption{
    \textbf{Impact of View Conditioning.}
    }
    \label{tab:modulation}
\end{table}
            \cref{tab:modulation} demonstrates the impact of introducing view conditioning into the CFM framework adopted in~\cite{costanzino2025sim3d} to more robustly handle one-to-many mappings caused by the multi-view setup. 
            Purposely, we introduce two modality-specific modulators that condition the features passed to the mapping networks based on the one-hot encoded view identity.  
            Without view conditioning, the method achieves $0.448$ I-AUROC and $0.538$ V-AUPRO@1\%. 
            Adding view conditioning via feature modulation does improve performance to $0.548$ I-AUROC and $0.663$ V-AUPRO@1\%, representing gains of $+10.0$\% in detection and $+12.5$\% in segmentation.

\begin{figure}[t]
    \centering
    \resizebox{\linewidth}{!}{
    \setlength{\tabcolsep}{2px}
    \begin{tabular}{ccc}
    
        Depth & DINO-v2 & DINO-Depth \\

        \includegraphics[width=0.33\linewidth,angle=180,origin=c]{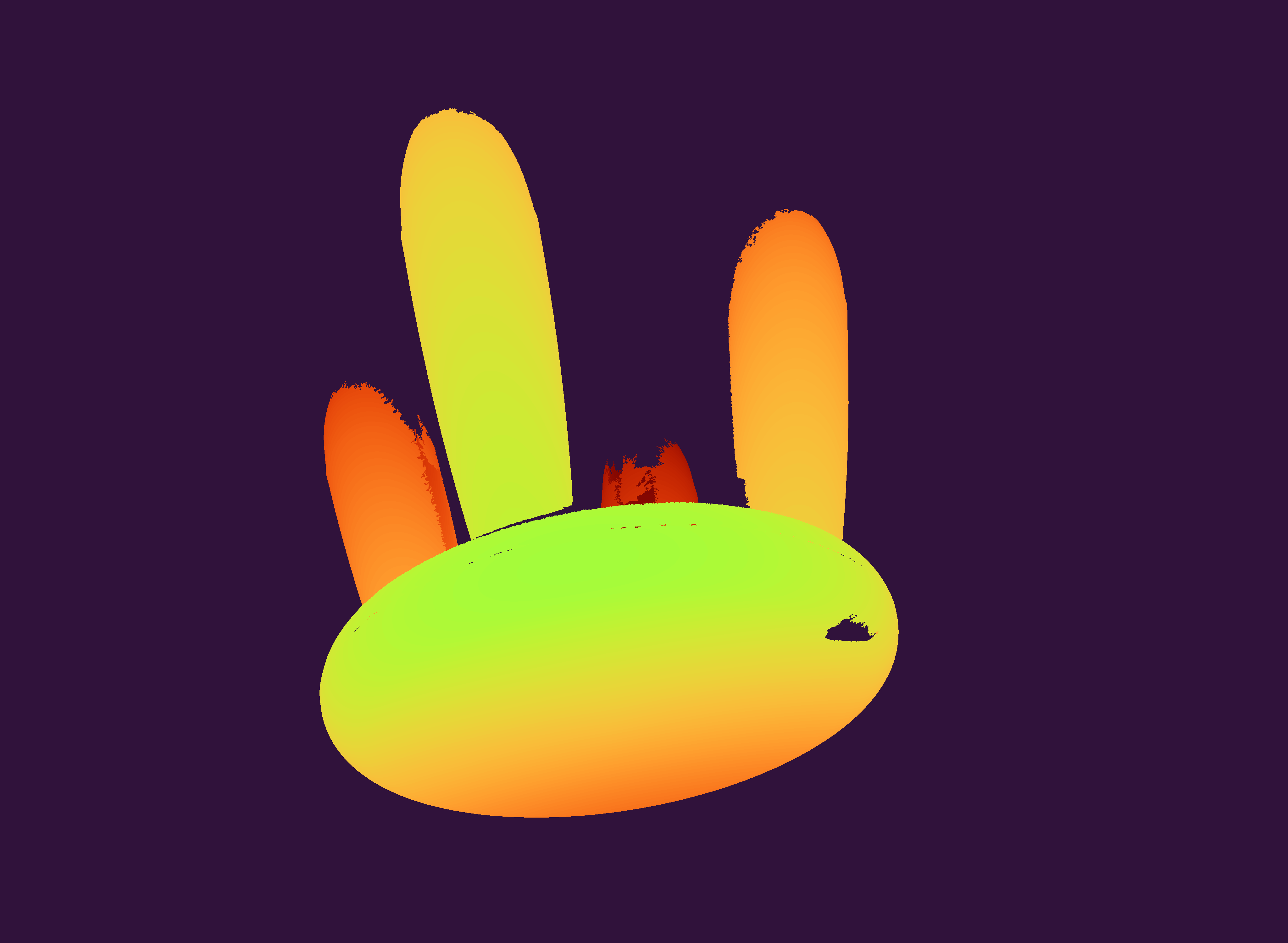} & 
        \includegraphics[width=0.33\linewidth,angle=180,origin=c]{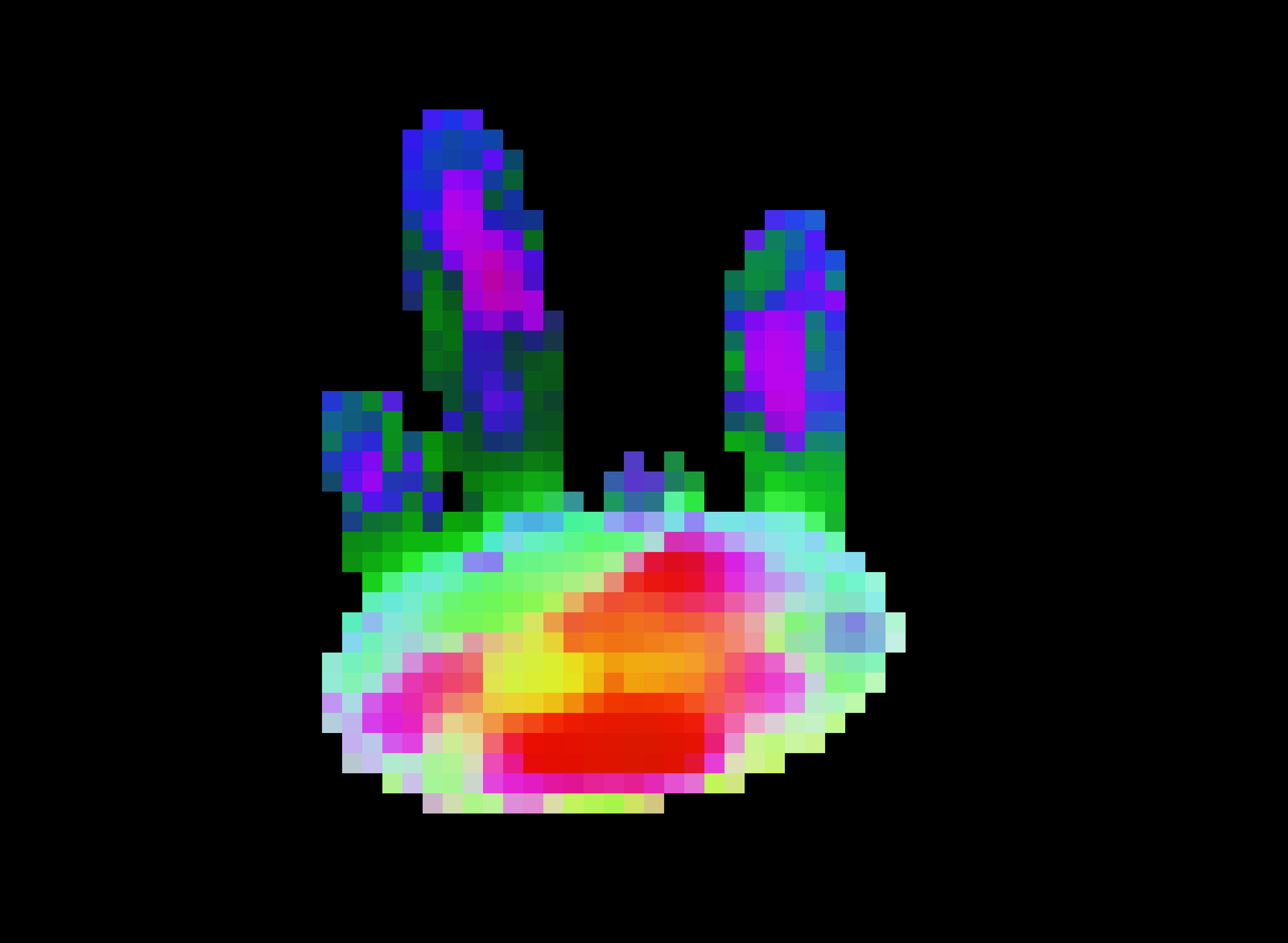} & 
        \includegraphics[width=0.33\linewidth,angle=180,origin=c]{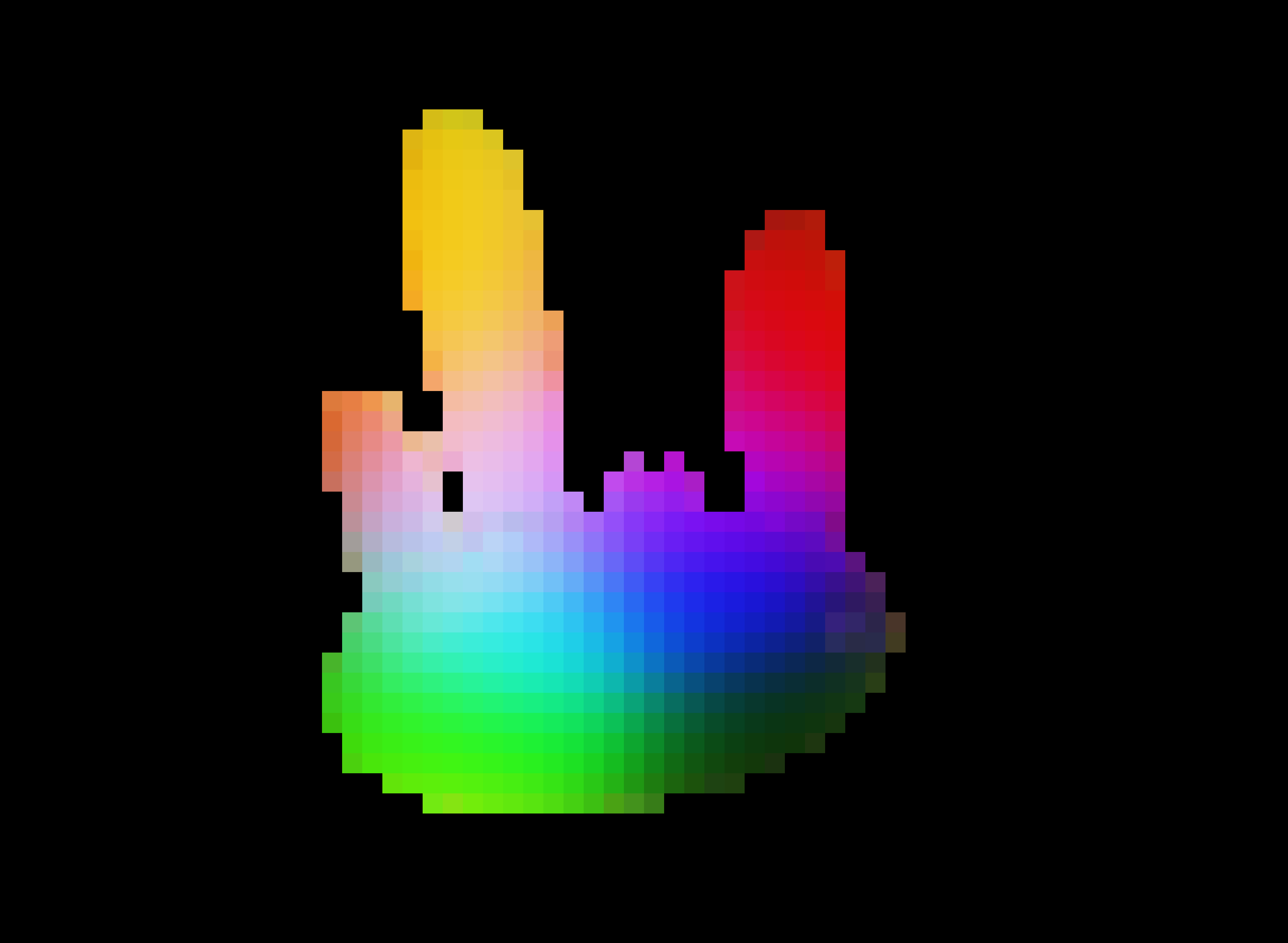} \\

        \includegraphics[width=0.33\linewidth,angle=180,origin=c]{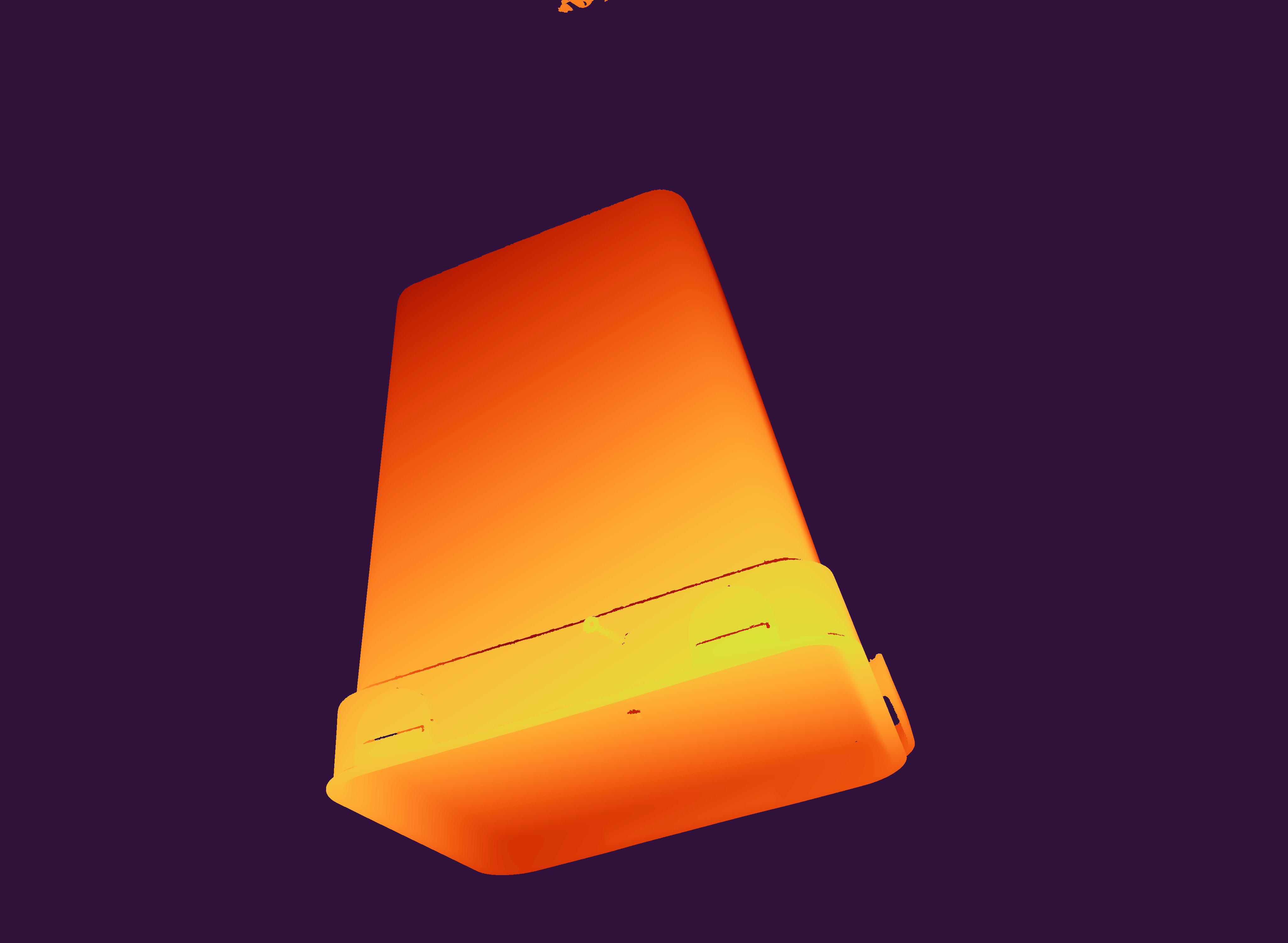} & 
        \includegraphics[width=0.33\linewidth,angle=180,origin=c]{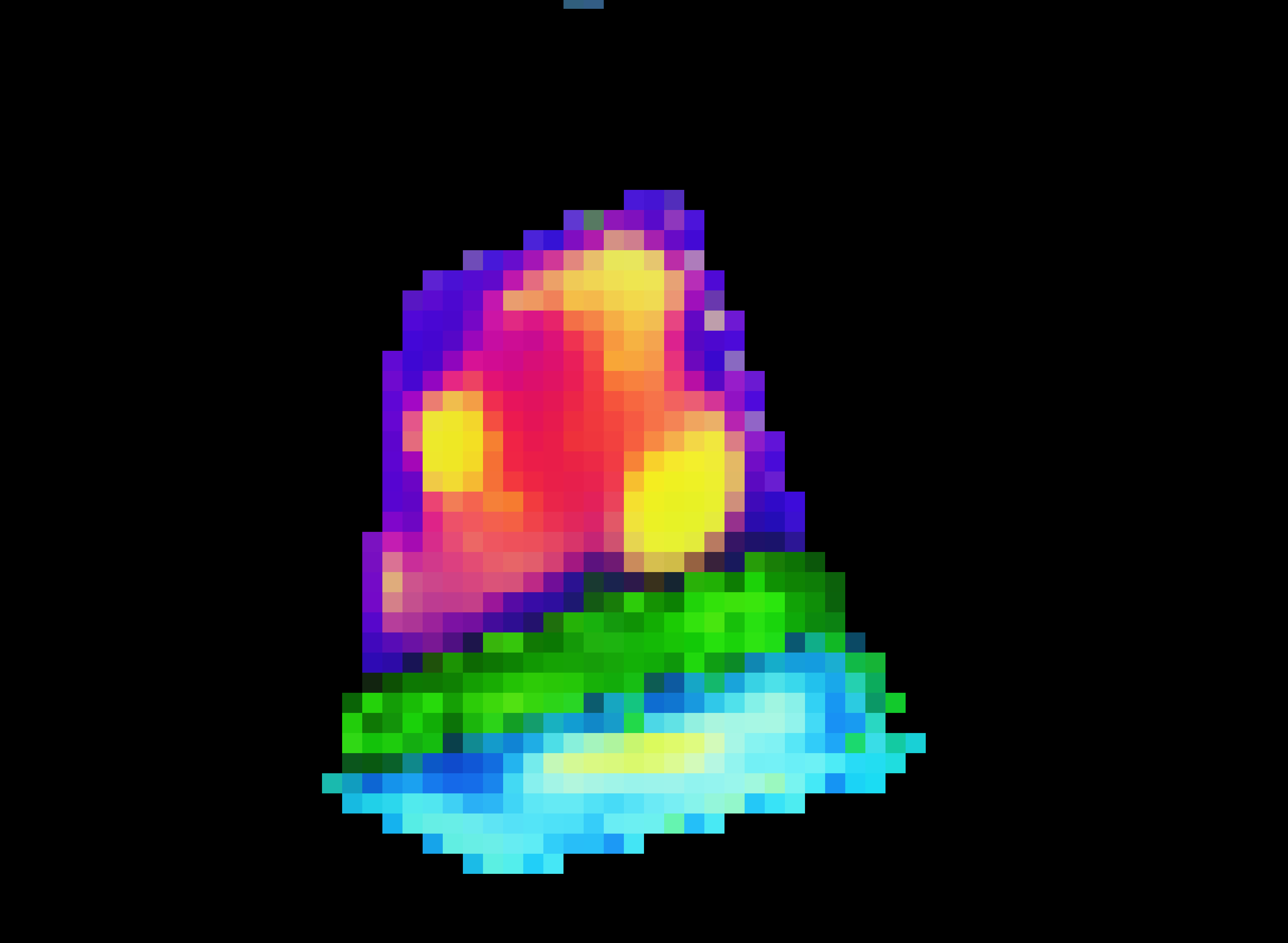} & 
        \includegraphics[width=0.33\linewidth,angle=180,origin=c]{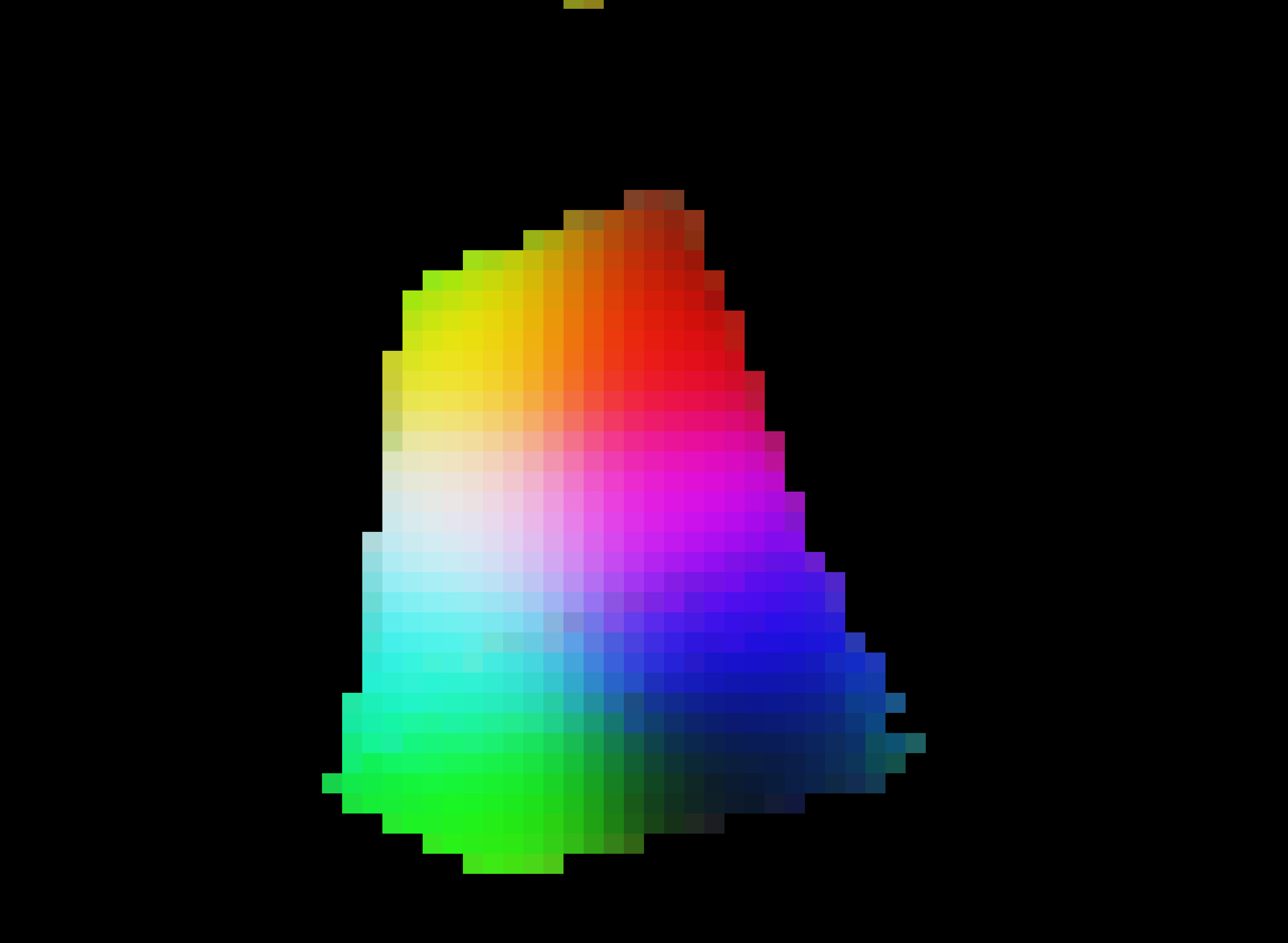} \\

    \end{tabular}
    }
    \caption{\textbf{PCA of Depth Features.}
    }
    \label{fig:depth_features}
\end{figure}

        \paragraph{3D Feature Extractor.}
            \begin{table}[ht]
    \centering
    \resizebox{\linewidth}{!}{
        \begin{tabular}{cc cc}

            \toprule
        
            \multicolumn{2}{c}{\textbf{Features}} & \multirow{2}{*}{\textbf{Detection}} & \multirow{2}{*}{\textbf{Segmentation}} \\
            
            \cmidrule(lr){1-2}
            
            Image & Depth / Point Cloud & & \\ 

            \cmidrule(lr){1-2} \cmidrule(lr){3-4}
                        
            \multirow{4}{*}{DINO-v2} & FPFH       & 0.548 & 0.663 \\
            & Point-MAE                           & 0.543 & 0.545 \\
            & DINO-v2                             & 0.575 & 0.684 \\
            & DINO-Depth                          & \textbf{0.804} & \textbf{0.735} \\

            \bottomrule
            
        \end{tabular}}
    \caption{
        \textbf{Effects of 3D Feature Extractor.} 
        }
    \label{tab:depth_encoder}
\end{table}
            Keeping the improved version of CFM that incorporates view conditioning, in~\cref{tab:depth_encoder} we compare different choices for the 3D feature extractor.  
            Using FPFH features, as proposed in~\cite{costanzino2025sim3d}, yields $0.548$ I-AUROC and $0.663$ V-AUPRO@1\%.
            Employing Point-MAE, as originally proposed in~\cite{costanzino2024cfm}, yields lower performance, i.e., $0.543$ I-AUROC and $0.545$ V-AUPRO@1\%. 
            Switching to foundational image features by feeding DINO-v2 with depth maps improves performance to $0.575$ I-AUROC and $0.684$ V-AUPRO@1\%. 
            However, DINO-Depth, our dedicated foundational depth encoder trained on industrial datasets, achieves by far the best results with $0.804$ I-AUROC and $0.735$ V-AUPRO@1\%, with improvements of $+22.9$\% in detection and $+5.1$\% in segmentation compared to DINO-v2.
            A comparison between DINO-v2 and DINO-Depth features is shown in~\cref{fig:depth_features}.
            The features learned by DINO-v2 are noisier and tend to over-segment depth maps, while those computed by DINO-Depth are significantly smoother and seem more amenable to capturing the semantics of depth maps.

        \paragraph{Depth Encoder Training Set.}
            We train DINO-Depth in a self-supervised manner, progressively expanding the composition of the training set, as shown in~\cref{tab:depth_encoder_training_set}.
            We begin with training split A, comprising MVTec 3D-AD~\cite{bergmann20223dad} and Eyecandies~\cite{bonfiglioli2022eyecandies}, namely the first two datasets for multimodal anomaly detection.
            Expanding to training split B, we incorporate Real-IAD D\textsuperscript{3}~\cite{wang2024real}, a more recent large-scale multimodal dataset, obtaining improvements of $+4.1$\% in detection and $+3.8$\% in segmentation over split A.
            Finally, we create training split C by adding depth maps obtained by using DepthAnything-v2~\cite{depth_anything_v2} on image-only datasets: MVTec AD~\cite{bergmann2019mvtec}, MVTec LOCO~\cite{mvtec_loco}, MVTec AD 2~\cite{heckler2025mvtecad2}, and ViSA~\cite{visa}. 
            This substantially increases training scale, delivering $+19.5$\% in detection and $+3.0$\% in segmentation relative to split B, demonstrating the value of large-scale pre-training even with pseudo-depth annotations.
            \begin{table}[ht]
    \centering
    \resizebox{\linewidth}{!}{
        \begin{tabular}{cccc}

            \toprule
        
            \textbf{Training Split} & \textbf{No. Samples} & \textbf{Detection} & \textbf{Segmentation} \\
            
            \cmidrule(lr){1-2} \cmidrule(lr){3-4}

            A & 19647 & 0.568 & 0.667 \\
            B & 23896 & 0.609 & 0.705 \\
            C & 46528 & \textbf{0.804} & \textbf{0.735} \\
                        
            \bottomrule
            
        \end{tabular}}
    \caption{
        \textbf{DINO-Depth Training Set.} 
        }
    \label{tab:depth_encoder_training_set}
\end{table}
            
        \paragraph{Cross-View Conditioning.}
            \cref{tab:cross_pairs} compares the CFM approach improved by view conditioning and DINO-Depth to our final proposal, namely \algoname{}, that deploys also cross-view conditioning along with the associated \emph{minimum}-based ensembling and \emph{maximum}-based aggregation, to pursue robustness to acquisition artefacts without sacrificing recall.
            Hence, CFM with view conditioning and DINO-Depth achieves $0.804$ I-AUROC and $0.735$ V-AUPRO@1\%, while \algoname{} yields $0.844$ I-AUROC and $0.804$ V-AUPRO@1\%, with a remarkable improvement of $+4.0$\% in detection and $+6.9$\% in segmentation.
            \begin{table}[ht]
    \centering
    \resizebox{0.7\linewidth}{!}{
        \begin{tabular}{c cc}
            \toprule
        
            \textbf{Cross-View} & \multirow{2}{*}{\textbf{Detection}} & \multirow{2}{*}{\textbf{Segmentation}} \\
            \textbf{Conditioning} & & \\
            
            \cmidrule(lr){1-1} \cmidrule(lr){2-3}

            \xmark{} & 0.804 & 0.735 \\
            \cmark{} & \textbf{0.844} & \textbf{0.804} \\
            
            \bottomrule
            
        \end{tabular}}
    \caption{
        \textbf{Cross-View Conditioning.}
        }
    \label{tab:cross_pairs}
\end{table}

    \subsection{Multi-class ADS.}        
        Recently, multi-class ADS, the task of training a single unified model to detect and segment anomalies across multiple object categories, has gained traction in literature, with proposals including both class-agnostic~\cite{he2024mambaad, guo2025dinomaly} and class-conditioned architectures~\cite{jiang2024multiclassanomalydetectionexploring, zhou2024vqflowtamingnormalizingflows}.
        This approach offers a particularly MLOps-friendly design as it allows a single model to be trained, deployed and maintained across multiple production lines.  
        \algoname{} can be seamlessly extended to multi-class ADS by incorporating one-hot class encodings into the modulators, thereby conditioning the crossmodal mapping networks on both the view pair and object category.
        We report the comparison between class-specific and multi-class formulations of \algoname{} in \cref{tab:omni_model}. 
        We observe that \algoname{} is particularly amenable to the SiM3D multi-class scenario, with a comparable performance to the class-specific models.
        \begin{table}[ht]
    \centering
    \resizebox{\linewidth}{!}{
        \begin{tabular}{c cc cc}
            
            \toprule

            \multirow{2}{*}{\textbf{Model}} & \multicolumn{2}{c}{\textbf{Real-to-real}} & \multicolumn{2}{c}{\textbf{Synthetic-to-real}} \\

            \cmidrule(lr){2-3} \cmidrule(lr){4-5}
            
            & Detection & Segmentation & Detection & Segmentation \\
            
            \cmidrule(lr){1-1} \cmidrule(lr){2-3} \cmidrule(lr){4-5}
            
            Class-specific & 0.844 & \textbf{0.804} & \textbf{0.623} & 0.755 \\
            Multi-class    & \textbf{0.850} & 0.801 & 0.618 & \textbf{0.758} \\
            
            \bottomrule
            
        \end{tabular}}
    \caption{
        \textbf{Class-specific vs. Multi-class.}
        }
    \label{tab:omni_model}
\end{table}
\section{Concluding Remarks}
\label{sec:discussion}
    We have presented \algoname{}, the first \emph{natively} multiview framework for multimodal 3D anomaly detection. 
    Our proposal extends the original crossmodal feature mapping formulation by three key contributions. 
    First, we introduce view-conditioning by feature-wise modulation to handle potential one-to-many crossmodal mappings. 
    Second, we train DINO-Depth, a foundational depth encoder tailored to industrial datasets, which enables processing of high-resolution 3D data. 
    Third, we propose cross-view conditioning with minimum-ensembling and maximum-aggregation, to optimise both robustness to sensing artefacts and sensitivity to defects.
    Ablation studies validate each contribution's effectiveness, with view-conditioning alone yielding  $\sim$10\% and $\sim$12\% improvements in detection and segmentation, DINO-Depth adding $\sim$26\% and $\sim$7\% on top of previous improvements, and, finally, cross-view conditioning with min-max further increasing performance by 4\% and $\sim$7\% in detection and segmentation.
    Experiments on both setups of SiM3D demonstrate state-of-the-art performance, with large margins versus previous methods.
    Moreover, \algoname{} lends itself to a natural and effective multi-class formulation, which may streamline adoption in real industrial settings.
    The main limitation of our work pertains to the relatively small variety of the experimental data, due to SiM3D being nowadays the only dataset for multiview and multimodal 3D anomaly detection. 
    Besides, the performance achieved by \algoname{} is far from saturating the benchmark, highlighting the need for further research on the challenging ADS setup proposed by SiM3D.

{
    \clearpage
    \small
    \bibliographystyle{ieeenat_fullname}
    \bibliography{main}
}

\clearpage
\maketitlesupplementary

\setcounter{section}{0}
\renewcommand{\thesection}{\Alph{section}}

\section{Experimental Settings}
    \subsection{Datasets and Metrics}
        We evaluate our method on the SiM3D benchmark~\cite{costanzino2025sim3d}, which consists of $8$ object categories with a total of $333$ instances.
        Each instance is captured from multiple calibrated viewpoints ($12$ or $36$ views, depending on the object type), providing both high-resolution grayscale images ($12$ Mpx) and dense 3D measurements ($5$-$7$M points) that can be accessed as either point clouds or depth maps.
        Following the benchmark protocol, we train on a single nominal instance per object category and test on all remaining instances, which include both nominal and anomalous samples.
        We evaluate our method in both setups defined by SiM3D: \textbf{real-to-real}, where training uses a single real nominal instance, and \textbf{synthetic-to-real}, where training uses data rendered from a CAD model; for both setups, the testing is conducted on real data.
        
        We adopt the evaluation metrics proposed by SiM3D.
        For anomaly detection, we report instance-level AUROC (I-AUROC) computed on the global anomaly scores.
        For anomaly segmentation, we report voxel-level AUPRO integrated up to $1$\% false positive rate (V-AUPRO@$1$\%), which better reflects the stringent requirements of industrial applications compared to the more commonly used $30$\% threshold.
        
    \subsection{Implementation Details}
        \paragraph{Feature Extractors.}
            For image feature extraction, we employ DINO-v2 ViT-B/14~\cite{oquab2023dinov2}, which provides features of dimension $c_{I}=768$.
            For depth feature extraction, we use DINO-Depth, our dedicated depth encoder, trained by a self-supervised learning objective similar to DINO-v2.
            Since the datasets selected to train DINO-Depth contain approximately $47$k samples, far from the large-scale training of DINO-v2, we train a ViT-S/14 to avert overfitting, yielding features of dimension $c_{D}=384$.
            Both feature extractors remain frozen during training of the modulator and crossmodal mapping networks.
        
        \paragraph{Network Architecture.}
            The crossmodal mapping networks $\mathcal{M}_{I \rightarrow D}$ and $\mathcal{M}_{D \rightarrow I}$ are implemented as three-layer MLPs with hidden dimensions $[768, 576, 384]$ and $[1152, 576, 384]$, respectively.
            Each hidden layer is followed by GeLU activation. 
            The view modulators $\Phi_{I}$ and $\Phi_{D}$ consist of two-layer MLPs with hidden dimension $128$. Both modulators take as input the concatenated one-hot encodings of source and target views and produce scale and shift parameters for feature-wise modulation.
        
        \paragraph{Training.}
            We resize images and depth maps to $896 \times 896$ pixels and train jointly $\mathcal{M}_{I \rightarrow D}$, $\mathcal{M}_{D \rightarrow I}$, $\Phi_{I}$, $\Phi_{D}$ for $200$ epochs using the Adam optimiser~\cite{adam} with an initial learning rate of $10^{-4}$.
            We employ the OneCycleLR scheduler~\cite{smith2019super} with maximum learning rate $5 \times 10^{-4}$, cosine annealing strategy, and $10$\% warm-up period. 
            During each epoch, we process all $N \times N$ source-target view pairs from the single training instance,  where $N$ is the number of views.
            The pairs are processed in batches of $48$. 
            This exhaustive pairing strategy ensures the network learns crossmodal mappings for all possible view relationships.
                
\section{Additional Experiments}

    \subsection{Computational Cost Analysis}
        To analyse the computational requirements of \algoname{}, we compare inference times across different architectural choices on the \emph{Sink Cabinet} class, which represents the most computationally demanding scenario in the dataset with $36$ available views and unfiltered background regions. 
        To ensure fair comparison, we measure inference times on the same machine for all architectural choices, computing the average across all the test samples from the \emph{Sink Cabinet} class. 
        For each sample, we record the elapsed time from data loading onto the GPU to the final computation of all the per-view anomaly maps. 
        All time measurements are performed after GPU warm-up, and we synchronise all CUDA threads before recording the total inference time to ensure accurate timing estimates.
        Note that these timings exclude the volume construction step, which is identical across all methods and thus does not affect relative comparisons.
        
        The original CFM~\cite{costanzino2024cfm} approach, using Point-MAE as the 3D backbone, requires $1654.792$ ms per sample for processing all $36$ views.
        Transitioning to the SiM3D configuration~\cite{costanzino2025sim3d} with DINO-v2 features for both images and depths, increases the inference time to $3815.606$ ms ($2.3\times$ overhead), primarily due to the higher-dimensional feature space for the 3D feature extractor.
        Despite incorporating cross-view feature aggregation, \algoname{} achieves $2886.768$ ms per sample -- $24$\% faster than the CFM SiM3D configuration -- while delivering superior performance through multi-view reasoning.
        This efficiency gain stems from our streamlined depth-based processing pipeline based on ViT-S/14\footnote{ViT-S/14 features $12$ layers, $384$ hidden size, $6$ heads and $1536$ MLP width, while ViT-B/14 features $12$ layers, $768$ hidden size, $12$ heads and $3072$ MLP width.}, demonstrating that cross-view modelling can be implemented without prohibitive computational penalties.
        Compared to the CFM SiM3D configuration, \algoname{} is advantageous in terms of both performance and cost, as it simultaneously improves detection and segmentation accuracy, while reducing inference time.
        \begin{table}[ht]
    \centering
    \resizebox{\linewidth}{!}{
        \begin{tabular}{c ccccccccc ccccccccc}

            \toprule
        
            \multirow{2}{*}{\textbf{{No. Views}}} & \multirow{2}{*}{\textbf{{Inference Time}}} & \textbf{Detection} & \textbf{Segmentation} \\
            
            \cmidrule(lr){3-4}
            
            [\#] & [ms] & \multicolumn{2}{c}{\emph{Sink Cabinet}} \\

            \cmidrule(lr){1-2} \cmidrule(lr){3-4}

            36 & 2886.768 & 1.000 & 0.785 \\
            18 & 1443.384 & 1.000 & 0.785 \\
            9  & 721.692  & 1.000 & 0.783 \\
            5  & 400.94   & 0.972 & 0.781 \\
            
            \bottomrule
            
        \end{tabular}}
    \caption{
        \textbf{No. Views vs. Inference Time vs. Performance}
        Comparison on \emph{Sink Cabinet}, i.e., the most computationally expensive class of SiM3D.
        }
    \label{tab:number_pair}
\end{table}
        Eventually, as each of the $N \times N$ anomaly maps can be individually aggregated into the anomaly volume, the runtime memory requirements pertain to $N$ feature maps and one anomaly map at the time, thus memory usage scales as $\mathcal{O}(N)$ and not $\mathcal{O}(N \times N)$.

        \paragraph{Cost Mitigation in Low-Resource Environments.}
            While incorporating cross-view feature mapping introduces additional computational overhead compared to single-view processing, this cost can be effectively mitigated through random view sampling without significant performance degradation.
            \cref{tab:number_pair} demonstrates this trade-off on the \emph{Sink Cabinet} class.
            Our analysis reveals that reducing the number of cross-view pairs from $36$ to $18$ halves inference time (from $2886.768$ ms to $1443.384$ ms per sample) while maintaining identical detection and segmentation performance (AUROC $1.000$, AUPRO $0.785$). 
            An even more aggressive reduction to $9$ views ($721.692$ ms) preserves perfect detection performance while yielding only a negligible 0.2\% drop in segmentation quality. 
            At $5$ views, the method achieves a $7.2\times$ speed-up over the full $36$-view configuration while retaining $97.2$\% detection accuracy and $78.1$\% segmentation performance. 
            This demonstrates that \algoname{} can be adapted to resource-constrained deployment scenarios by sampling a subset of available views, offering a tunable balance between computational efficiency and anomaly detection and segmentation accuracy.

    \subsection{Ablations}
        \paragraph{Preliminary Study on the Image Feature Extractor.}
            Before developing our dedicated depth encoder, we conducted a preliminary study to determine whether existing pre-trained vision models could effectively extract features from both modalities. 
            As shown in ~\cref{tab:image_encoder}, we compared using DINO-v3 and DINO-v2 as feature extractors for both images and depth maps (after treating them as single-channel images).
            
            DINO-v2 demonstrates superior performance, achieving $0.575$ I-AUROC and $0.684$ V-AUPRO@$1$\% compared to DINO-v3's $0.534$ I-AUROC and $0.569$ V-AUPRO@$1$\%, representing improvements of $7.7$\% in detection and $20.2$\% in segmentation. 
            We attribute this performance gap to differences in the training data composition. 
            While DINO-v3's training corpus consists predominantly of social media images, which may lack industrial imagery, DINO-v2 was trained on a more diverse dataset that likely includes a broader range of visual domains.
            
            This finding informed two key design decisions: (1) we adopted DINO-v2 as our image feature extractor, and (2) we employed the same Vision Transformer architecture\footnote{In particular, same patch size and positional encoding.} and training methodology for our dedicated depth encoder (DINO-Depth), ensuring architectural consistency while specialising the model for depth-based industrial anomaly detection through targeted pre-training on industrial datasets.
            \begin{table}[ht]
    \centering
    \resizebox{0.9\linewidth}{!}{
        \begin{tabular}{cc cc}

            \toprule
        
            \multicolumn{2}{c}{\textbf{Features}} & \multirow{2}{*}{\textbf{Detection}} & \multirow{2}{*}{\textbf{Segmentation}} \\
            
            \cmidrule(lr){1-2}
            
            Image & Depth & & \\ 

            \midrule
            
            DINO-v3 & DINO-v3    & 0.534 & 0.569 \\
            DINO-v2 & DINO-v2    & \textbf{0.575} & \textbf{0.684} \\
            
            \bottomrule
            
        \end{tabular}}
    \caption{
        \textbf{Effects of Image Feature Extractor.} 
        }
    \label{tab:image_encoder}
\end{table}
            
        \paragraph{Aggregation Function.}
            In~\cref{subsec:volume_aggregation} of the main paper, we aggregate the per-view anomaly maps $\Psi^t_I$ and $\Psi^t_D$ by projecting them separately into 3D space and taking the maximum score at each voxel.
            Here, we investigate alternative strategies to aggregate the two modality maps before 3D projection. 
            Specifically, for each view $t$, we compute a unified anomaly map $\Psi^t$ by combining $\Psi^t_I$ and $\Psi^t_D$ using different aggregation functions $\Psi^t = \Xi(\Psi^t_I, \Psi^t_D)$, where $\Xi \in \{\max, \min, \product, \mean \}$ operates element-wise on the 2D maps. 
            The unified maps are then projected into 3D space to construct the anomaly volume.
            
            \begin{table*}[ht]
    \centering
    \resizebox{\linewidth}{!}{
        \begin{tabular}{c ccccccccc ccccccccc}

            \toprule
        
            \multirow{2}{*}{\textbf{{Aggregation}}} & \multicolumn{9}{c}{\textbf{Detection}} & \multicolumn{9}{c}{\textbf{Segmentation}} \\
            
            \cmidrule(lr){2-10} \cmidrule(lr){11-19}
            
            & Pl. Stool & Rub. Bin & W. Vase & B. Furn. & Cont. & Pl. Vase & W. Stool & Sink Cab. & Mean & Pl. Stool & Rub. Bin & W. Vase & B. Furn. & Cont. & Pl. Vase & W. Stool & Sink Cab. & Mean \\ 

            \cmidrule(lr){1-1} \cmidrule(lr){2-10} \cmidrule(lr){11-19}
                        
            $\max$  & 0.909 & 0.990 & 0.945 & 0.647 & 0.740 & 0.607 & 0.916 & 1.000 & \textbf{0.844} & 0.855 & 0.707 & 0.863 & 0.791 & 0.831 & 0.885 & 0.711 & 0.785 & \textbf{0.804} \\
            $\min$ & 0.690 & 1.000 & 1.000 & 0.500 & 0.455 & 0.606 & 0.520 & 1.000 & 0.721 & 0.292 & 0.412 & 0.847 & 0.408 & 0.623 & 0.874 & 0.089 & 0.061 & 0.451 \\
            $\product$ & 0.700 & 1.000 & 0.972 & 0.568 & 0.520 & 0.618 & 0.583 & 1.000 & 0.745 & 0.563 & 0.634 & 0.881 & 0.635 & 0.710 & 0.869 & 0.365 & 0.471 & 0.641 \\
            $\mean$ & 0.736 & 1.000 & 0.972 & 0.556 & 0.669 & 0.613 & 0.708 & 1.000 & 0.782 & 0.775 & 0.677 & 0.878 & 0.749 & 0.794 & 0.865 & 0.569 & 0.740 & 0.756 \\
            
            \bottomrule
            
        \end{tabular}}
    \caption{
        \textbf{Aggregation Function.} 
        }
    \label{tab:aggregation_fcn}

\end{table*}
            \cref{tab:aggregation_fcn} presents the results on the real-to-real setup. 
            The maximum aggregation achieves the best performance with $0.844$ I-AUROC and $0.804$ V-AUPRO@$1$\%, validating our design choice. 
            The minimum aggregation performs poorly ($0.721$ I-AUROC, $0.451$ V-AUPRO@$1$\%), as it requires both modalities to detect an anomaly, significantly reducing sensitivity. 
            This is particularly problematic for anomalies visible in only one modality (e.g., colour defects in the image or geometric defects in depth). 
            The product and average aggregations show intermediate performance but still underperform the maximum by $9.9$\% and $6.2$\% in detection, and $16.3$\% and $4.8$\% in segmentation, respectively.
            
            These results confirm that the maximum aggregation optimally balances the complementary information from both modalities: an anomaly is flagged if detected by either modality, maintaining high sensitivity while leveraging the precision gained from our minimum-based cross-view ensembling.
            
    \subsection{Visualisations}

        \paragraph{Failure Cases}
            \begin{figure}[t]
    \centering
        \includegraphics[width=\linewidth]{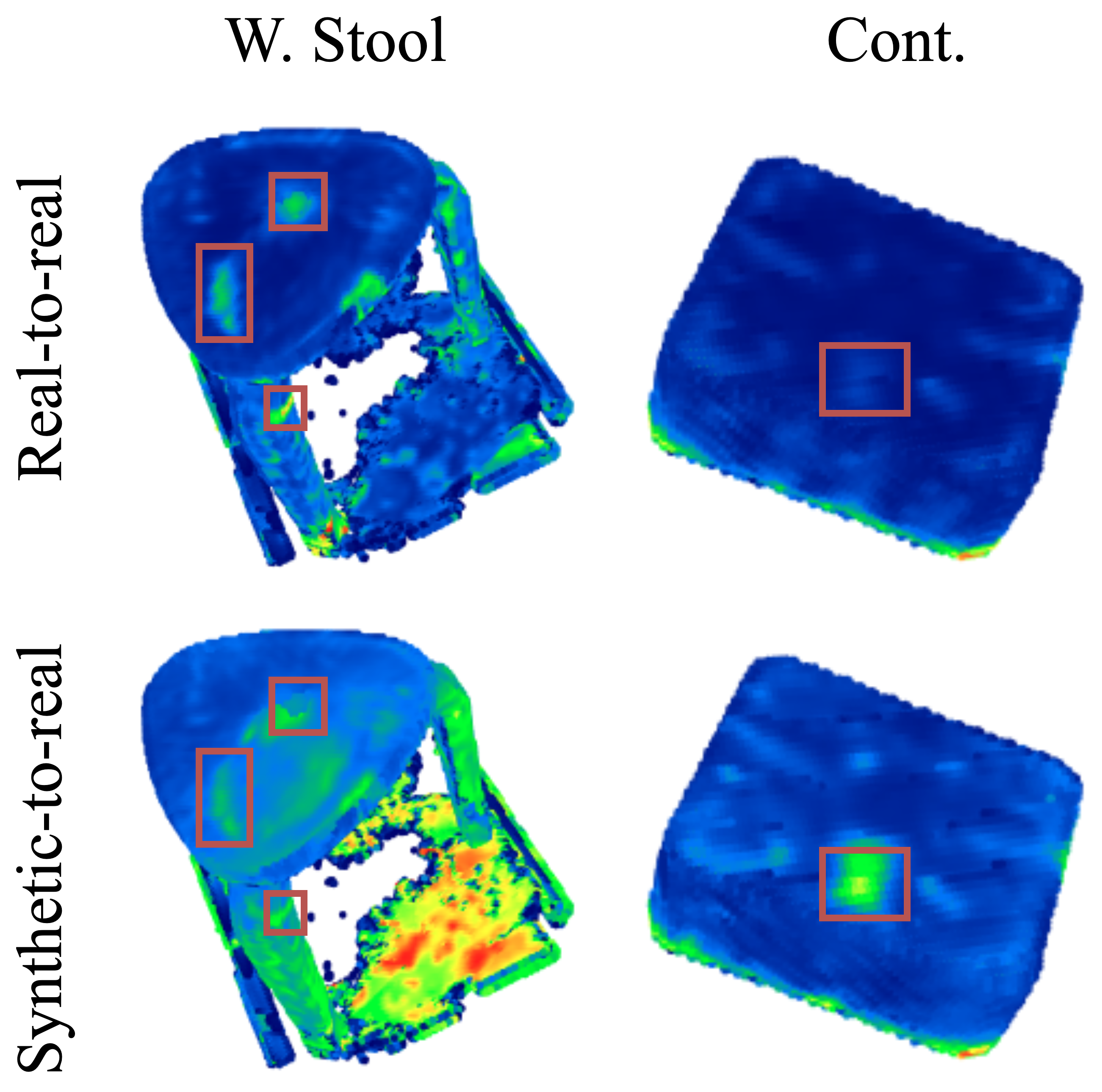}
    \caption{
        \textbf{\algoname{} Failure Cases.}
    }
    \label{fig:failure}
\end{figure}

            \cref{fig:failure} illustrates failure cases of \algoname{}. 
            In the left example, genuine anomalies (\textcolor{custom_red}{\textbf{red boxes}}) are overshadowed by spurious peaks arising from unfiltered background regions. 
            This phenomenon is exacerbated in the synthetic-to-real scenario (bottom-left), where the model exhibits particularly pronounced false activations in the unfiltered background due to the domain gap.
            Indeed, the synthetic training data lacks realistic background appearance, causing the model to misinterpret background variations as anomalies.
            The right example demonstrates the opposite behaviour: while the real-to-real model (top-right) entirely misses the anomaly, the synthetic-to-real model (bottom-right) correctly identifies and localises the defect.
            This contrasting behaviour highlights the trade-off between training data fidelity and generalisation. 
            The real-to-real model achieves lower false positive rates by tightly fitting the real normal distribution, but may consequently miss subtle anomalies. 
            The synthetic-to-real model, trained without access to real appearance patterns, maintains higher sensitivity to deviations but struggles to discriminate variations from true defects.
            
        %\paragraph{Additional Qualitatives}
            %We provide extensive qualitative comparisons between real-to-real and synthetic-to-real scenarios in the Supplementary Materials. 
            %Interactive visualisations can be found in \texttt{qualitatives\_A.html} and \texttt{qualitatives\_B.html}, which showcase additional segmentation results across various object categories and anomaly types.
            %These visualisations can be freely zoomed, rotated, and explored to examine anomaly predictions from multiple perspectives.
            %We plan to embed these visualisations on the project page upon acceptance.

\begin{figure*}[t]
    \centering
    \resizebox{0.75\linewidth}{!}{
    \setlength{\tabcolsep}{2px}
    \begin{tabular}{ccc}
    
        Depth & DINO-v2 & DINO-Depth \\

        \includegraphics[width=0.33\linewidth,angle=180,origin=c]{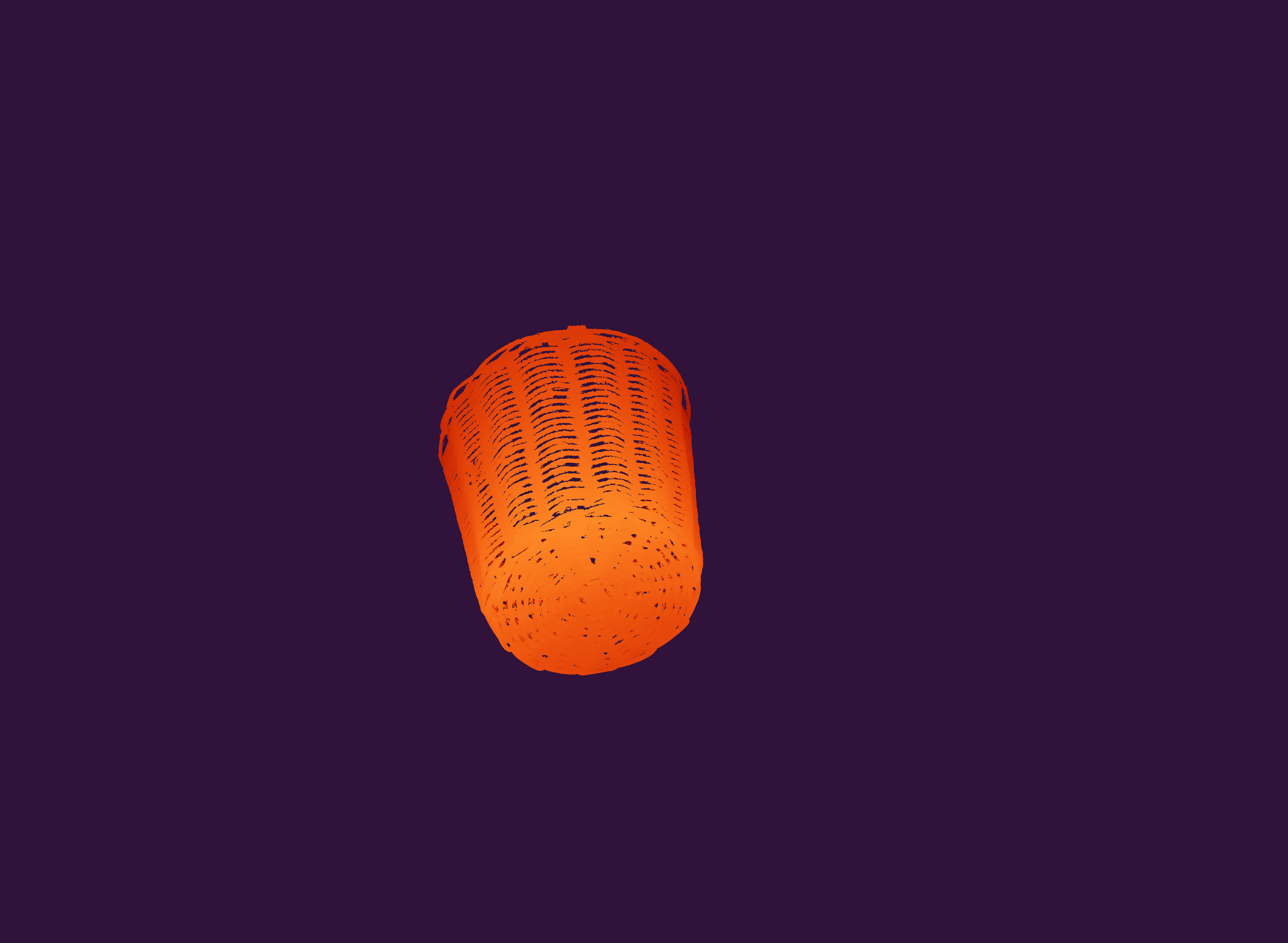} & 
        \includegraphics[width=0.33\linewidth,angle=180,origin=c]{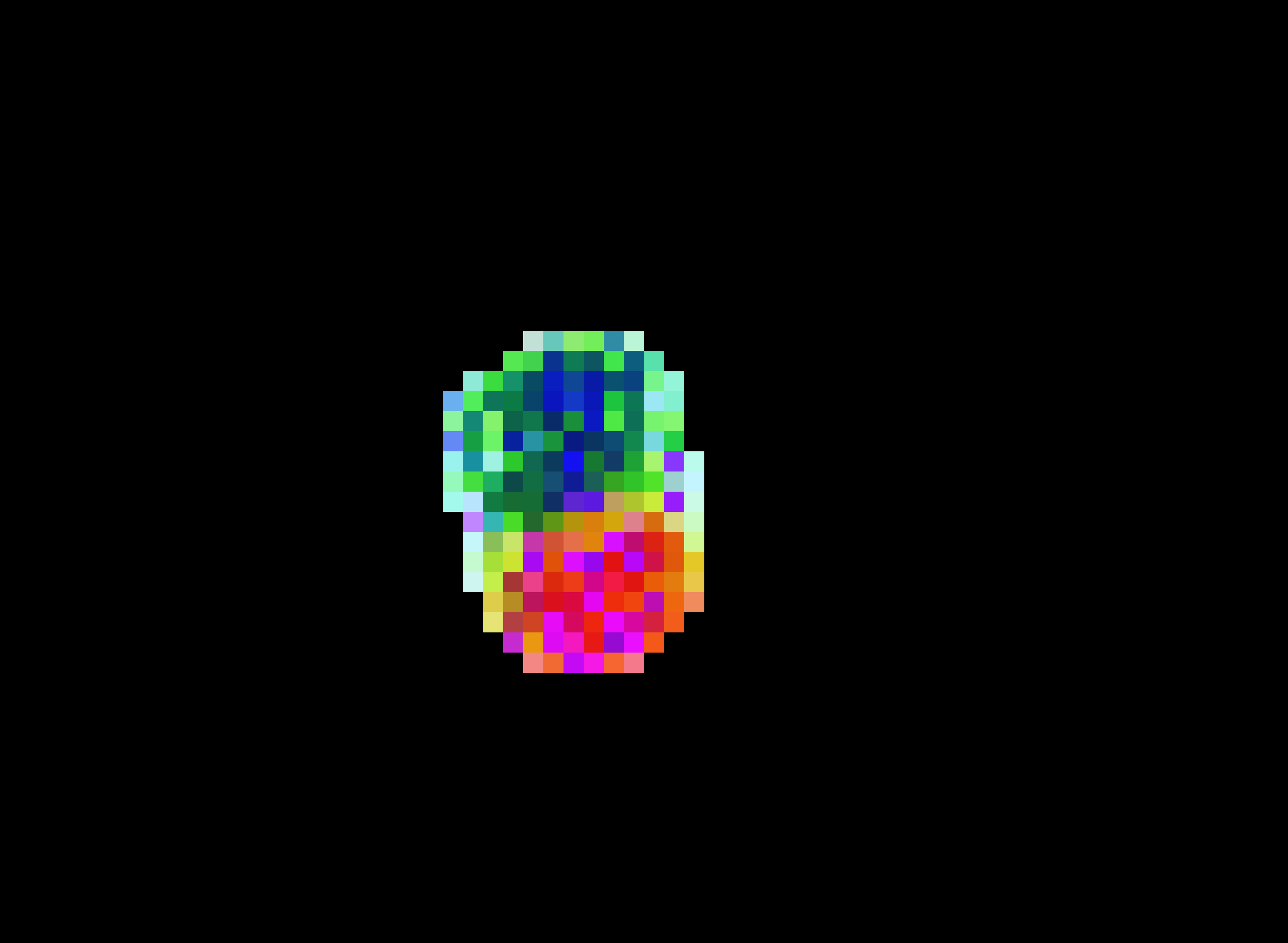} & 
        \includegraphics[width=0.33\linewidth,angle=180,origin=c]{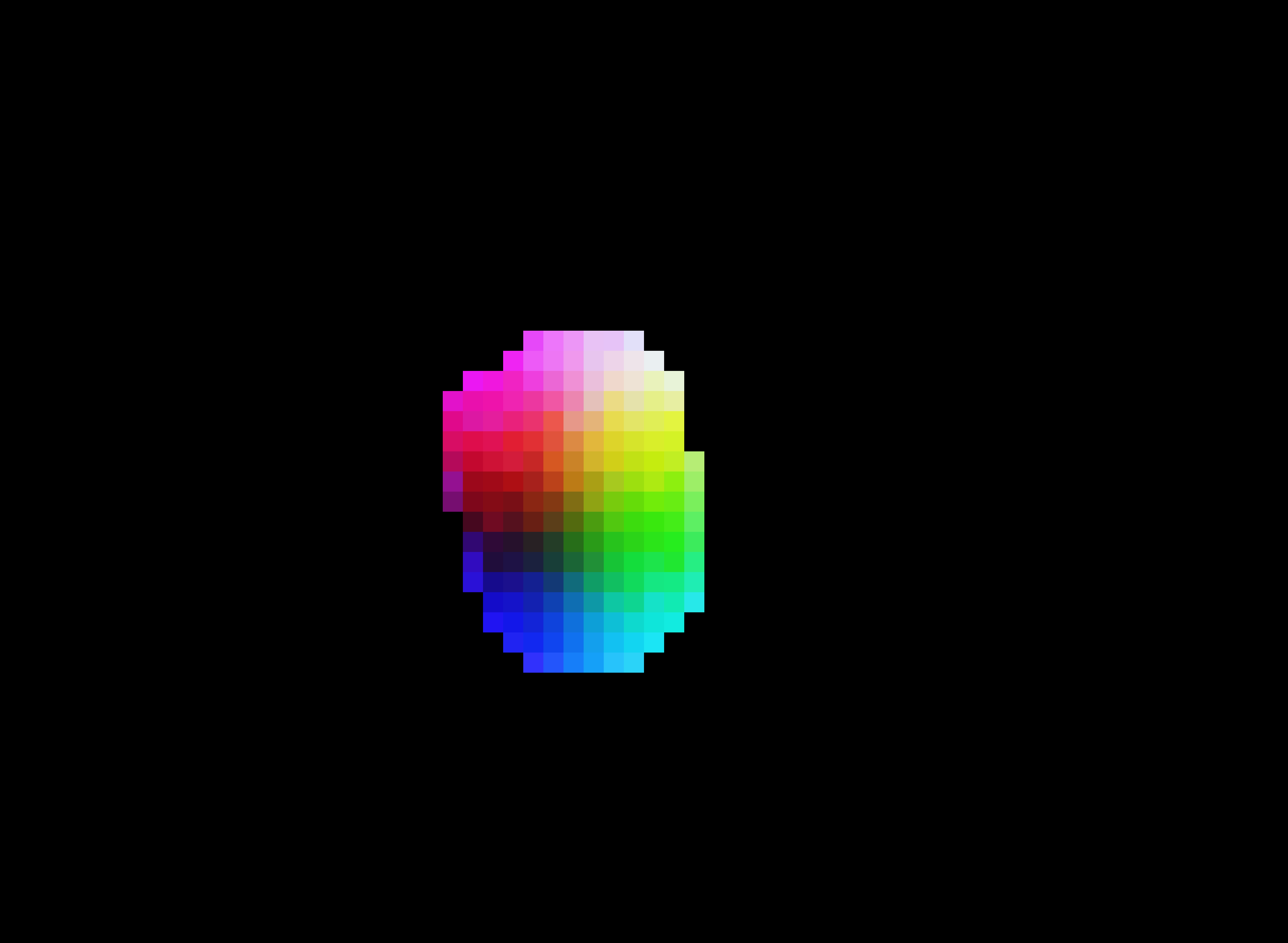} \\

        \includegraphics[width=0.33\linewidth]{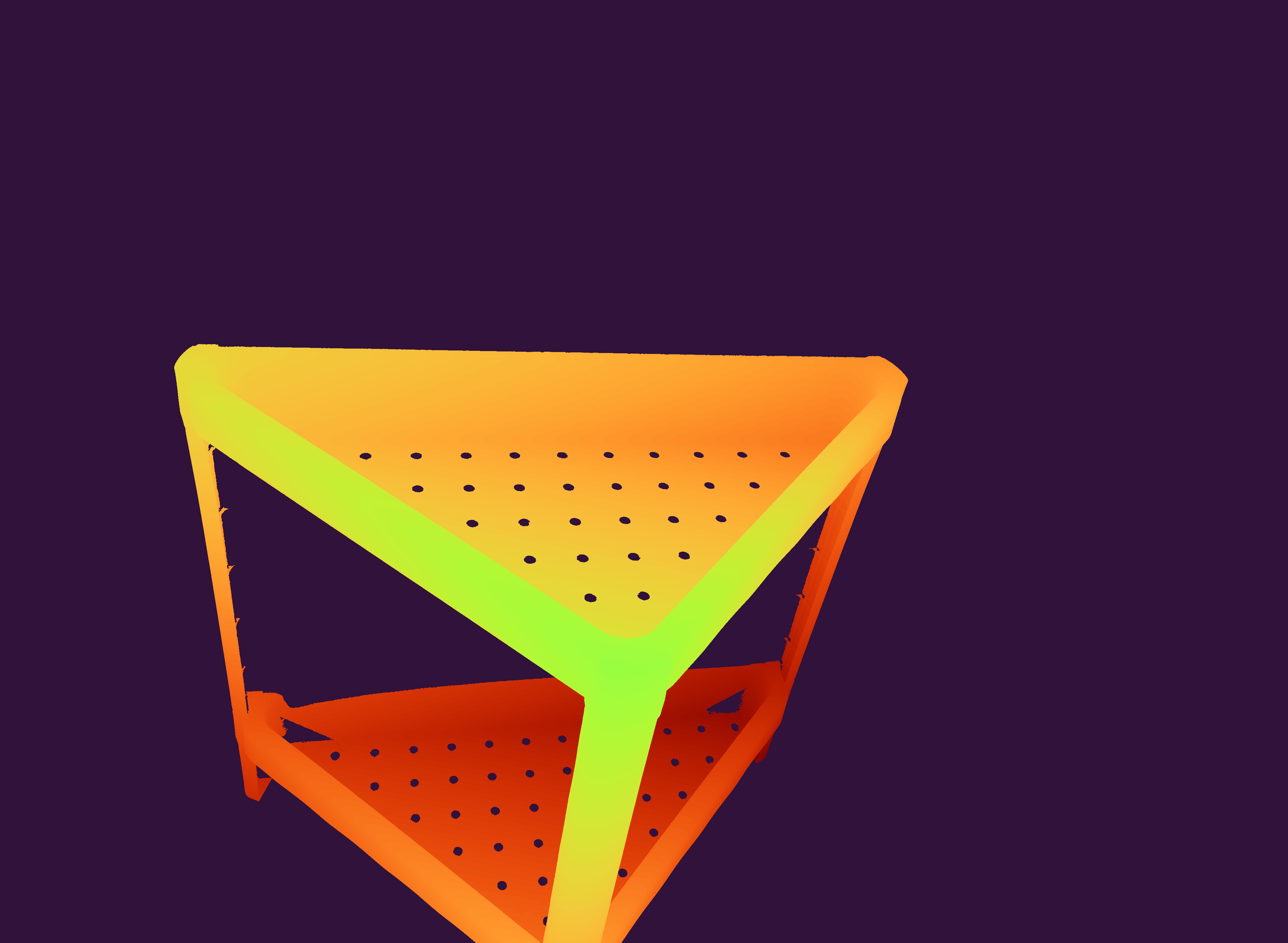} & 
        \includegraphics[width=0.33\linewidth]{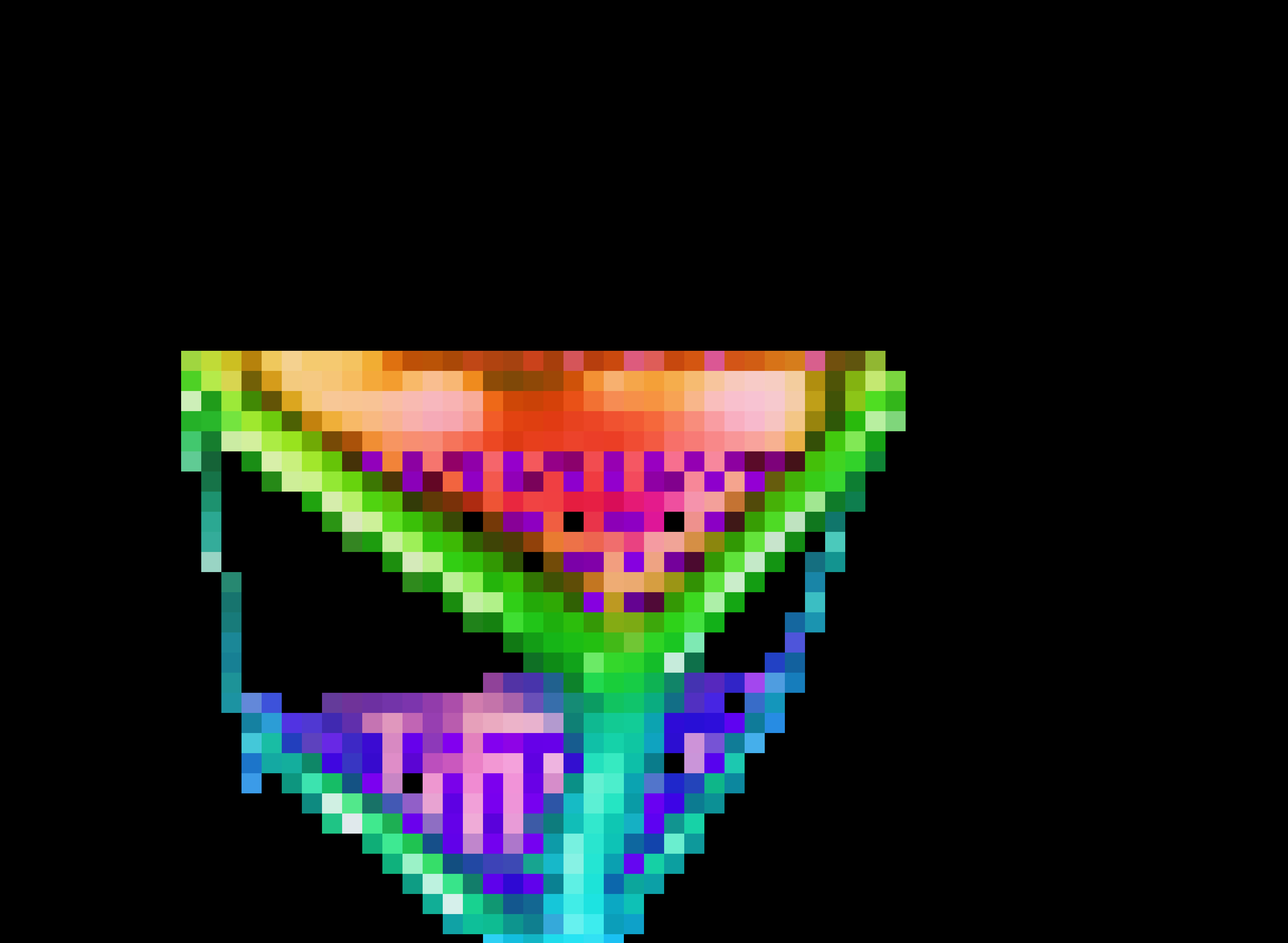} & 
        \includegraphics[width=0.33\linewidth]{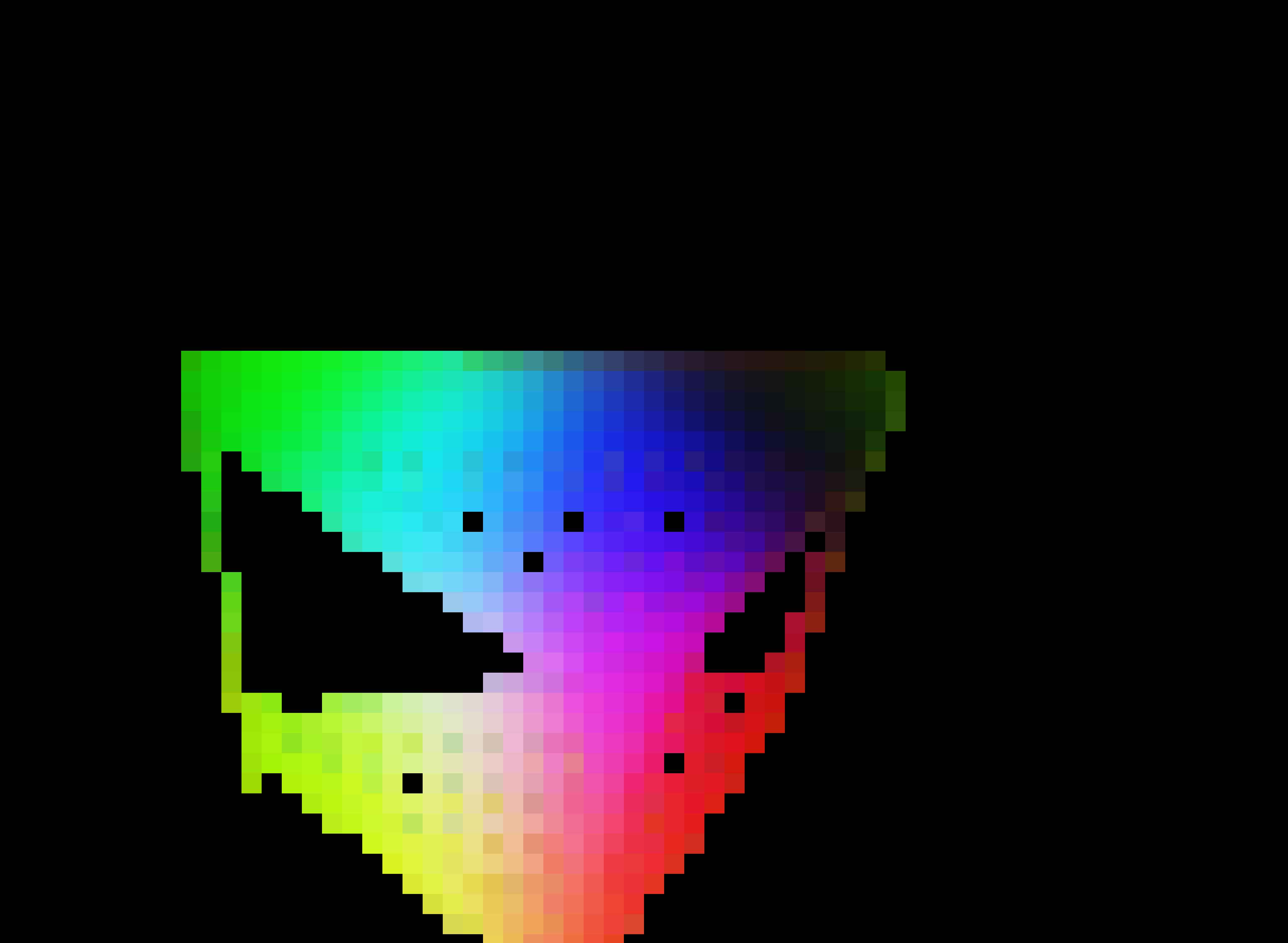} \\

        \includegraphics[width=0.33\linewidth,angle=180,origin=c]{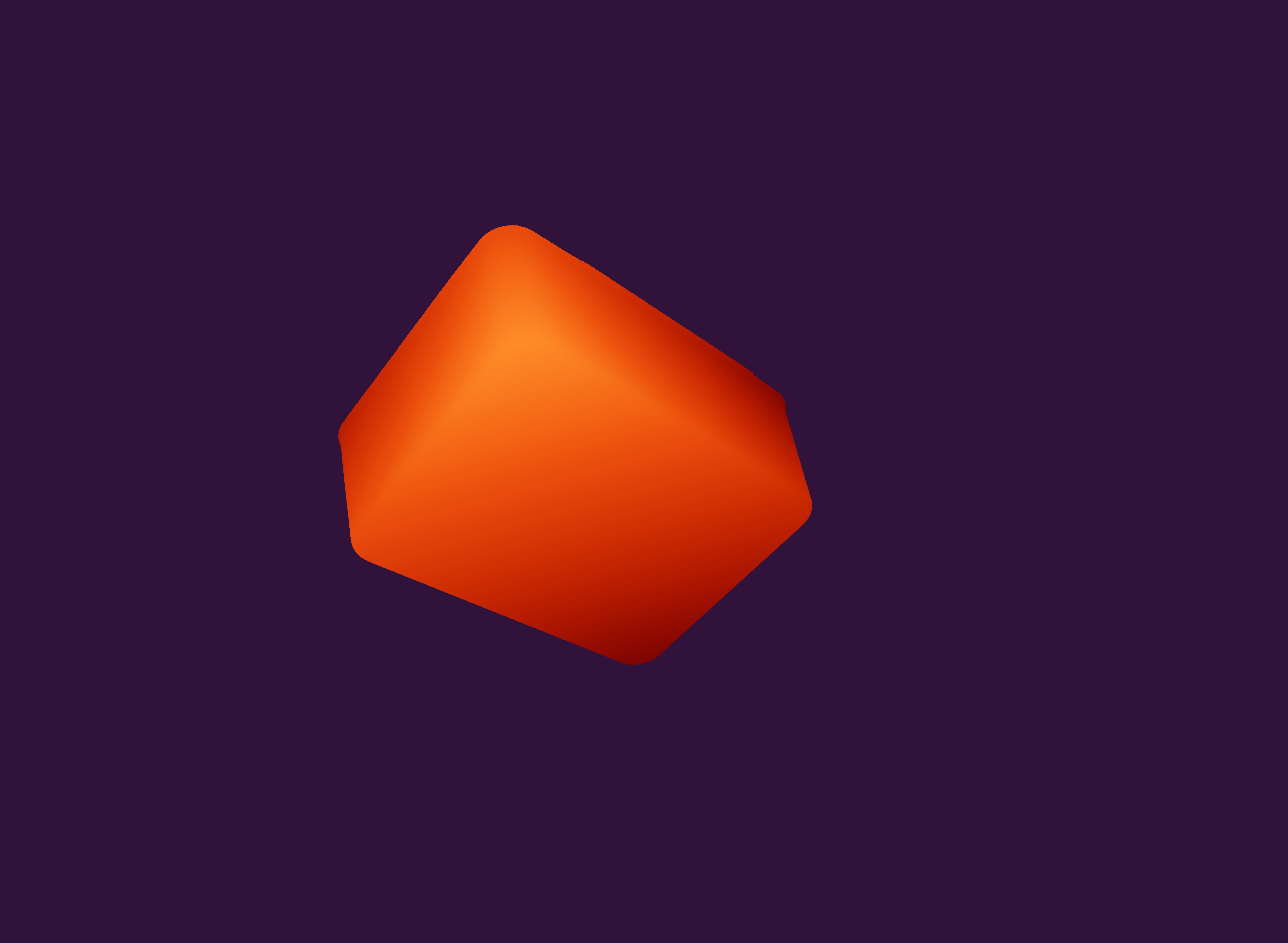} & 
        \includegraphics[width=0.33\linewidth,angle=180,origin=c]{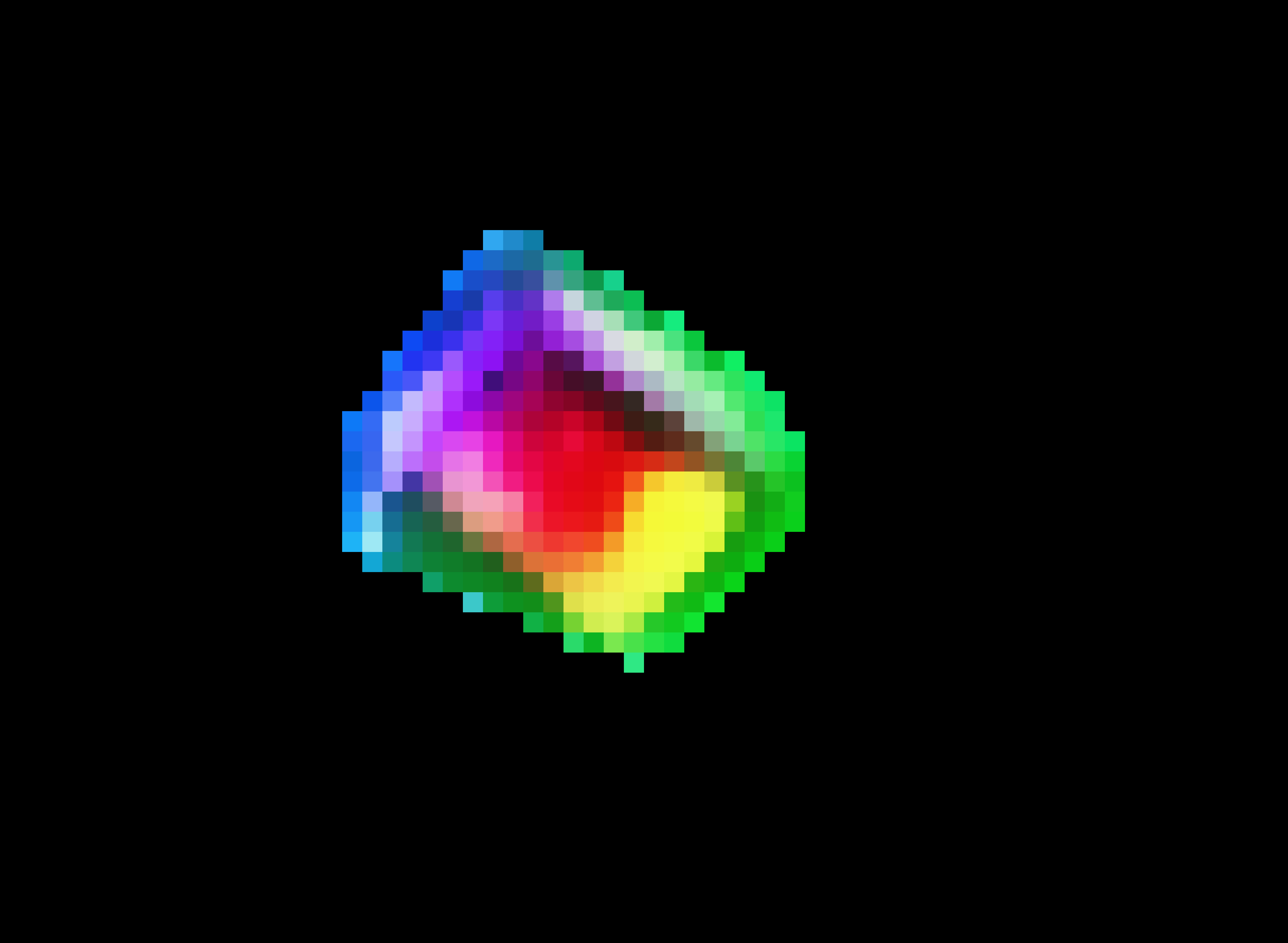} & 
        \includegraphics[width=0.33\linewidth,angle=180,origin=c]{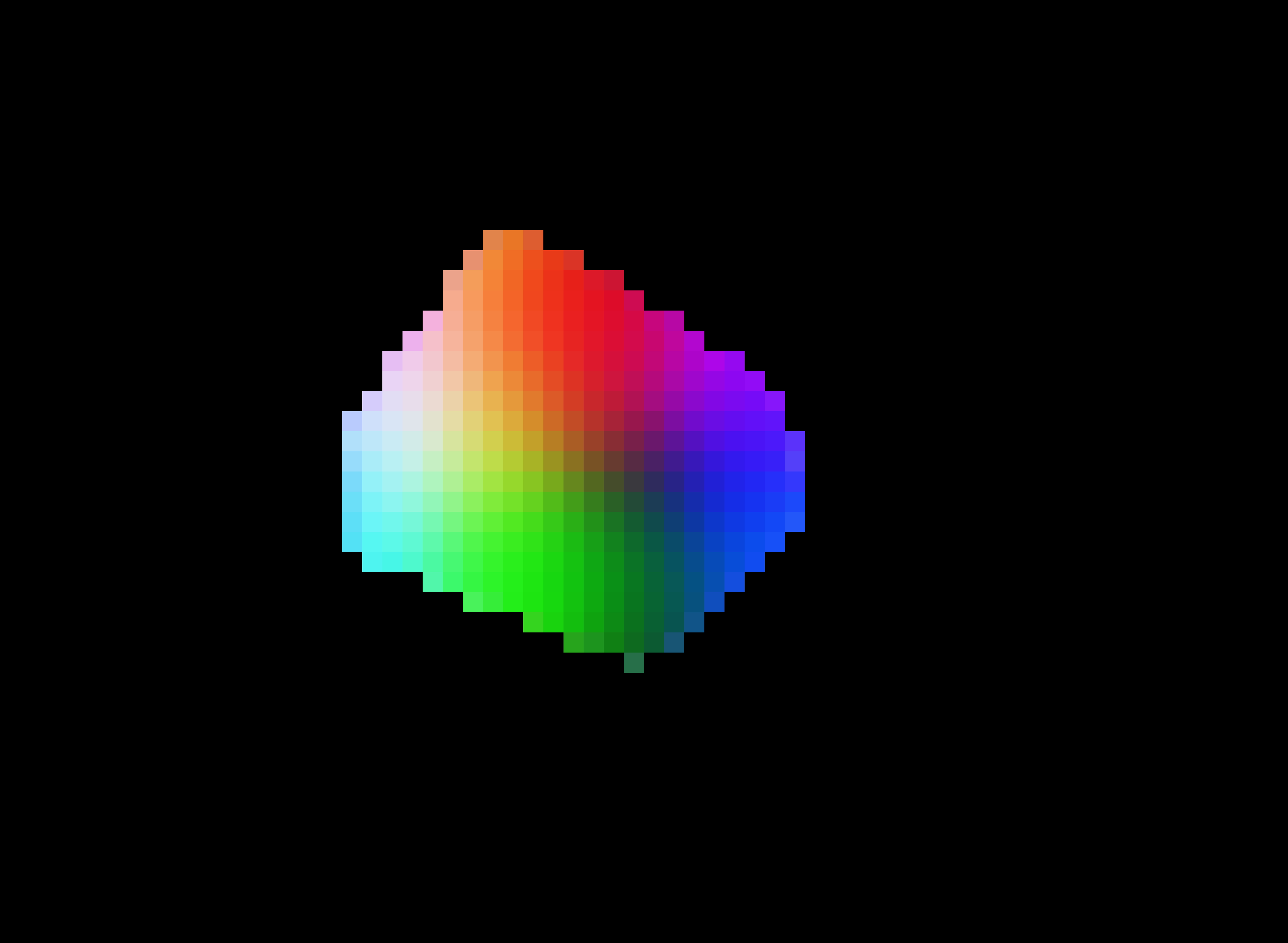} \\

        \includegraphics[width=0.33\linewidth,angle=180,origin=c]{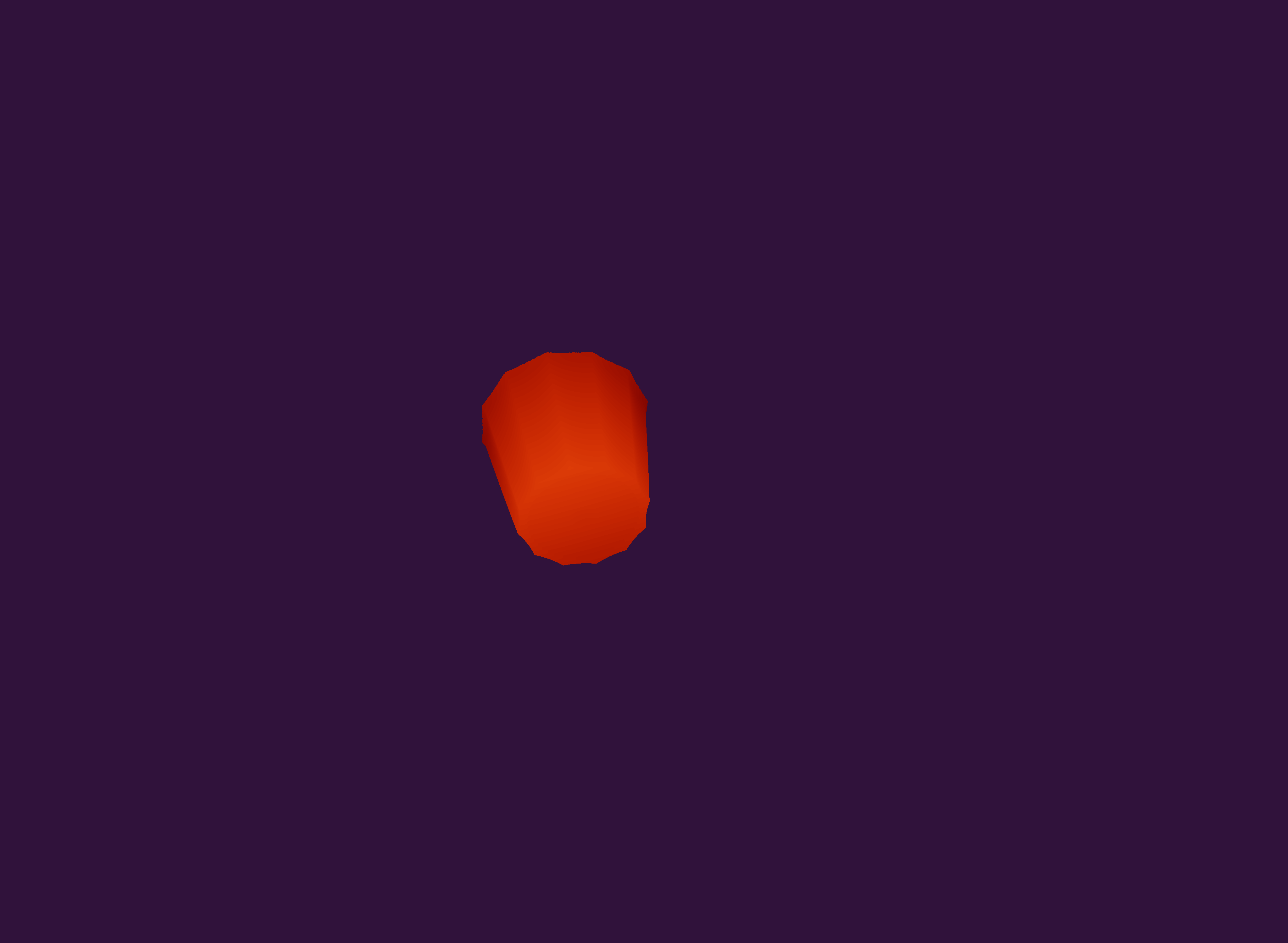} & 
        \includegraphics[width=0.33\linewidth,angle=180,origin=c]{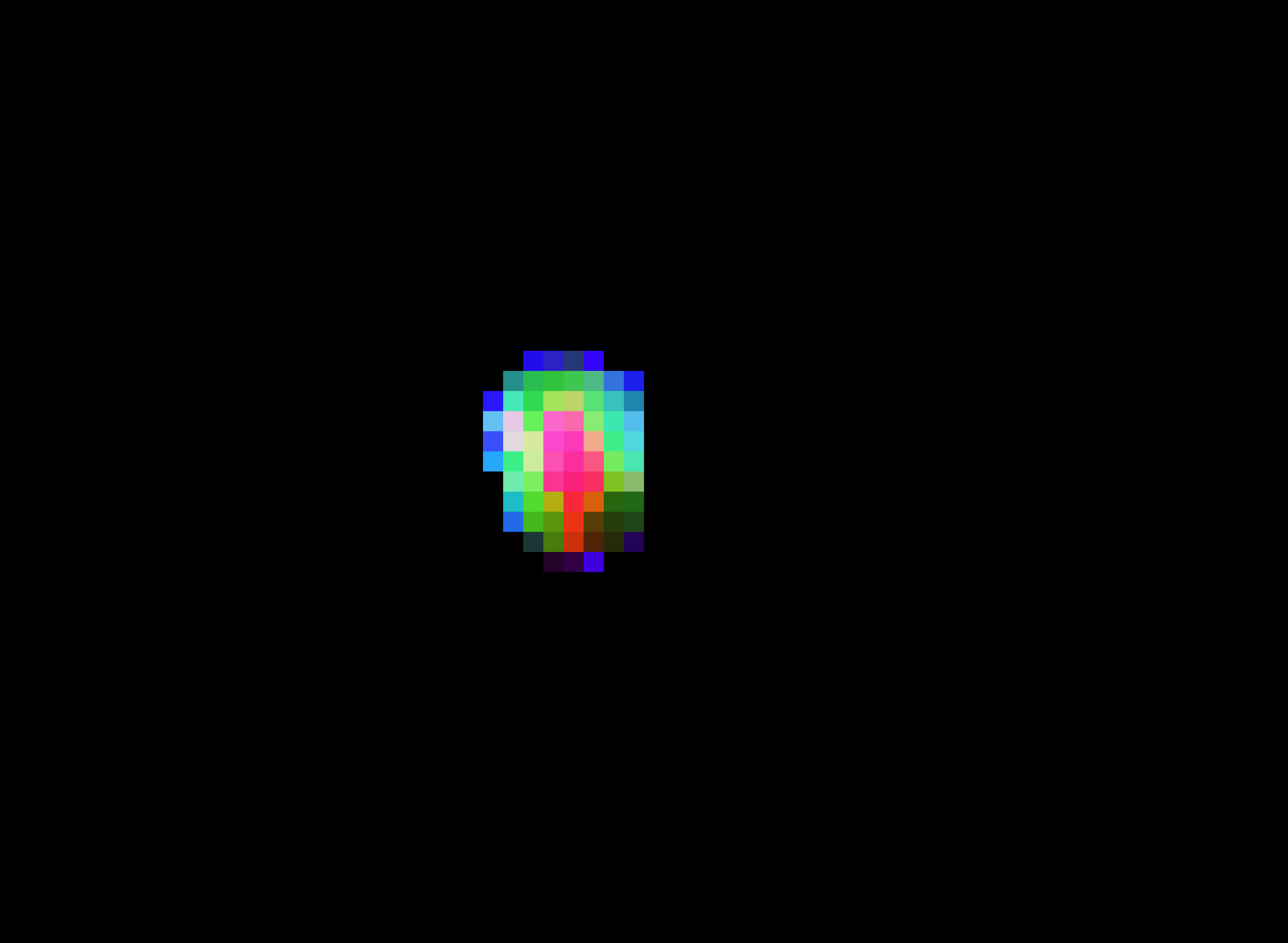} & 
        \includegraphics[width=0.33\linewidth,angle=180,origin=c]{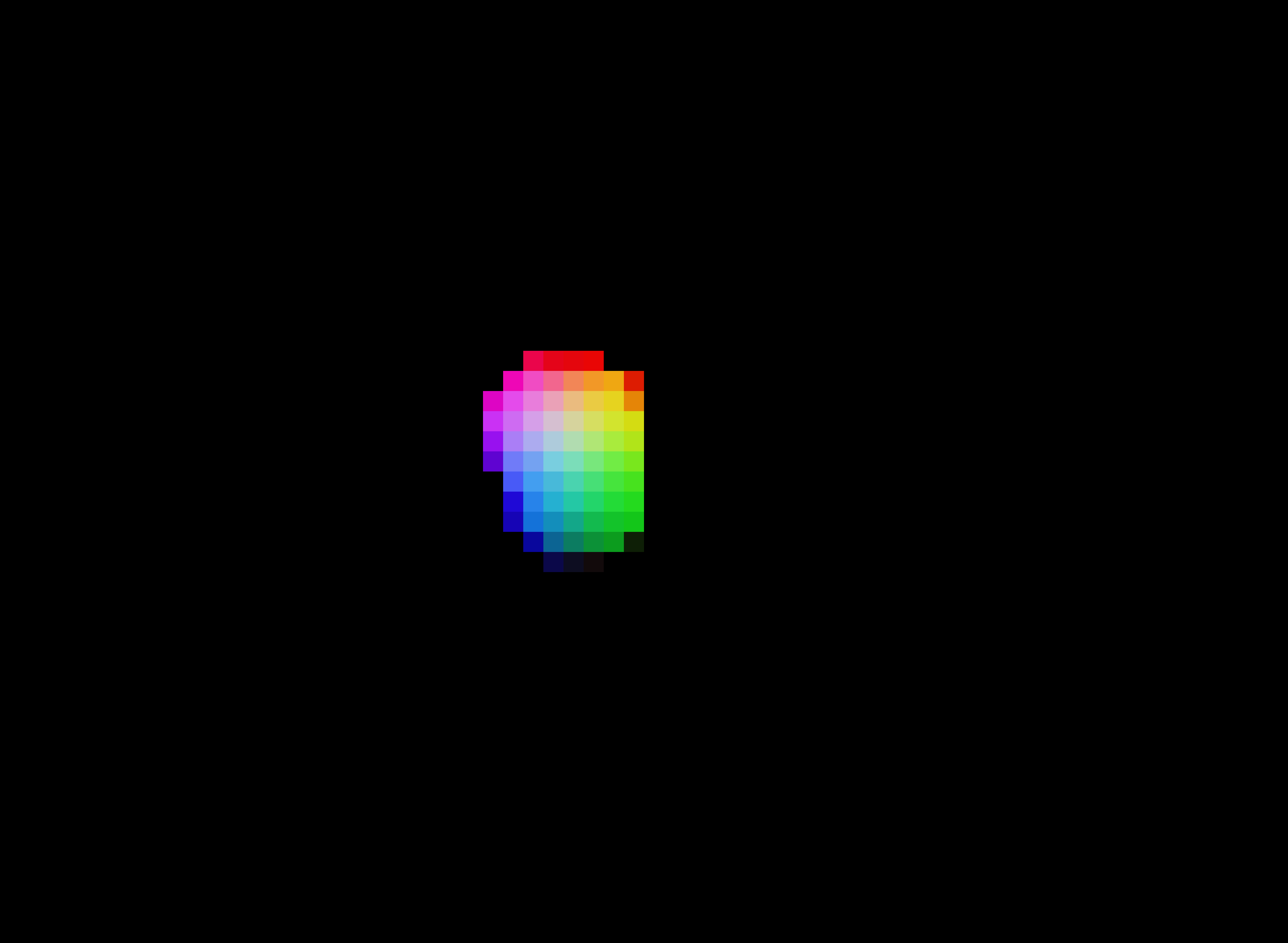} \\

        \includegraphics[width=0.33\linewidth,angle=180,origin=c]{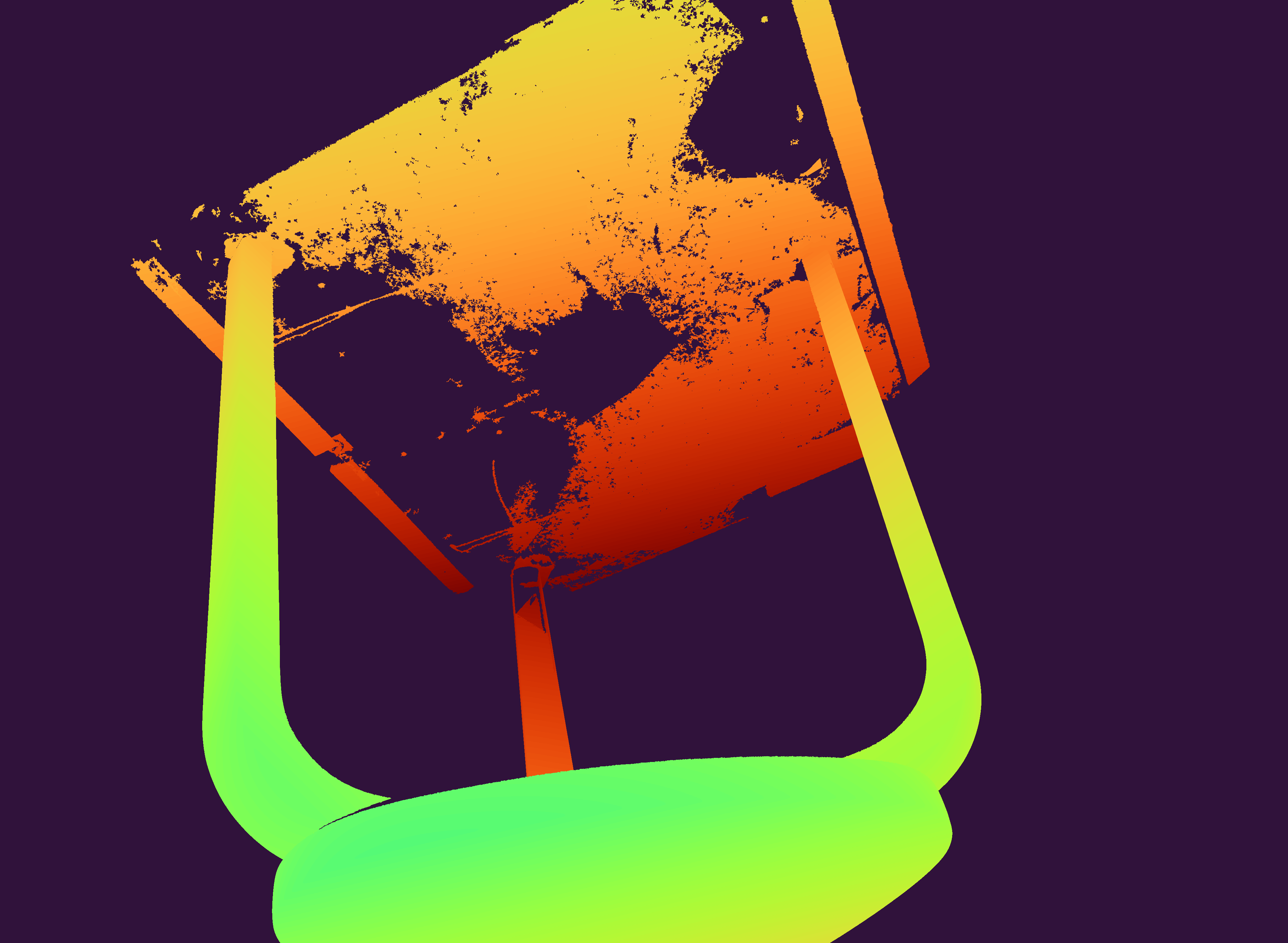} & 
        \includegraphics[width=0.33\linewidth,angle=180,origin=c]{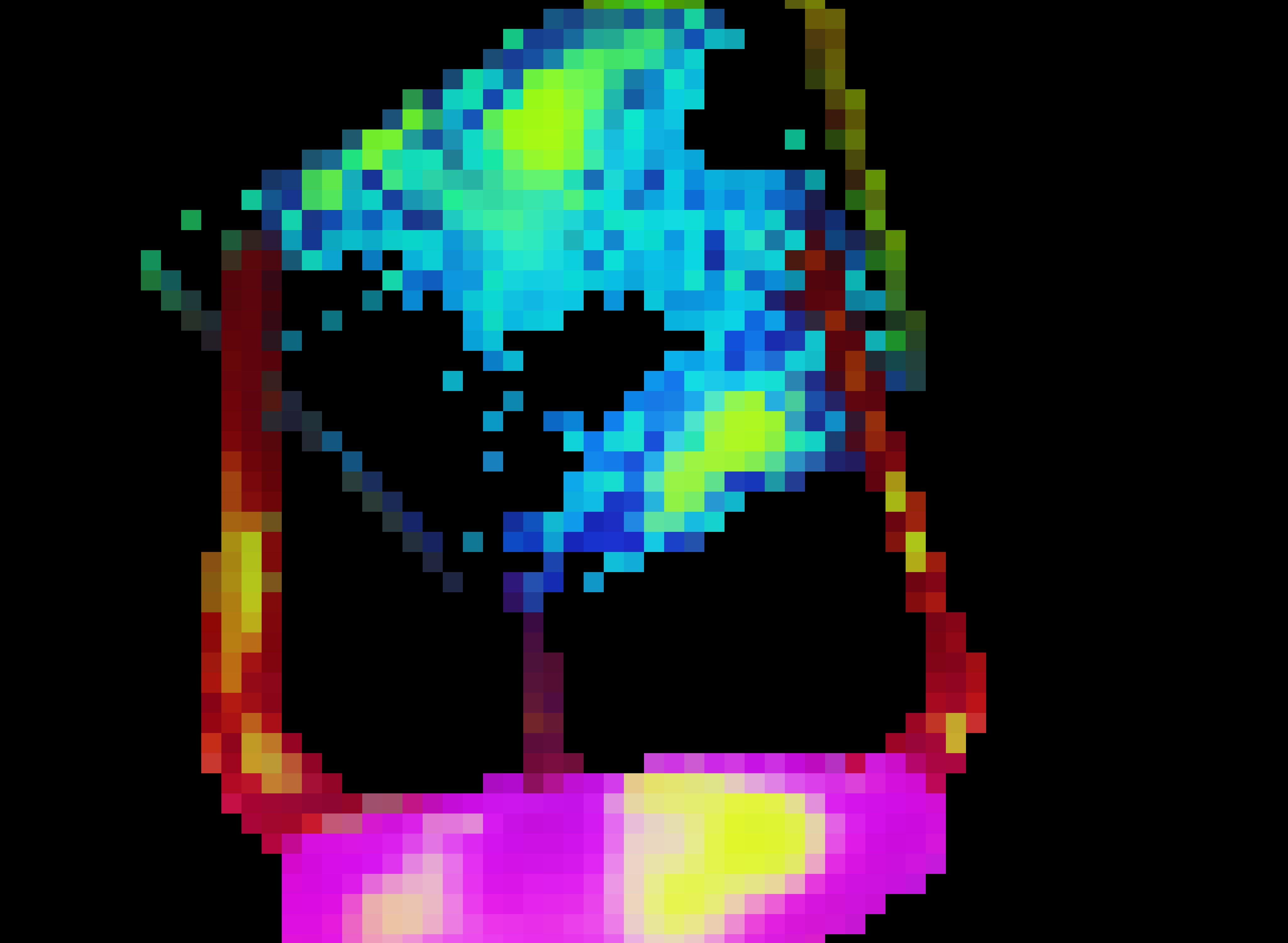} & 
        \includegraphics[width=0.33\linewidth,angle=180,origin=c]{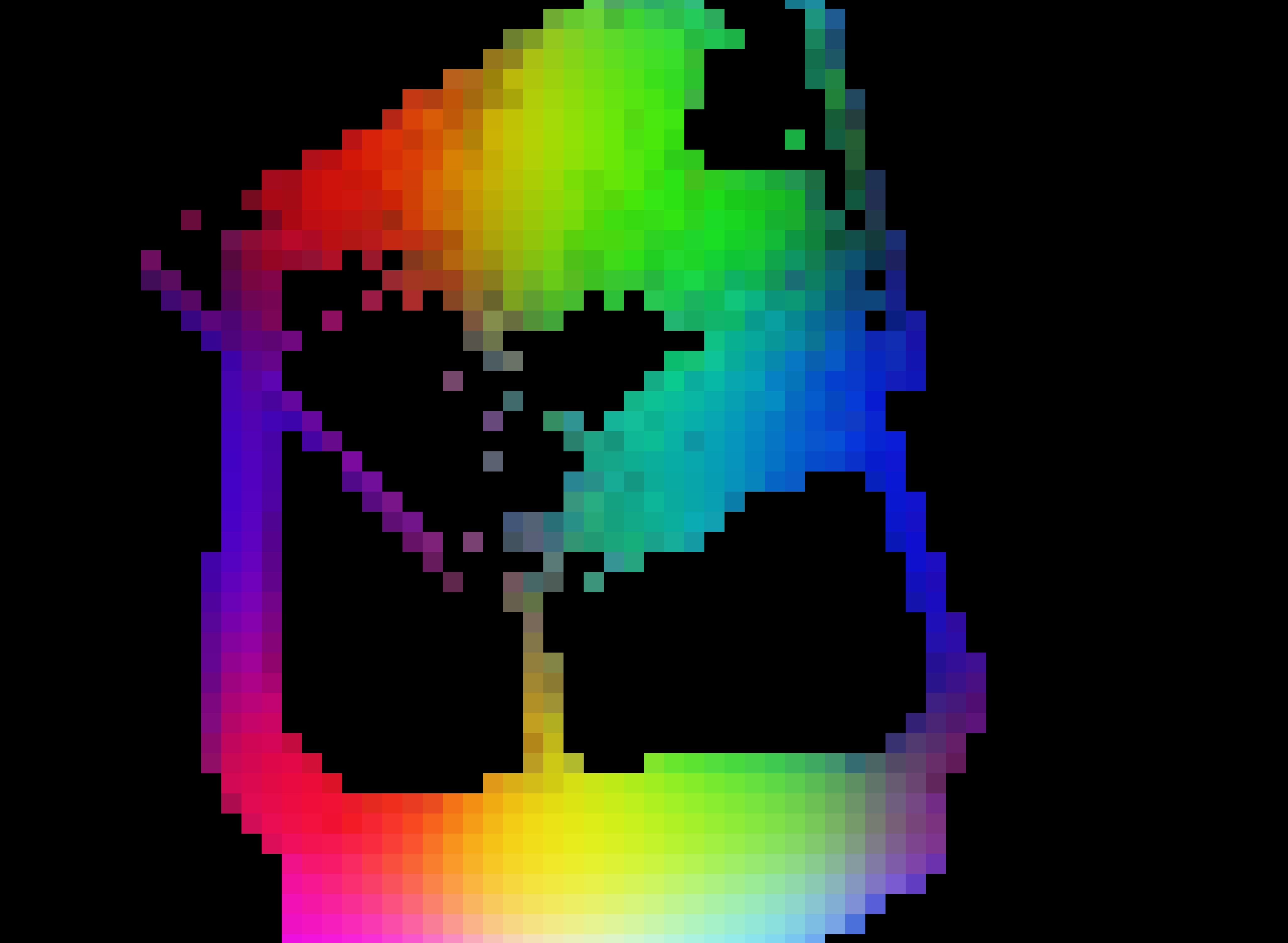} \\

        \includegraphics[width=0.33\linewidth,angle=180,origin=c]{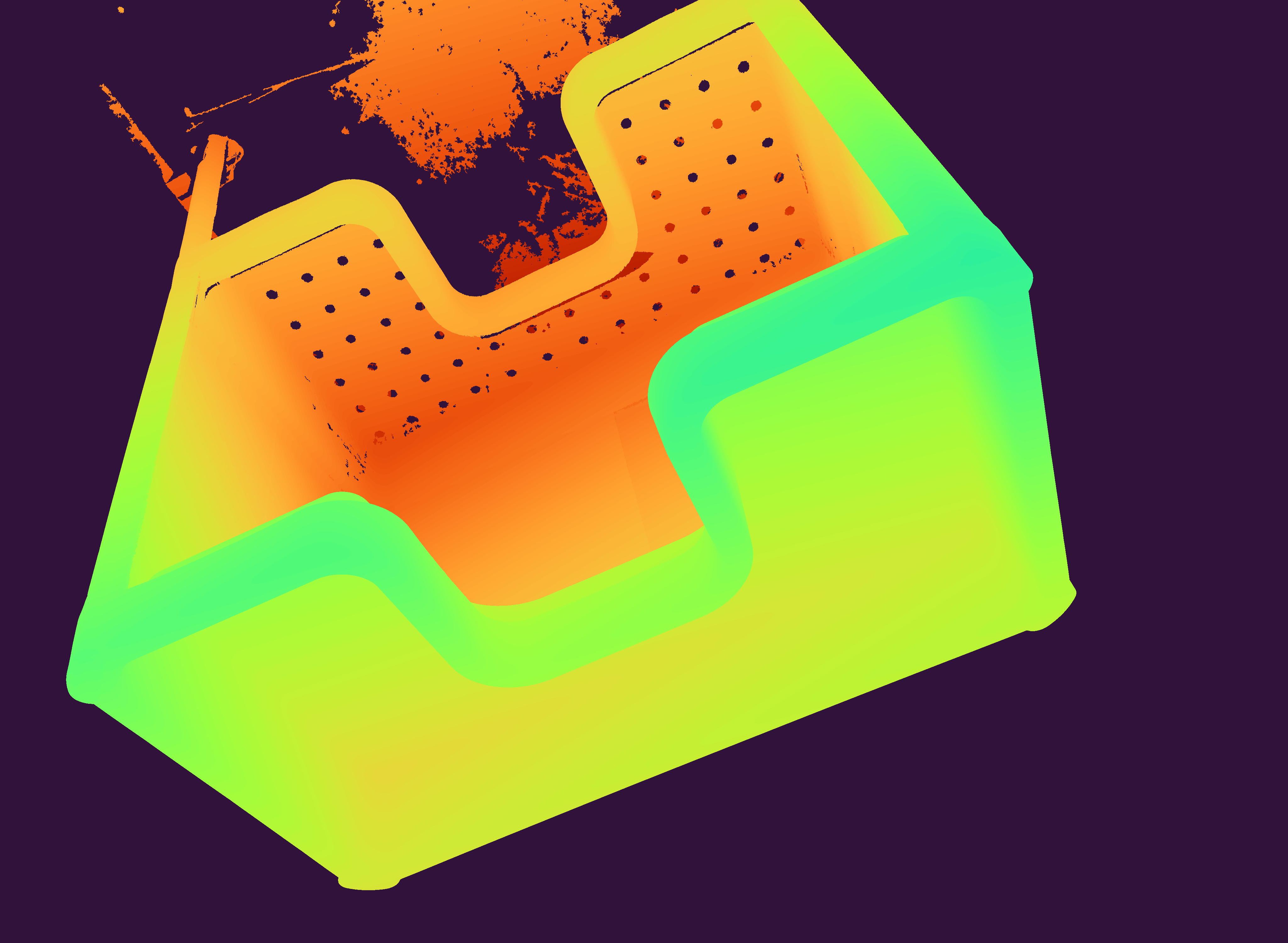} & 
        \includegraphics[width=0.33\linewidth,angle=180,origin=c]{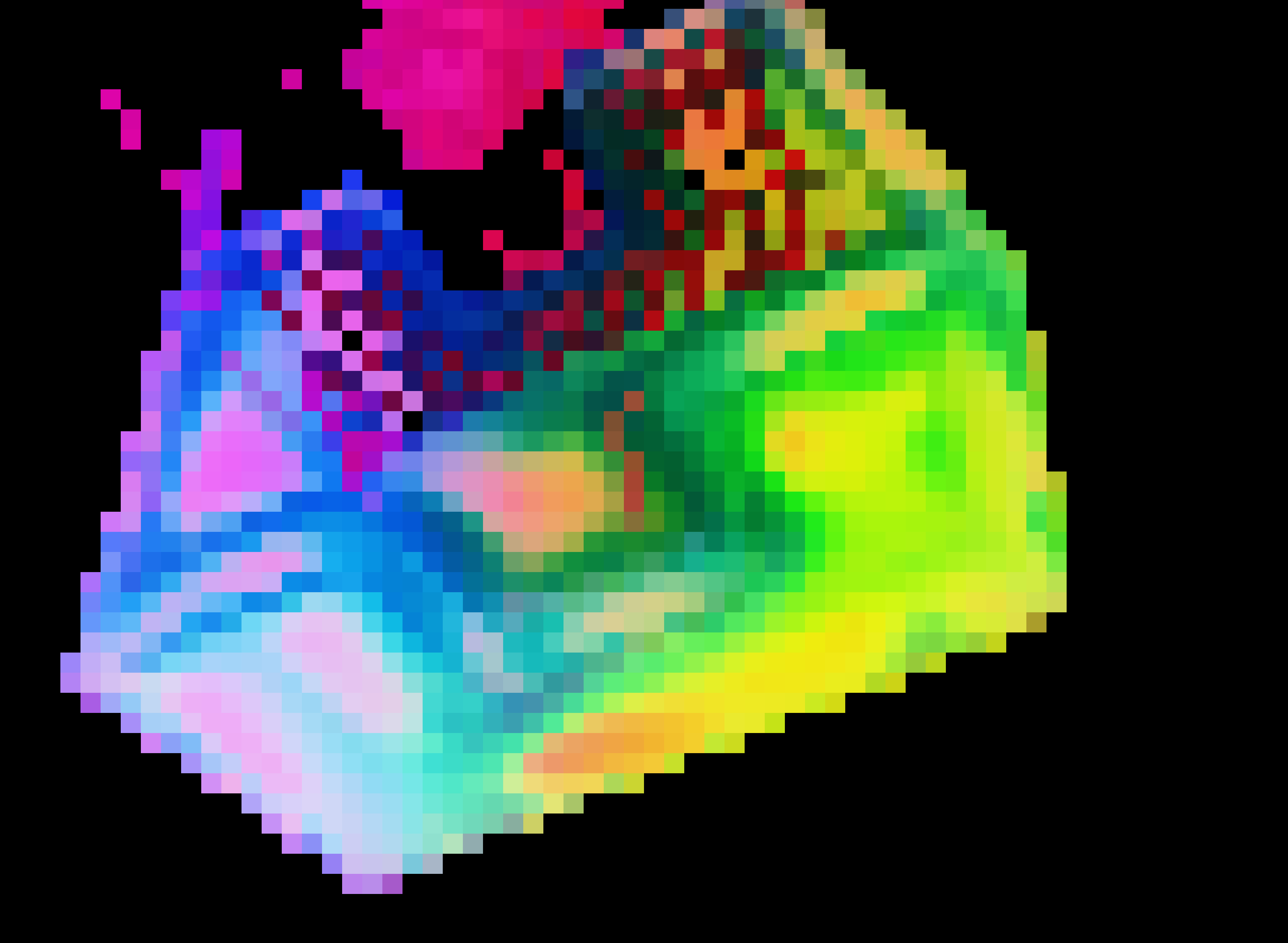} & 
        \includegraphics[width=0.33\linewidth,angle=180,origin=c]{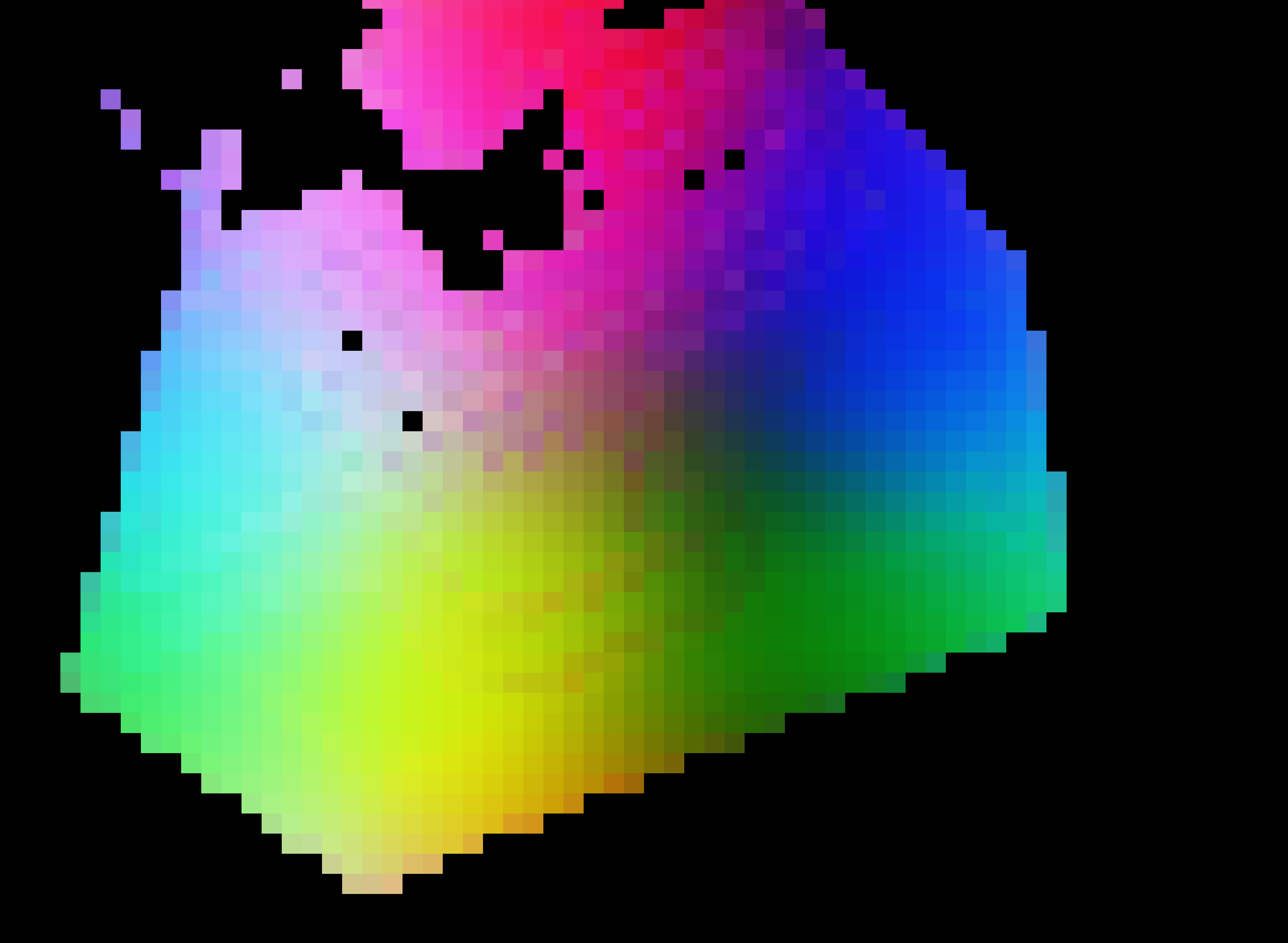} \\

    \end{tabular}
    }
    \caption{\textbf{PCA of Depth Features.}
    }
    \label{fig:depth_features_full}
\end{figure*}

        \paragraph{Additional Depth Features Visualisations}
            We report in \cref{fig:depth_features_full} the comparison between DINO-v2 and DINO-Depth features, also for the classes missing from the main paper.

\begin{figure*}[t]
    \centering
    \resizebox{\linewidth}{!}{
        \setlength{\tabcolsep}{1px}
        \begin{tabular}{cc cccccccccccc}
            & & \multicolumn{12}{c}{\textbf{Target Views}} \\
            & & $v_{1}$ & $v_{2}$ & $v_{3}$ & $v_{4}$ & $v_{5}$ & $v_{6}$ & $v_{7}$ & $v_{8}$ & $v_{9}$ & $v_{10}$ & $v_{11}$ & $v_{12}$ \\

            \multirow{12}{*}{\rotatebox{90}{\textbf{Source Views} \quad \quad \quad \quad \quad \quad \quad \quad \quad \quad \quad \quad \quad \quad \quad \quad \quad \quad \quad \quad}} & \rotatebox{90}{\quad $v_{1}$} &
            \includegraphics[width=0.08\linewidth,angle=180,origin=c]{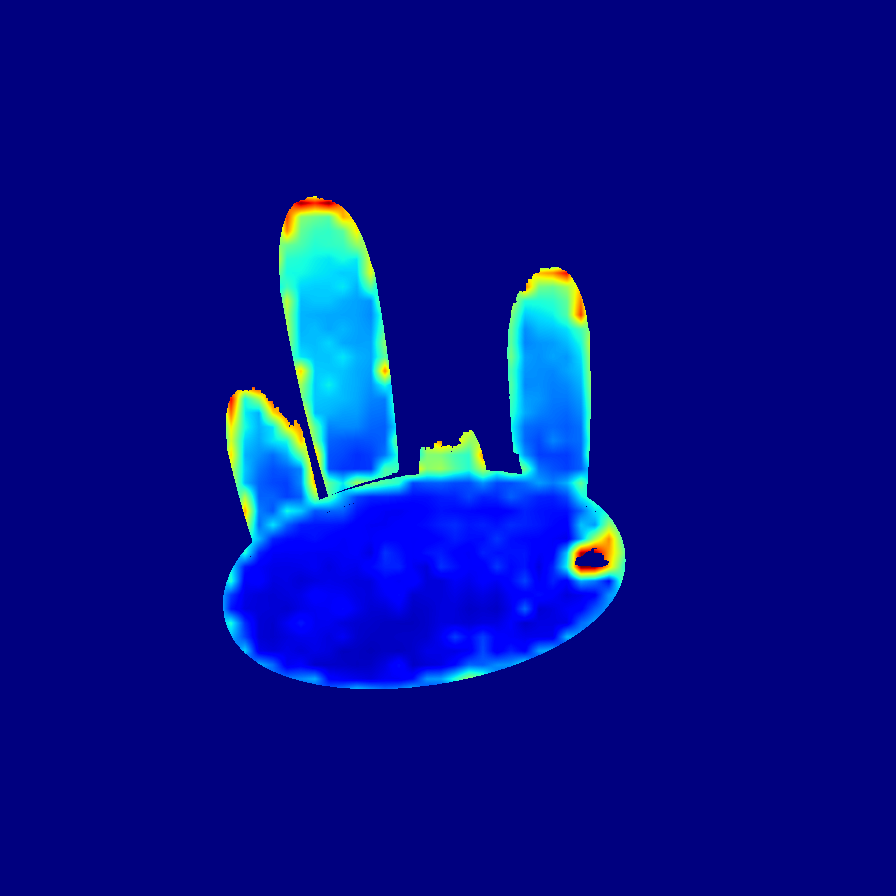} & 
            \includegraphics[width=0.08\linewidth,angle=180,origin=c]{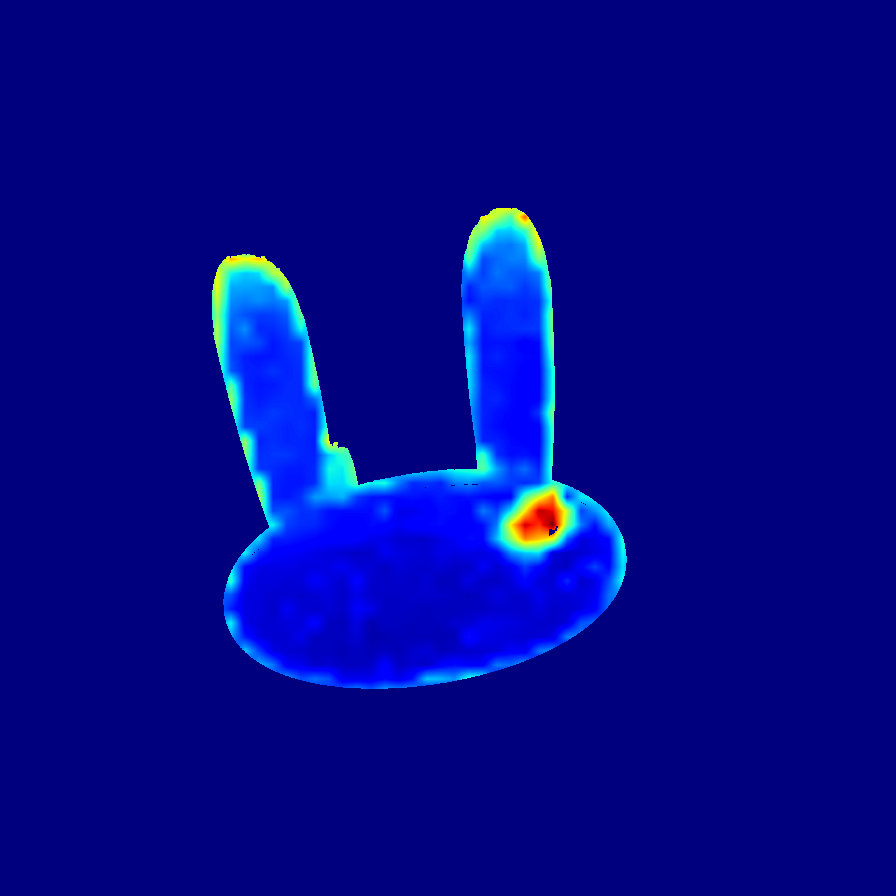} & 
            \includegraphics[width=0.08\linewidth,angle=180,origin=c]{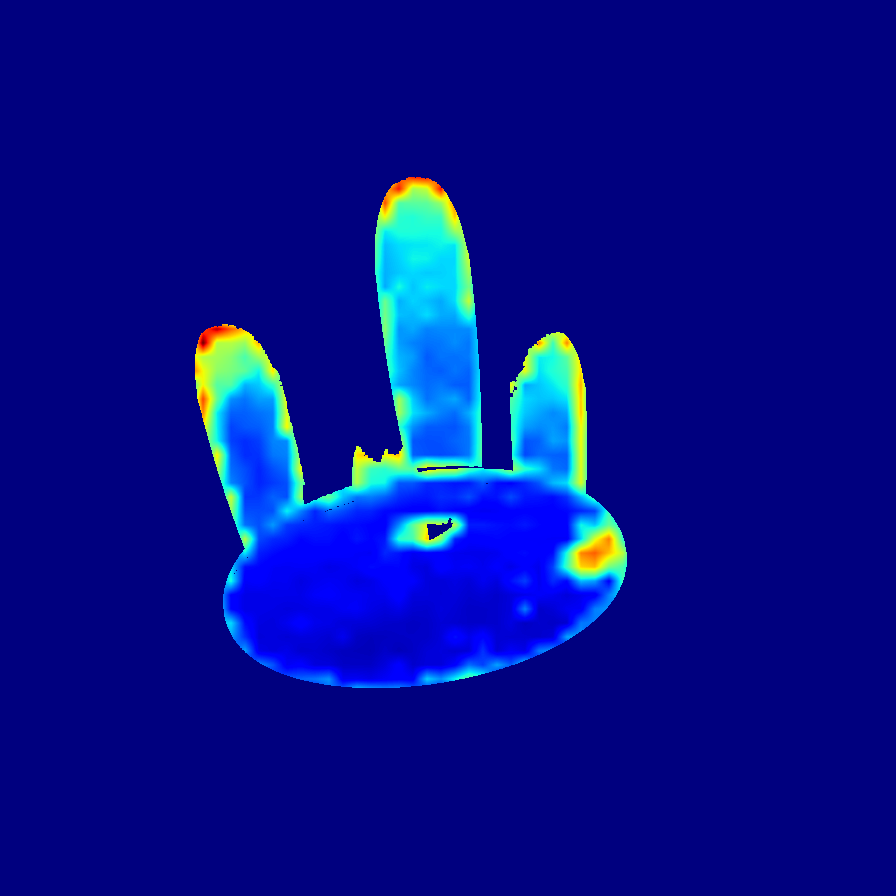} & 
            \includegraphics[width=0.08\linewidth,angle=180,origin=c]{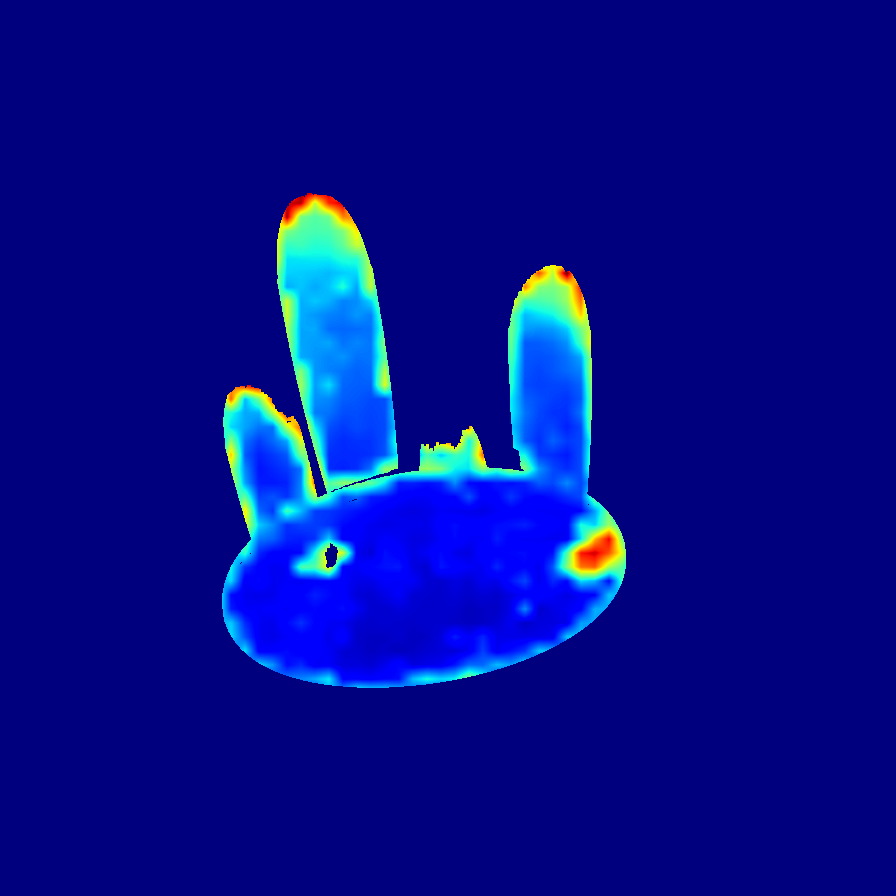} & 
            \includegraphics[width=0.08\linewidth,angle=180,origin=c]{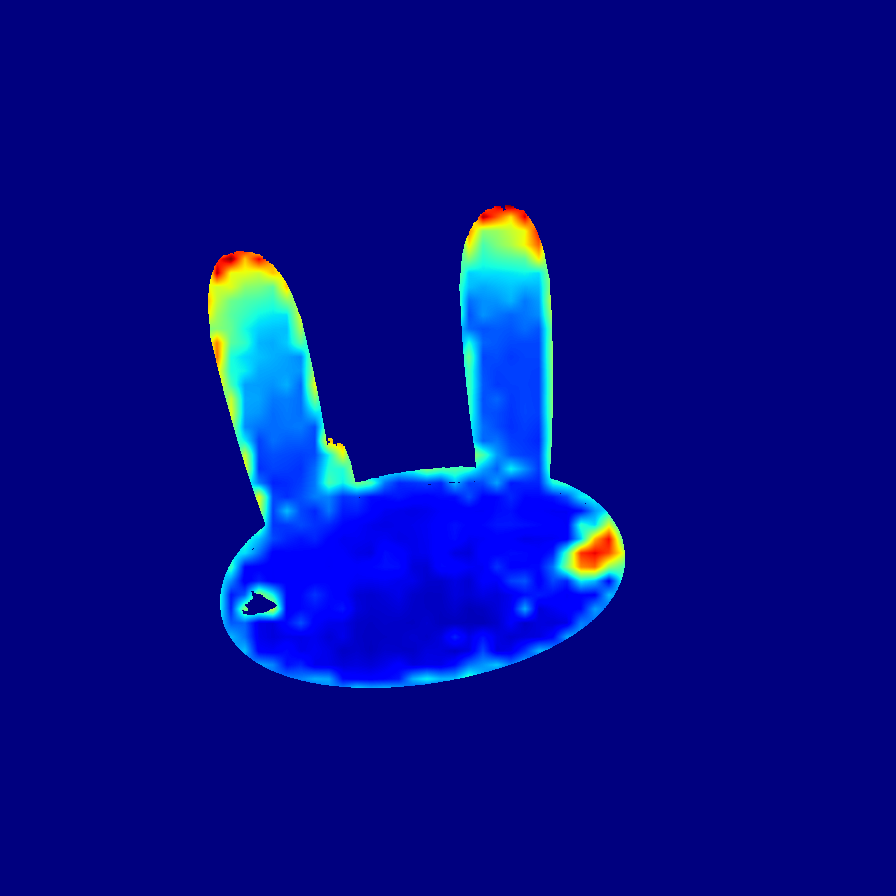} & 
            \includegraphics[width=0.08\linewidth,angle=180,origin=c]{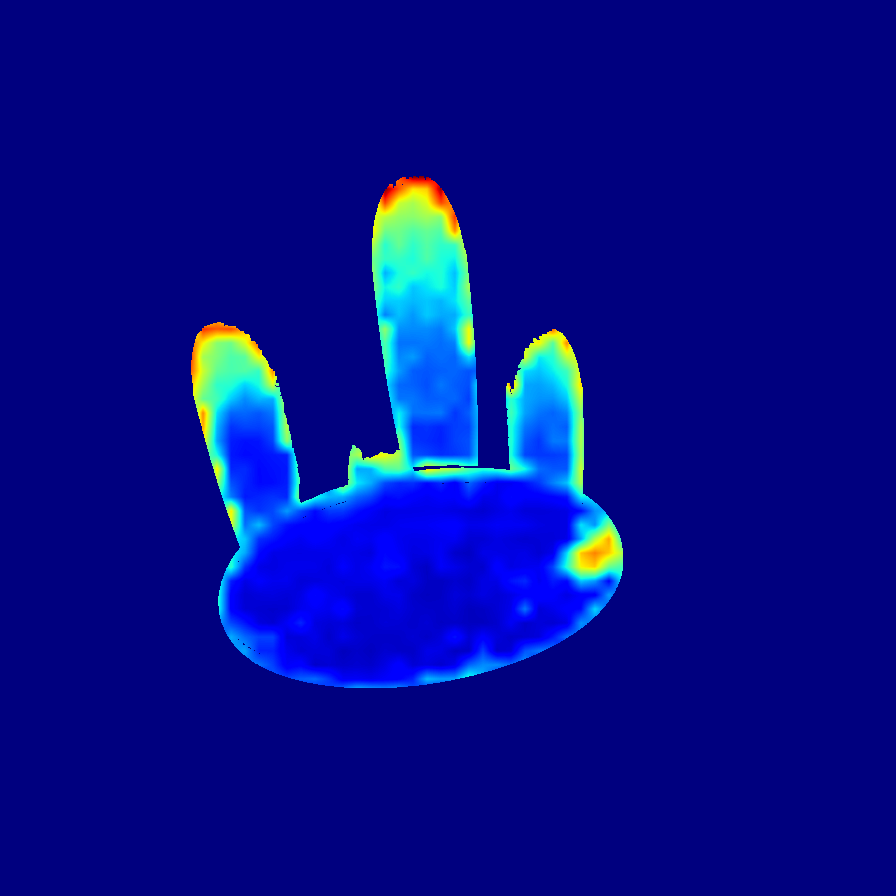} & 
            \includegraphics[width=0.08\linewidth,angle=180,origin=c]{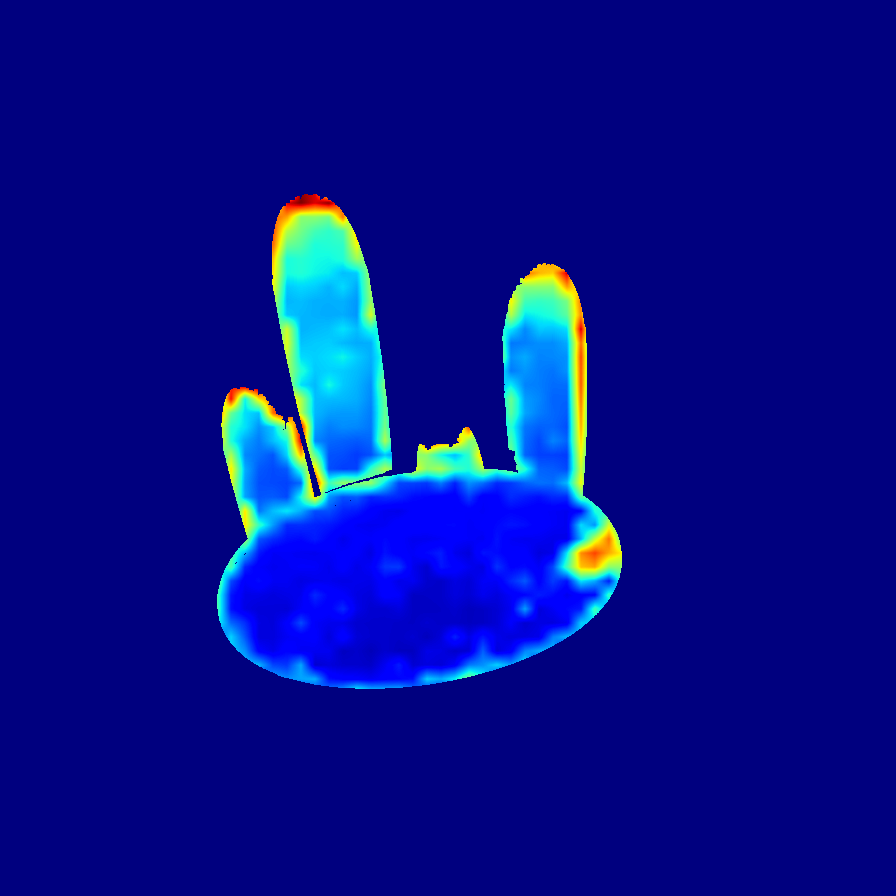} & 
            \includegraphics[width=0.08\linewidth,angle=180,origin=c]{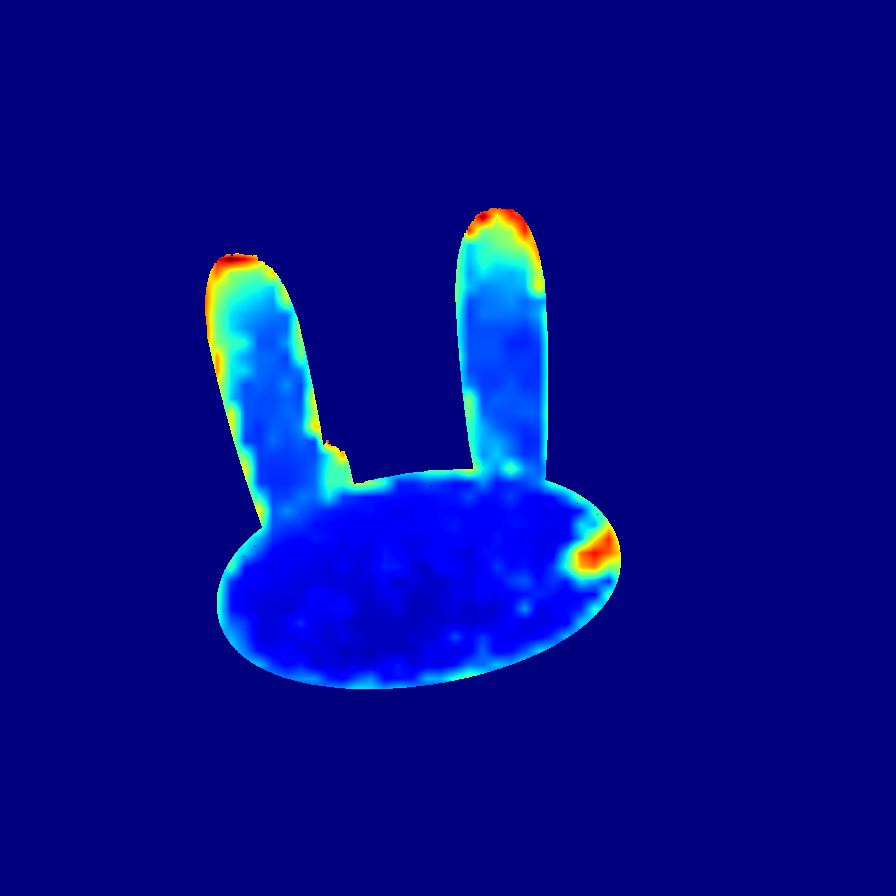} & 
            \includegraphics[width=0.08\linewidth,angle=180,origin=c]{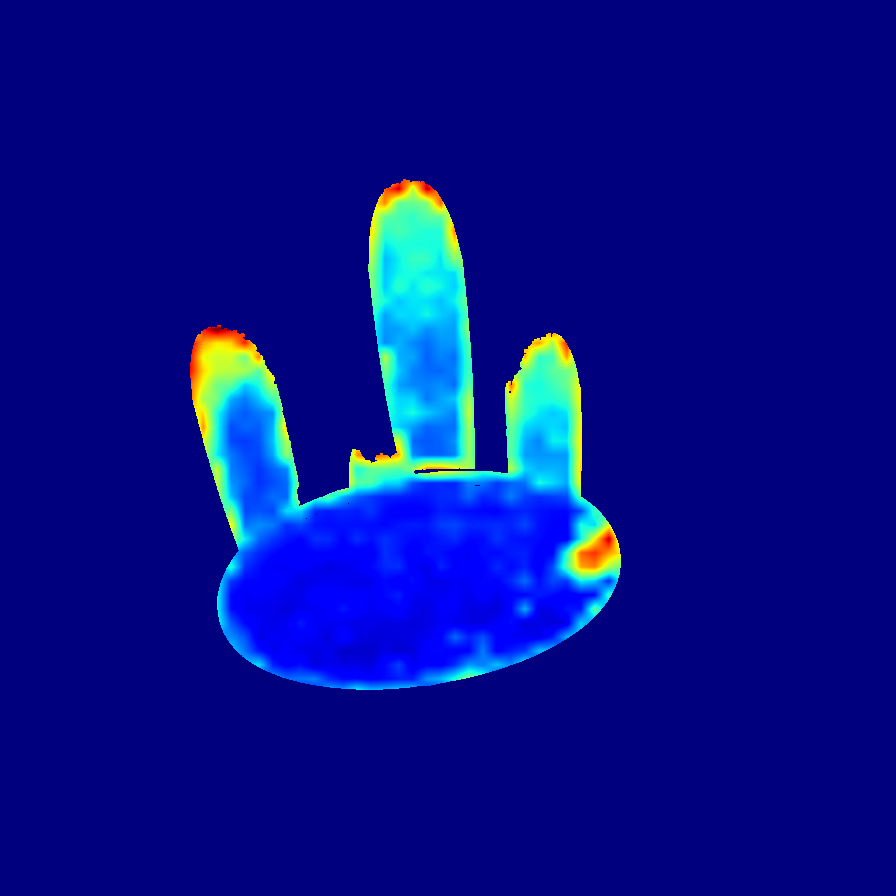} & 
            \includegraphics[width=0.08\linewidth,angle=180,origin=c]{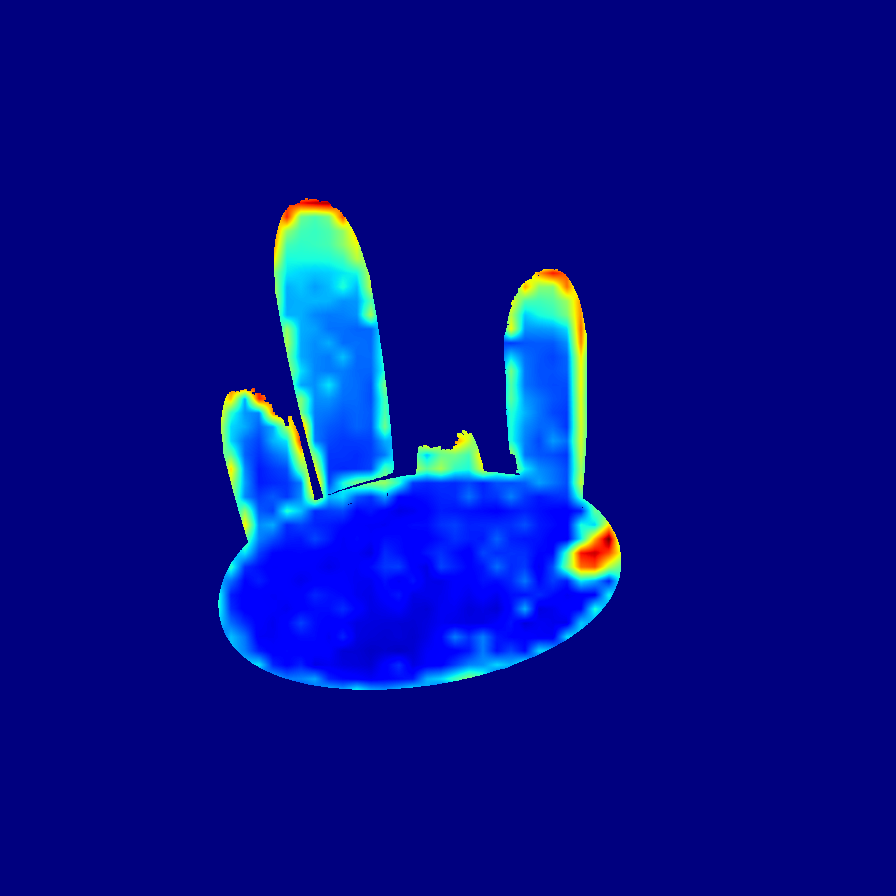} & 
            \includegraphics[width=0.08\linewidth,angle=180,origin=c]{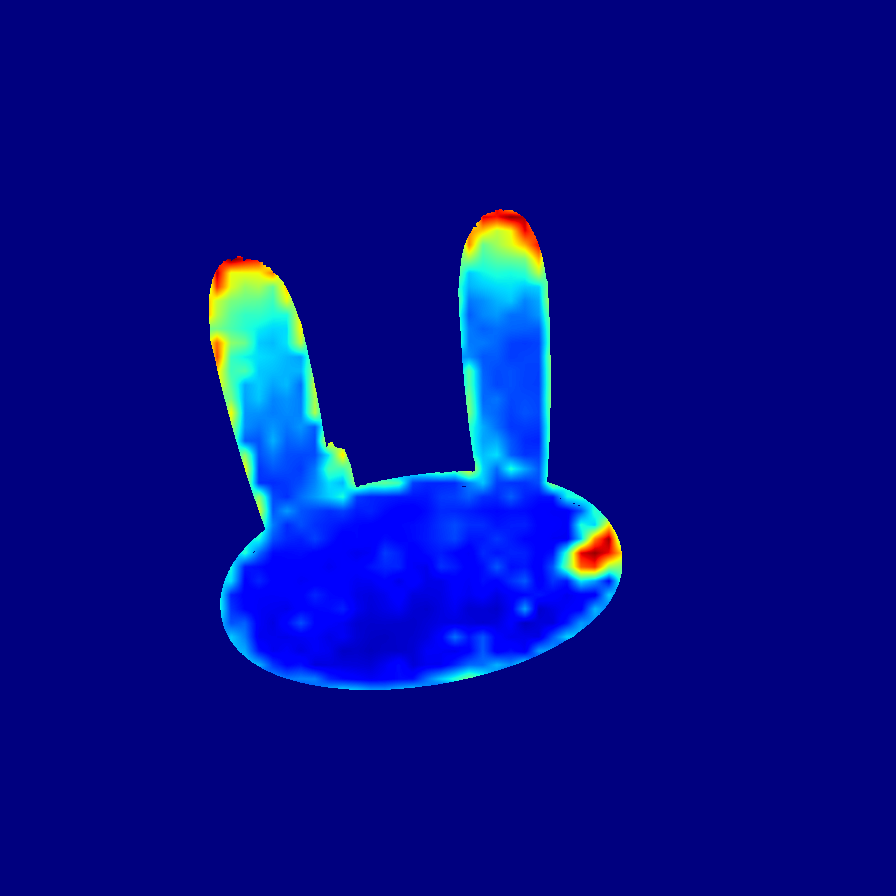} &
            \includegraphics[width=0.08\linewidth,angle=180,origin=c]{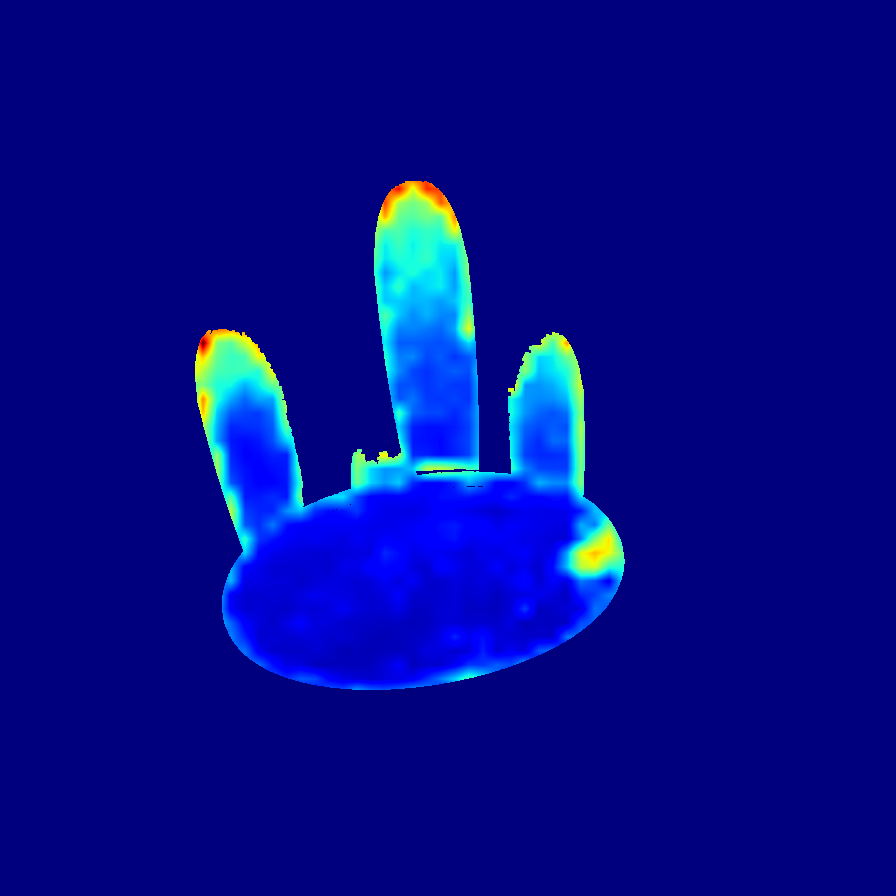} \\

            & \rotatebox{90}{\quad $v_{2}$} & 
            \includegraphics[width=0.08\linewidth,angle=180,origin=c]{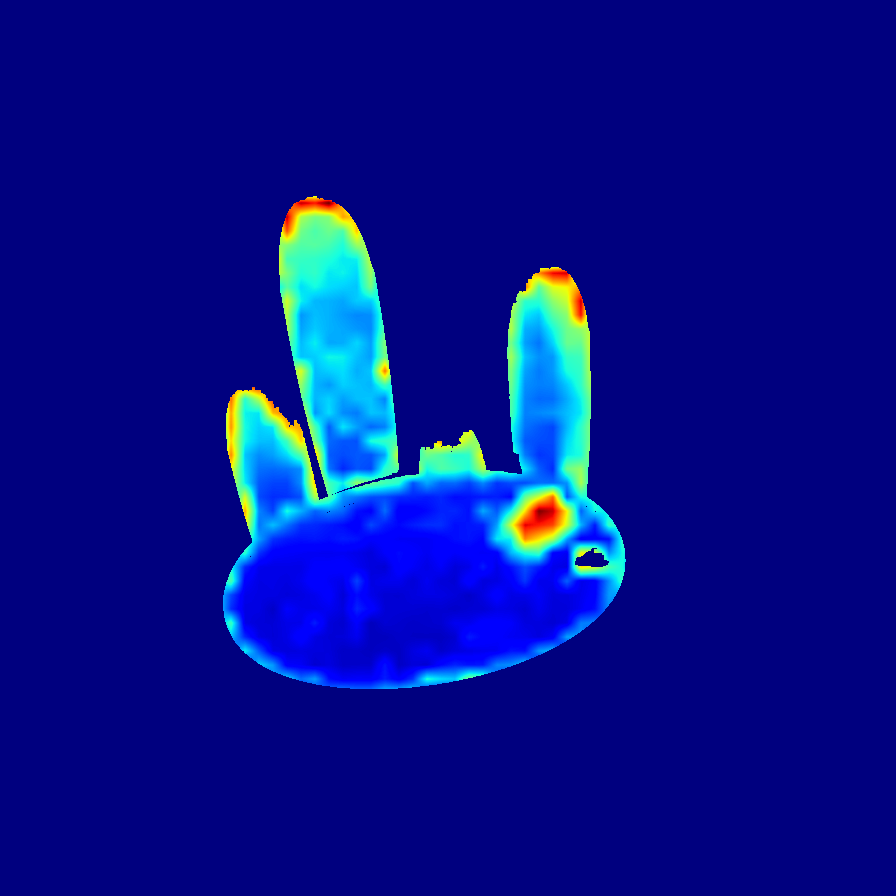} & 
            \includegraphics[width=0.08\linewidth,angle=180,origin=c]{images/cross_views_image2depth/02_am_C2_C2.png} & 
            \includegraphics[width=0.08\linewidth,angle=180,origin=c]{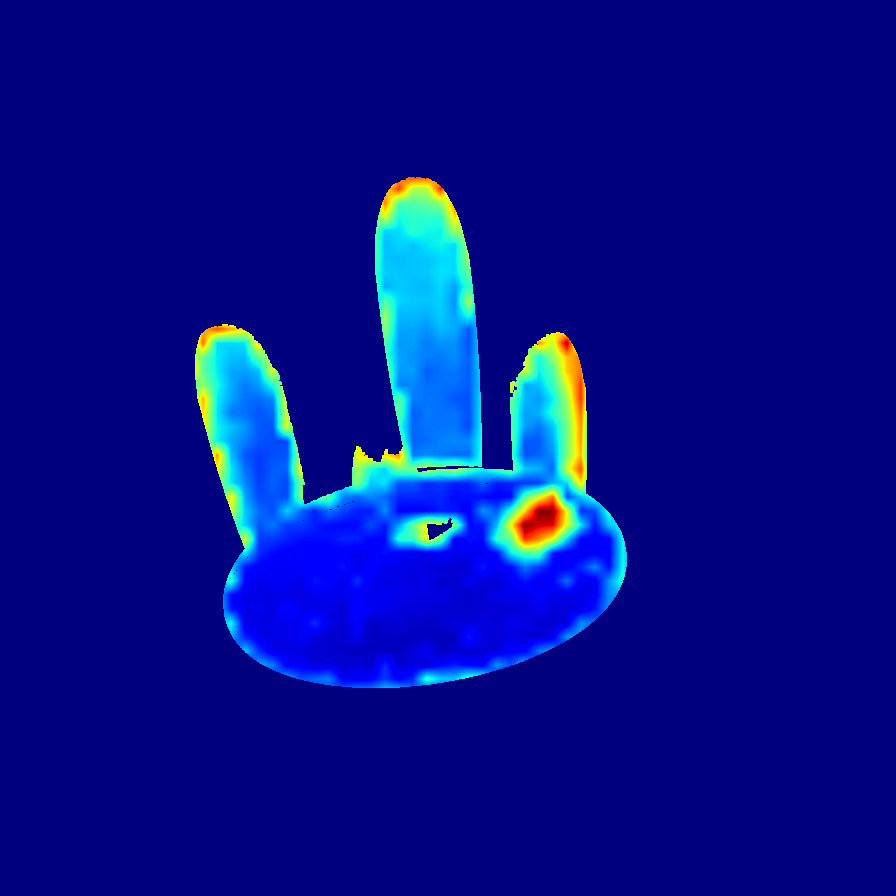} & 
            \includegraphics[width=0.08\linewidth,angle=180,origin=c]{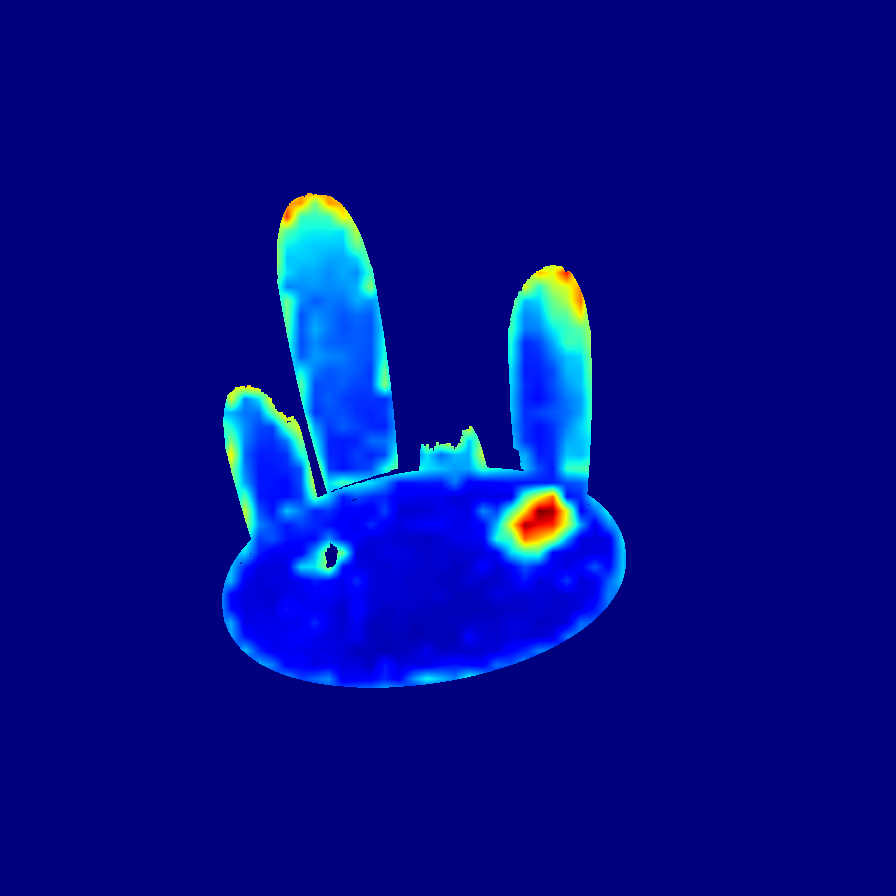} & 
            \includegraphics[width=0.08\linewidth,angle=180,origin=c]{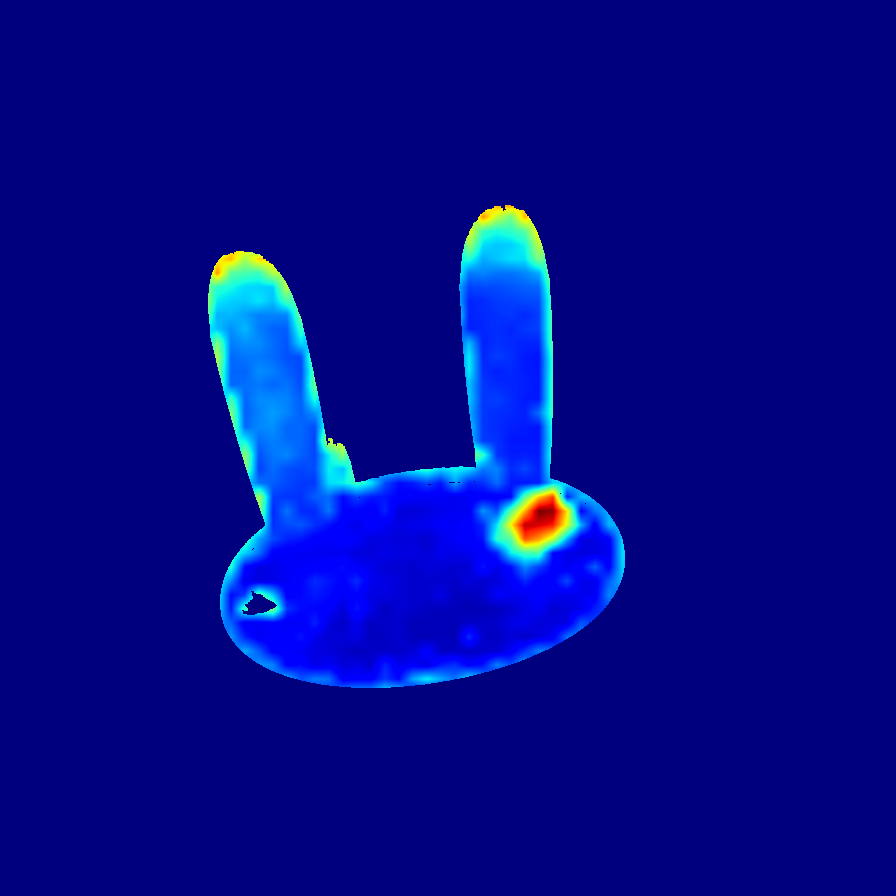} & 
            \includegraphics[width=0.08\linewidth,angle=180,origin=c]{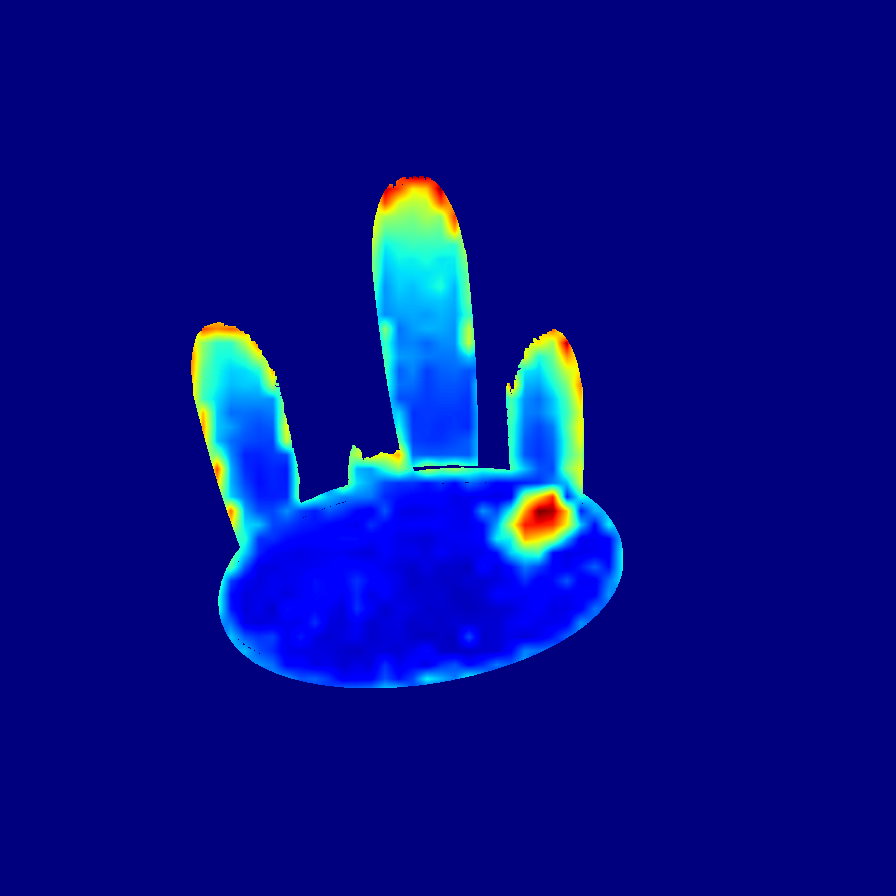} & 
            \includegraphics[width=0.08\linewidth,angle=180,origin=c]{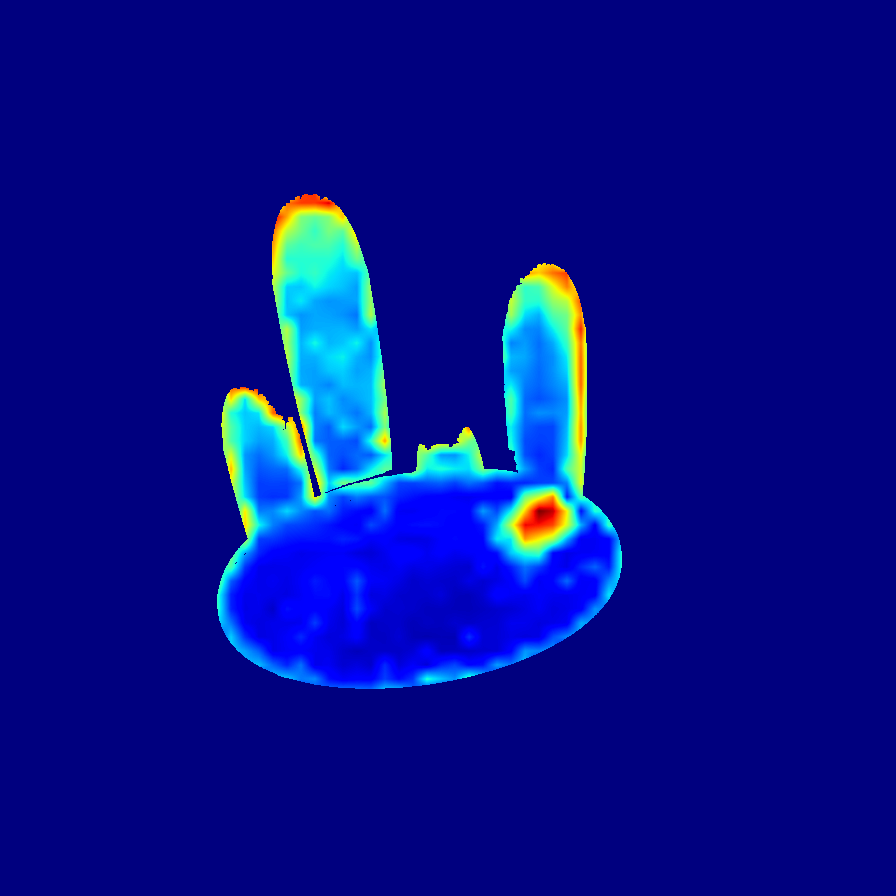} & 
            \includegraphics[width=0.08\linewidth,angle=180,origin=c]{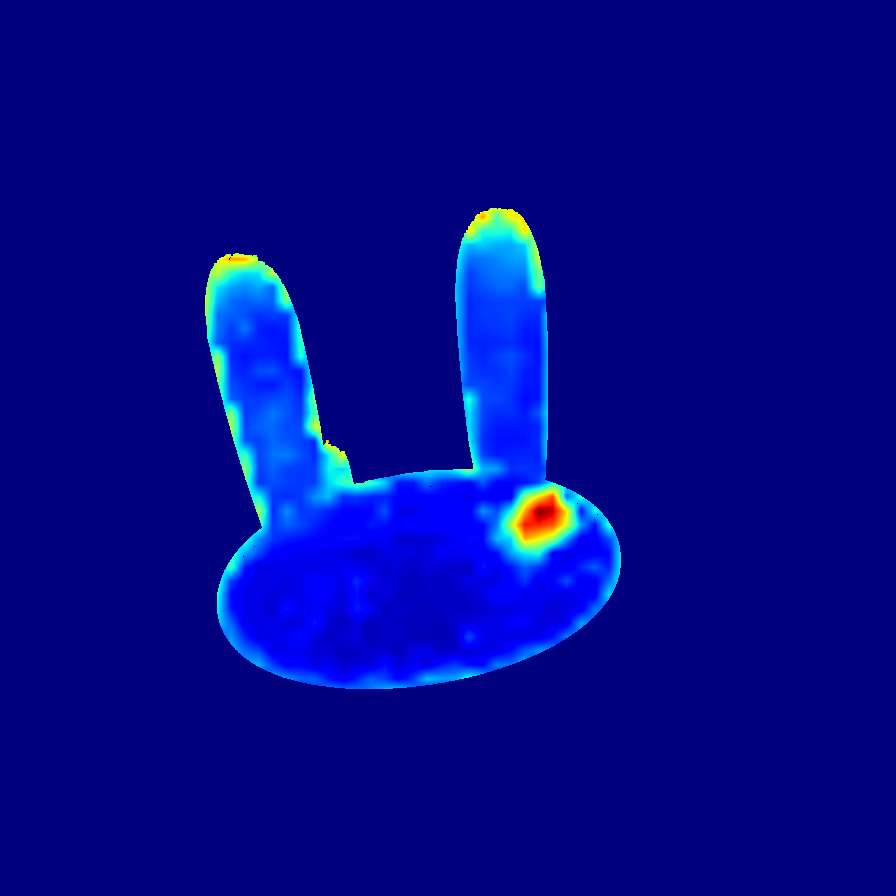} & 
            \includegraphics[width=0.08\linewidth,angle=180,origin=c]{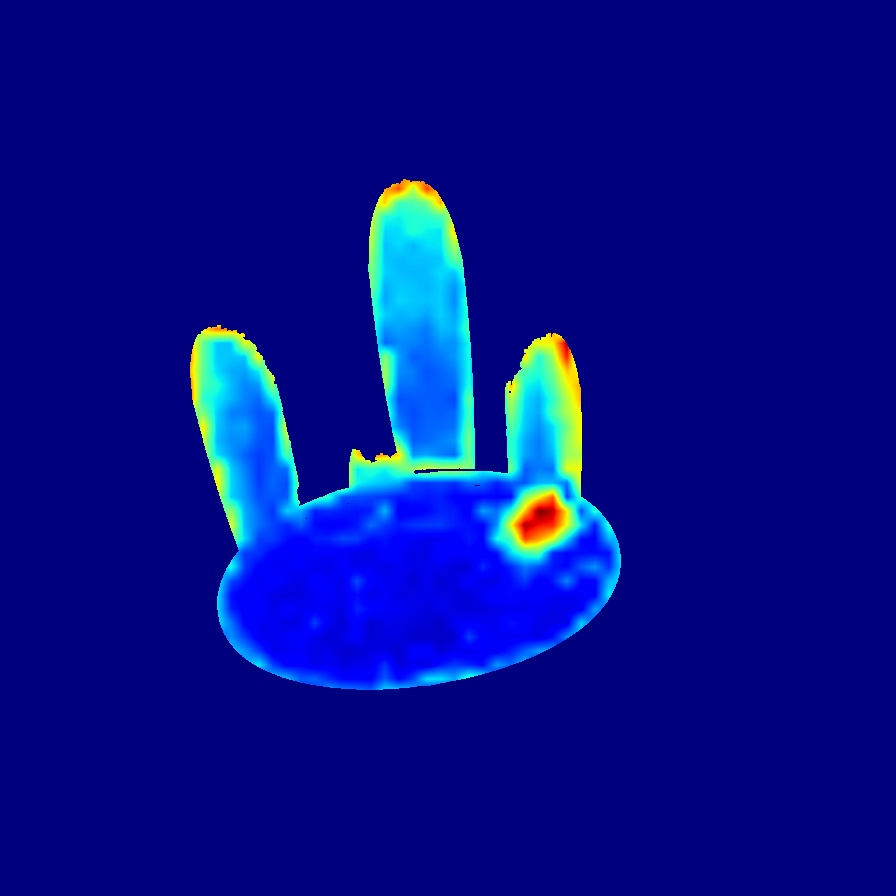} & 
            \includegraphics[width=0.08\linewidth,angle=180,origin=c]{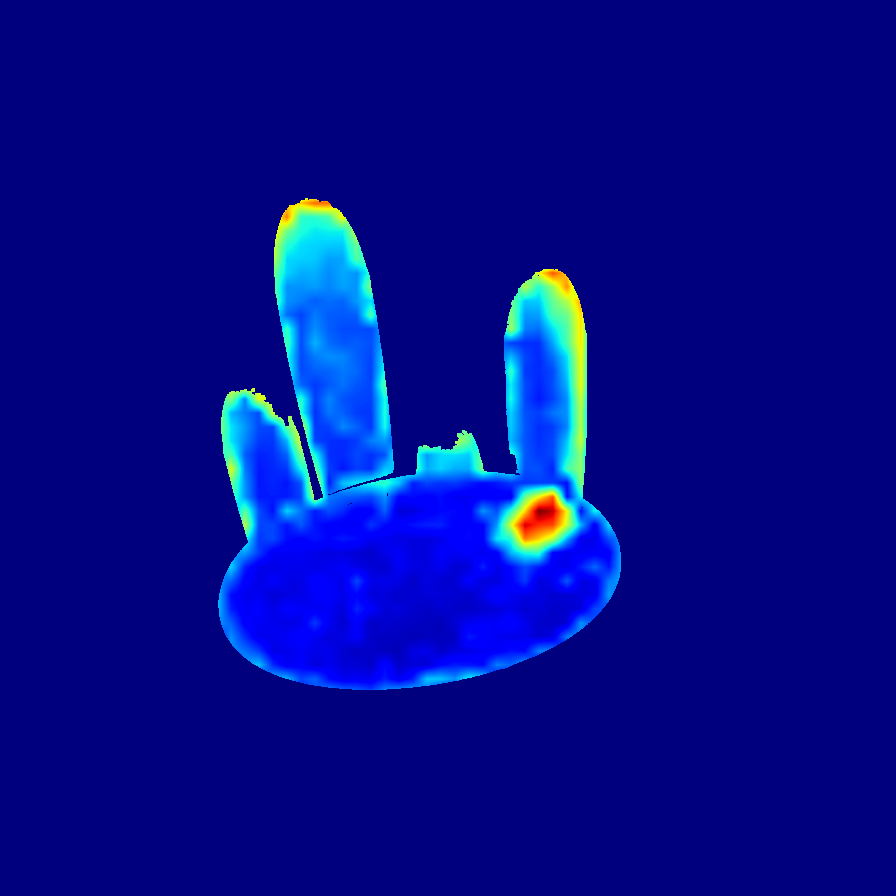} & 
            \includegraphics[width=0.08\linewidth,angle=180,origin=c]{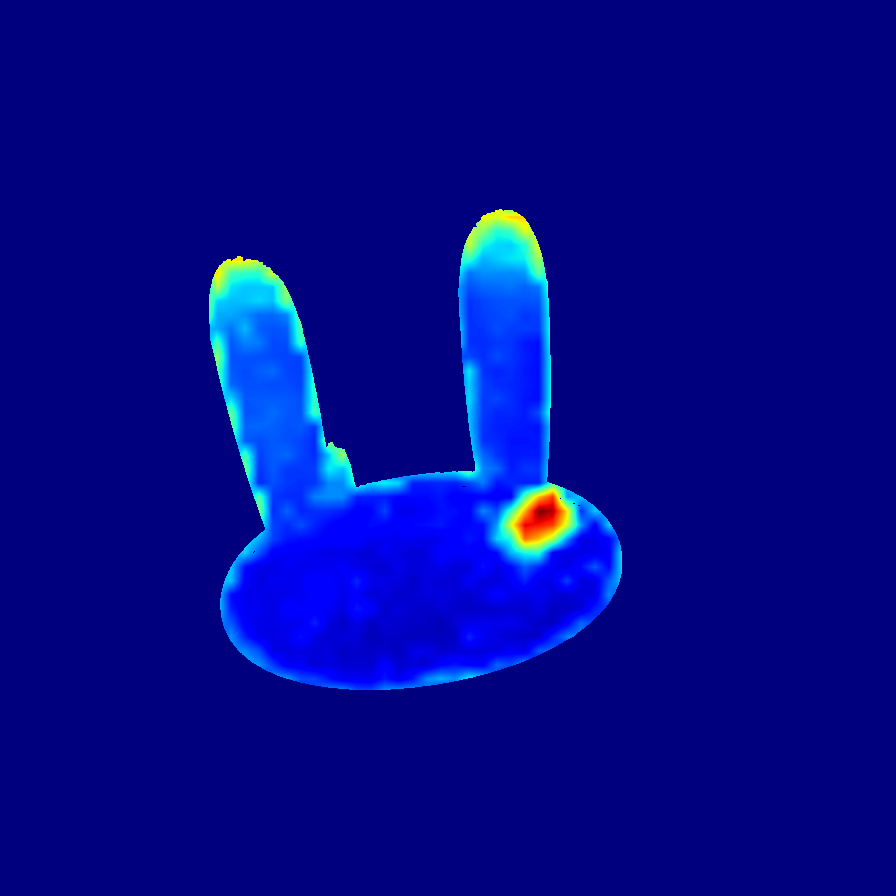} &
            \includegraphics[width=0.08\linewidth,angle=180,origin=c]{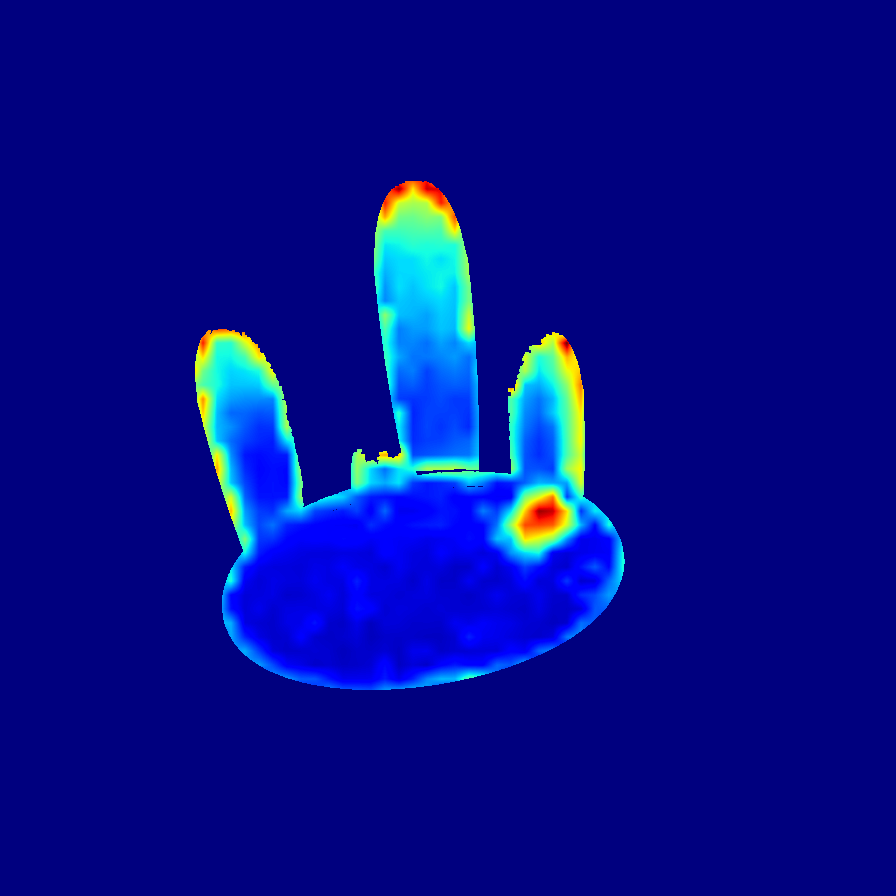} \\
            
            & \rotatebox{90}{\quad $v_{3}$} &
            \includegraphics[width=0.08\linewidth,angle=180,origin=c]{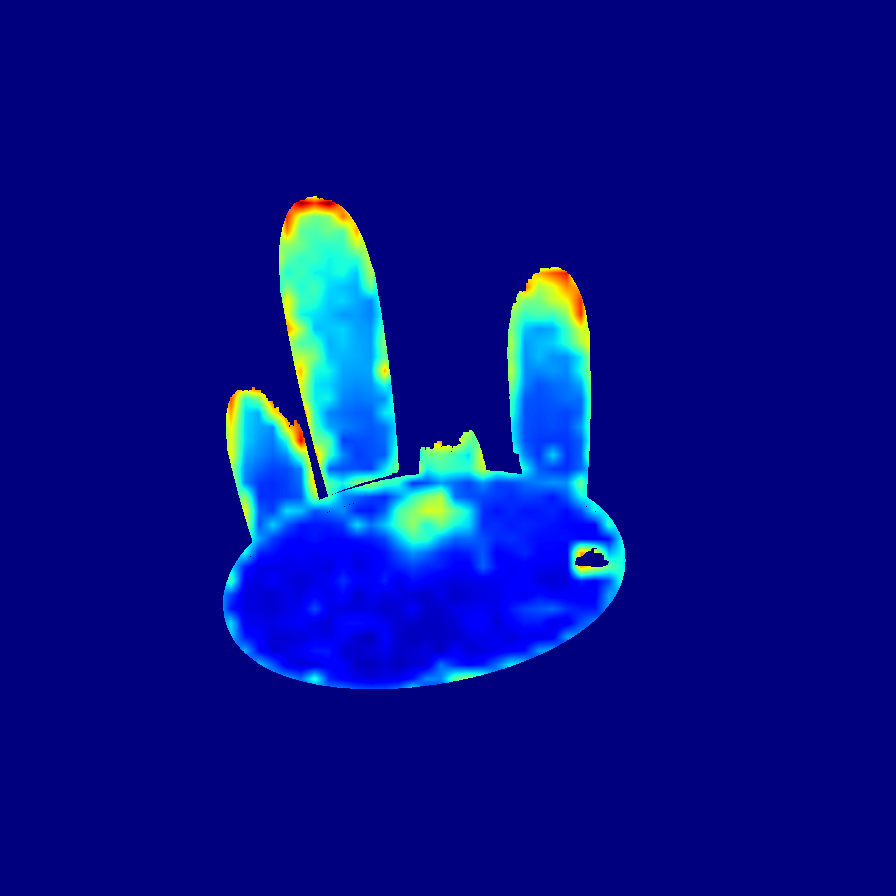} & 
            \includegraphics[width=0.08\linewidth,angle=180,origin=c]{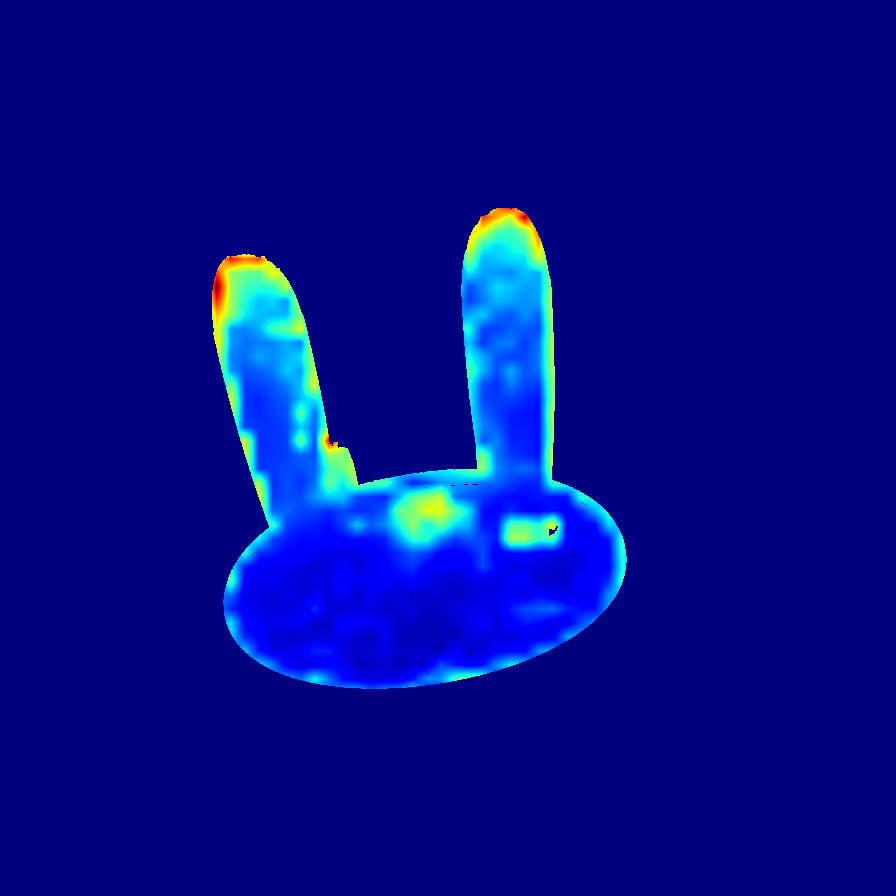} & 
            \includegraphics[width=0.08\linewidth,angle=180,origin=c]{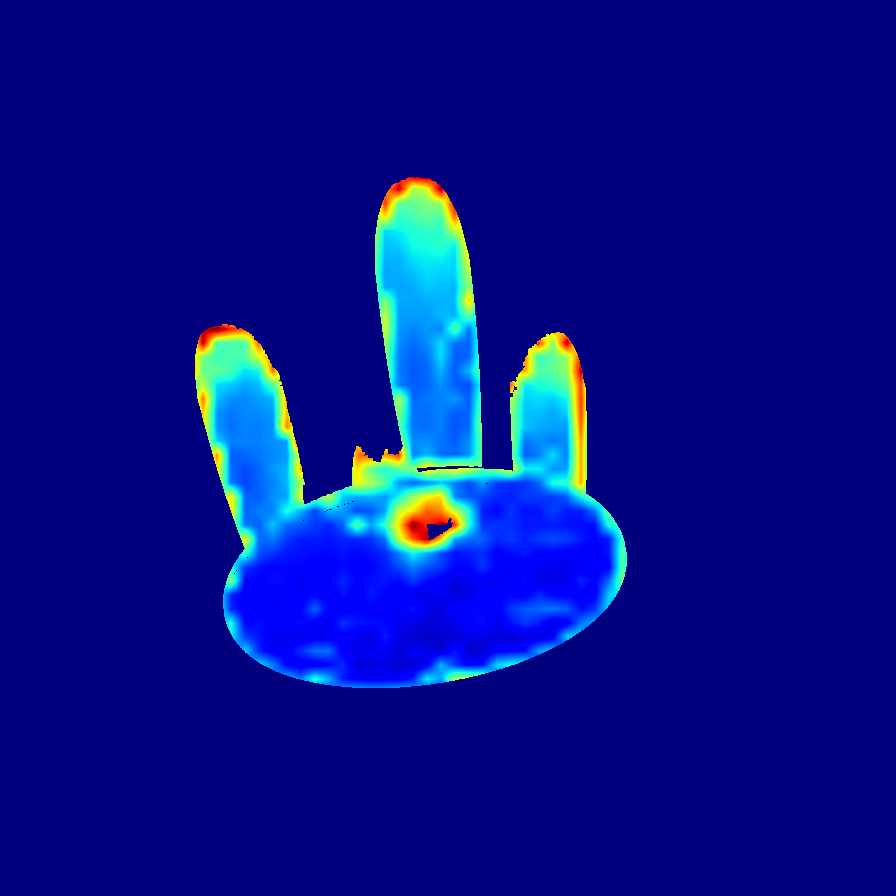} & 
            \includegraphics[width=0.08\linewidth,angle=180,origin=c]{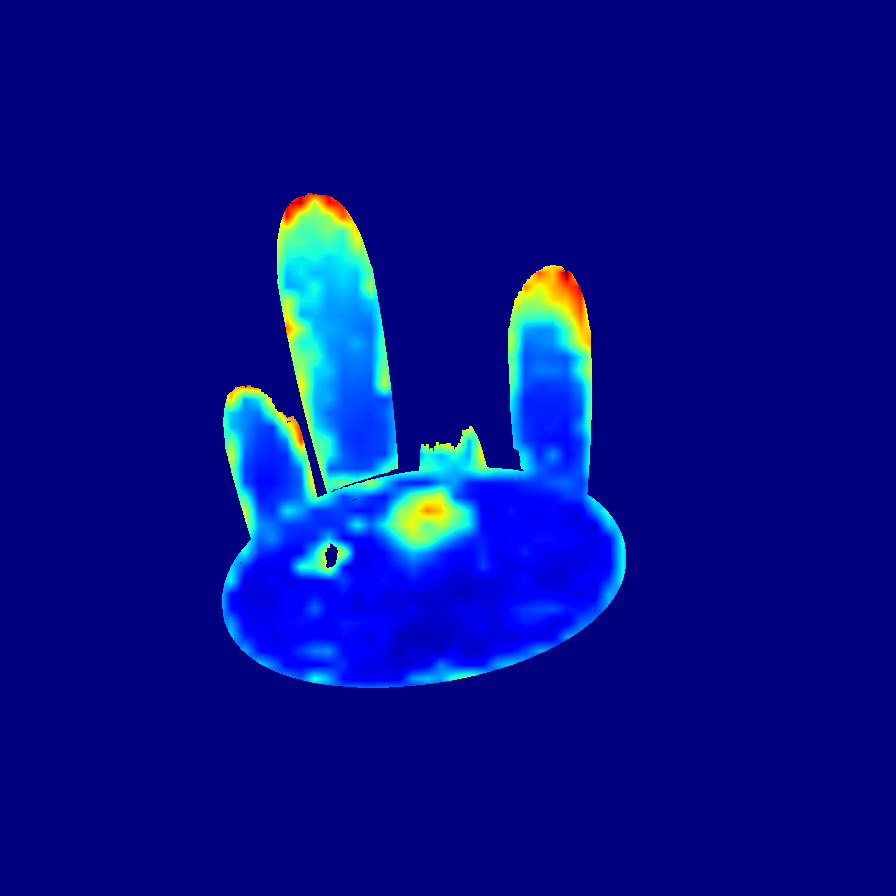} & 
            \includegraphics[width=0.08\linewidth,angle=180,origin=c]{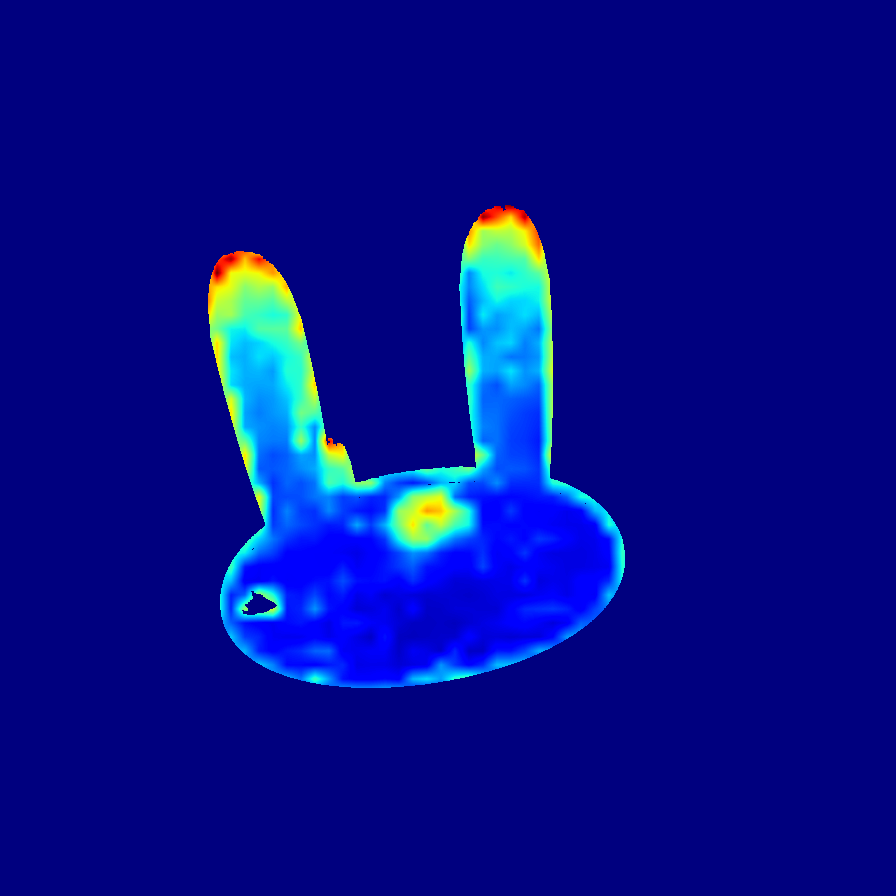} & 
            \includegraphics[width=0.08\linewidth,angle=180,origin=c]{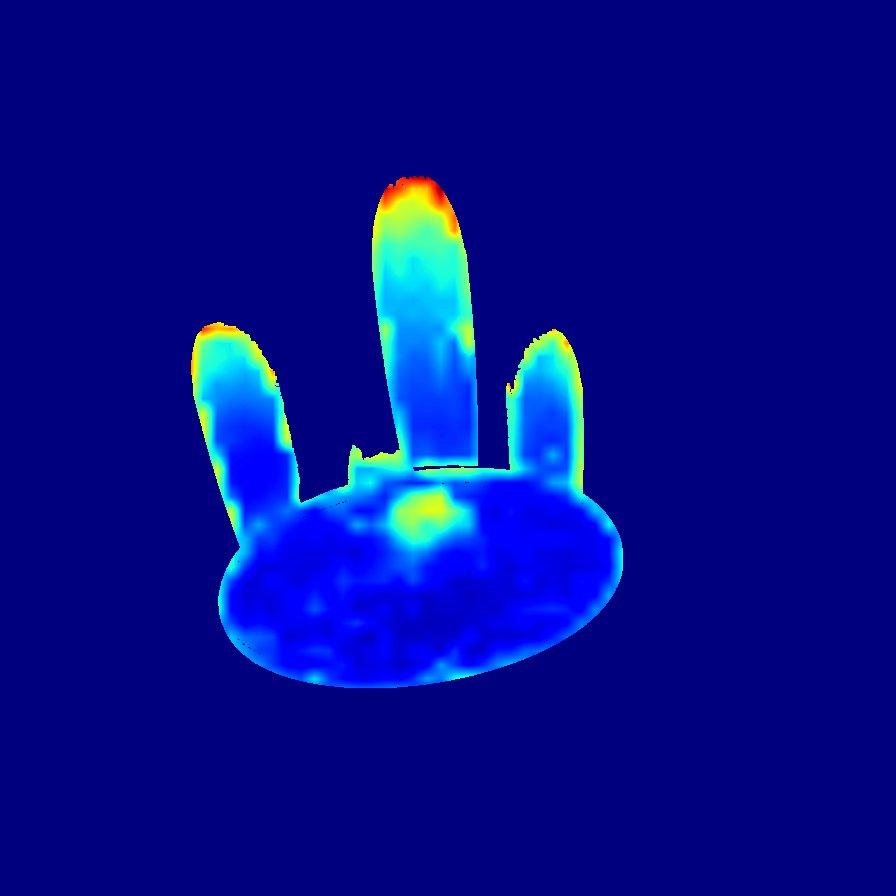} & 
            \includegraphics[width=0.08\linewidth,angle=180,origin=c]{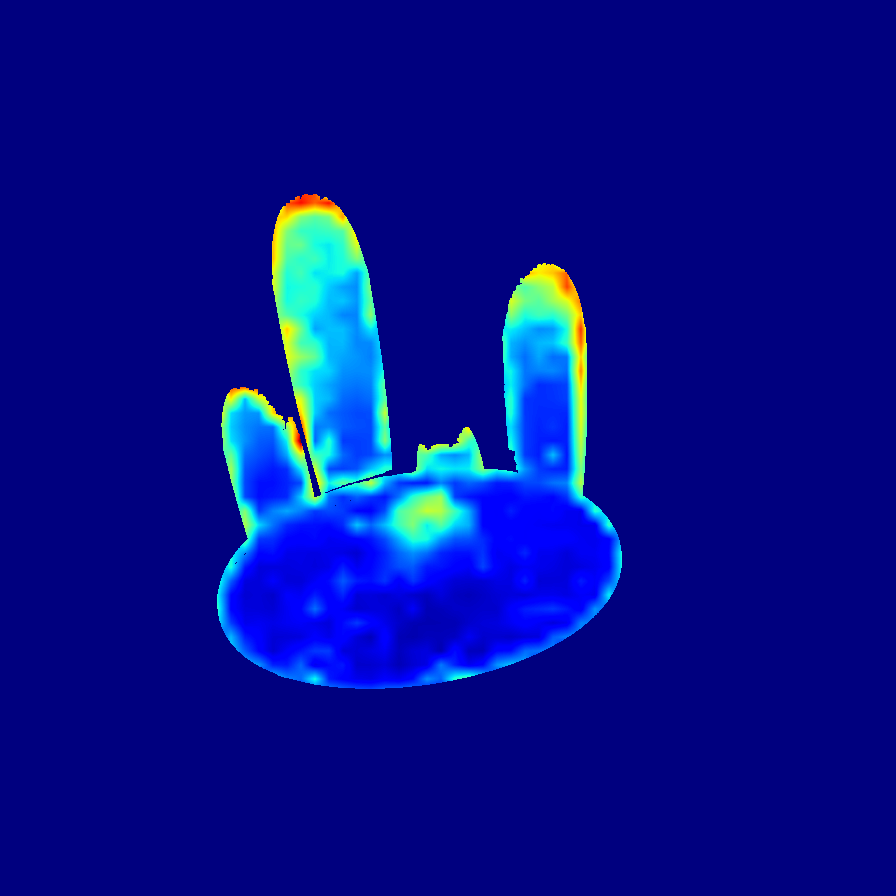} & 
            \includegraphics[width=0.08\linewidth,angle=180,origin=c]{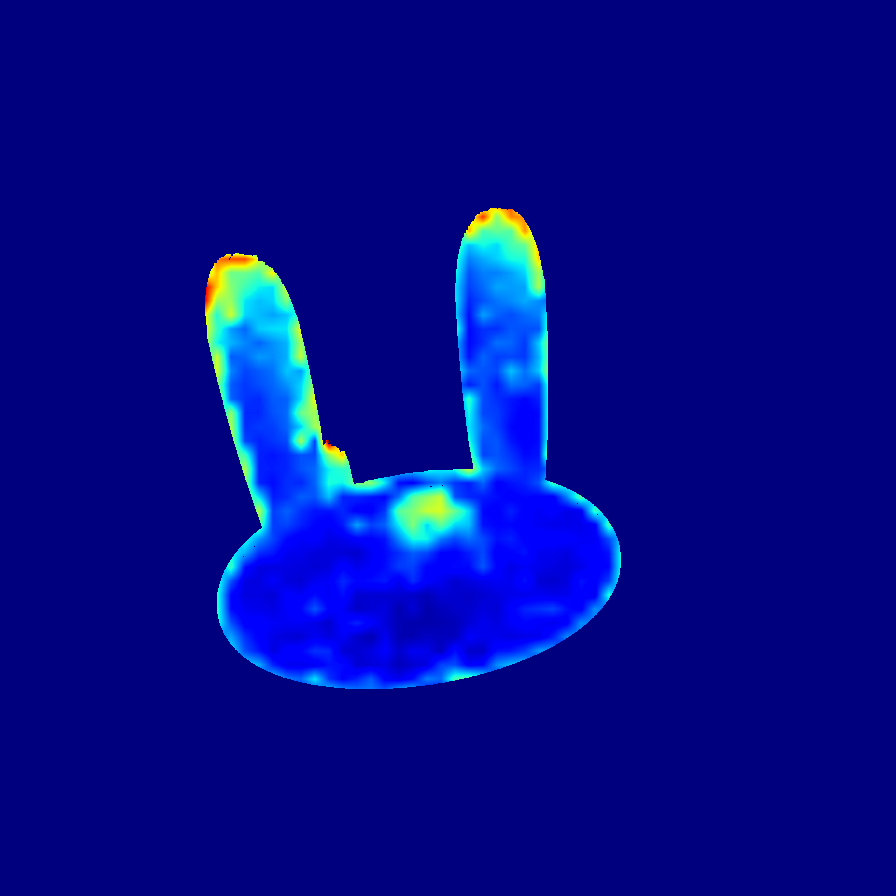} & 
            \includegraphics[width=0.08\linewidth,angle=180,origin=c]{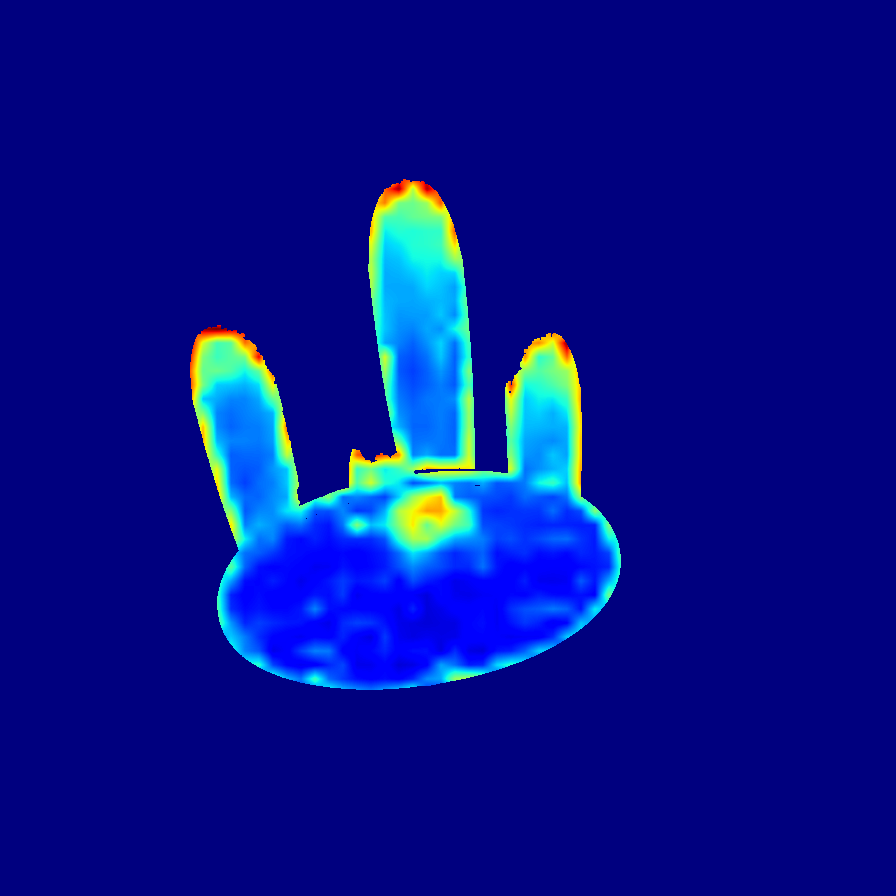} & 
            \includegraphics[width=0.08\linewidth,angle=180,origin=c]{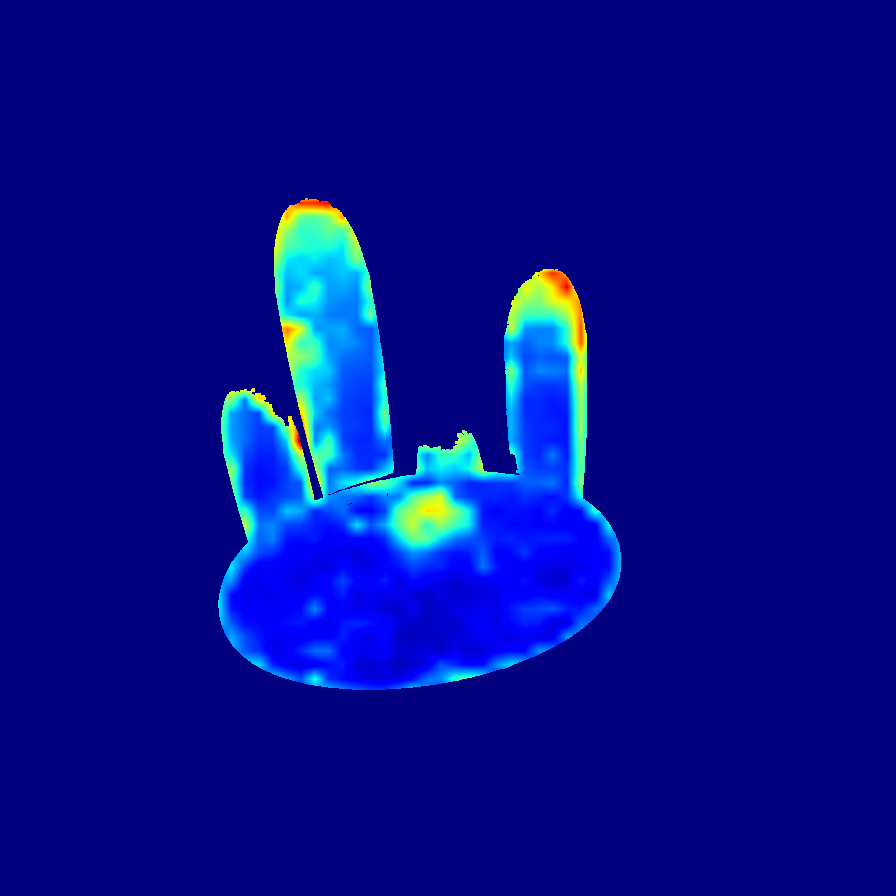} & 
            \includegraphics[width=0.08\linewidth,angle=180,origin=c]{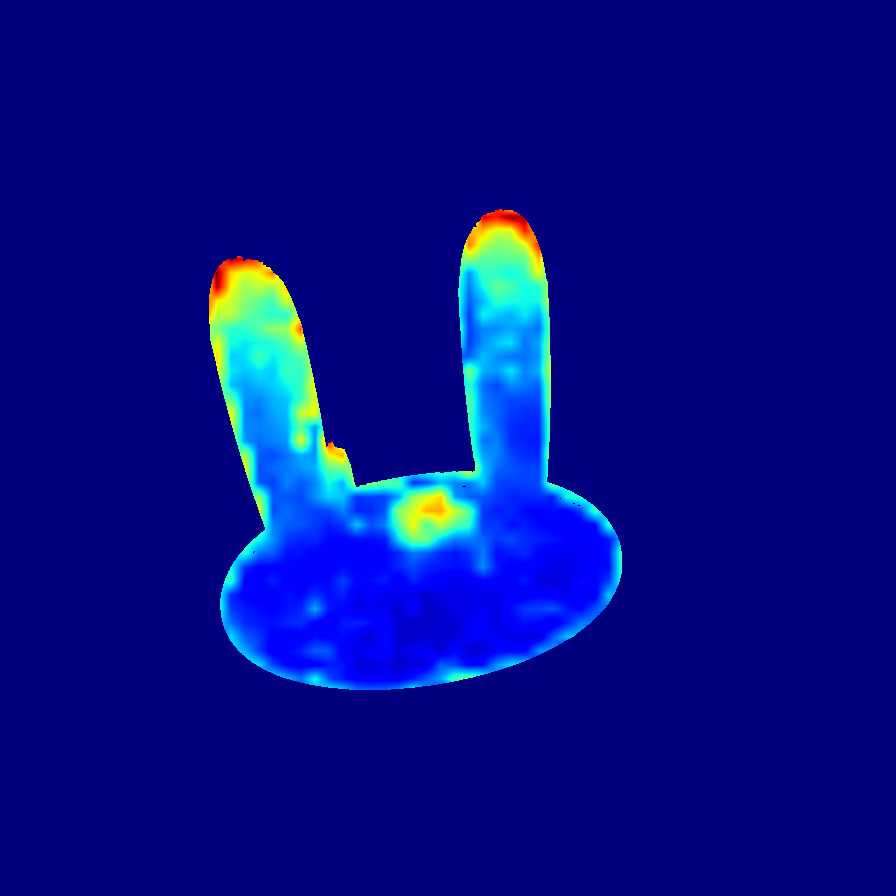} &
            \includegraphics[width=0.08\linewidth,angle=180,origin=c]{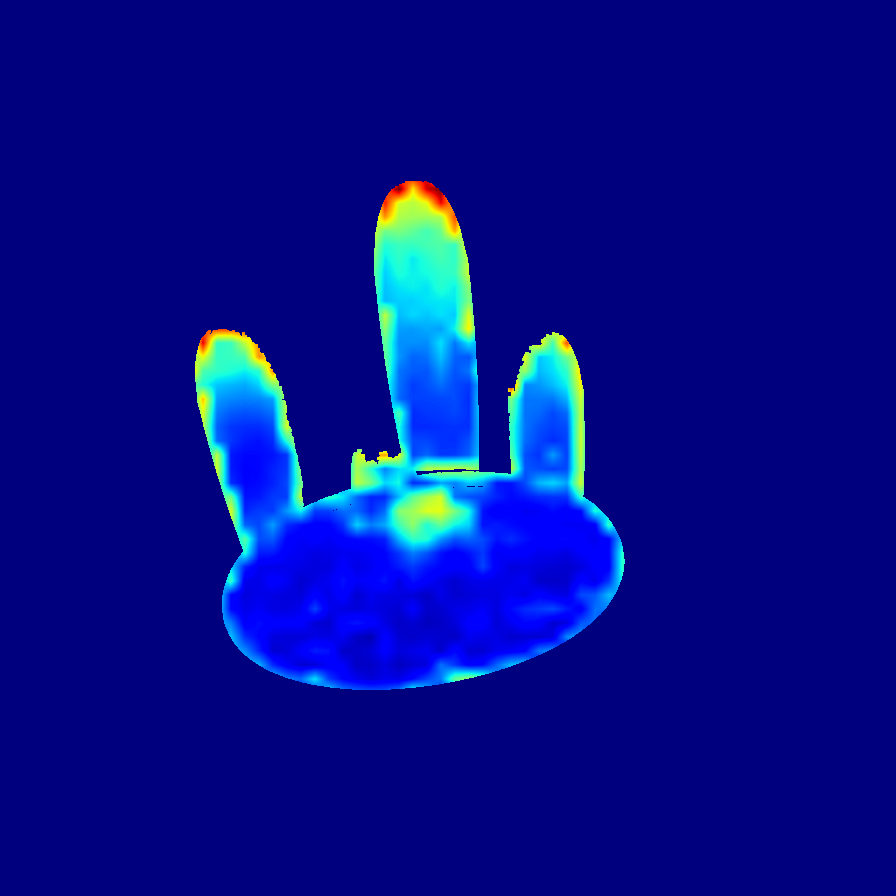} \\
            
            & \rotatebox{90}{\quad $v_{4}$} &
            \includegraphics[width=0.08\linewidth,angle=180,origin=c]{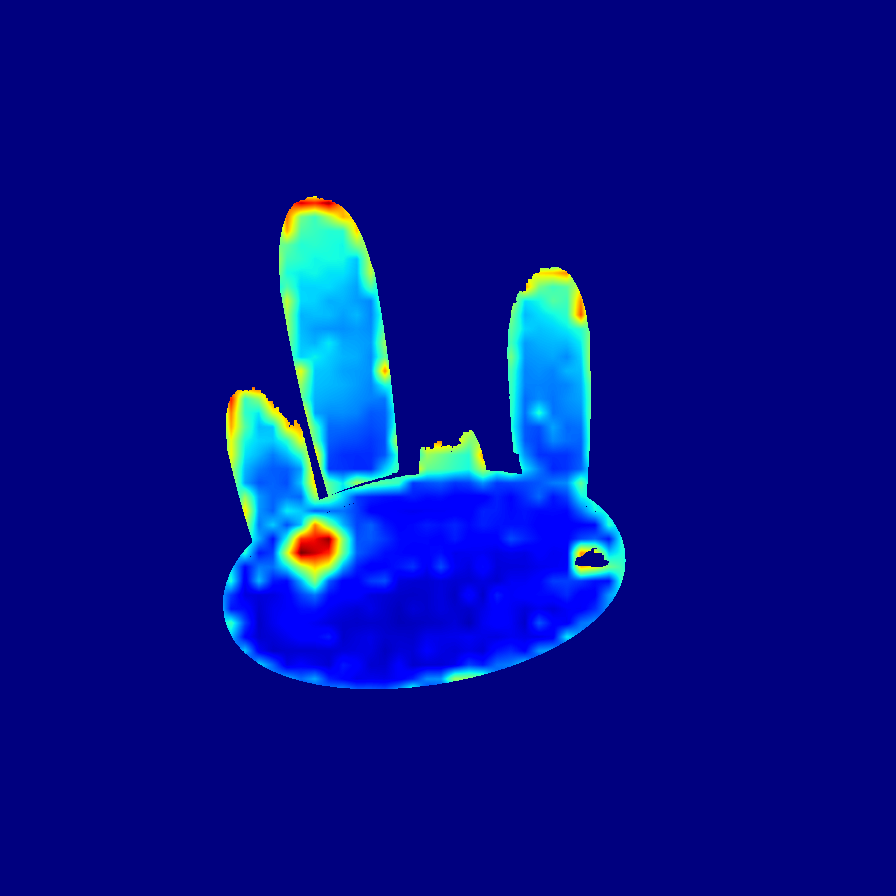} & 
            \includegraphics[width=0.08\linewidth,angle=180,origin=c]{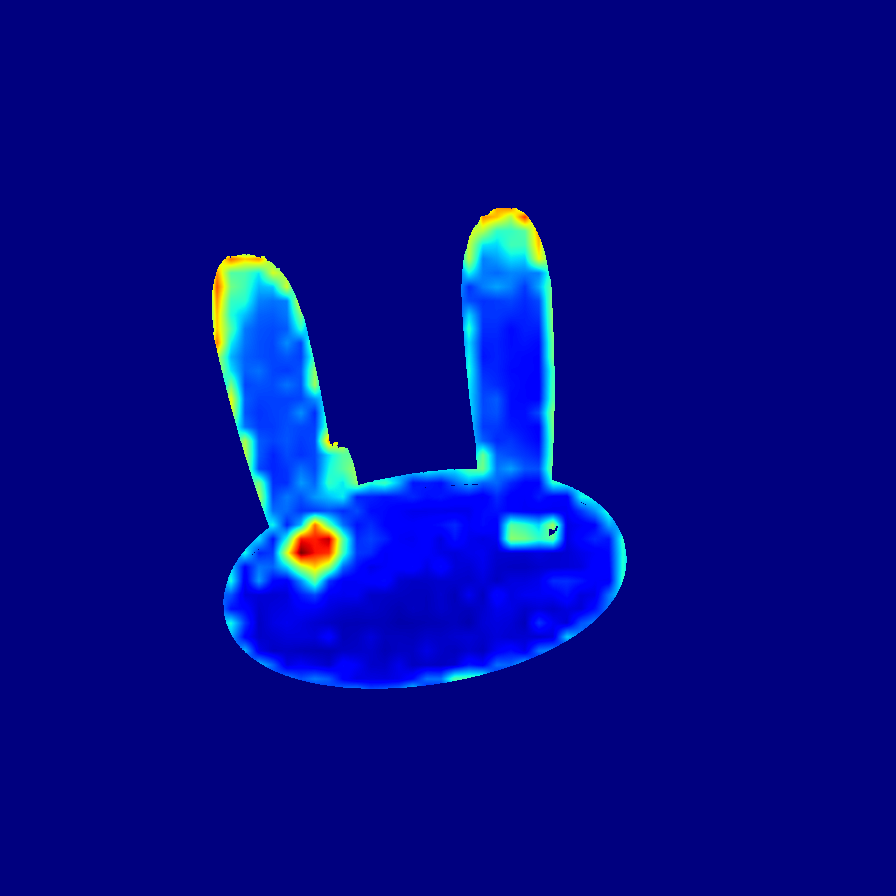} & 
            \includegraphics[width=0.08\linewidth,angle=180,origin=c]{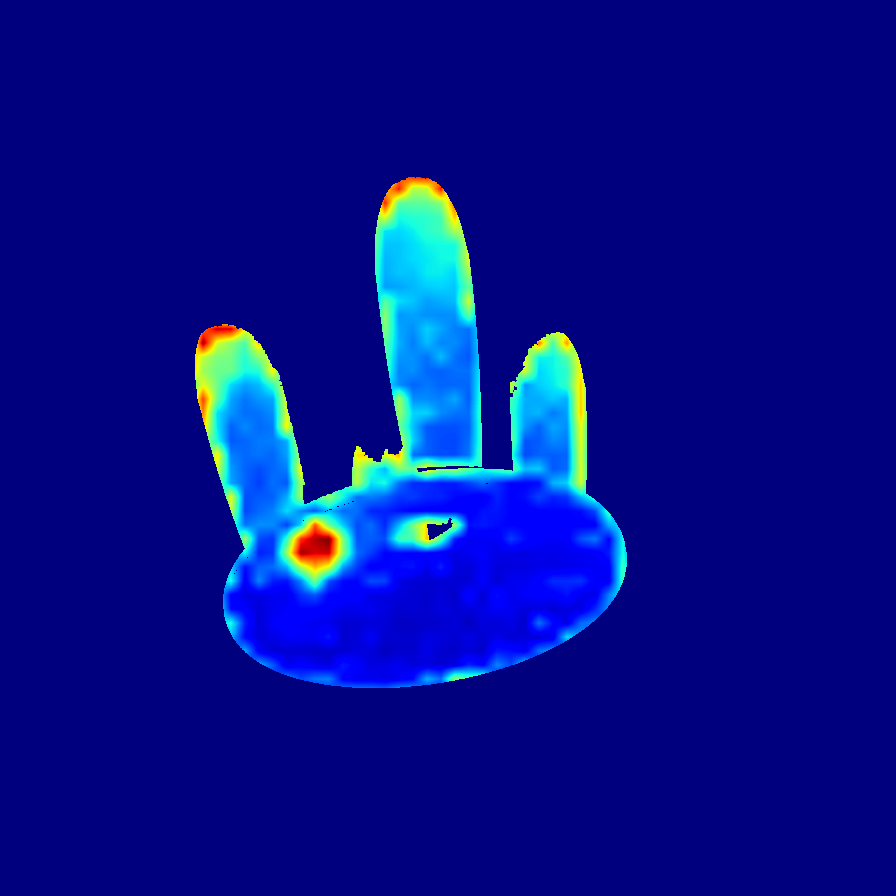} & 
            \includegraphics[width=0.08\linewidth,angle=180,origin=c]{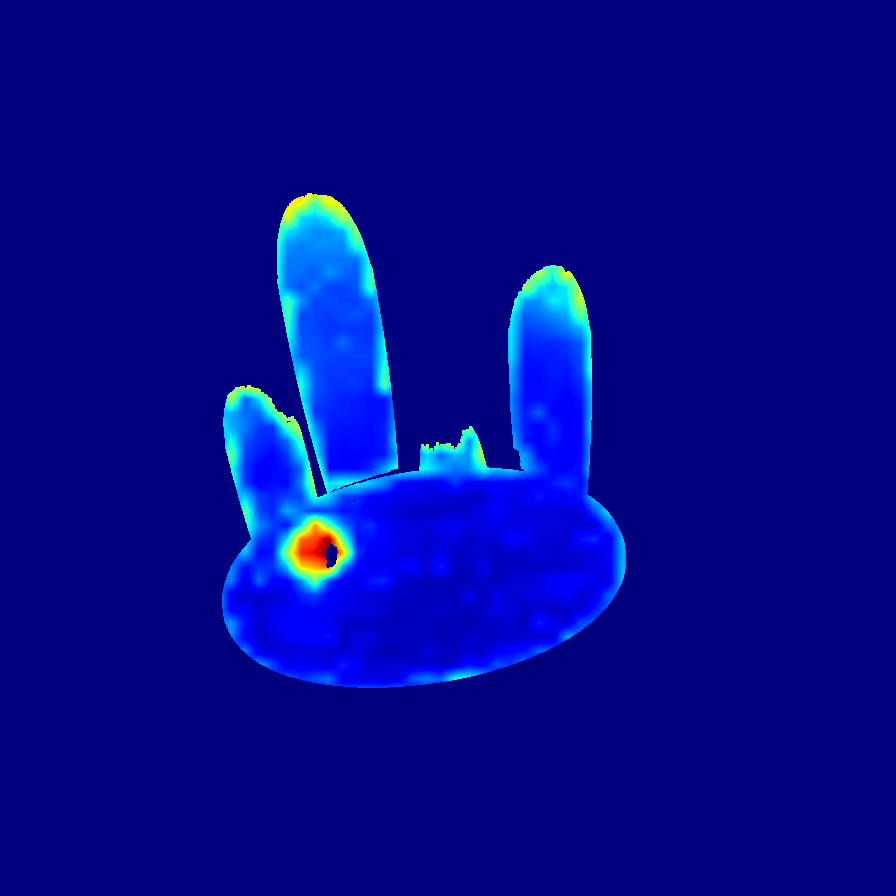} & 
            \includegraphics[width=0.08\linewidth,angle=180,origin=c]{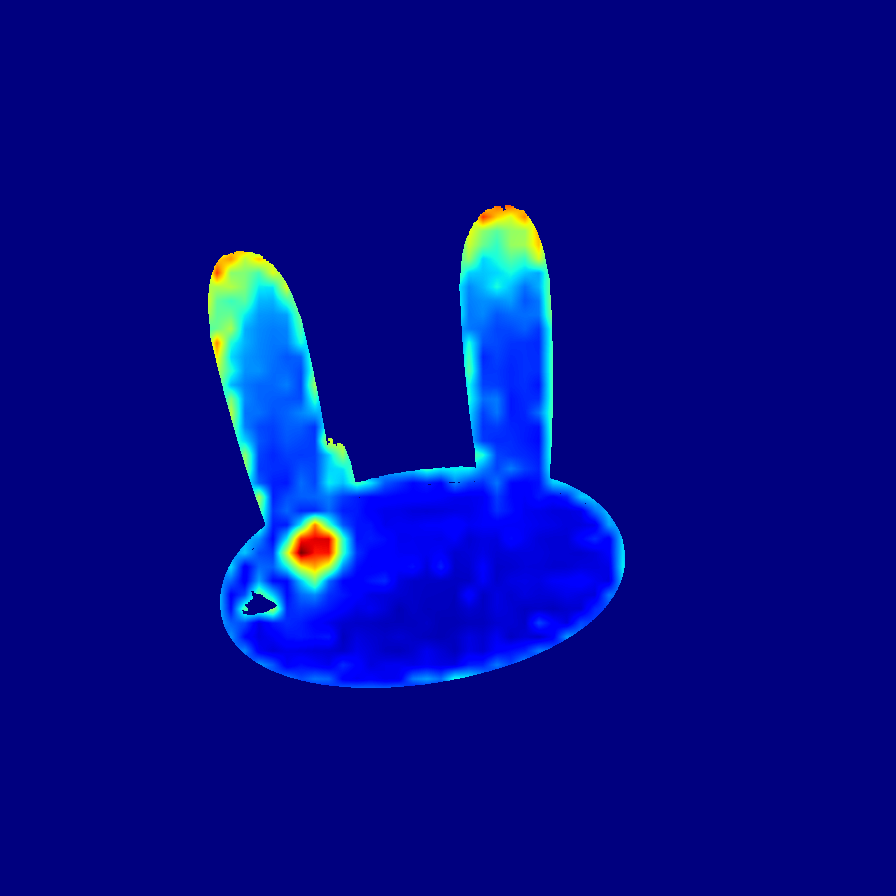} & 
            \includegraphics[width=0.08\linewidth,angle=180,origin=c]{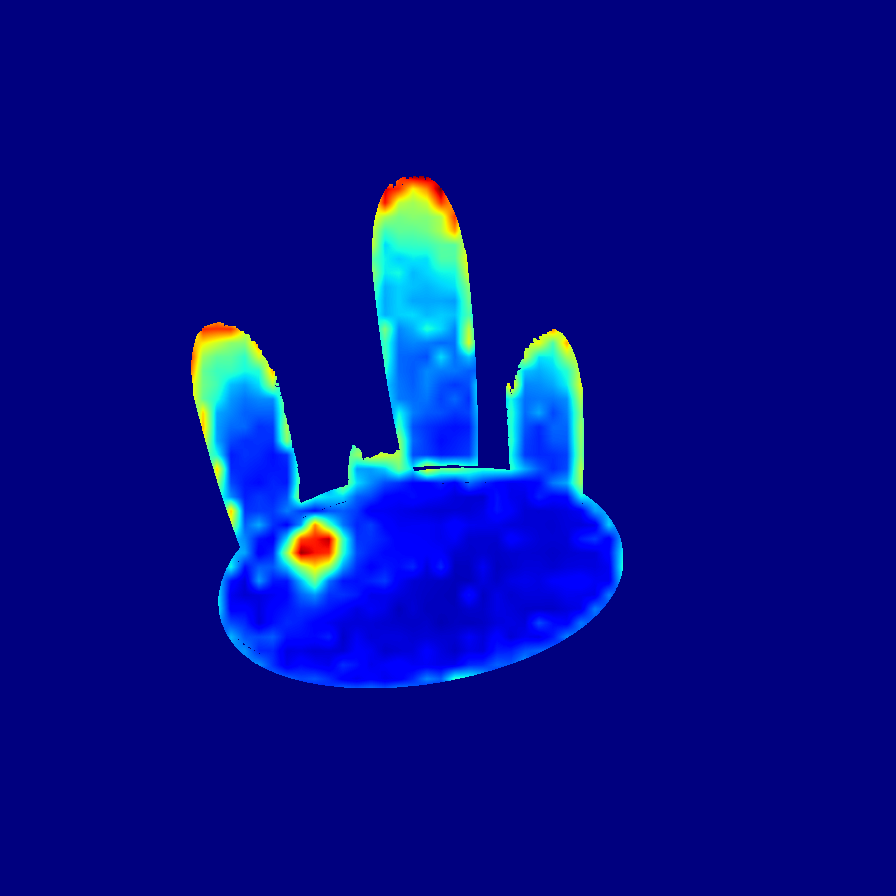} & 
            \includegraphics[width=0.08\linewidth,angle=180,origin=c]{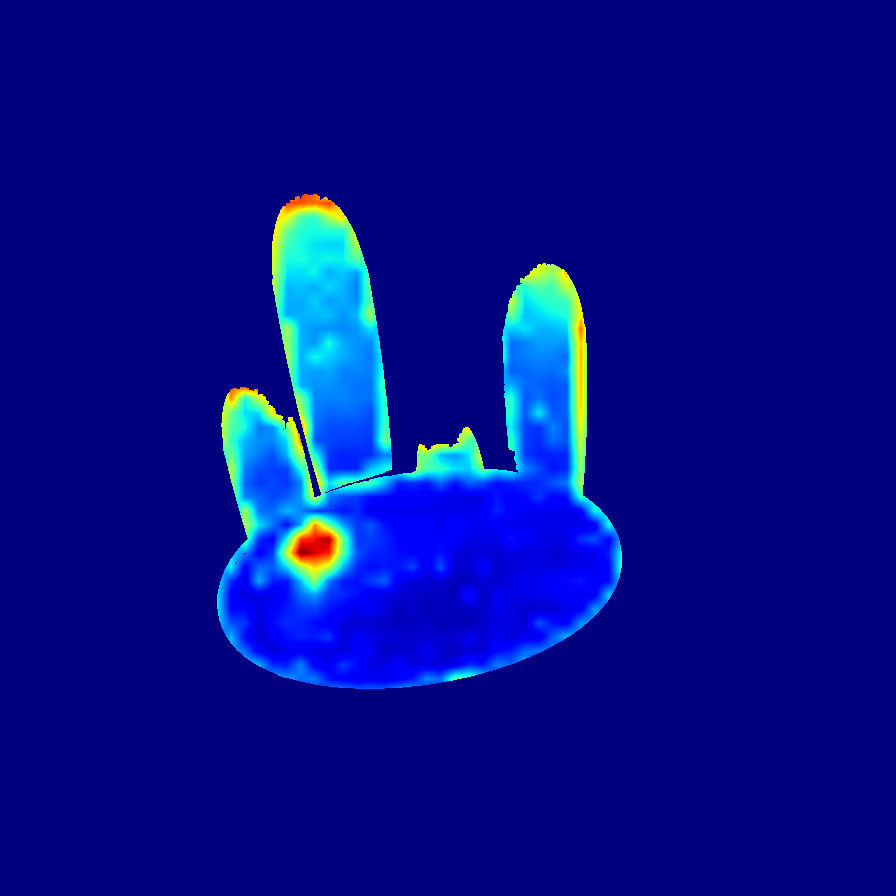} & 
            \includegraphics[width=0.08\linewidth,angle=180,origin=c]{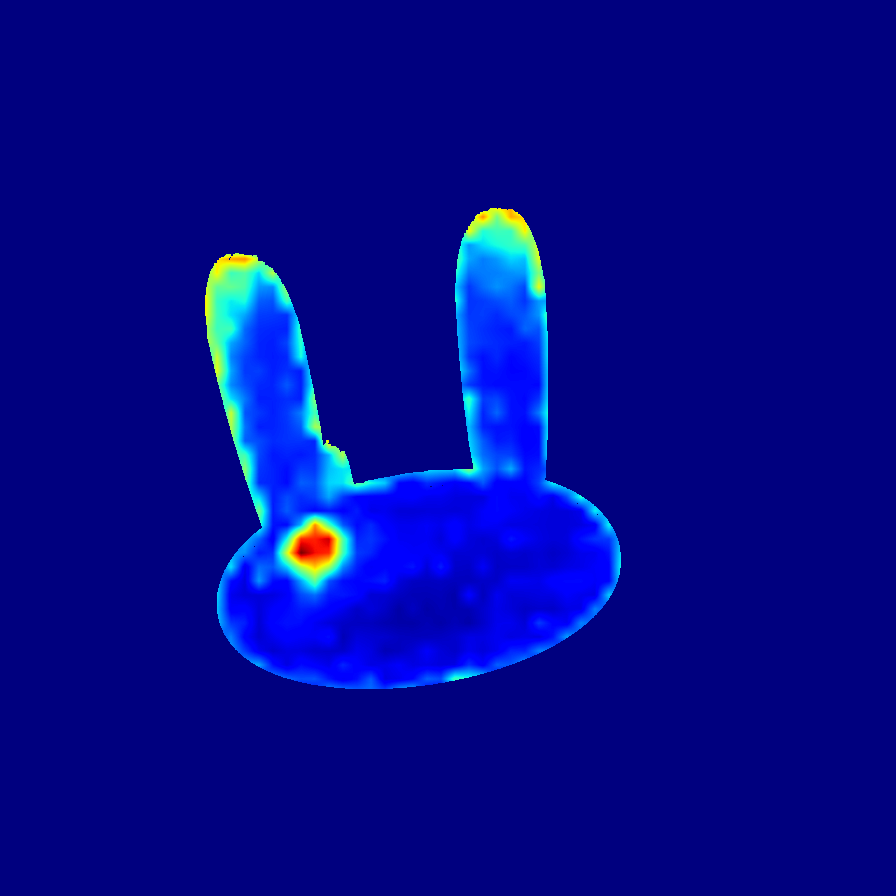} & 
            \includegraphics[width=0.08\linewidth,angle=180,origin=c]{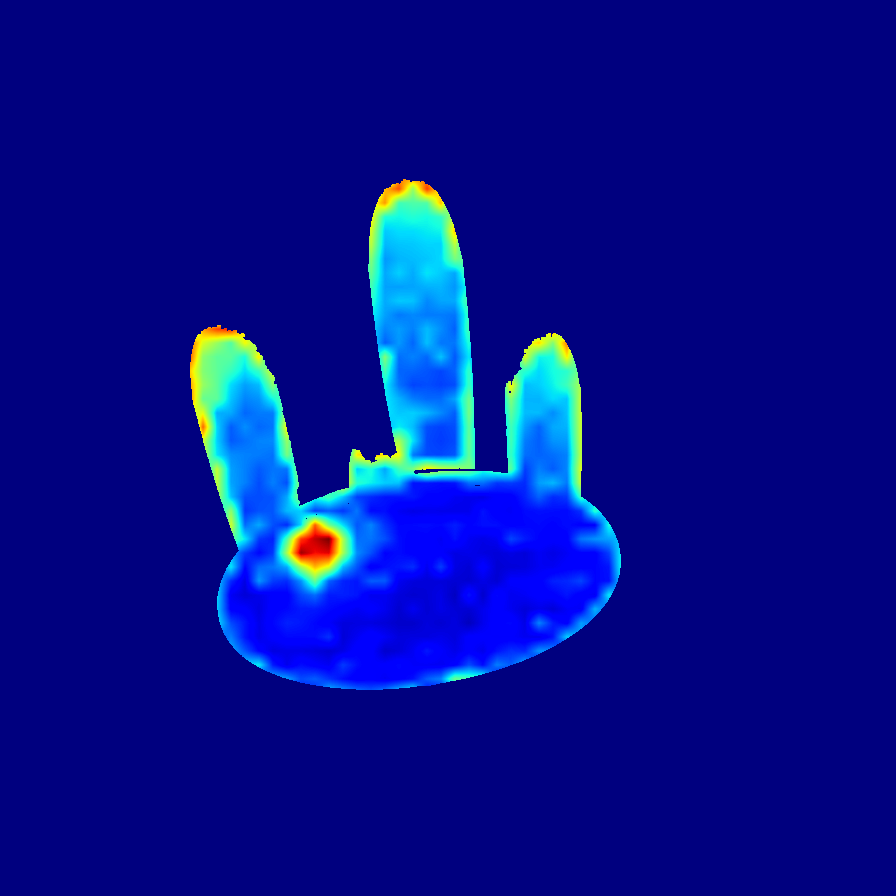} & 
            \includegraphics[width=0.08\linewidth,angle=180,origin=c]{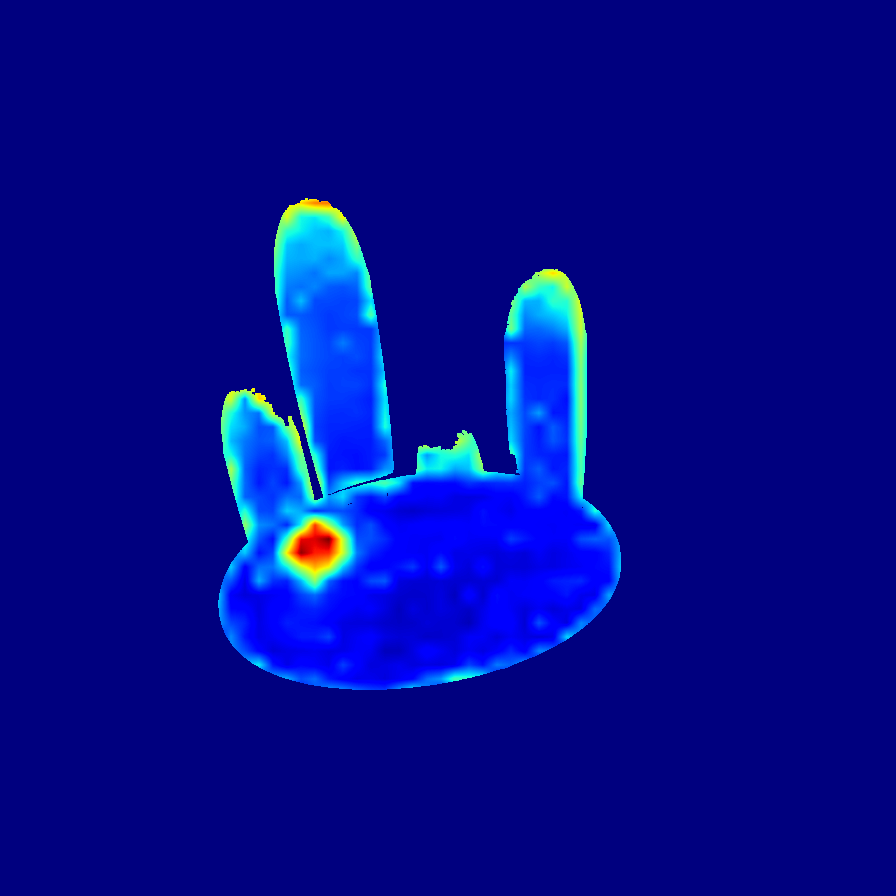} & 
            \includegraphics[width=0.08\linewidth,angle=180,origin=c]{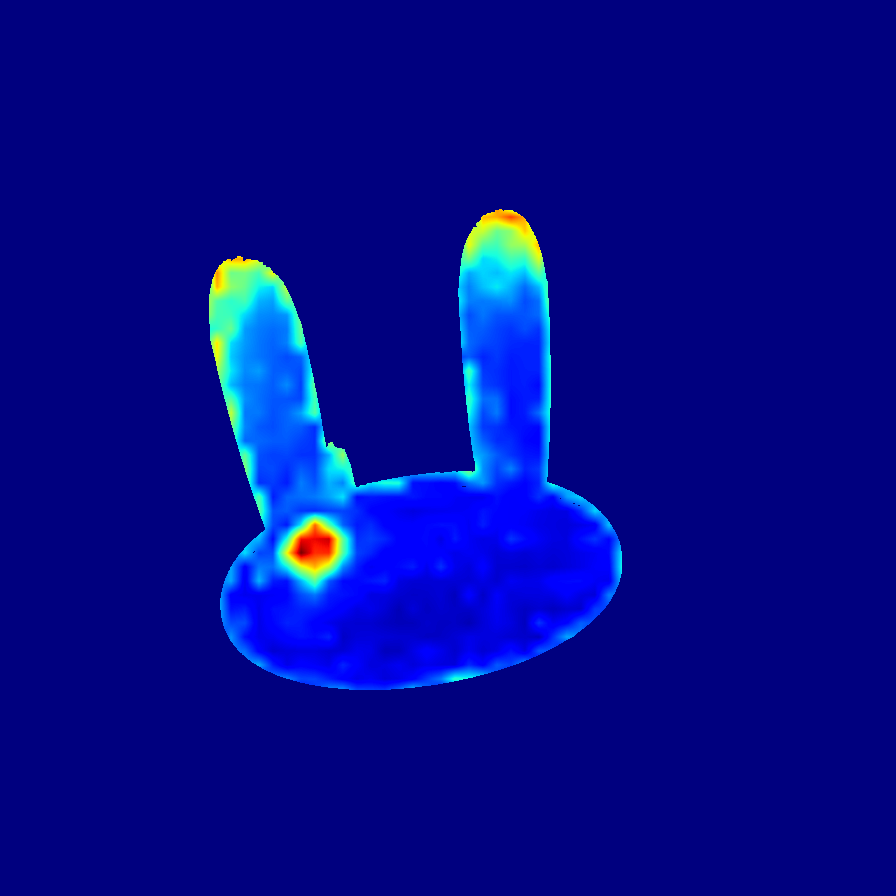} &
            \includegraphics[width=0.08\linewidth,angle=180,origin=c]{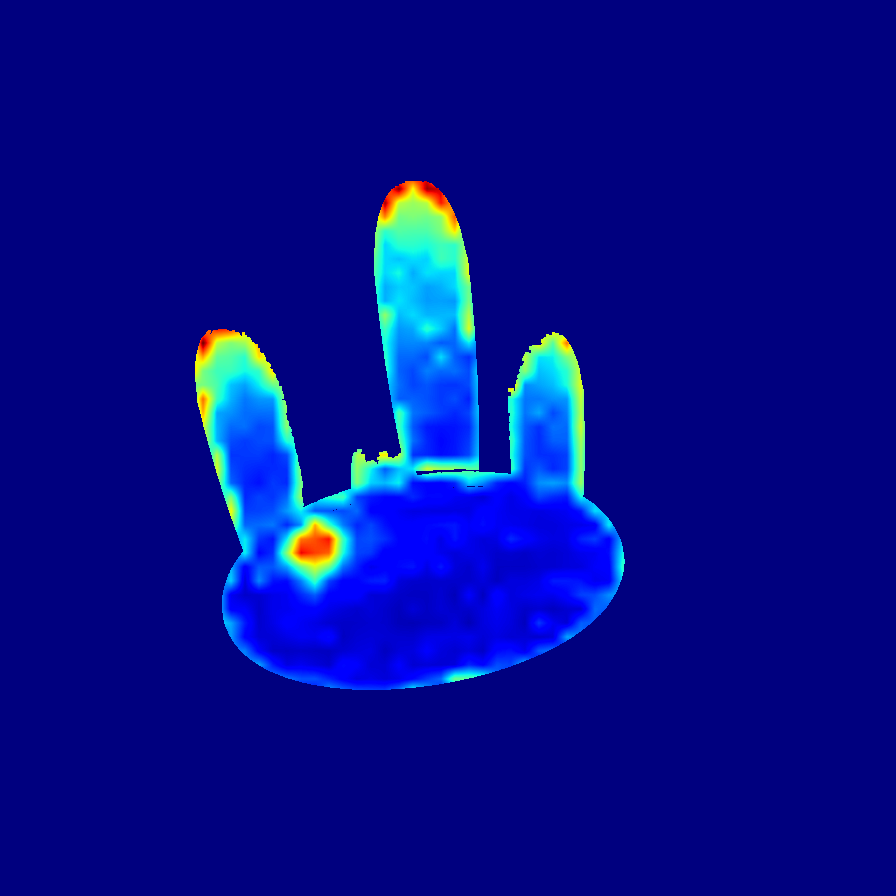} \\

            & \rotatebox{90}{\quad $v_{5}$} &
            \includegraphics[width=0.08\linewidth,angle=180,origin=c]{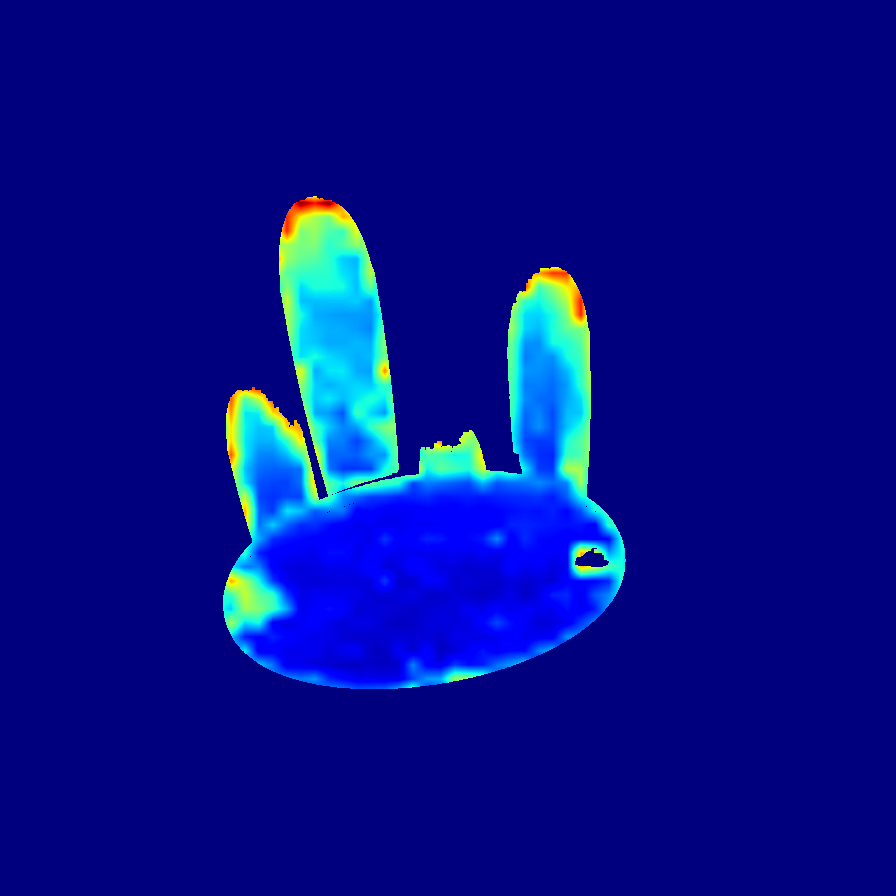} & 
            \includegraphics[width=0.08\linewidth,angle=180,origin=c]{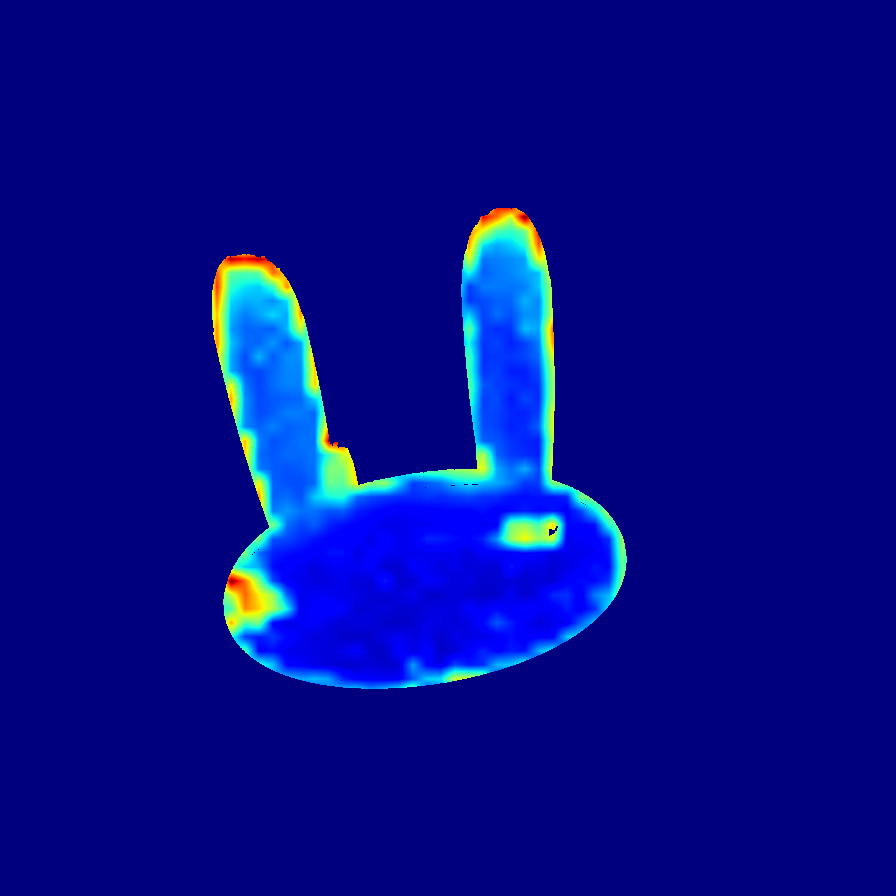} & 
            \includegraphics[width=0.08\linewidth,angle=180,origin=c]{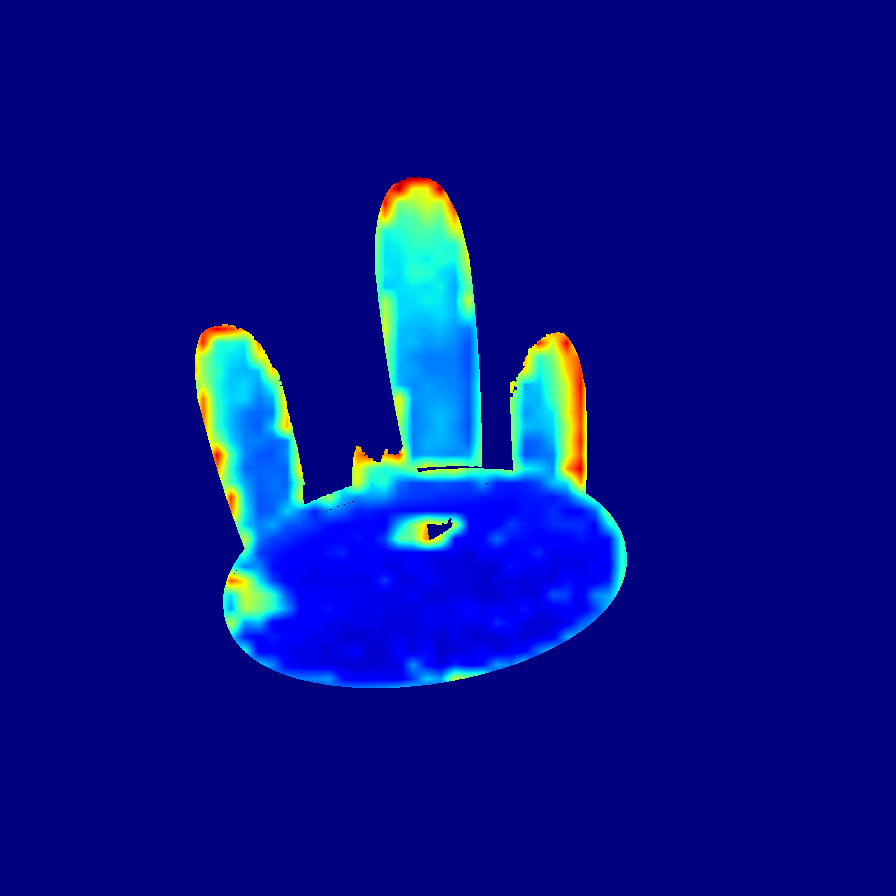} & 
            \includegraphics[width=0.08\linewidth,angle=180,origin=c]{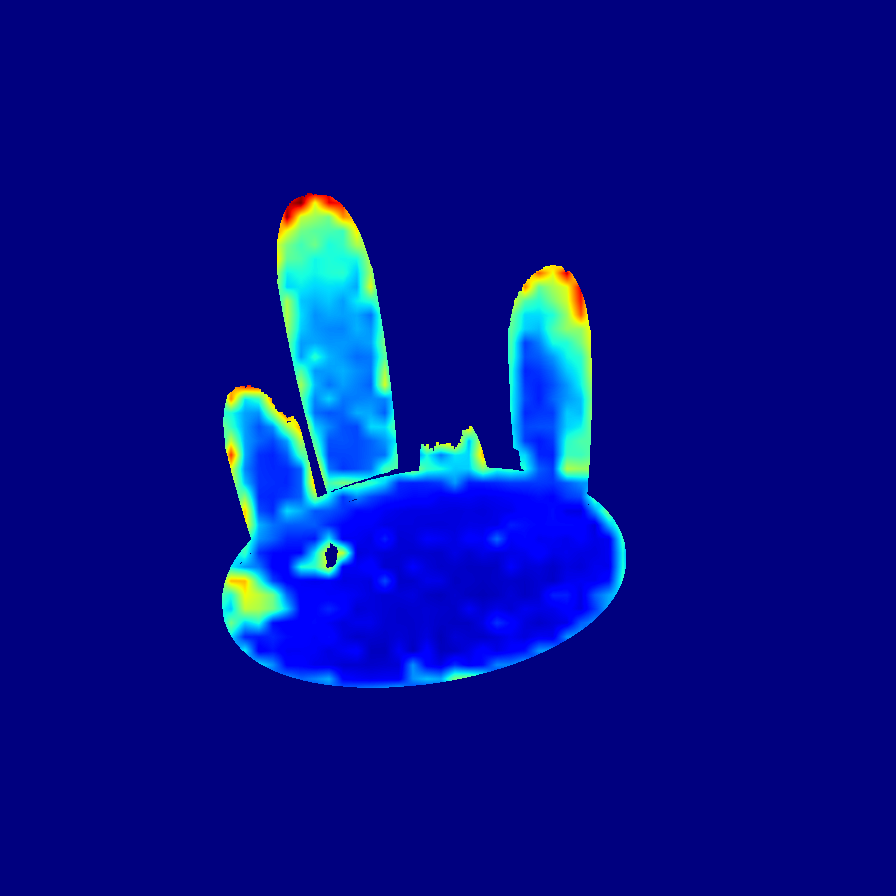} & 
            \includegraphics[width=0.08\linewidth,angle=180,origin=c]{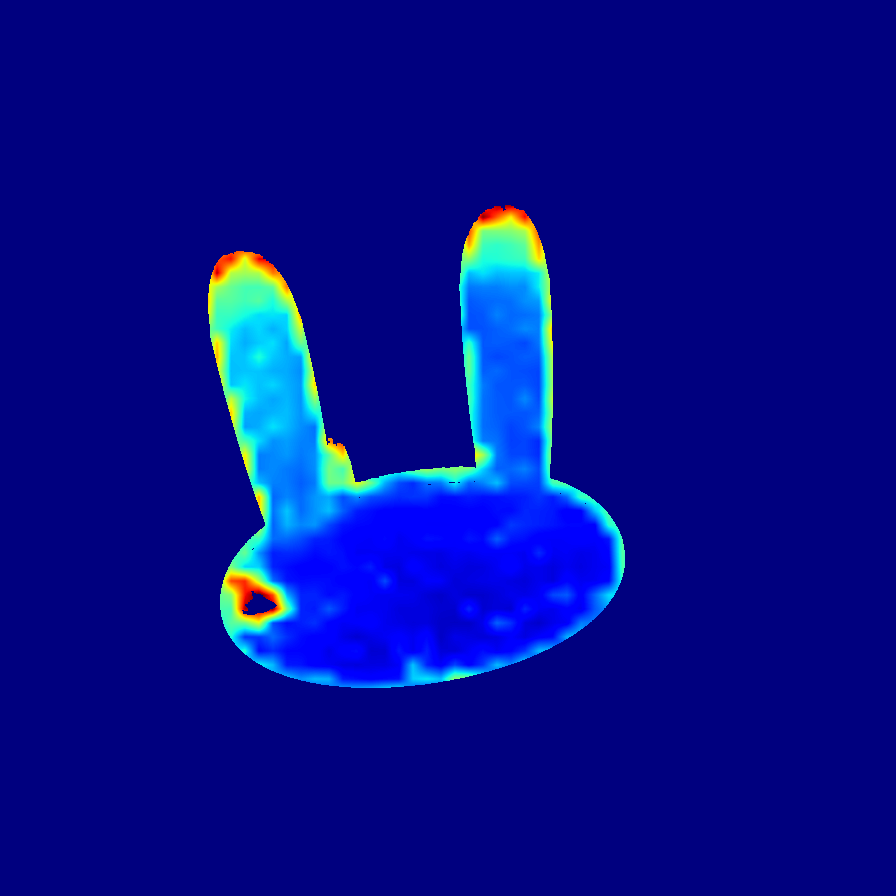} & 
            \includegraphics[width=0.08\linewidth,angle=180,origin=c]{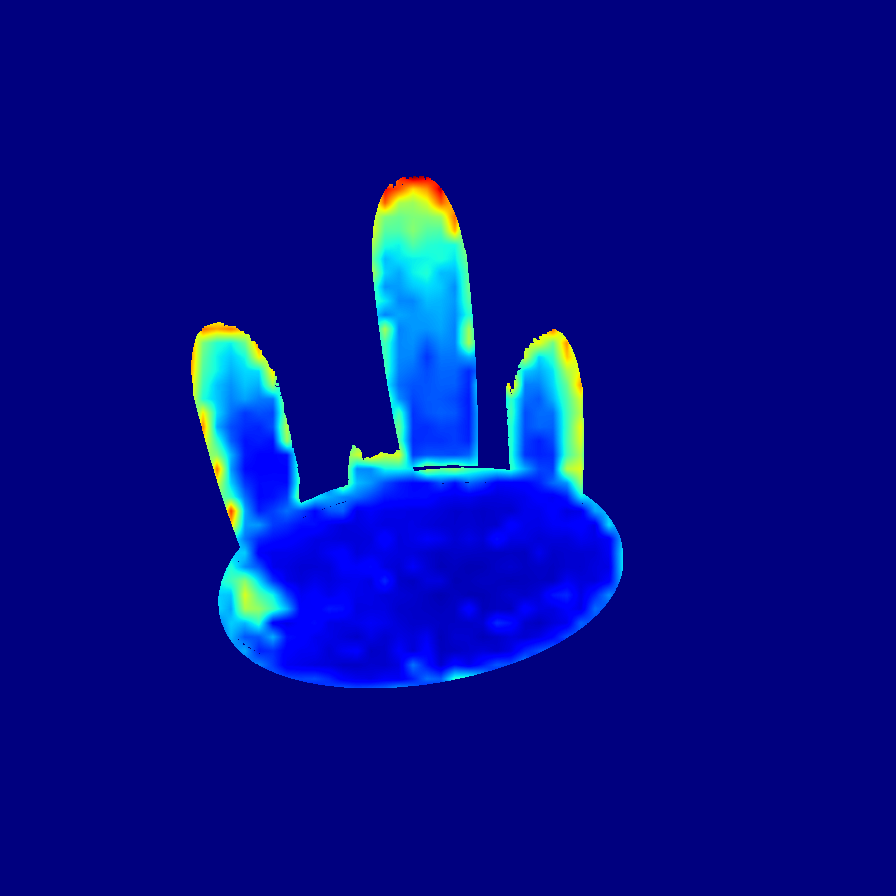} & 
            \includegraphics[width=0.08\linewidth,angle=180,origin=c]{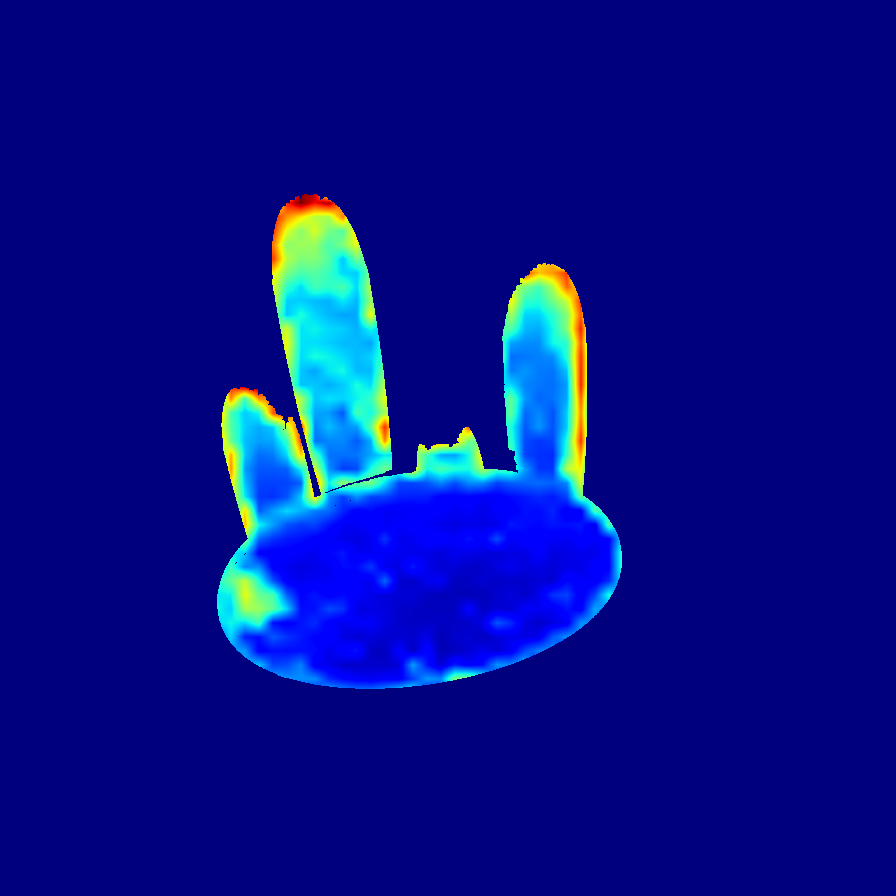} & 
            \includegraphics[width=0.08\linewidth,angle=180,origin=c]{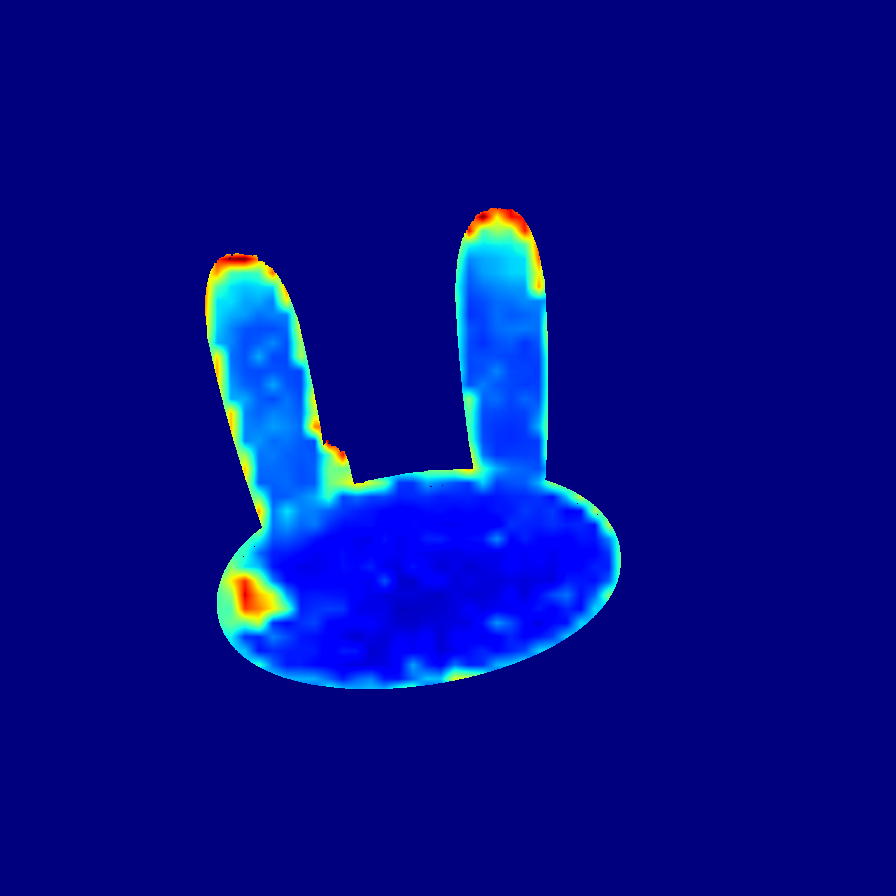} & 
            \includegraphics[width=0.08\linewidth,angle=180,origin=c]{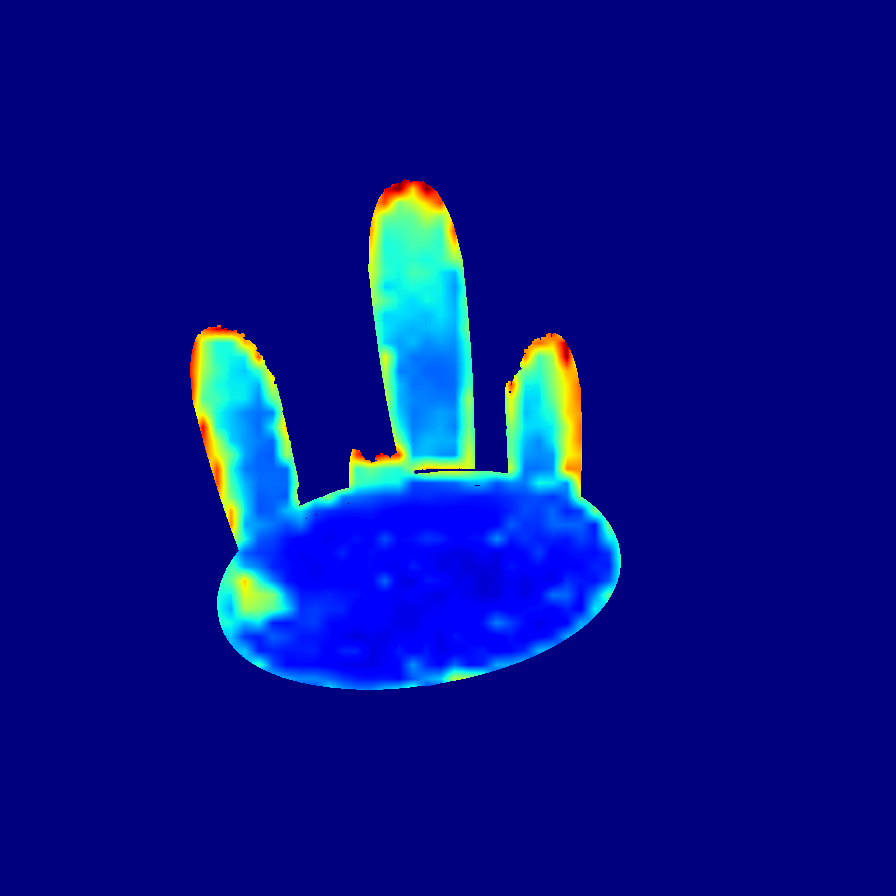} & 
            \includegraphics[width=0.08\linewidth,angle=180,origin=c]{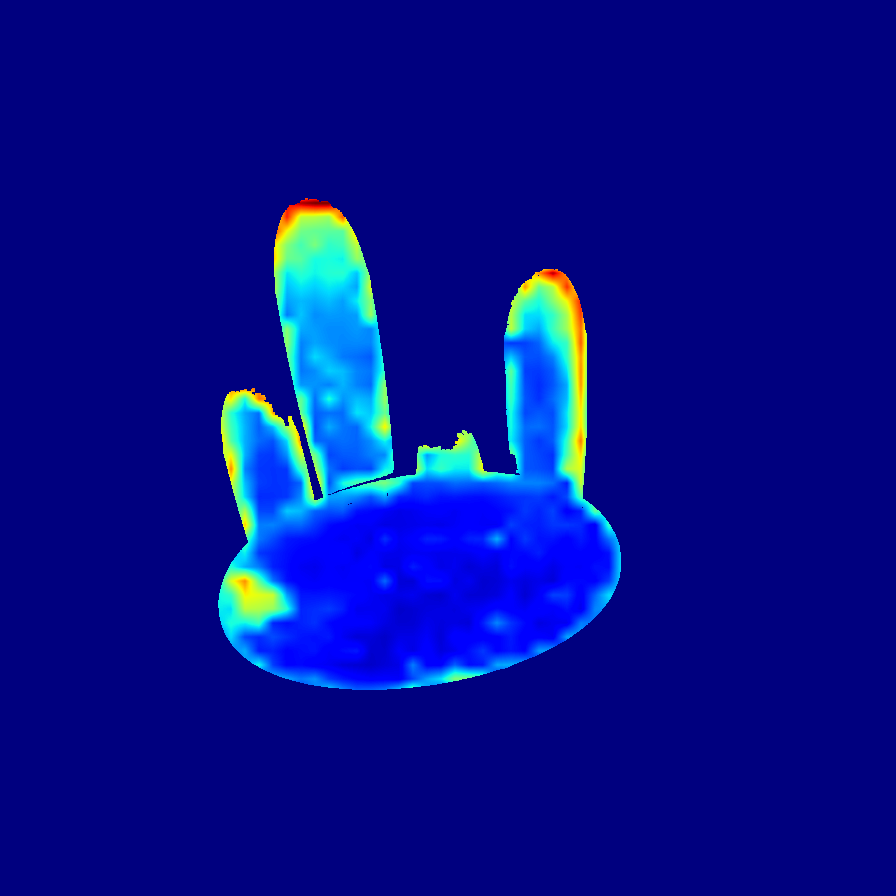} & 
            \includegraphics[width=0.08\linewidth,angle=180,origin=c]{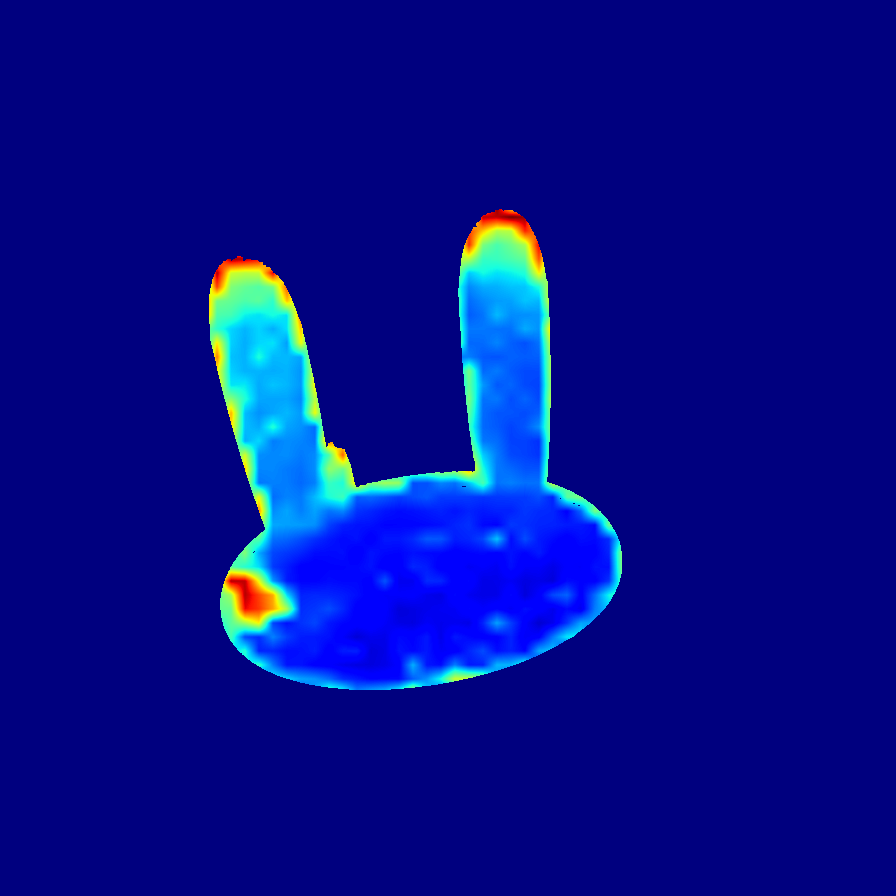} &
            \includegraphics[width=0.08\linewidth,angle=180,origin=c]{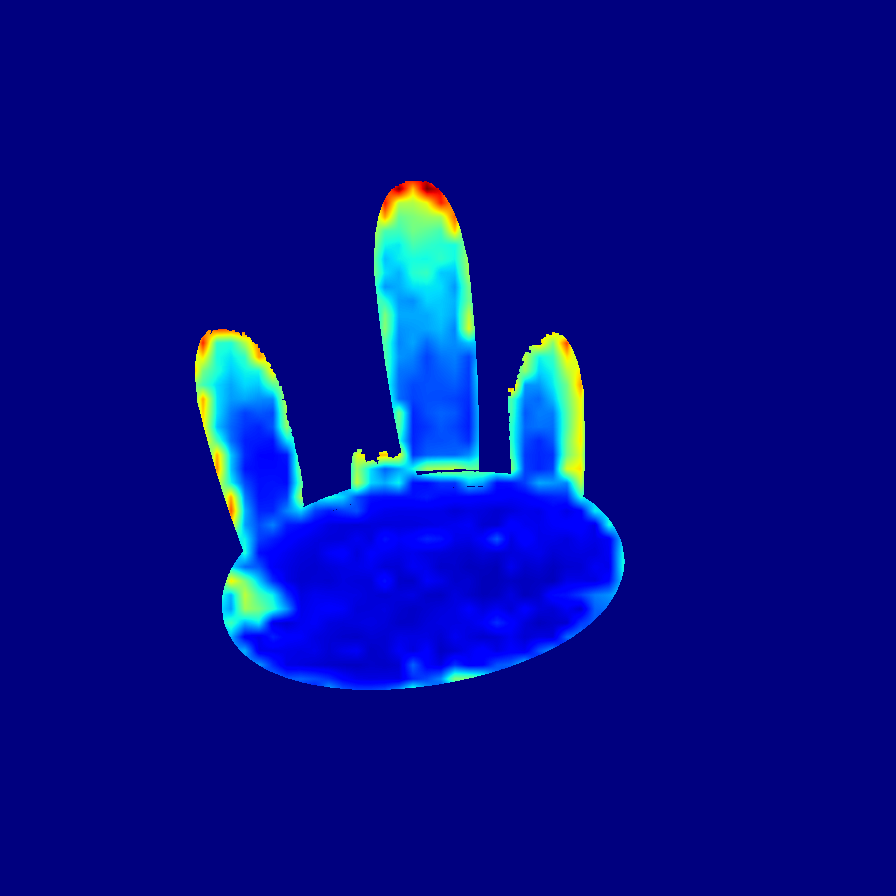} \\

            & \rotatebox{90}{\quad $v_{6}$} &
            \includegraphics[width=0.08\linewidth,angle=180,origin=c]{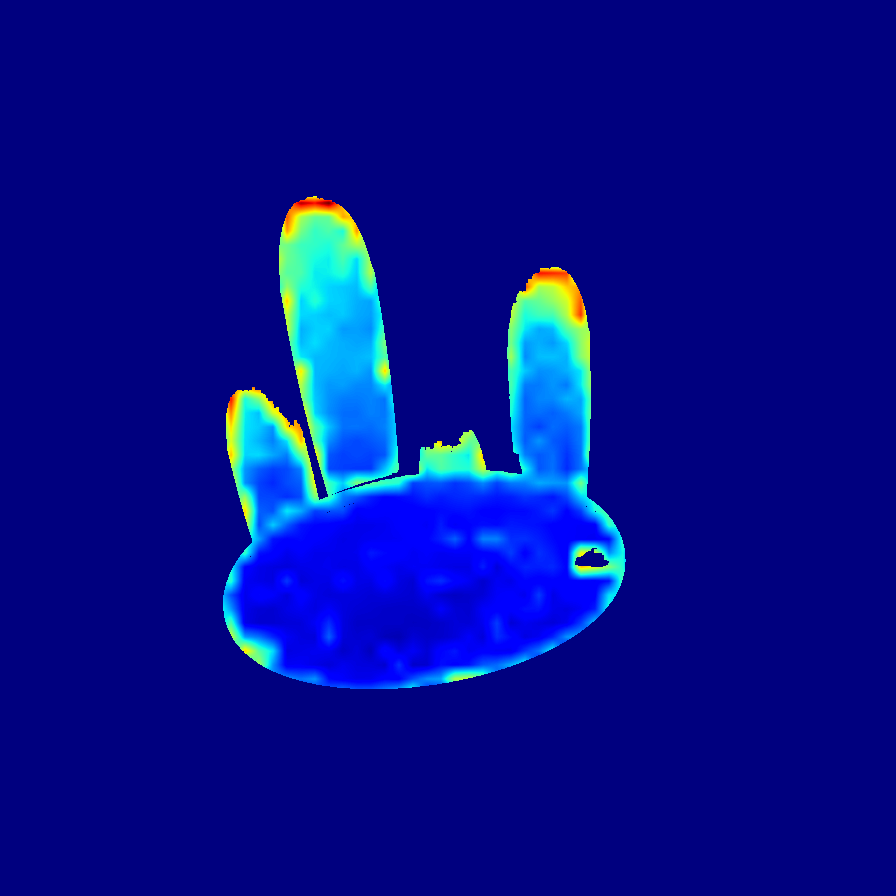} & 
            \includegraphics[width=0.08\linewidth,angle=180,origin=c]{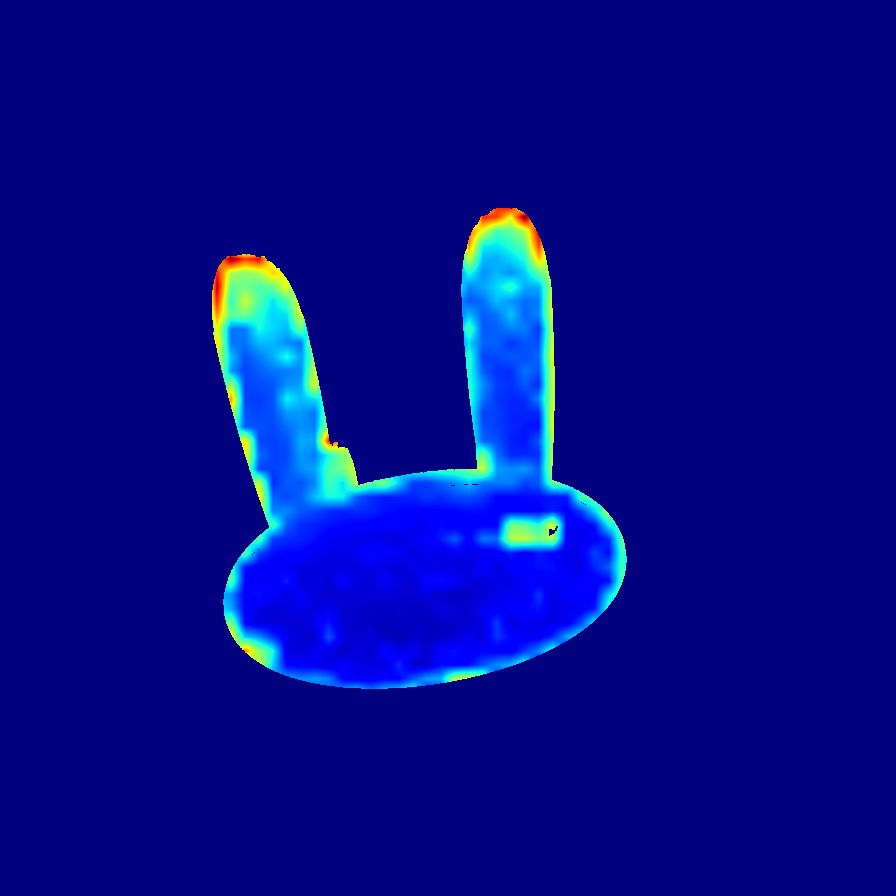} & 
            \includegraphics[width=0.08\linewidth,angle=180,origin=c]{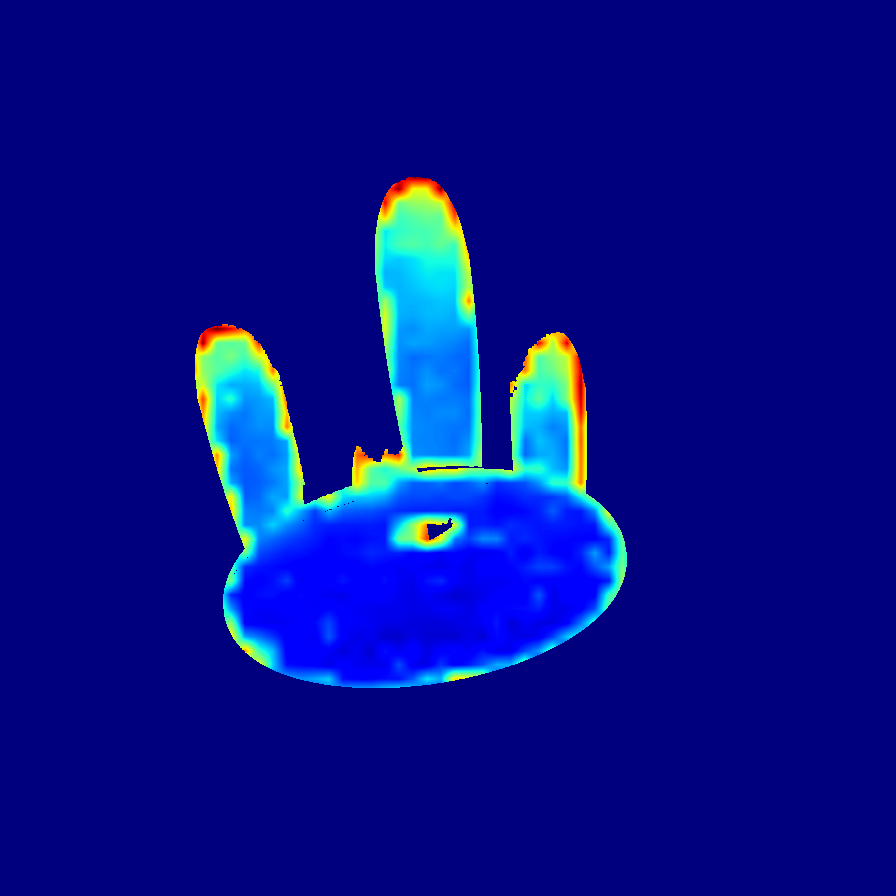} & 
            \includegraphics[width=0.08\linewidth,angle=180,origin=c]{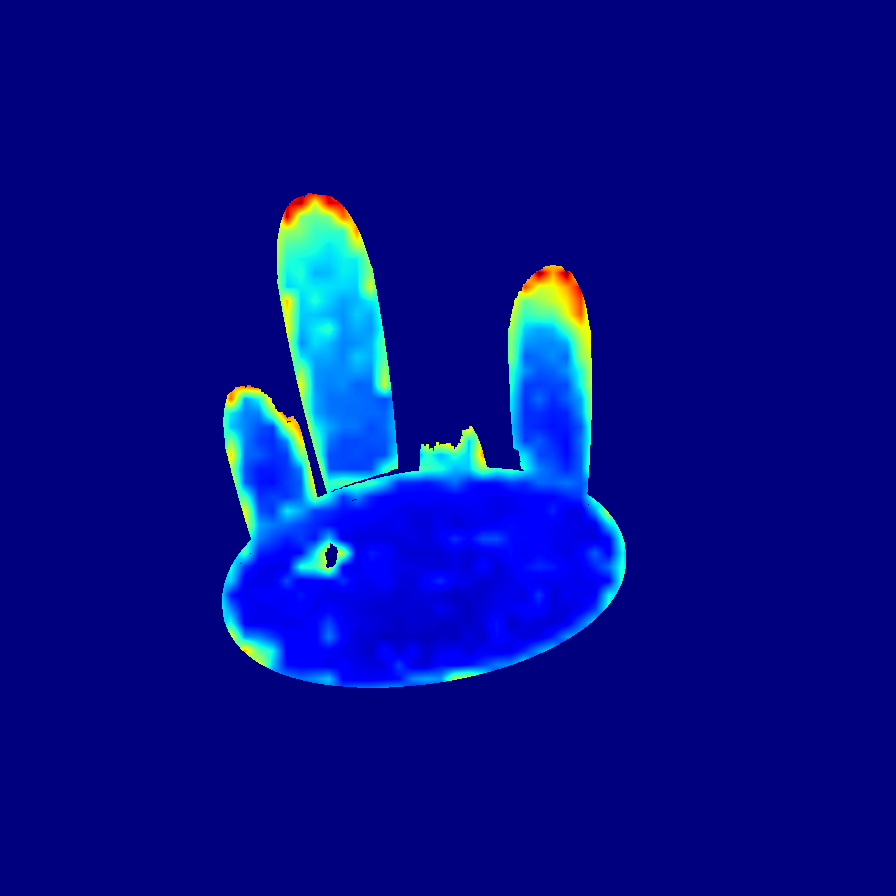} & 
            \includegraphics[width=0.08\linewidth,angle=180,origin=c]{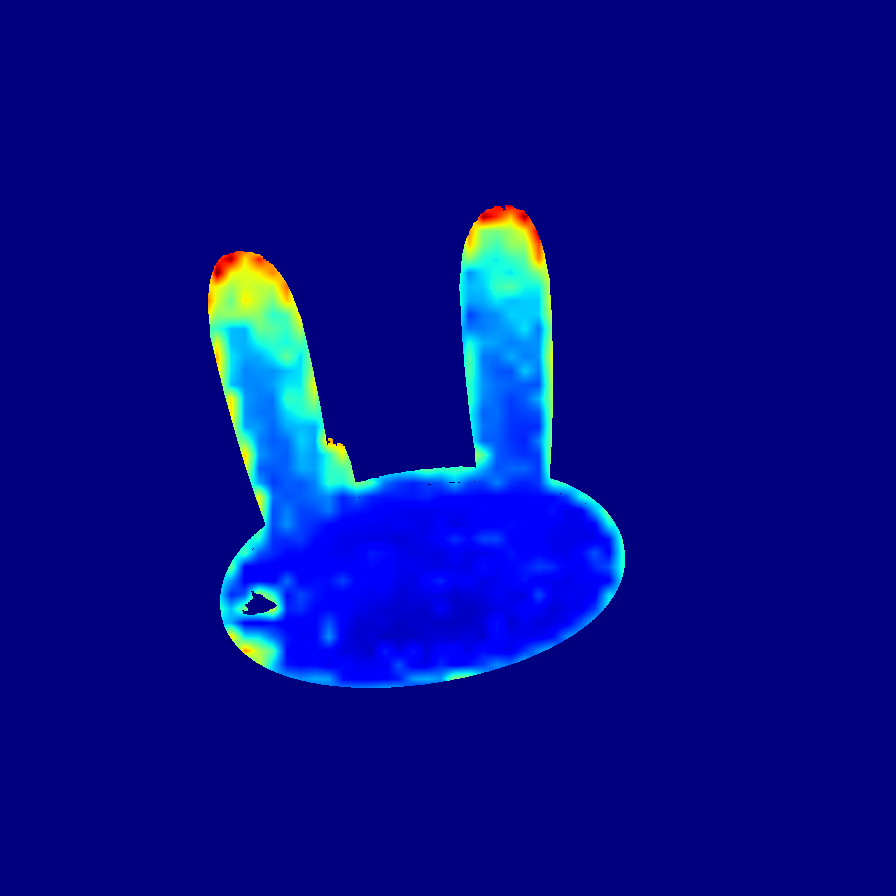} & 
            \includegraphics[width=0.08\linewidth,angle=180,origin=c]{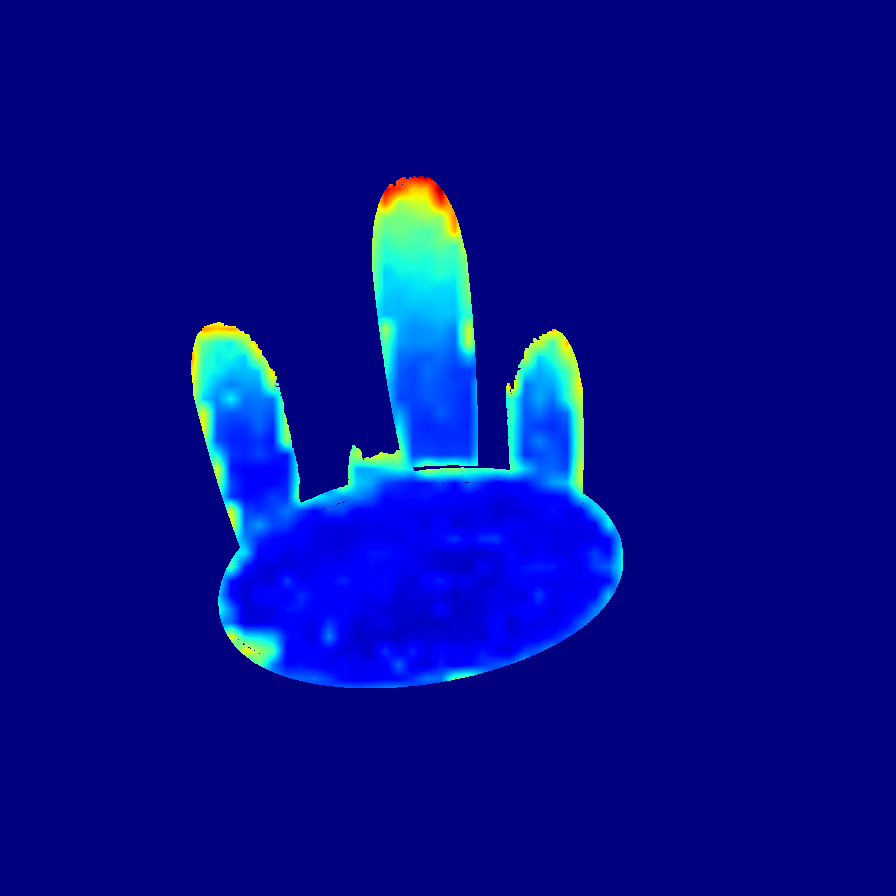} & 
            \includegraphics[width=0.08\linewidth,angle=180,origin=c]{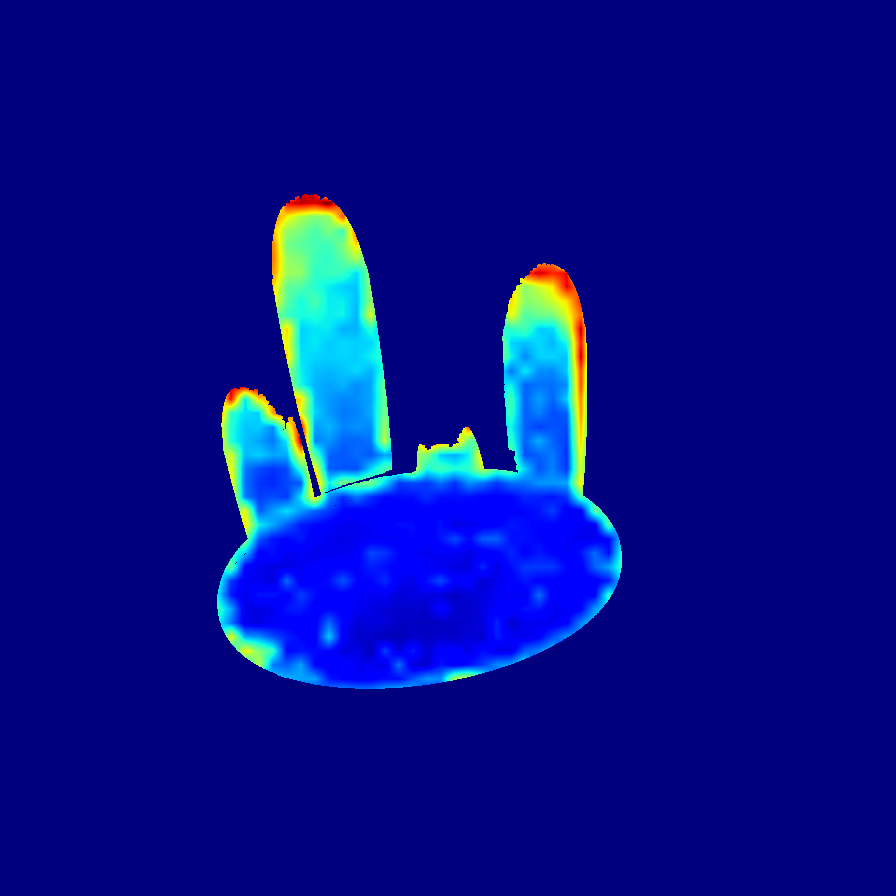} & 
            \includegraphics[width=0.08\linewidth,angle=180,origin=c]{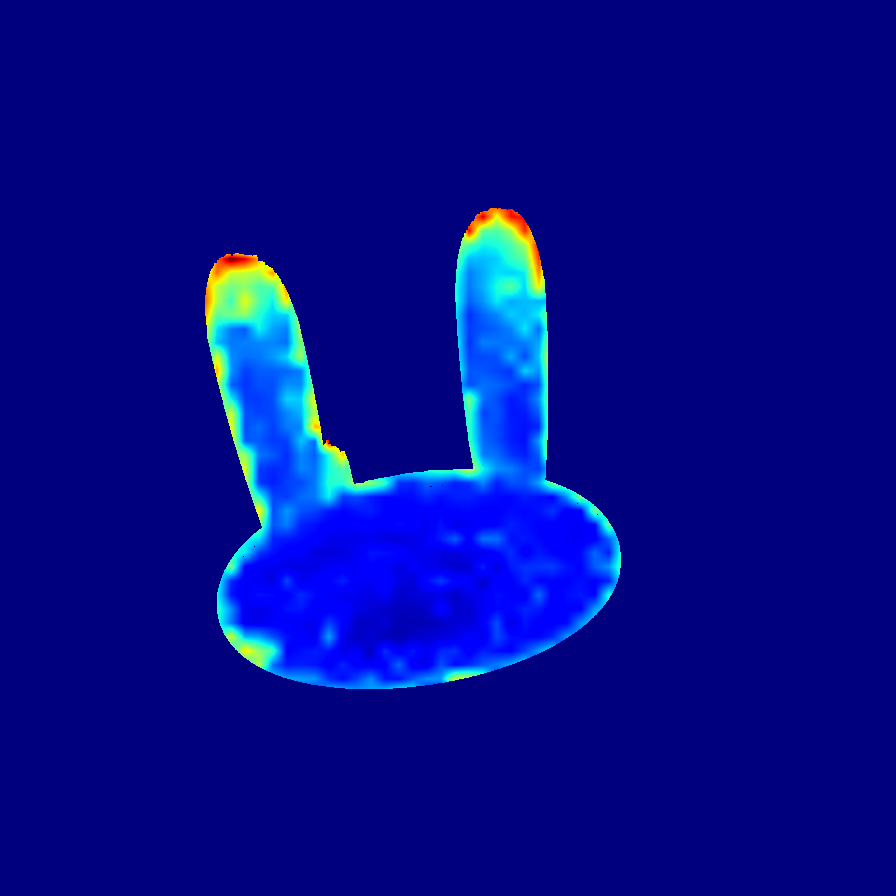} & 
            \includegraphics[width=0.08\linewidth,angle=180,origin=c]{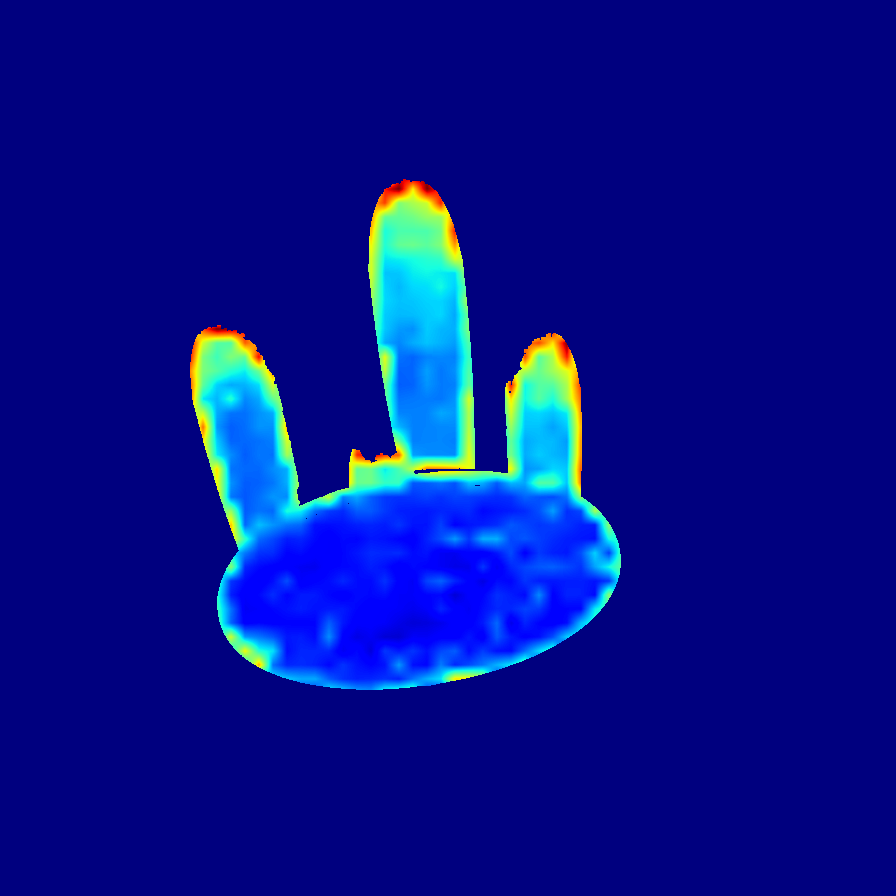} & 
            \includegraphics[width=0.08\linewidth,angle=180,origin=c]{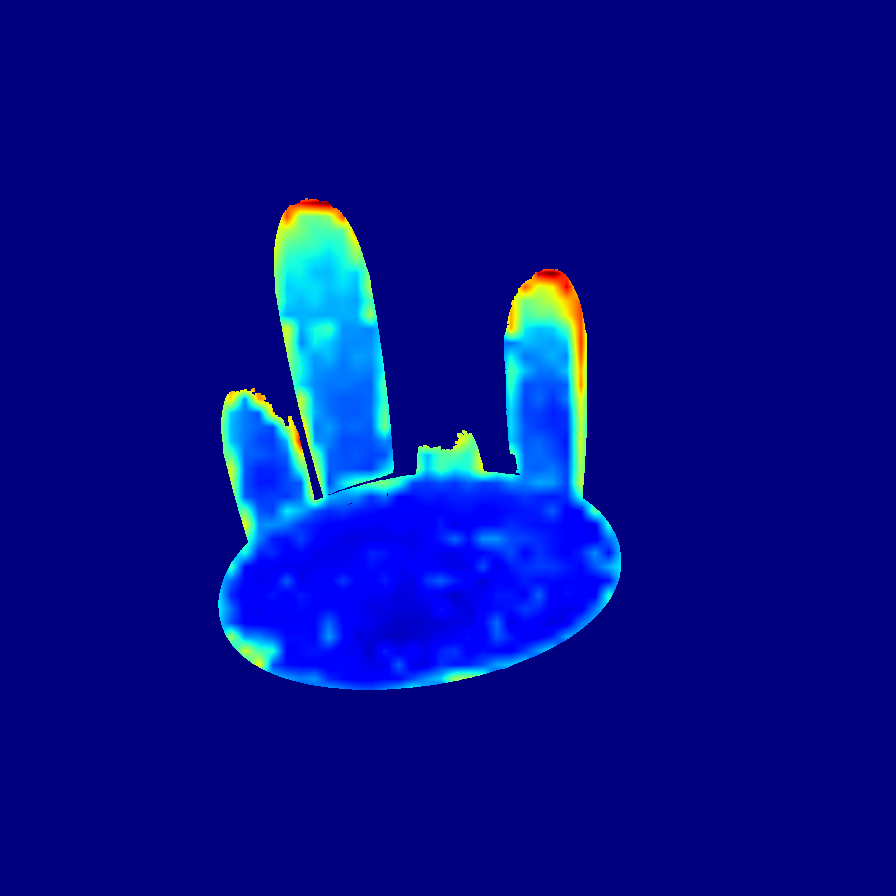} & 
            \includegraphics[width=0.08\linewidth,angle=180,origin=c]{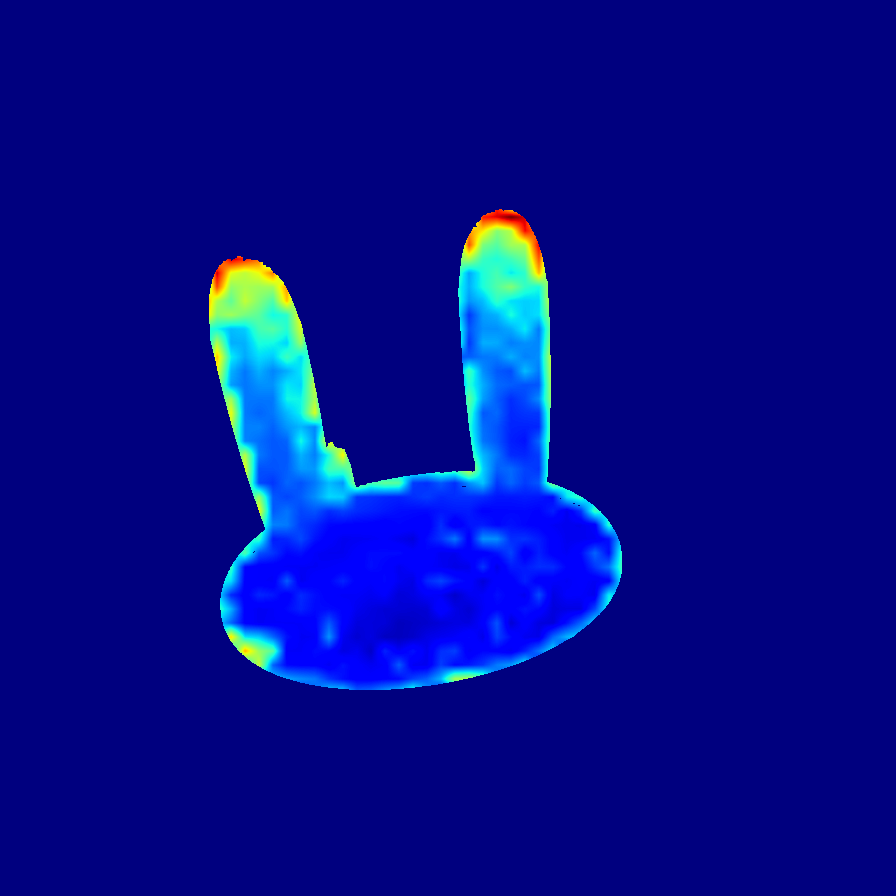} &
            \includegraphics[width=0.08\linewidth,angle=180,origin=c]{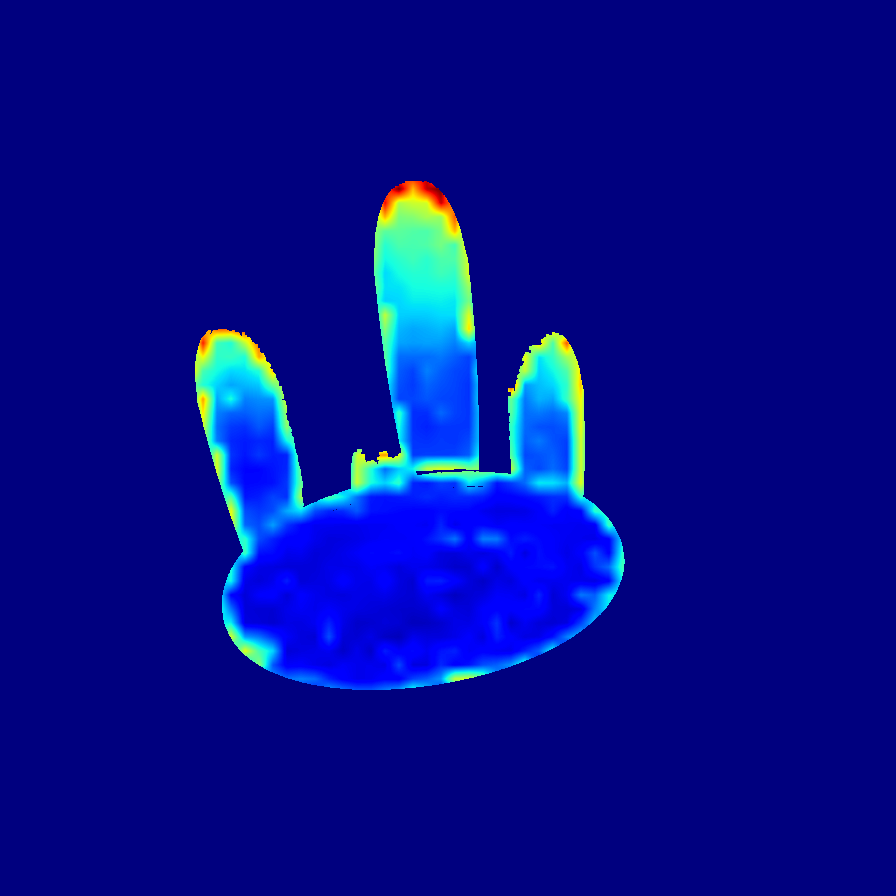} \\

            & \rotatebox{90}{\quad $v_{7}$} &
            \includegraphics[width=0.08\linewidth,angle=180,origin=c]{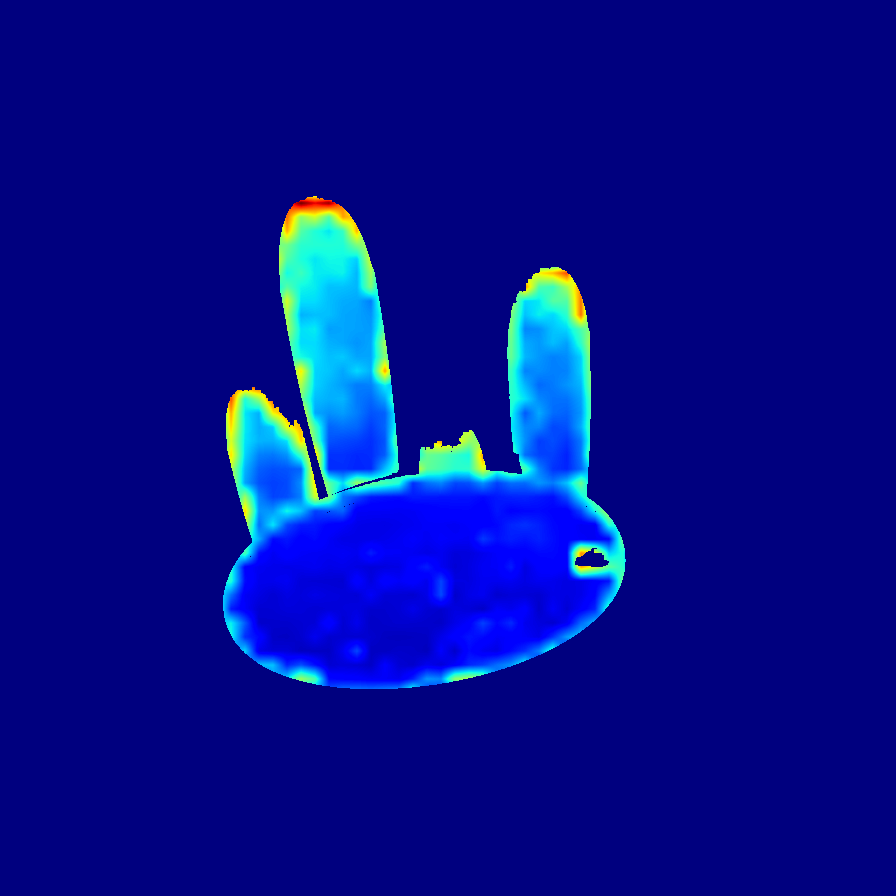} & 
            \includegraphics[width=0.08\linewidth,angle=180,origin=c]{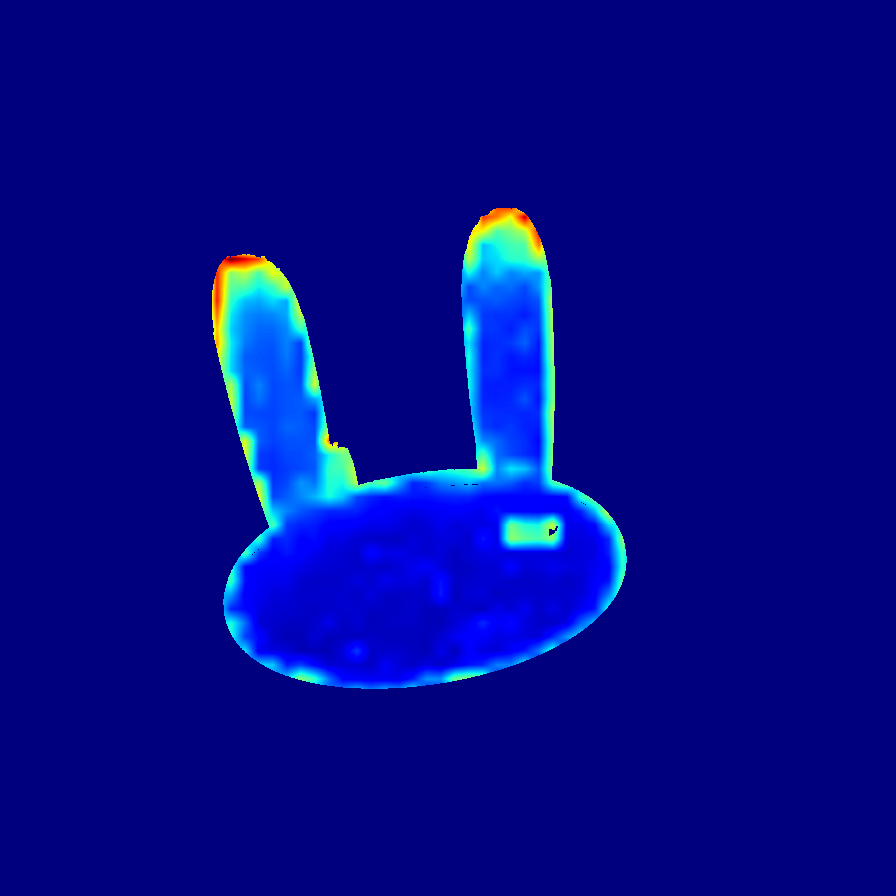} & 
            \includegraphics[width=0.08\linewidth,angle=180,origin=c]{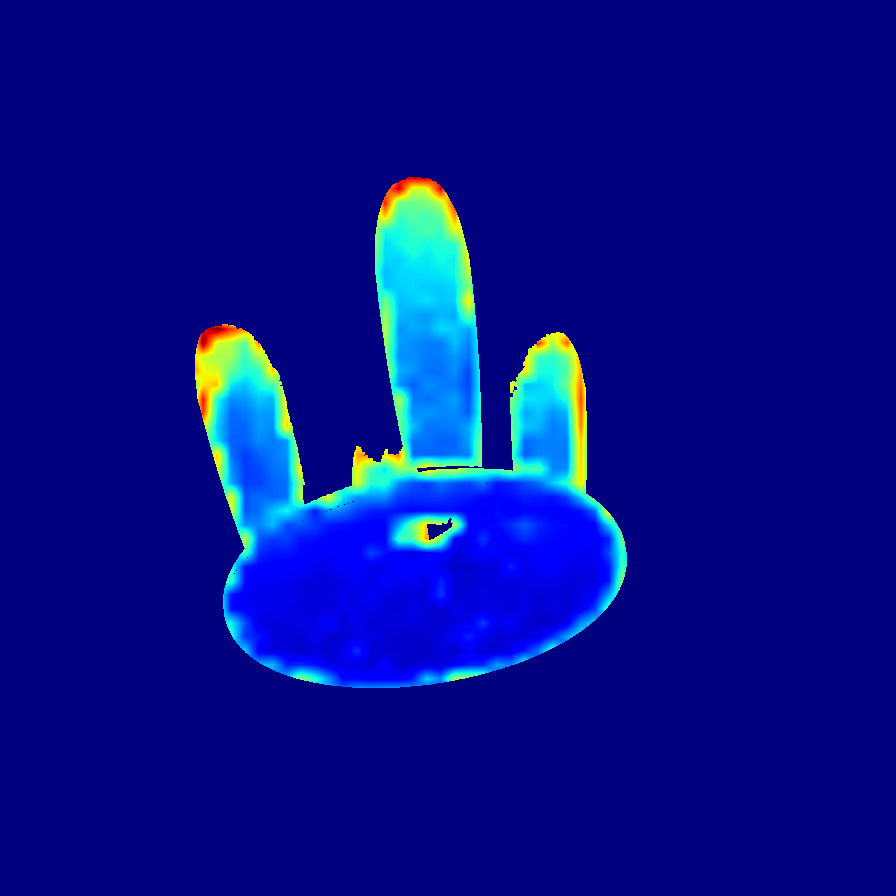} & 
            \includegraphics[width=0.08\linewidth,angle=180,origin=c]{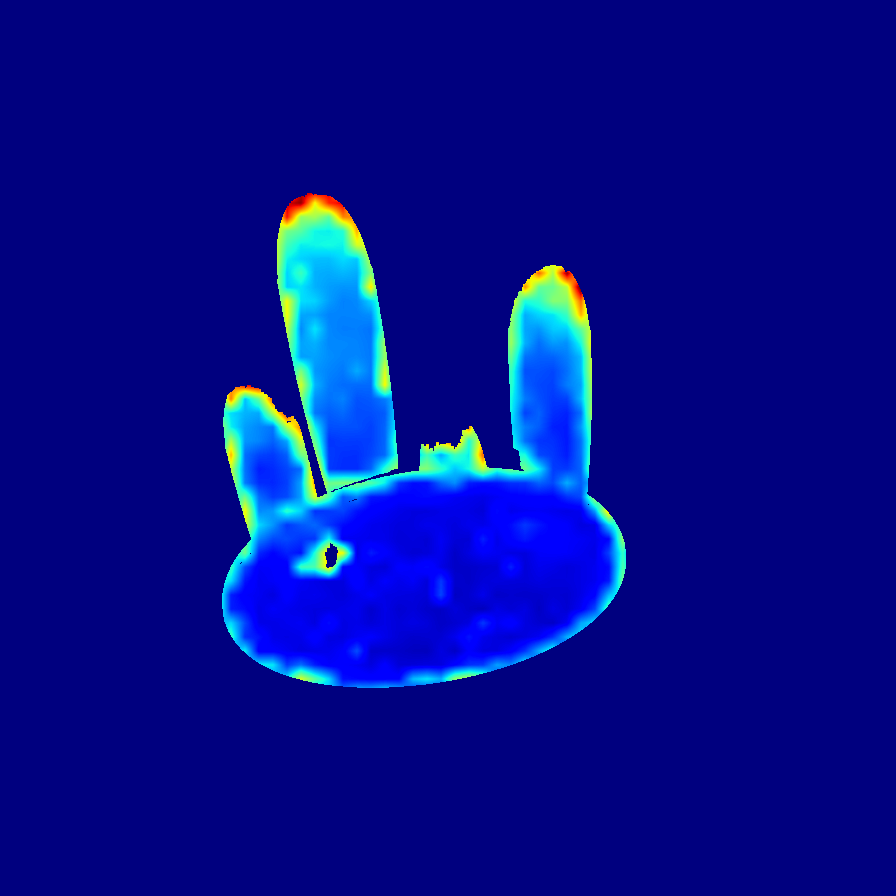} & 
            \includegraphics[width=0.08\linewidth,angle=180,origin=c]{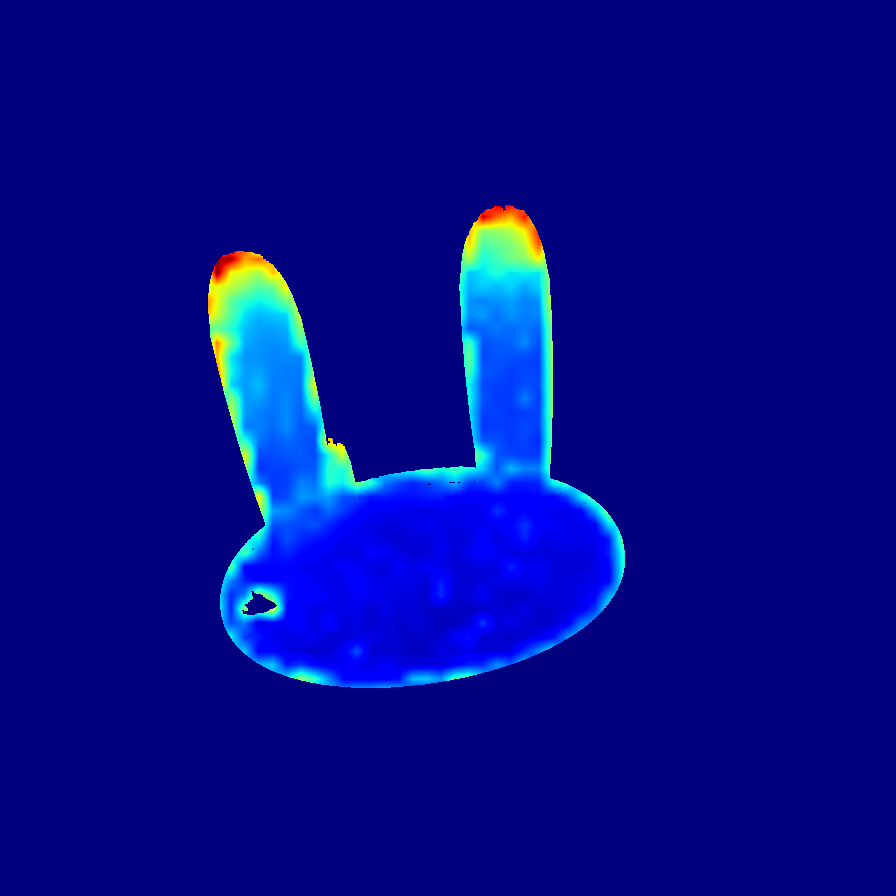} & 
            \includegraphics[width=0.08\linewidth,angle=180,origin=c]{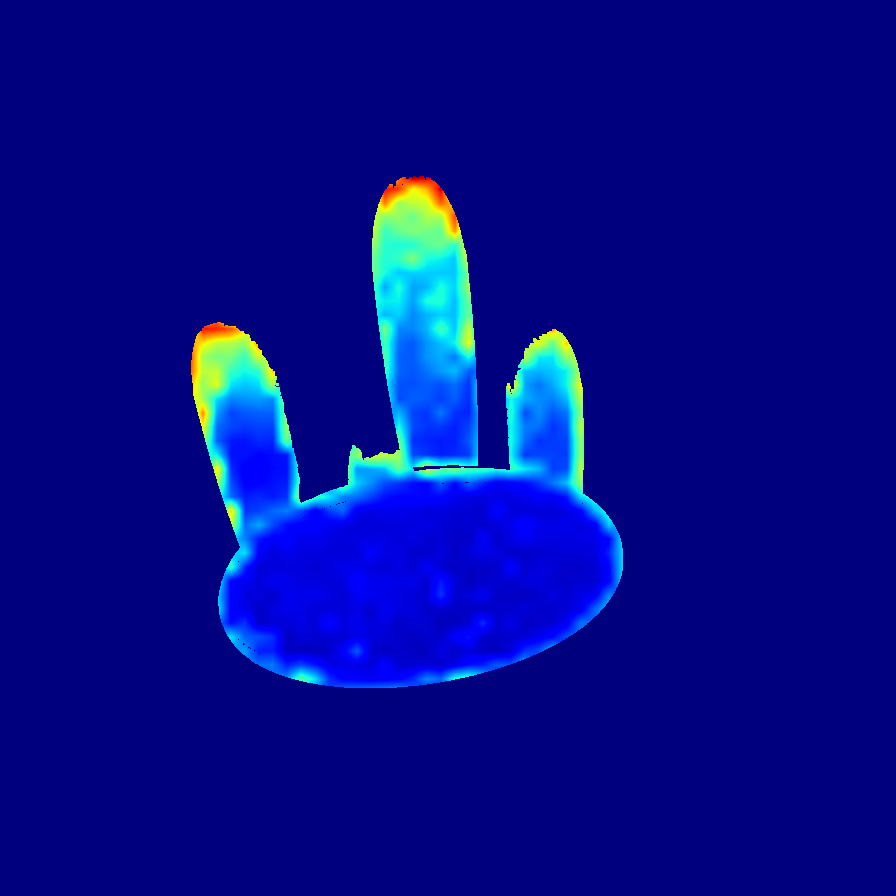} & 
            \includegraphics[width=0.08\linewidth,angle=180,origin=c]{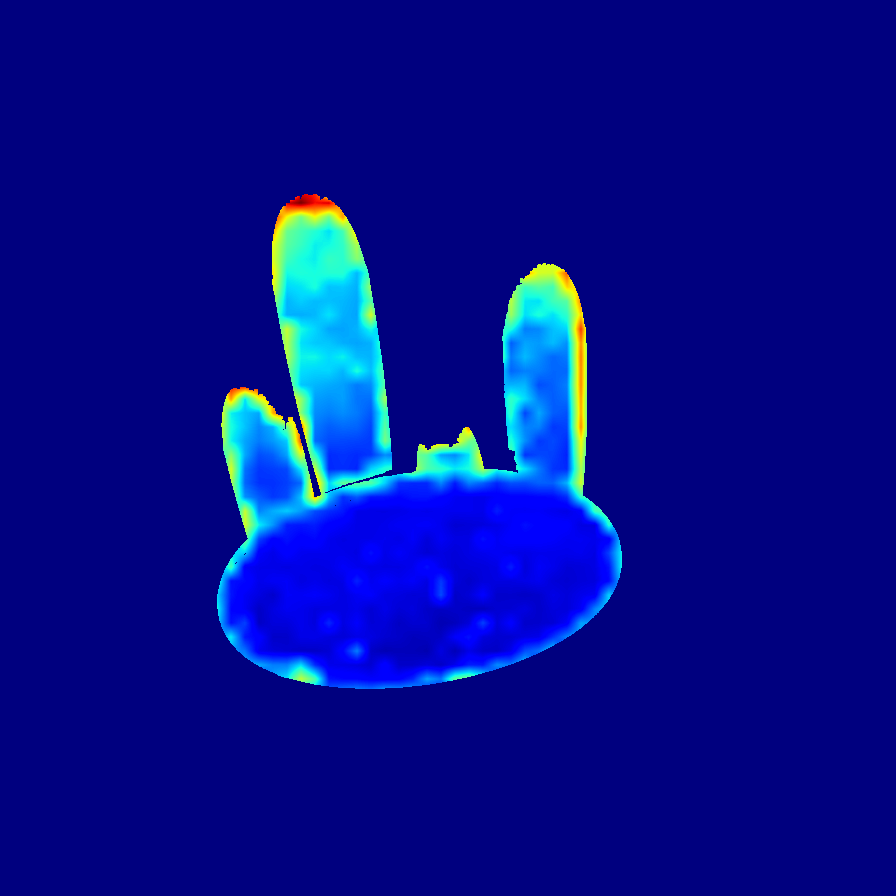} & 
            \includegraphics[width=0.08\linewidth,angle=180,origin=c]{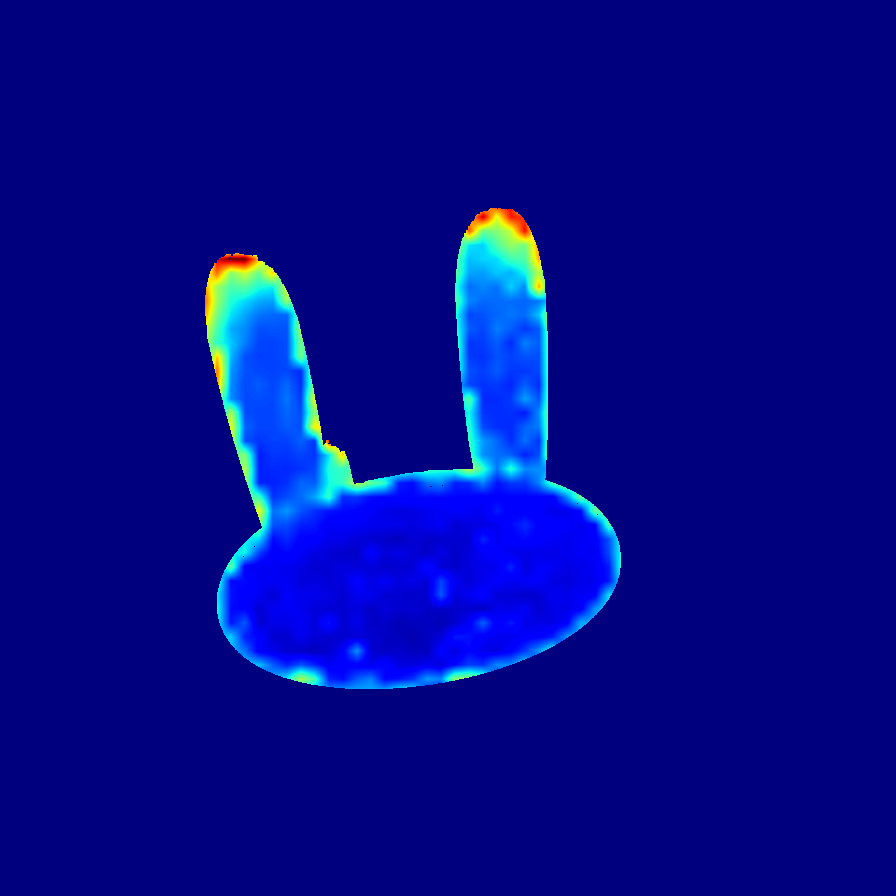} & 
            \includegraphics[width=0.08\linewidth,angle=180,origin=c]{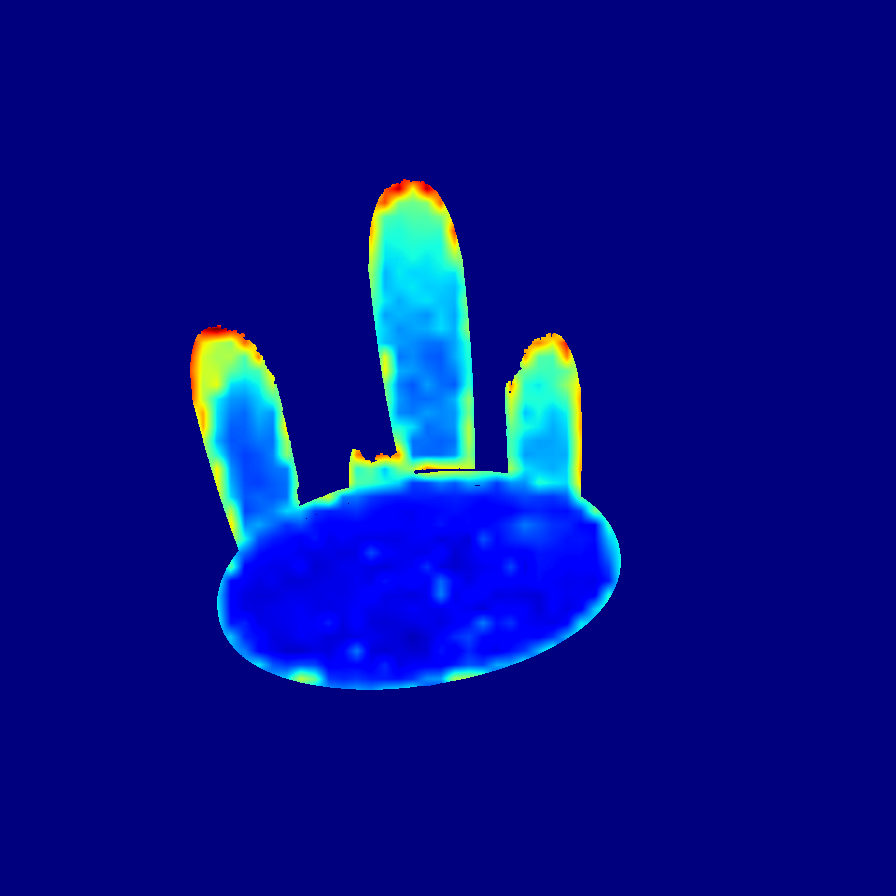} & 
            \includegraphics[width=0.08\linewidth,angle=180,origin=c]{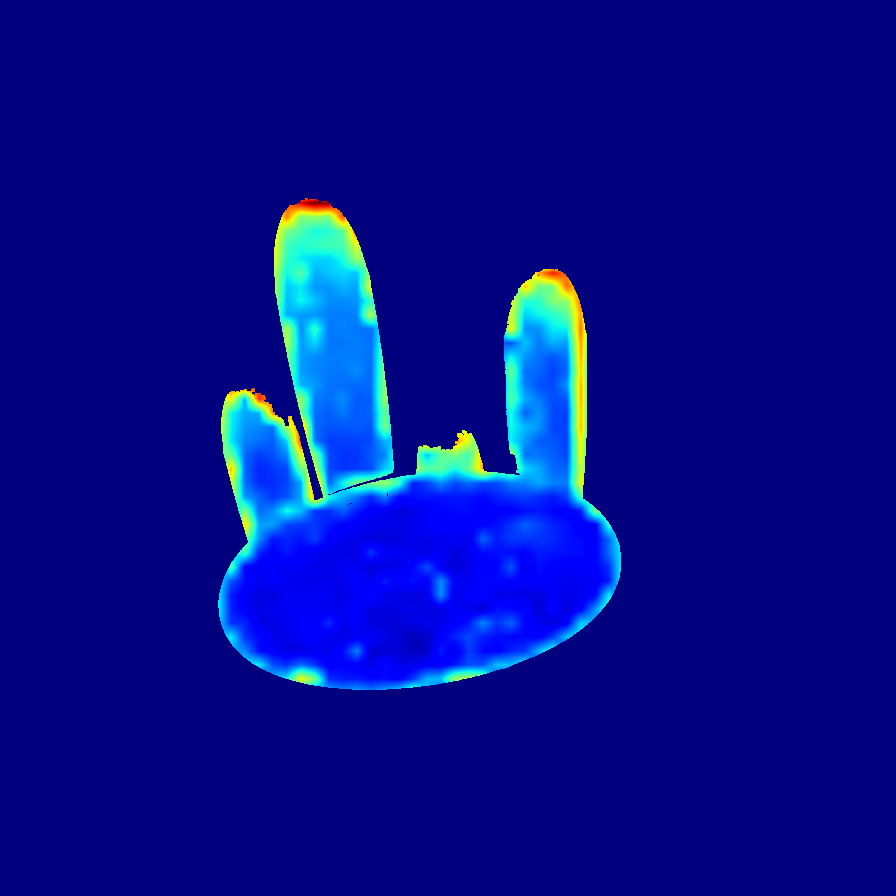} & 
            \includegraphics[width=0.08\linewidth,angle=180,origin=c]{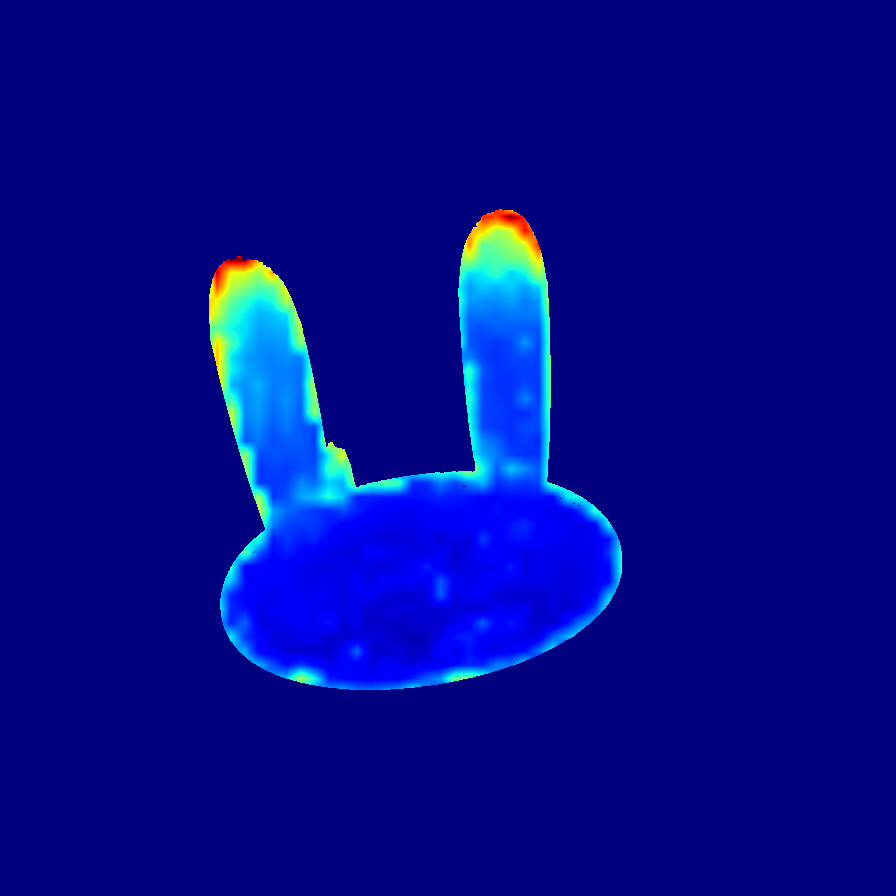} &
            \includegraphics[width=0.08\linewidth,angle=180,origin=c]{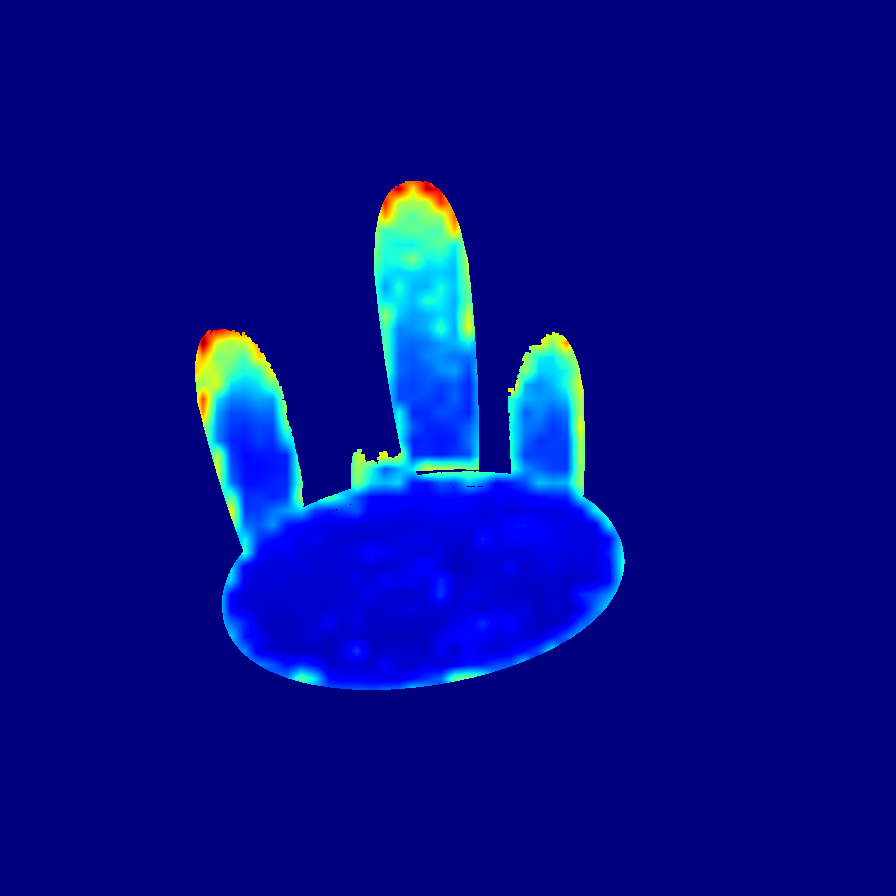} \\

            & \rotatebox{90}{\quad $v_{8}$} &
            \includegraphics[width=0.08\linewidth,angle=180,origin=c]{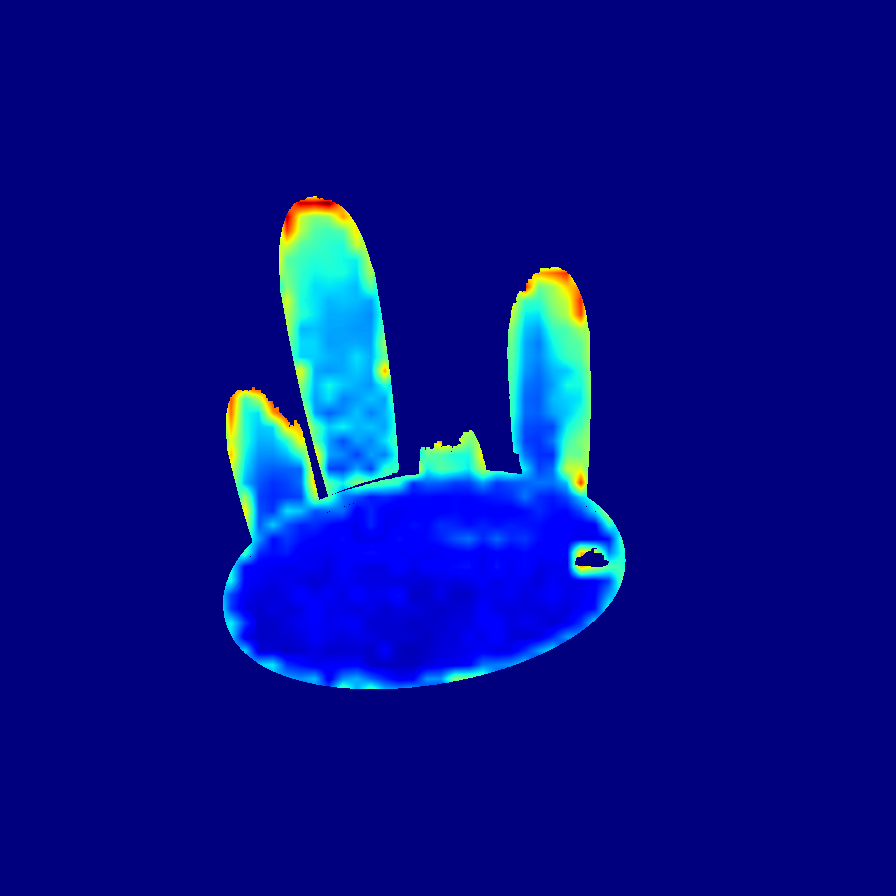} & 
            \includegraphics[width=0.08\linewidth,angle=180,origin=c]{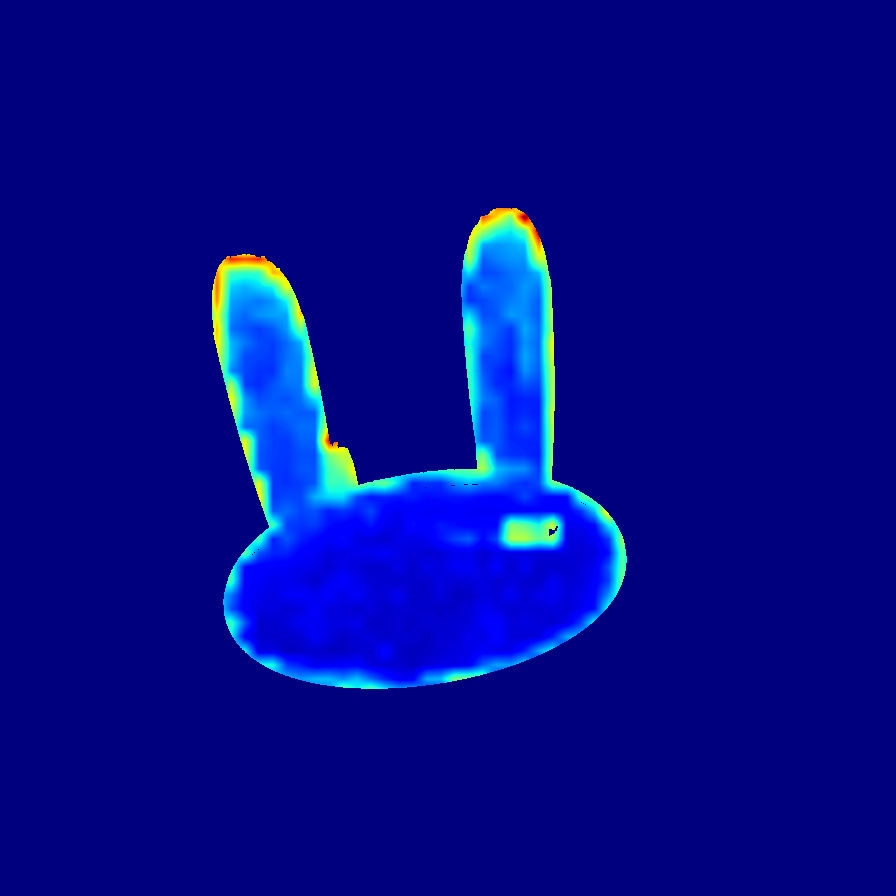} & 
            \includegraphics[width=0.08\linewidth,angle=180,origin=c]{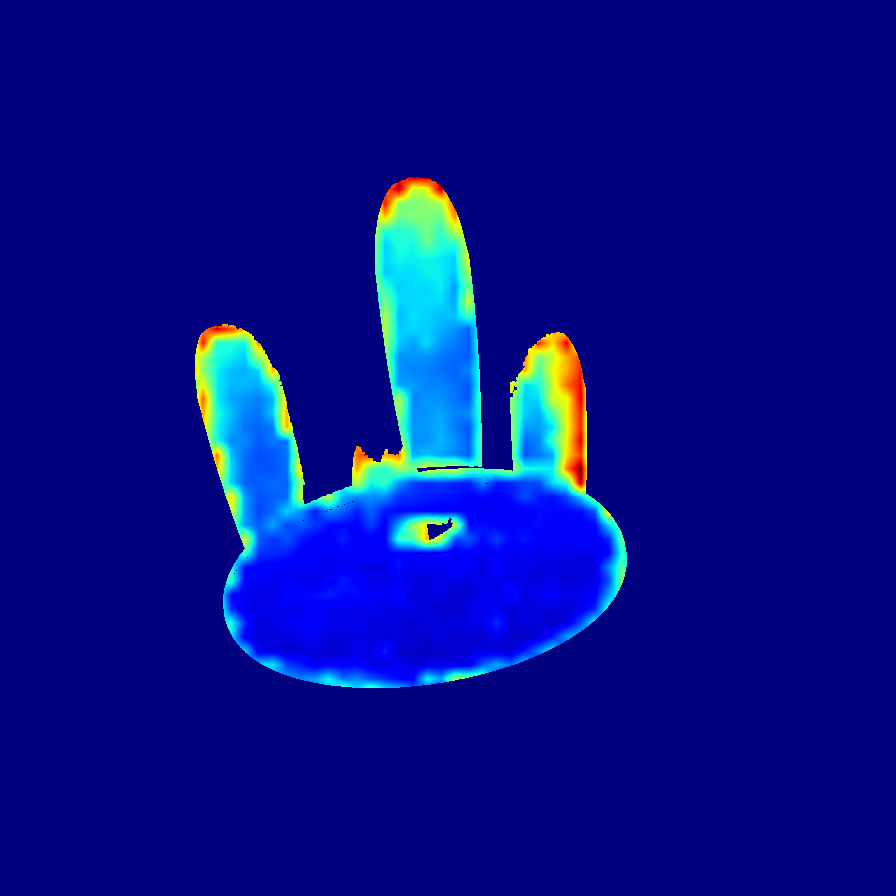} & 
            \includegraphics[width=0.08\linewidth,angle=180,origin=c]{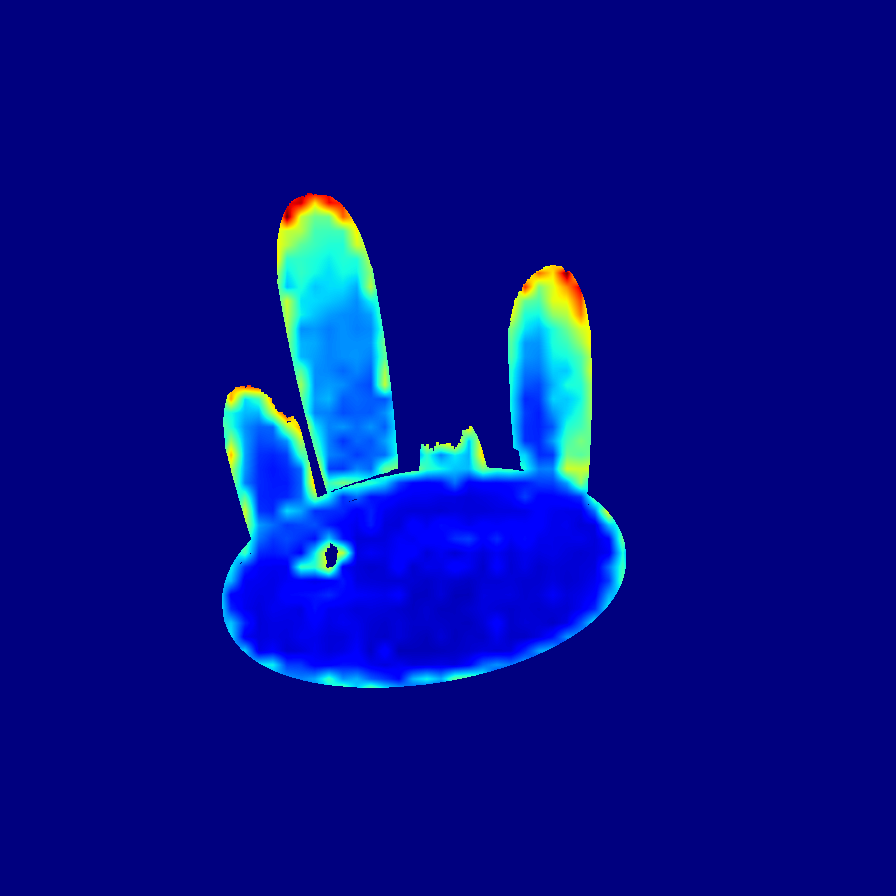} & 
            \includegraphics[width=0.08\linewidth,angle=180,origin=c]{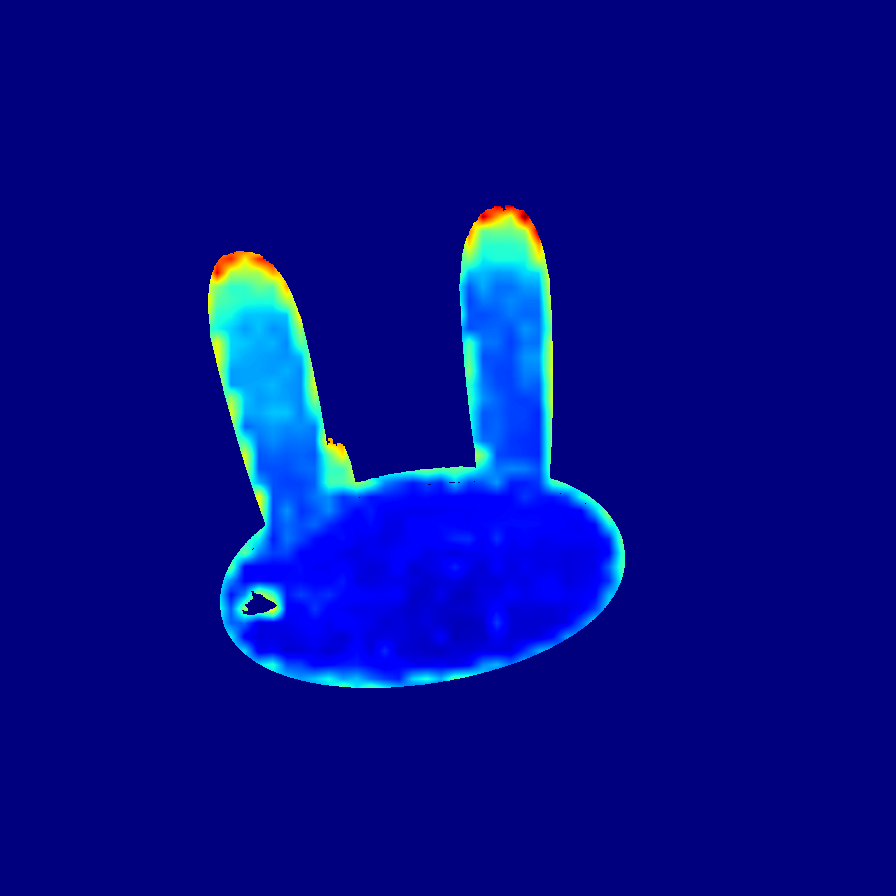} & 
            \includegraphics[width=0.08\linewidth,angle=180,origin=c]{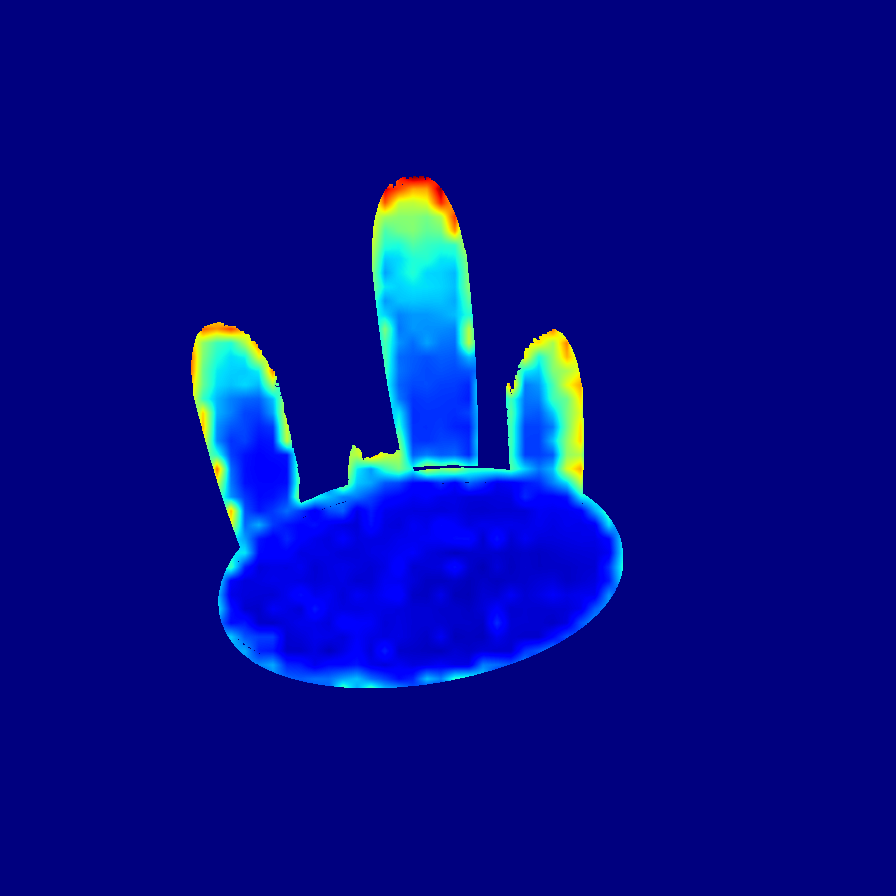} & 
            \includegraphics[width=0.08\linewidth,angle=180,origin=c]{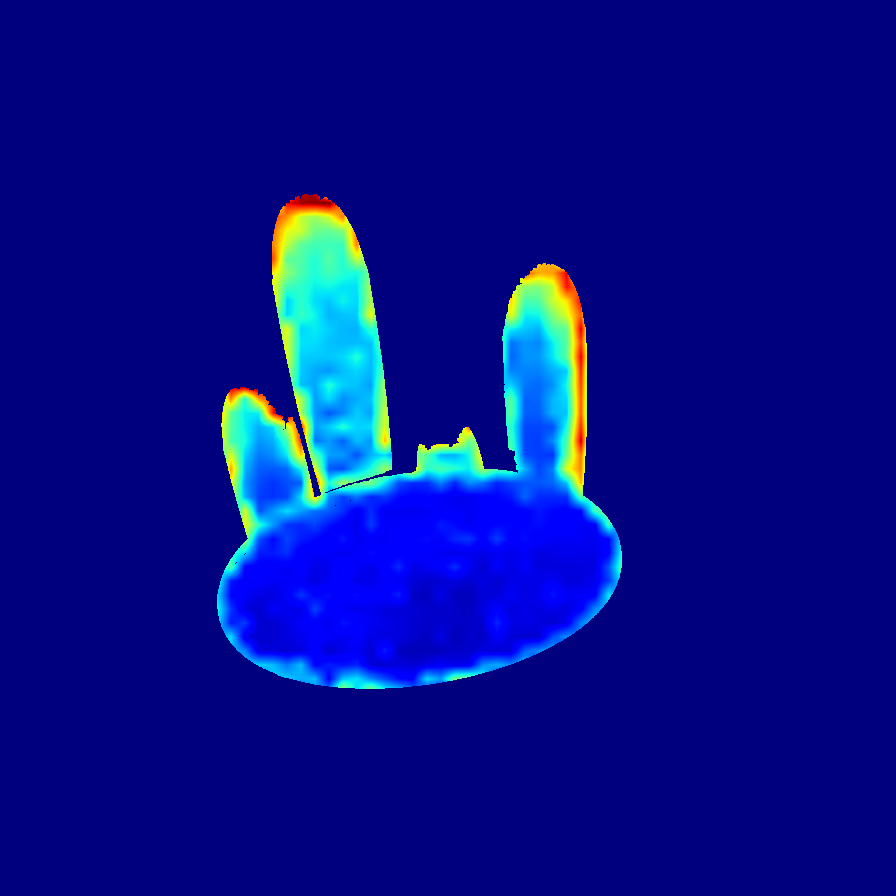} & 
            \includegraphics[width=0.08\linewidth,angle=180,origin=c]{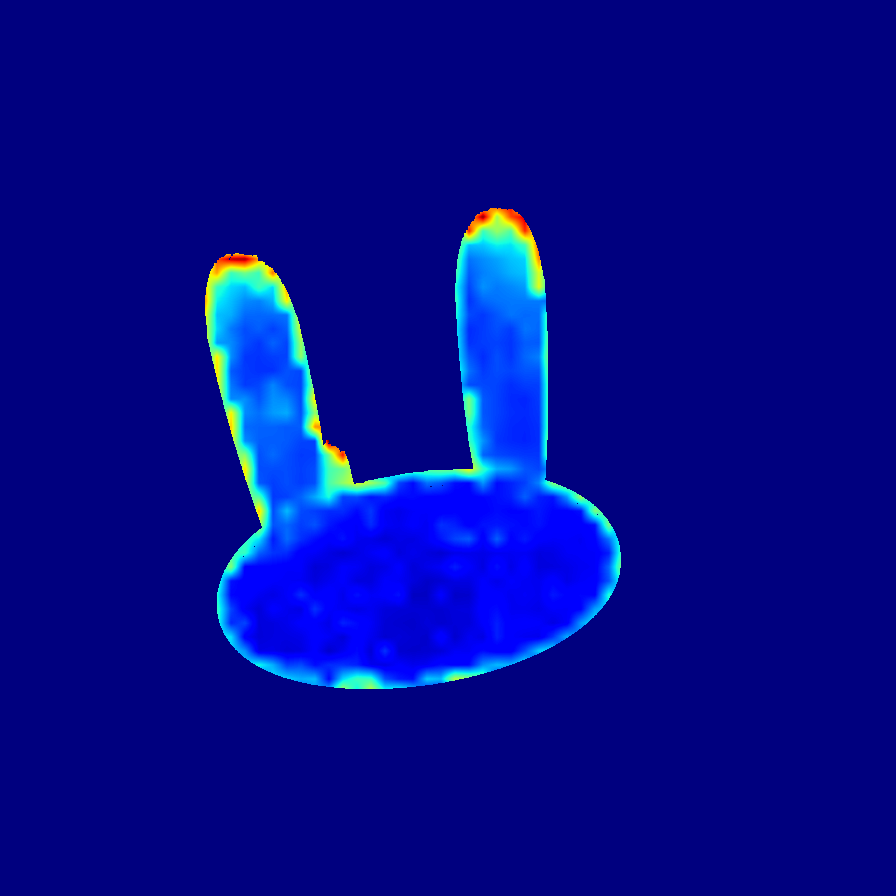} & 
            \includegraphics[width=0.08\linewidth,angle=180,origin=c]{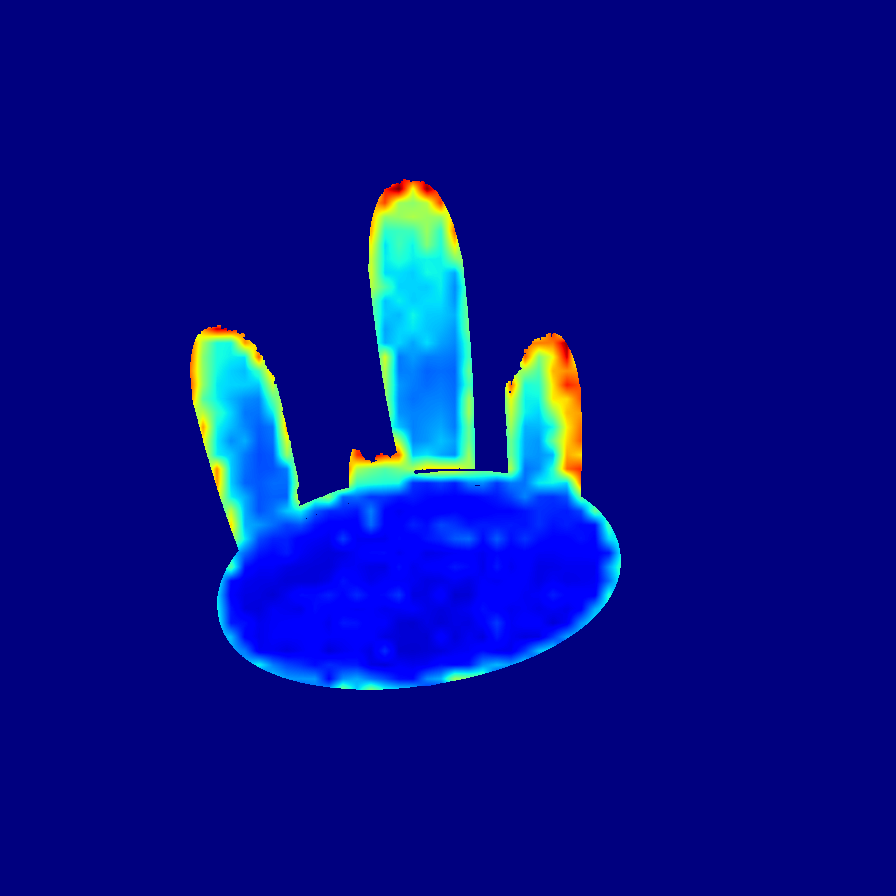} & 
            \includegraphics[width=0.08\linewidth,angle=180,origin=c]{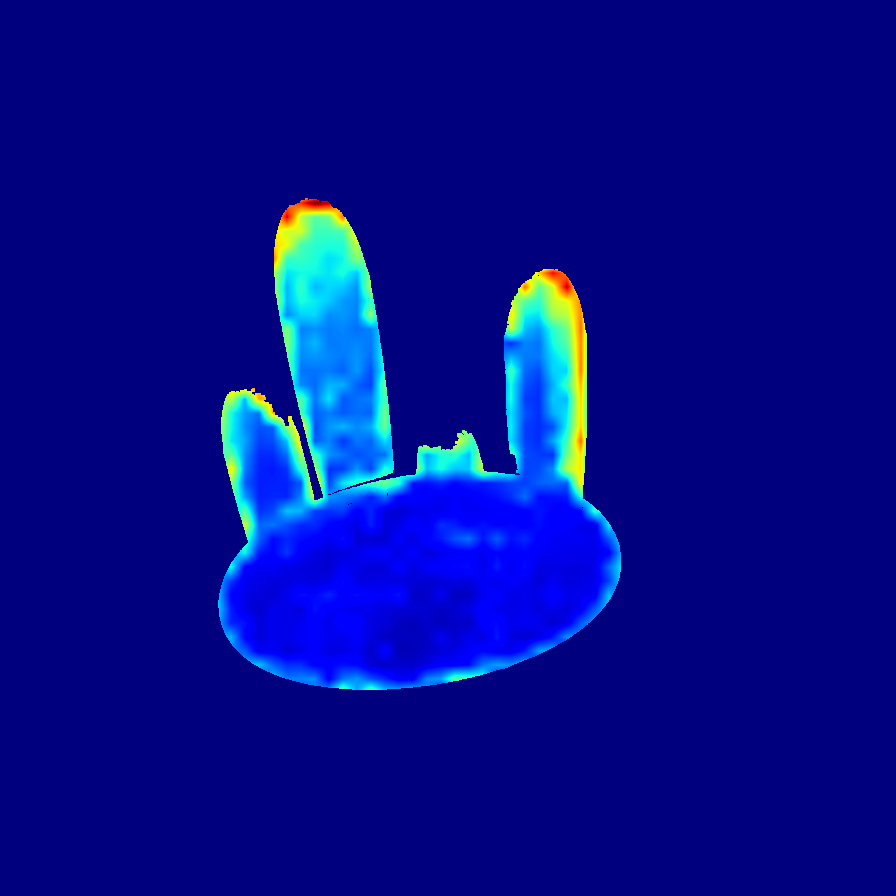} & 
            \includegraphics[width=0.08\linewidth,angle=180,origin=c]{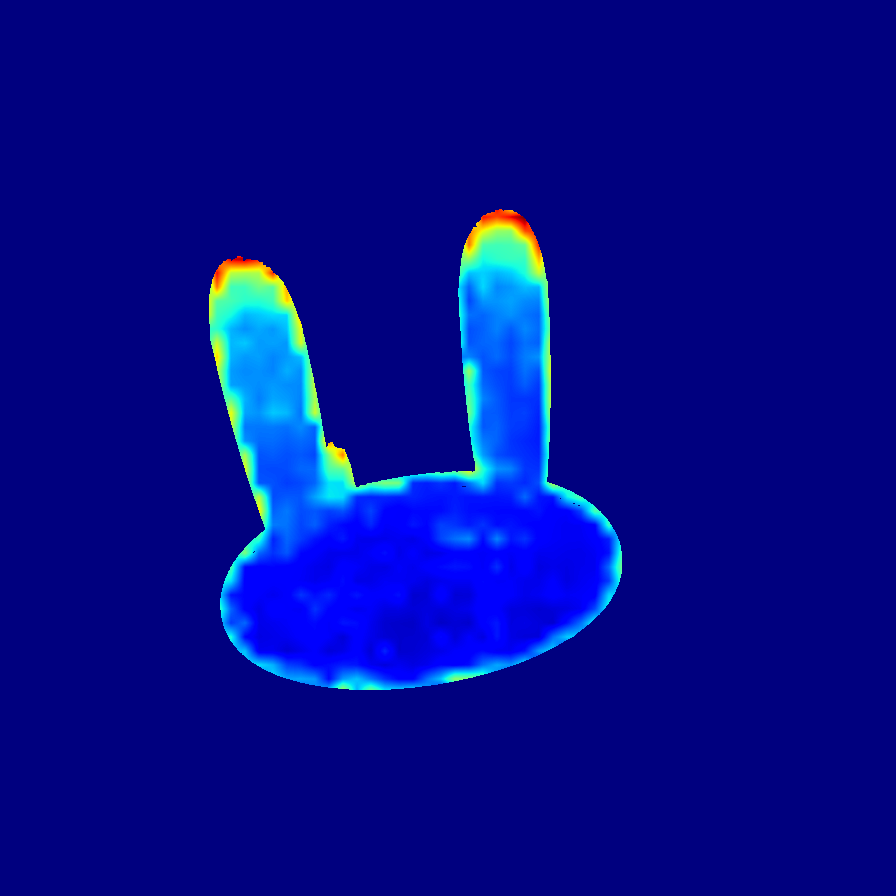} &
            \includegraphics[width=0.08\linewidth,angle=180,origin=c]{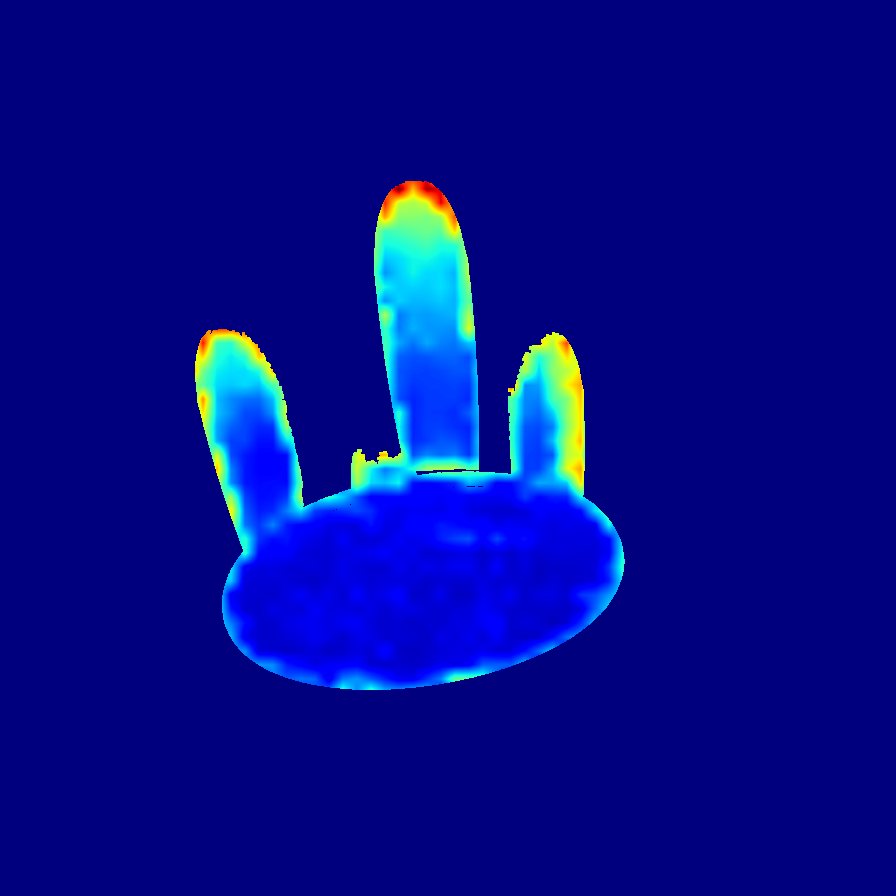} \\

            & \rotatebox{90}{\quad $v_{9}$} &
            \includegraphics[width=0.08\linewidth,angle=180,origin=c]{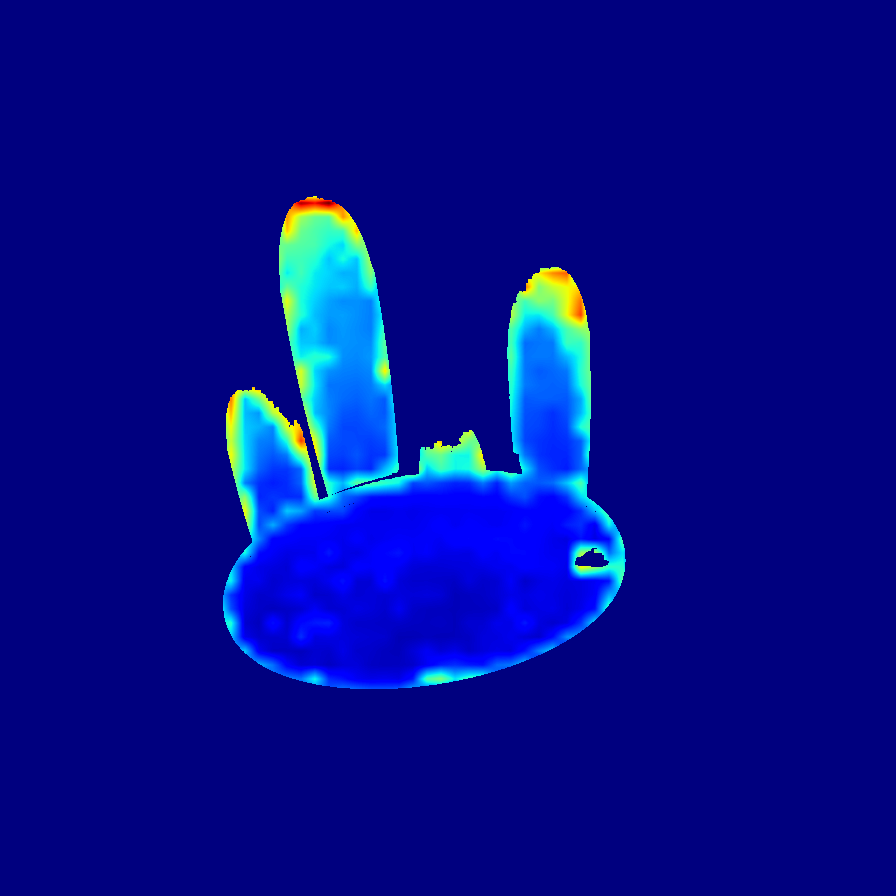} & 
            \includegraphics[width=0.08\linewidth,angle=180,origin=c]{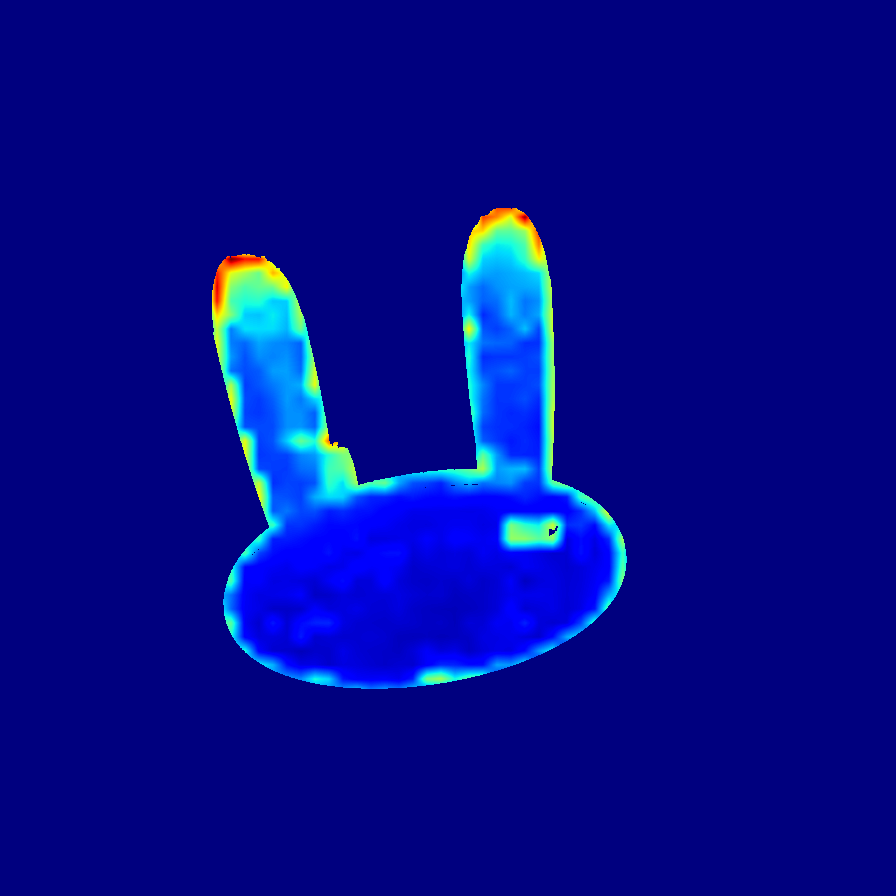} & 
            \includegraphics[width=0.08\linewidth,angle=180,origin=c]{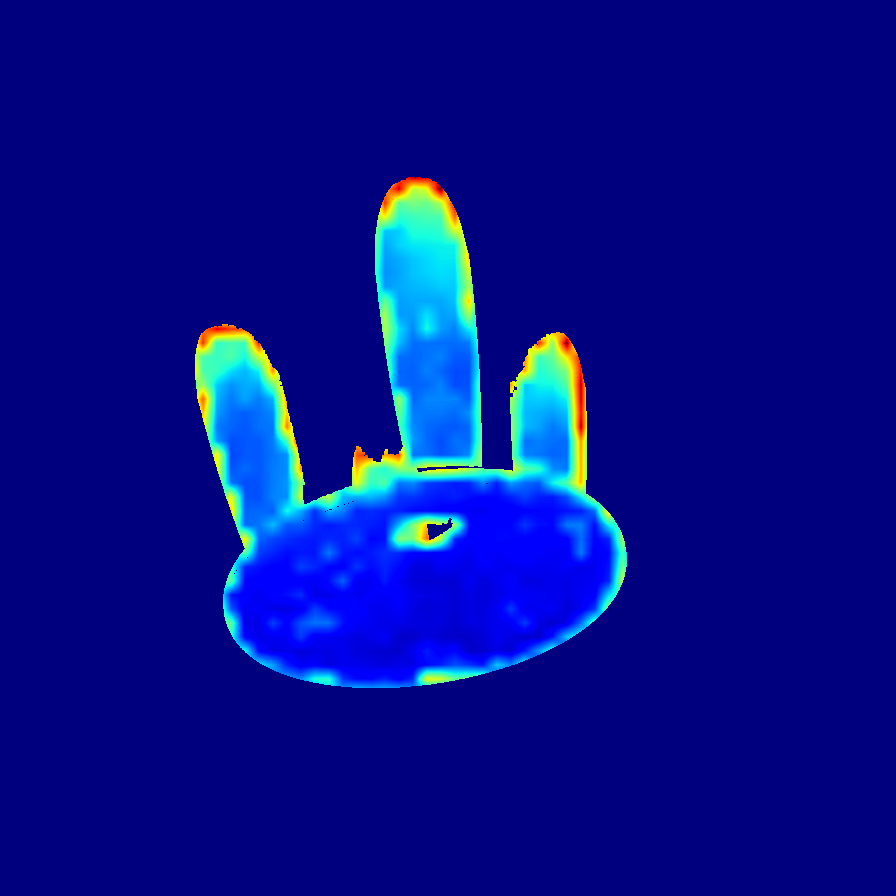} & 
            \includegraphics[width=0.08\linewidth,angle=180,origin=c]{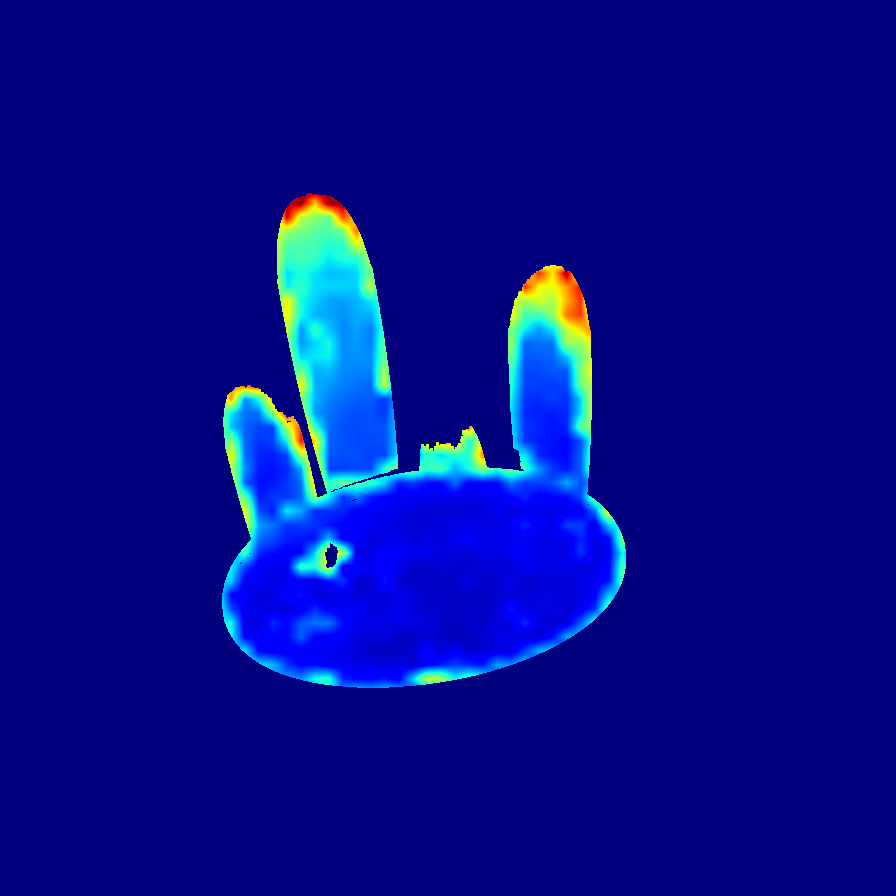} & 
            \includegraphics[width=0.08\linewidth,angle=180,origin=c]{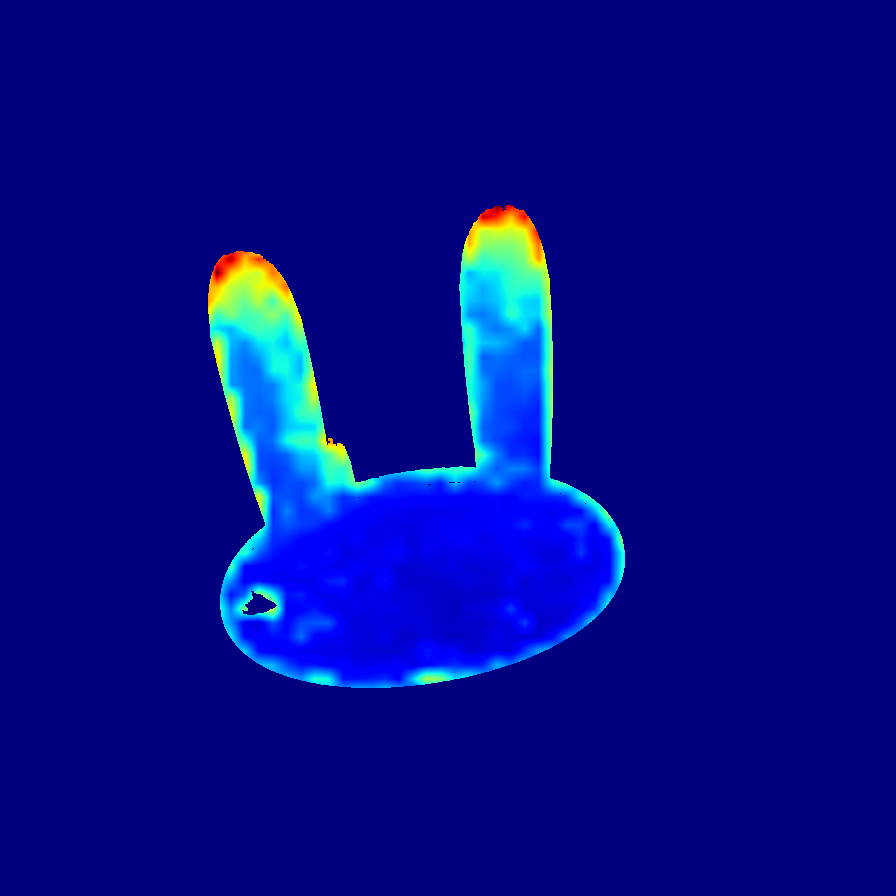} & 
            \includegraphics[width=0.08\linewidth,angle=180,origin=c]{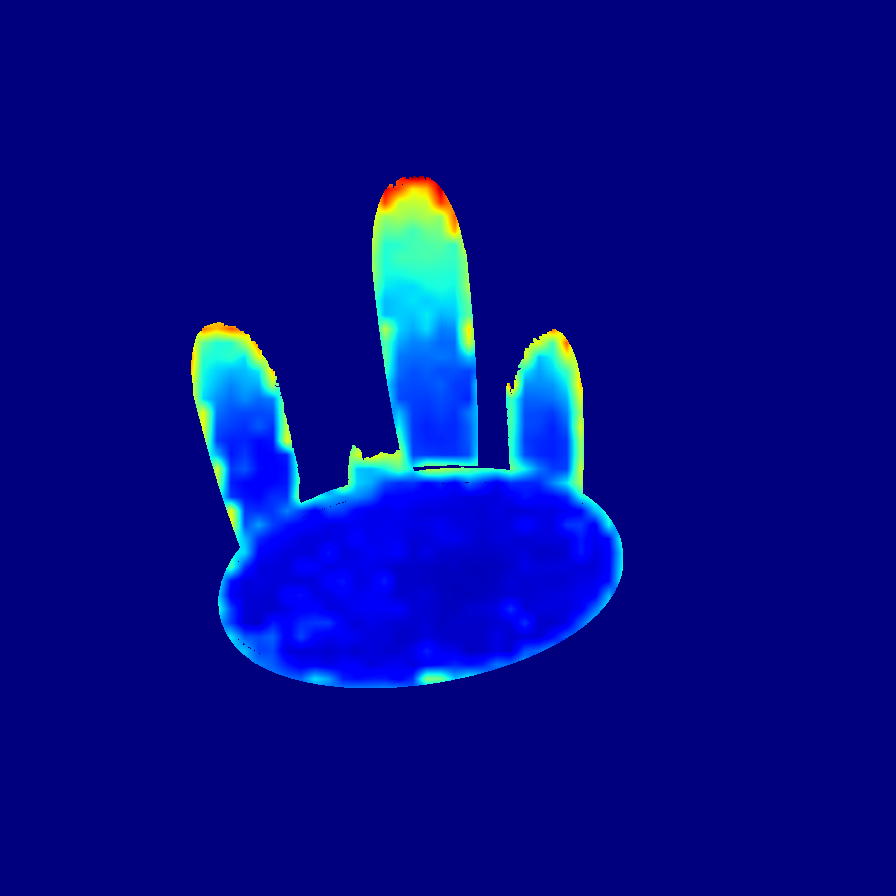} & 
            \includegraphics[width=0.08\linewidth,angle=180,origin=c]{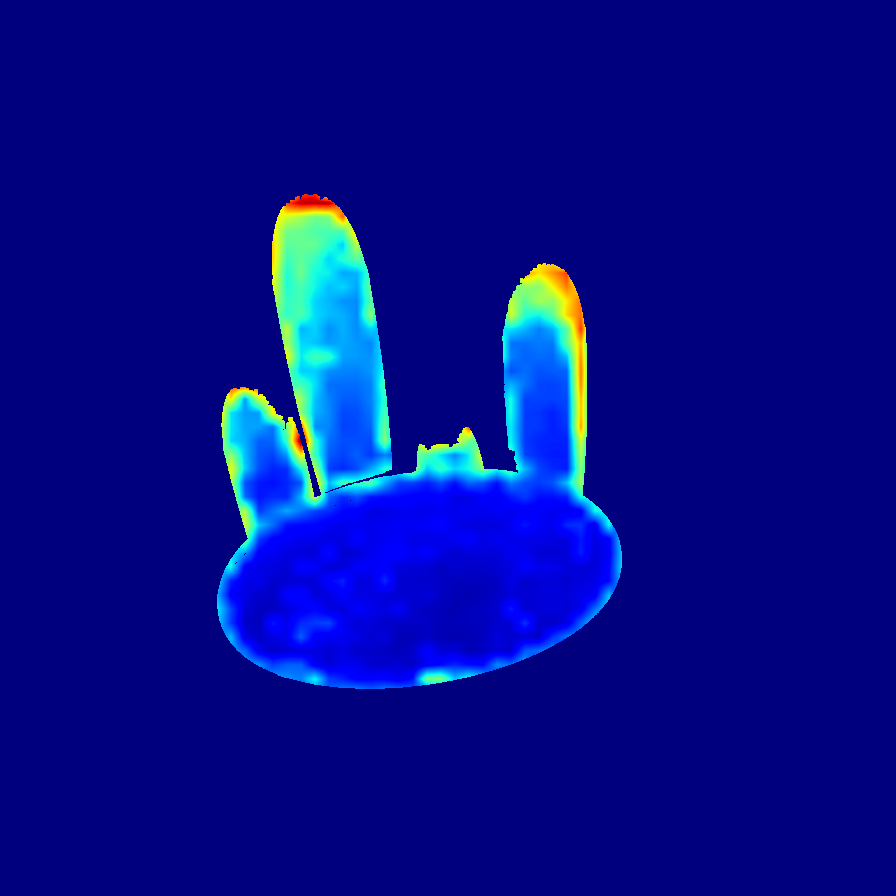} & 
            \includegraphics[width=0.08\linewidth,angle=180,origin=c]{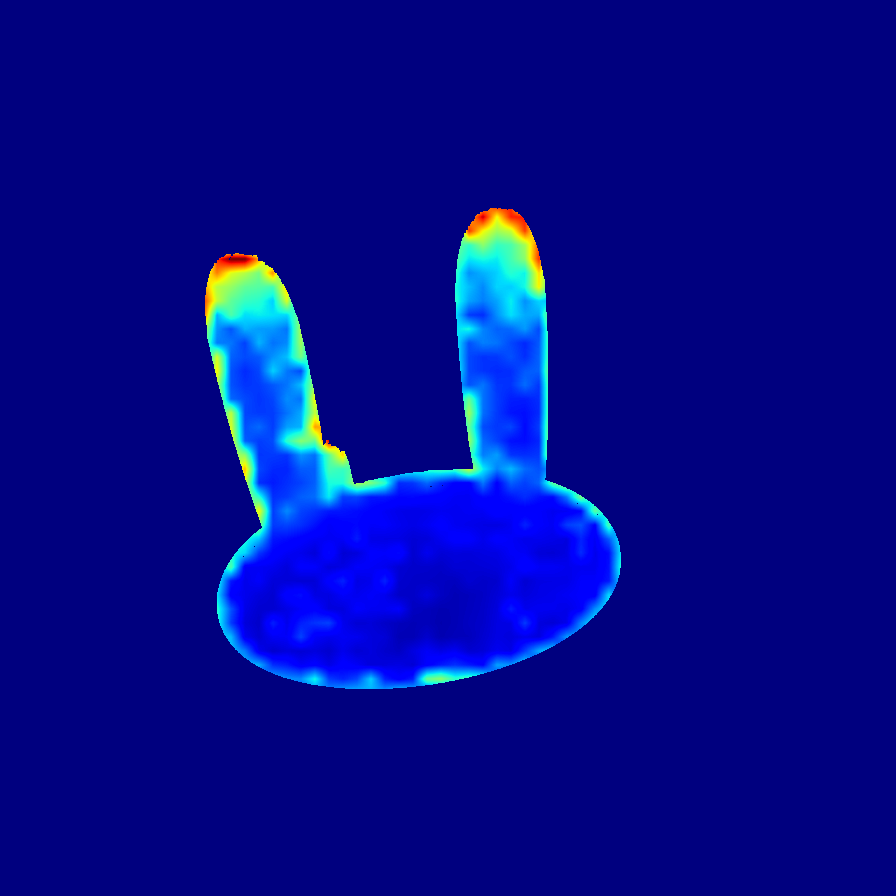} & 
            \includegraphics[width=0.08\linewidth,angle=180,origin=c]{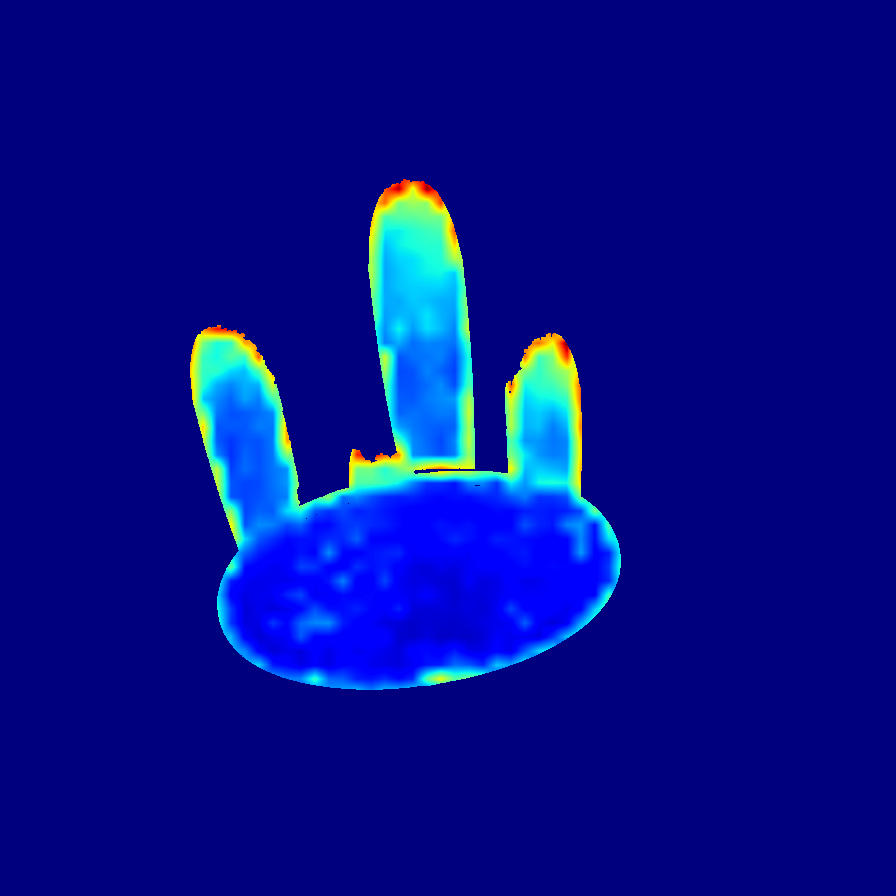} & 
            \includegraphics[width=0.08\linewidth,angle=180,origin=c]{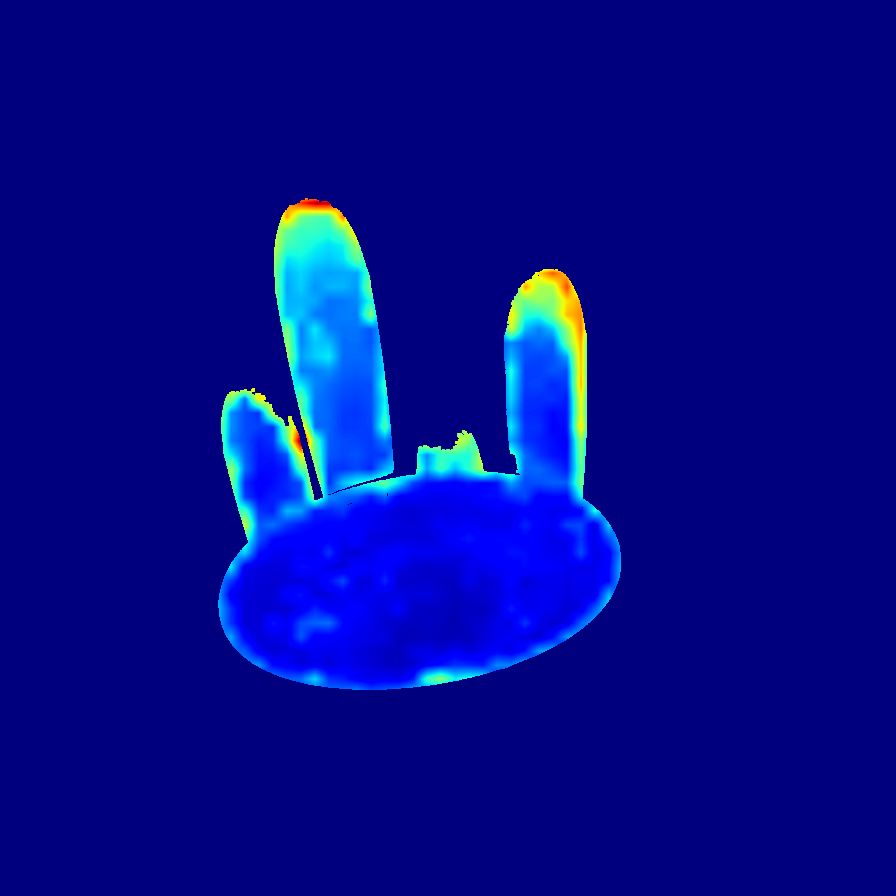} & 
            \includegraphics[width=0.08\linewidth,angle=180,origin=c]{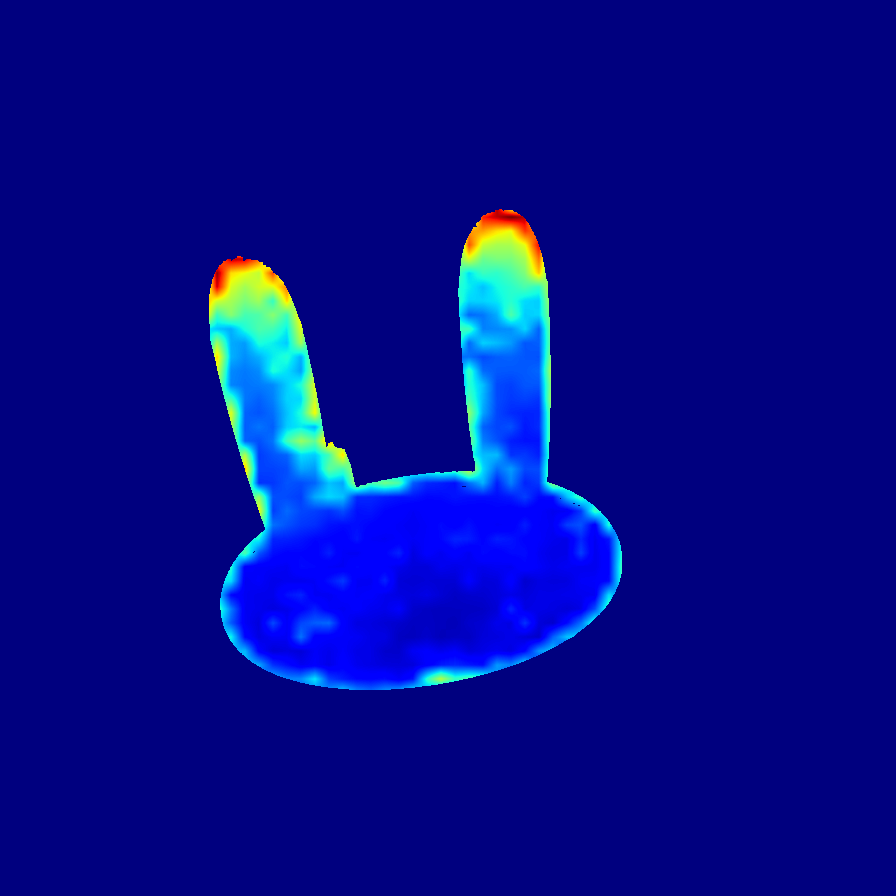} &
            \includegraphics[width=0.08\linewidth,angle=180,origin=c]{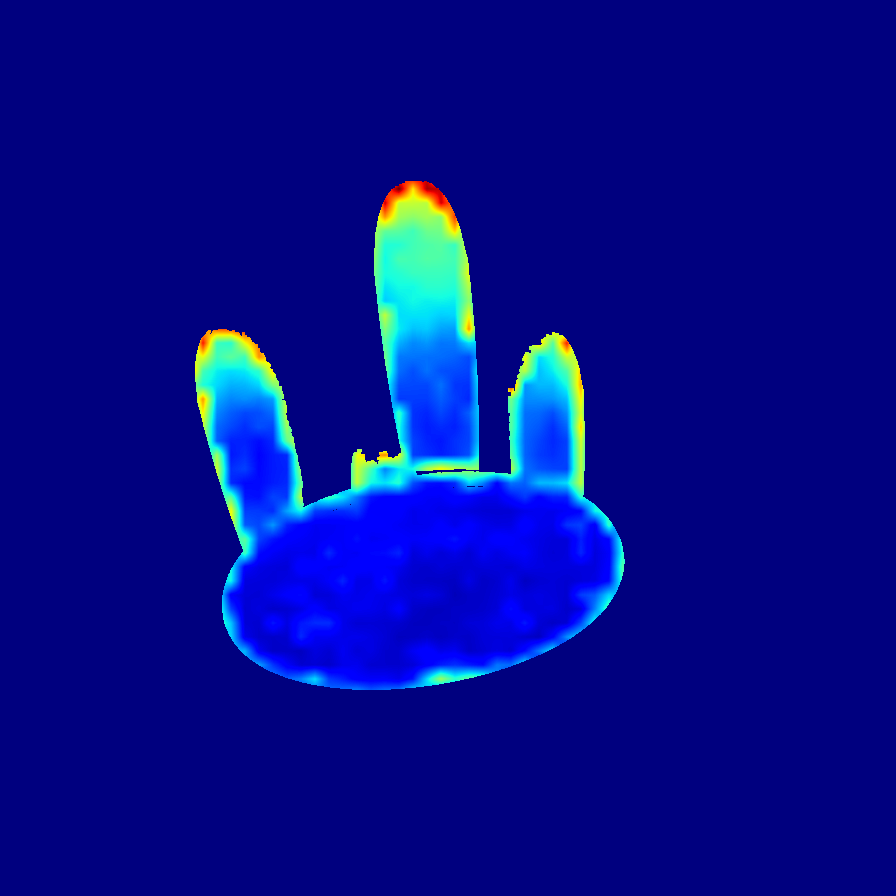} \\

            & \rotatebox{90}{\quad $v_{10}$} &
            \includegraphics[width=0.08\linewidth,angle=180,origin=c]{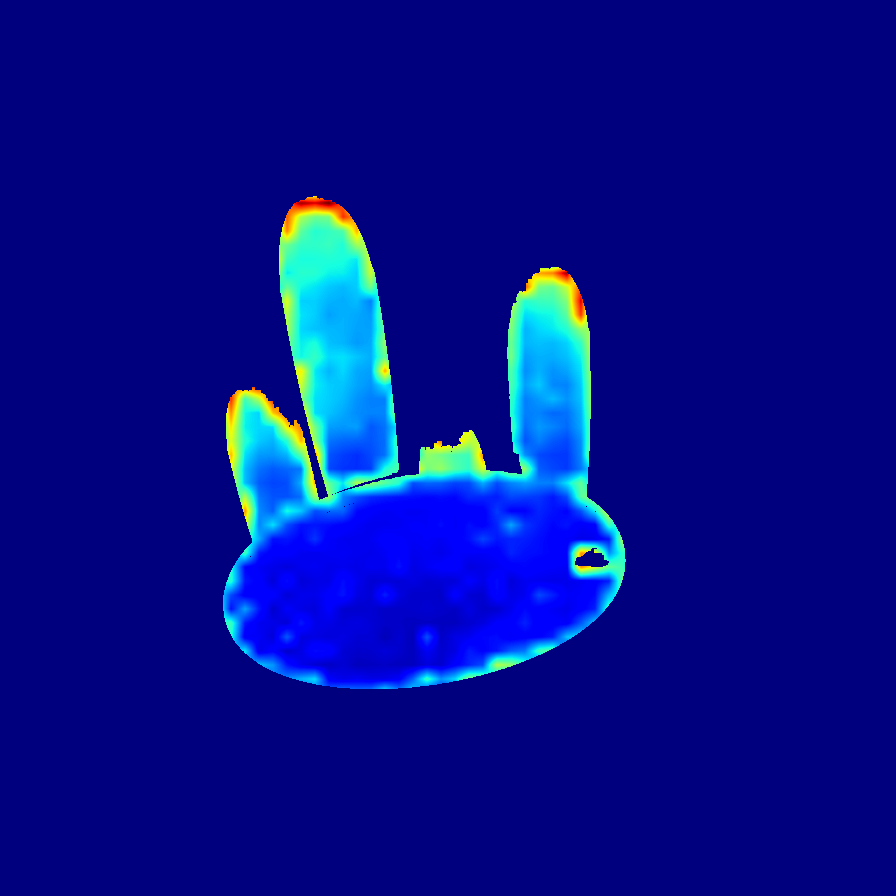} & 
            \includegraphics[width=0.08\linewidth,angle=180,origin=c]{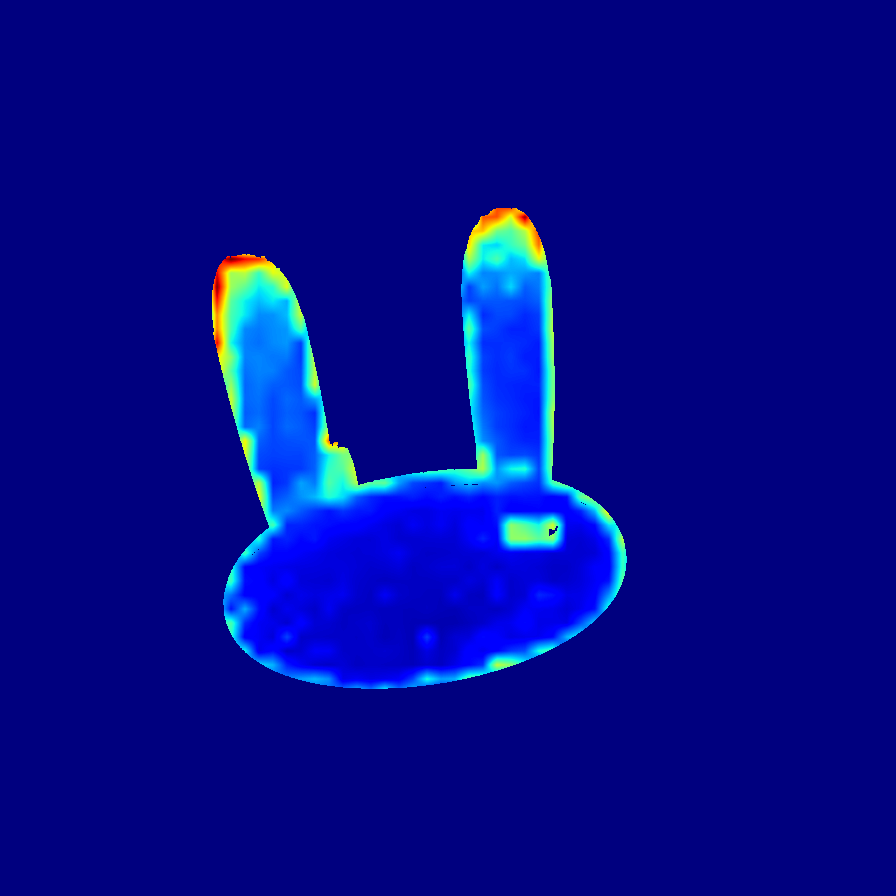} & 
            \includegraphics[width=0.08\linewidth,angle=180,origin=c]{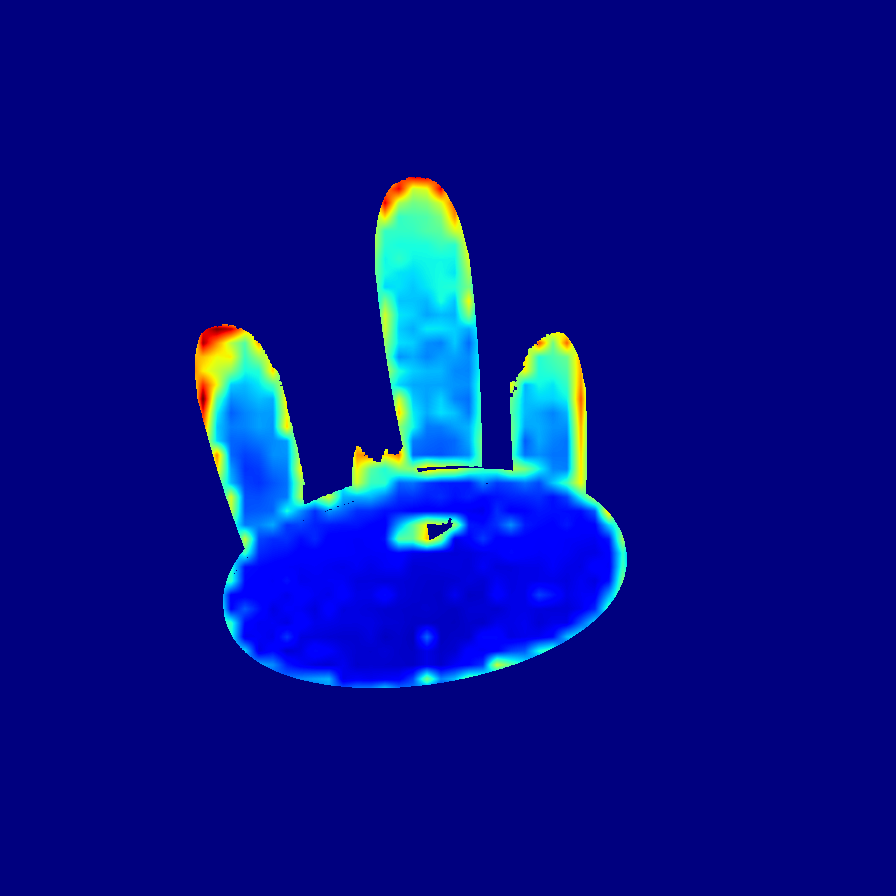} & 
            \includegraphics[width=0.08\linewidth,angle=180,origin=c]{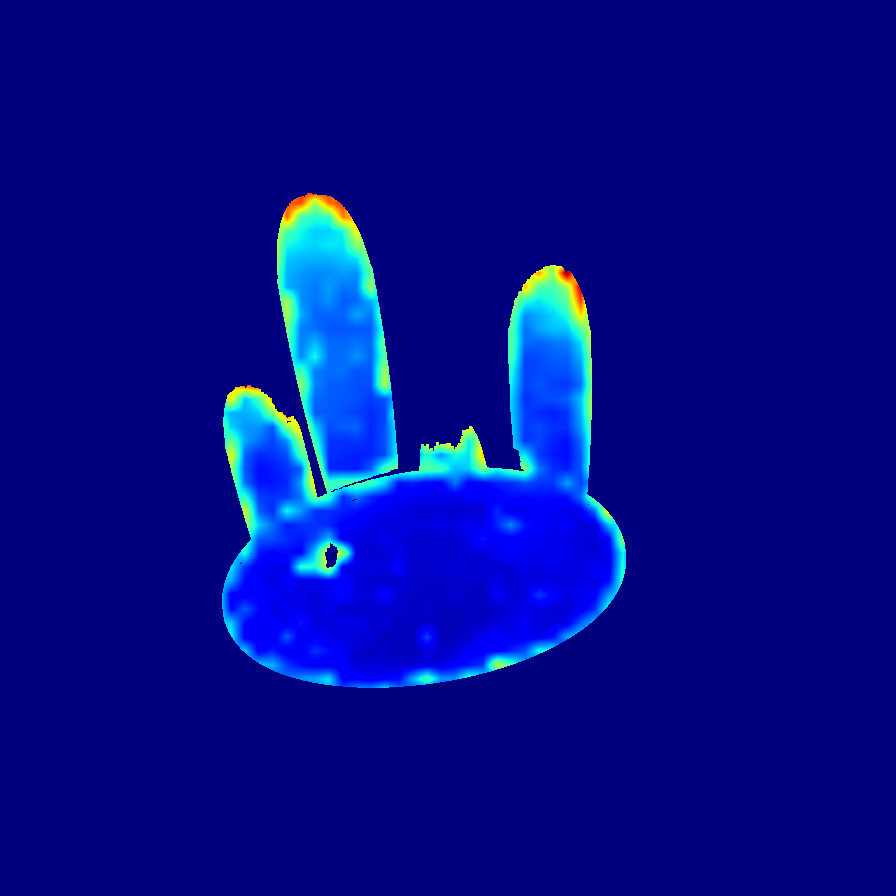} & 
            \includegraphics[width=0.08\linewidth,angle=180,origin=c]{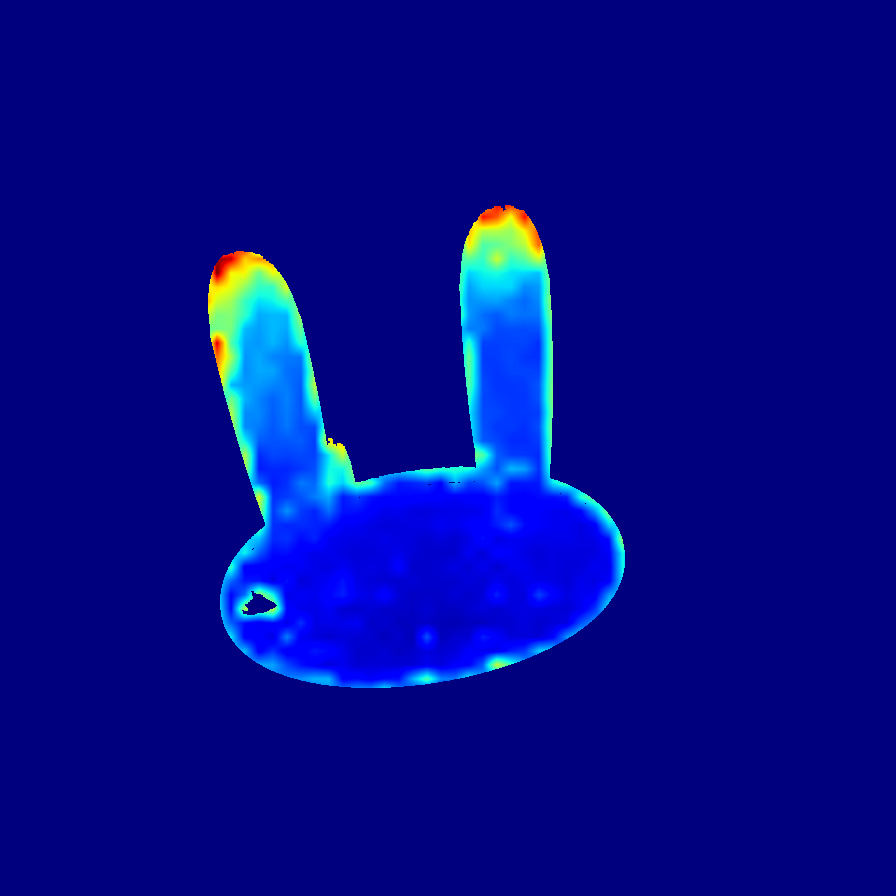} & 
            \includegraphics[width=0.08\linewidth,angle=180,origin=c]{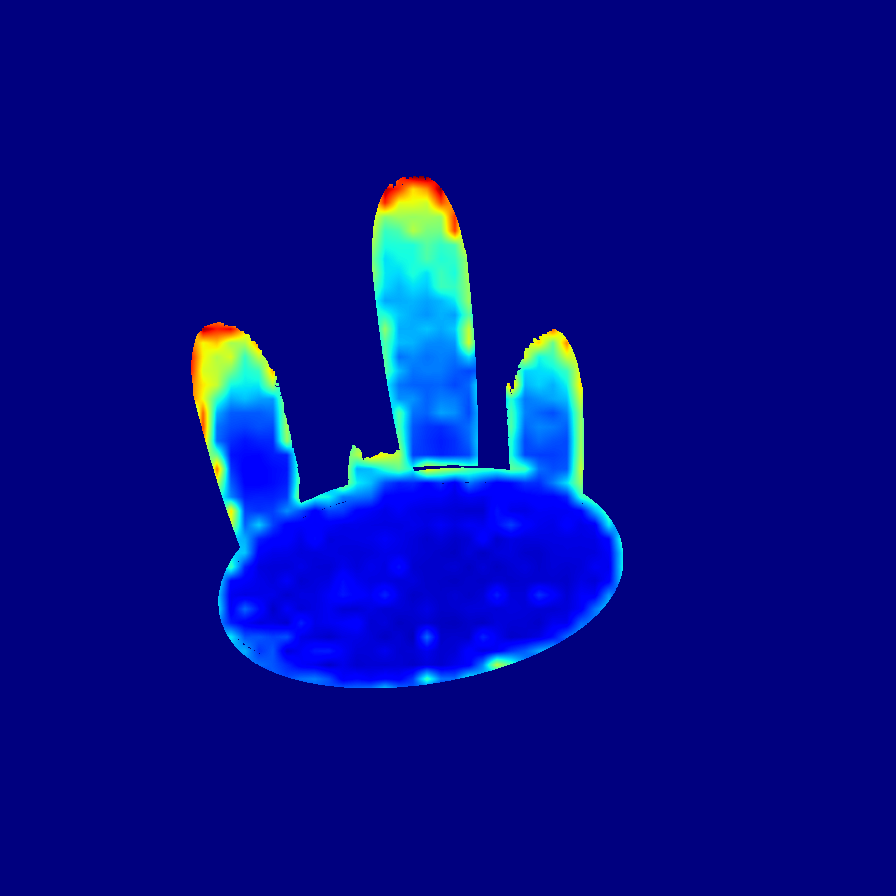} & 
            \includegraphics[width=0.08\linewidth,angle=180,origin=c]{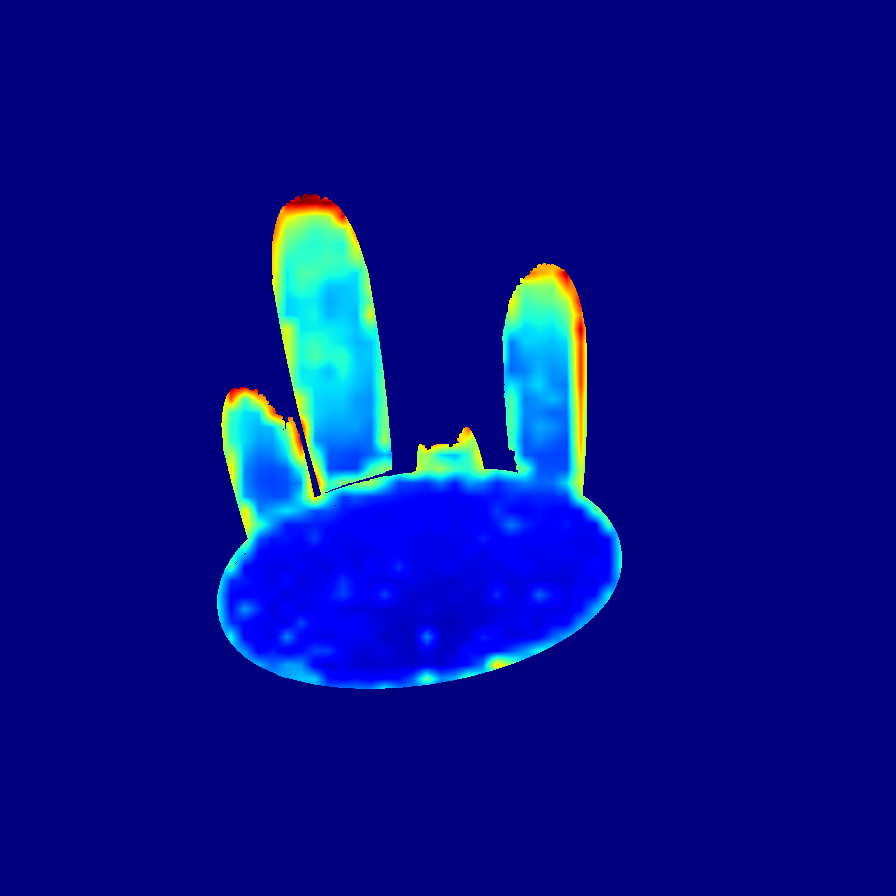} & 
            \includegraphics[width=0.08\linewidth,angle=180,origin=c]{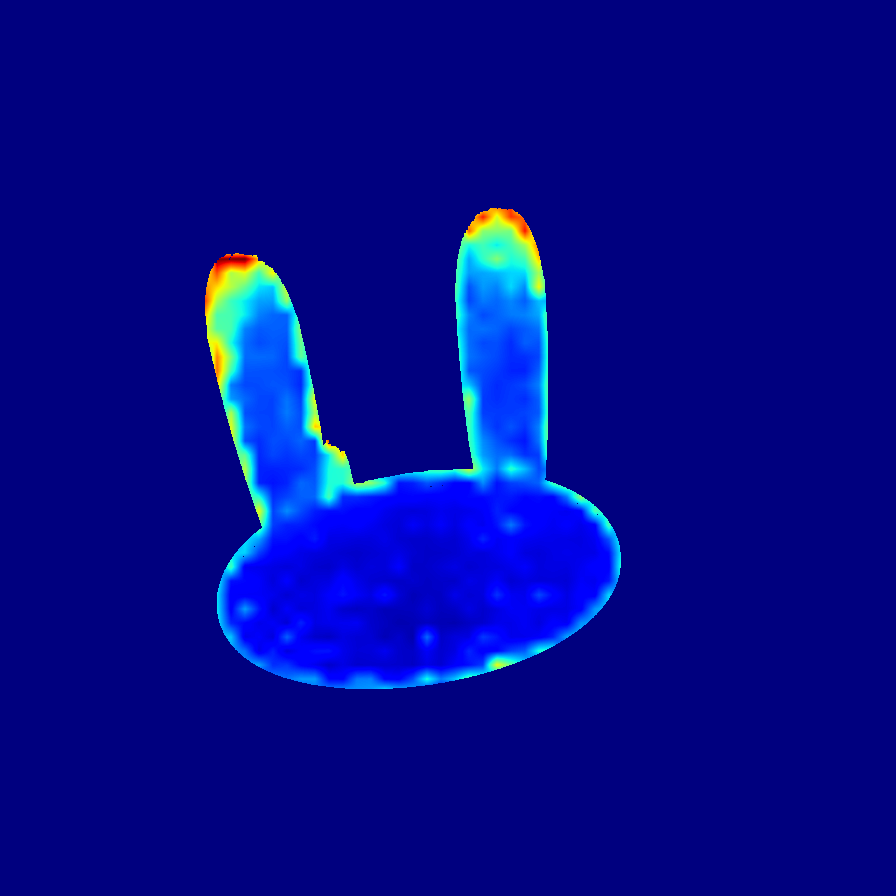} & 
            \includegraphics[width=0.08\linewidth,angle=180,origin=c]{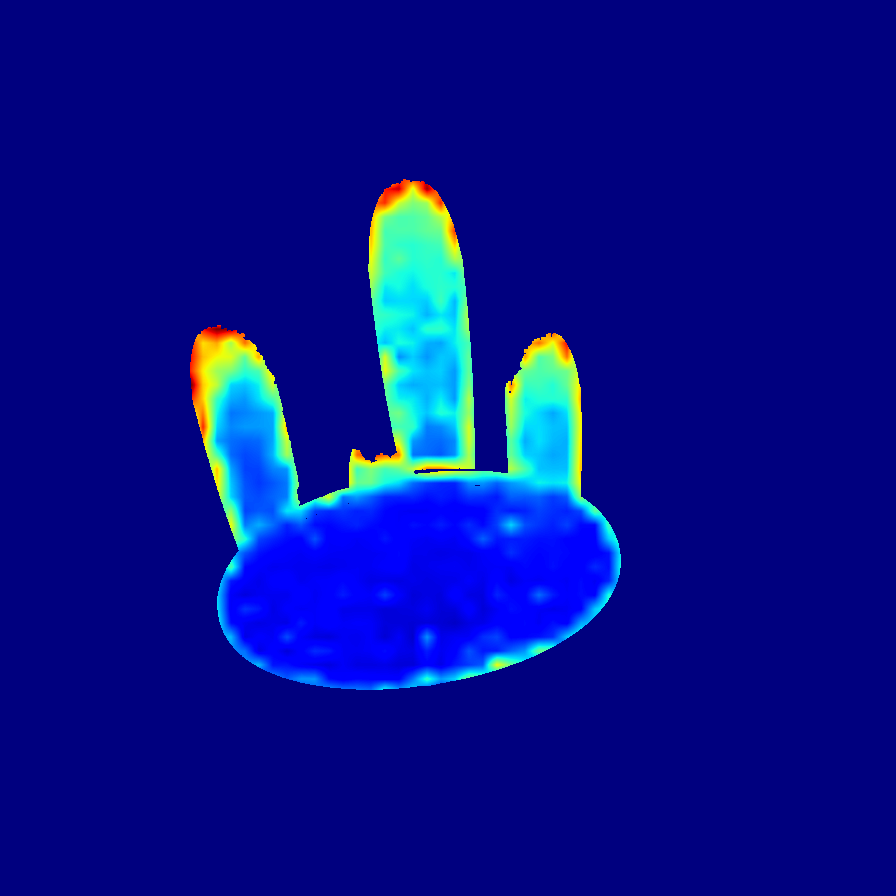} & 
            \includegraphics[width=0.08\linewidth,angle=180,origin=c]{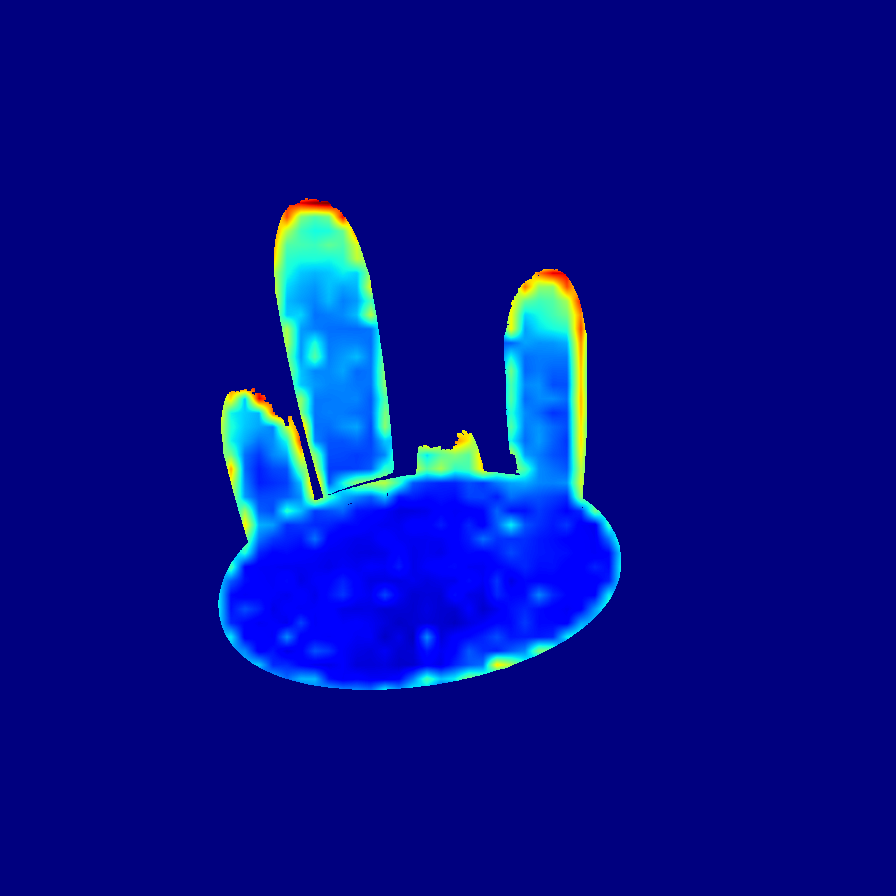} & 
            \includegraphics[width=0.08\linewidth,angle=180,origin=c]{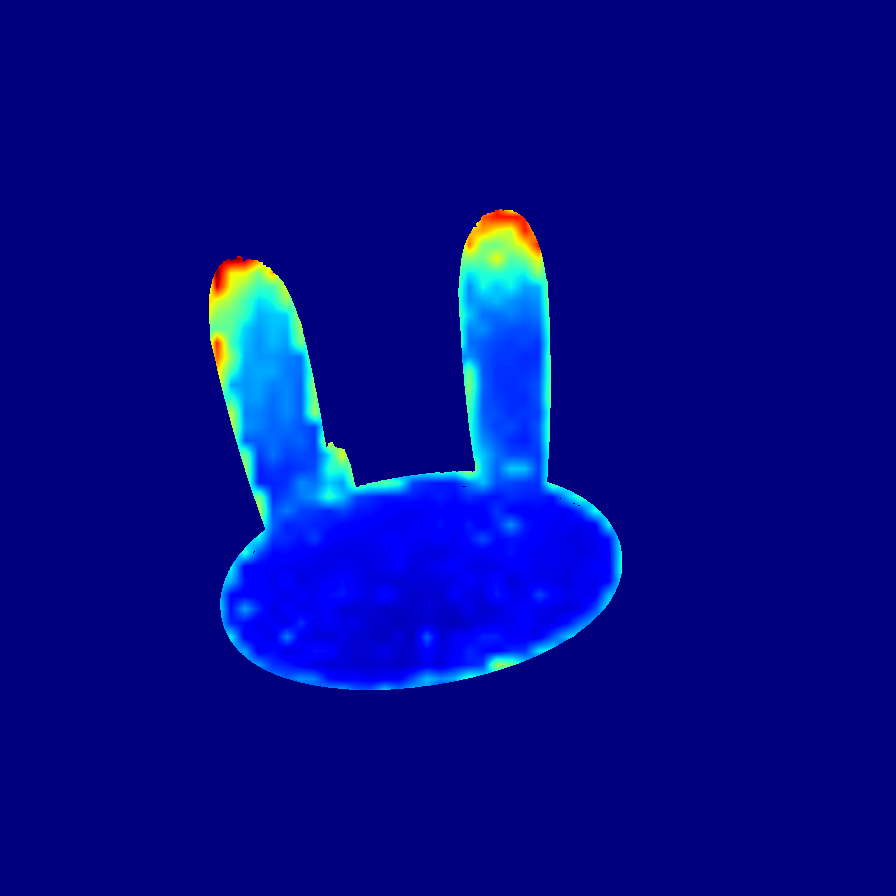} &
            \includegraphics[width=0.08\linewidth,angle=180,origin=c]{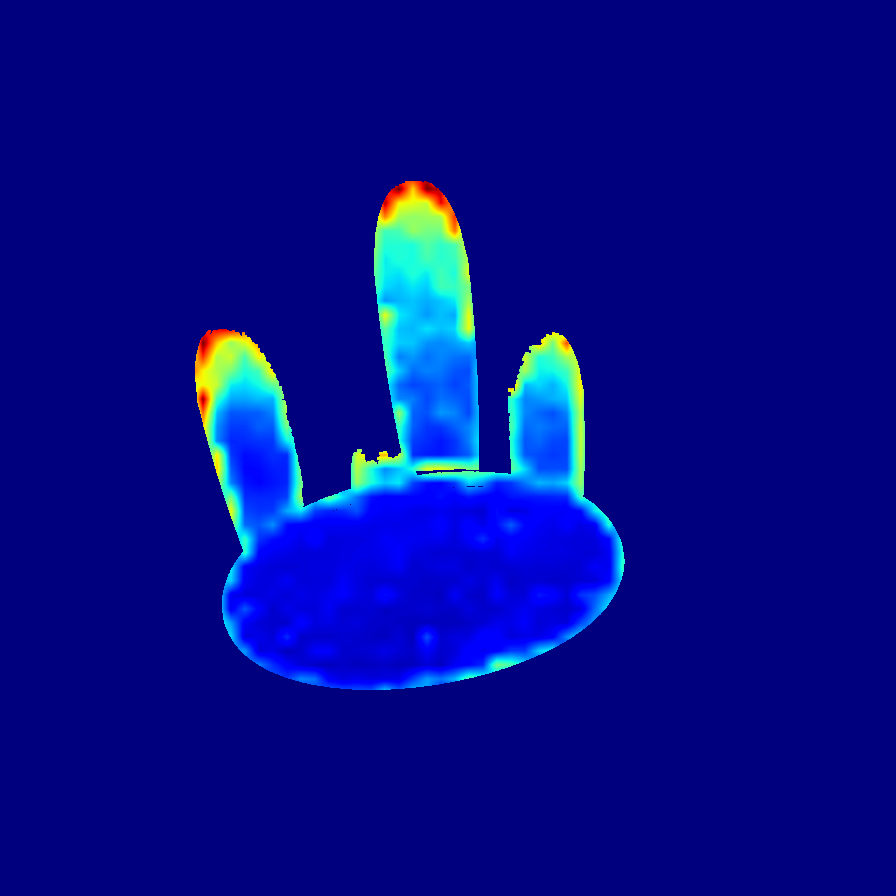} \\

            & \rotatebox{90}{\quad $v_{11}$} &
            \includegraphics[width=0.08\linewidth,angle=180,origin=c]{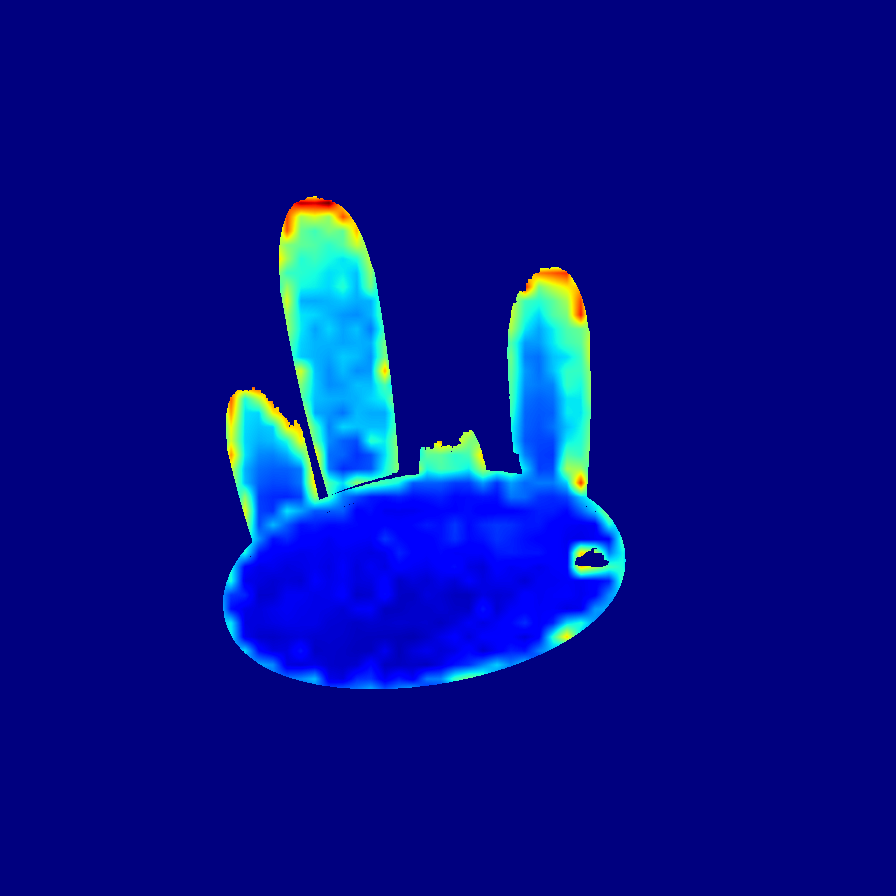} & 
            \includegraphics[width=0.08\linewidth,angle=180,origin=c]{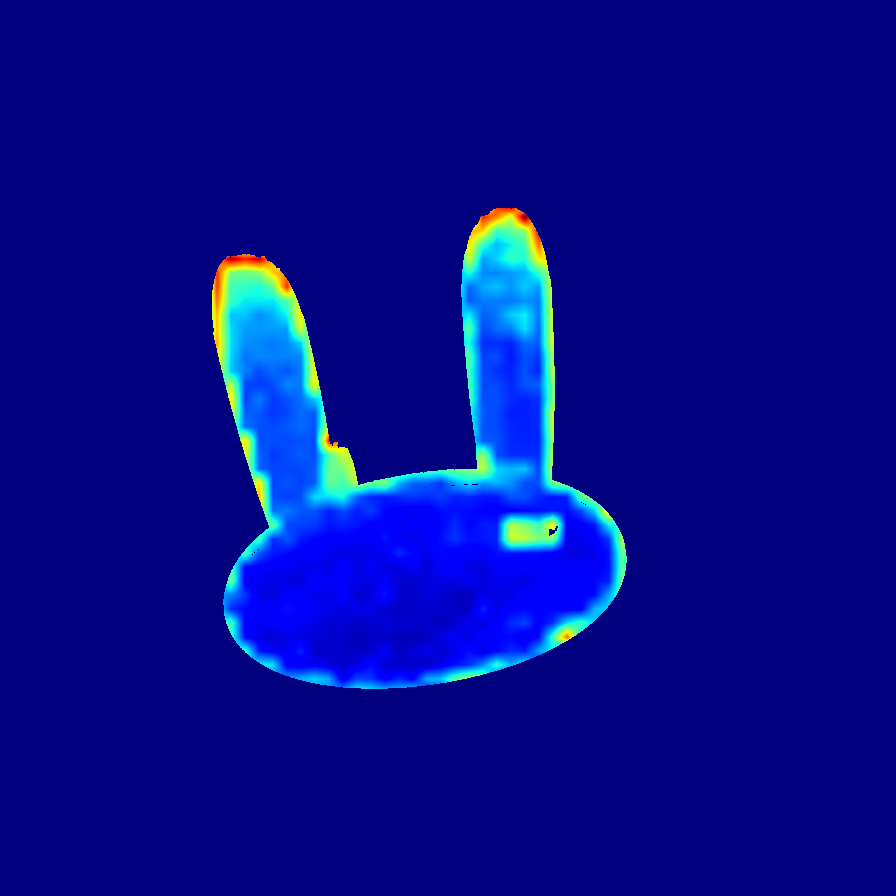} & 
            \includegraphics[width=0.08\linewidth,angle=180,origin=c]{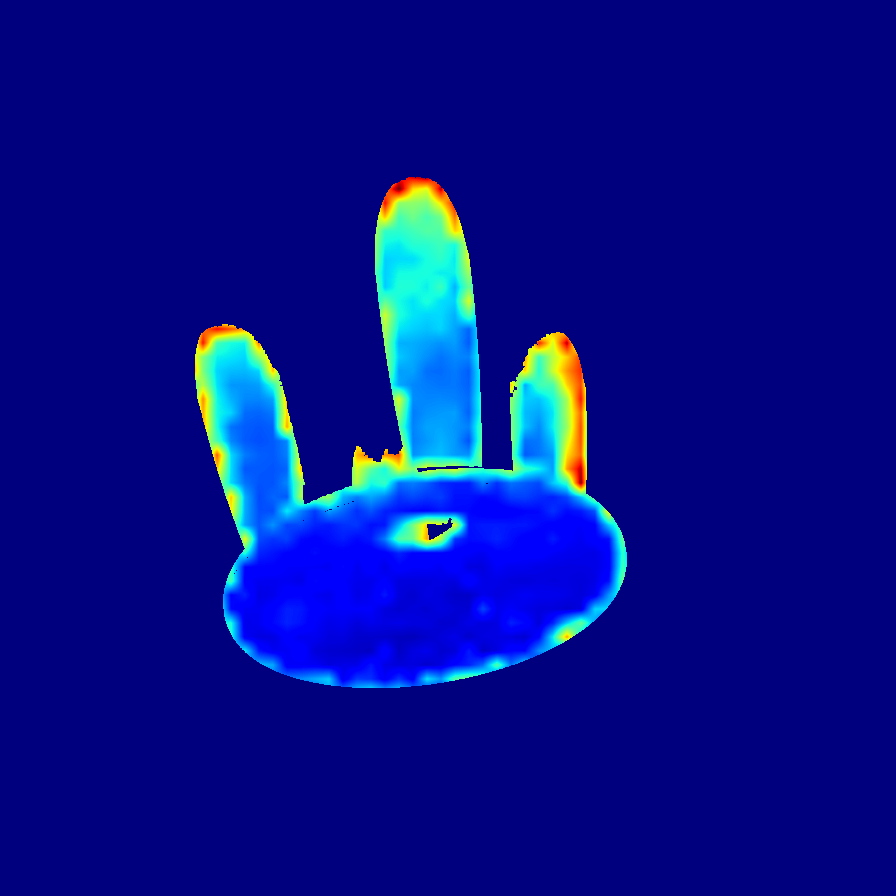} & 
            \includegraphics[width=0.08\linewidth,angle=180,origin=c]{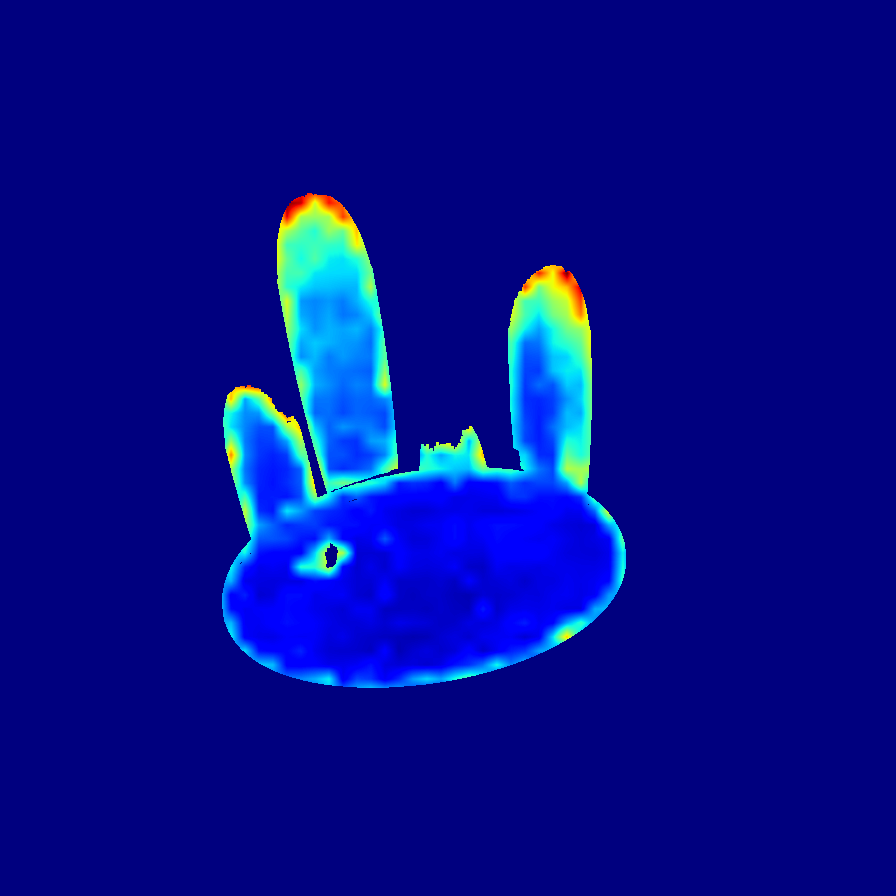} & 
            \includegraphics[width=0.08\linewidth,angle=180,origin=c]{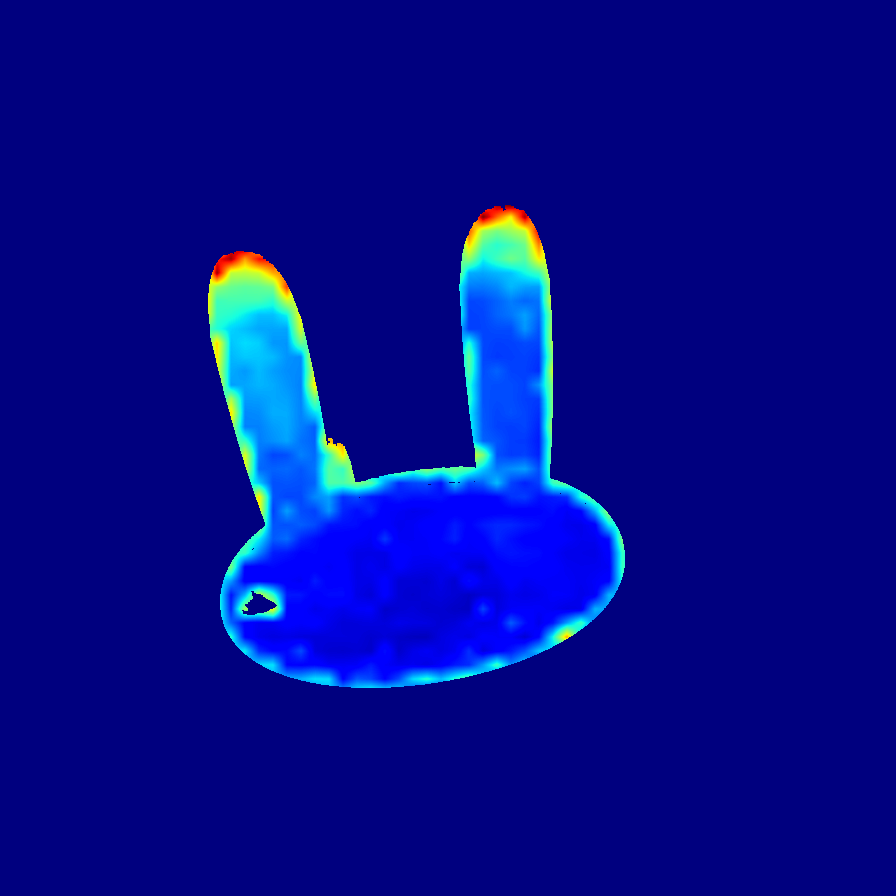} & 
            \includegraphics[width=0.08\linewidth,angle=180,origin=c]{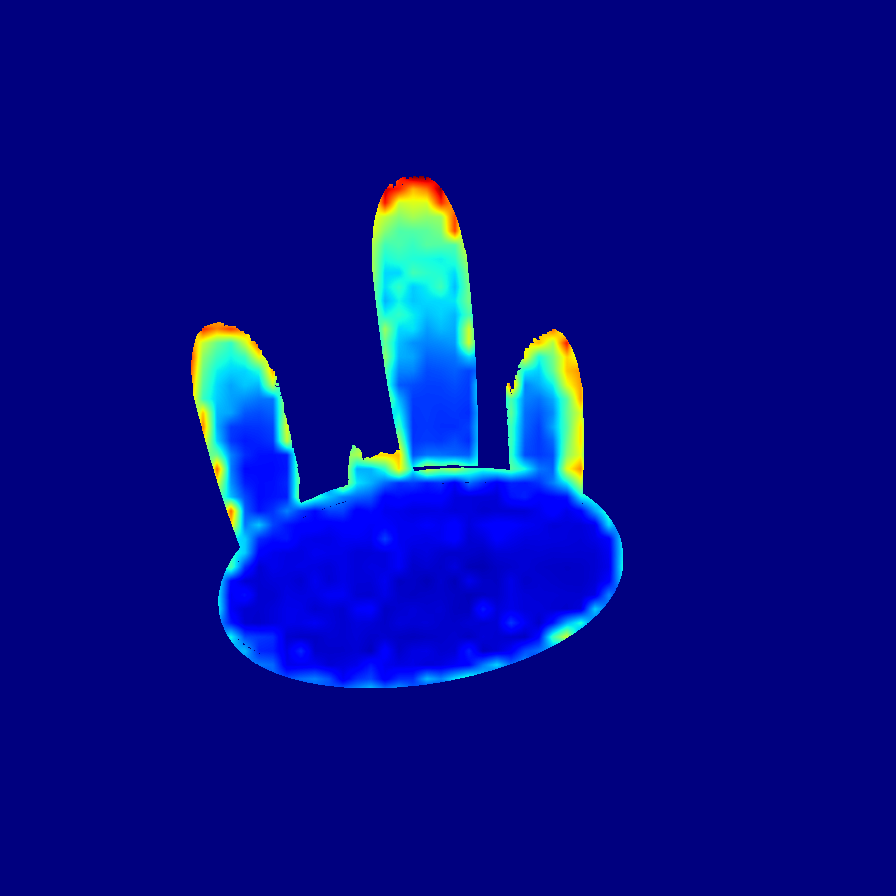} & 
            \includegraphics[width=0.08\linewidth,angle=180,origin=c]{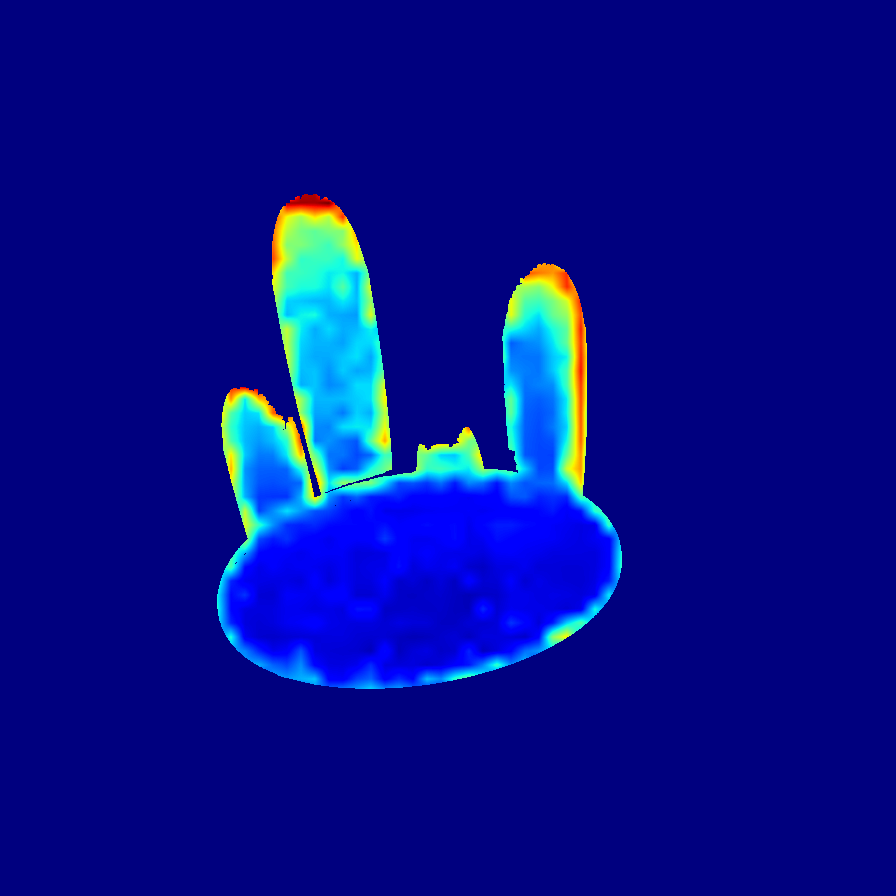} & 
            \includegraphics[width=0.08\linewidth,angle=180,origin=c]{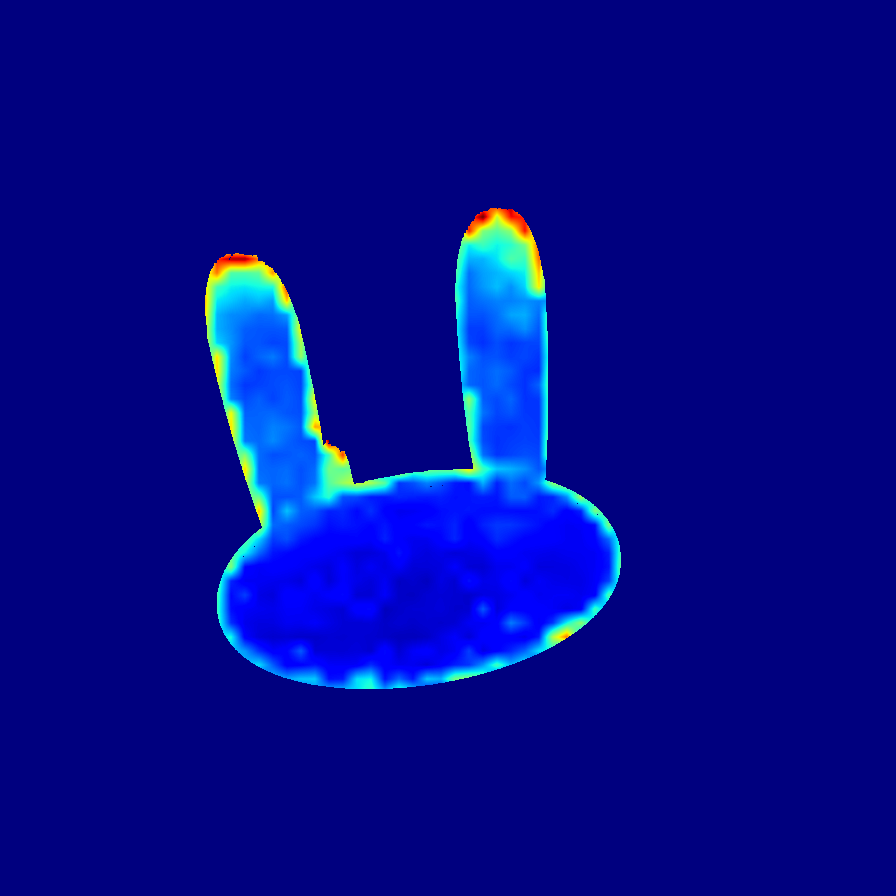} & 
            \includegraphics[width=0.08\linewidth,angle=180,origin=c]{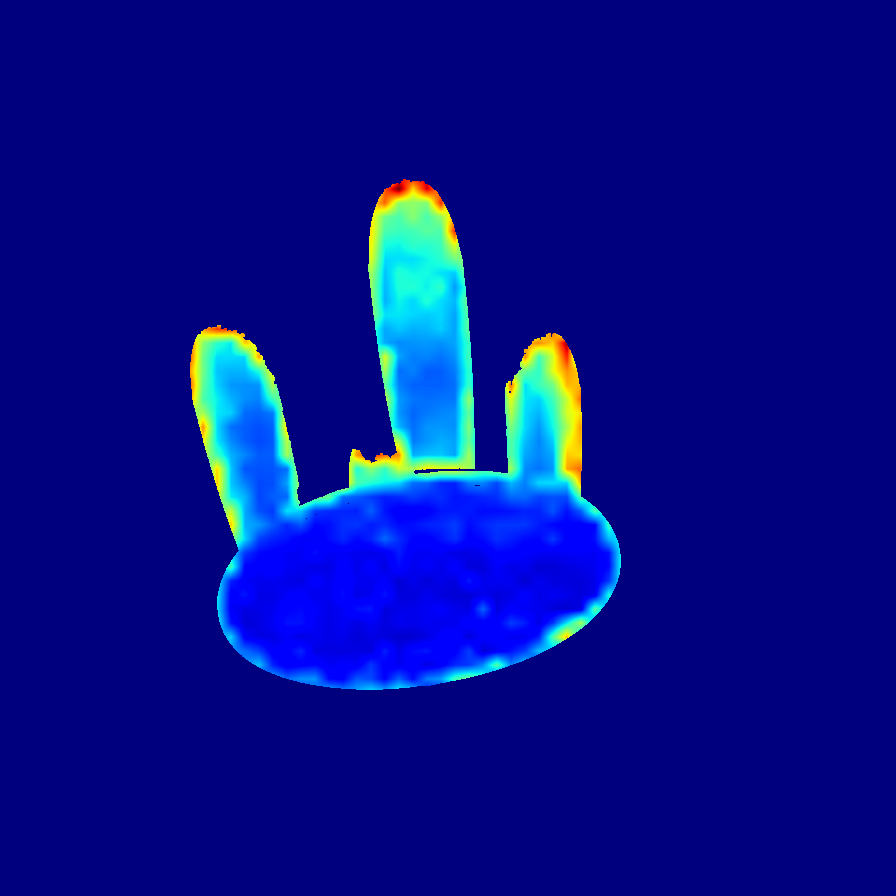} & 
            \includegraphics[width=0.08\linewidth,angle=180,origin=c]{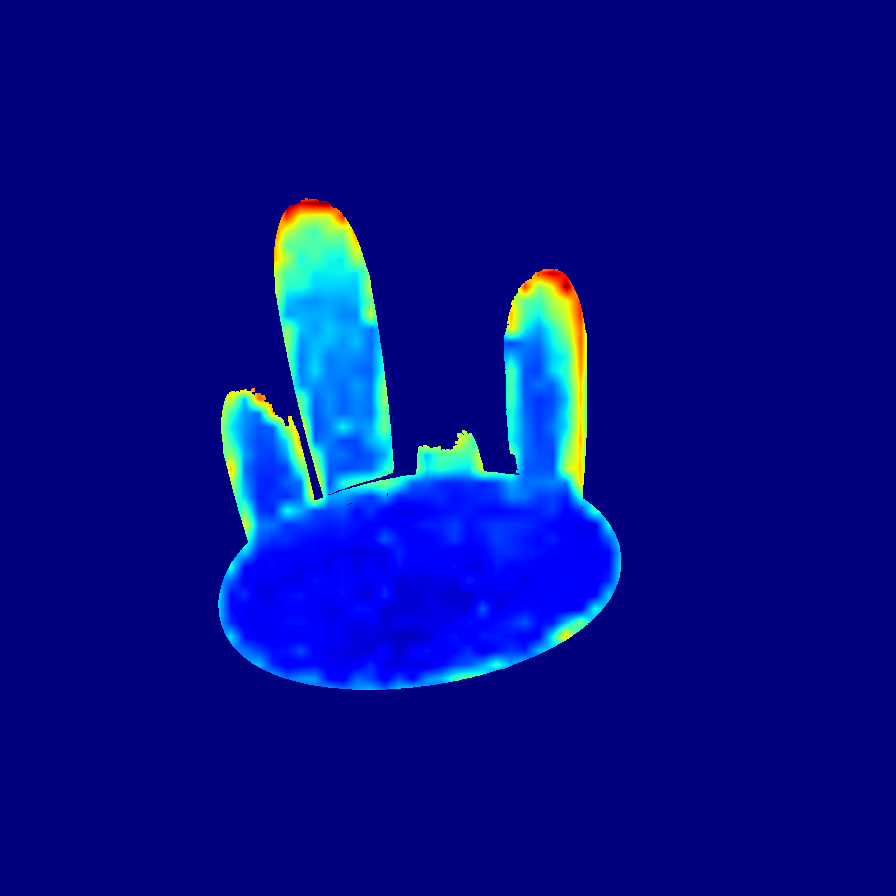} & 
            \includegraphics[width=0.08\linewidth,angle=180,origin=c]{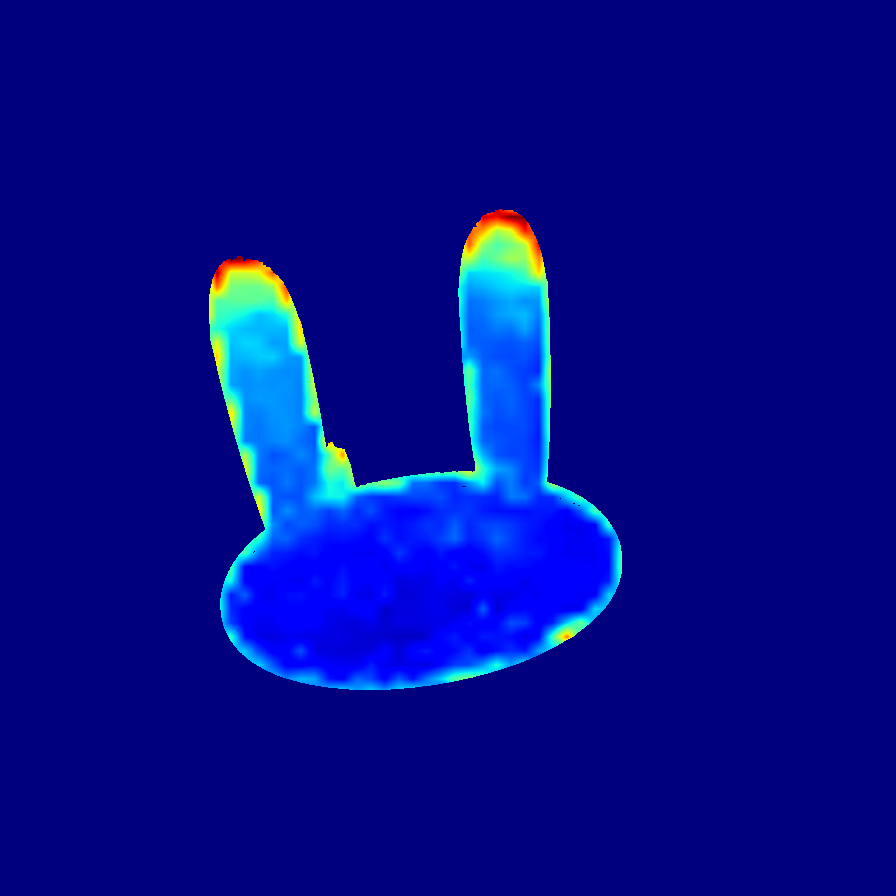} &
            \includegraphics[width=0.08\linewidth,angle=180,origin=c]{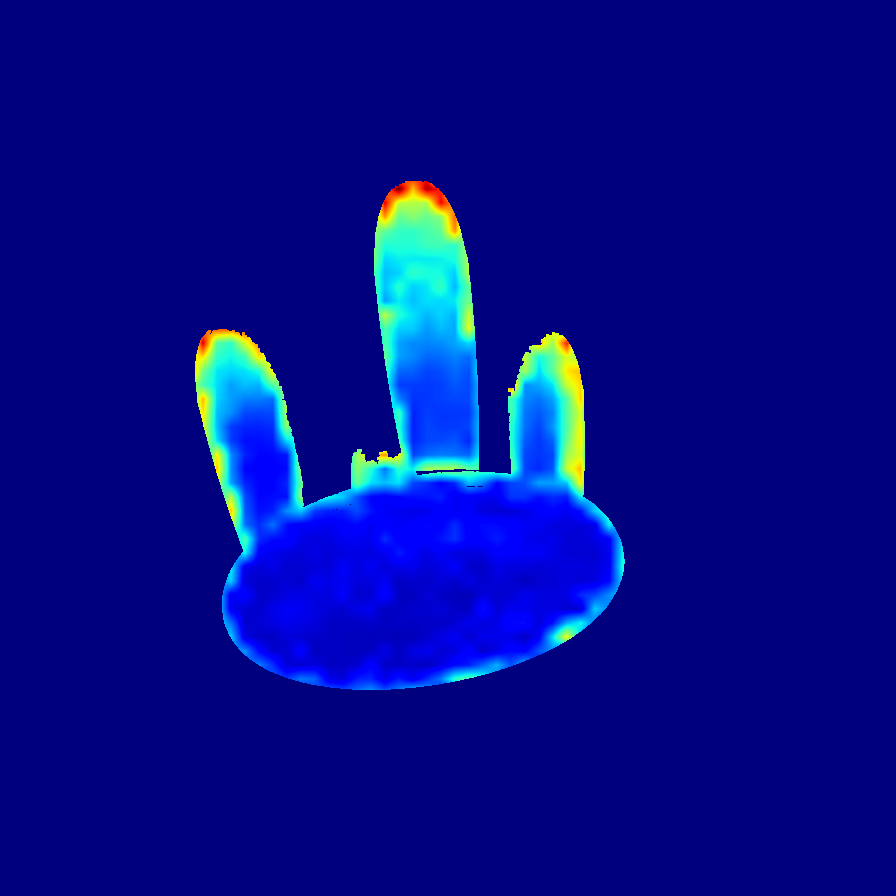} \\

            & \rotatebox{90}{\quad $v_{12}$} &
            \includegraphics[width=0.08\linewidth,angle=180,origin=c]{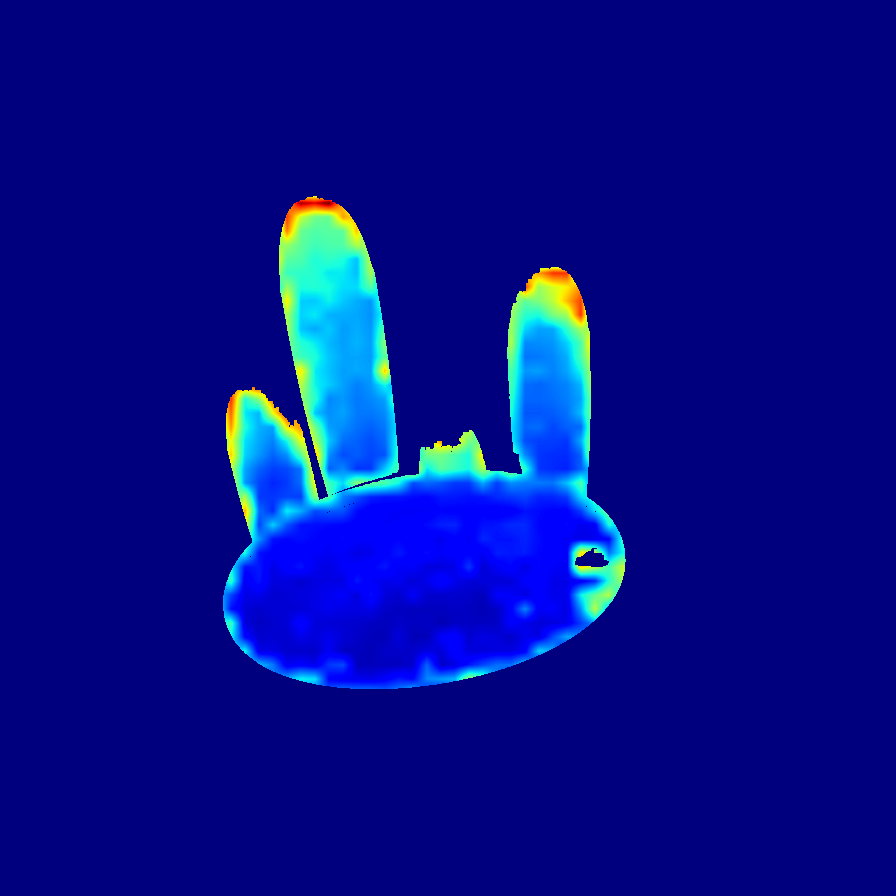} & 
            \includegraphics[width=0.08\linewidth,angle=180,origin=c]{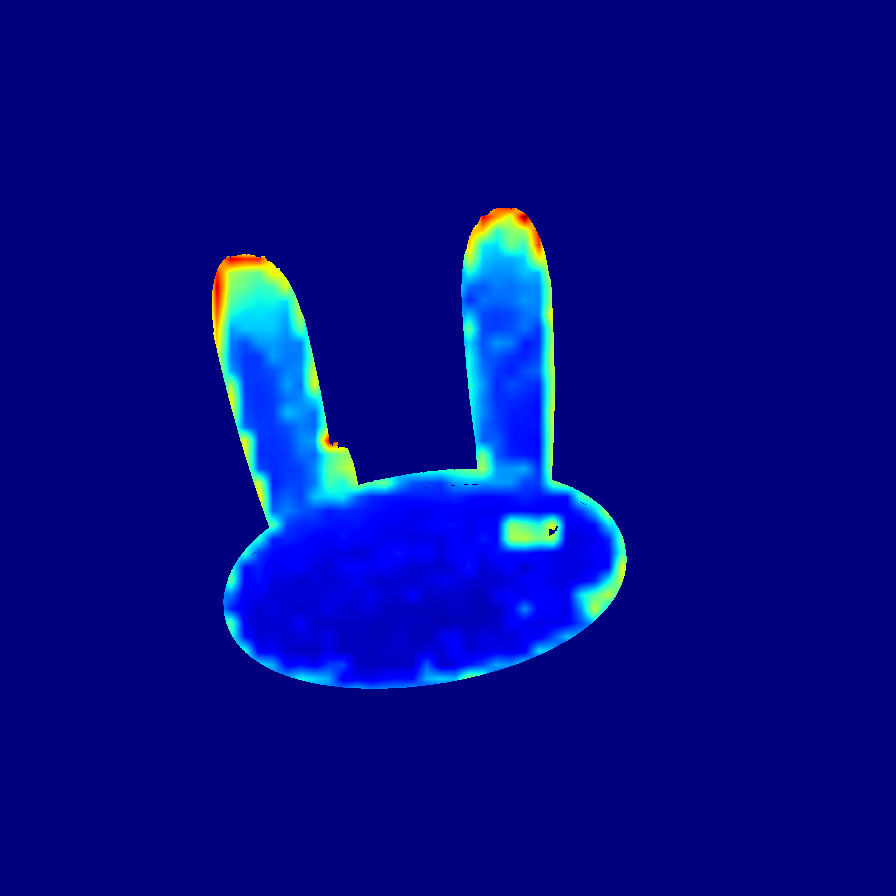} & 
            \includegraphics[width=0.08\linewidth,angle=180,origin=c]{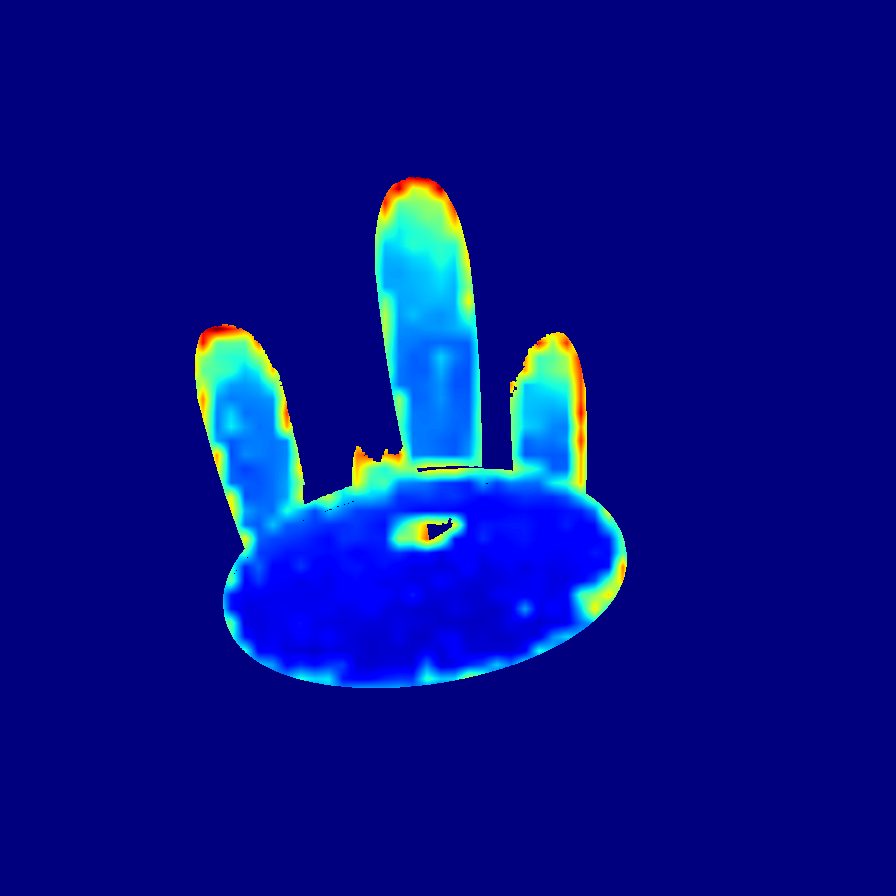} & 
            \includegraphics[width=0.08\linewidth,angle=180,origin=c]{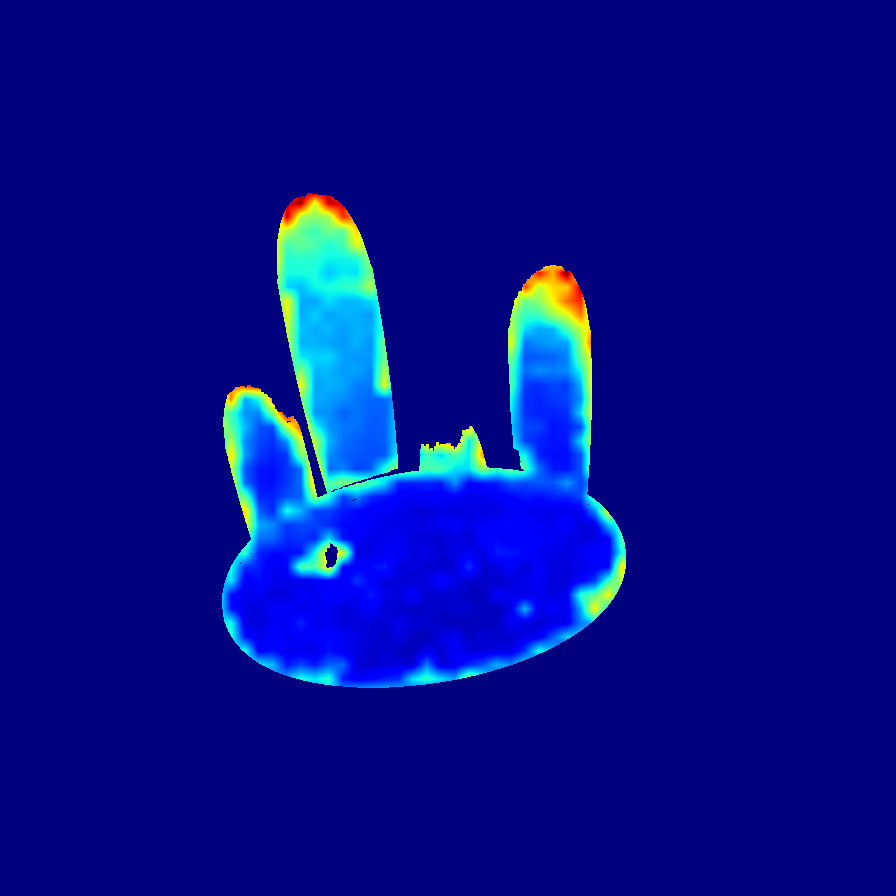} & 
            \includegraphics[width=0.08\linewidth,angle=180,origin=c]{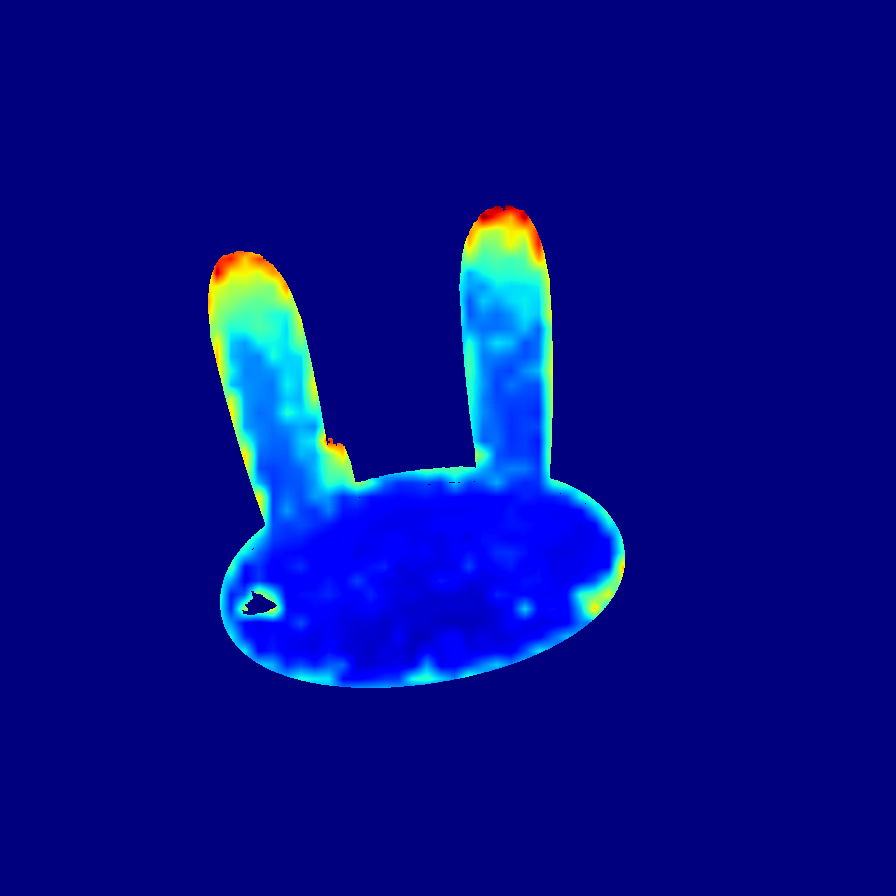} & 
            \includegraphics[width=0.08\linewidth,angle=180,origin=c]{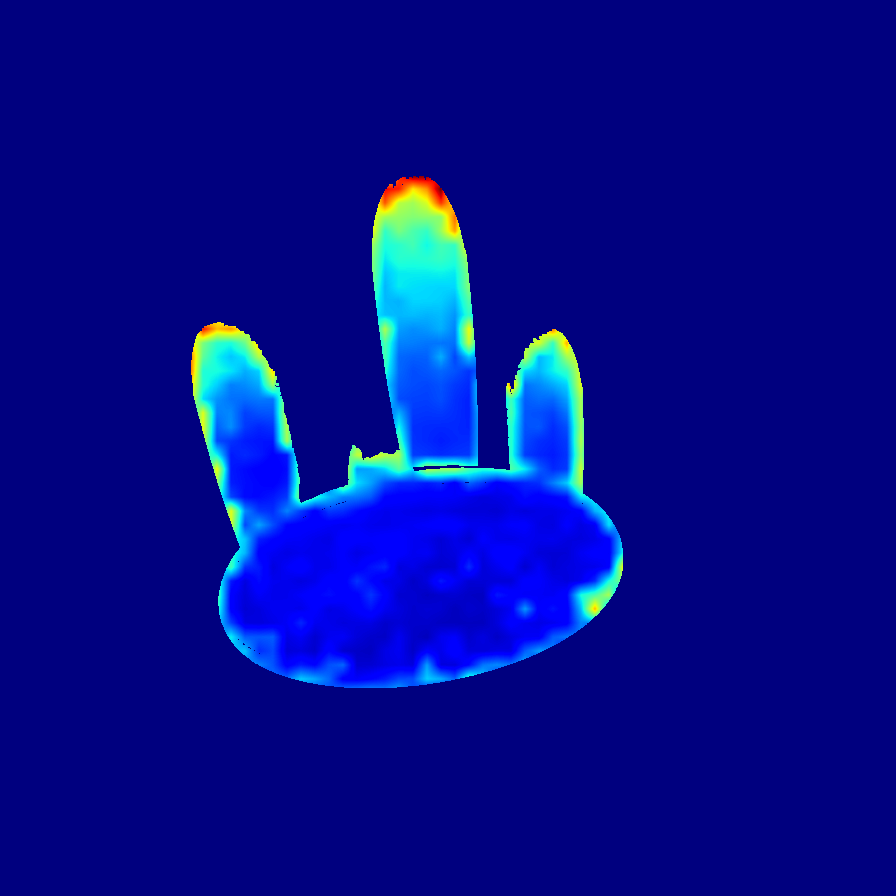} & 
            \includegraphics[width=0.08\linewidth,angle=180,origin=c]{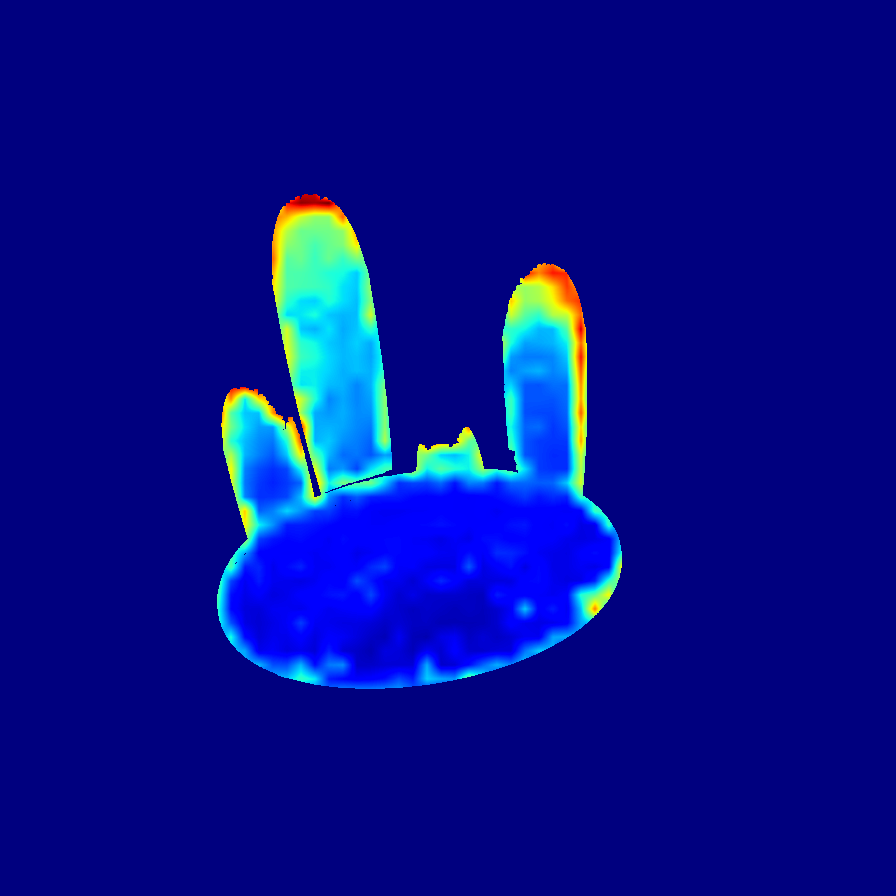} & 
            \includegraphics[width=0.08\linewidth,angle=180,origin=c]{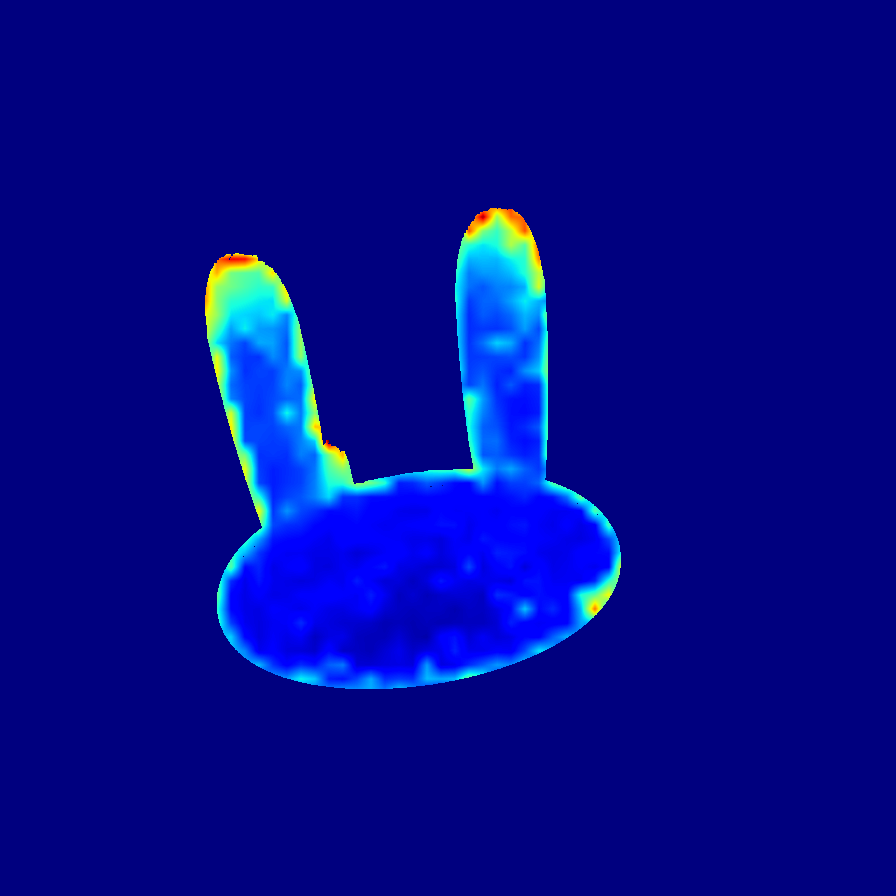} & 
            \includegraphics[width=0.08\linewidth,angle=180,origin=c]{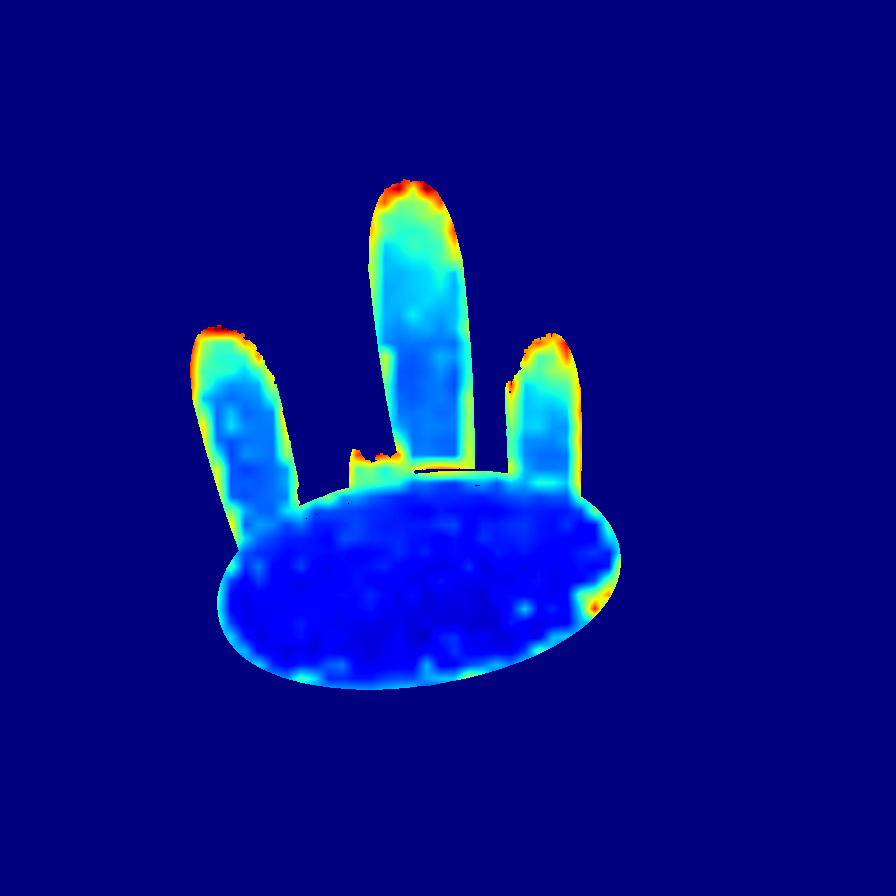} & 
            \includegraphics[width=0.08\linewidth,angle=180,origin=c]{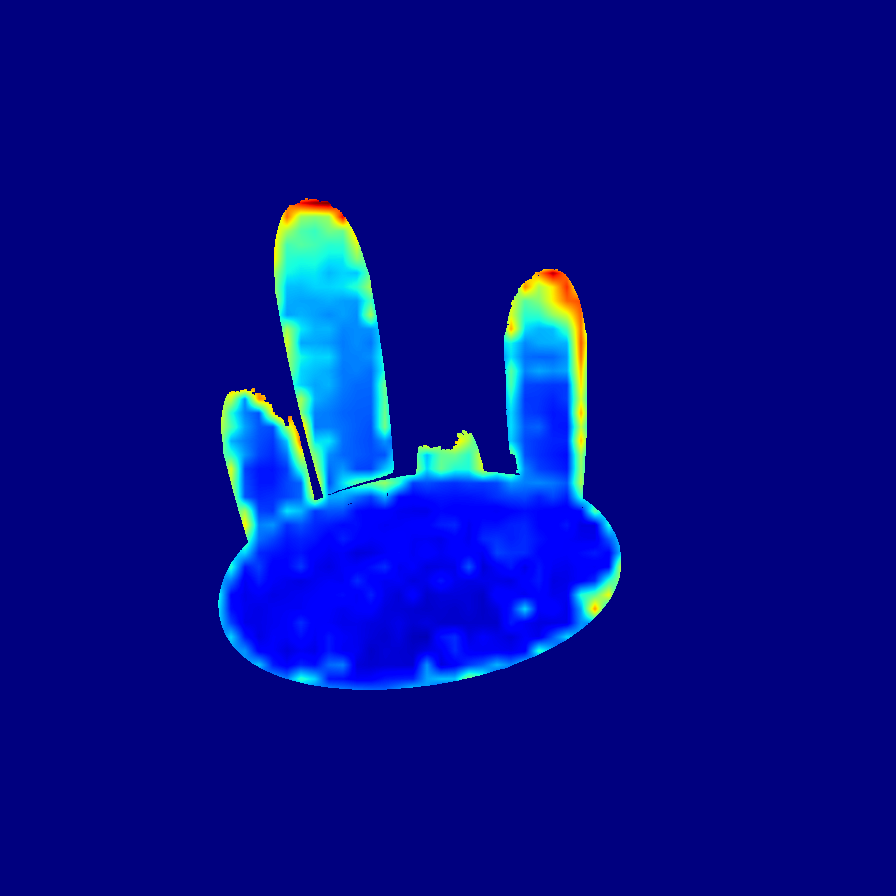} & 
            \includegraphics[width=0.08\linewidth,angle=180,origin=c]{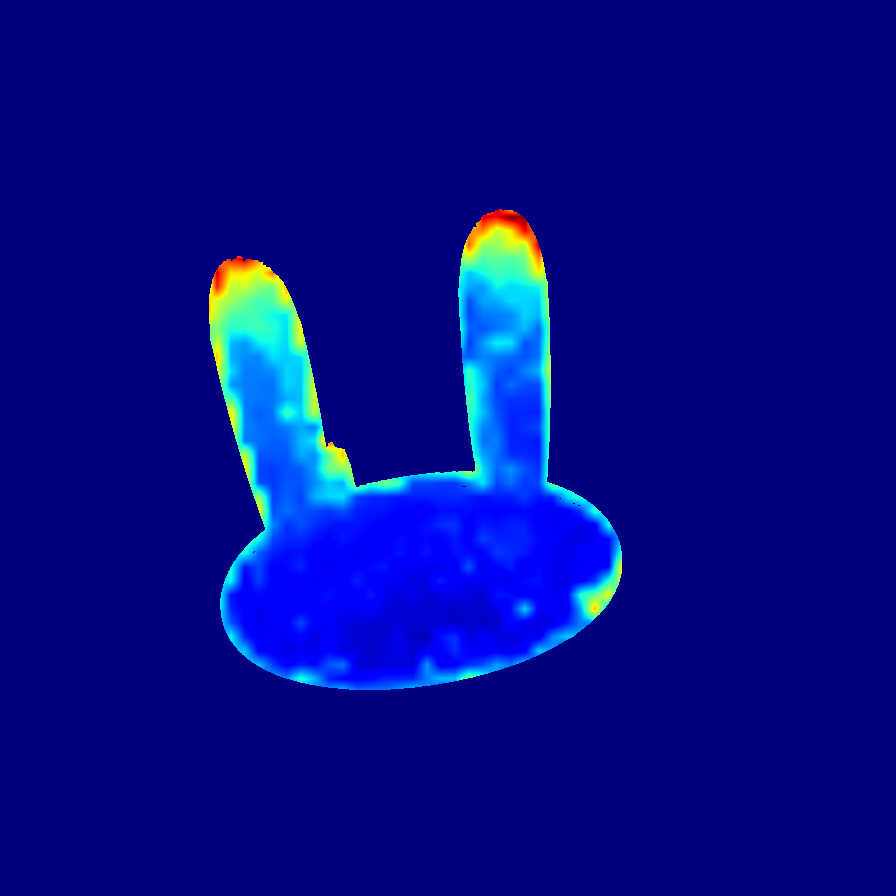} &
            \includegraphics[width=0.08\linewidth,angle=180,origin=c]{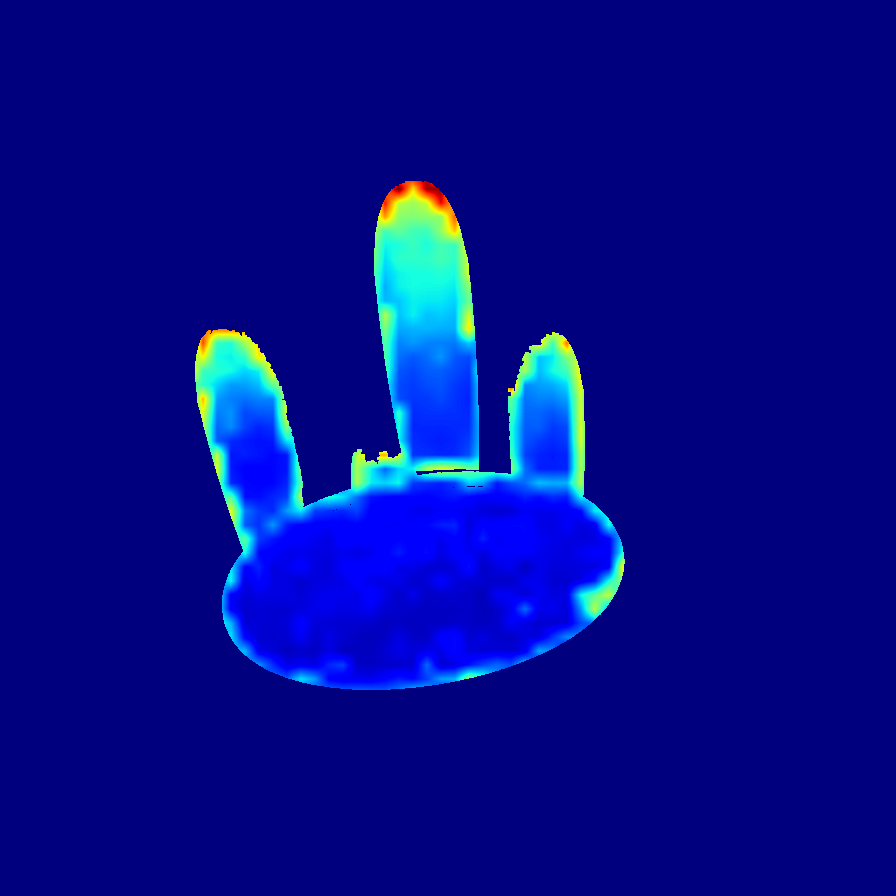} \\

            & \rotatebox{90}{\quad Ens.} &
            \includegraphics[width=0.08\linewidth,angle=180,origin=c]{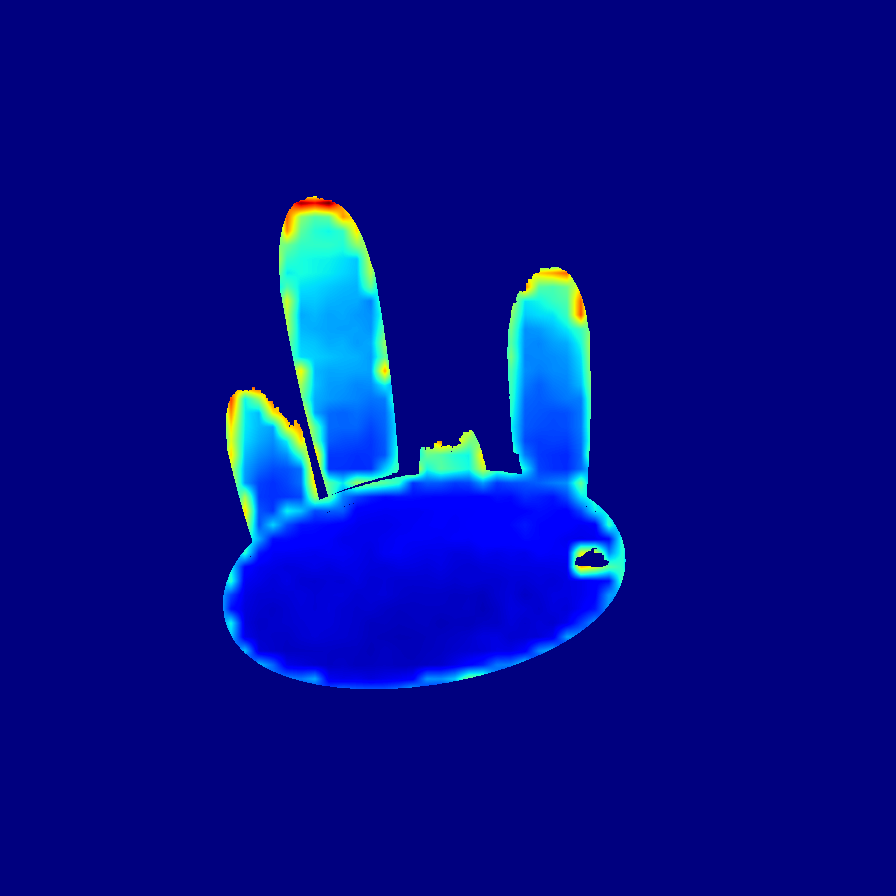} & 
            \includegraphics[width=0.08\linewidth,angle=180,origin=c]{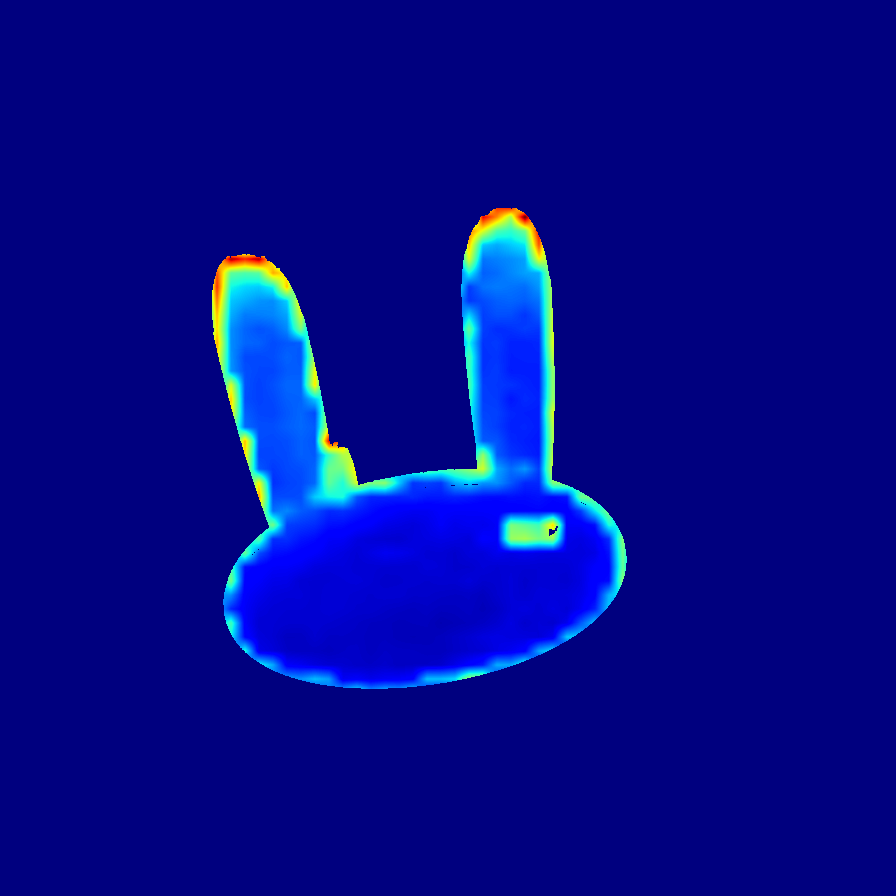} & 
            \includegraphics[width=0.08\linewidth,angle=180,origin=c]{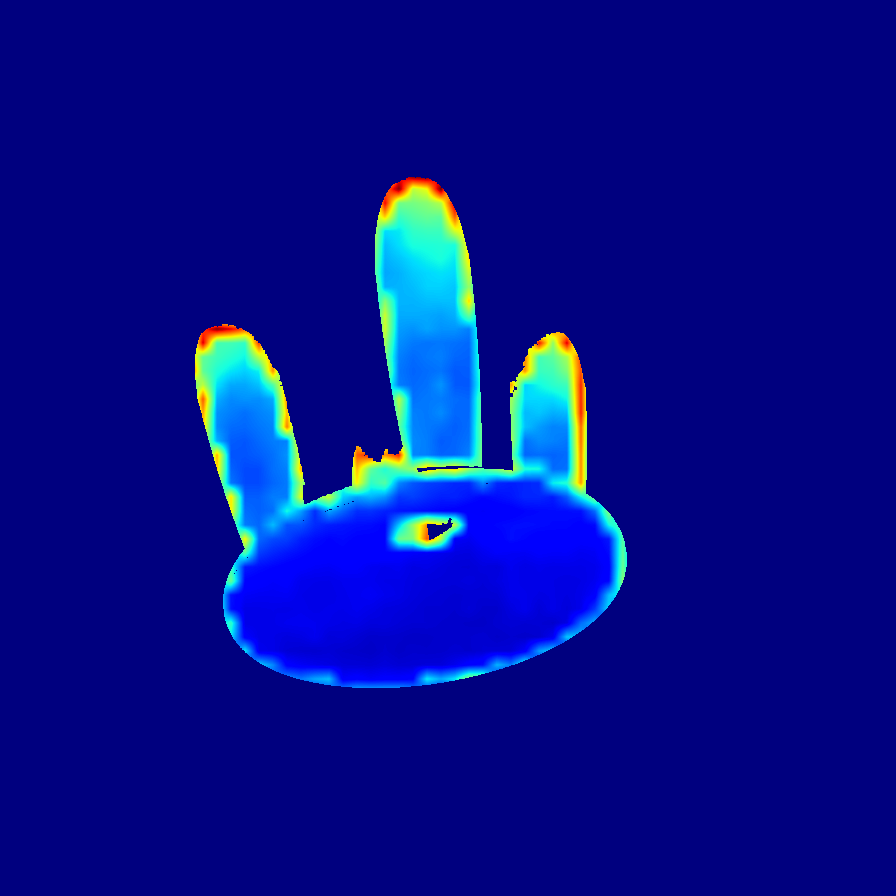} & 
            \includegraphics[width=0.08\linewidth,angle=180,origin=c]{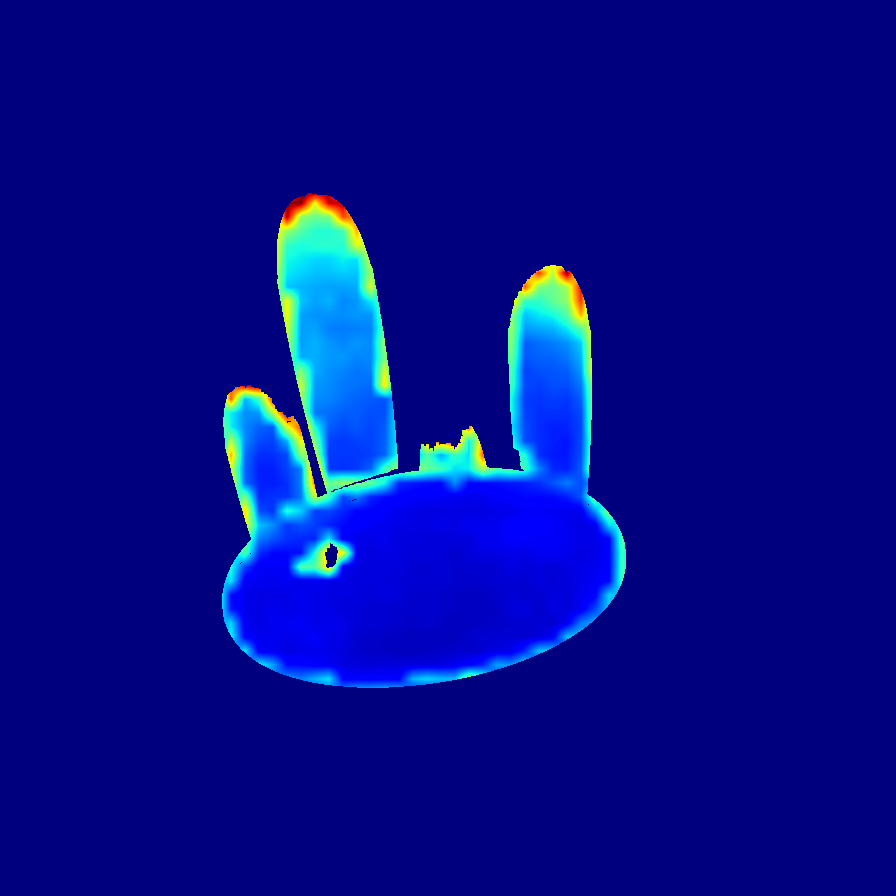} & 
            \includegraphics[width=0.08\linewidth,angle=180,origin=c]{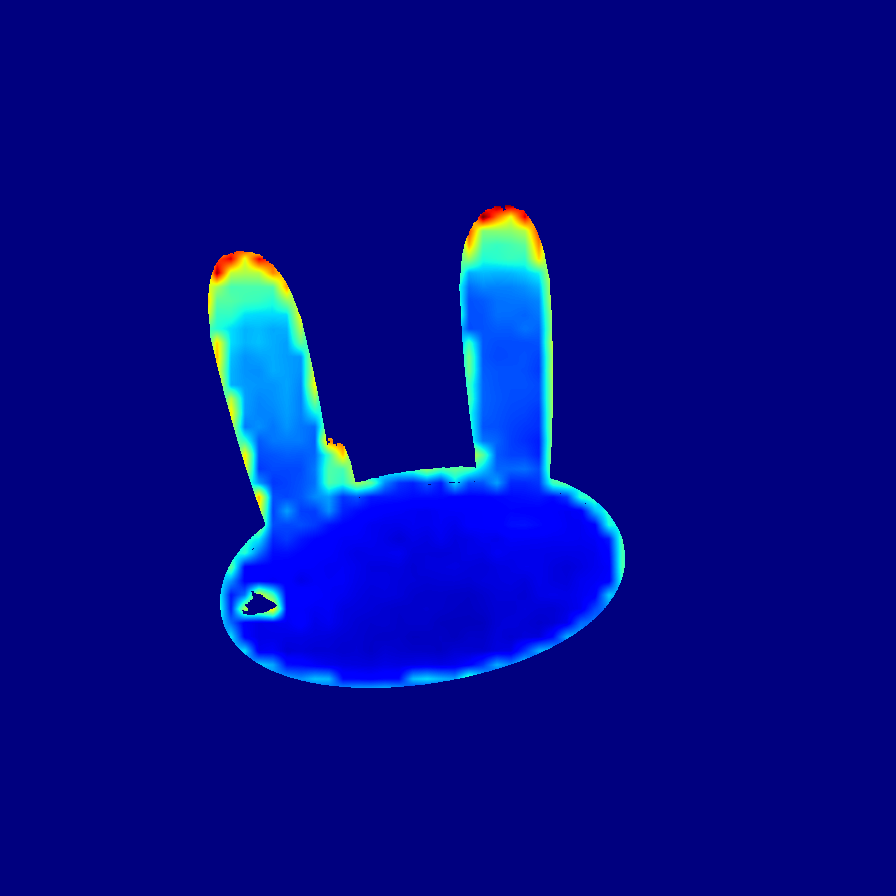} & 
            \includegraphics[width=0.08\linewidth,angle=180,origin=c]{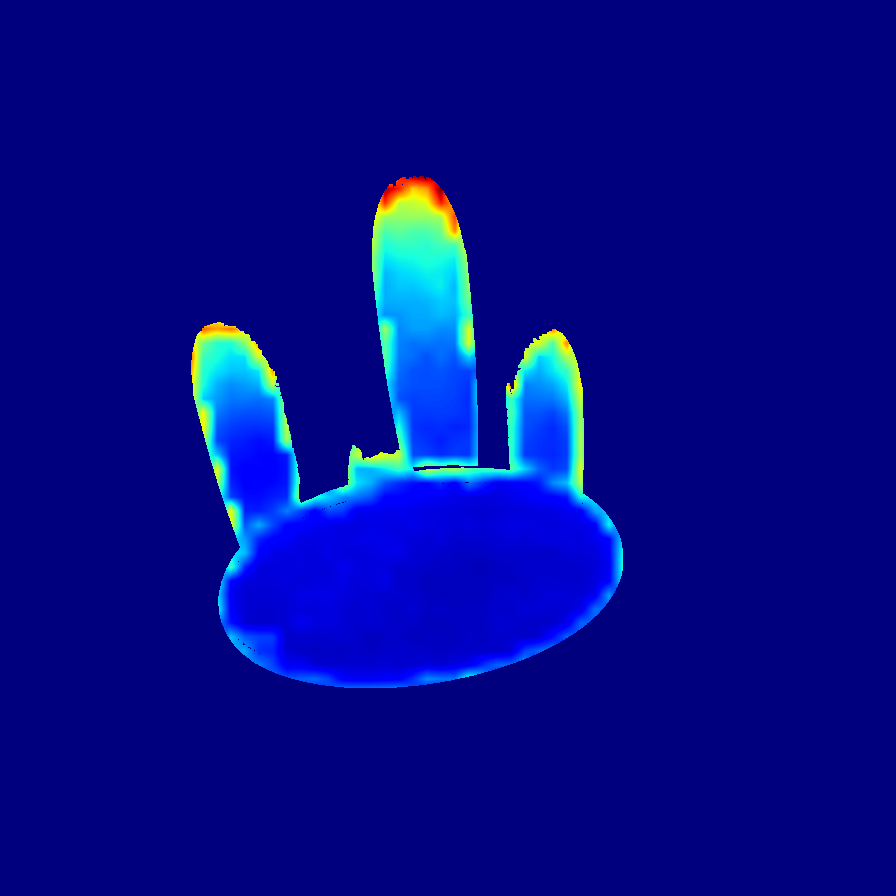} & 
            \includegraphics[width=0.08\linewidth,angle=180,origin=c]{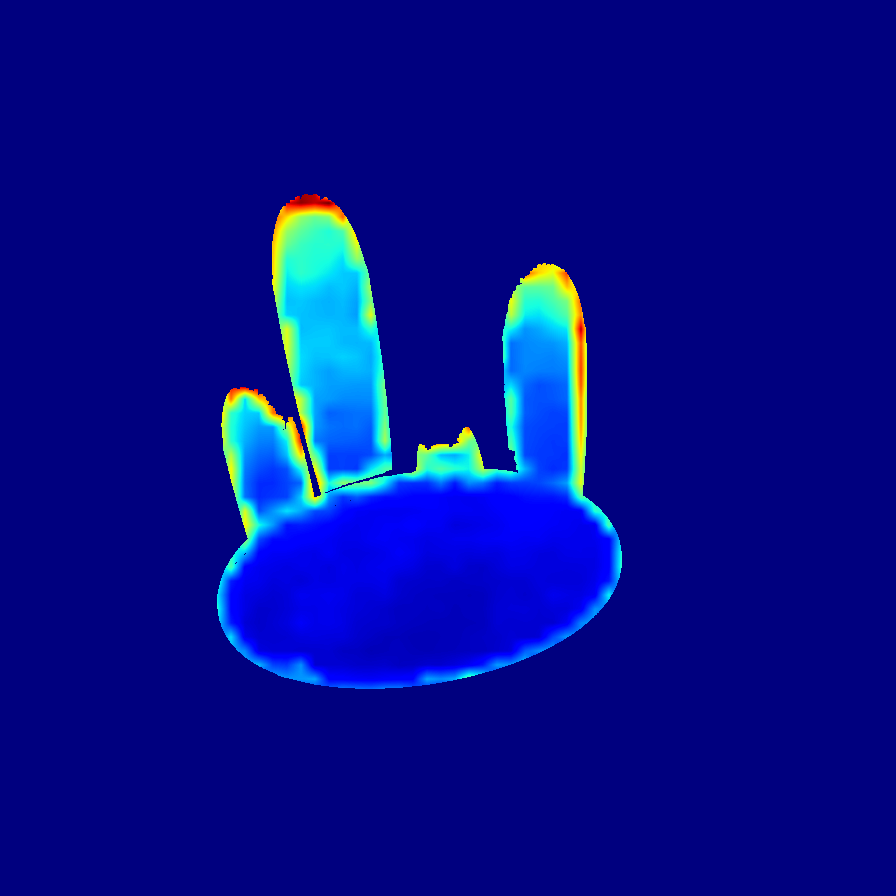} & 
            \includegraphics[width=0.08\linewidth,angle=180,origin=c]{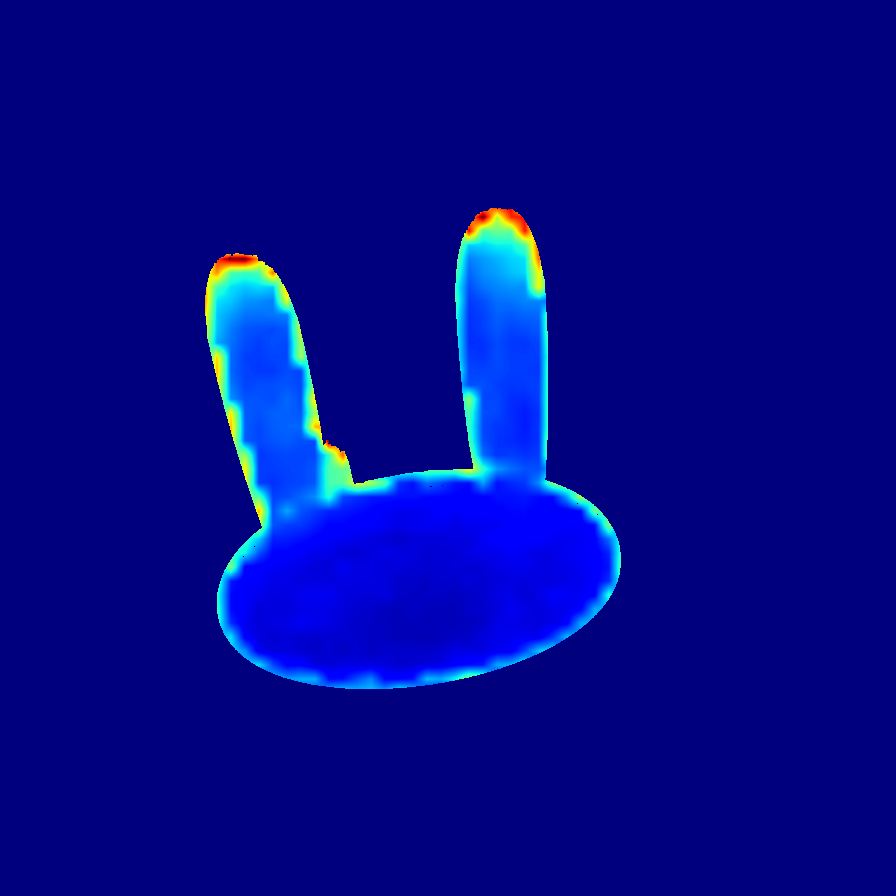} & 
            \includegraphics[width=0.08\linewidth,angle=180,origin=c]{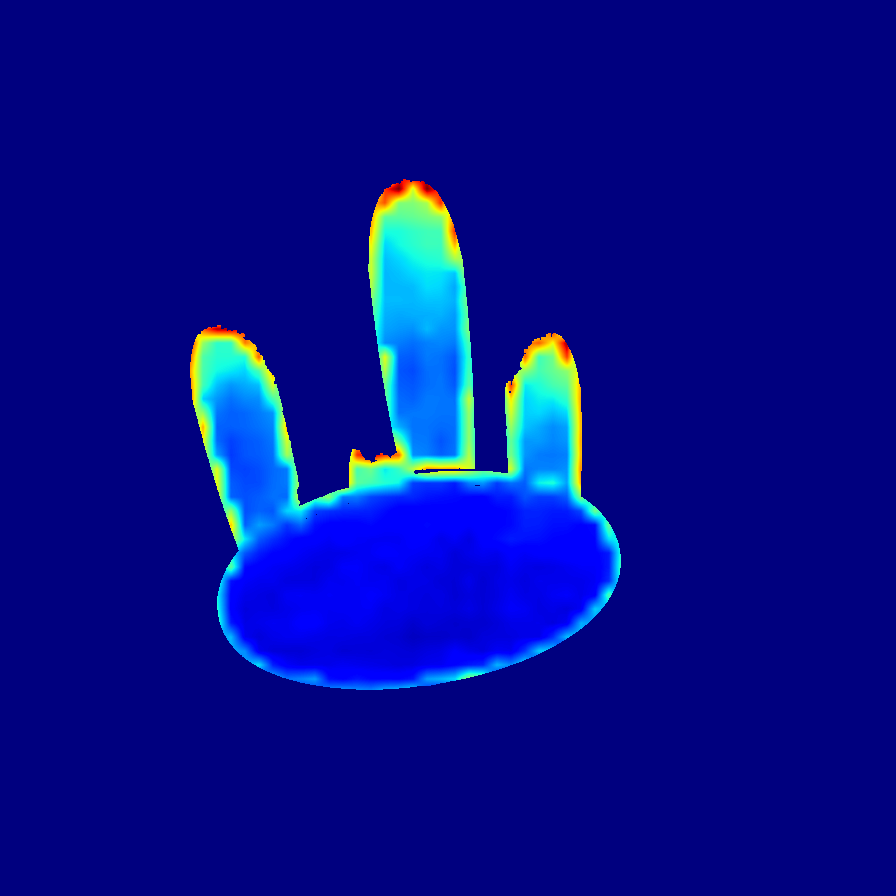} & 
            \includegraphics[width=0.08\linewidth,angle=180,origin=c]{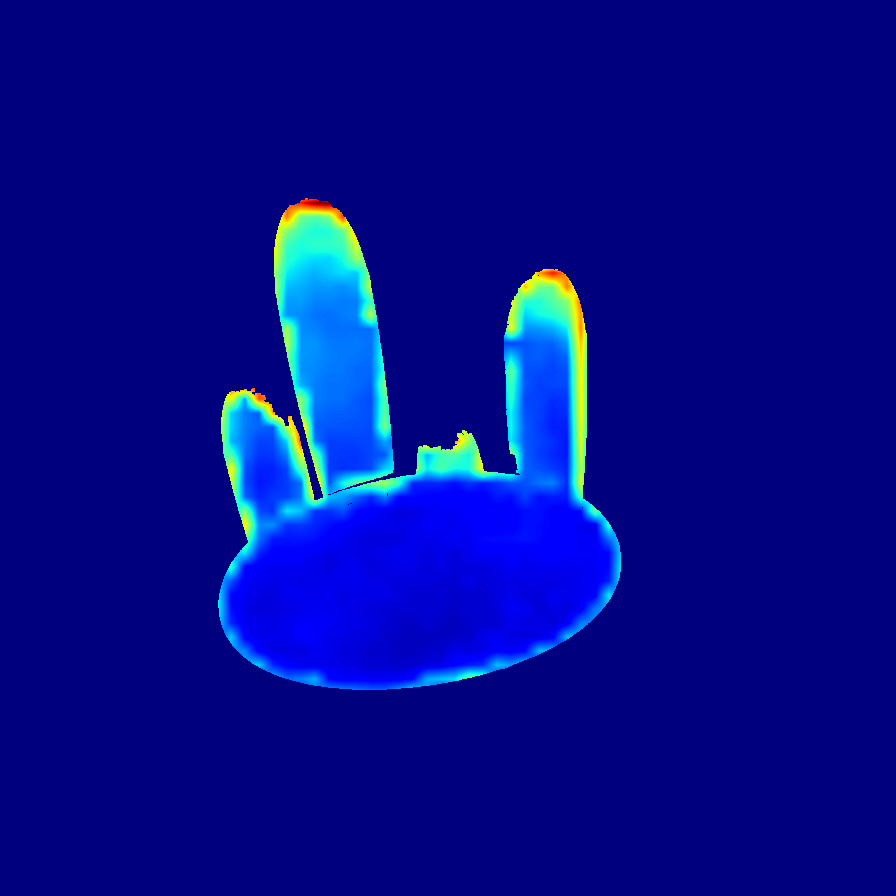} & 
            \includegraphics[width=0.08\linewidth,angle=180,origin=c]{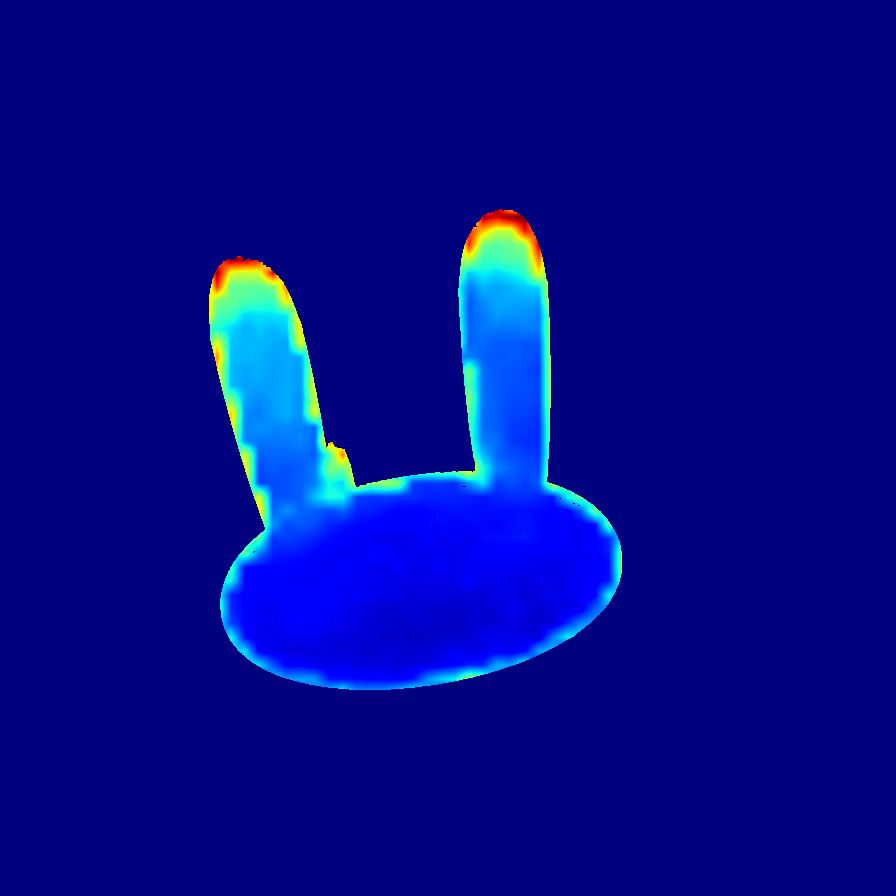} & 
            \includegraphics[width=0.08\linewidth,angle=180,origin=c]{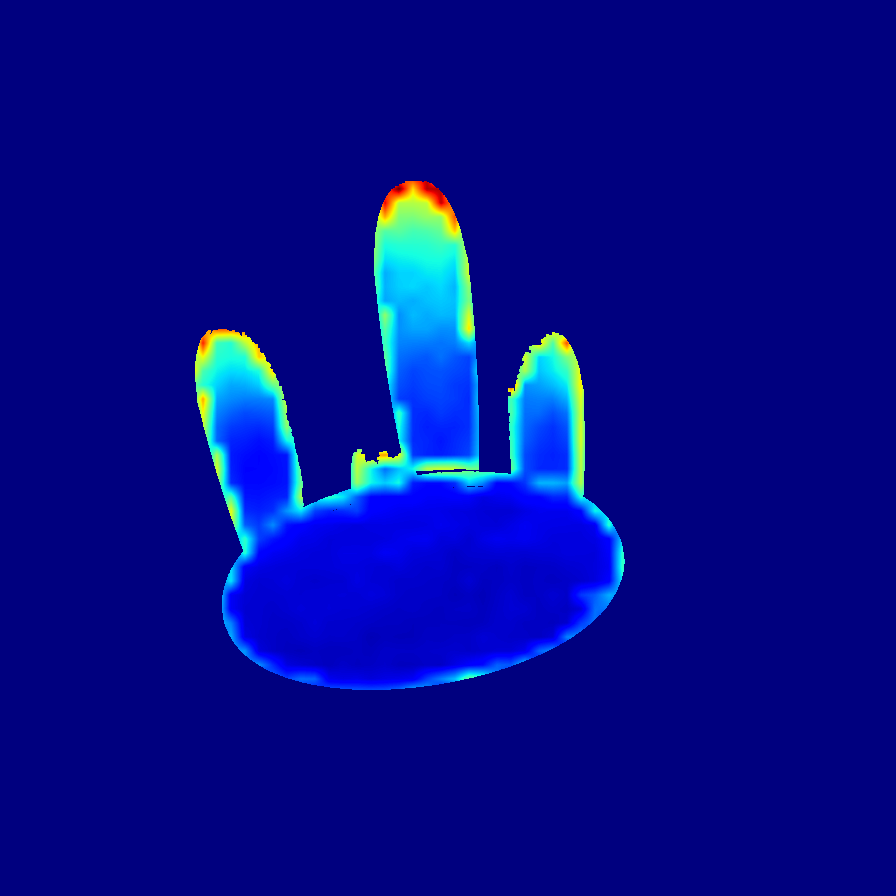} \\
            
        \end{tabular}
    }
    \caption{
        \textbf{Image-to-Depth Cross-Views.}
        }
    \label{fig:all_views_image2depth}
\end{figure*}

\begin{figure*}[t]
    \centering
    \resizebox{\linewidth}{!}{
        \setlength{\tabcolsep}{1px}
        \begin{tabular}{cc cccccccccccc}
            & & \multicolumn{12}{c}{\textbf{Target Views}} \\
            & & $v_{1}$ & $v_{2}$ & $v_{3}$ & $v_{4}$ & $v_{5}$ & $v_{6}$ & $v_{7}$ & $v_{8}$ & $v_{9}$ & $v_{10}$ & $v_{11}$ & $v_{12}$ \\

            \multirow{12}{*}{\rotatebox{90}{\textbf{Source Views} \quad \quad \quad \quad \quad \quad \quad \quad \quad \quad \quad \quad \quad \quad \quad \quad \quad \quad \quad \quad}} & \rotatebox{90}{\quad $v_{1}$} &
            \includegraphics[width=0.08\linewidth,angle=180,origin=c]{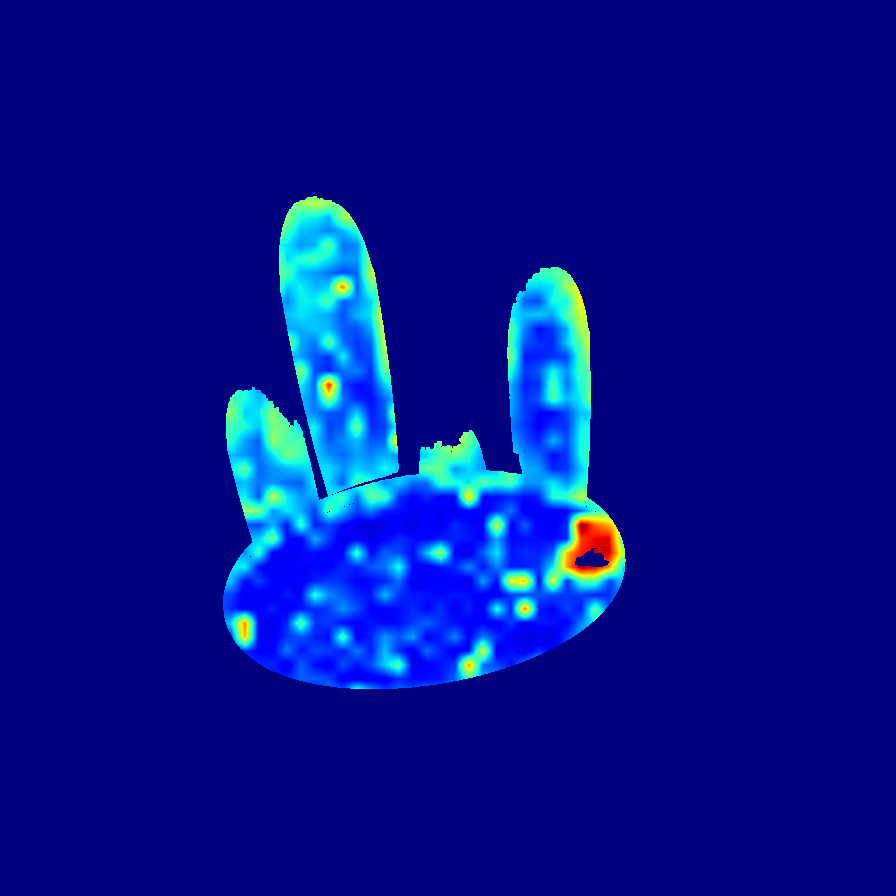} & 
            \includegraphics[width=0.08\linewidth,angle=180,origin=c]{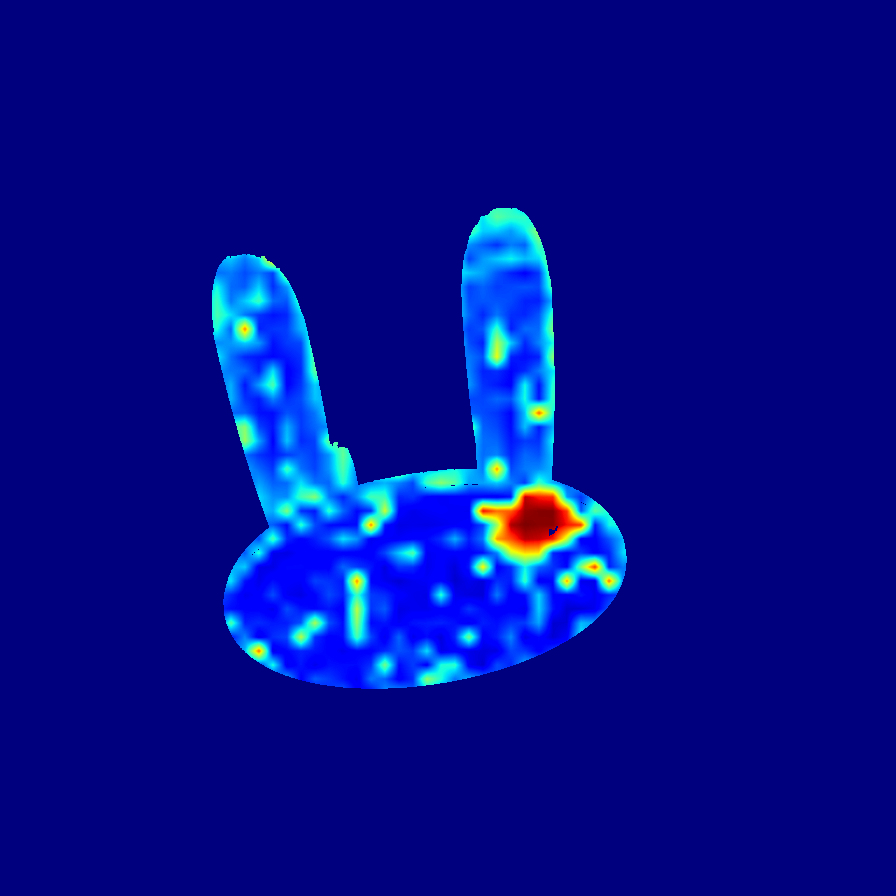} & 
            \includegraphics[width=0.08\linewidth,angle=180,origin=c]{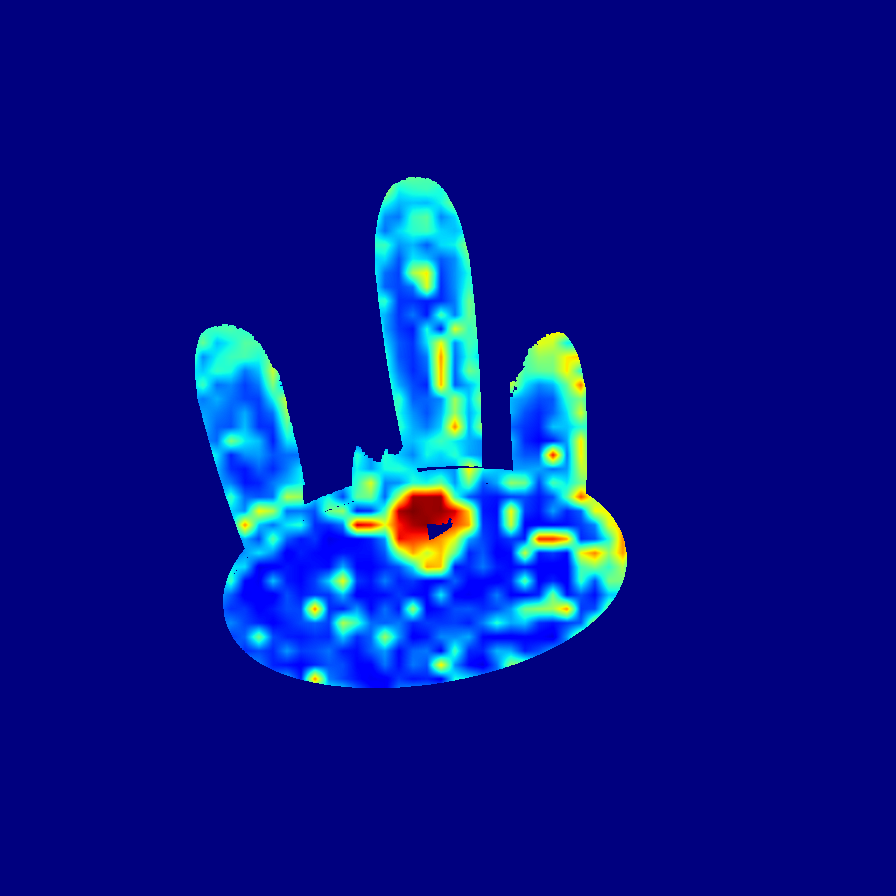} & 
            \includegraphics[width=0.08\linewidth,angle=180,origin=c]{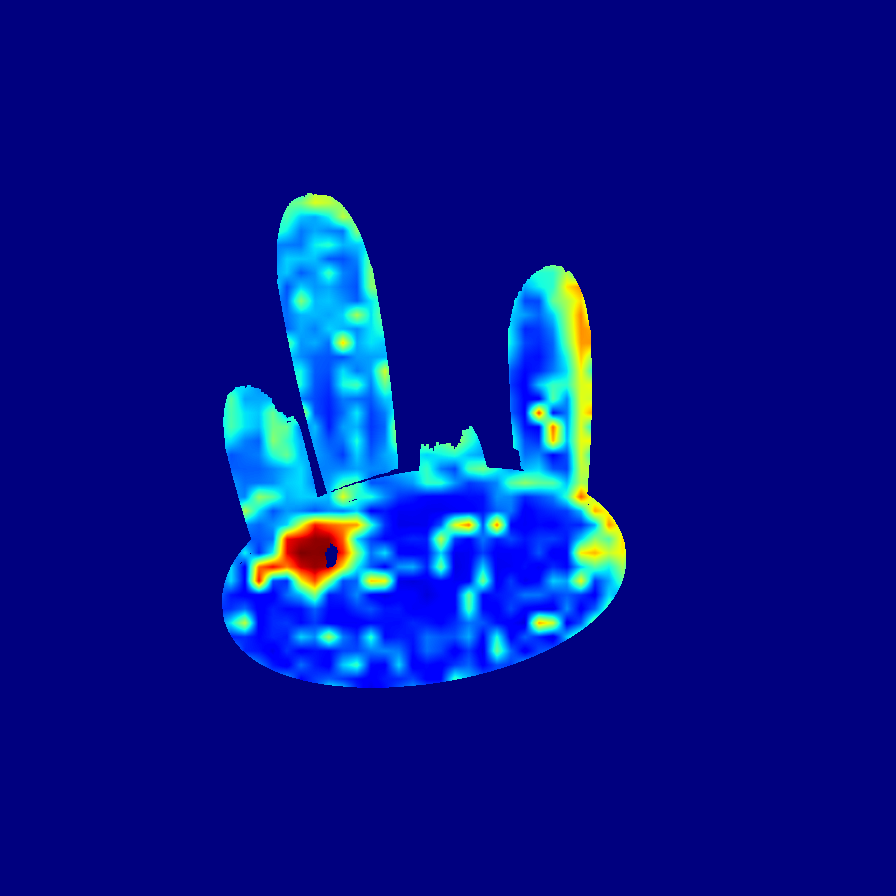} & 
            \includegraphics[width=0.08\linewidth,angle=180,origin=c]{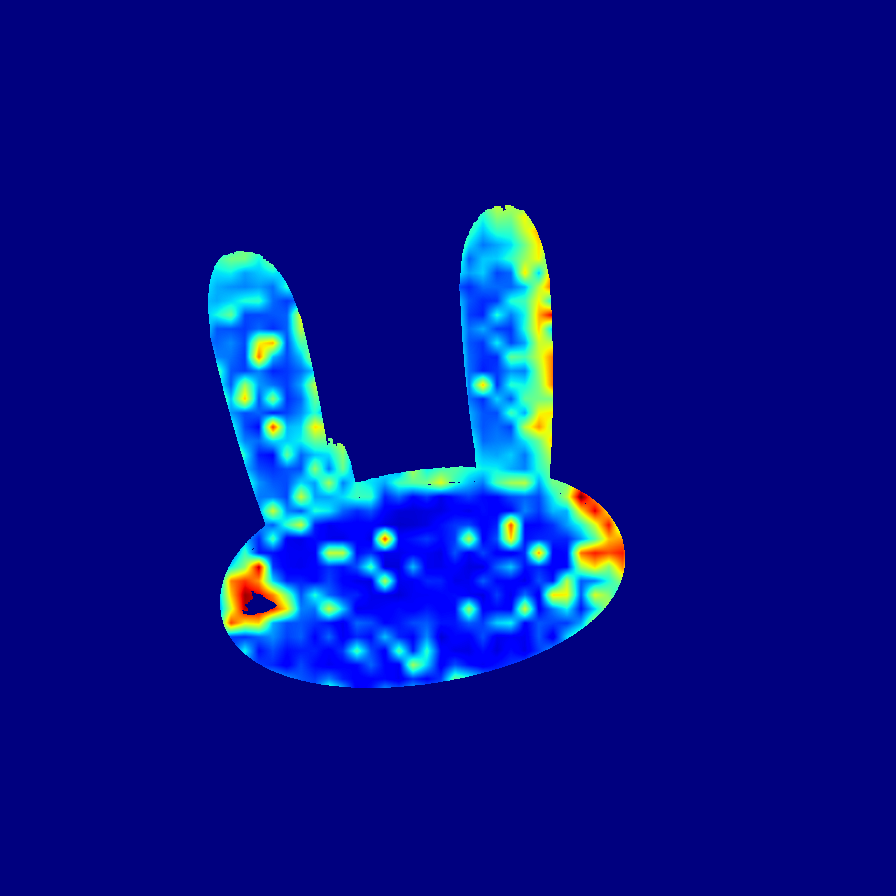} & 
            \includegraphics[width=0.08\linewidth,angle=180,origin=c]{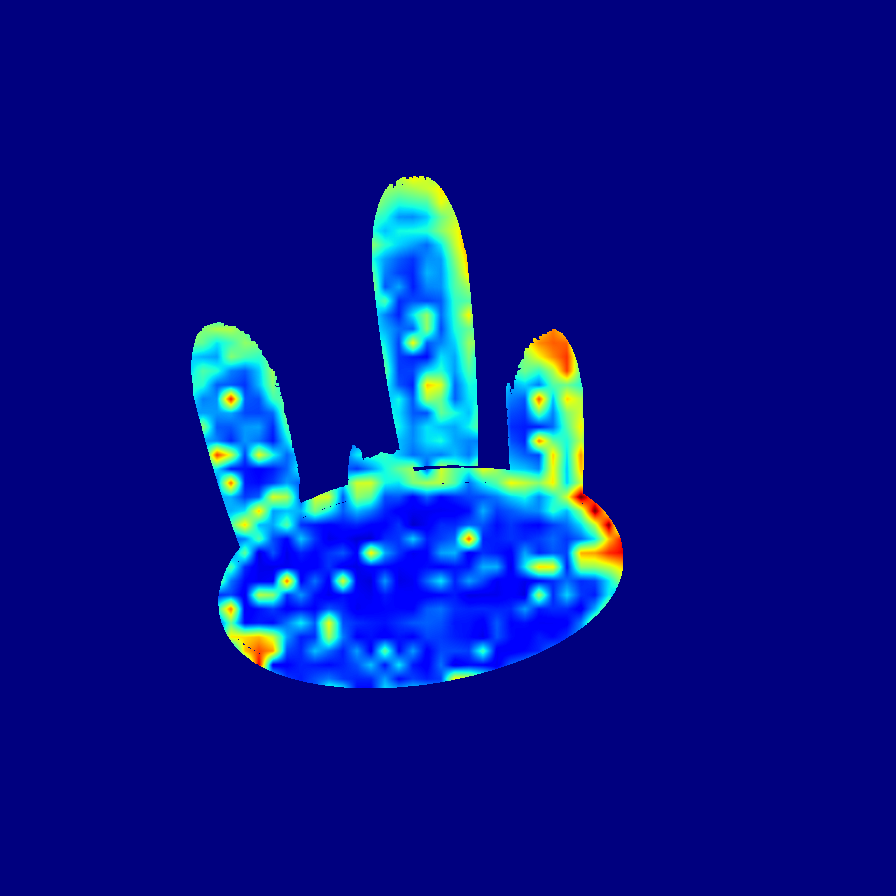} & 
            \includegraphics[width=0.08\linewidth,angle=180,origin=c]{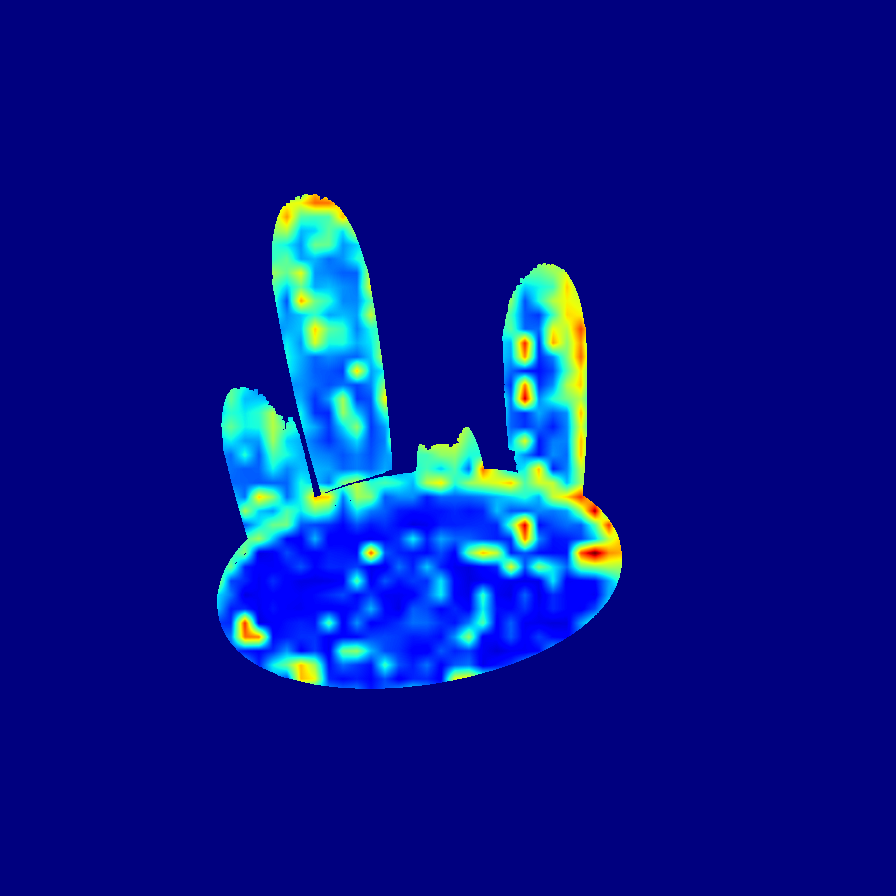} & 
            \includegraphics[width=0.08\linewidth,angle=180,origin=c]{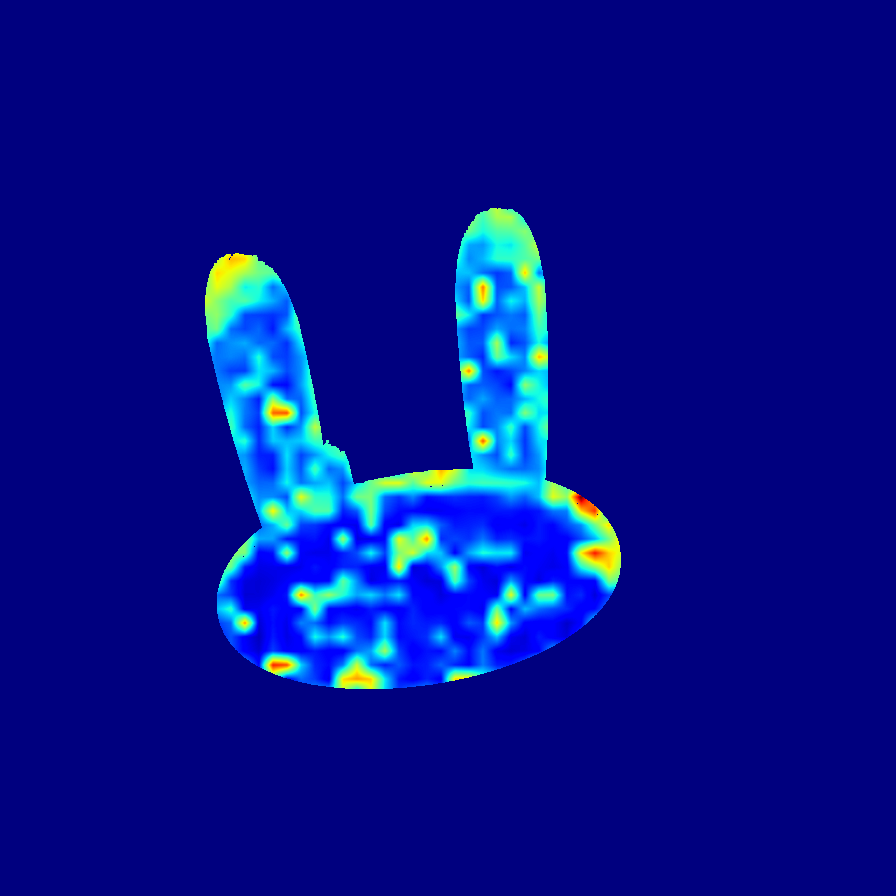} & 
            \includegraphics[width=0.08\linewidth,angle=180,origin=c]{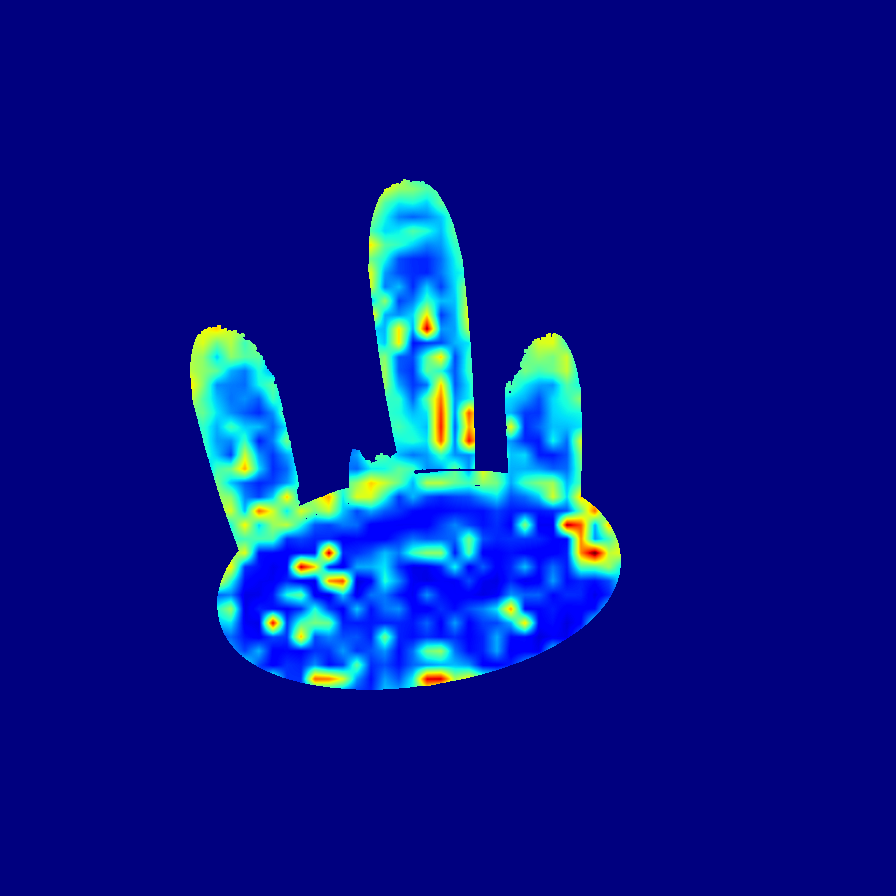} & 
            \includegraphics[width=0.08\linewidth,angle=180,origin=c]{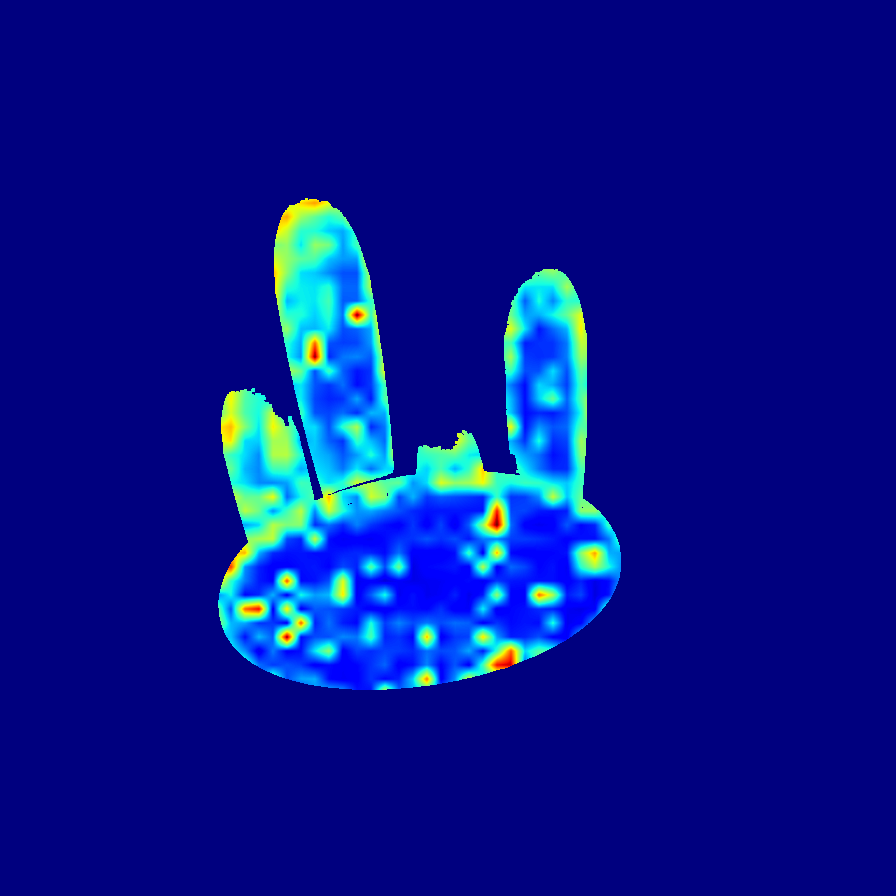} & 
            \includegraphics[width=0.08\linewidth,angle=180,origin=c]{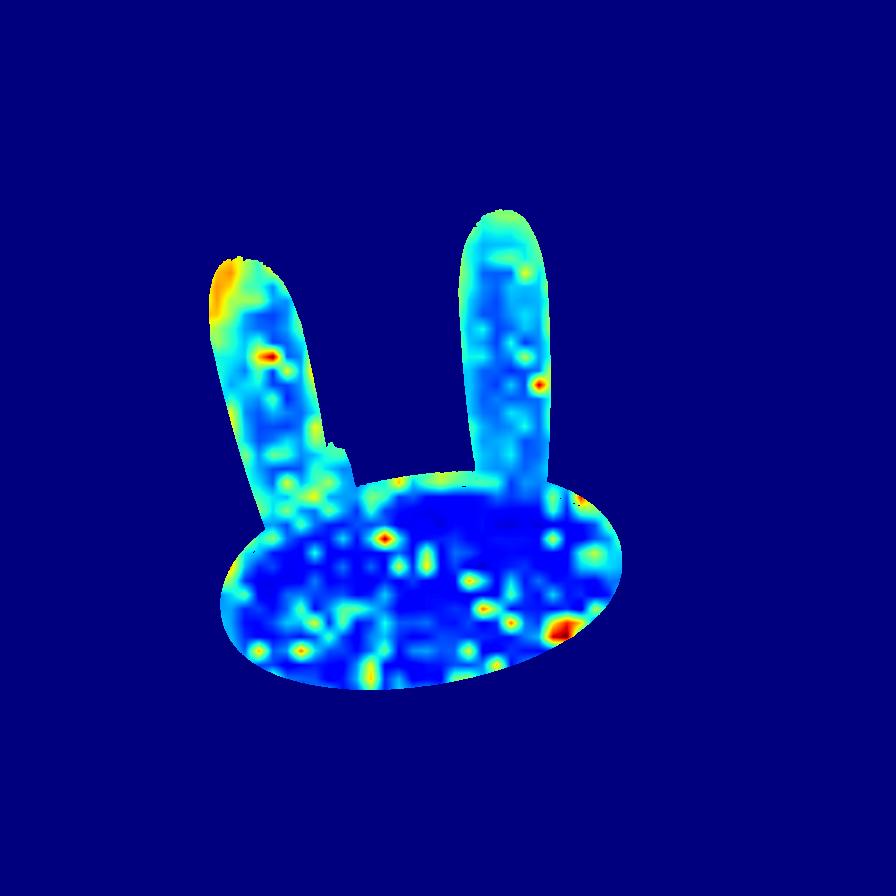} &
            \includegraphics[width=0.08\linewidth,angle=180,origin=c]{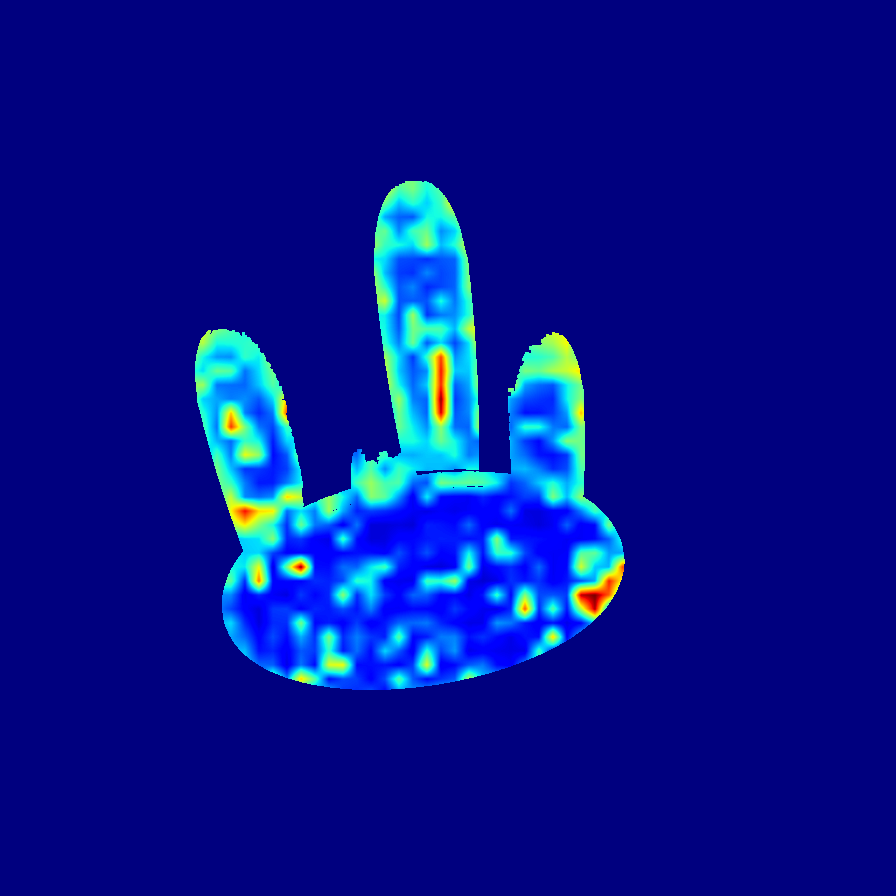} \\

            & \rotatebox{90}{\quad $v_{2}$} & 
            \includegraphics[width=0.08\linewidth,angle=180,origin=c]{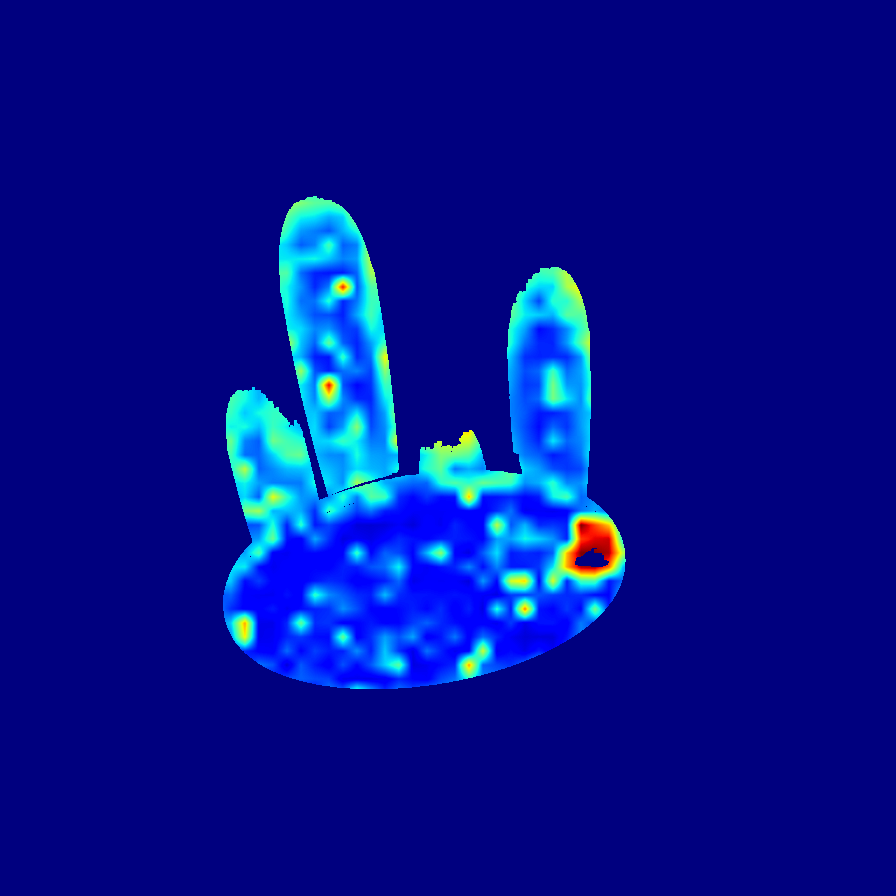} & 
            \includegraphics[width=0.08\linewidth,angle=180,origin=c]{images/cross_views_depth2image/02_am_C2_C2.png} & 
            \includegraphics[width=0.08\linewidth,angle=180,origin=c]{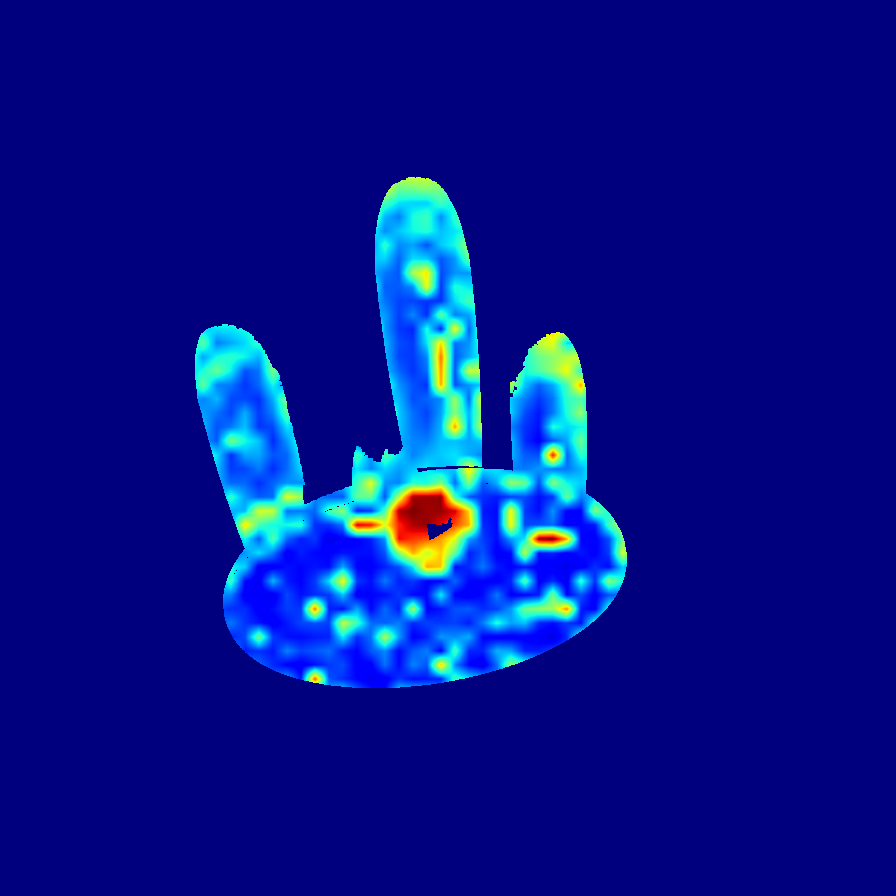} & 
            \includegraphics[width=0.08\linewidth,angle=180,origin=c]{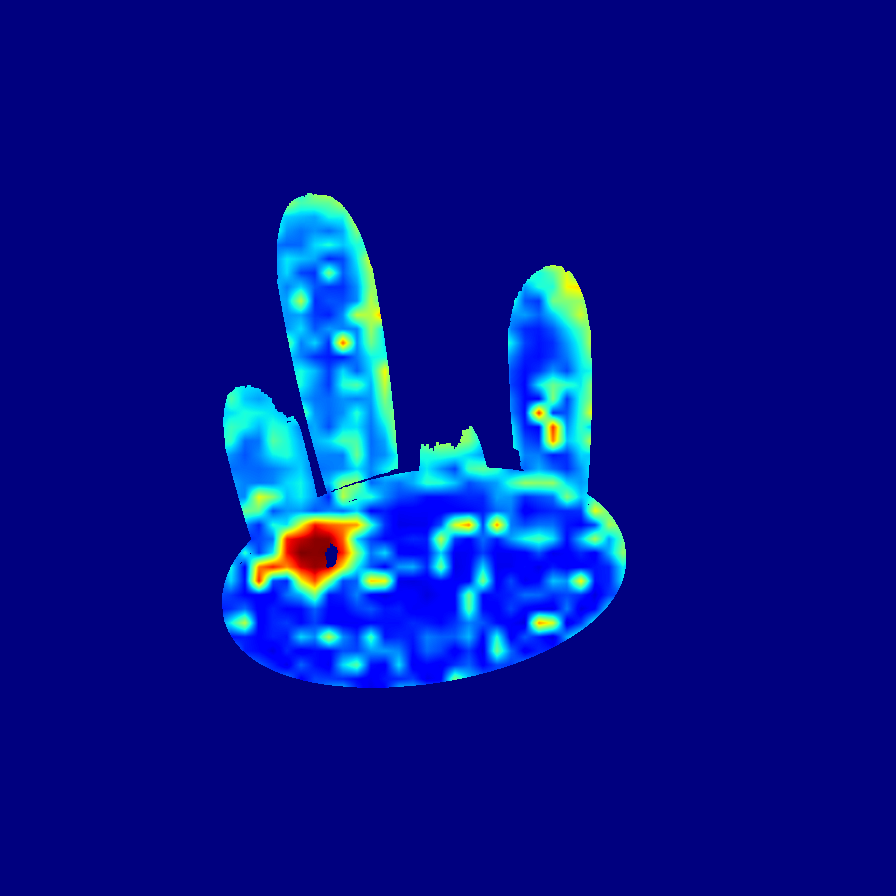} & 
            \includegraphics[width=0.08\linewidth,angle=180,origin=c]{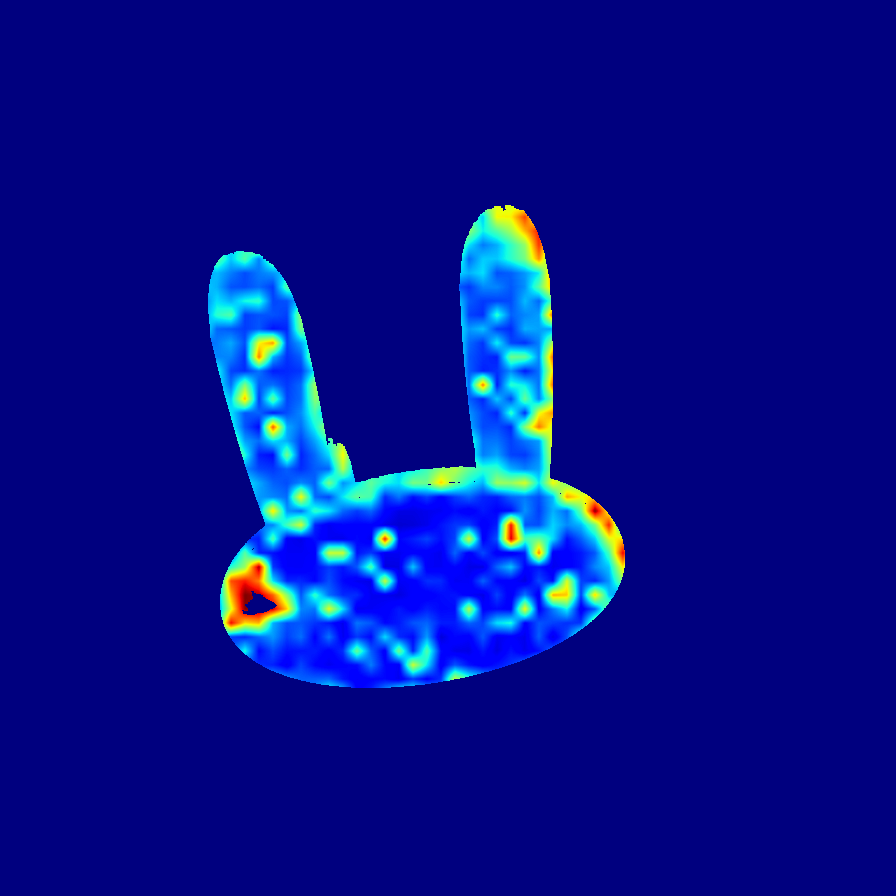} & 
            \includegraphics[width=0.08\linewidth,angle=180,origin=c]{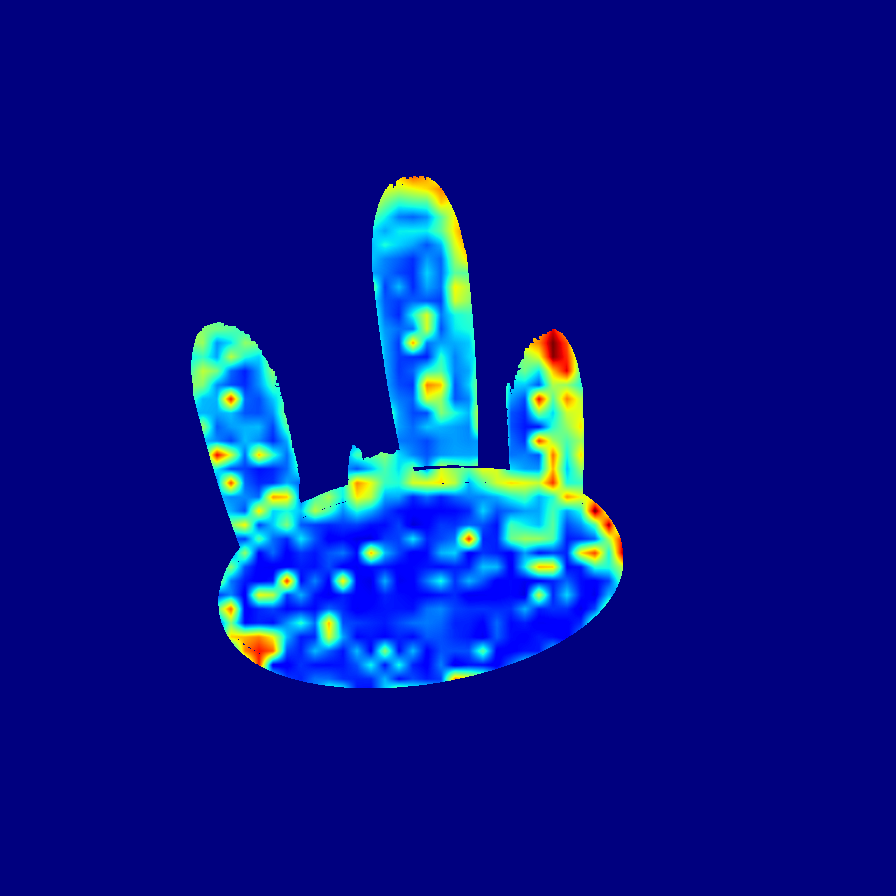} & 
            \includegraphics[width=0.08\linewidth,angle=180,origin=c]{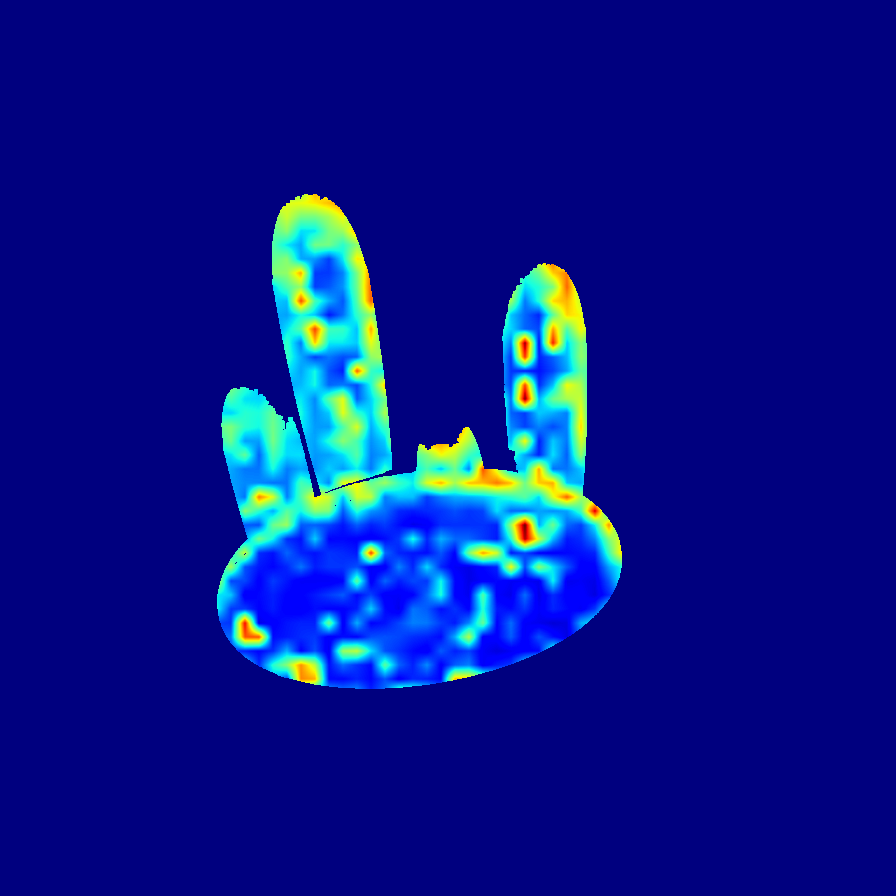} & 
            \includegraphics[width=0.08\linewidth,angle=180,origin=c]{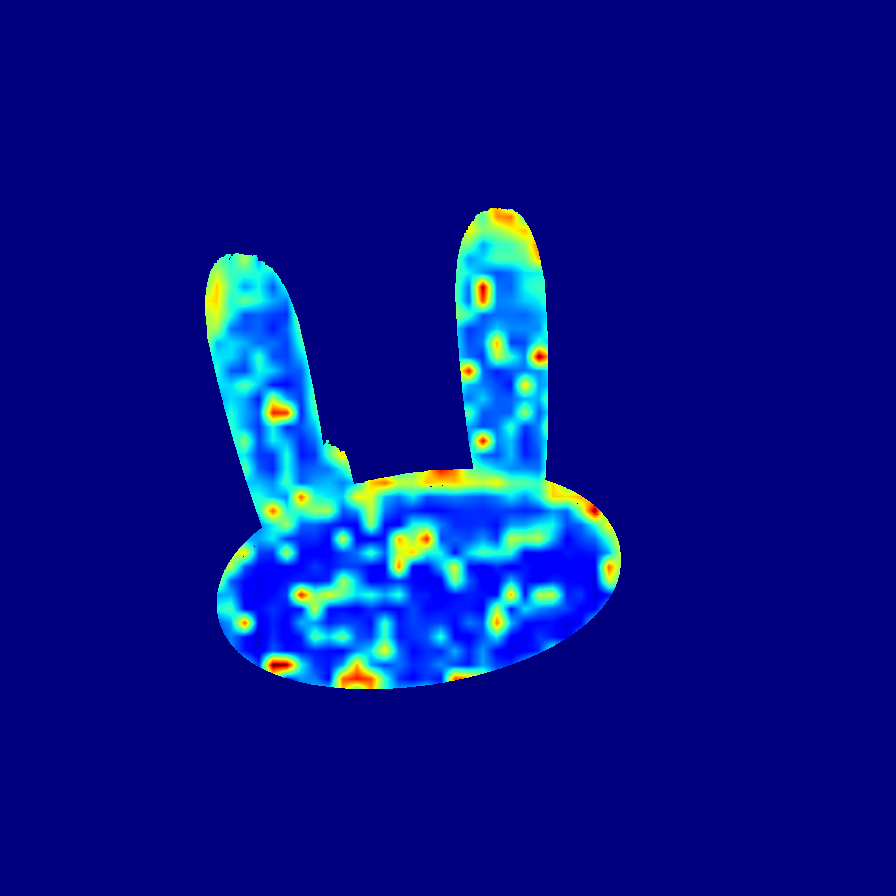} & 
            \includegraphics[width=0.08\linewidth,angle=180,origin=c]{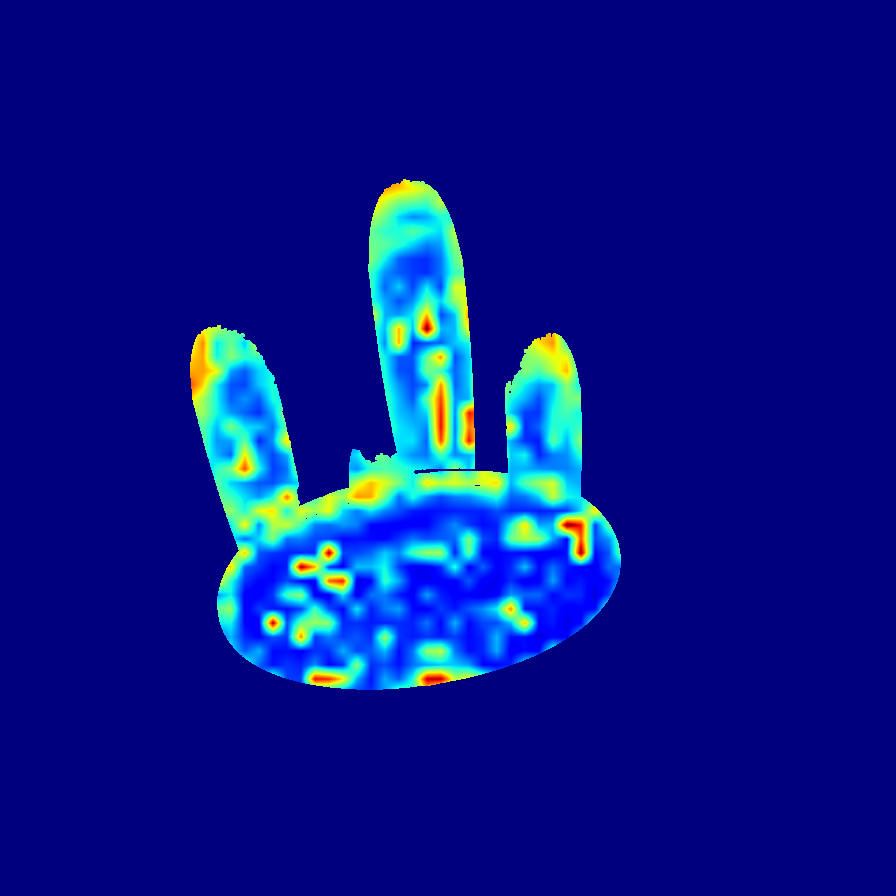} & 
            \includegraphics[width=0.08\linewidth,angle=180,origin=c]{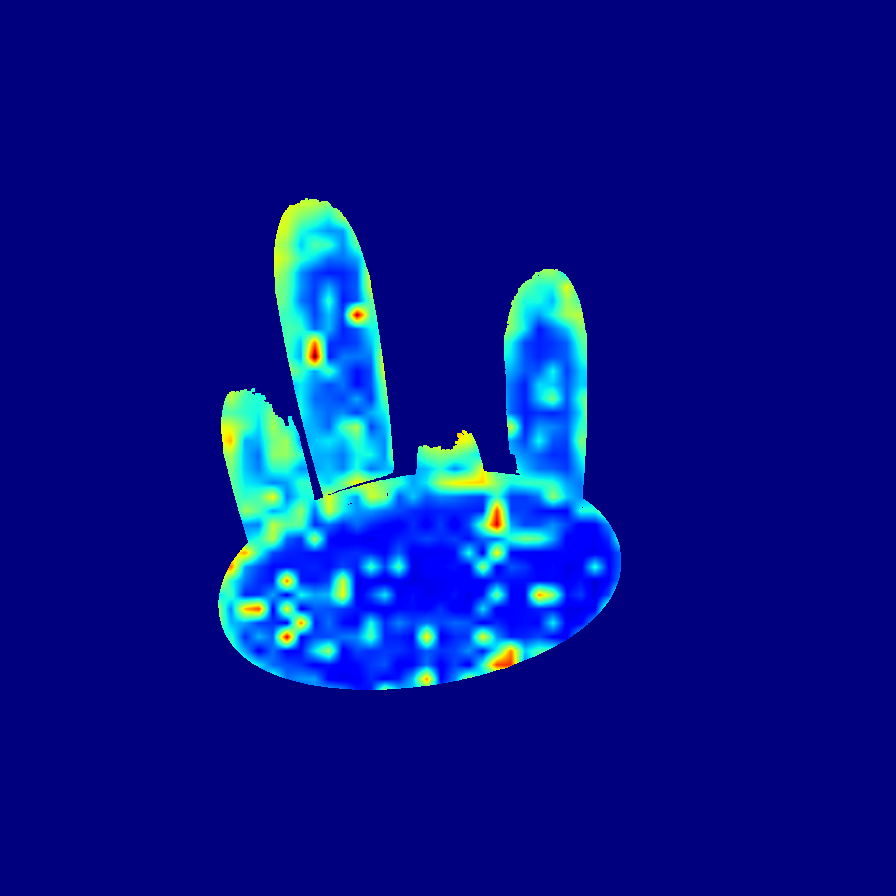} & 
            \includegraphics[width=0.08\linewidth,angle=180,origin=c]{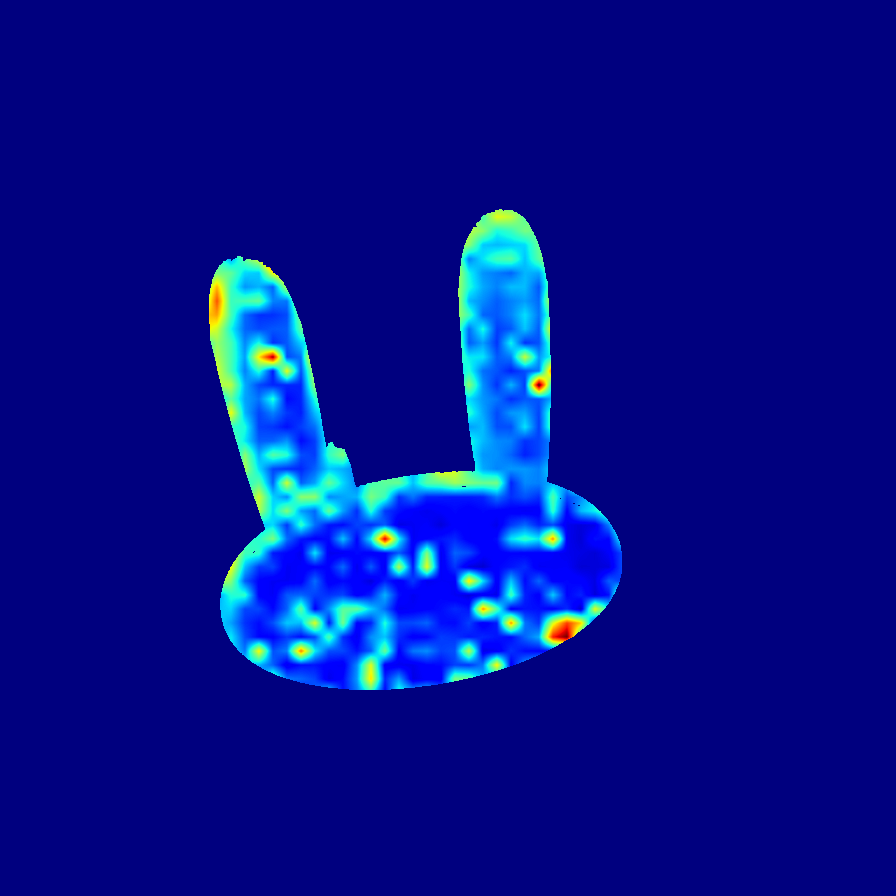} &
            \includegraphics[width=0.08\linewidth,angle=180,origin=c]{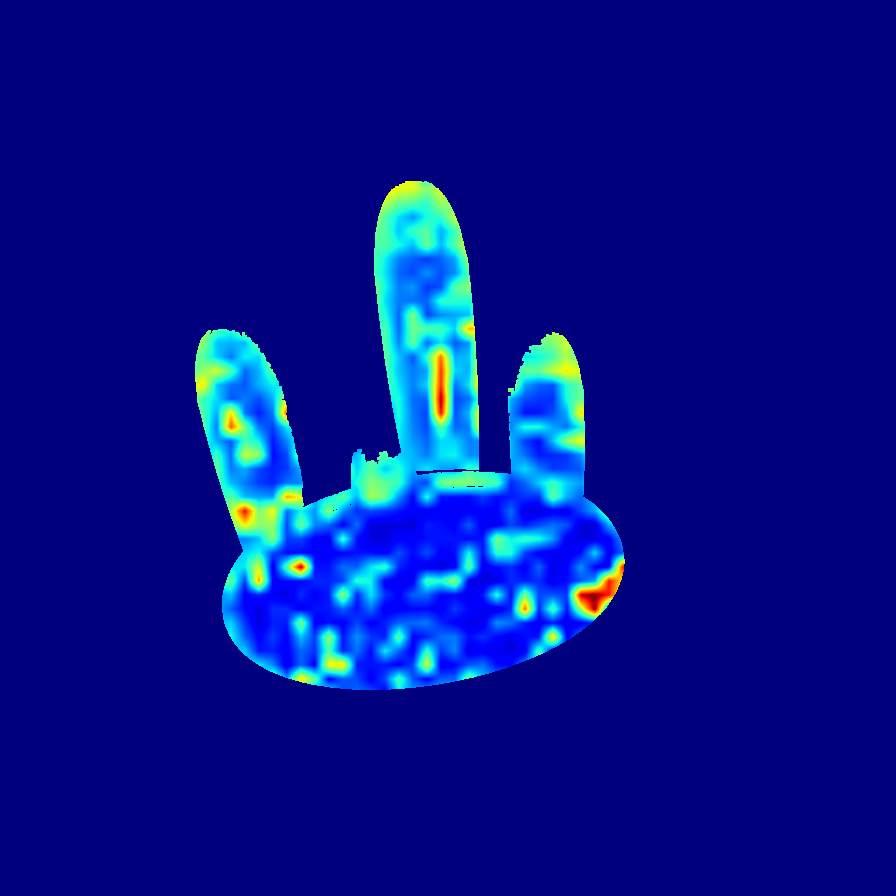} \\
            
            & \rotatebox{90}{\quad $v_{3}$} &
            \includegraphics[width=0.08\linewidth,angle=180,origin=c]{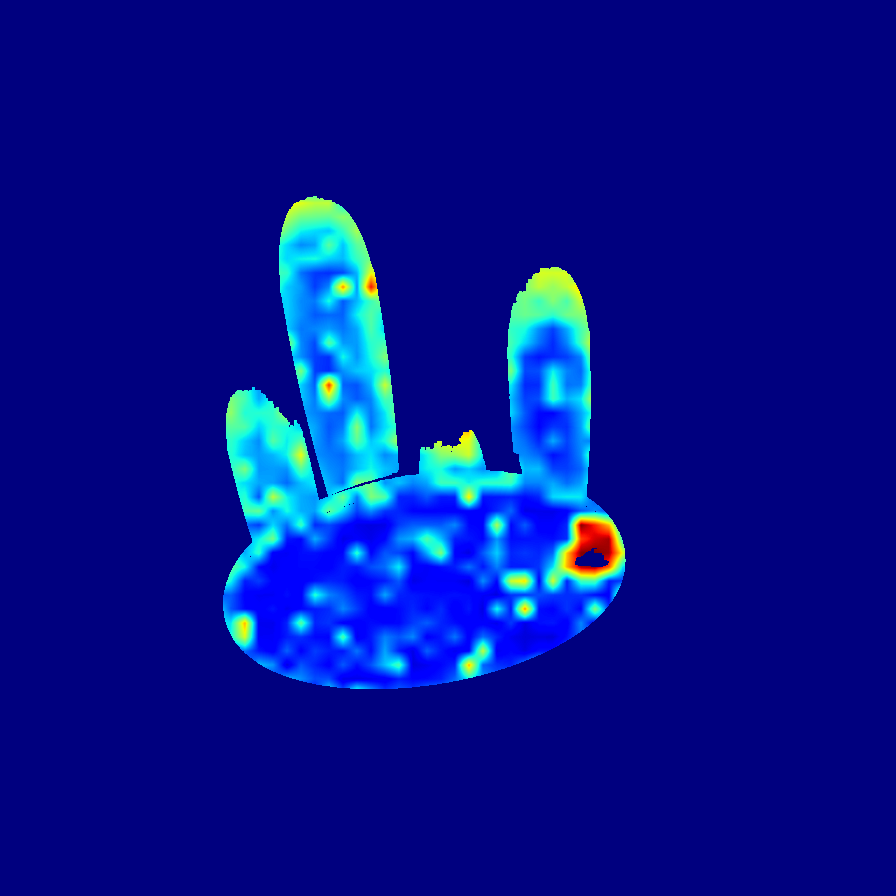} & 
            \includegraphics[width=0.08\linewidth,angle=180,origin=c]{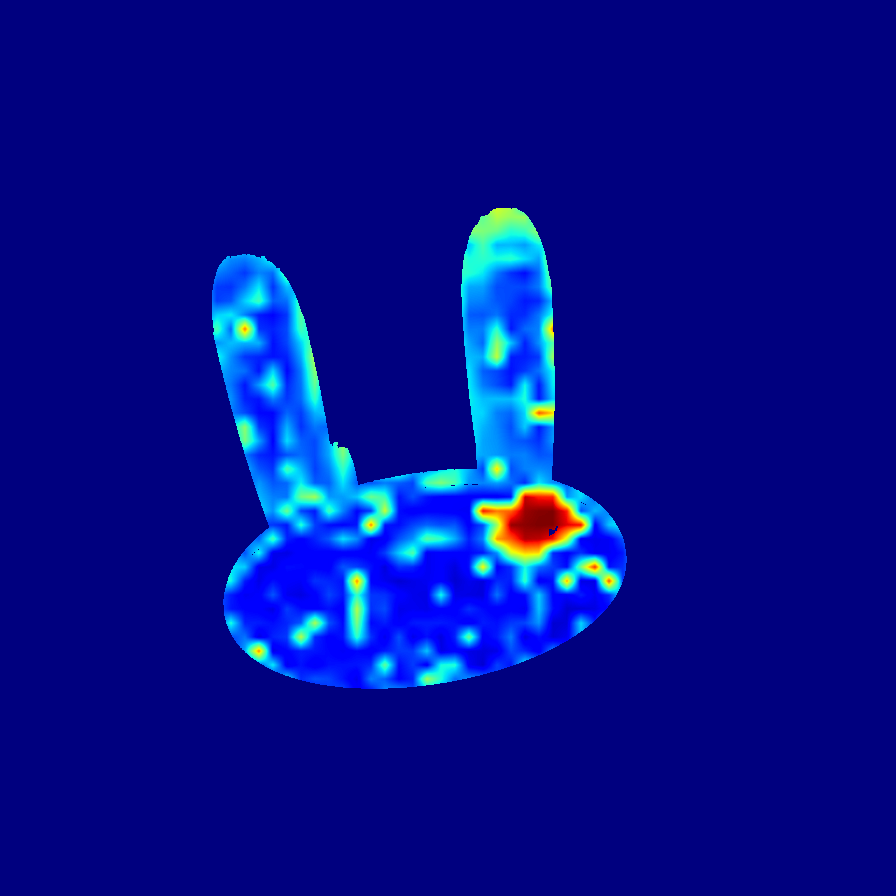} & 
            \includegraphics[width=0.08\linewidth,angle=180,origin=c]{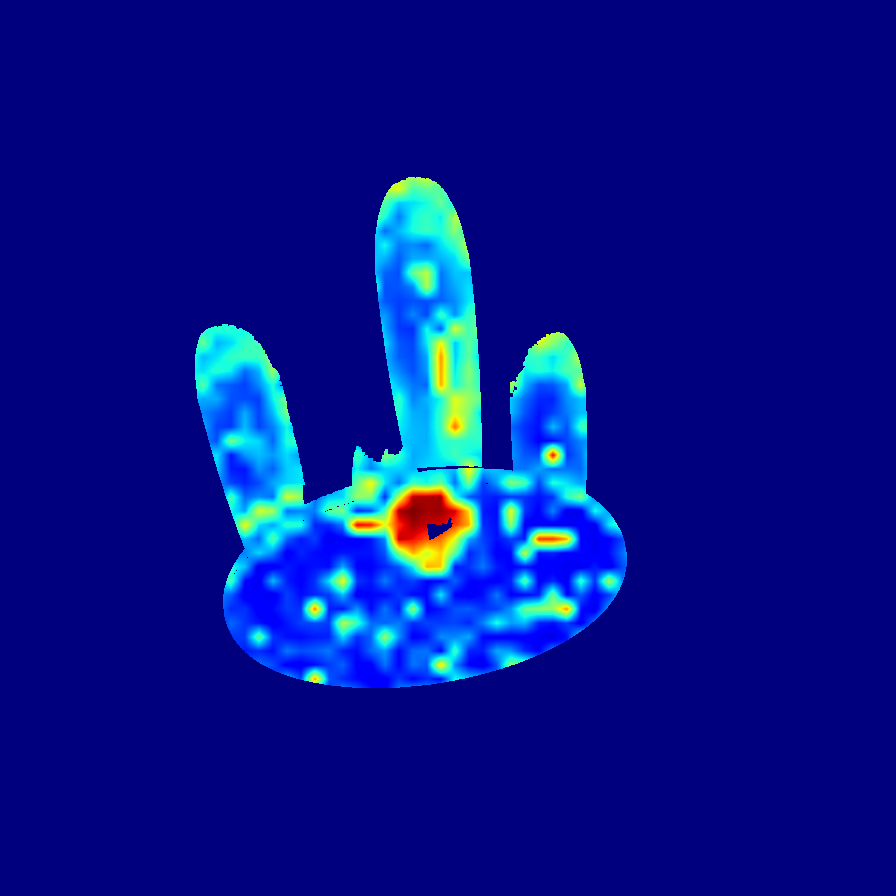} & 
            \includegraphics[width=0.08\linewidth,angle=180,origin=c]{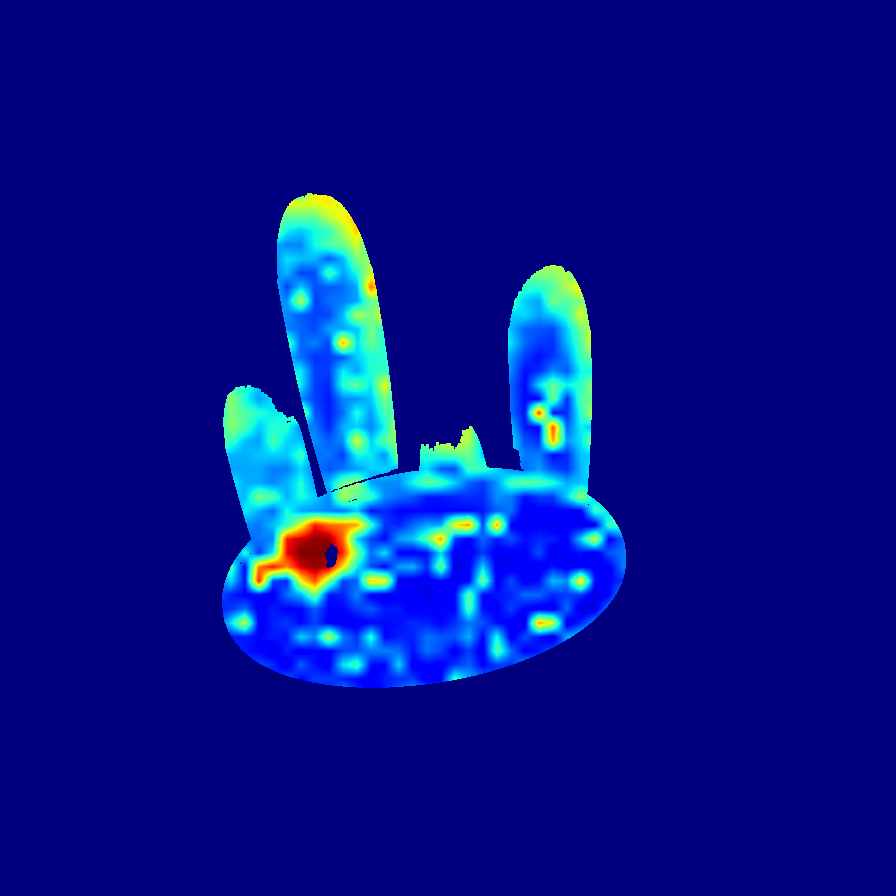} & 
            \includegraphics[width=0.08\linewidth,angle=180,origin=c]{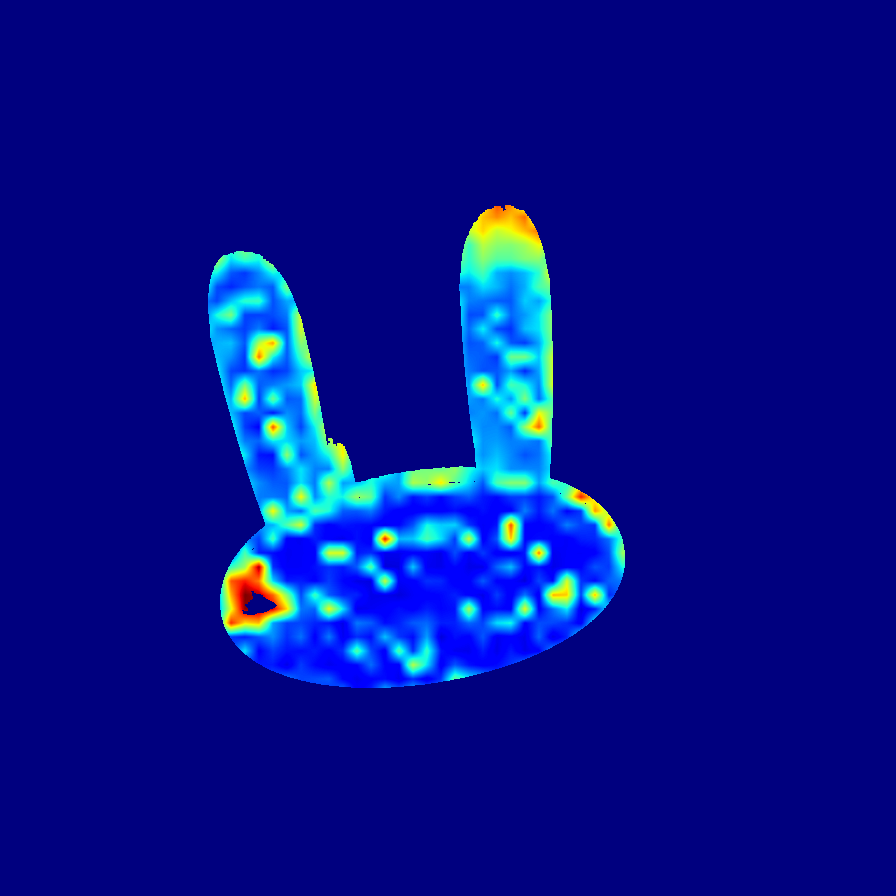} & 
            \includegraphics[width=0.08\linewidth,angle=180,origin=c]{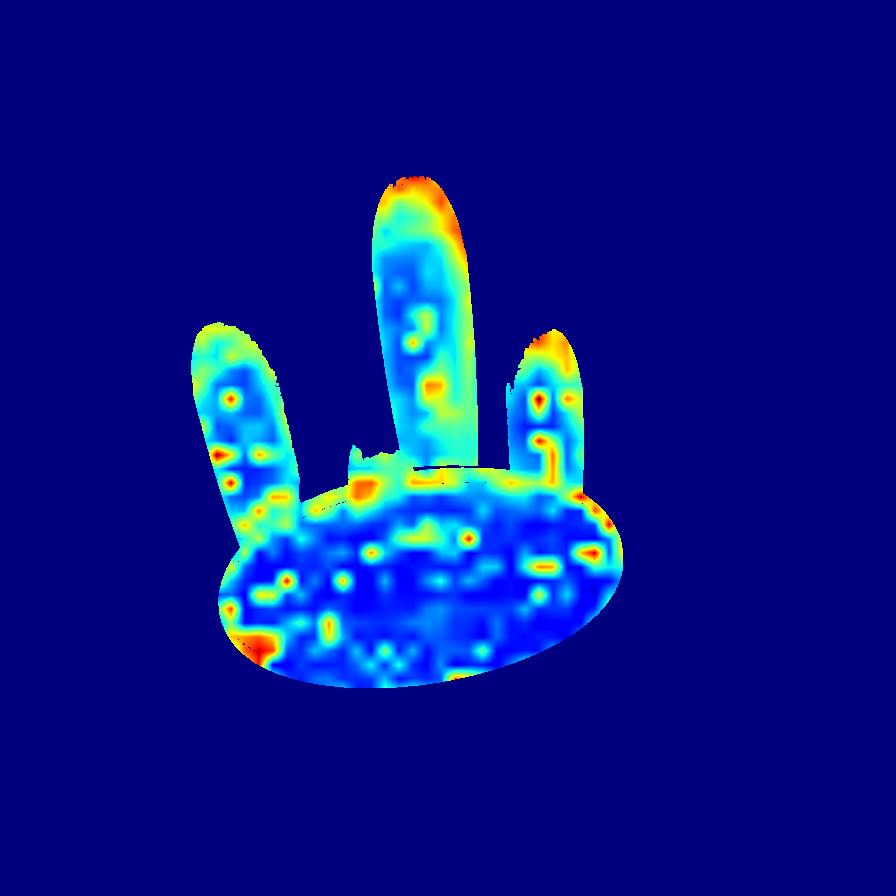} & 
            \includegraphics[width=0.08\linewidth,angle=180,origin=c]{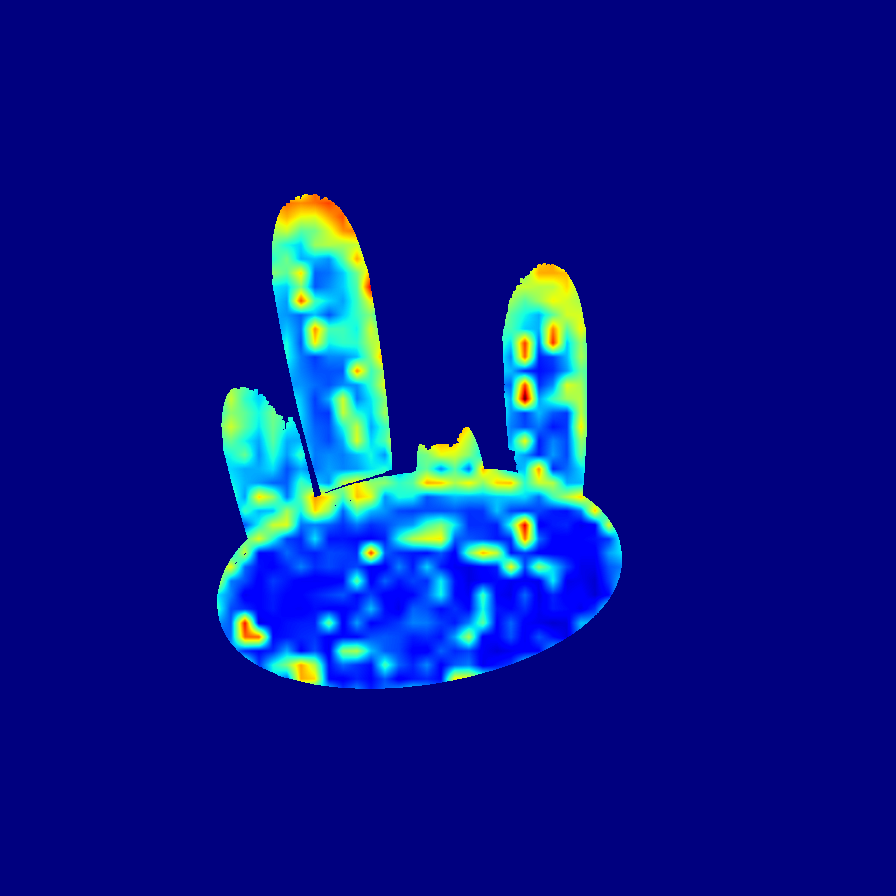} & 
            \includegraphics[width=0.08\linewidth,angle=180,origin=c]{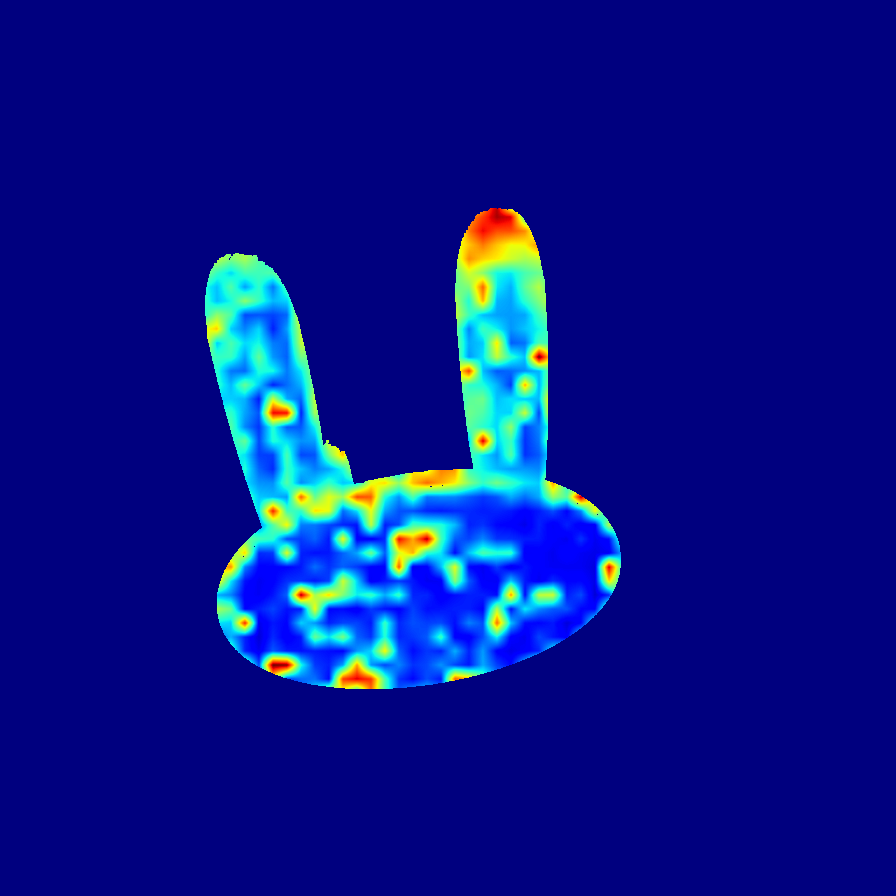} & 
            \includegraphics[width=0.08\linewidth,angle=180,origin=c]{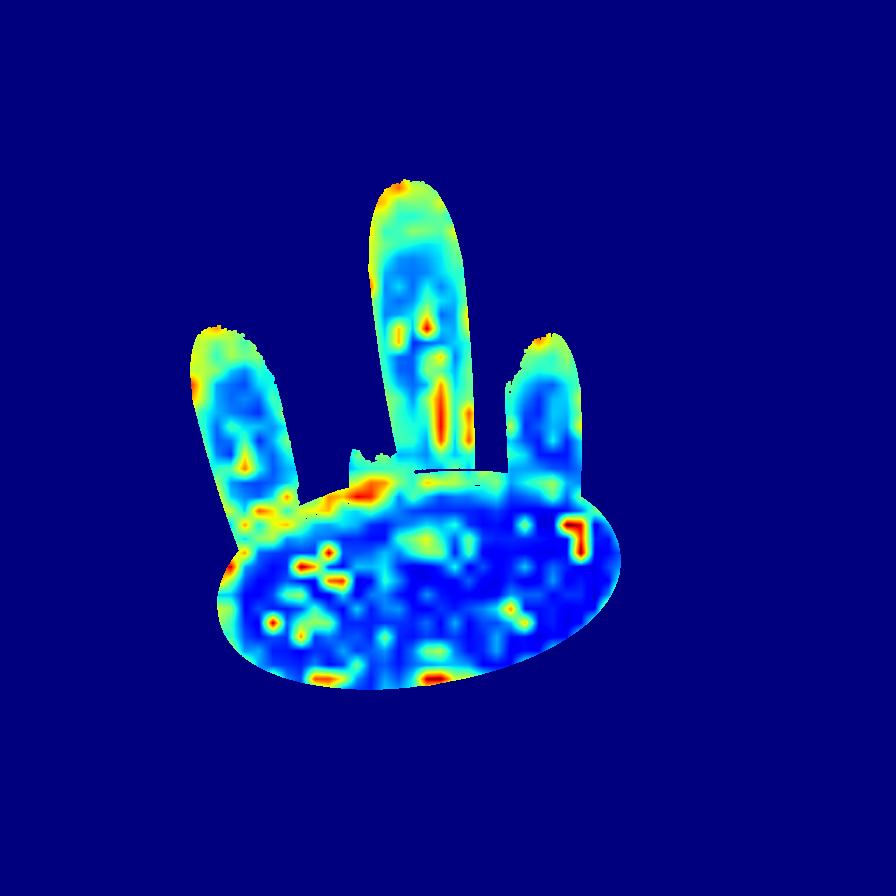} & 
            \includegraphics[width=0.08\linewidth,angle=180,origin=c]{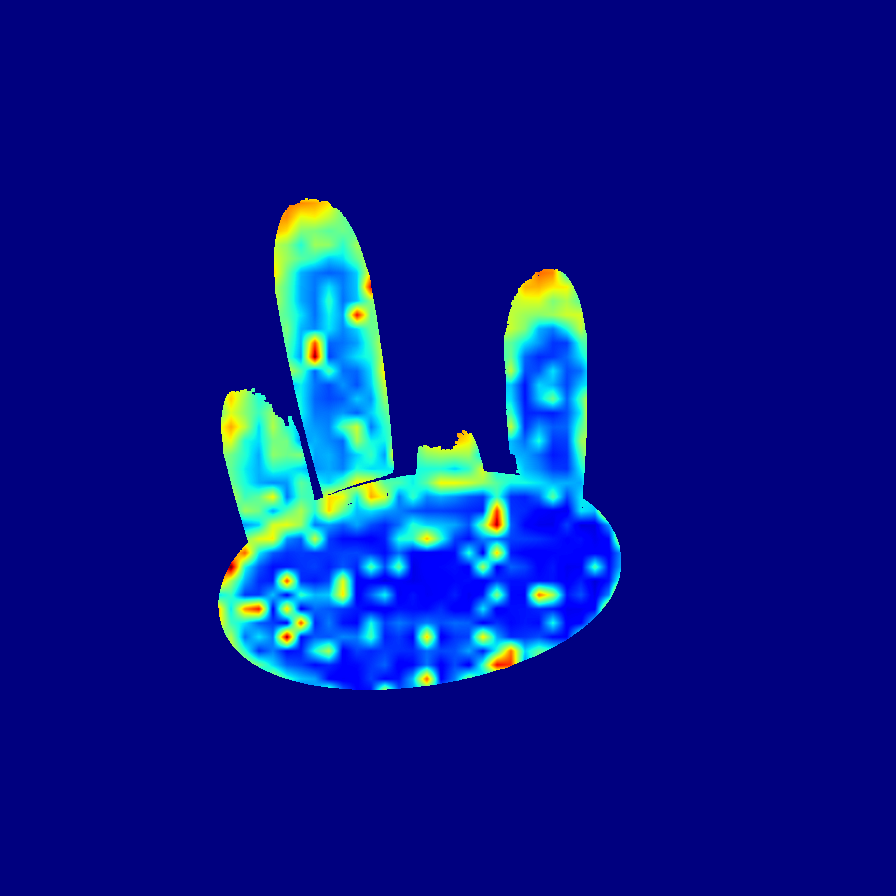} & 
            \includegraphics[width=0.08\linewidth,angle=180,origin=c]{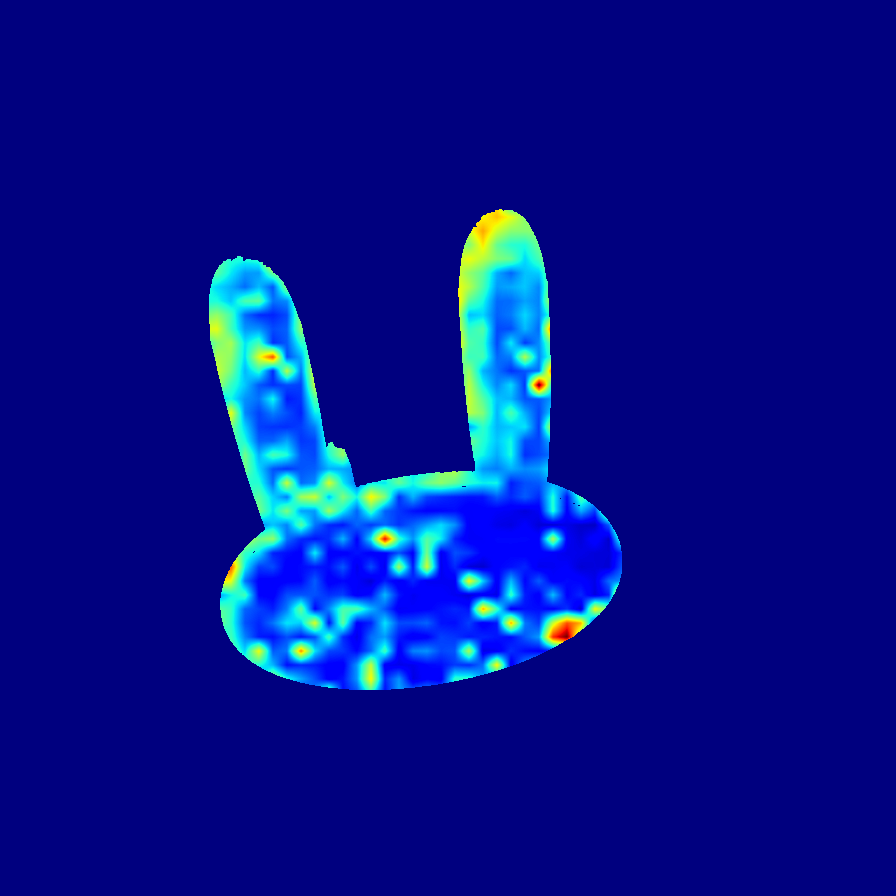} &
            \includegraphics[width=0.08\linewidth,angle=180,origin=c]{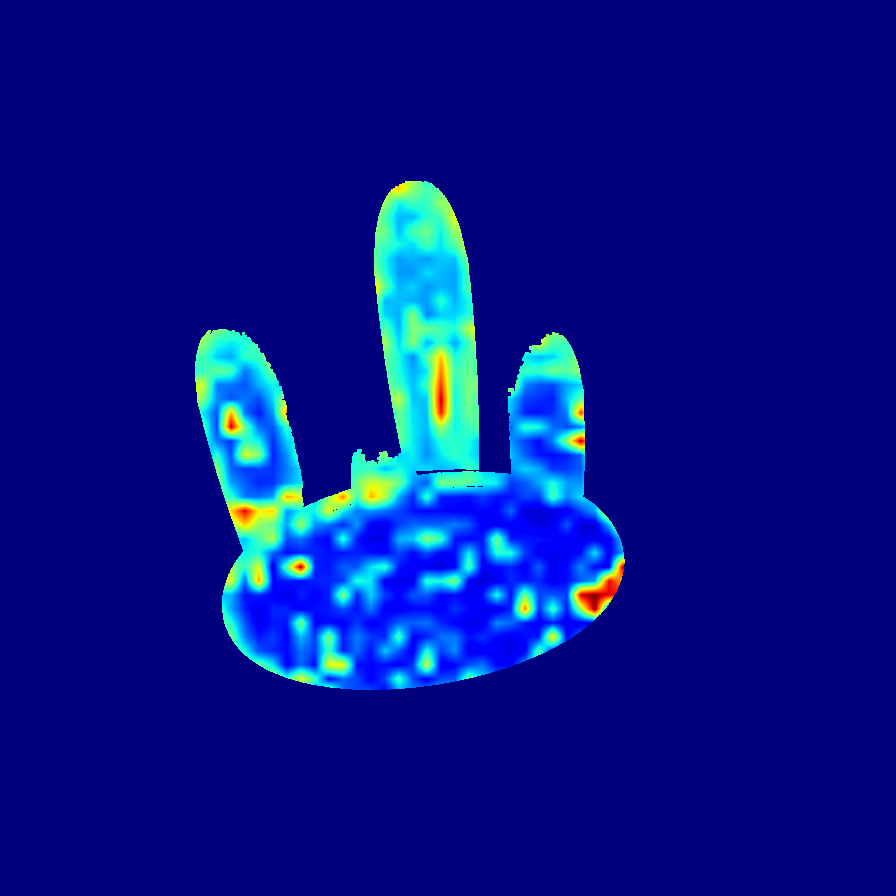} \\
            
            & \rotatebox{90}{\quad $v_{4}$} &
            \includegraphics[width=0.08\linewidth,angle=180,origin=c]{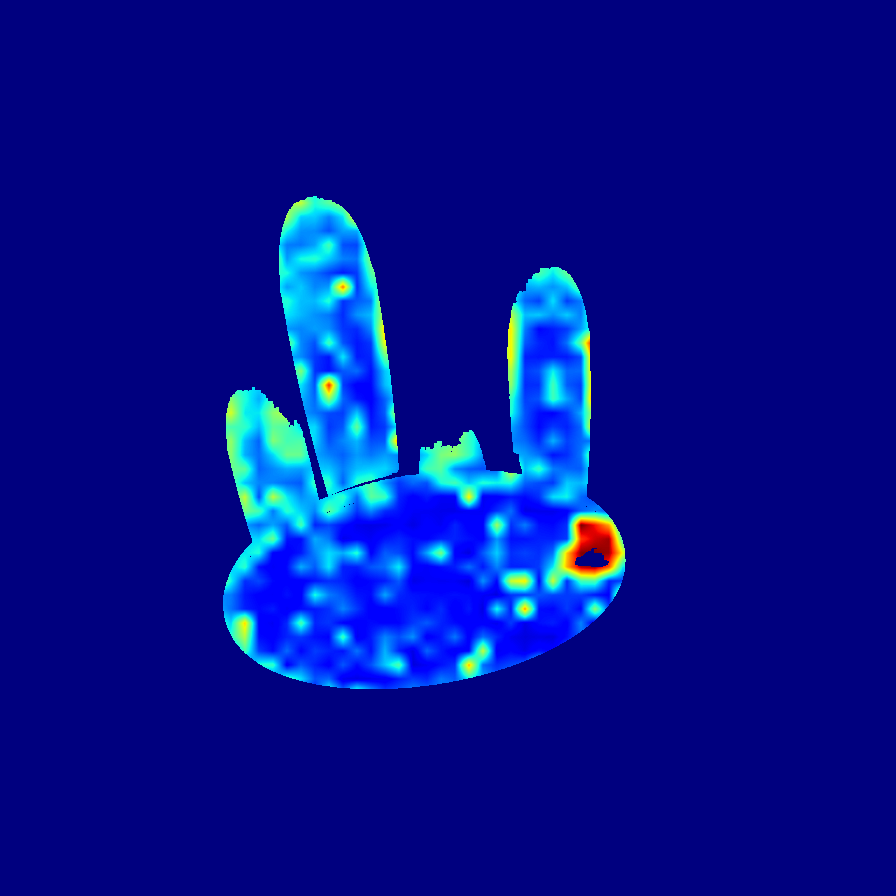} & 
            \includegraphics[width=0.08\linewidth,angle=180,origin=c]{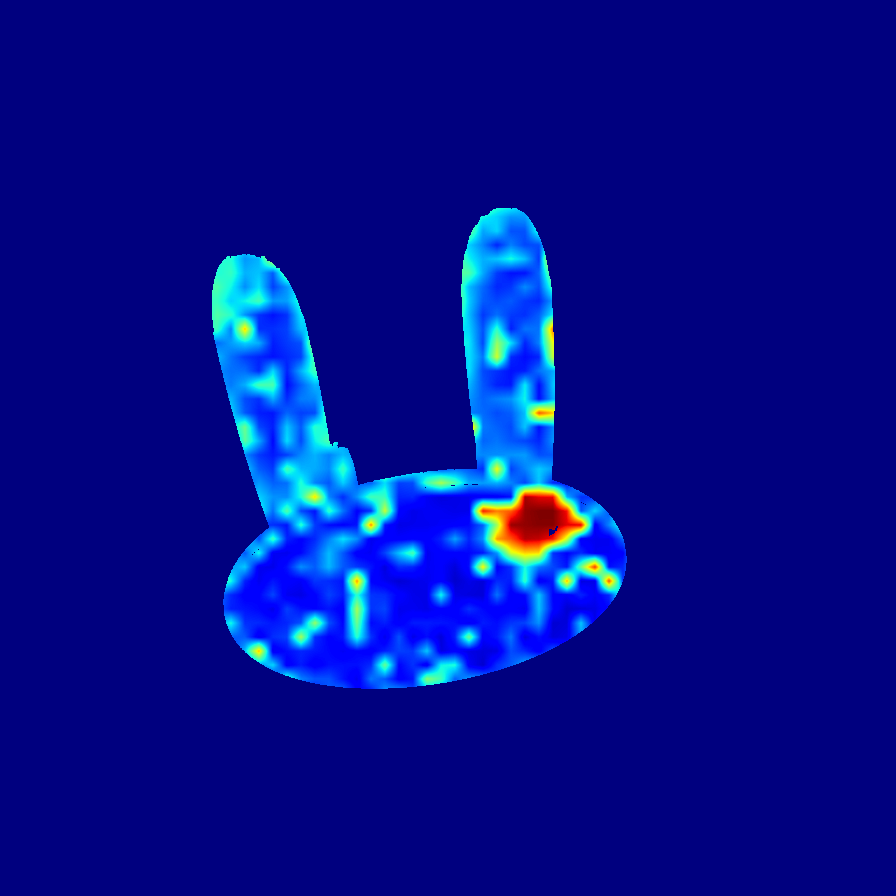} & 
            \includegraphics[width=0.08\linewidth,angle=180,origin=c]{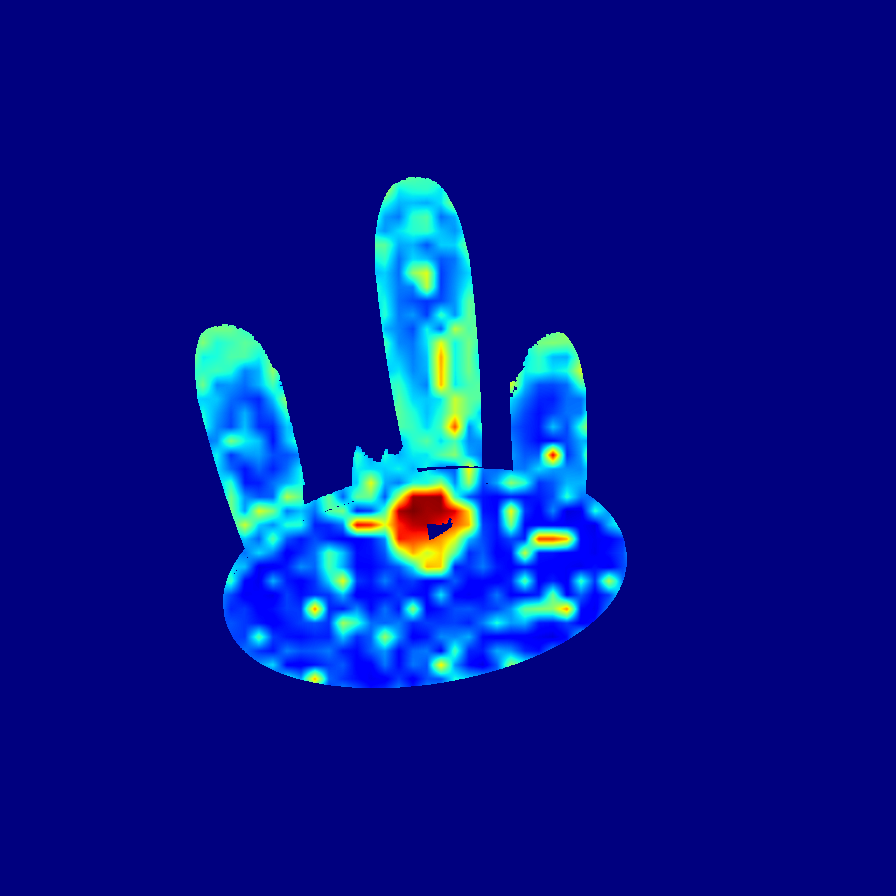} & 
            \includegraphics[width=0.08\linewidth,angle=180,origin=c]{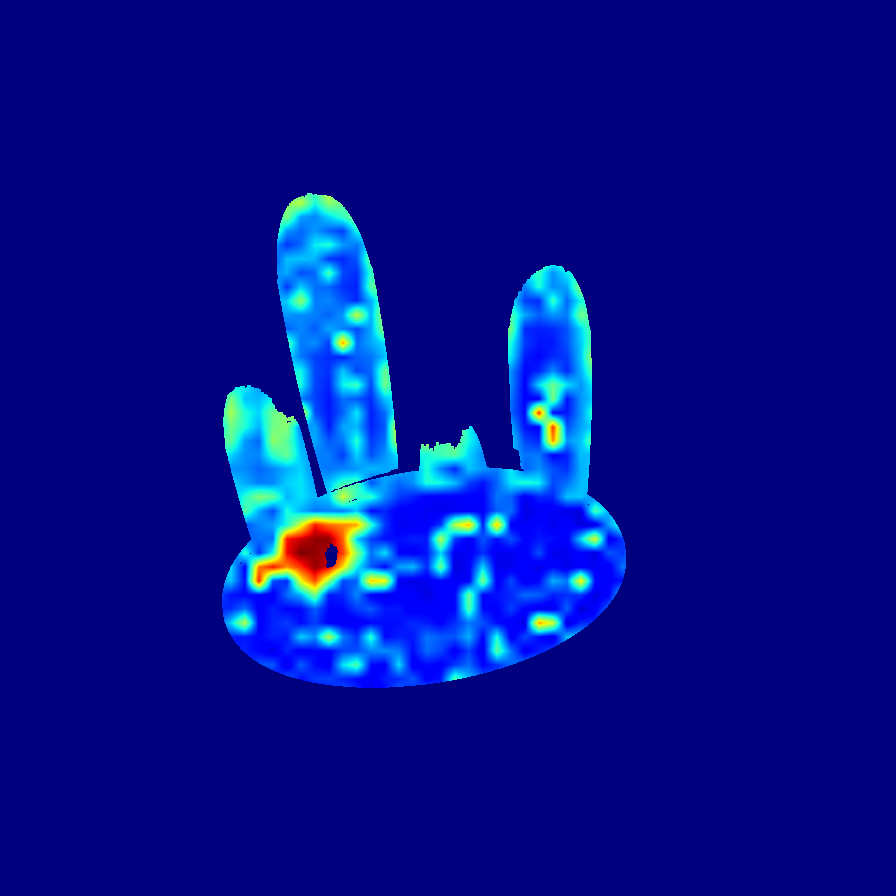} & 
            \includegraphics[width=0.08\linewidth,angle=180,origin=c]{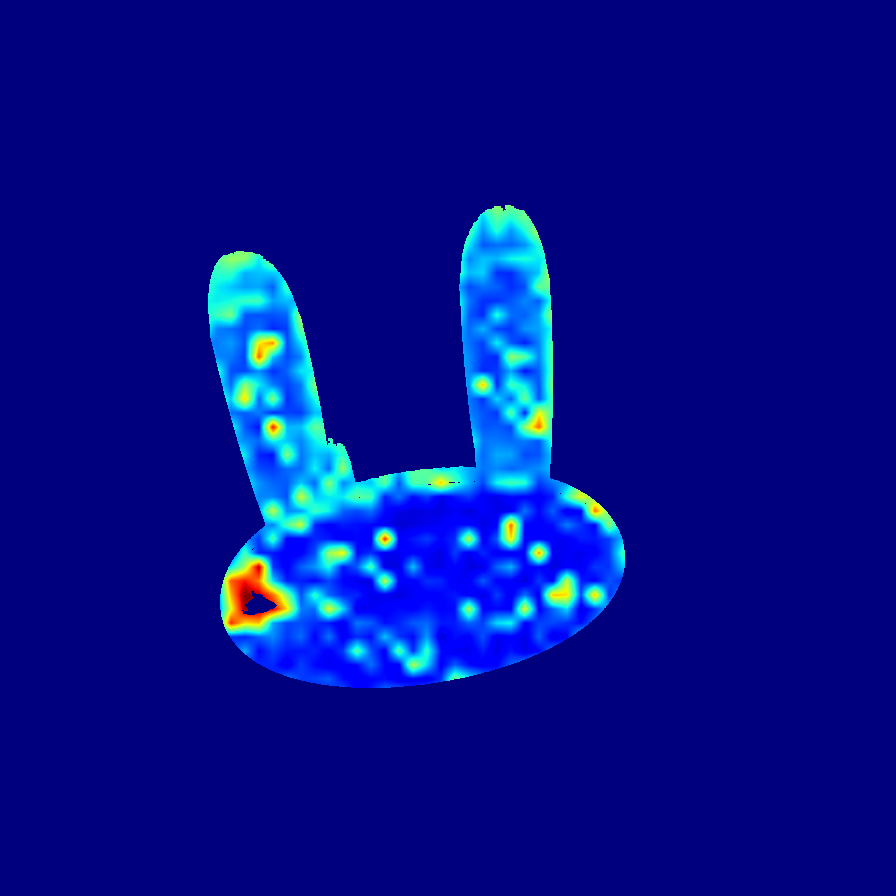} & 
            \includegraphics[width=0.08\linewidth,angle=180,origin=c]{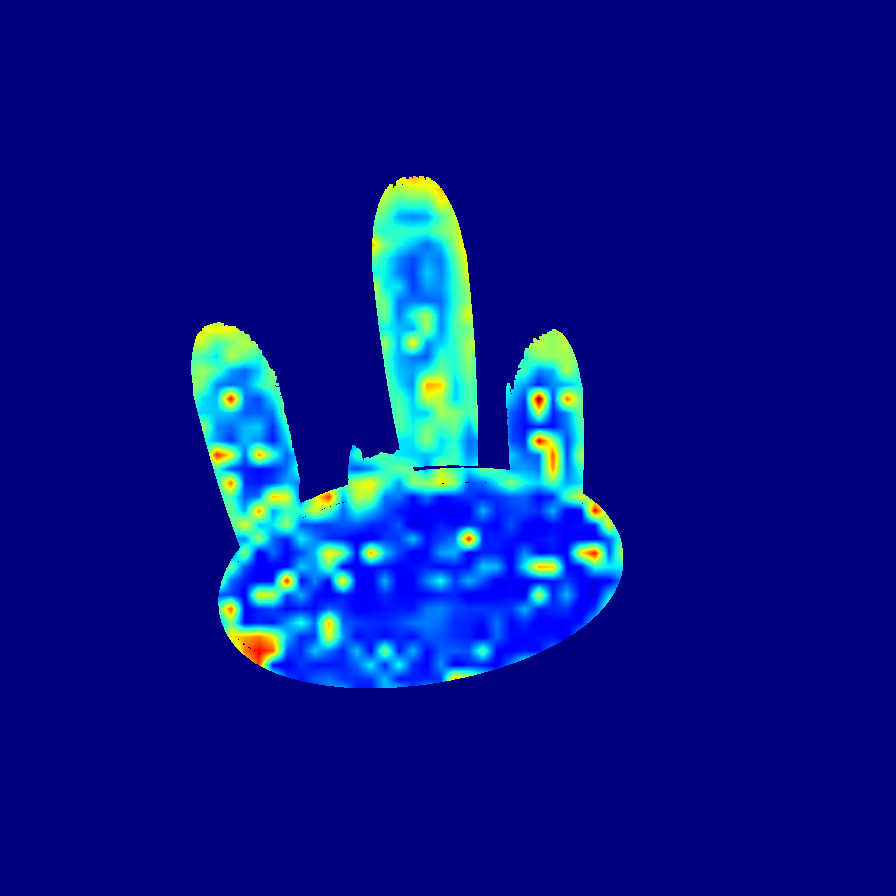} & 
            \includegraphics[width=0.08\linewidth,angle=180,origin=c]{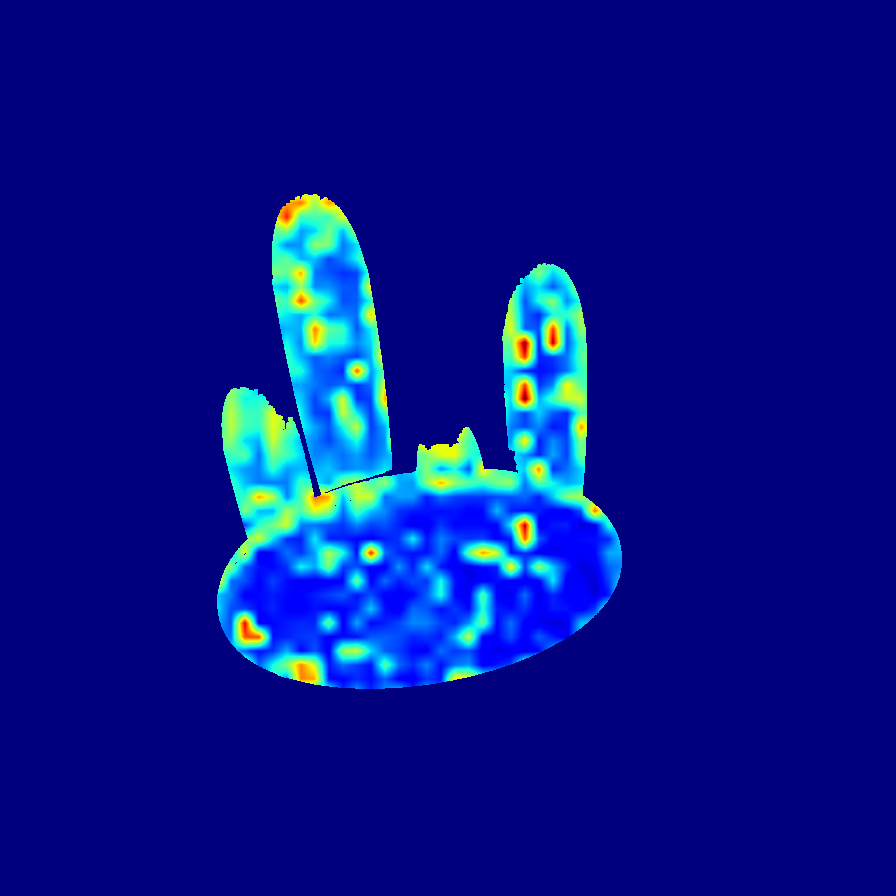} & 
            \includegraphics[width=0.08\linewidth,angle=180,origin=c]{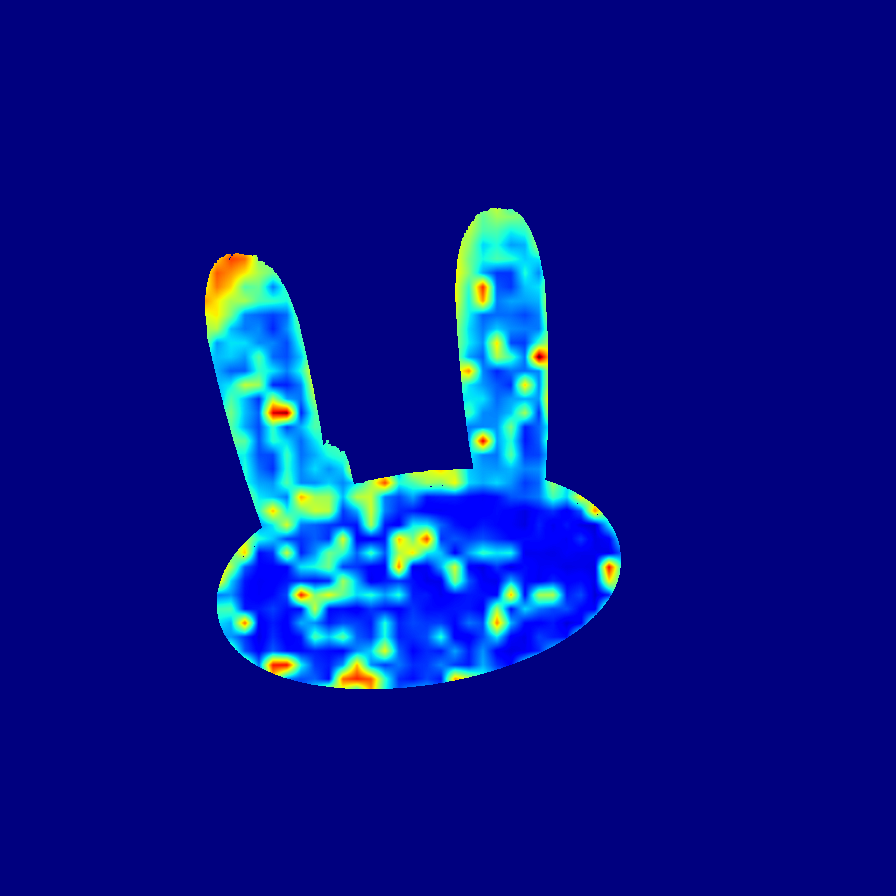} & 
            \includegraphics[width=0.08\linewidth,angle=180,origin=c]{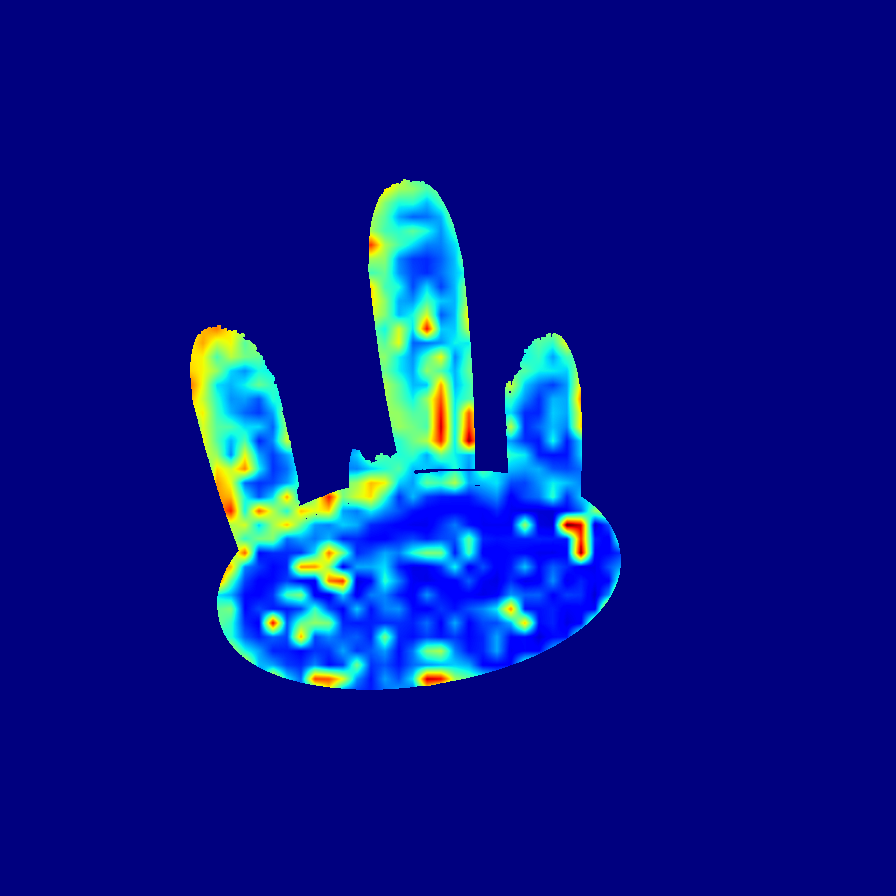} & 
            \includegraphics[width=0.08\linewidth,angle=180,origin=c]{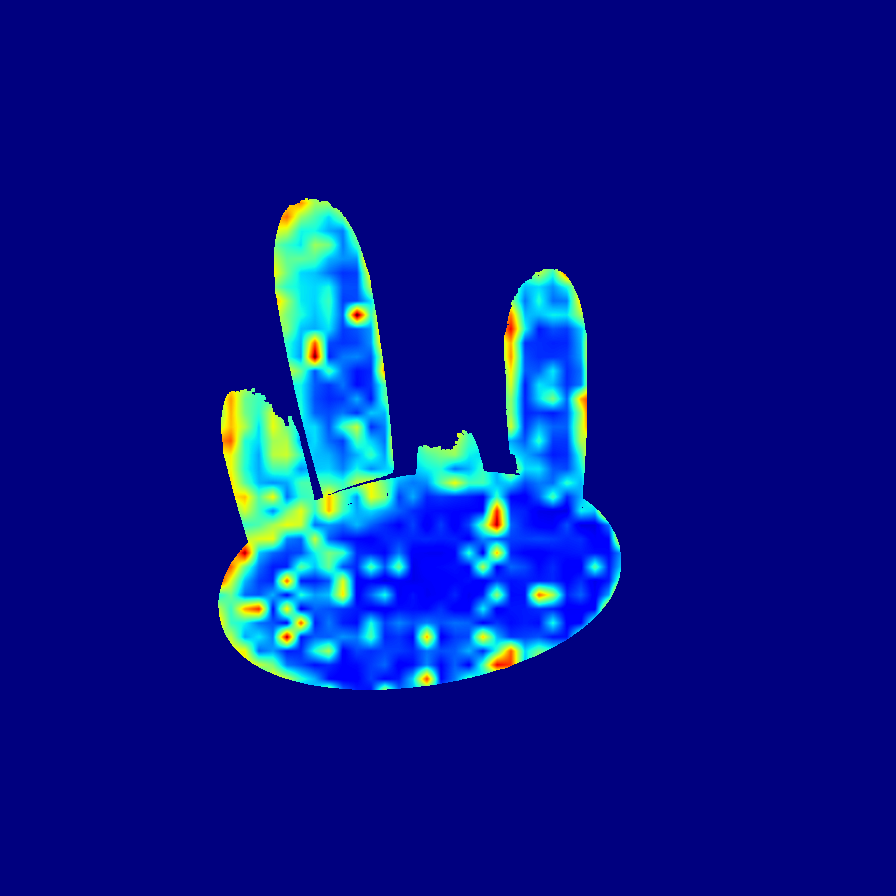} & 
            \includegraphics[width=0.08\linewidth,angle=180,origin=c]{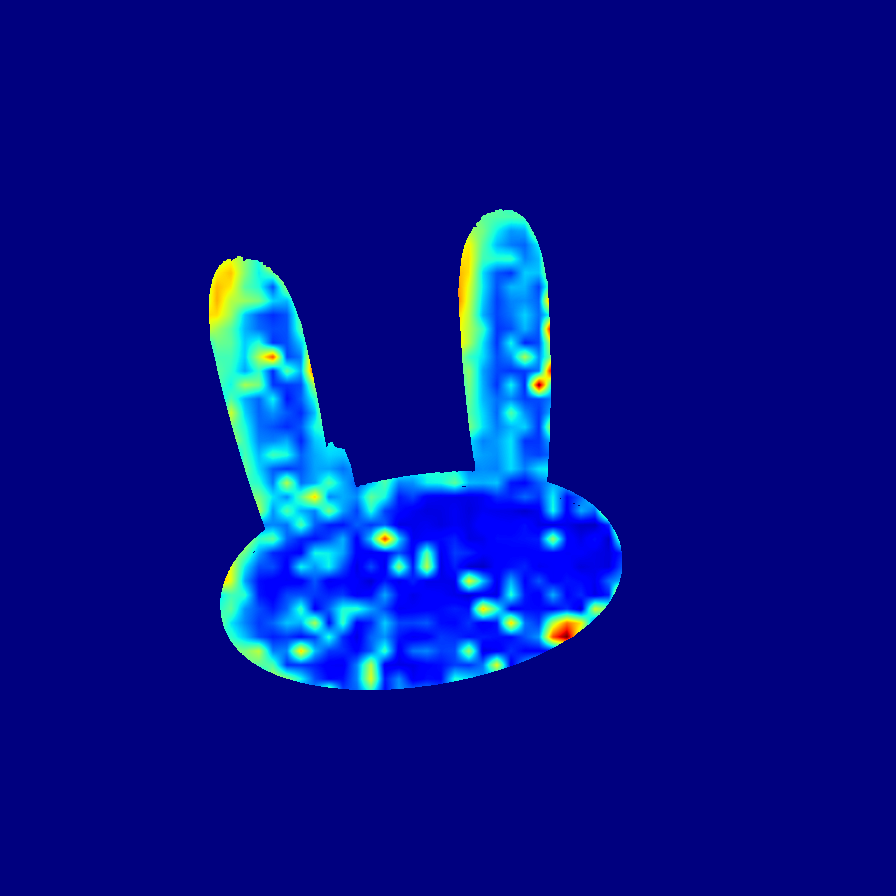} &
            \includegraphics[width=0.08\linewidth,angle=180,origin=c]{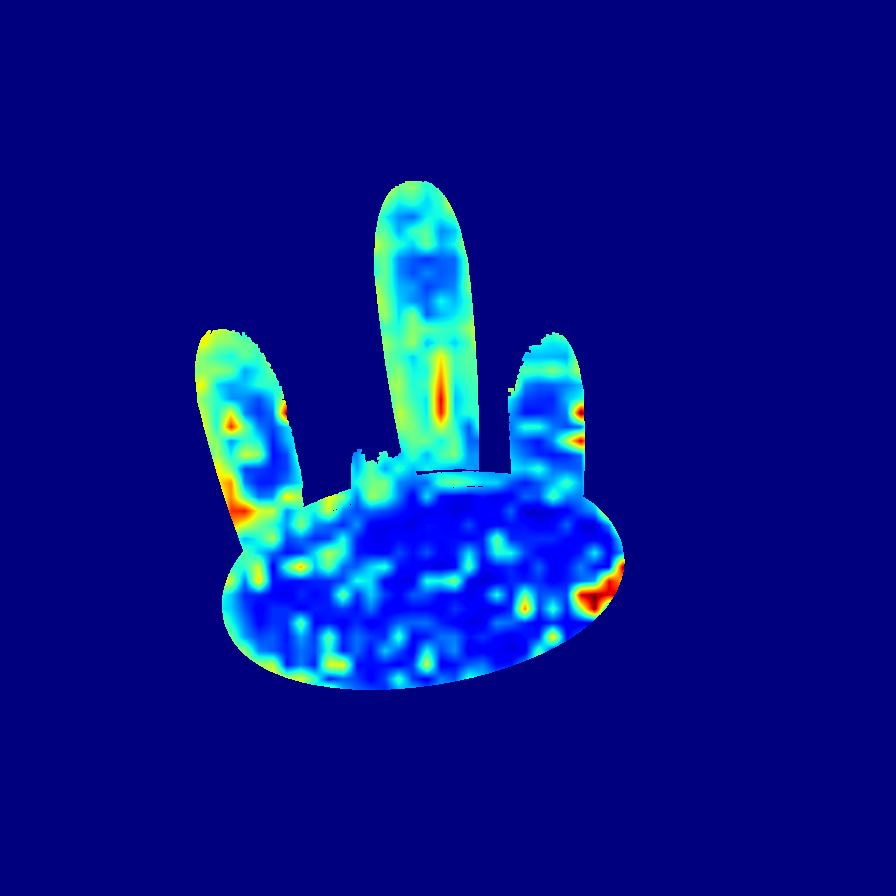} \\

            & \rotatebox{90}{\quad $v_{5}$} &
            \includegraphics[width=0.08\linewidth,angle=180,origin=c]{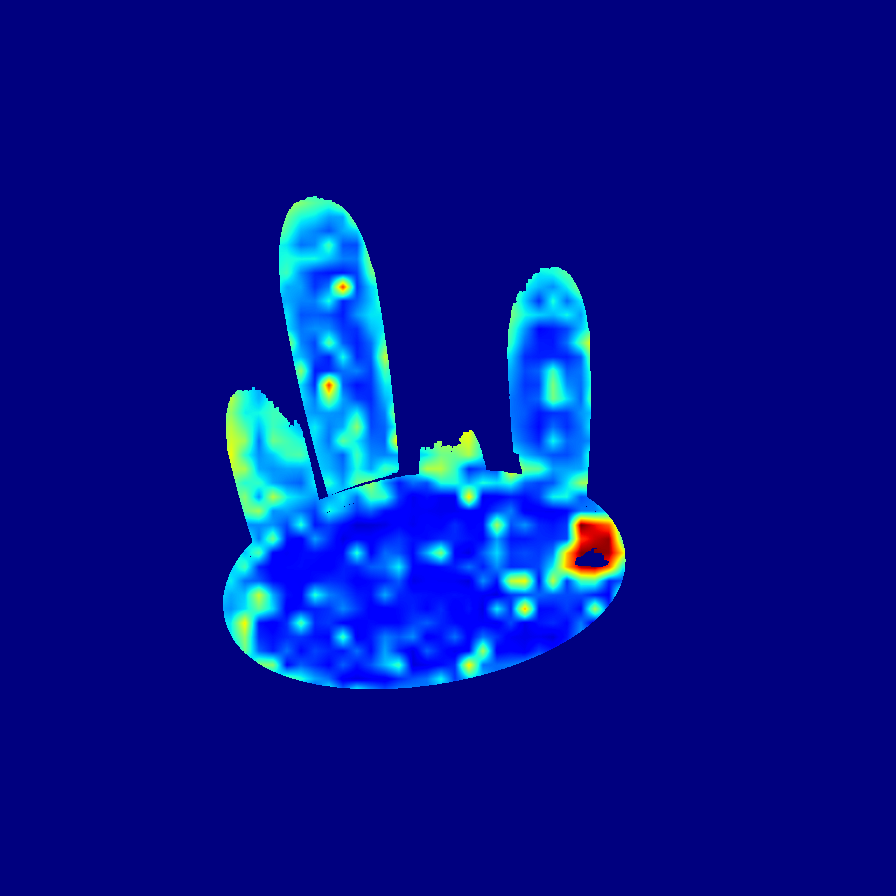} & 
            \includegraphics[width=0.08\linewidth,angle=180,origin=c]{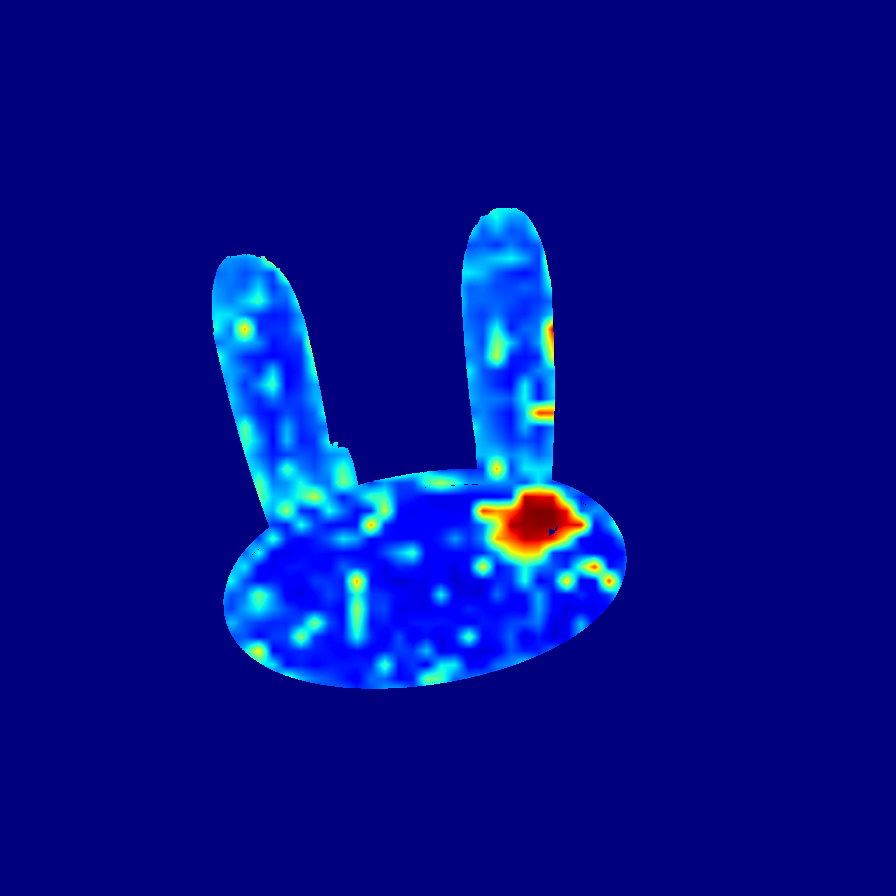} & 
            \includegraphics[width=0.08\linewidth,angle=180,origin=c]{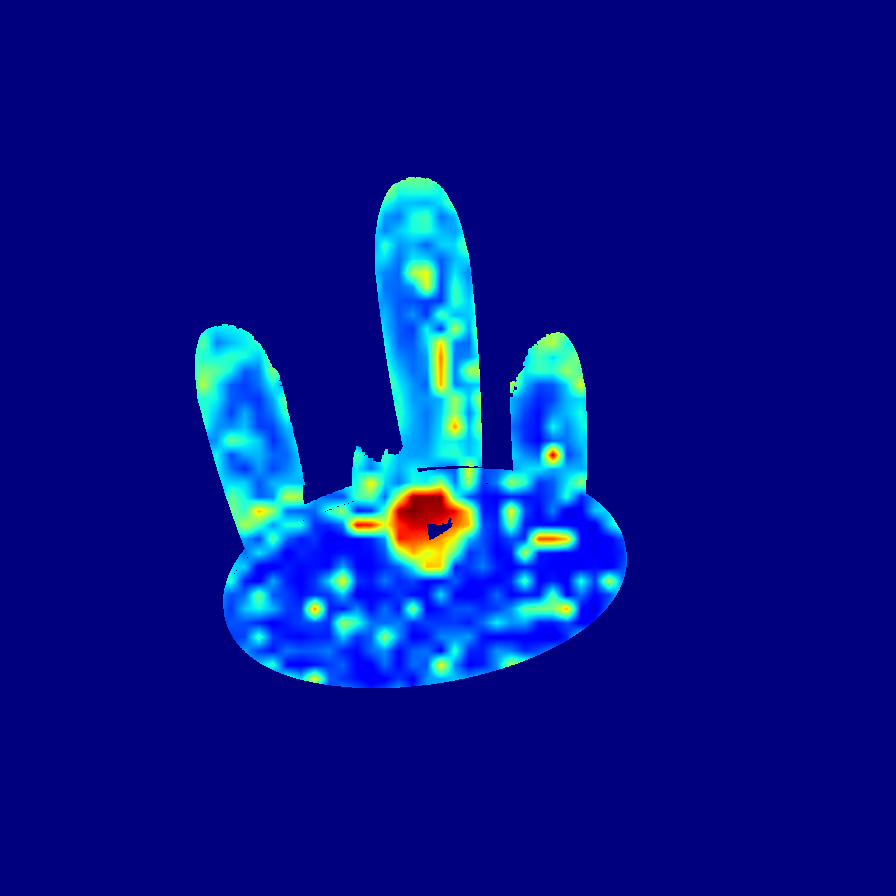} & 
            \includegraphics[width=0.08\linewidth,angle=180,origin=c]{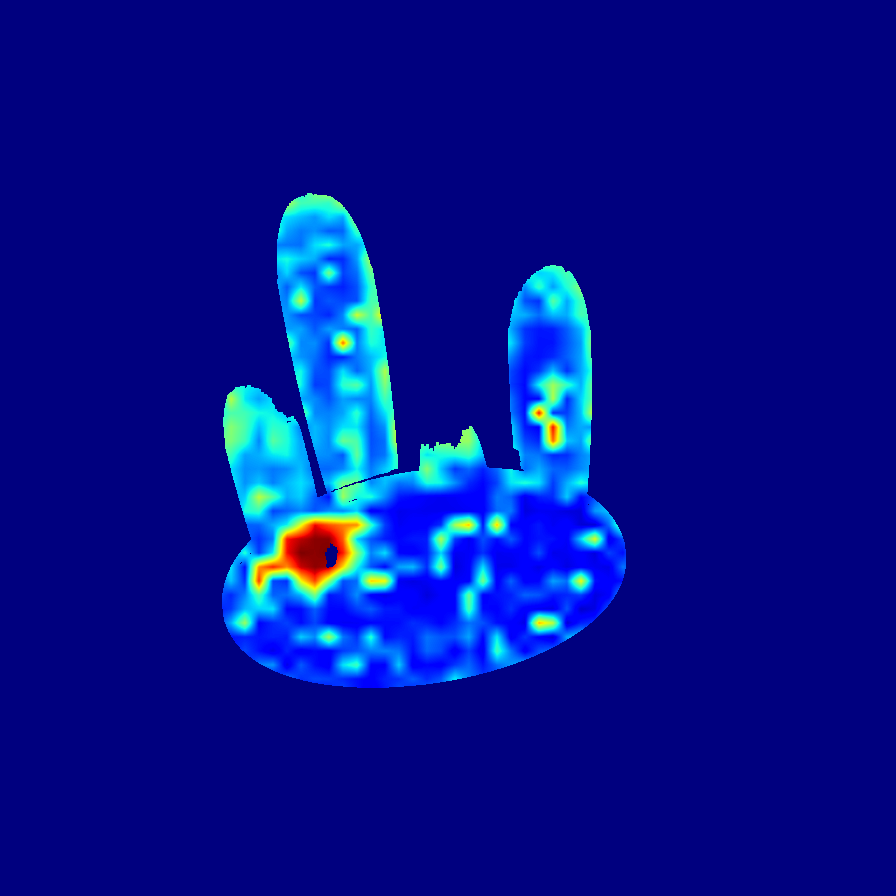} & 
            \includegraphics[width=0.08\linewidth,angle=180,origin=c]{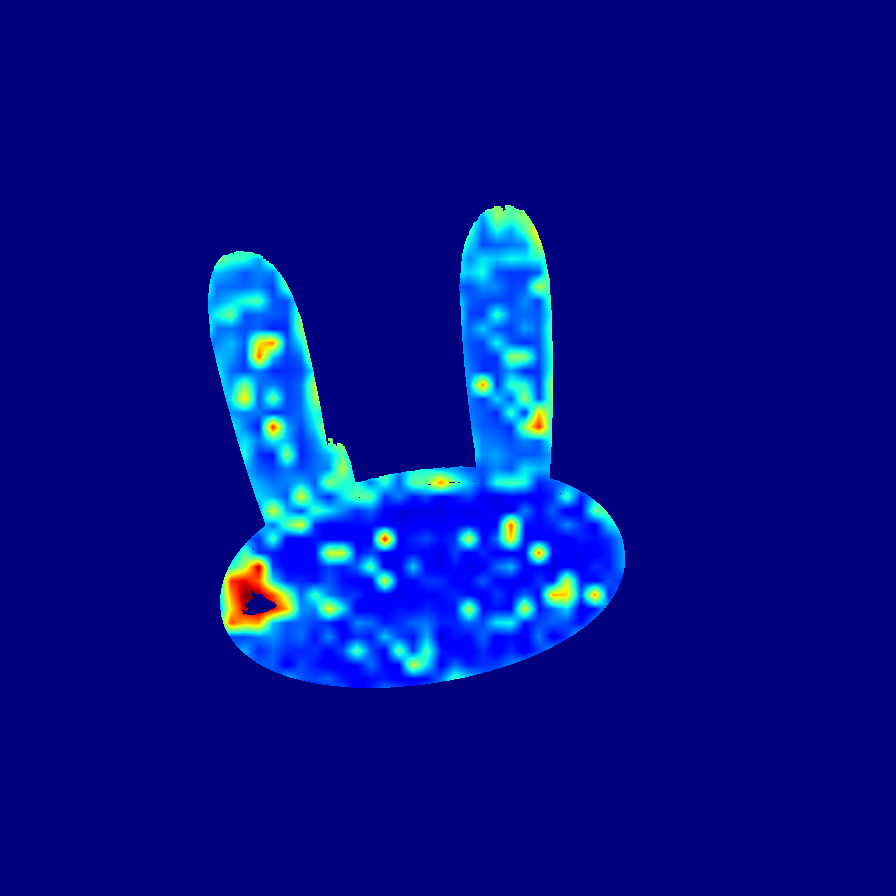} & 
            \includegraphics[width=0.08\linewidth,angle=180,origin=c]{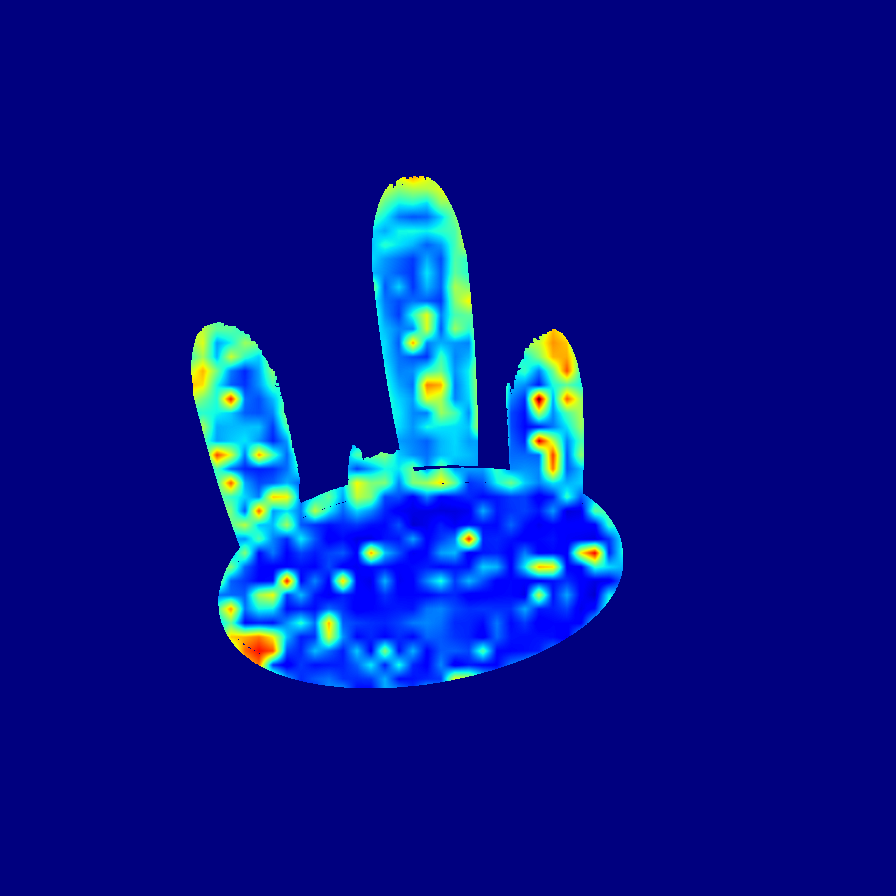} & 
            \includegraphics[width=0.08\linewidth,angle=180,origin=c]{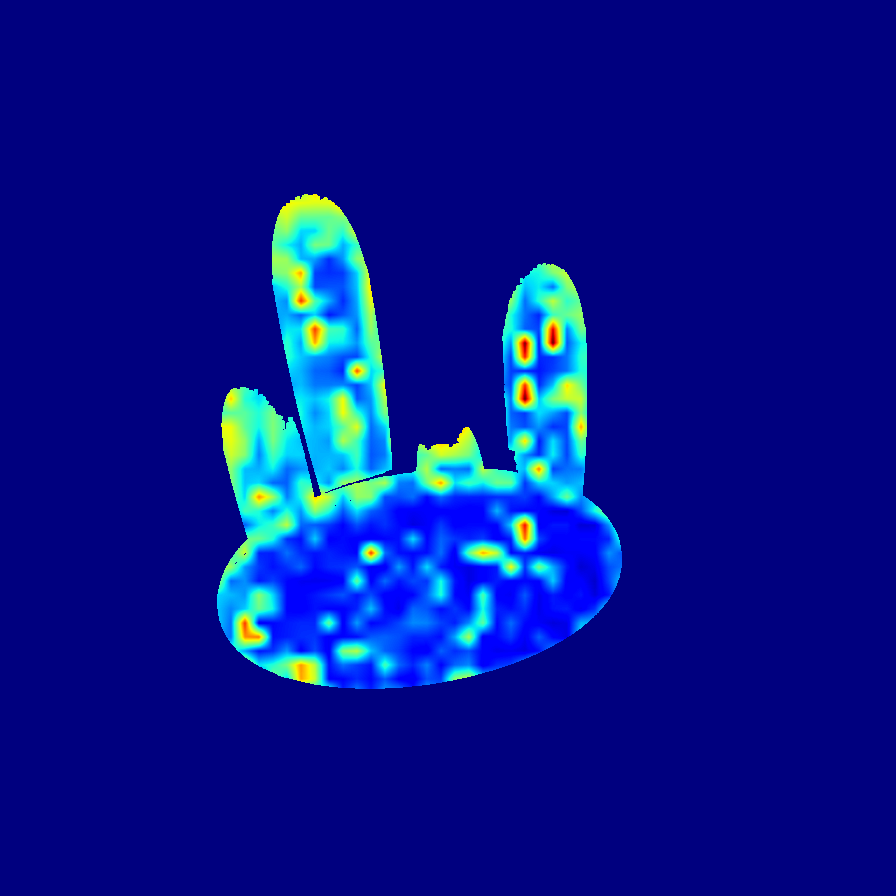} & 
            \includegraphics[width=0.08\linewidth,angle=180,origin=c]{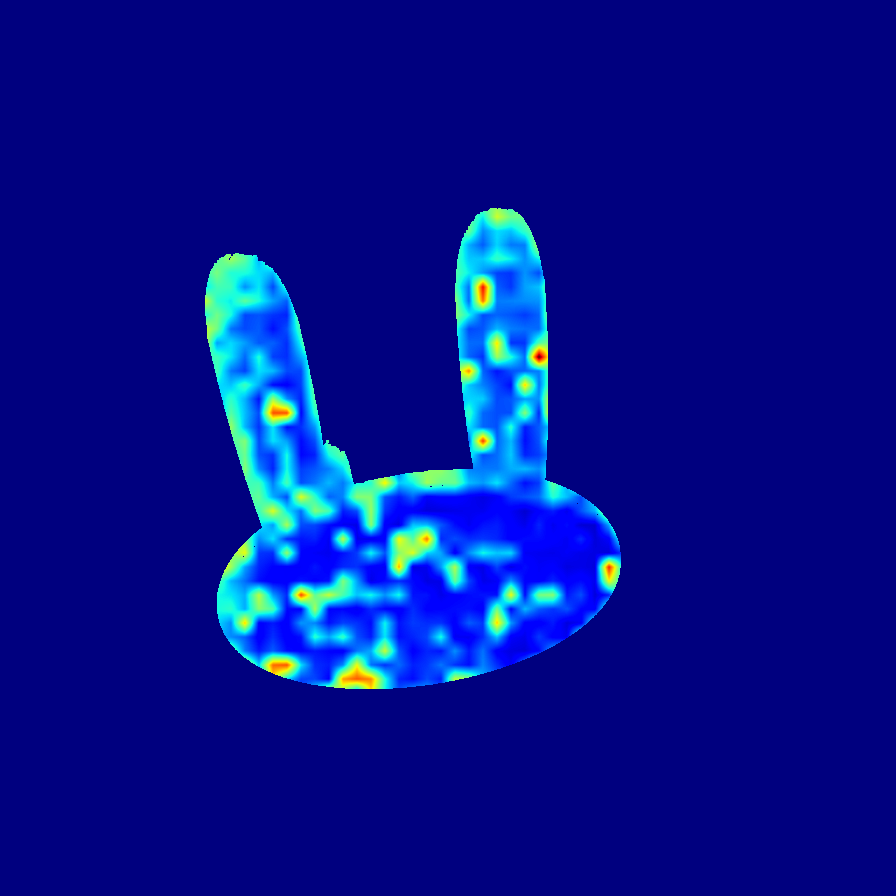} & 
            \includegraphics[width=0.08\linewidth,angle=180,origin=c]{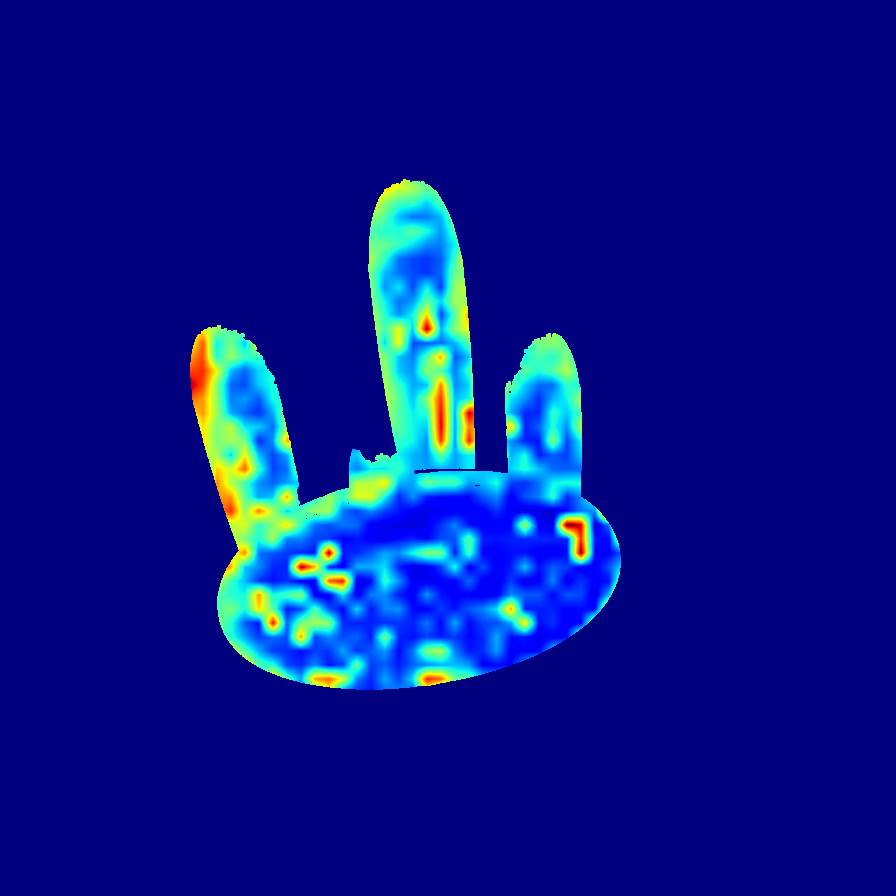} & 
            \includegraphics[width=0.08\linewidth,angle=180,origin=c]{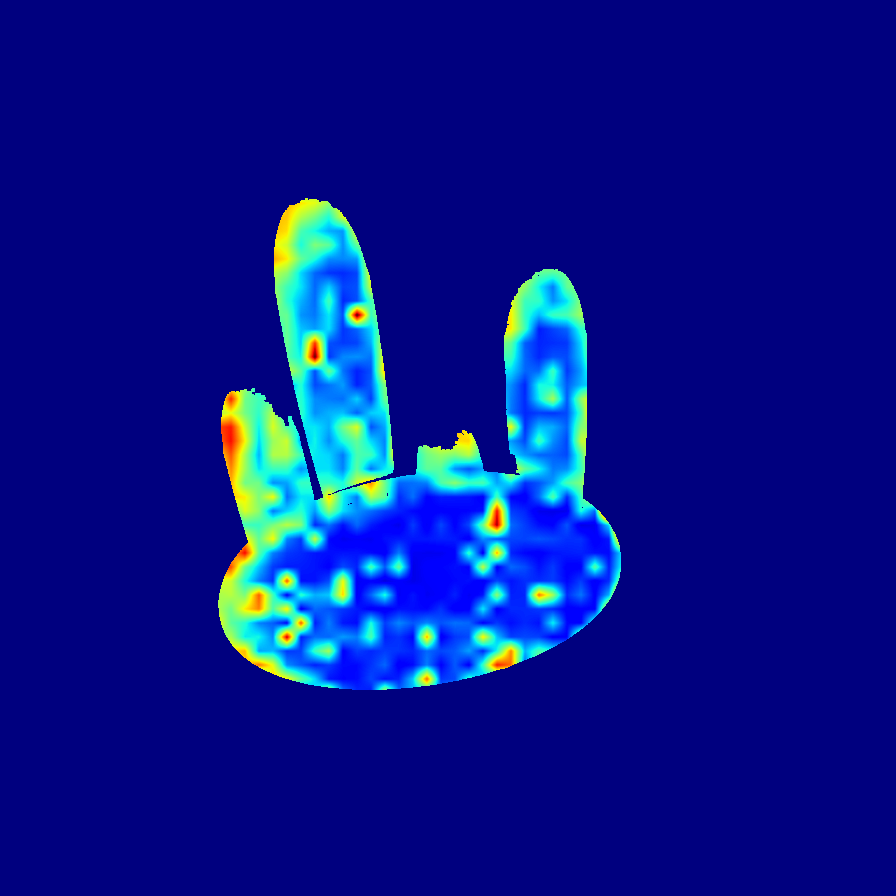} & 
            \includegraphics[width=0.08\linewidth,angle=180,origin=c]{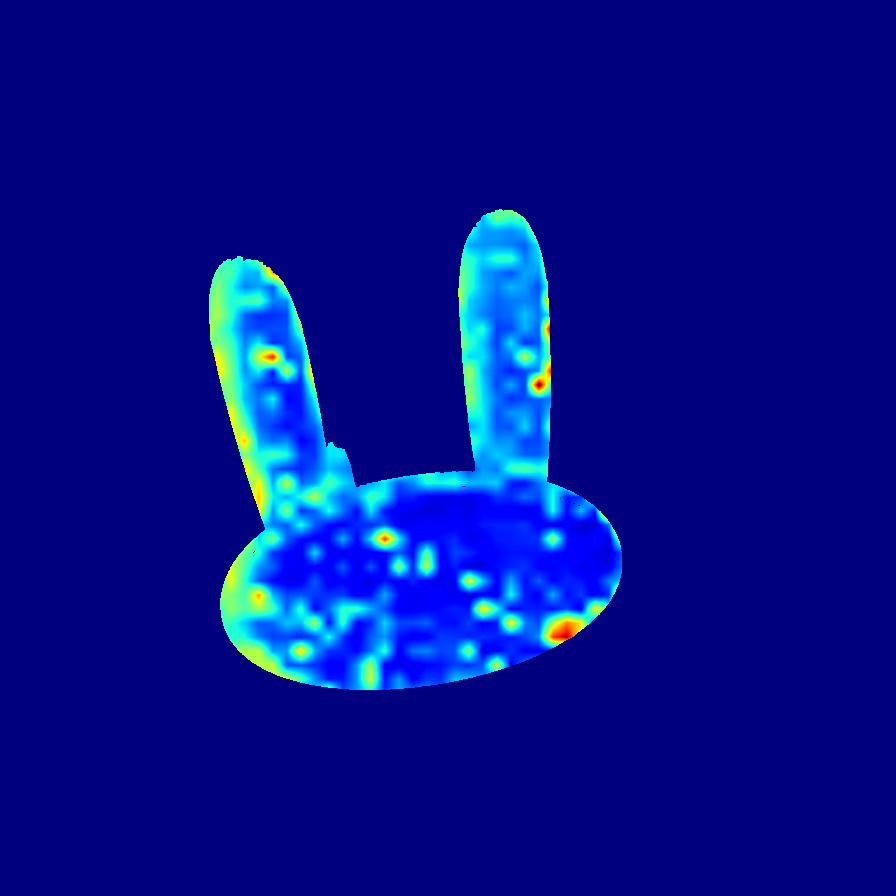} &
            \includegraphics[width=0.08\linewidth,angle=180,origin=c]{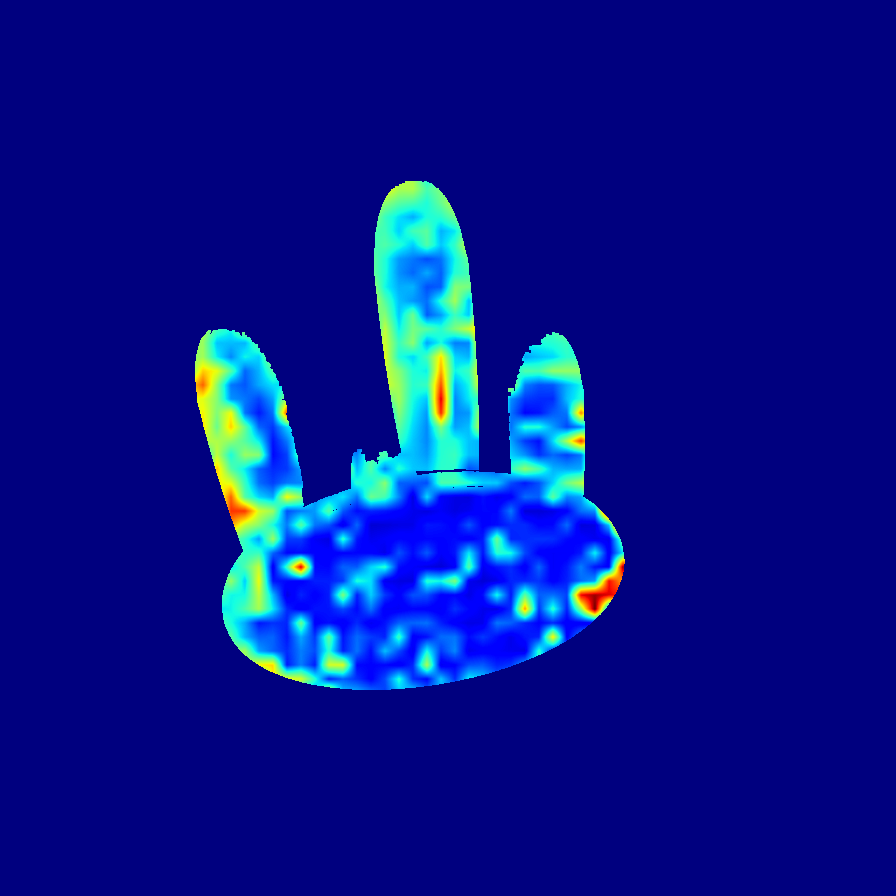} \\

            & \rotatebox{90}{\quad $v_{6}$} &
            \includegraphics[width=0.08\linewidth,angle=180,origin=c]{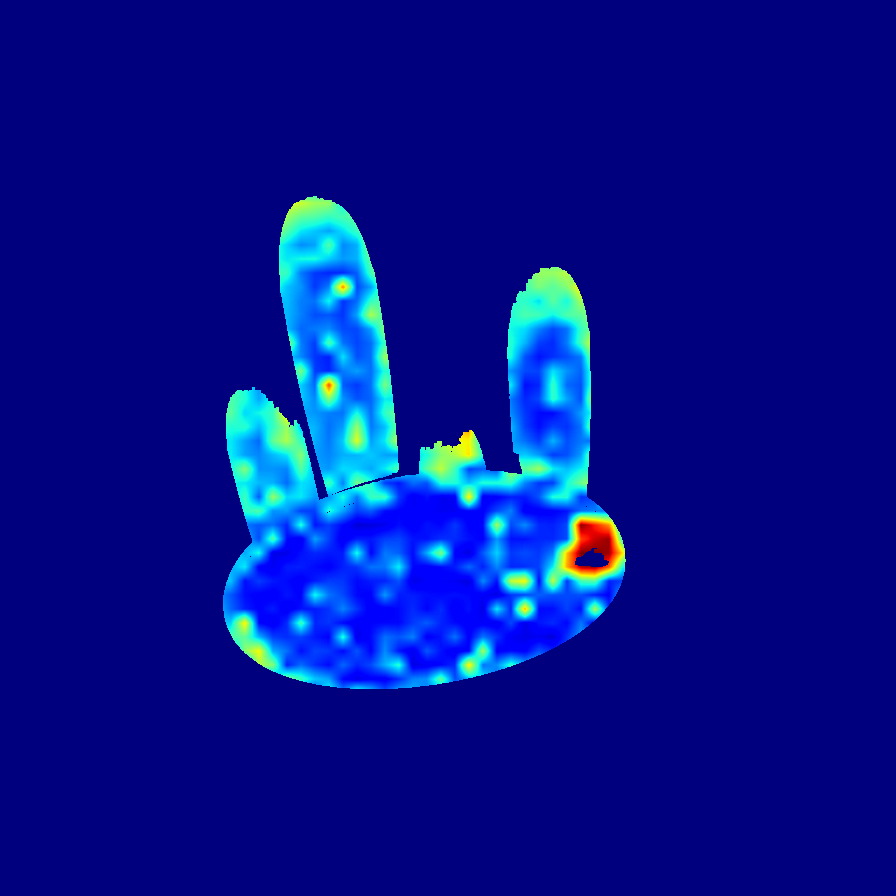} & 
            \includegraphics[width=0.08\linewidth,angle=180,origin=c]{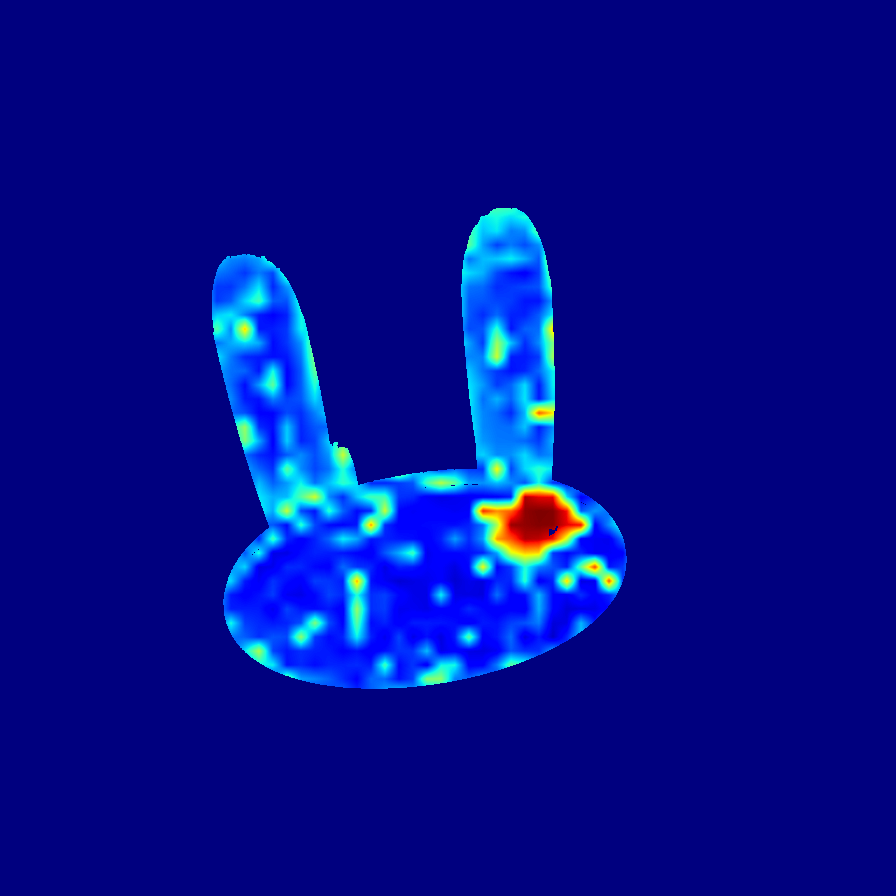} & 
            \includegraphics[width=0.08\linewidth,angle=180,origin=c]{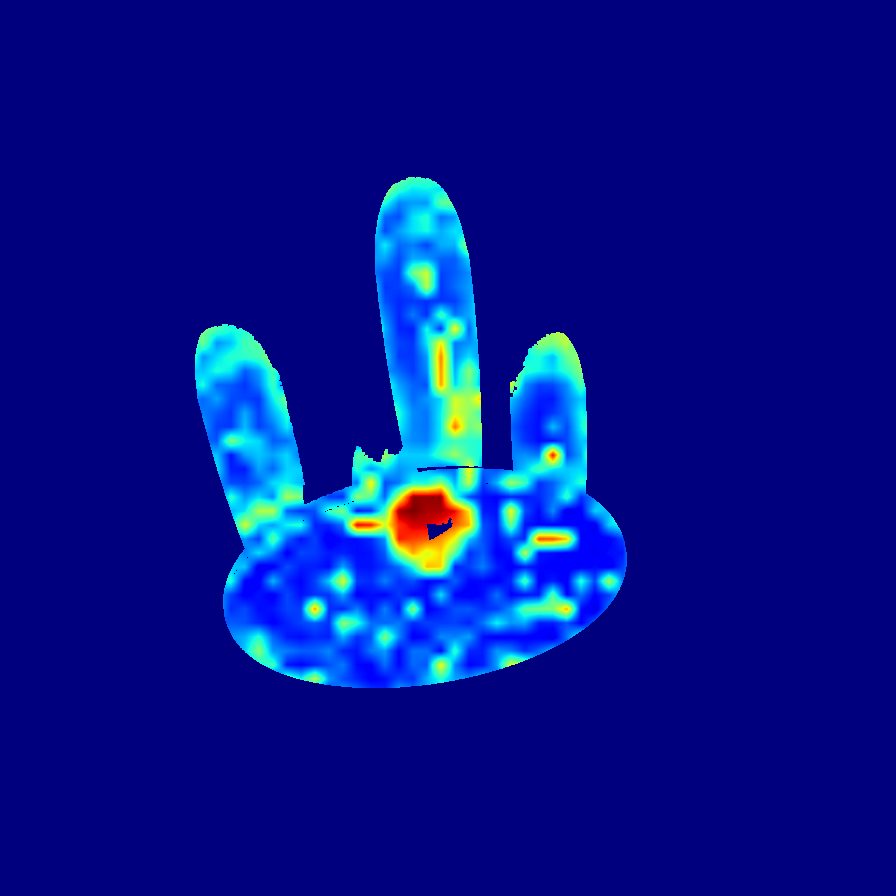} & 
            \includegraphics[width=0.08\linewidth,angle=180,origin=c]{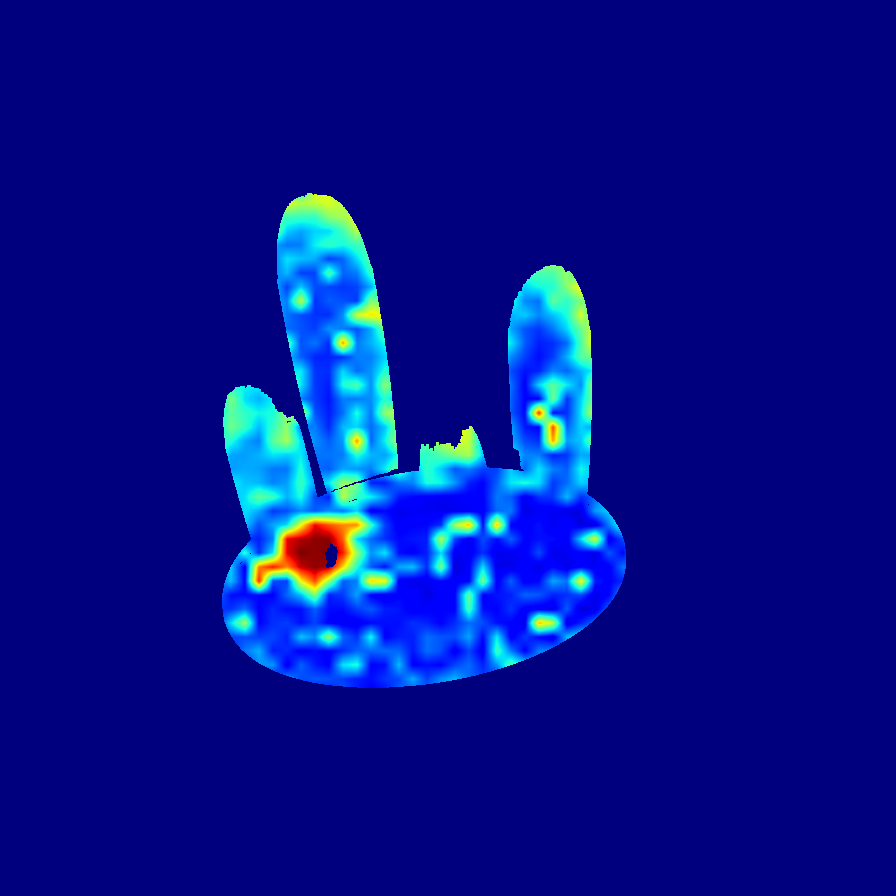} & 
            \includegraphics[width=0.08\linewidth,angle=180,origin=c]{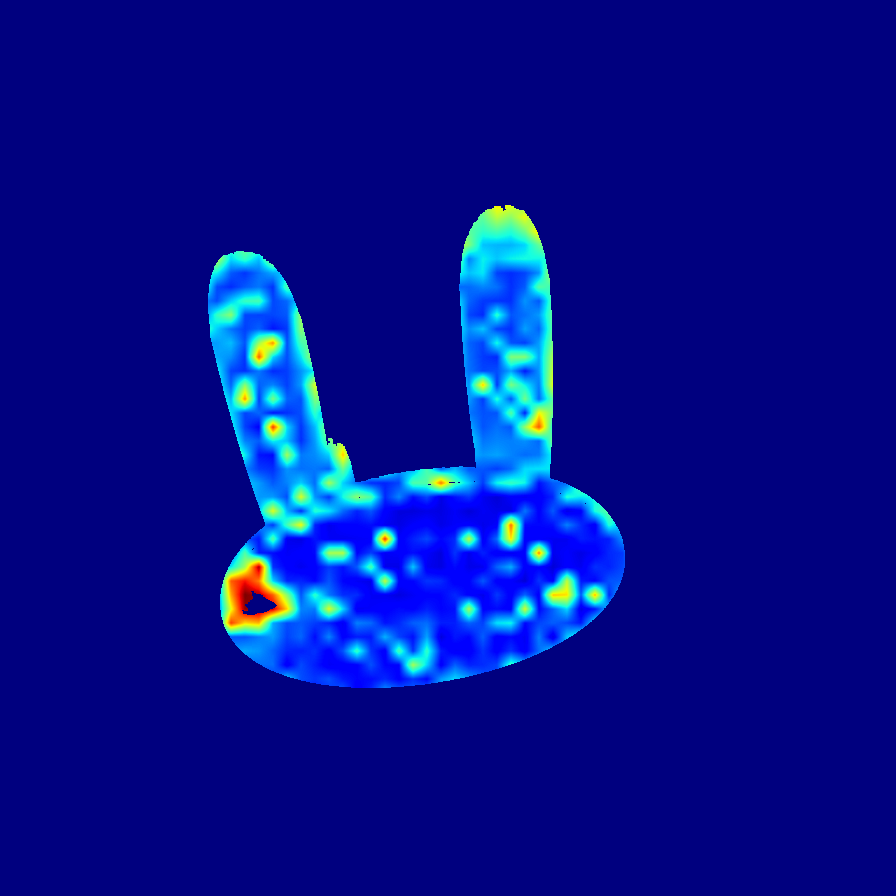} & 
            \includegraphics[width=0.08\linewidth,angle=180,origin=c]{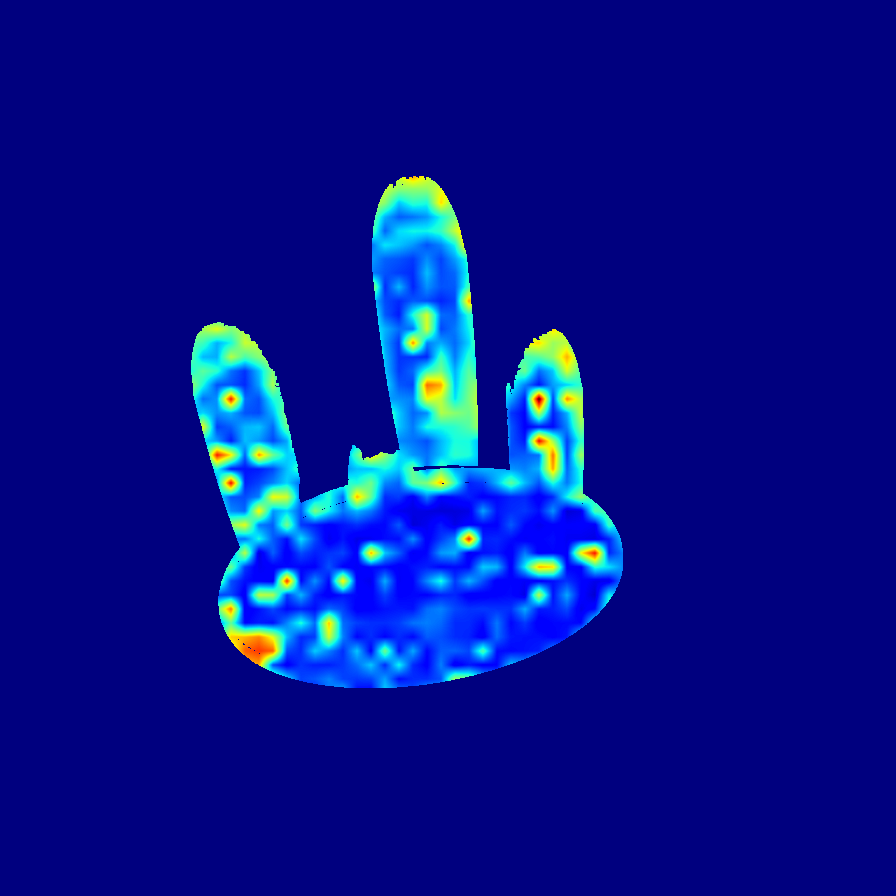} & 
            \includegraphics[width=0.08\linewidth,angle=180,origin=c]{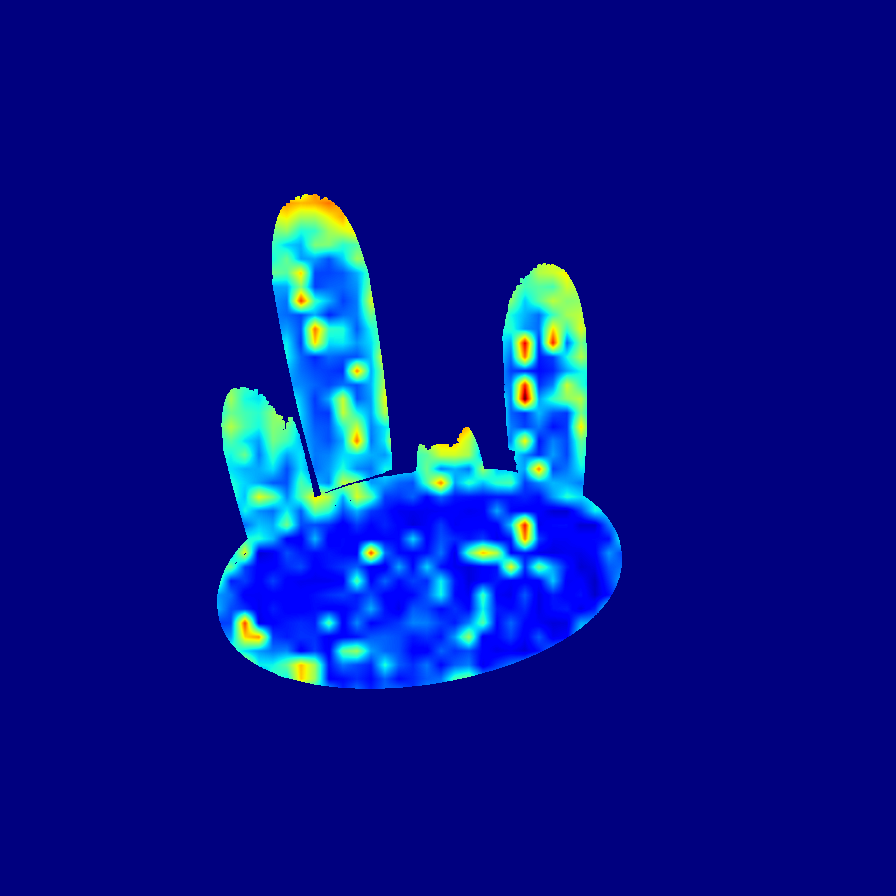} & 
            \includegraphics[width=0.08\linewidth,angle=180,origin=c]{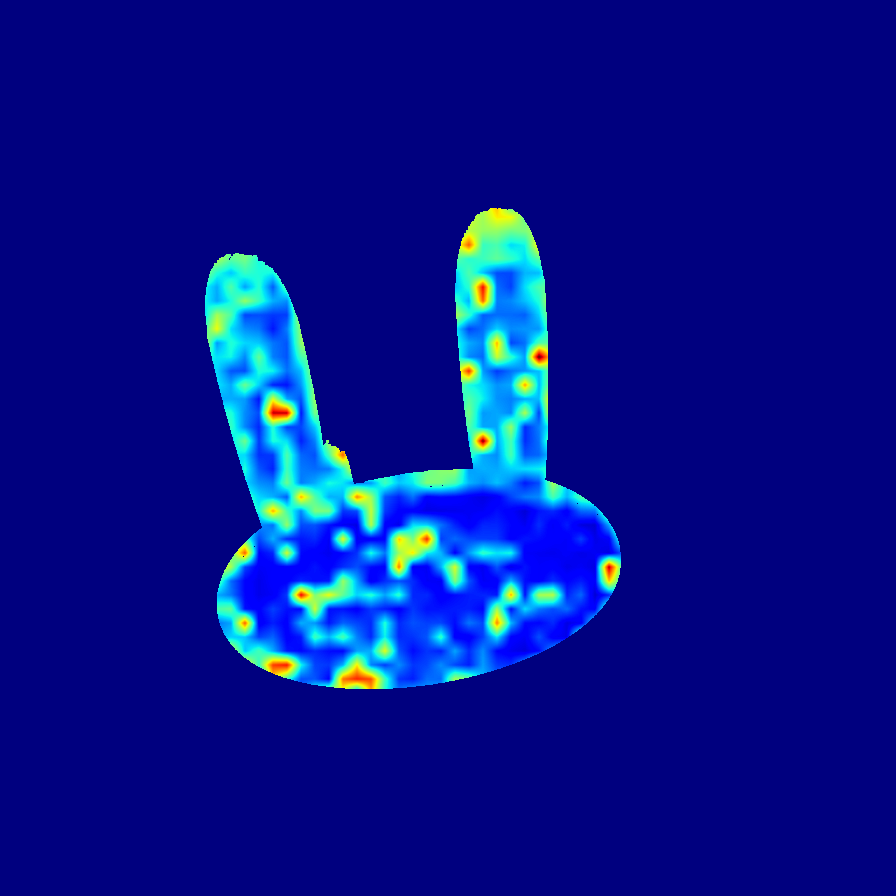} & 
            \includegraphics[width=0.08\linewidth,angle=180,origin=c]{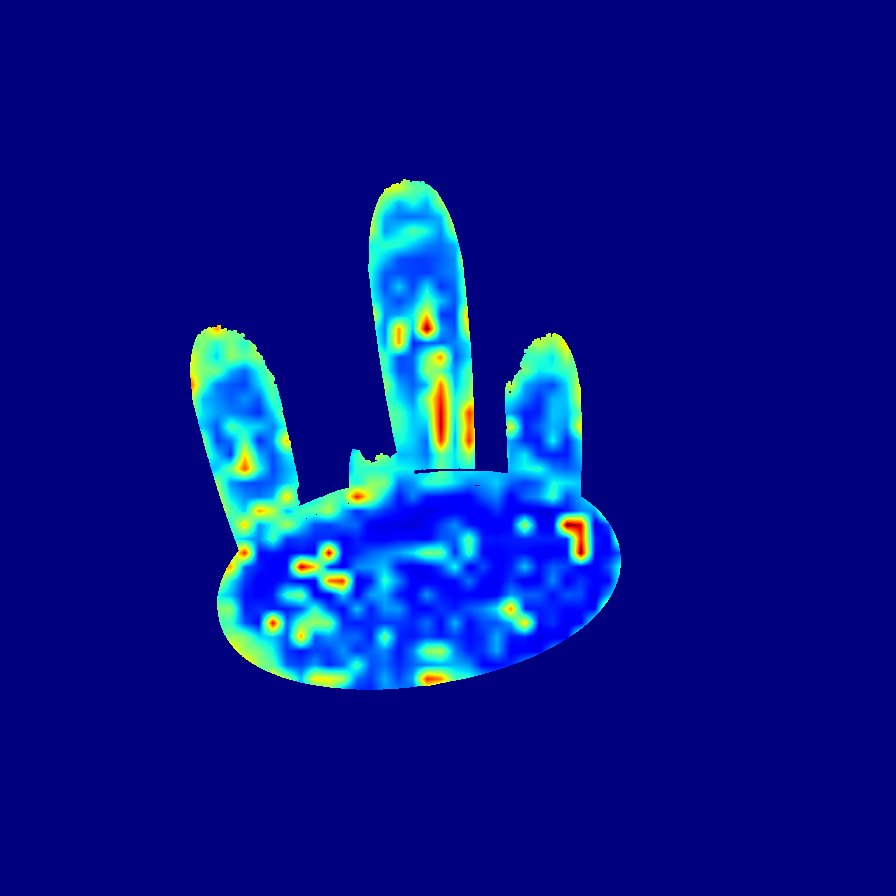} & 
            \includegraphics[width=0.08\linewidth,angle=180,origin=c]{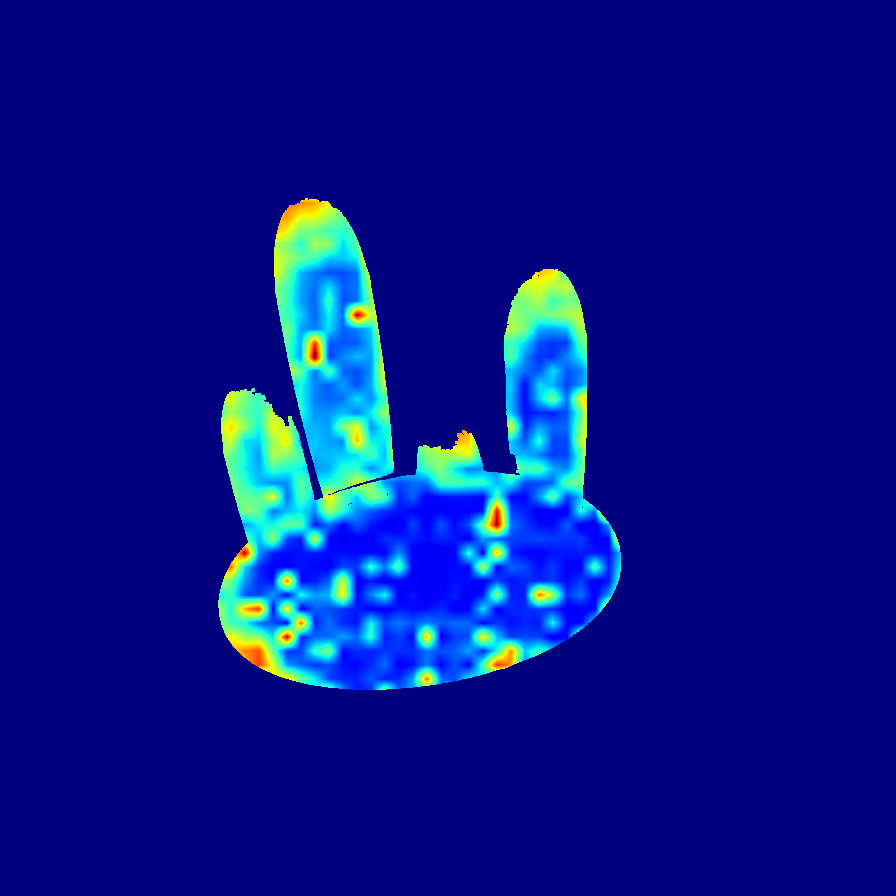} & 
            \includegraphics[width=0.08\linewidth,angle=180,origin=c]{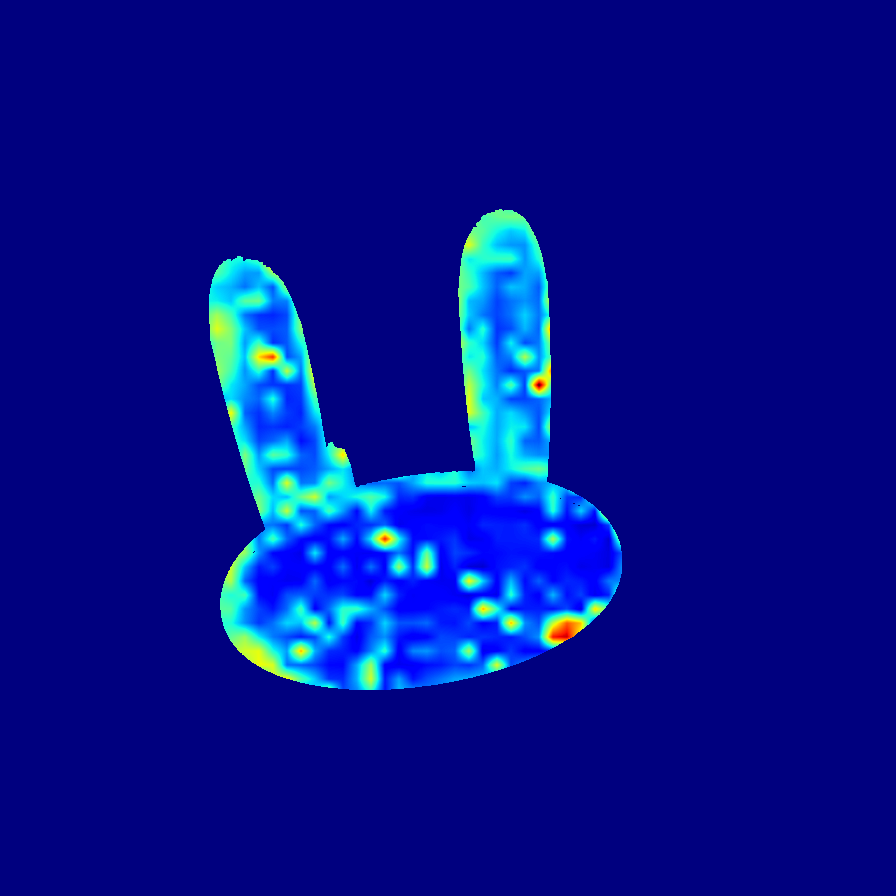} &
            \includegraphics[width=0.08\linewidth,angle=180,origin=c]{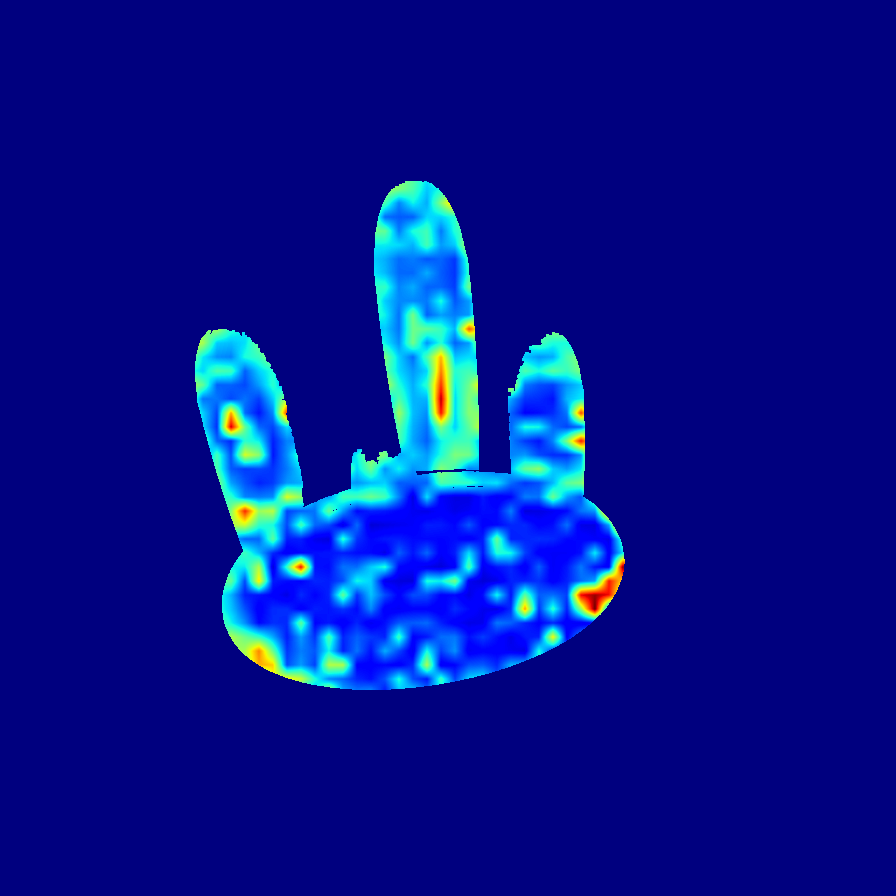} \\

            & \rotatebox{90}{\quad $v_{7}$} &
            \includegraphics[width=0.08\linewidth,angle=180,origin=c]{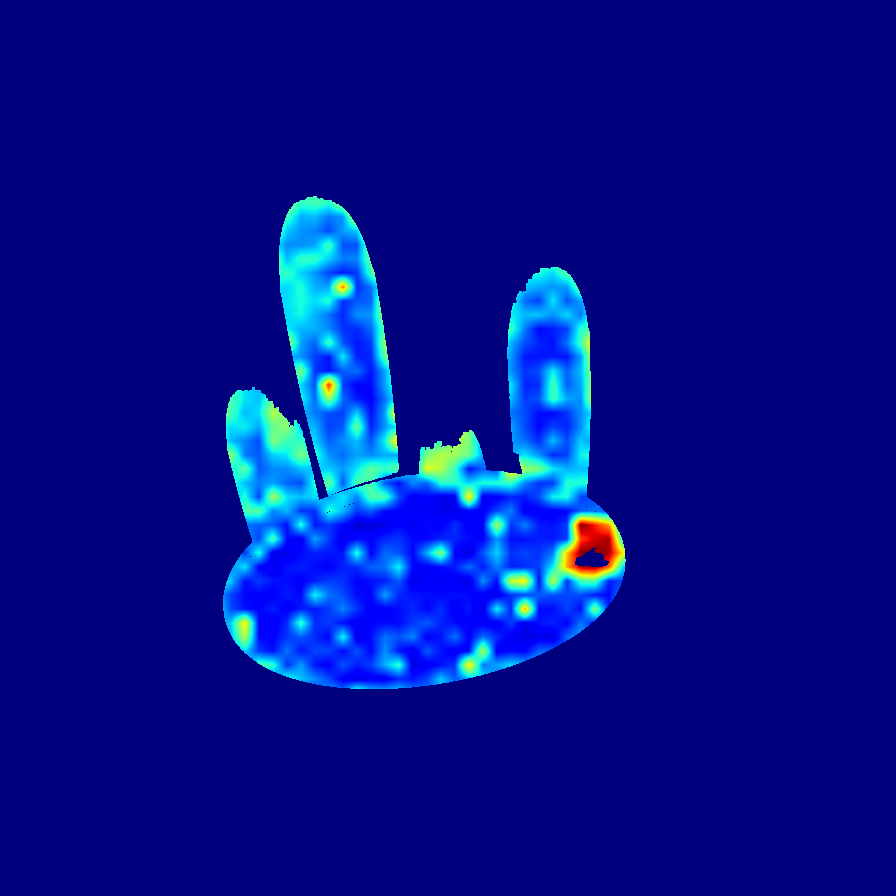} & 
            \includegraphics[width=0.08\linewidth,angle=180,origin=c]{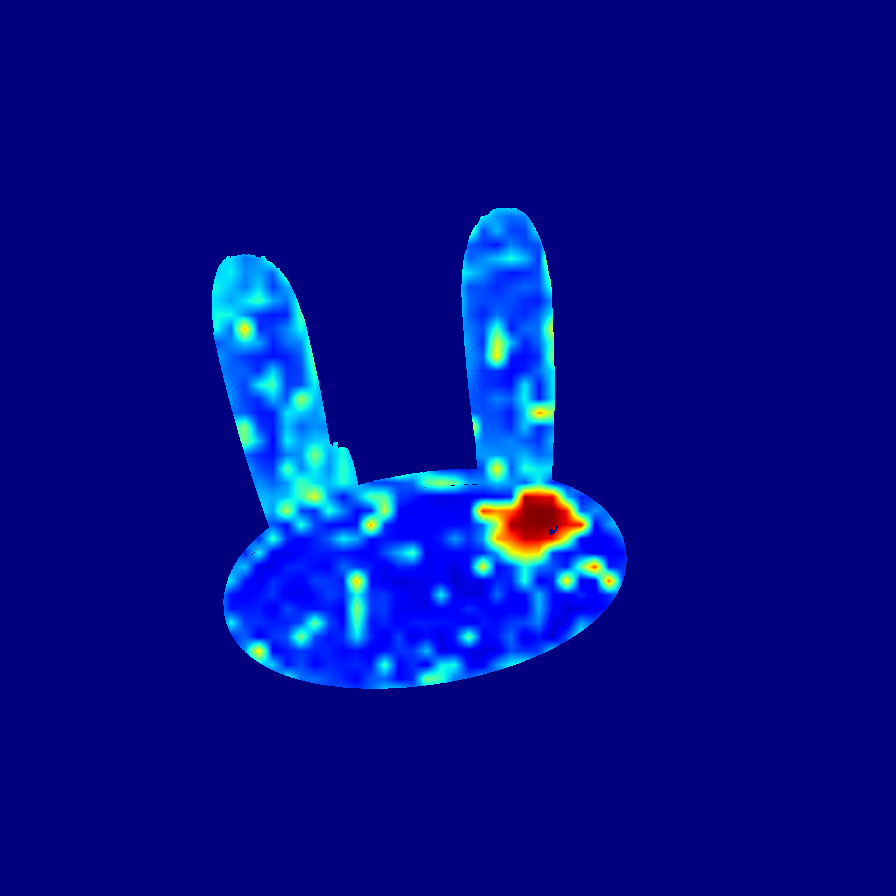} & 
            \includegraphics[width=0.08\linewidth,angle=180,origin=c]{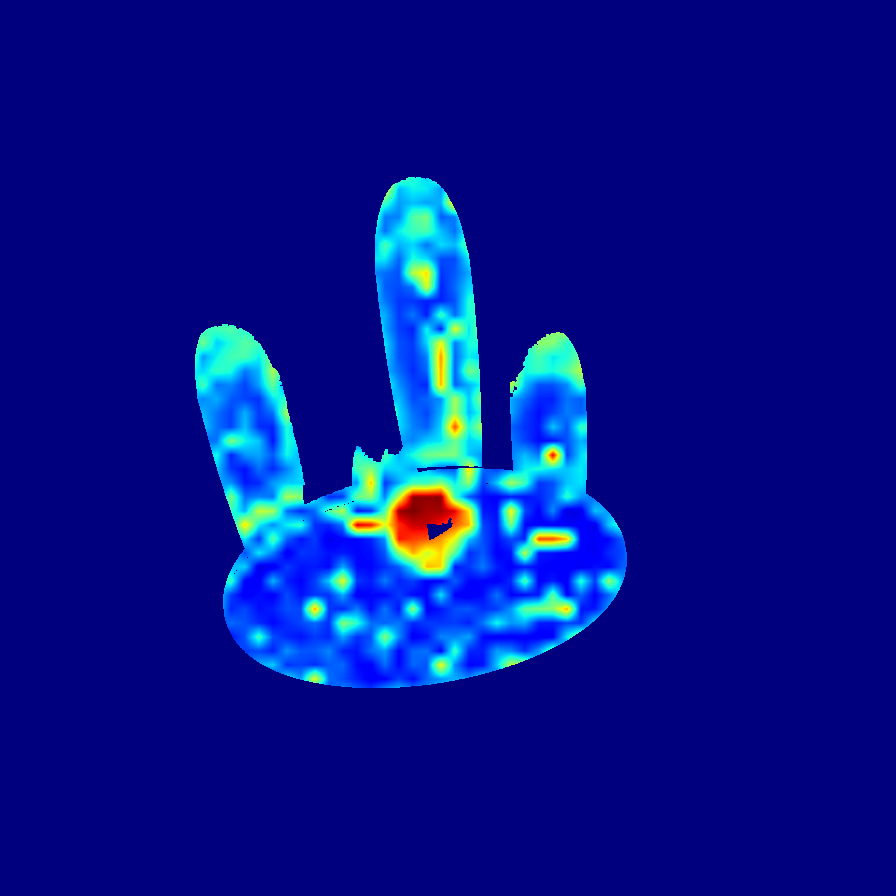} & 
            \includegraphics[width=0.08\linewidth,angle=180,origin=c]{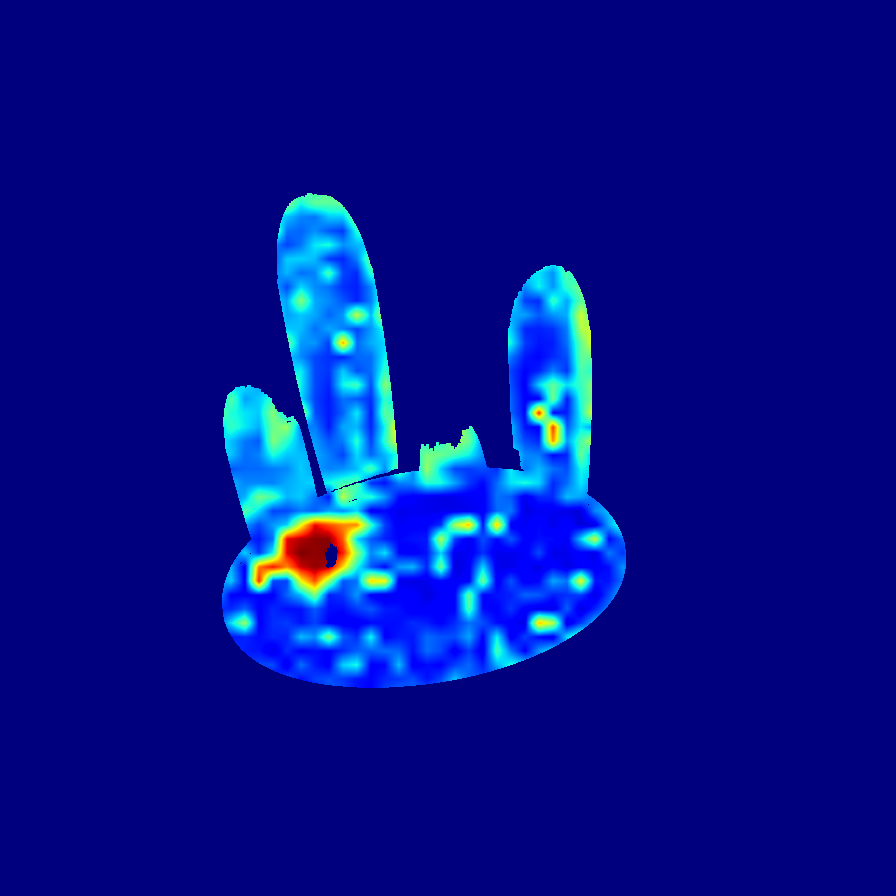} & 
            \includegraphics[width=0.08\linewidth,angle=180,origin=c]{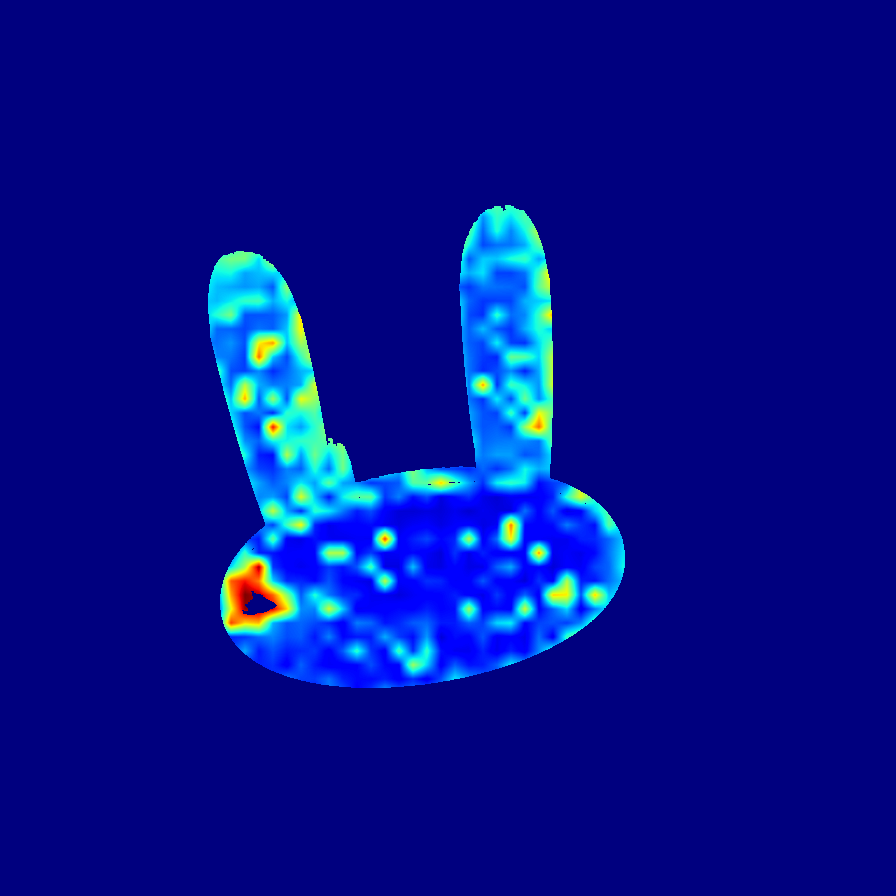} & 
            \includegraphics[width=0.08\linewidth,angle=180,origin=c]{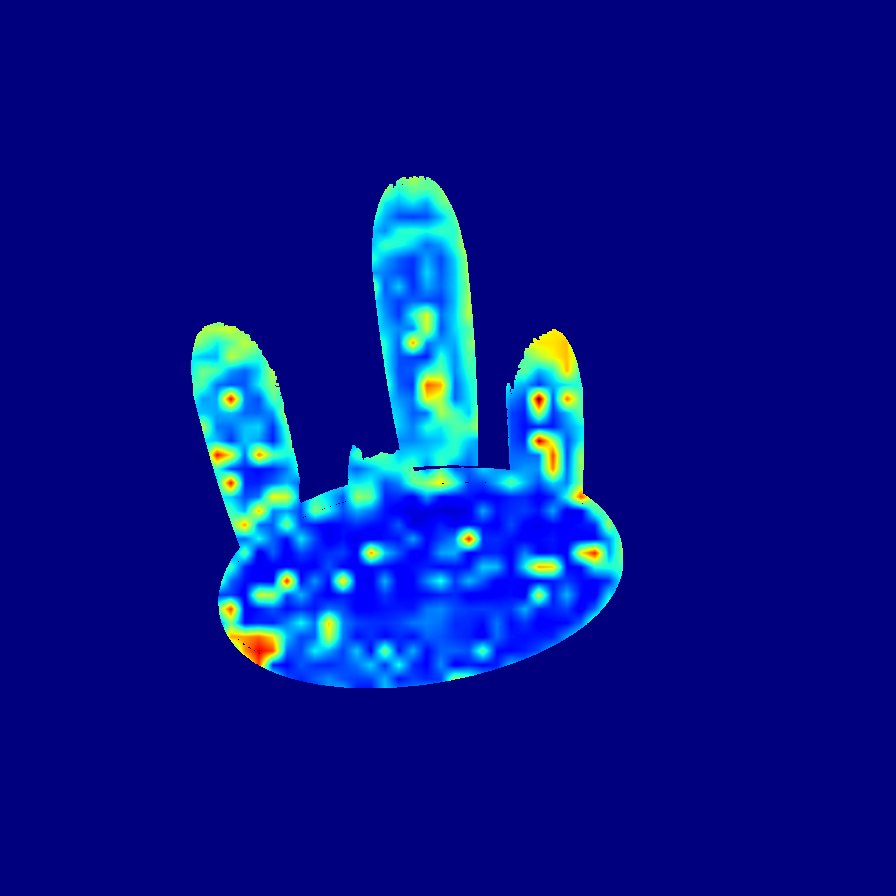} & 
            \includegraphics[width=0.08\linewidth,angle=180,origin=c]{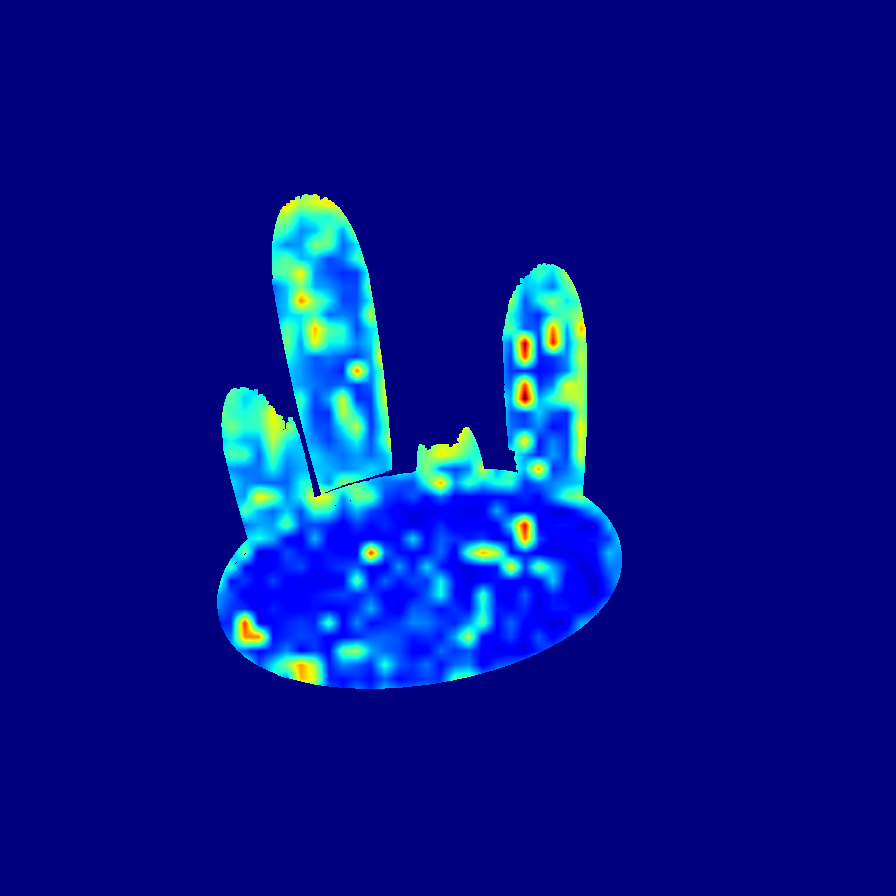} & 
            \includegraphics[width=0.08\linewidth,angle=180,origin=c]{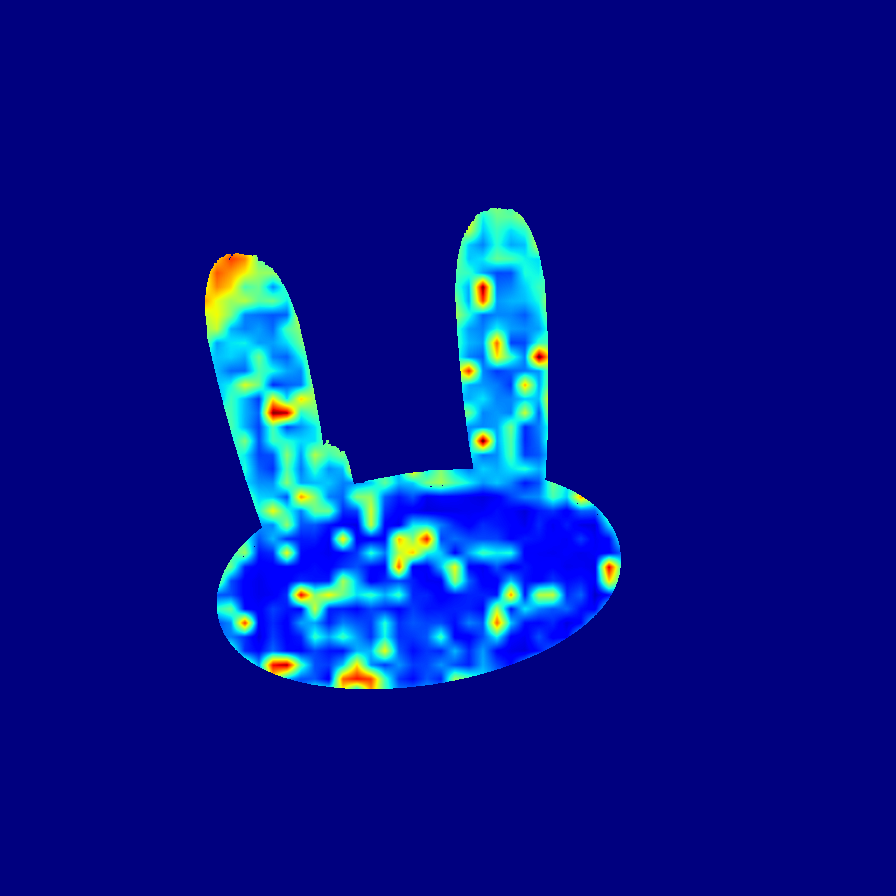} & 
            \includegraphics[width=0.08\linewidth,angle=180,origin=c]{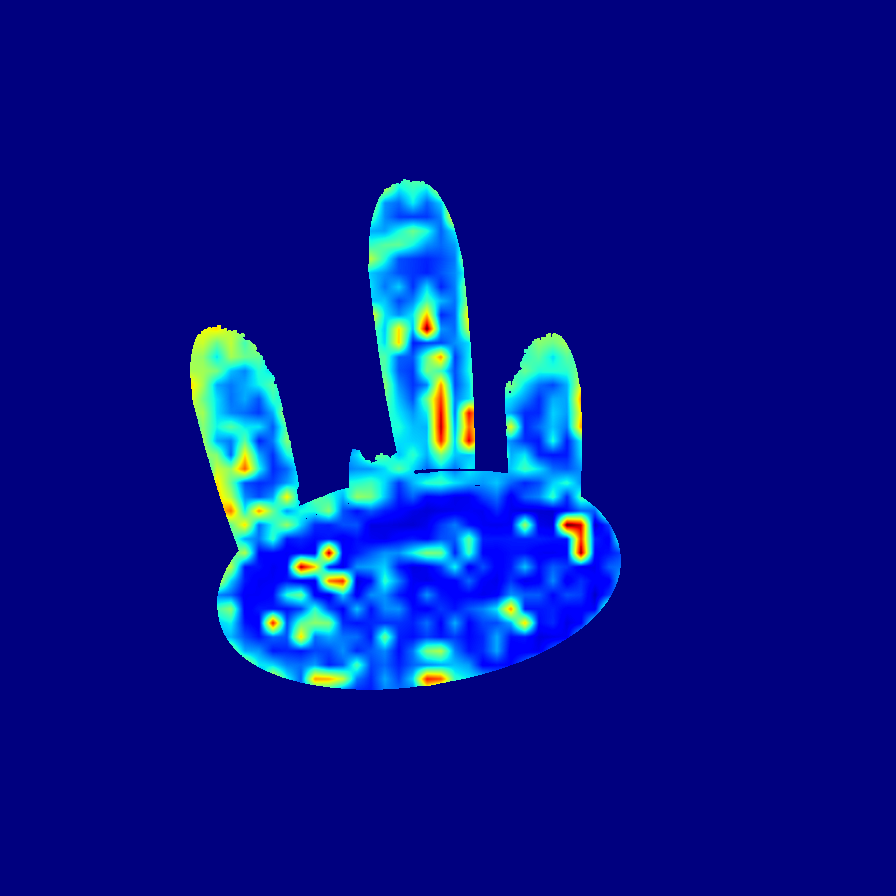} & 
            \includegraphics[width=0.08\linewidth,angle=180,origin=c]{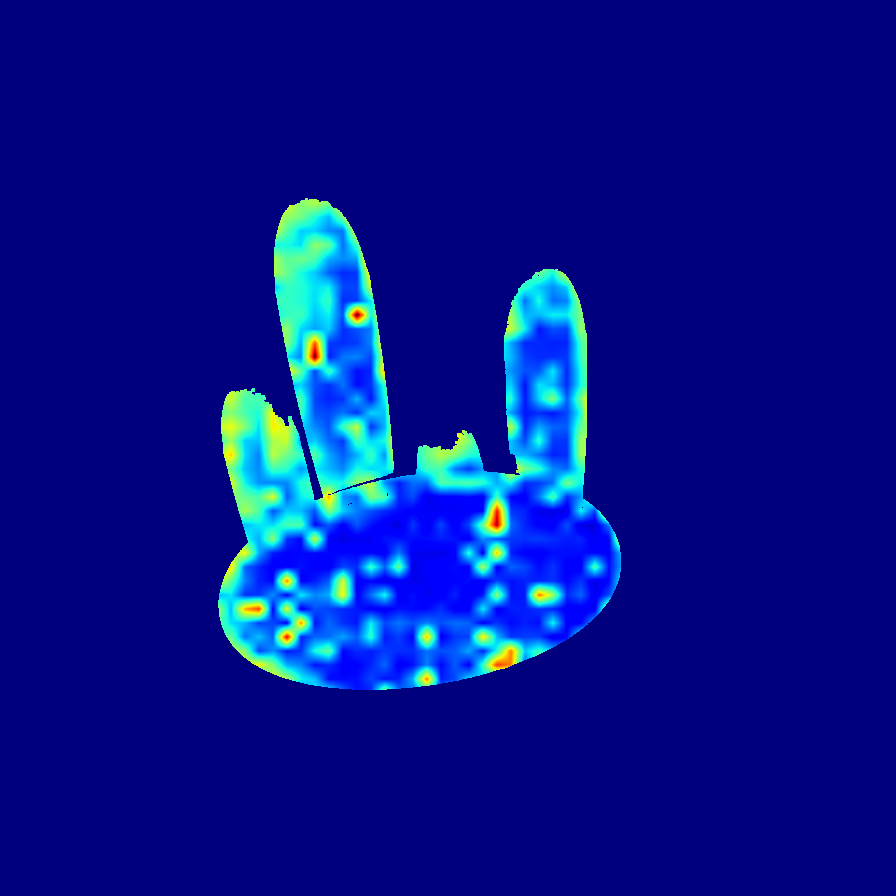} & 
            \includegraphics[width=0.08\linewidth,angle=180,origin=c]{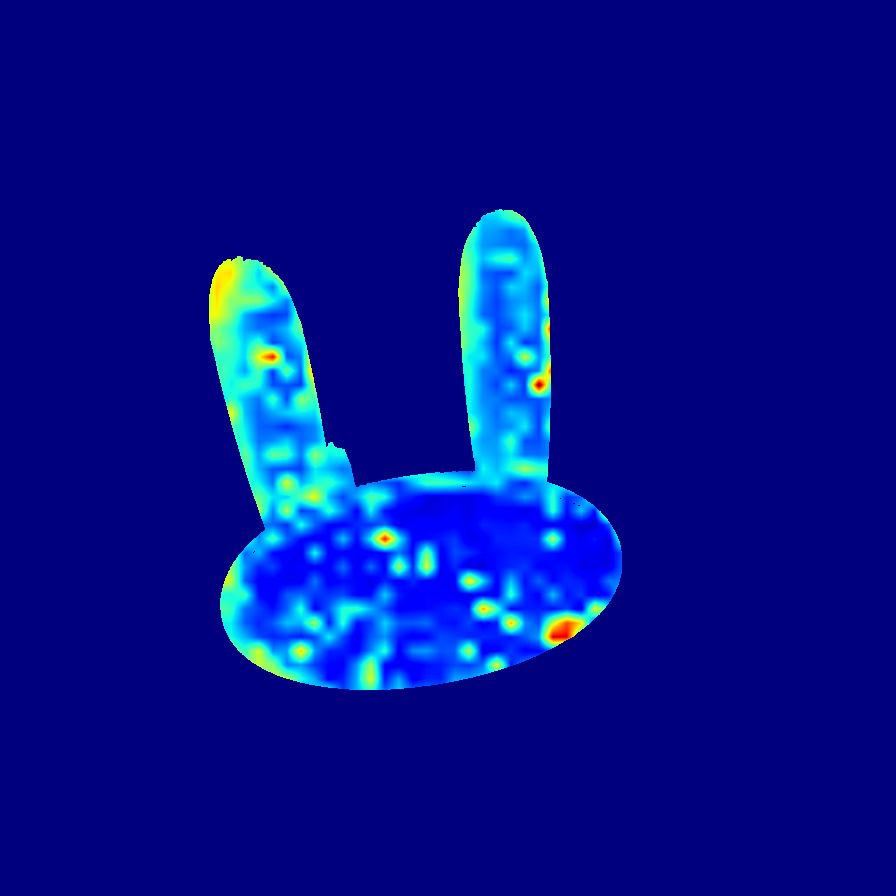} &
            \includegraphics[width=0.08\linewidth,angle=180,origin=c]{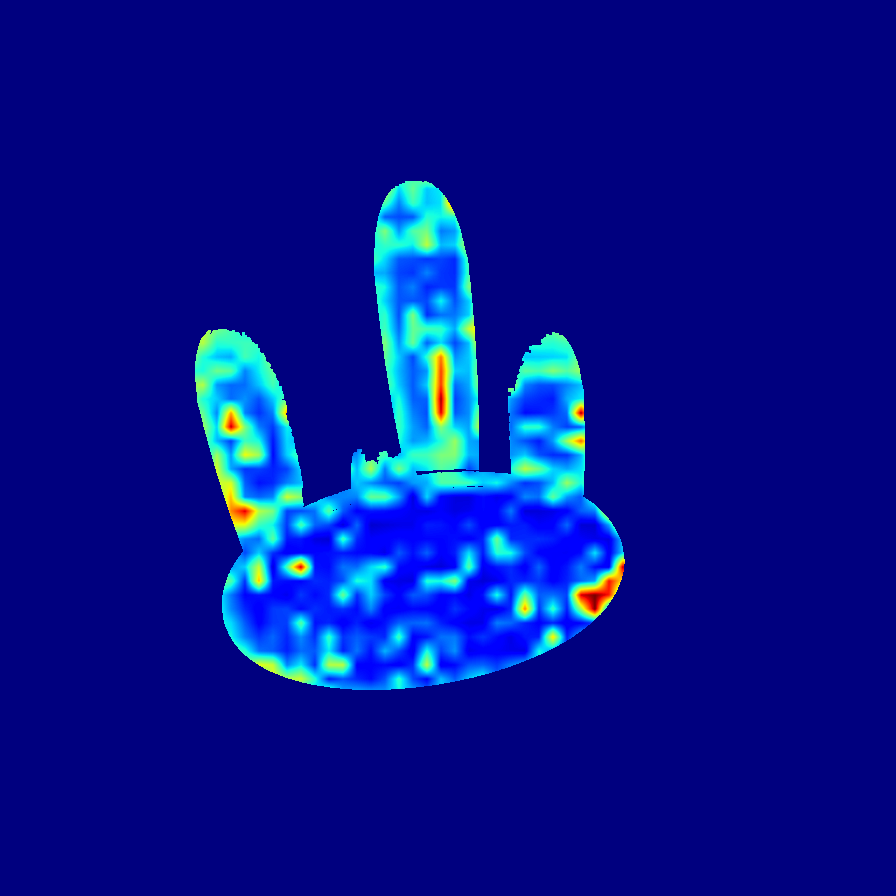} \\

            & \rotatebox{90}{\quad $v_{8}$} &
            \includegraphics[width=0.08\linewidth,angle=180,origin=c]{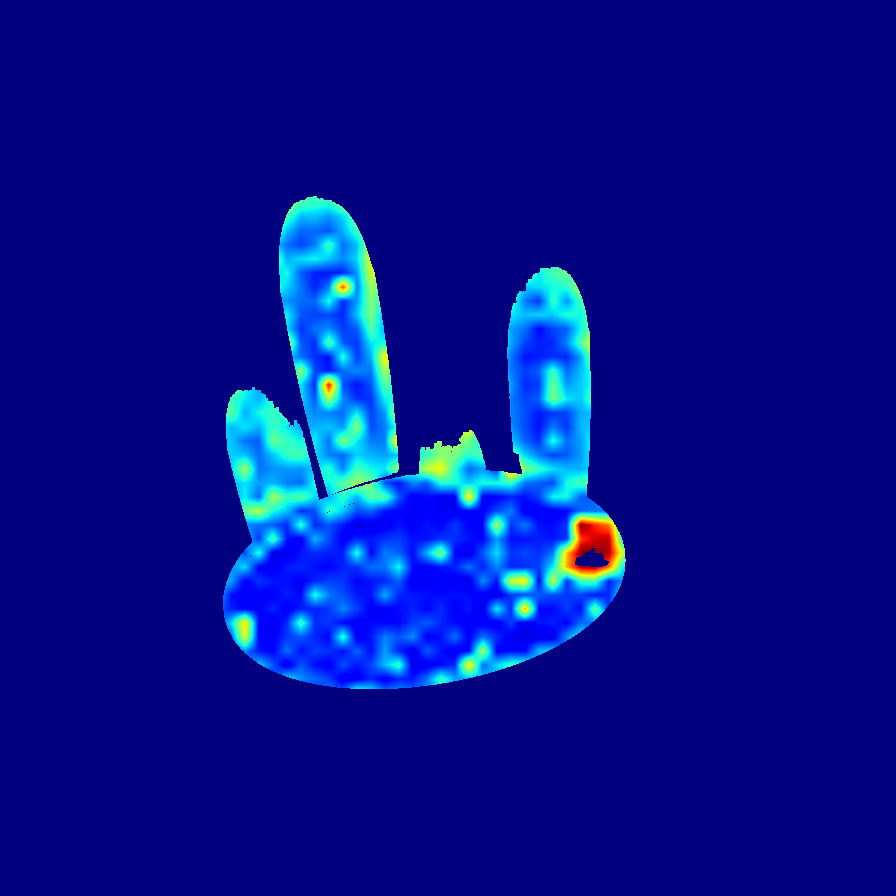} & 
            \includegraphics[width=0.08\linewidth,angle=180,origin=c]{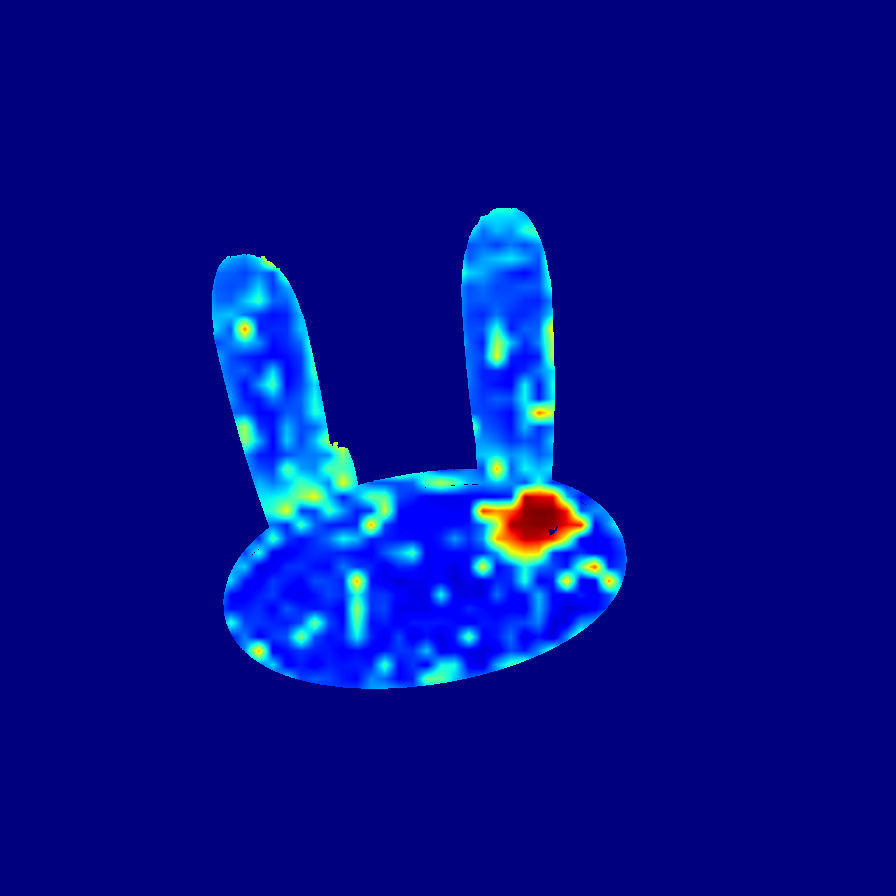} & 
            \includegraphics[width=0.08\linewidth,angle=180,origin=c]{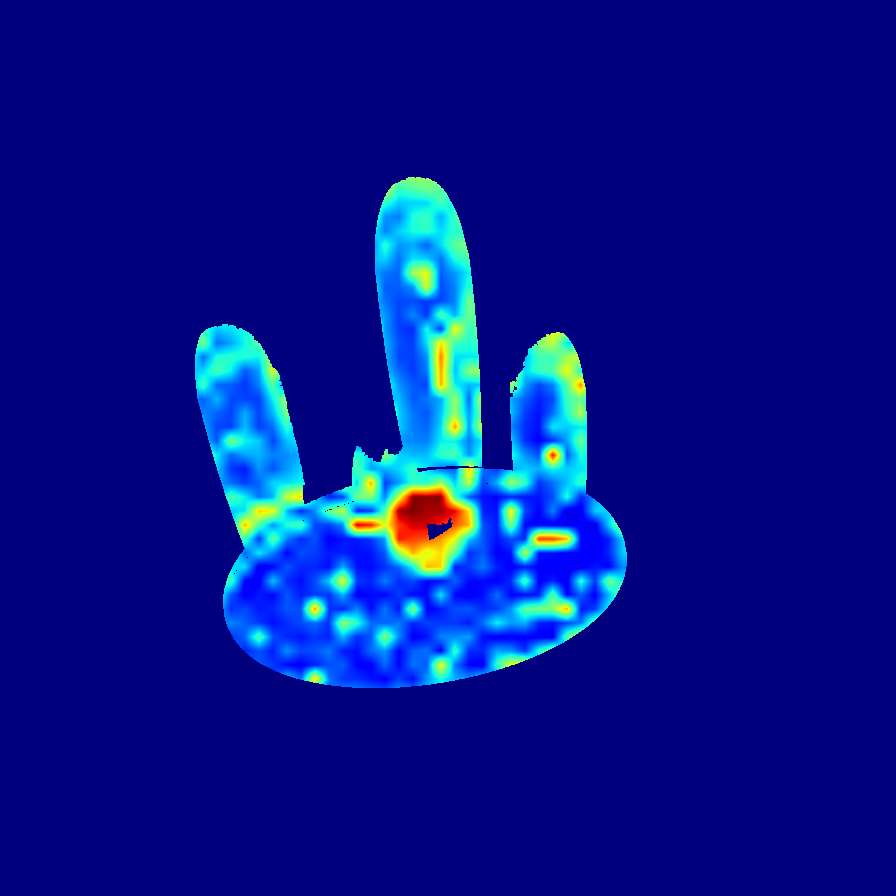} & 
            \includegraphics[width=0.08\linewidth,angle=180,origin=c]{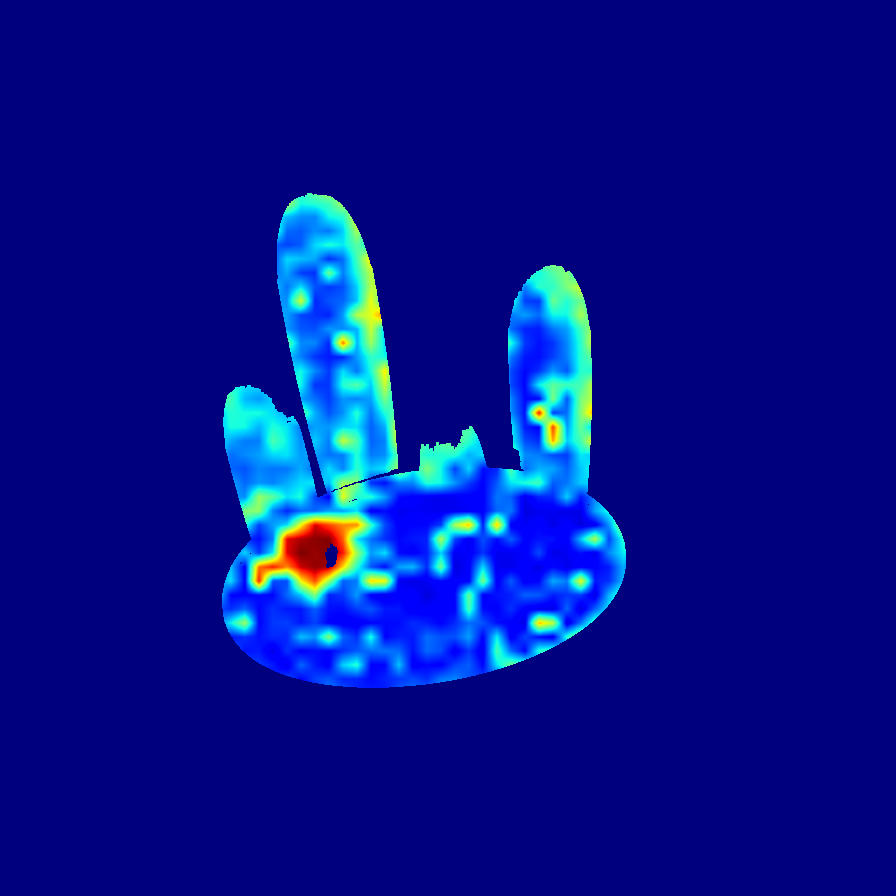} & 
            \includegraphics[width=0.08\linewidth,angle=180,origin=c]{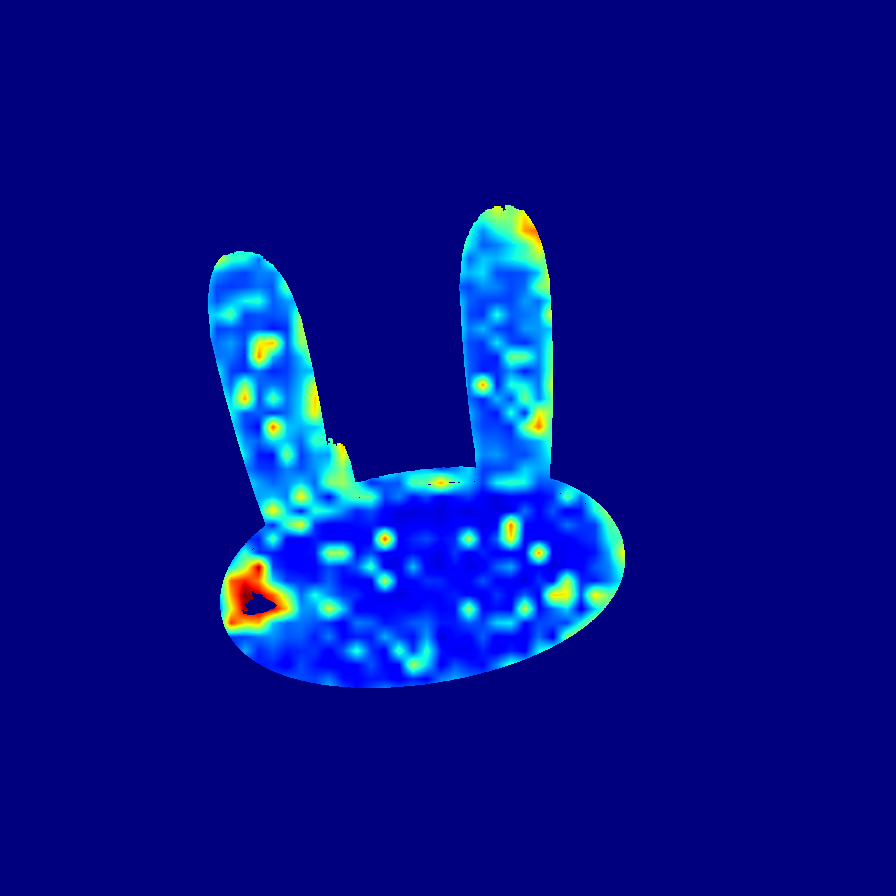} & 
            \includegraphics[width=0.08\linewidth,angle=180,origin=c]{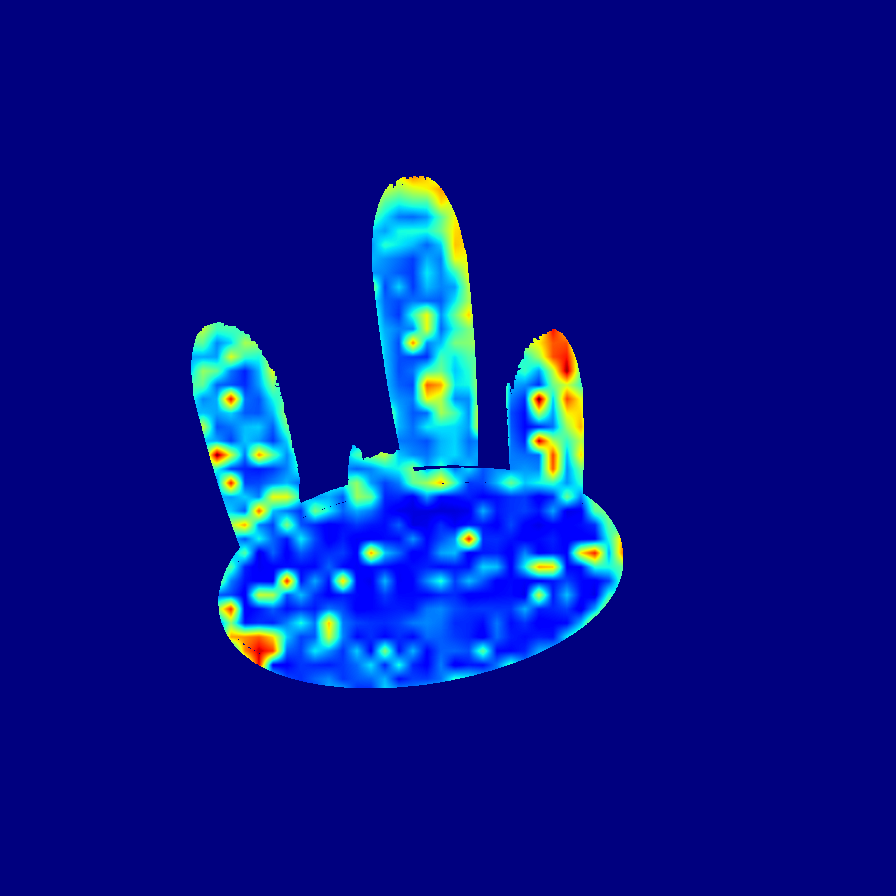} & 
            \includegraphics[width=0.08\linewidth,angle=180,origin=c]{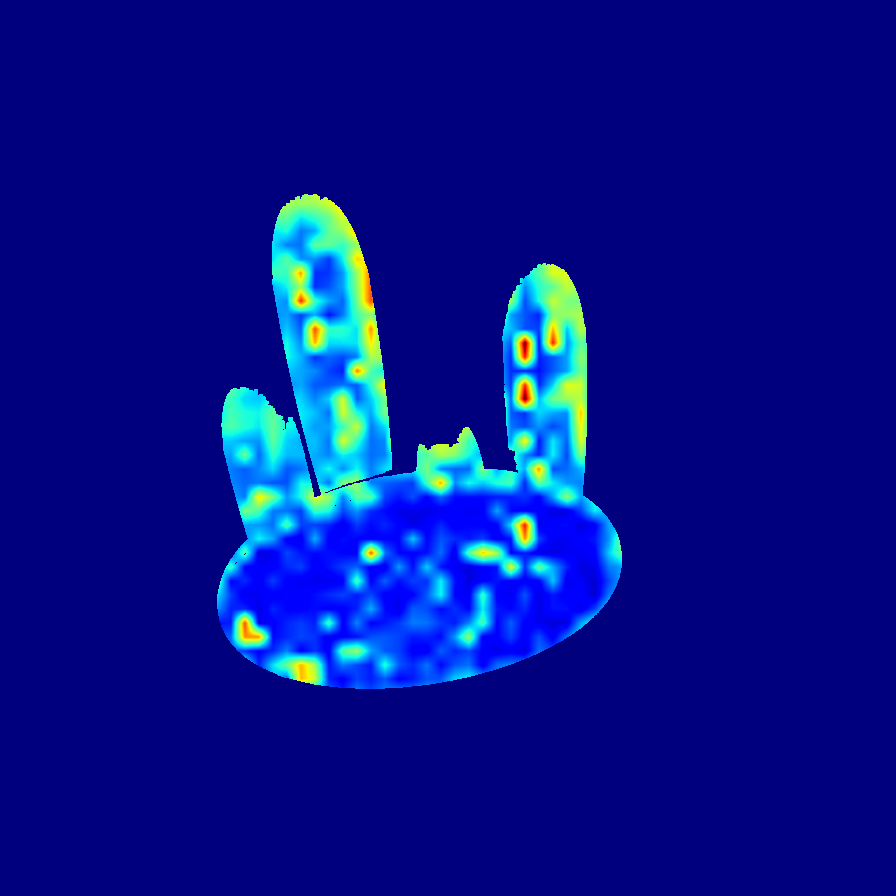} & 
            \includegraphics[width=0.08\linewidth,angle=180,origin=c]{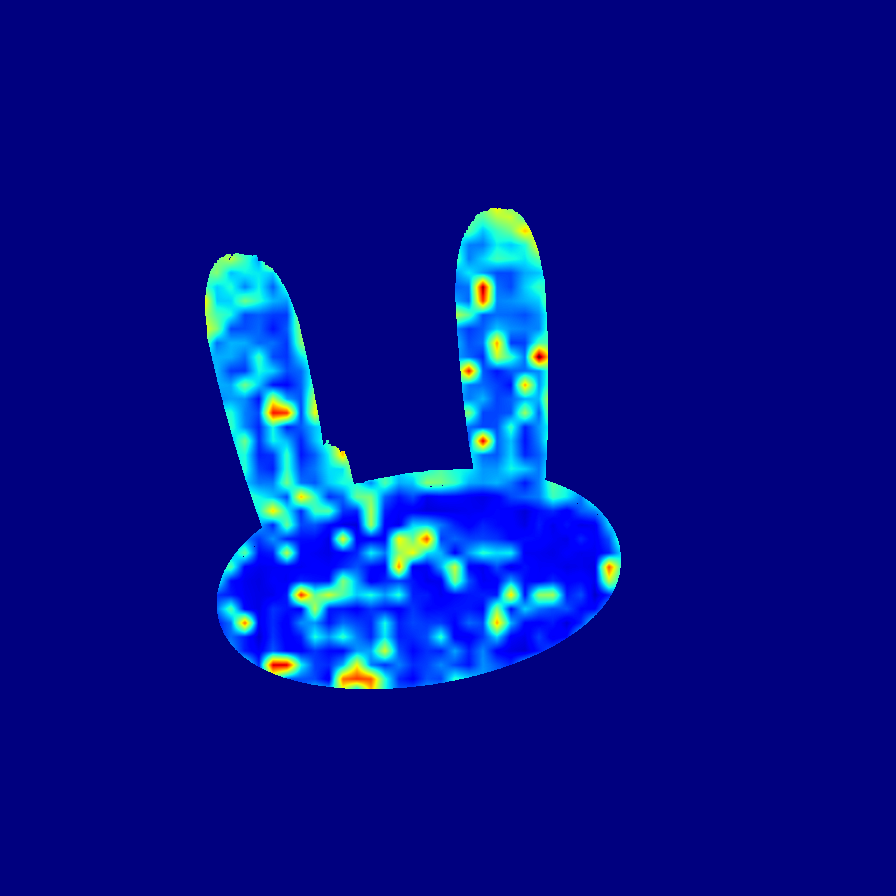} & 
            \includegraphics[width=0.08\linewidth,angle=180,origin=c]{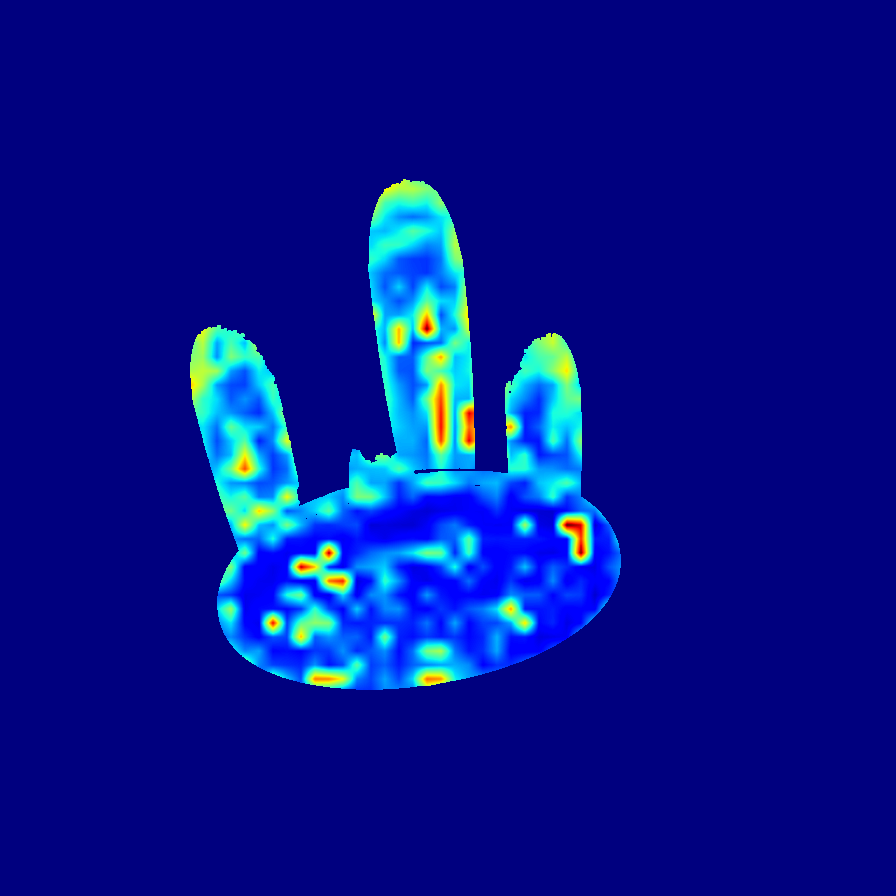} & 
            \includegraphics[width=0.08\linewidth,angle=180,origin=c]{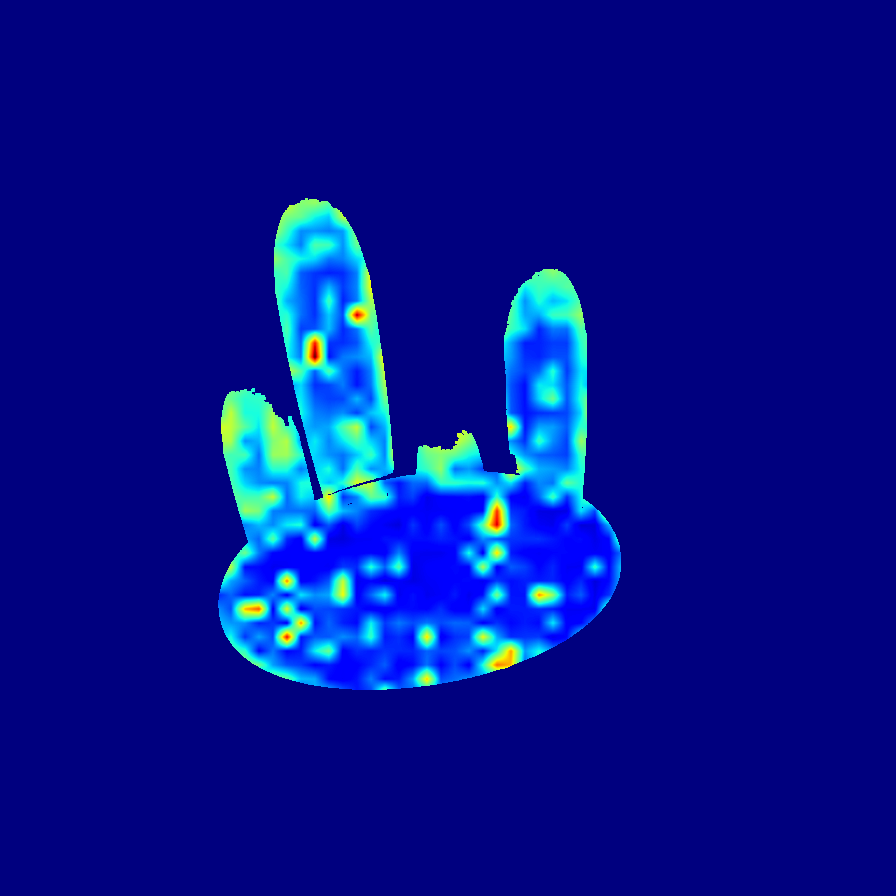} & 
            \includegraphics[width=0.08\linewidth,angle=180,origin=c]{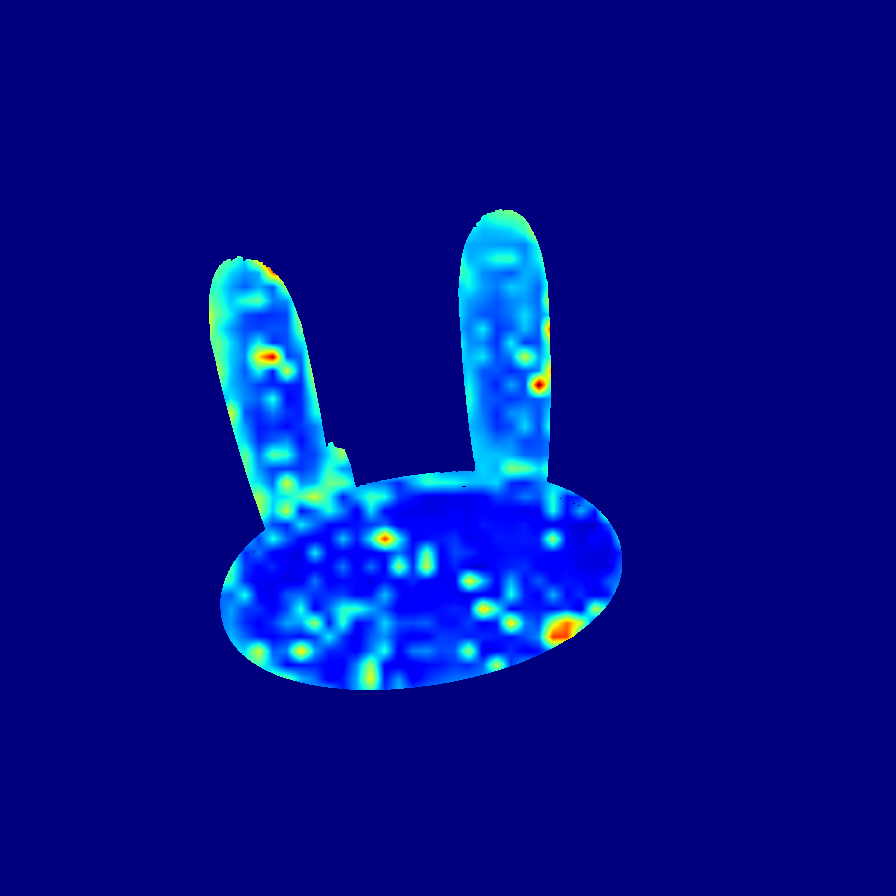} &
            \includegraphics[width=0.08\linewidth,angle=180,origin=c]{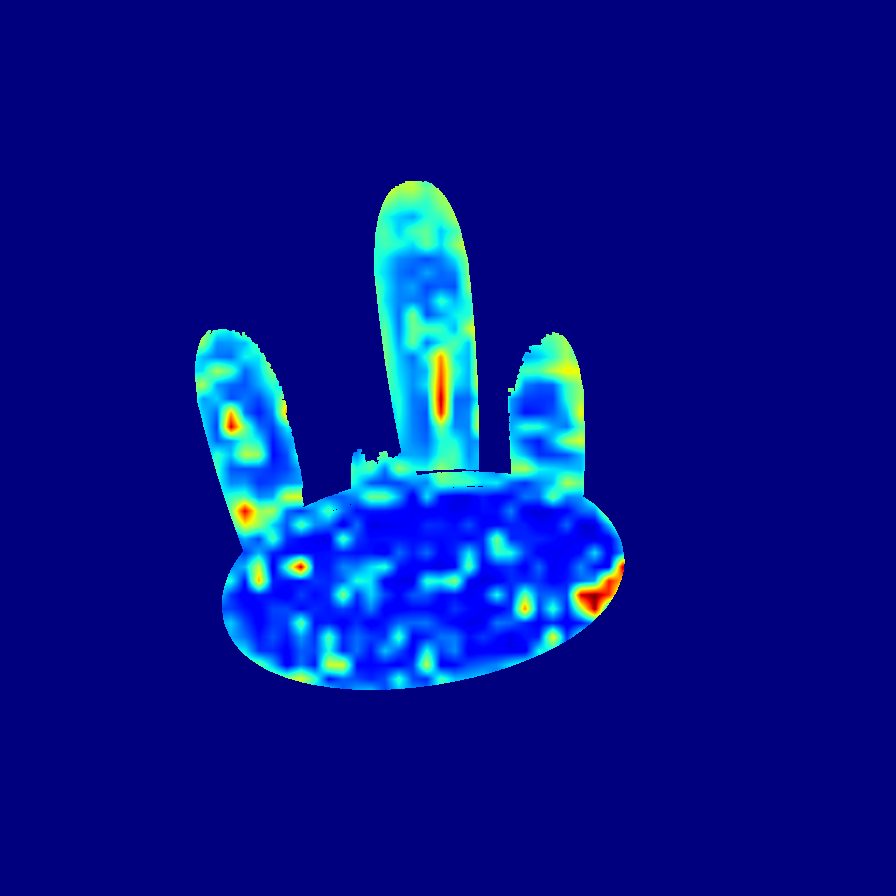} \\

            & \rotatebox{90}{\quad $v_{9}$} &
            \includegraphics[width=0.08\linewidth,angle=180,origin=c]{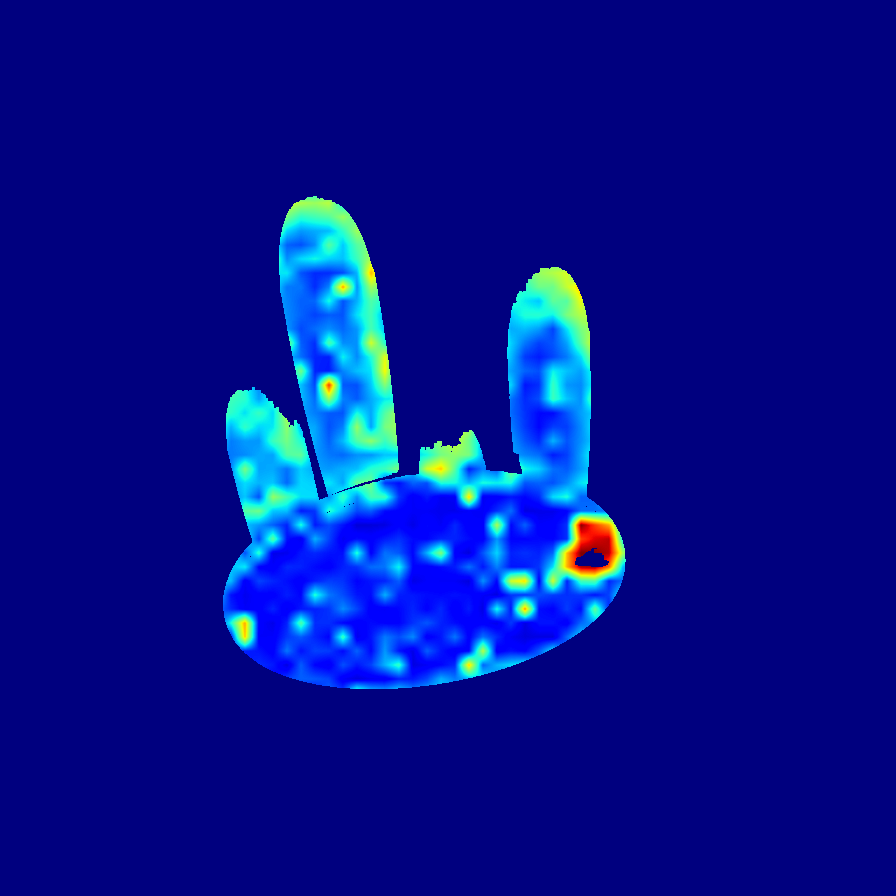} & 
            \includegraphics[width=0.08\linewidth,angle=180,origin=c]{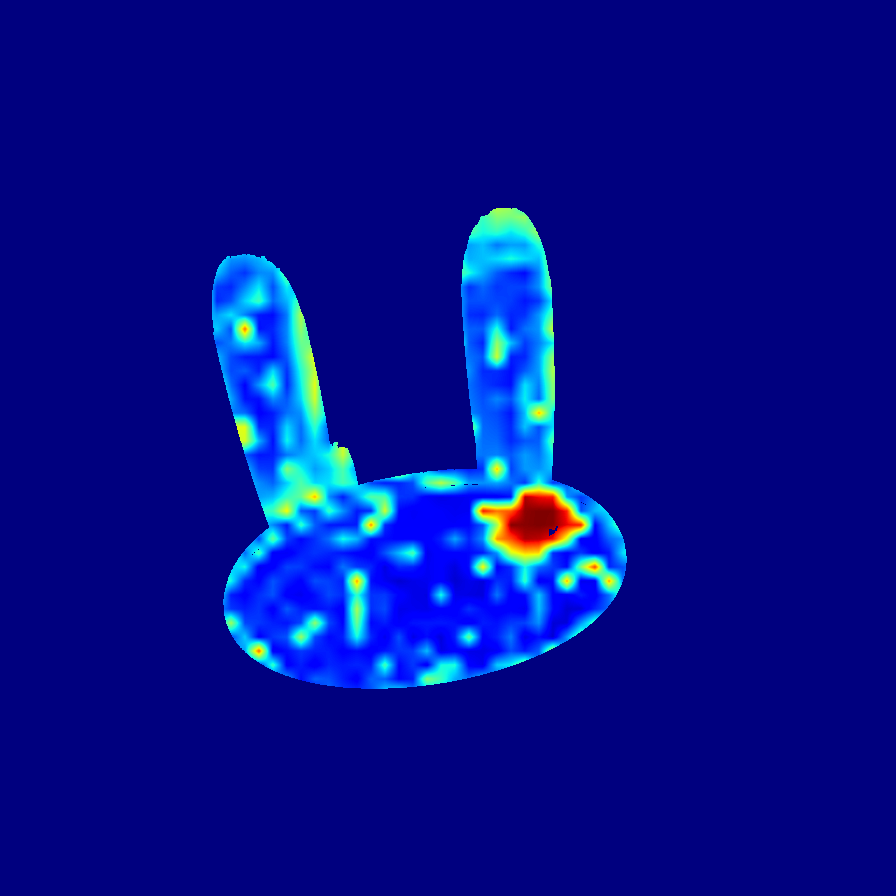} & 
            \includegraphics[width=0.08\linewidth,angle=180,origin=c]{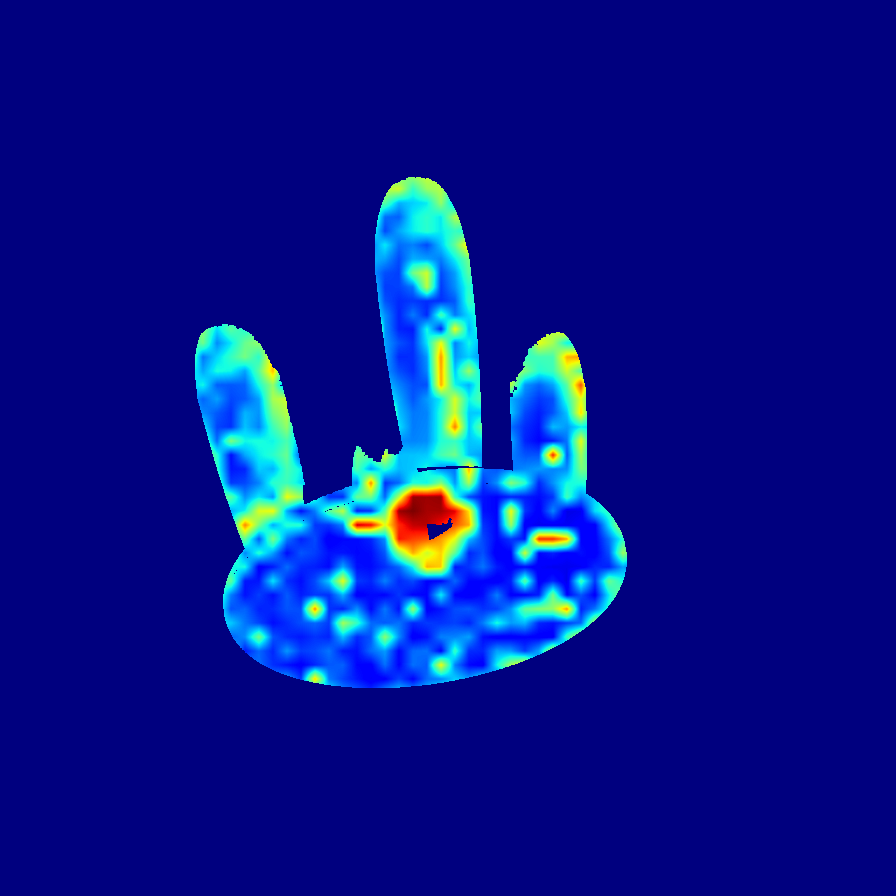} & 
            \includegraphics[width=0.08\linewidth,angle=180,origin=c]{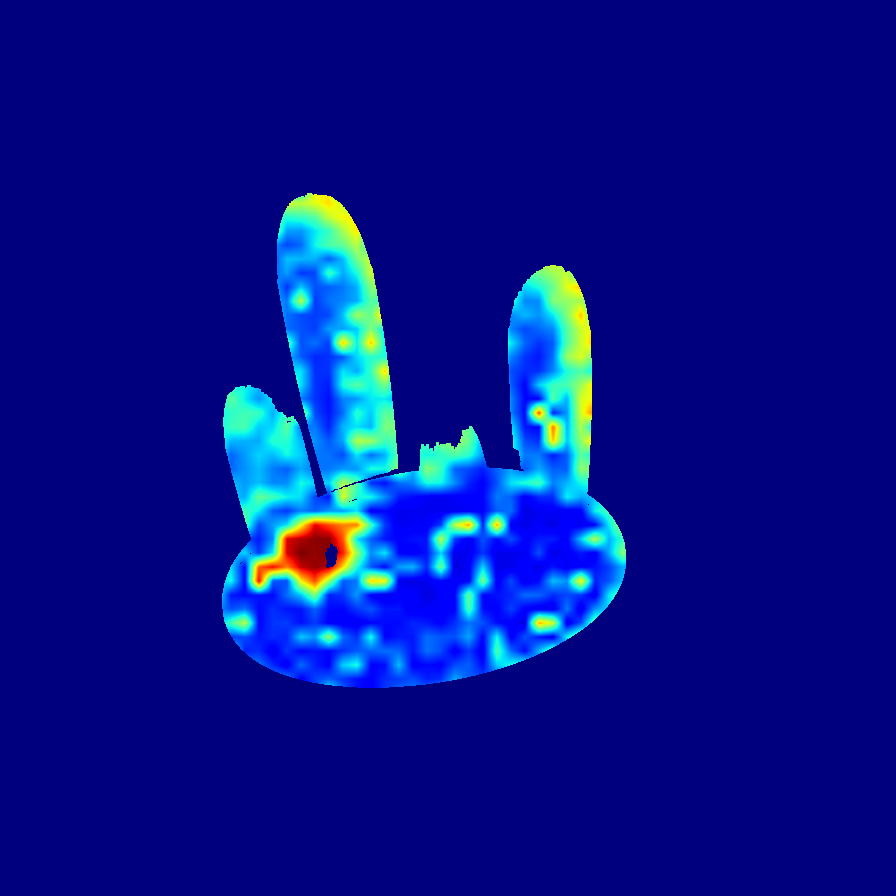} & 
            \includegraphics[width=0.08\linewidth,angle=180,origin=c]{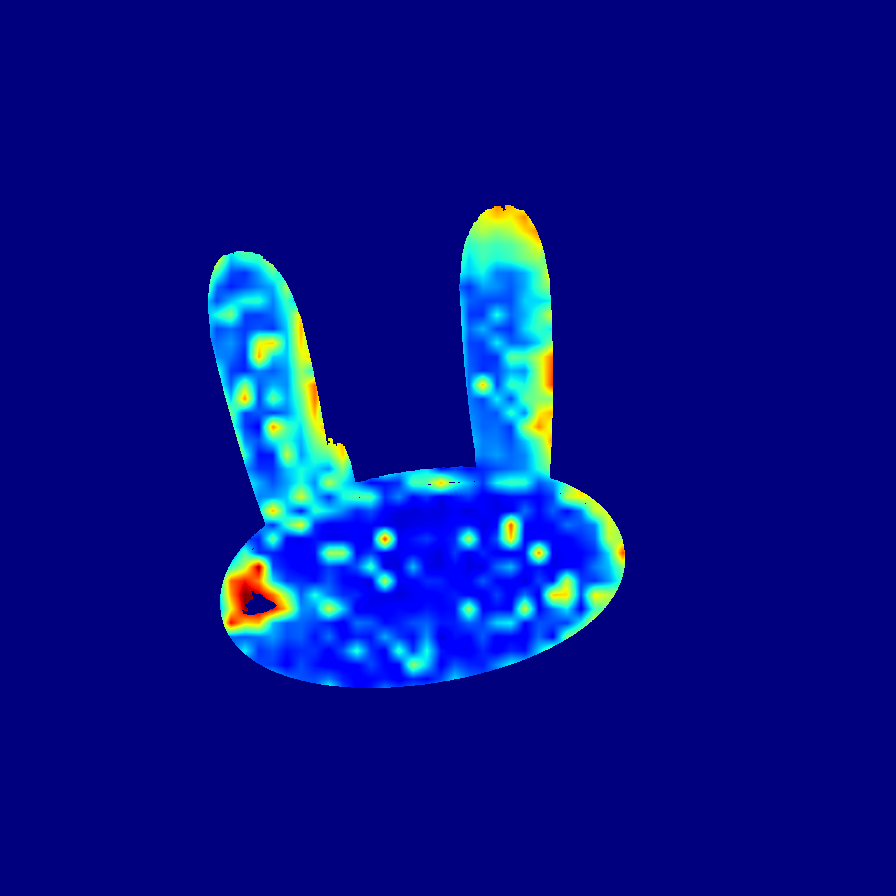} & 
            \includegraphics[width=0.08\linewidth,angle=180,origin=c]{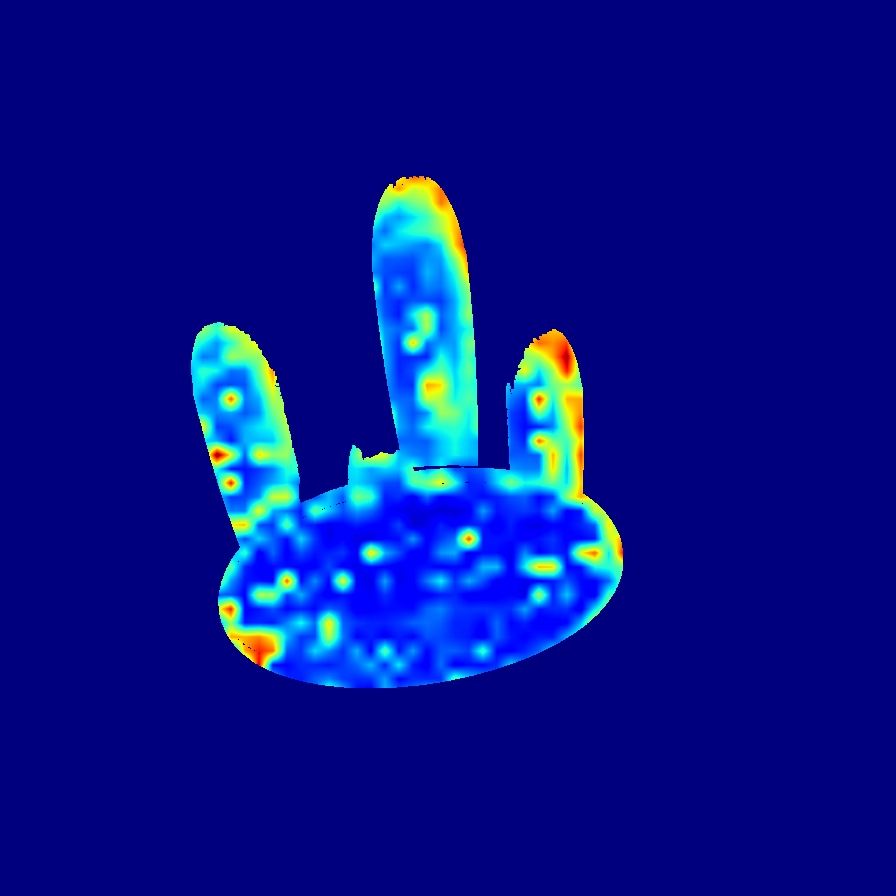} & 
            \includegraphics[width=0.08\linewidth,angle=180,origin=c]{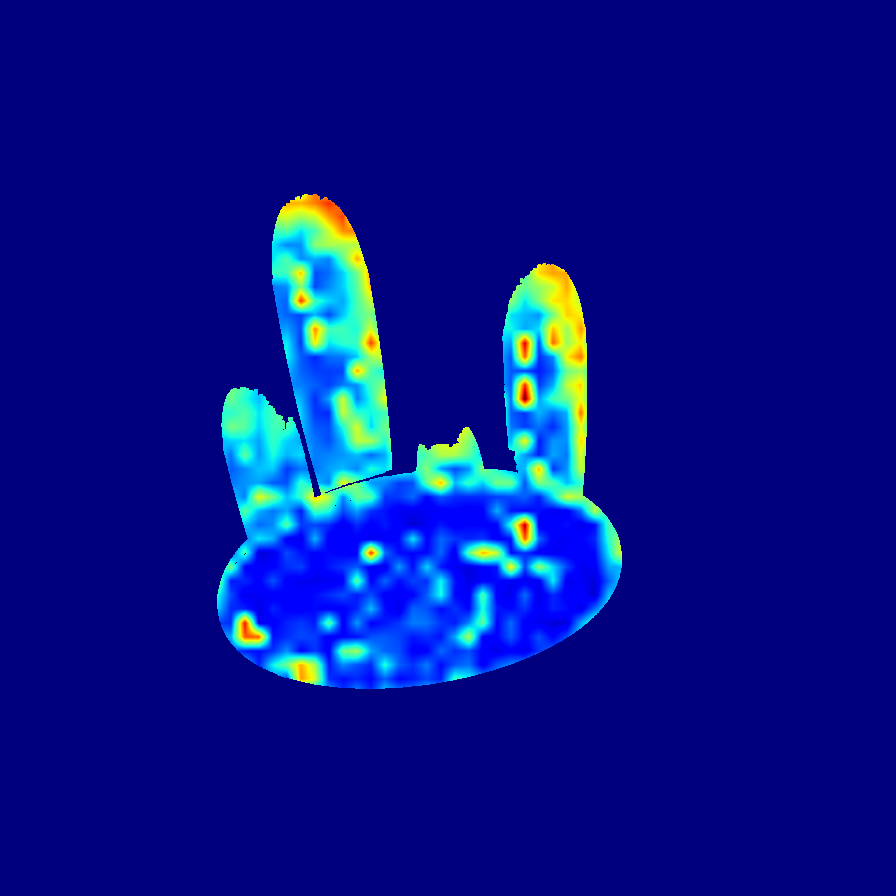} & 
            \includegraphics[width=0.08\linewidth,angle=180,origin=c]{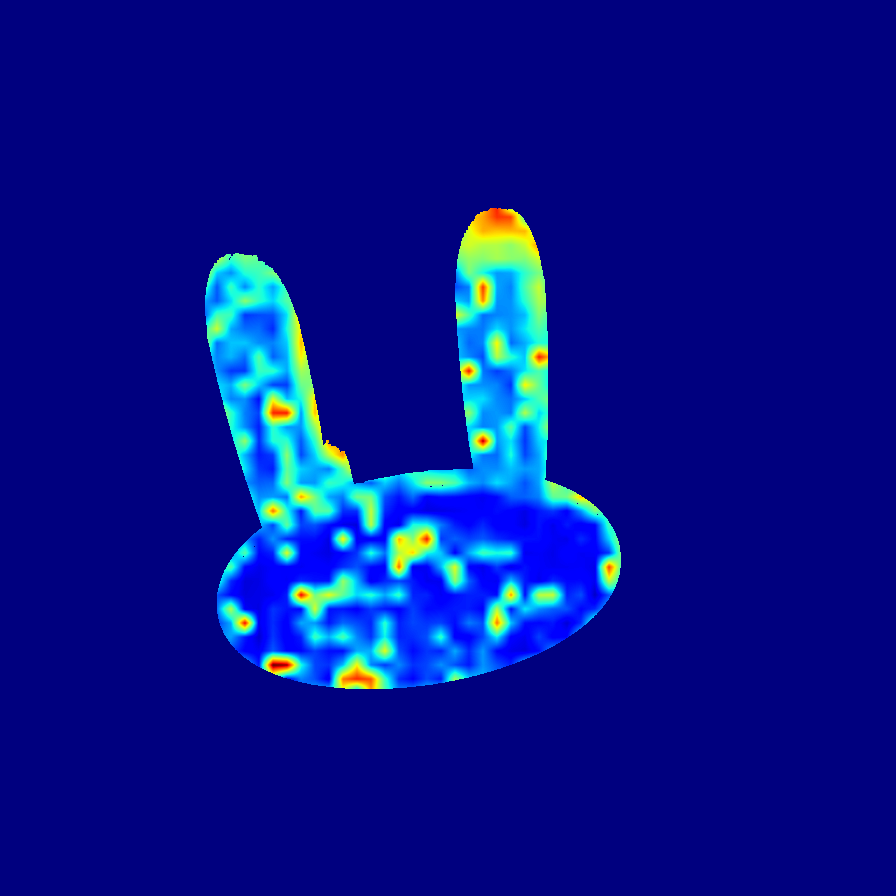} & 
            \includegraphics[width=0.08\linewidth,angle=180,origin=c]{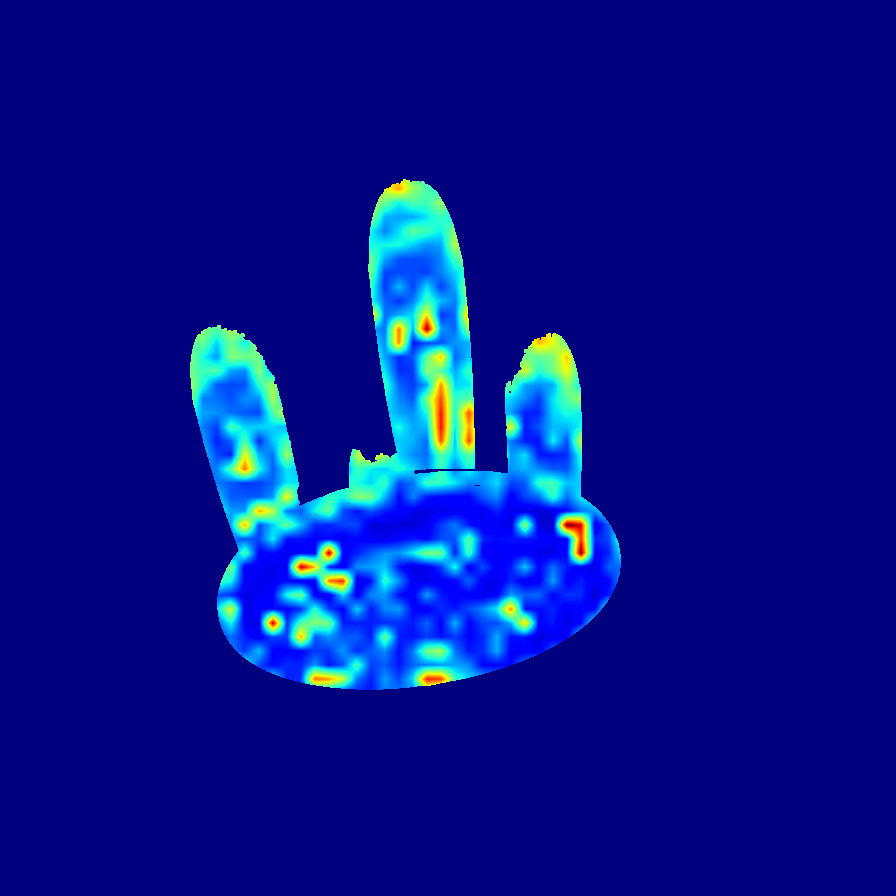} & 
            \includegraphics[width=0.08\linewidth,angle=180,origin=c]{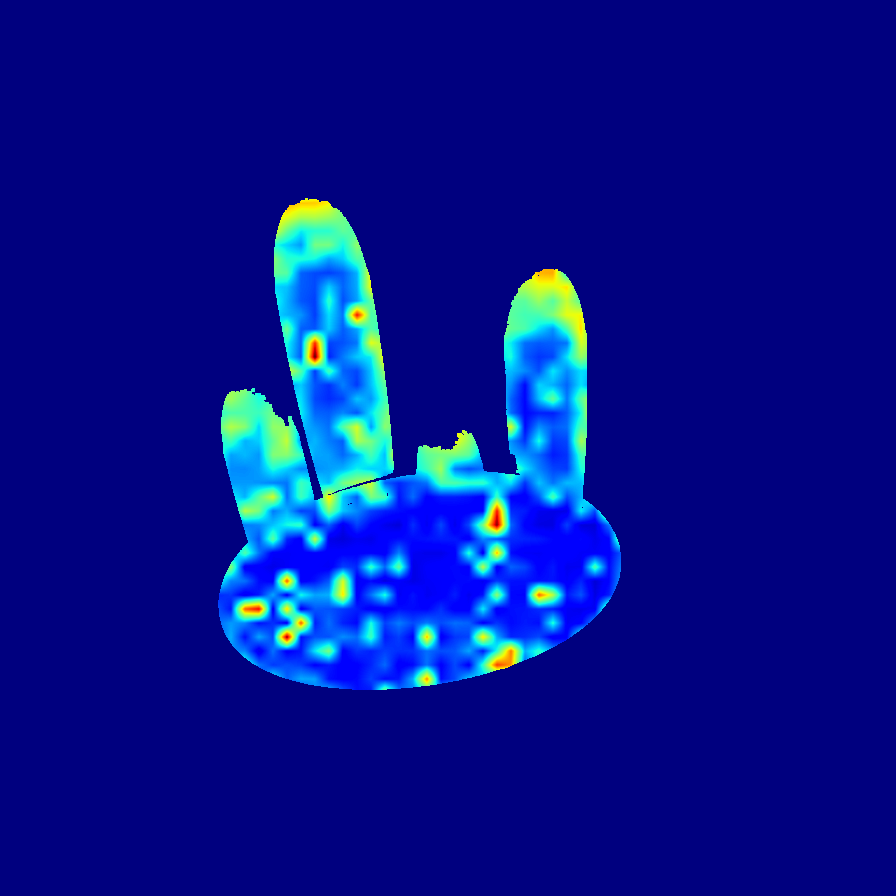} & 
            \includegraphics[width=0.08\linewidth,angle=180,origin=c]{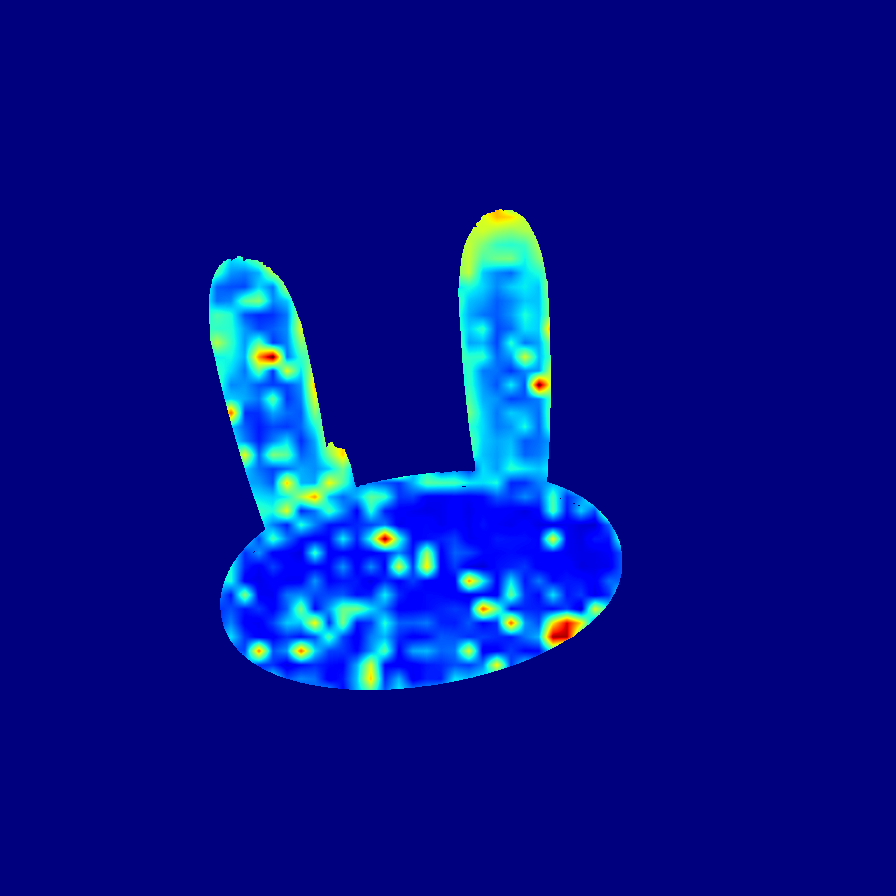} &
            \includegraphics[width=0.08\linewidth,angle=180,origin=c]{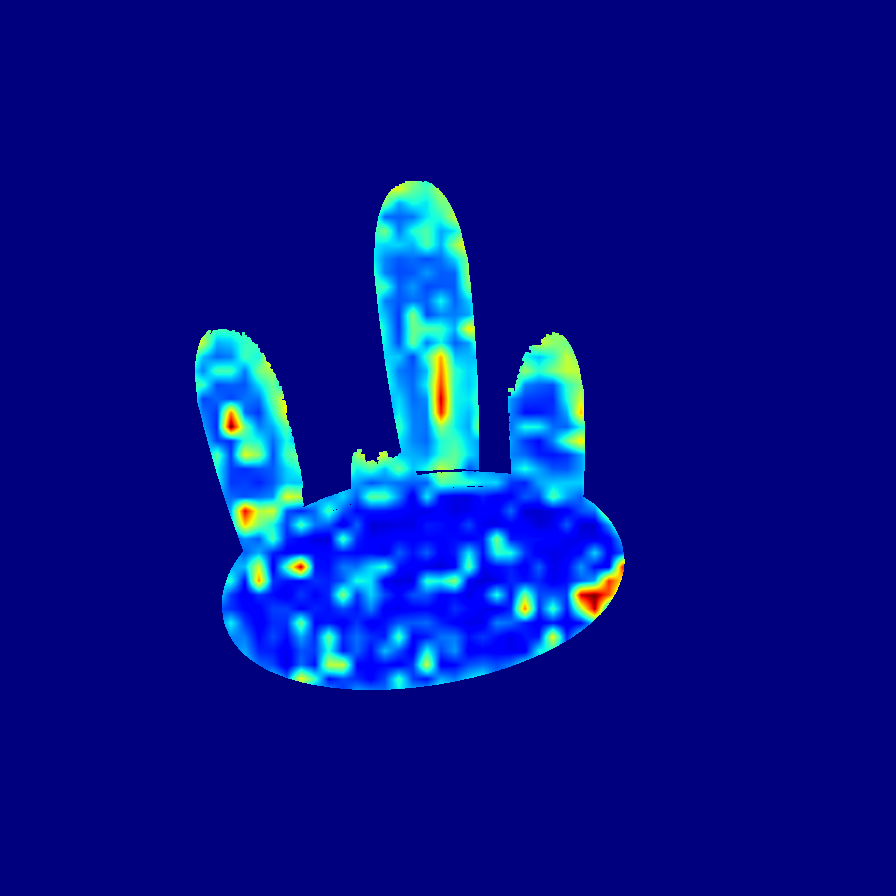} \\

            & \rotatebox{90}{\quad $v_{10}$} &
            \includegraphics[width=0.08\linewidth,angle=180,origin=c]{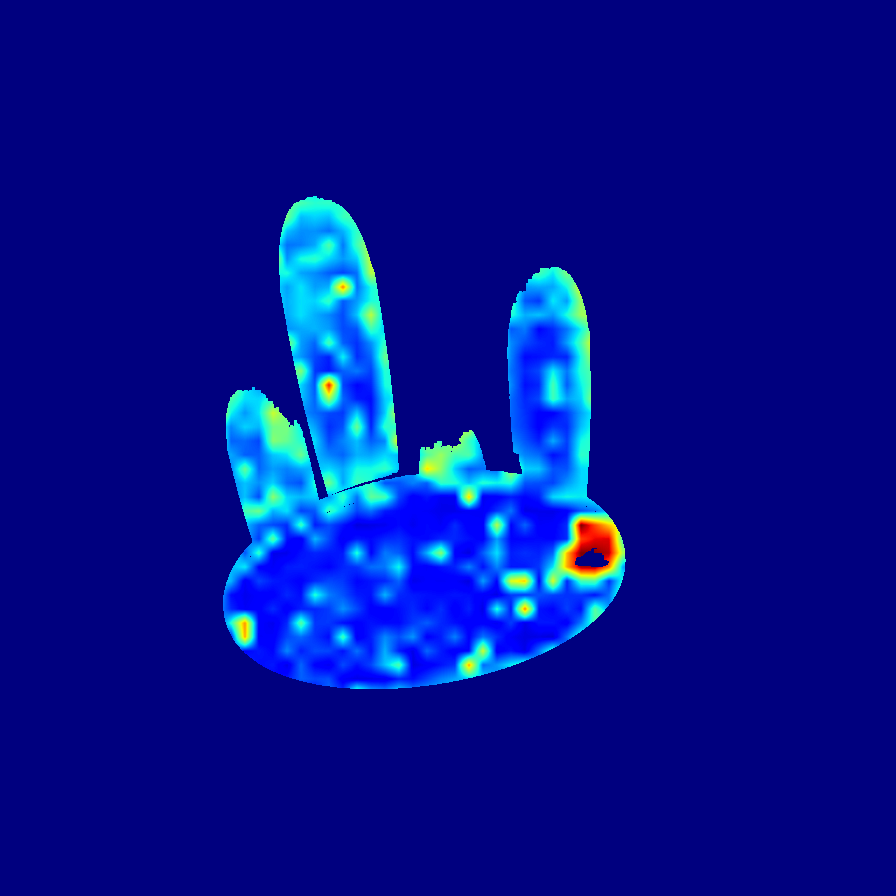} & 
            \includegraphics[width=0.08\linewidth,angle=180,origin=c]{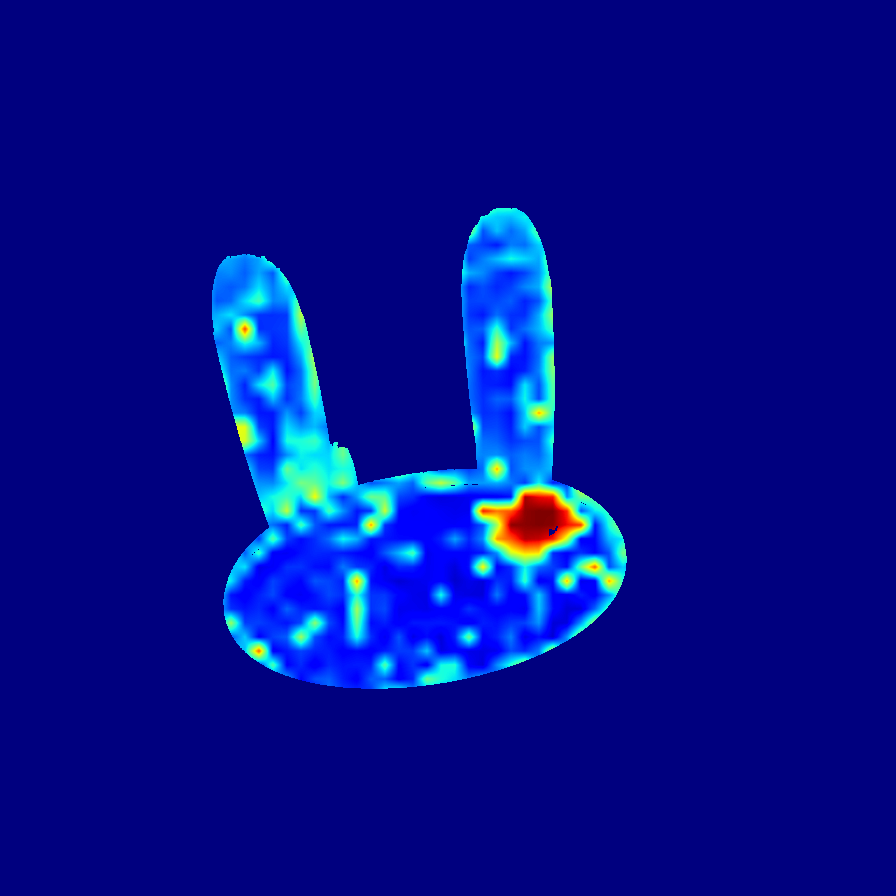} & 
            \includegraphics[width=0.08\linewidth,angle=180,origin=c]{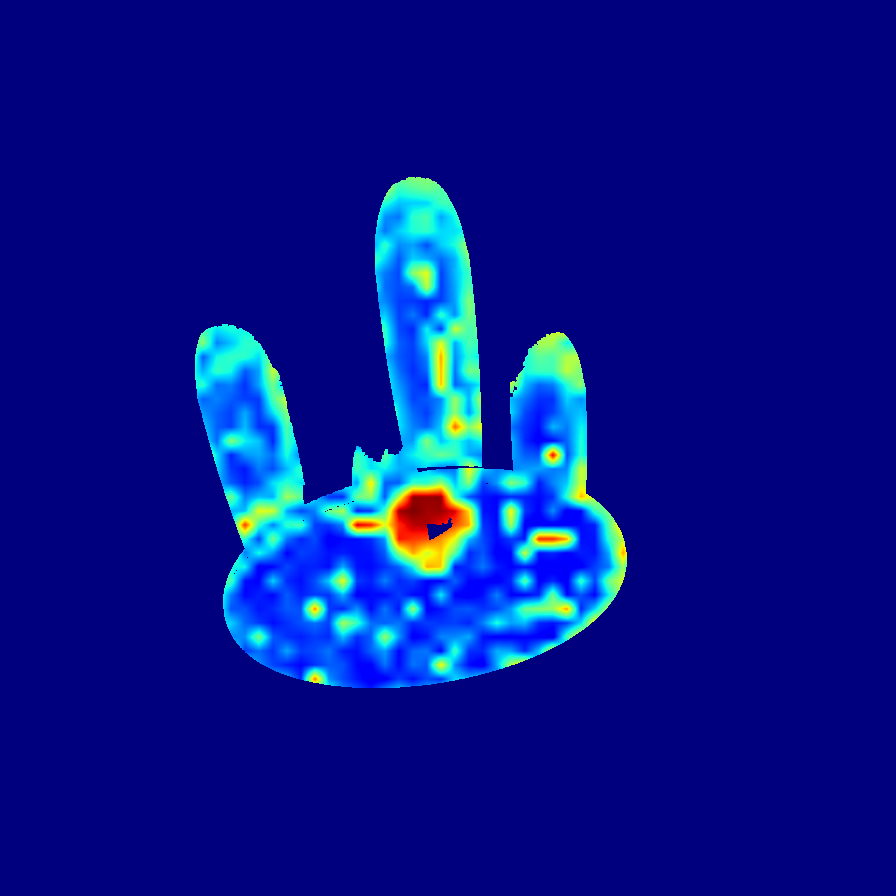} & 
            \includegraphics[width=0.08\linewidth,angle=180,origin=c]{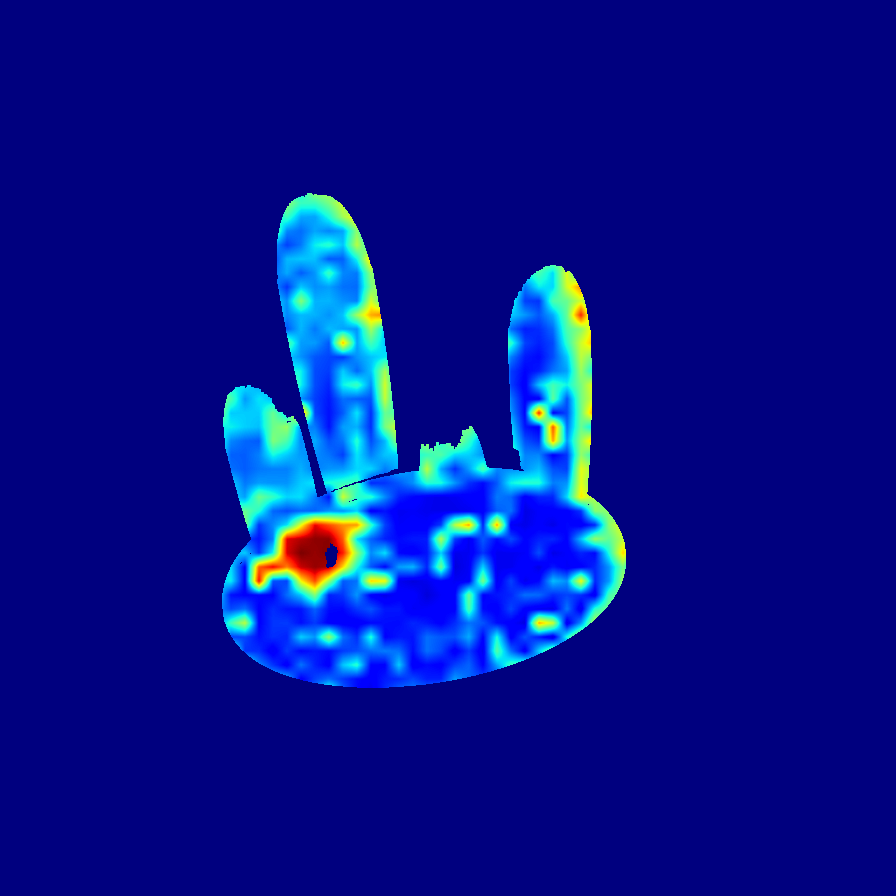} & 
            \includegraphics[width=0.08\linewidth,angle=180,origin=c]{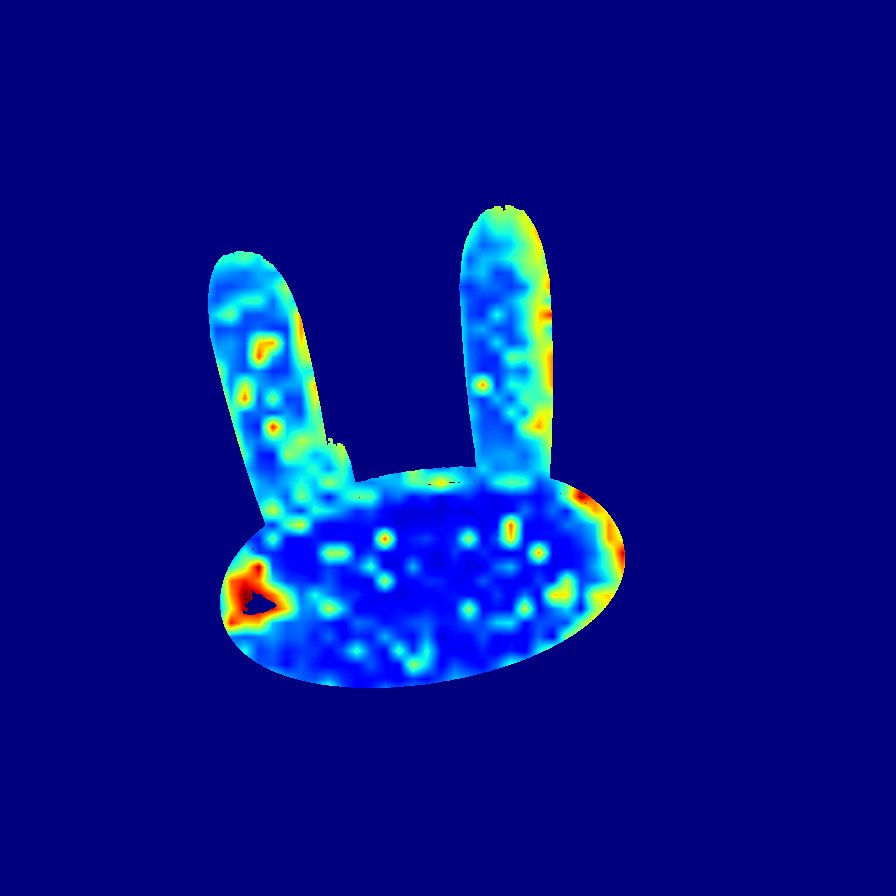} & 
            \includegraphics[width=0.08\linewidth,angle=180,origin=c]{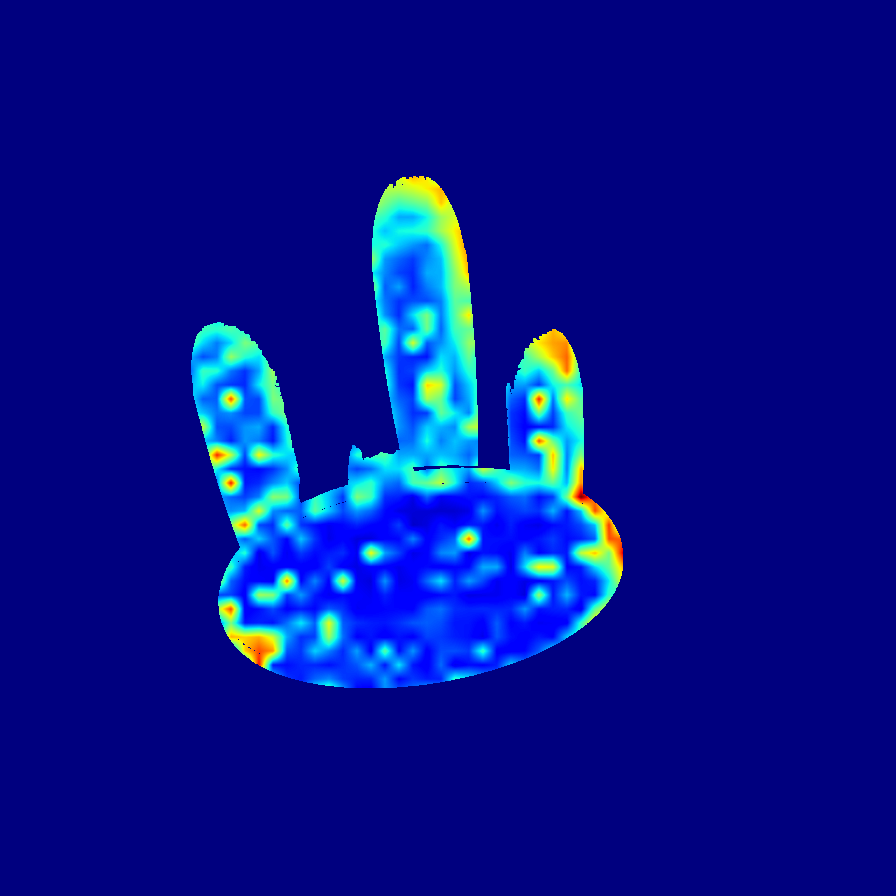} & 
            \includegraphics[width=0.08\linewidth,angle=180,origin=c]{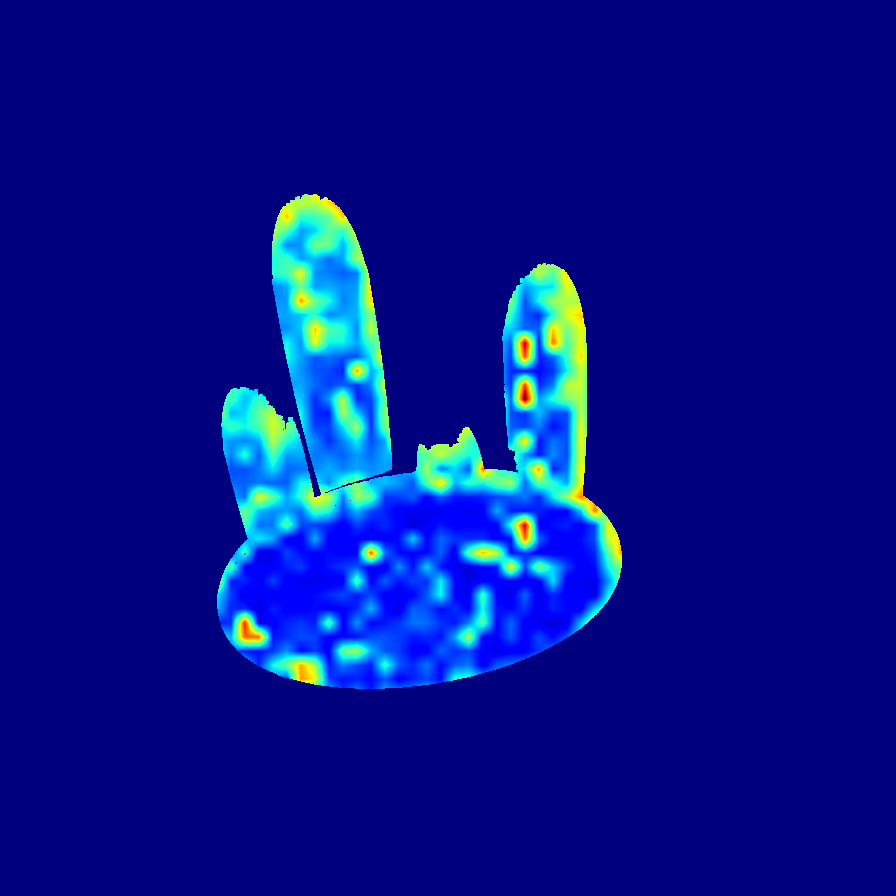} & 
            \includegraphics[width=0.08\linewidth,angle=180,origin=c]{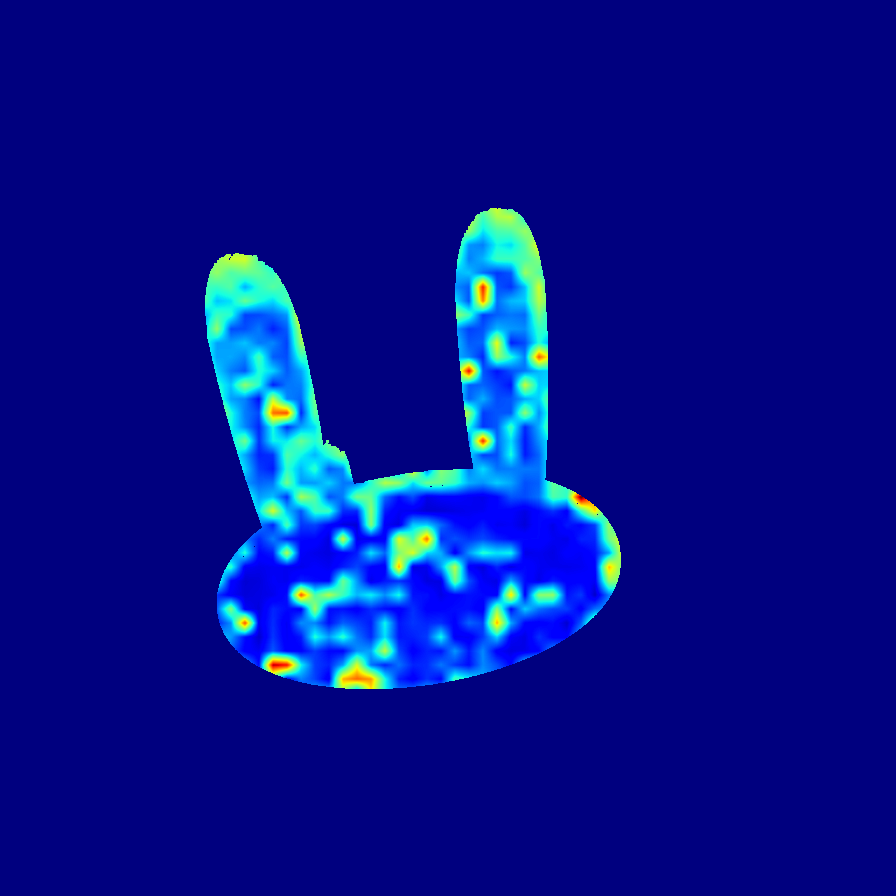} & 
            \includegraphics[width=0.08\linewidth,angle=180,origin=c]{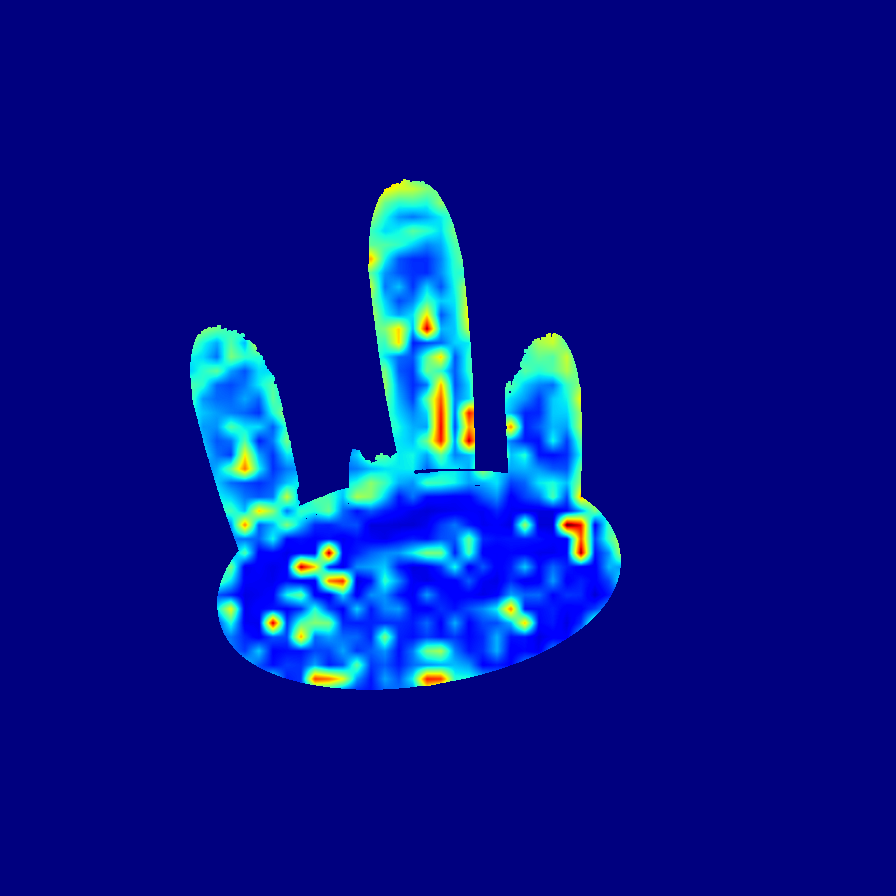} & 
            \includegraphics[width=0.08\linewidth,angle=180,origin=c]{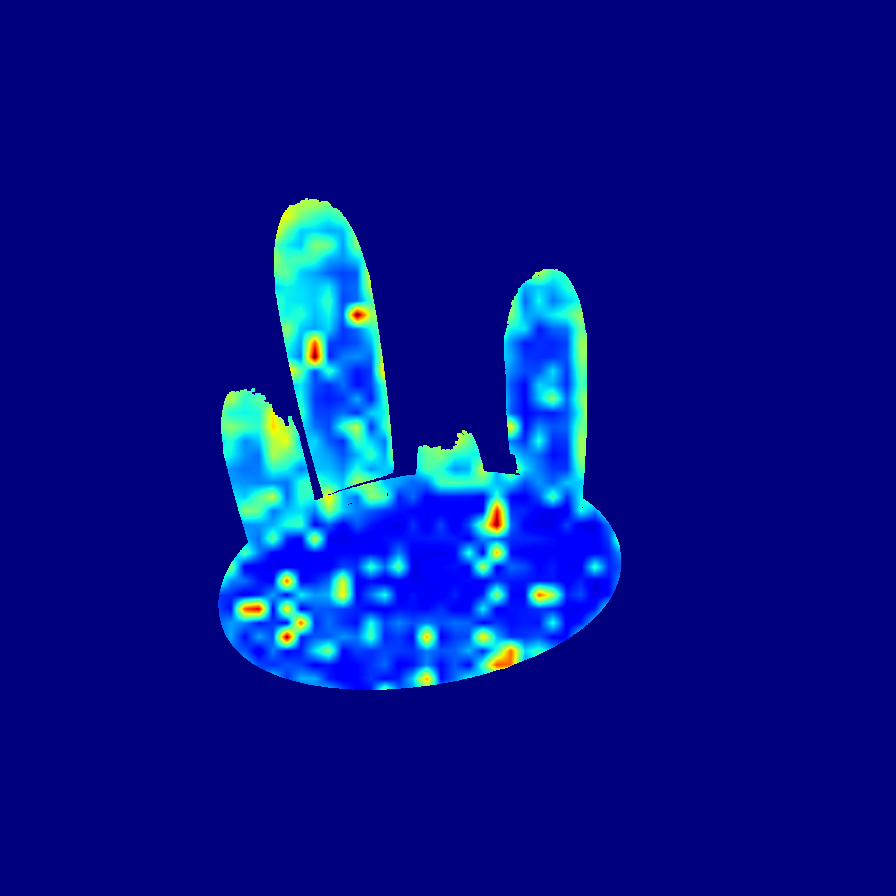} & 
            \includegraphics[width=0.08\linewidth,angle=180,origin=c]{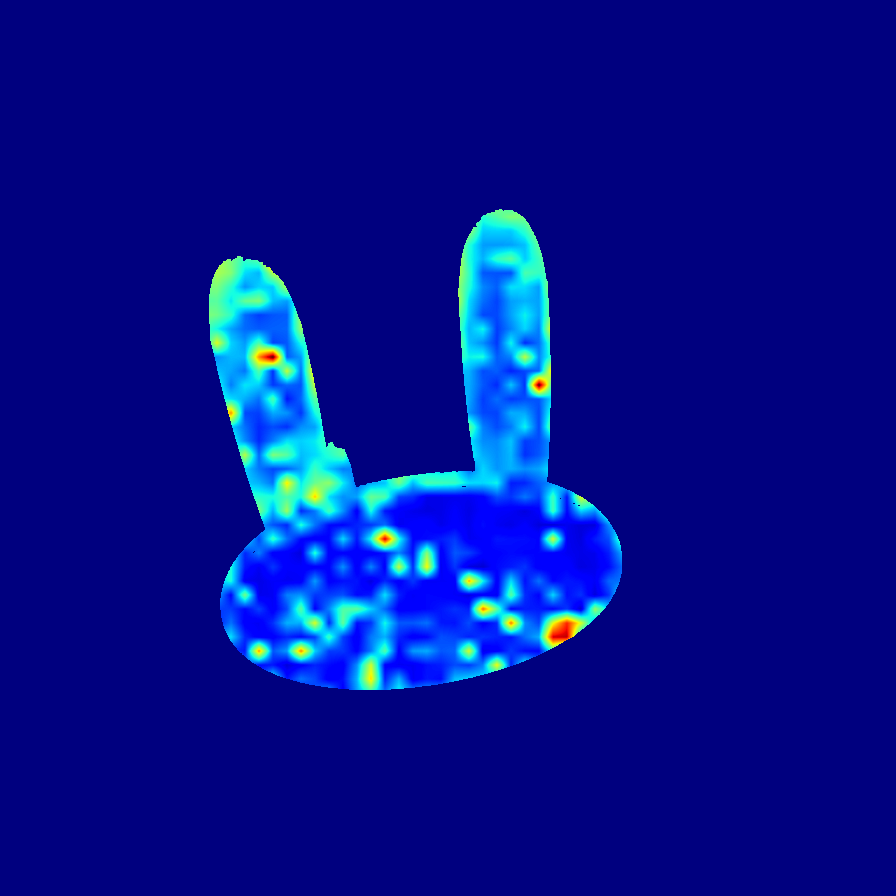} &
            \includegraphics[width=0.08\linewidth,angle=180,origin=c]{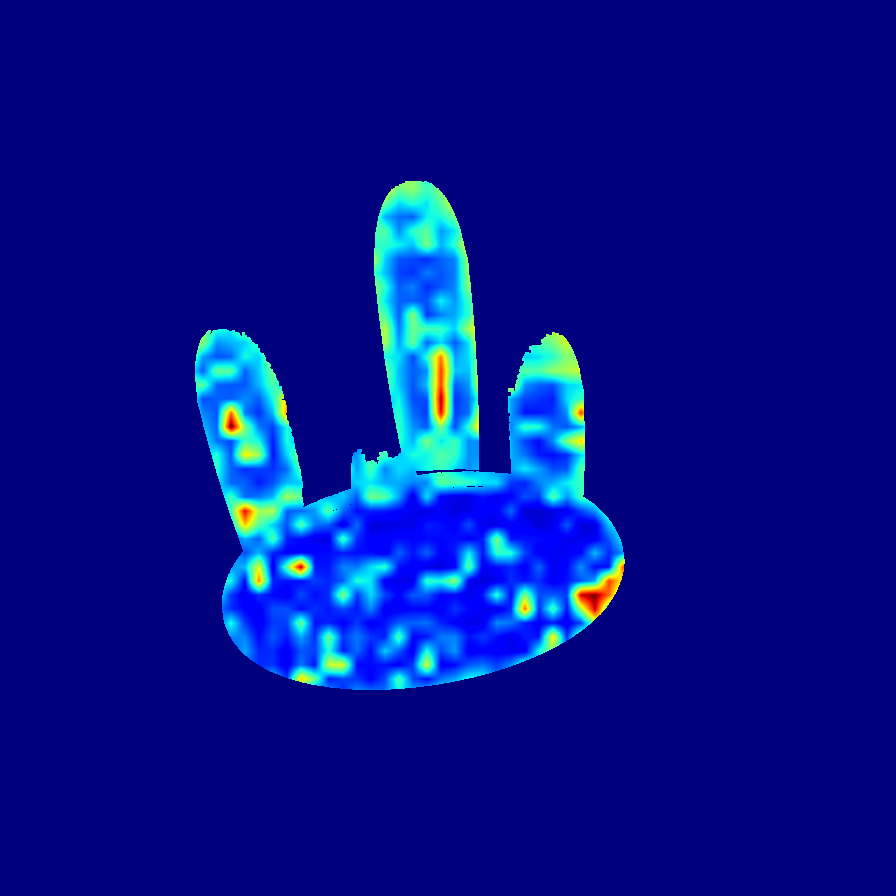} \\

            & \rotatebox{90}{\quad $v_{11}$} &
            \includegraphics[width=0.08\linewidth,angle=180,origin=c]{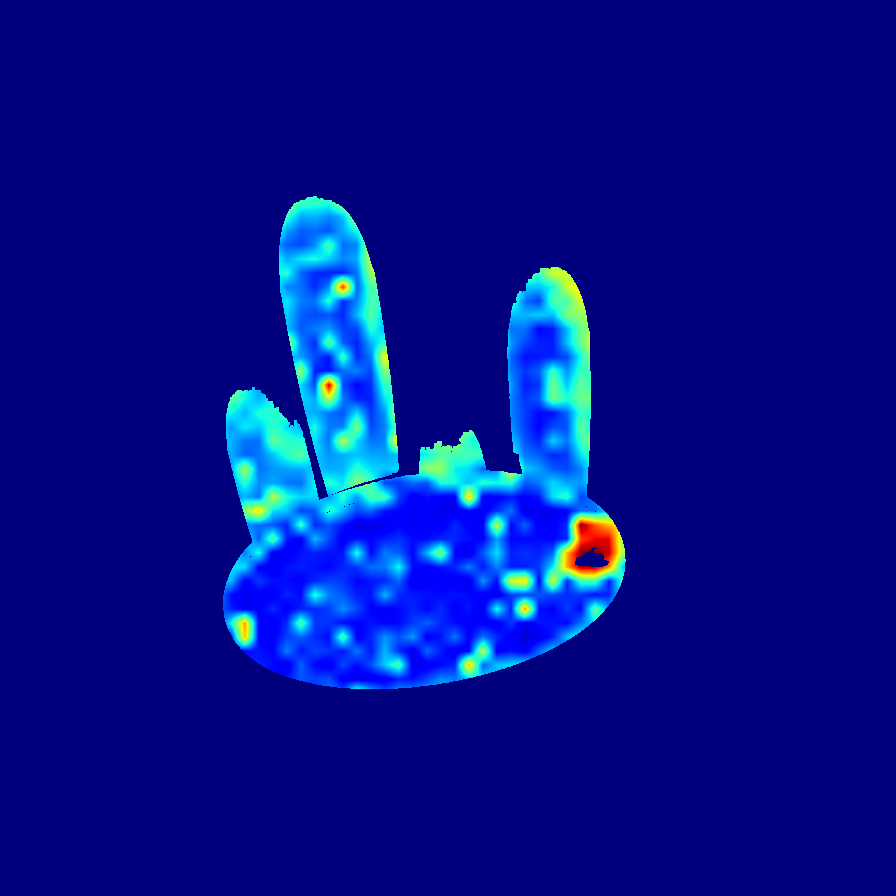} & 
            \includegraphics[width=0.08\linewidth,angle=180,origin=c]{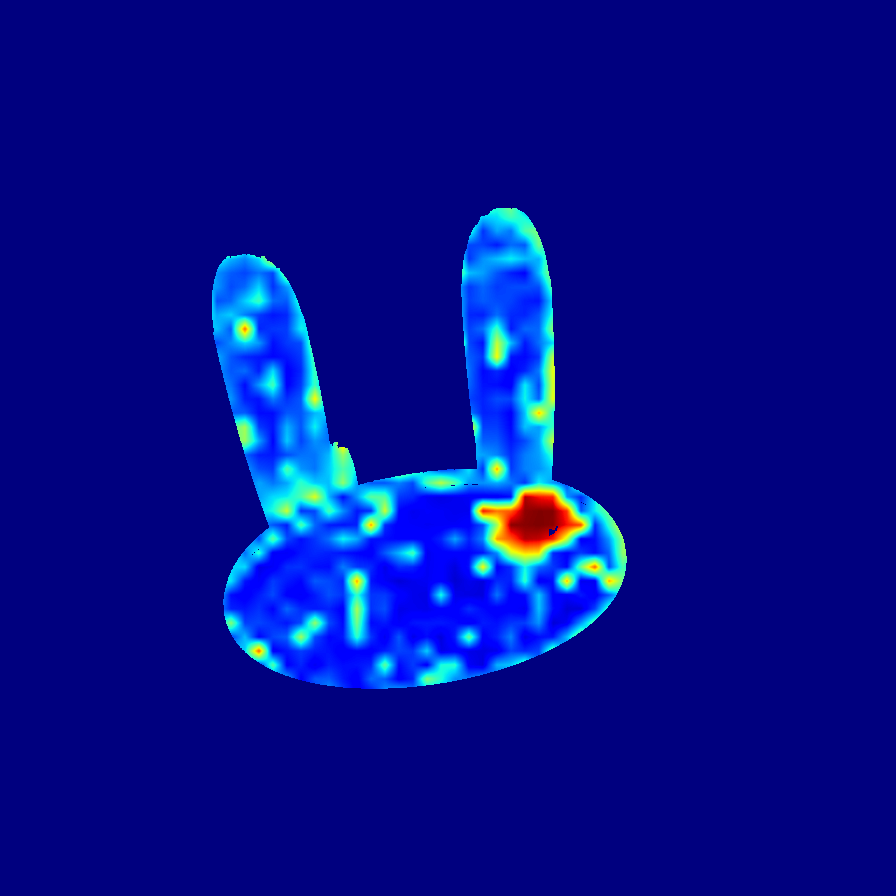} & 
            \includegraphics[width=0.08\linewidth,angle=180,origin=c]{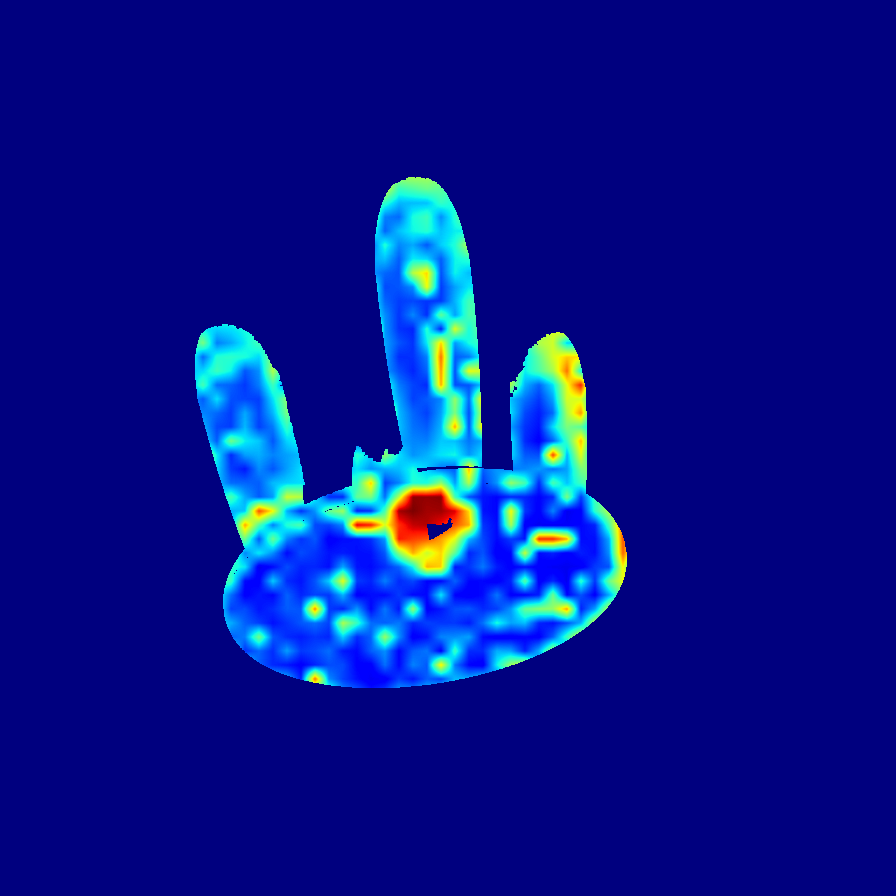} & 
            \includegraphics[width=0.08\linewidth,angle=180,origin=c]{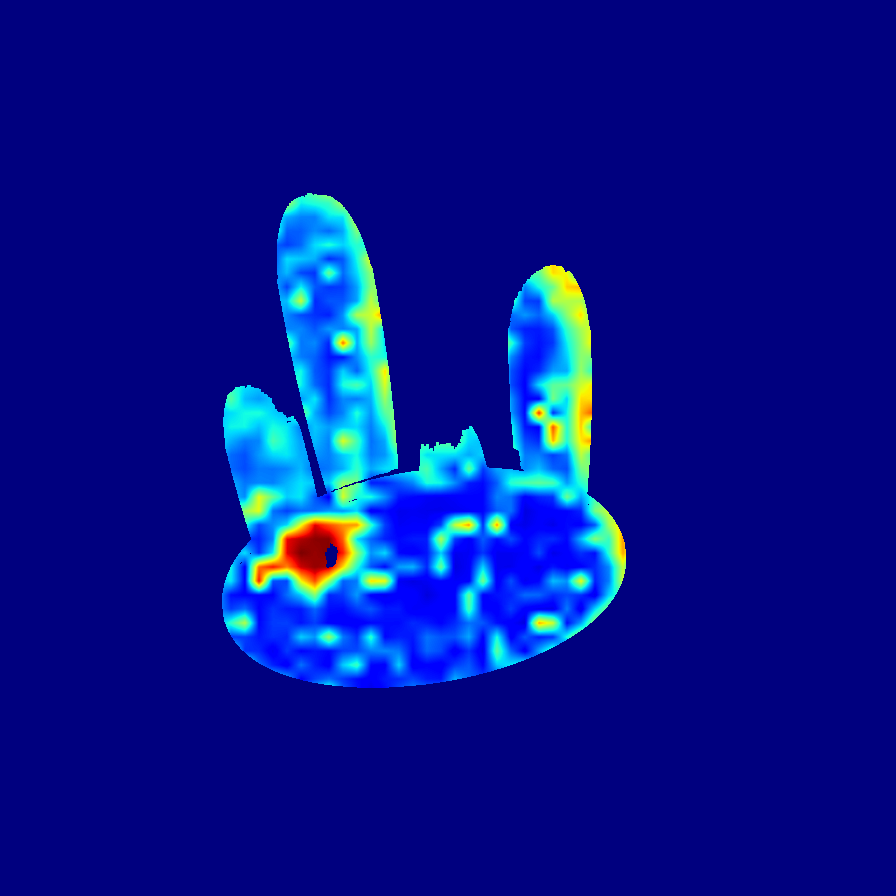} & 
            \includegraphics[width=0.08\linewidth,angle=180,origin=c]{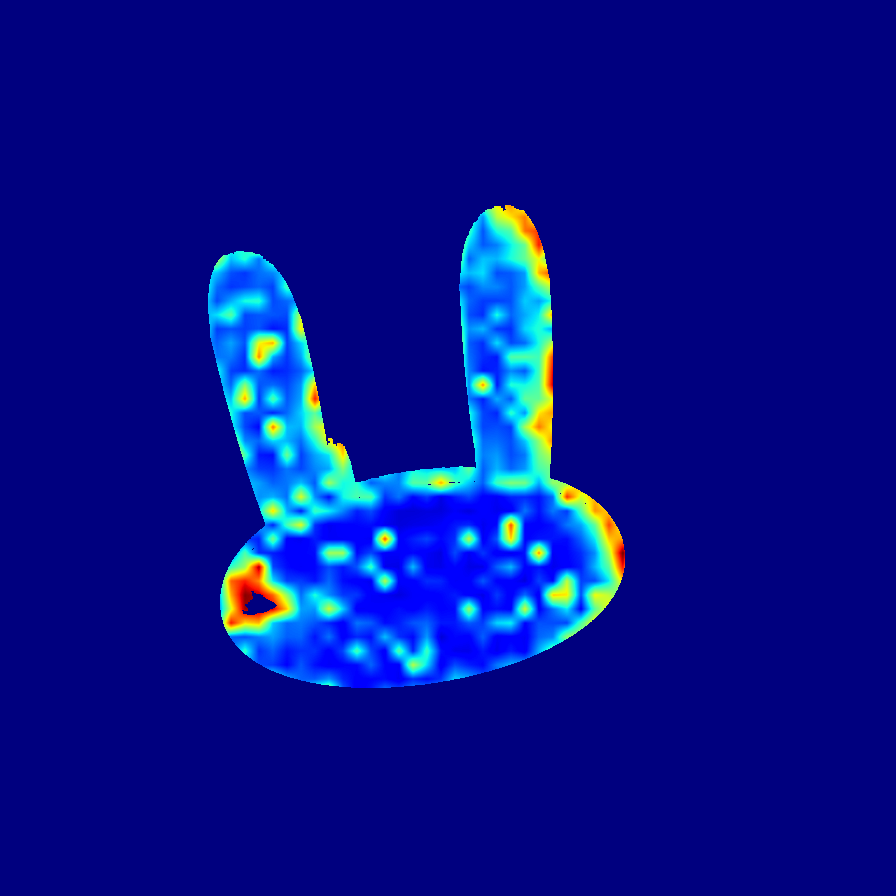} & 
            \includegraphics[width=0.08\linewidth,angle=180,origin=c]{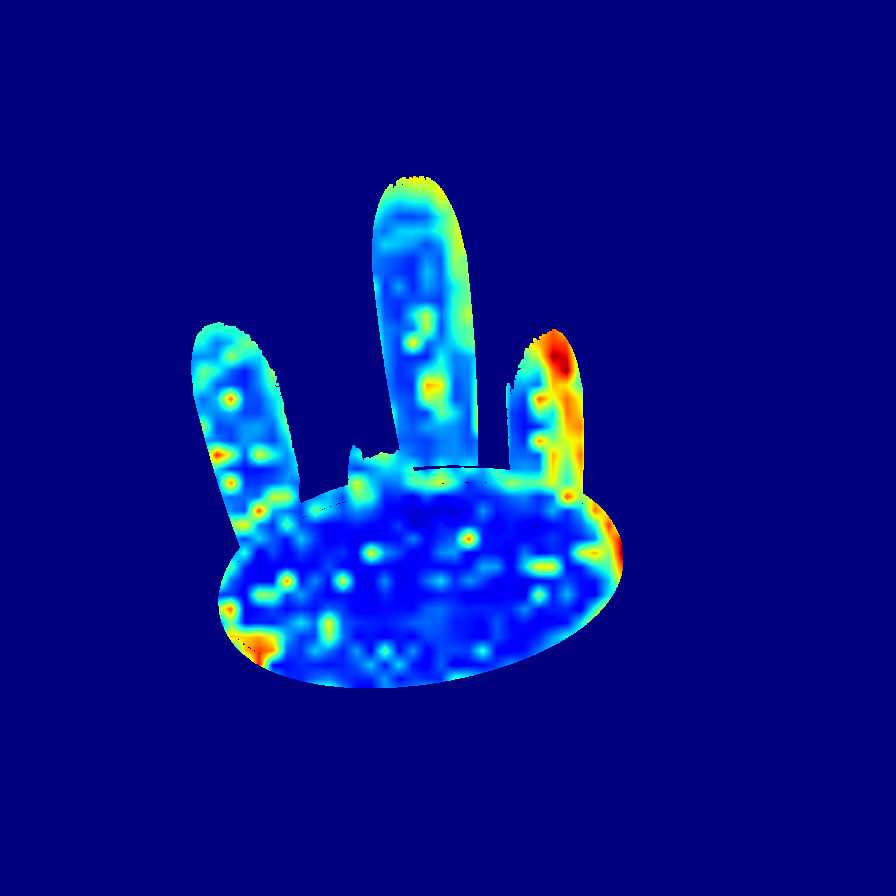} & 
            \includegraphics[width=0.08\linewidth,angle=180,origin=c]{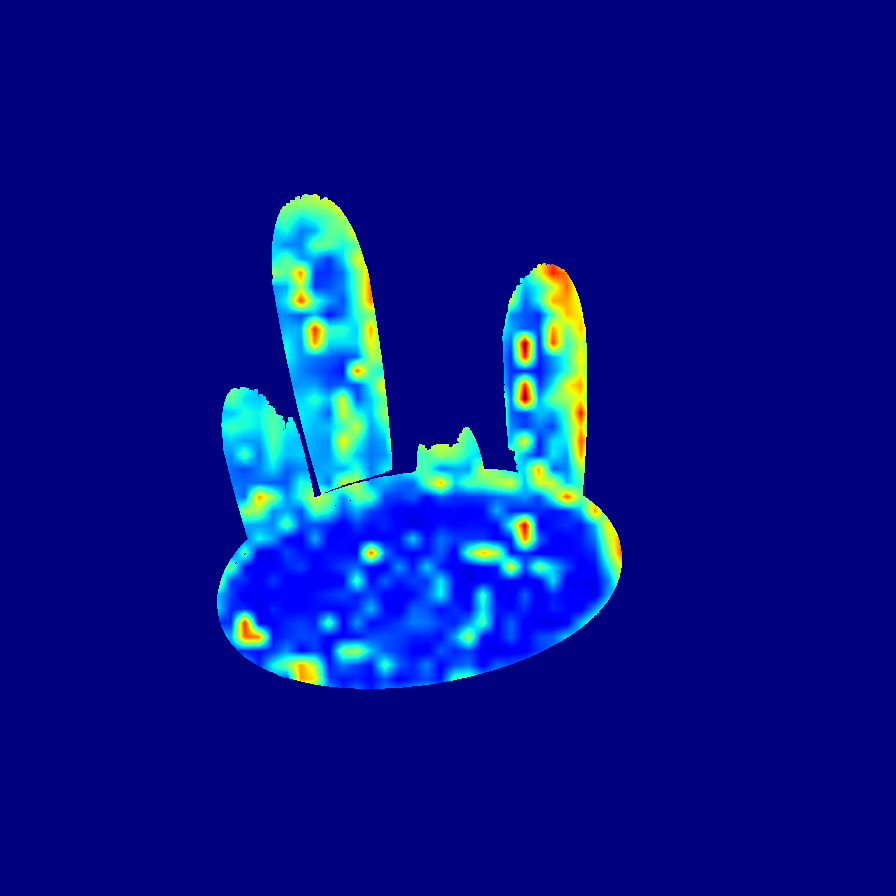} & 
            \includegraphics[width=0.08\linewidth,angle=180,origin=c]{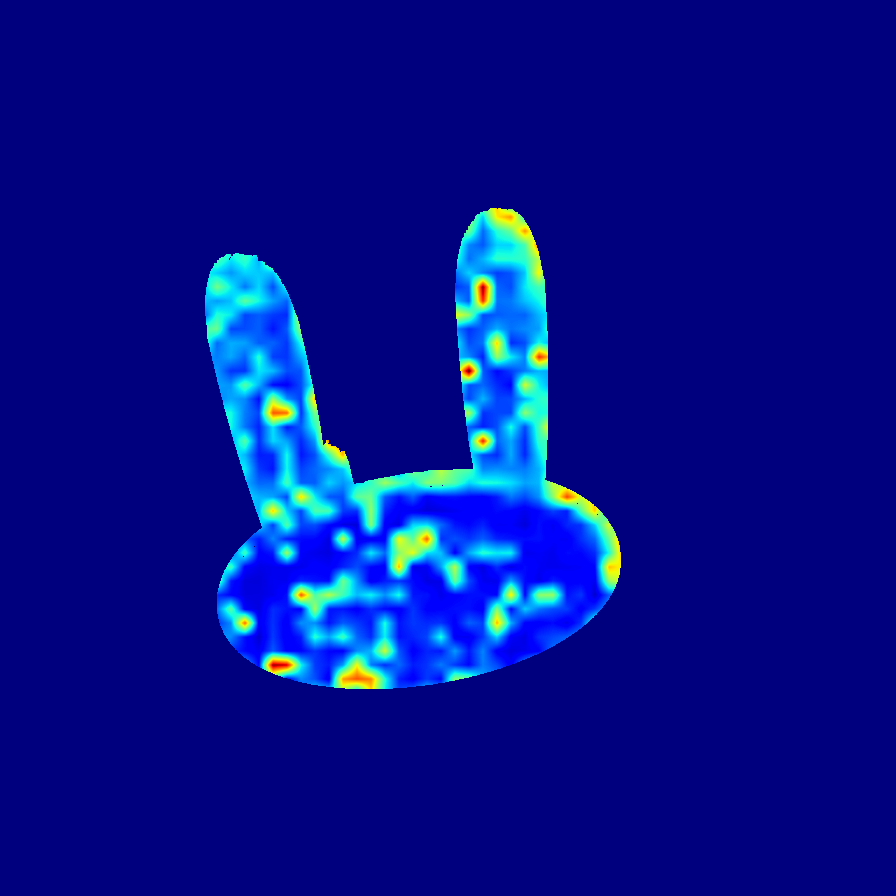} & 
            \includegraphics[width=0.08\linewidth,angle=180,origin=c]{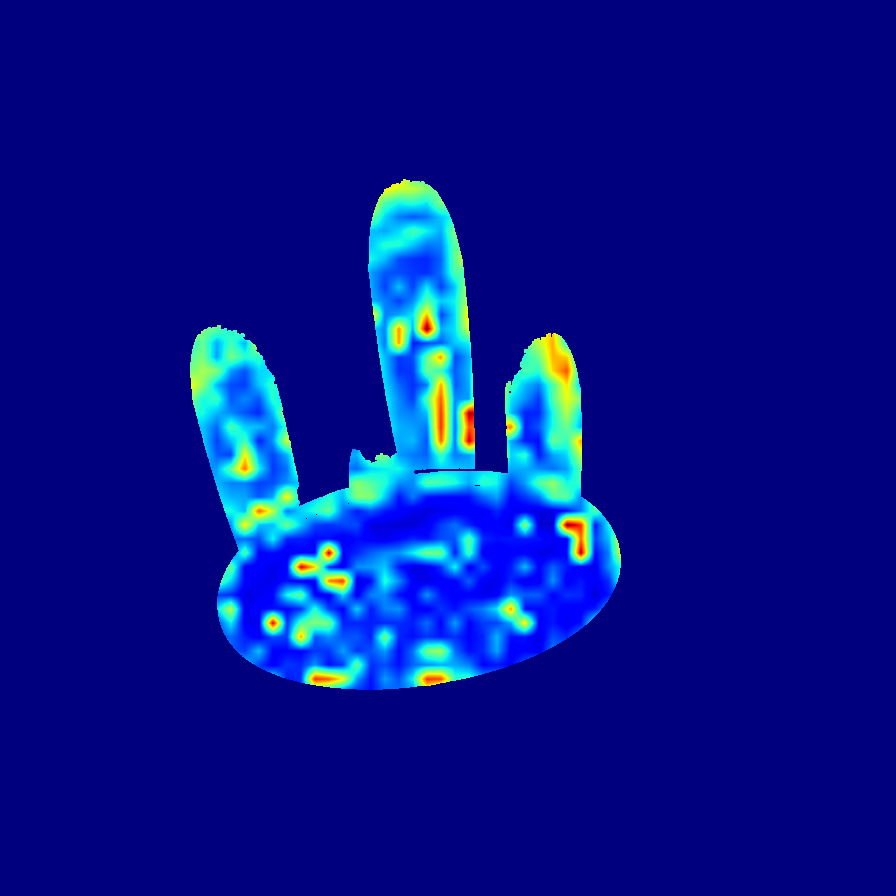} & 
            \includegraphics[width=0.08\linewidth,angle=180,origin=c]{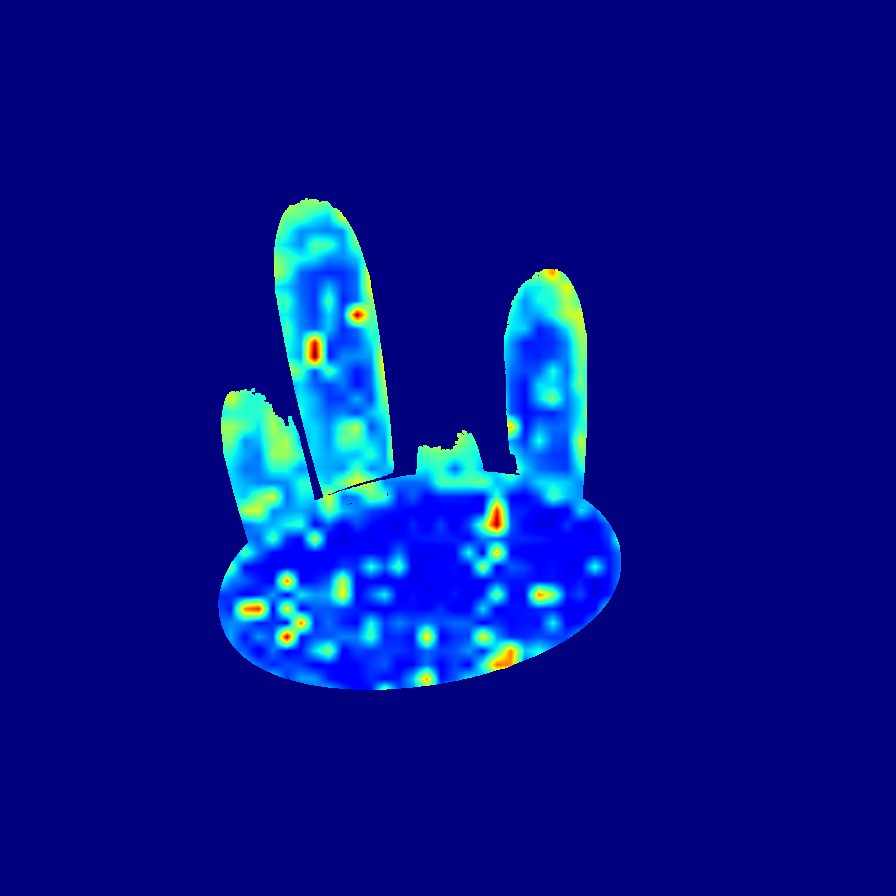} & 
            \includegraphics[width=0.08\linewidth,angle=180,origin=c]{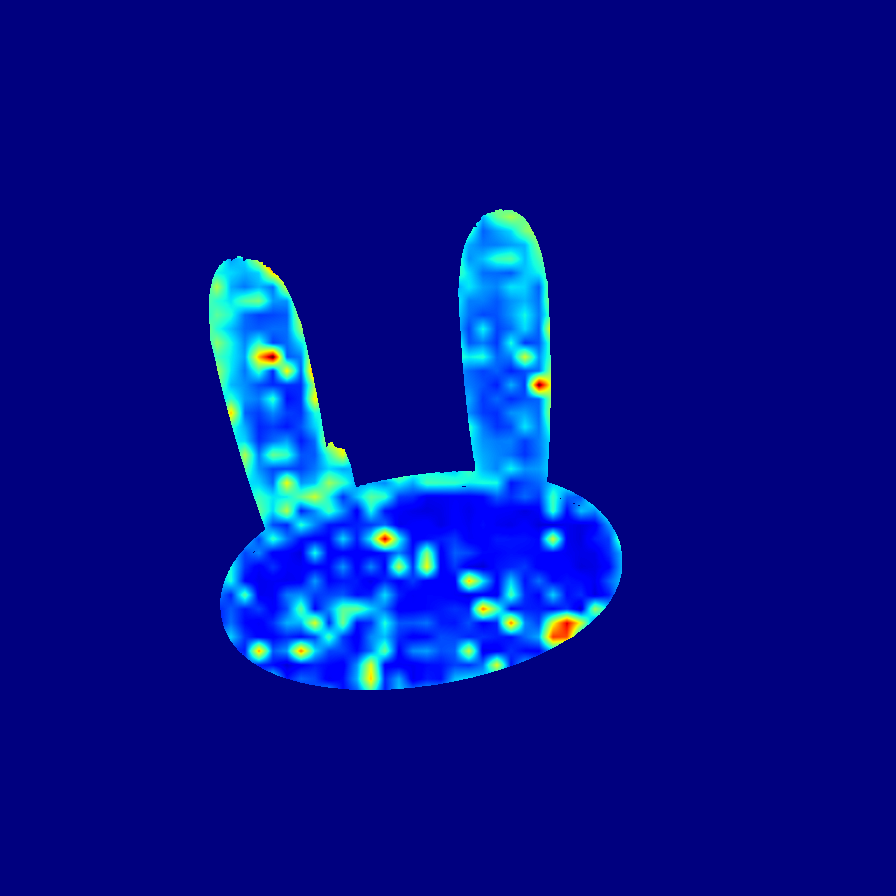} &
            \includegraphics[width=0.08\linewidth,angle=180,origin=c]{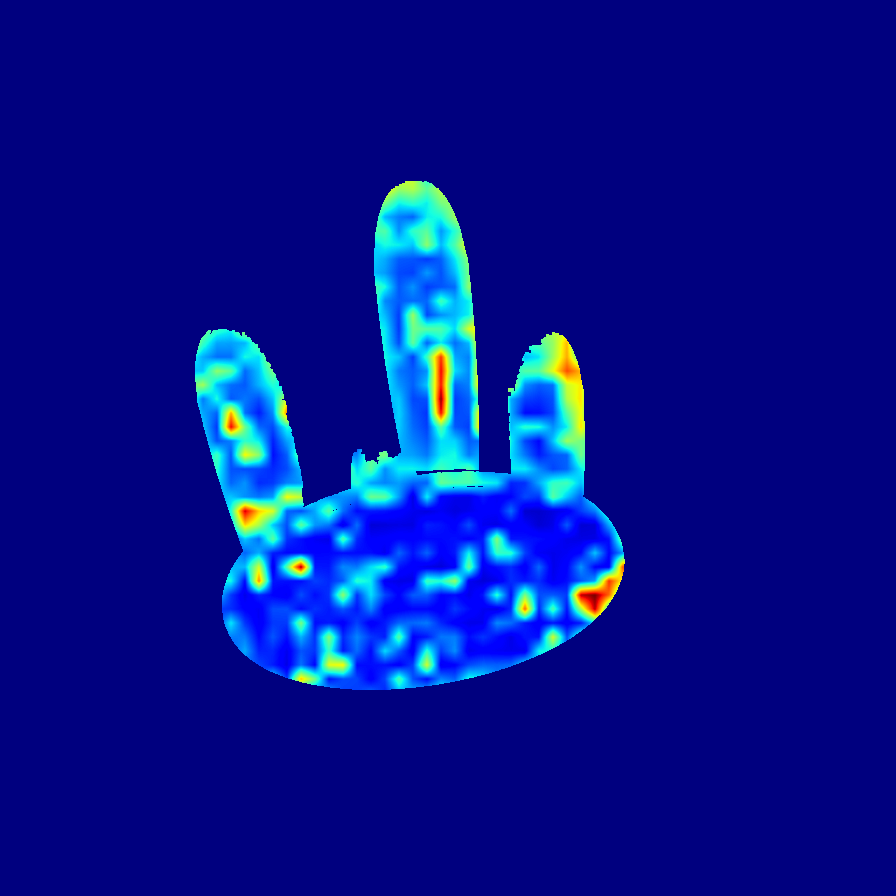} \\

            & \rotatebox{90}{\quad $v_{12}$} &
            \includegraphics[width=0.08\linewidth,angle=180,origin=c]{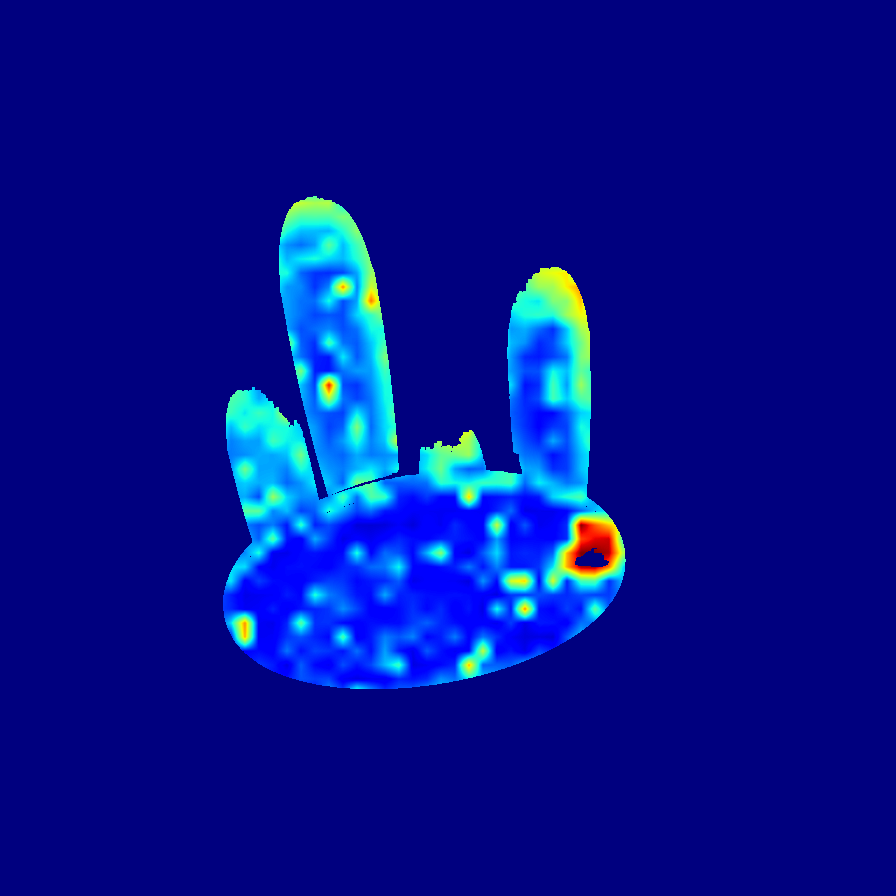} & 
            \includegraphics[width=0.08\linewidth,angle=180,origin=c]{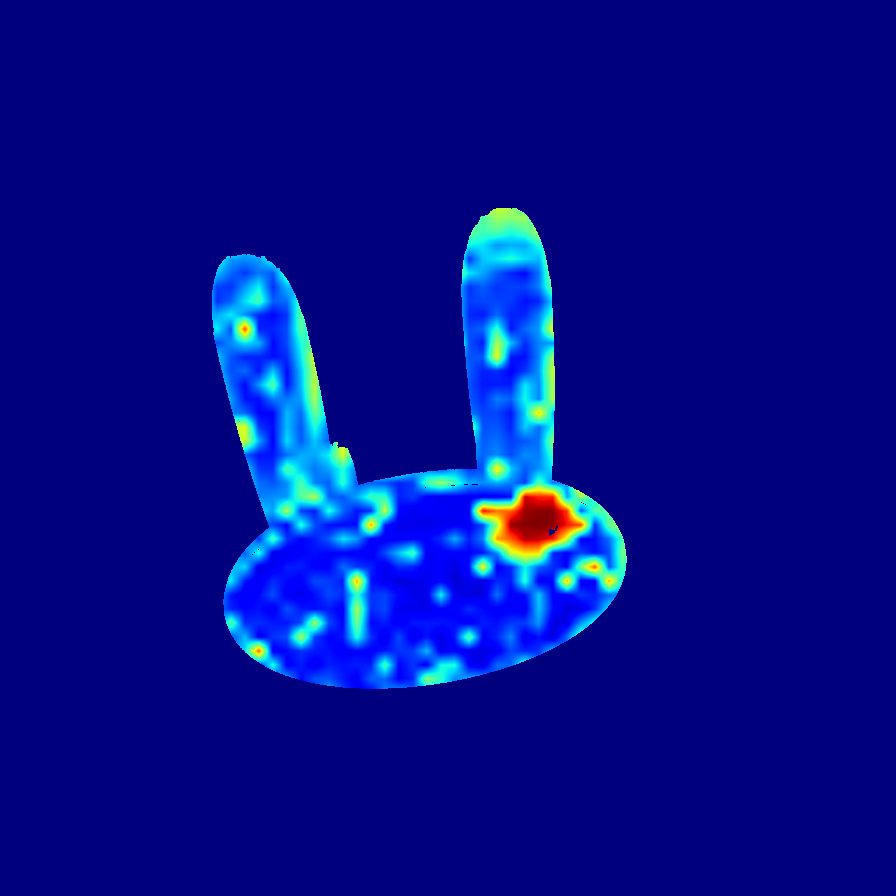} & 
            \includegraphics[width=0.08\linewidth,angle=180,origin=c]{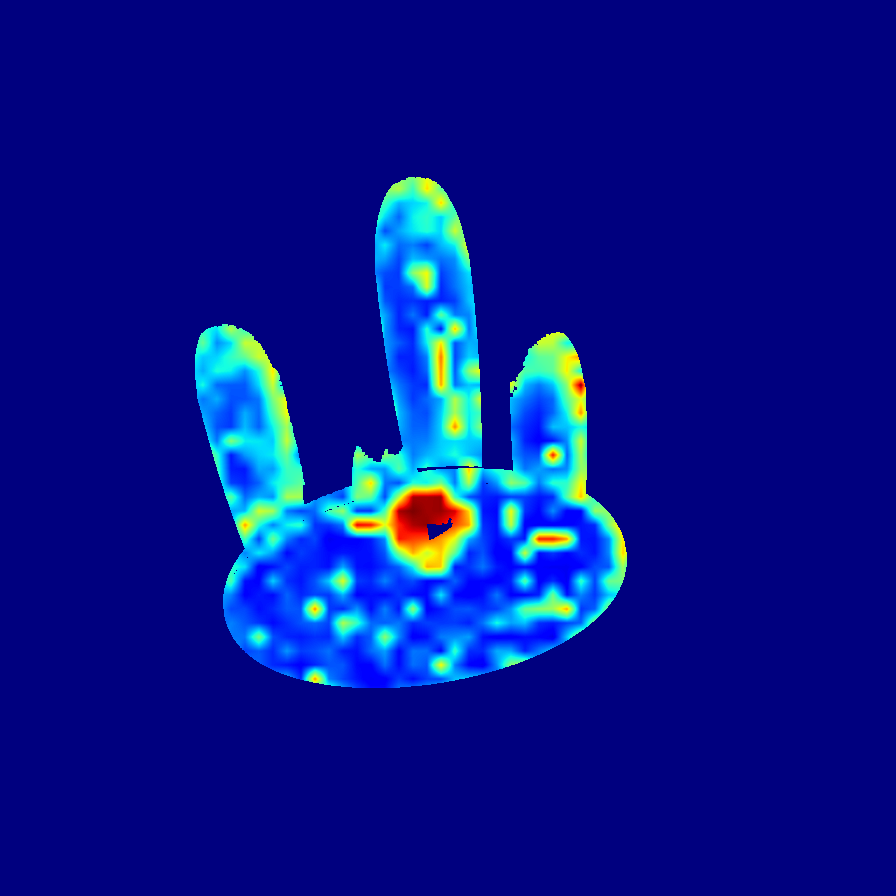} & 
            \includegraphics[width=0.08\linewidth,angle=180,origin=c]{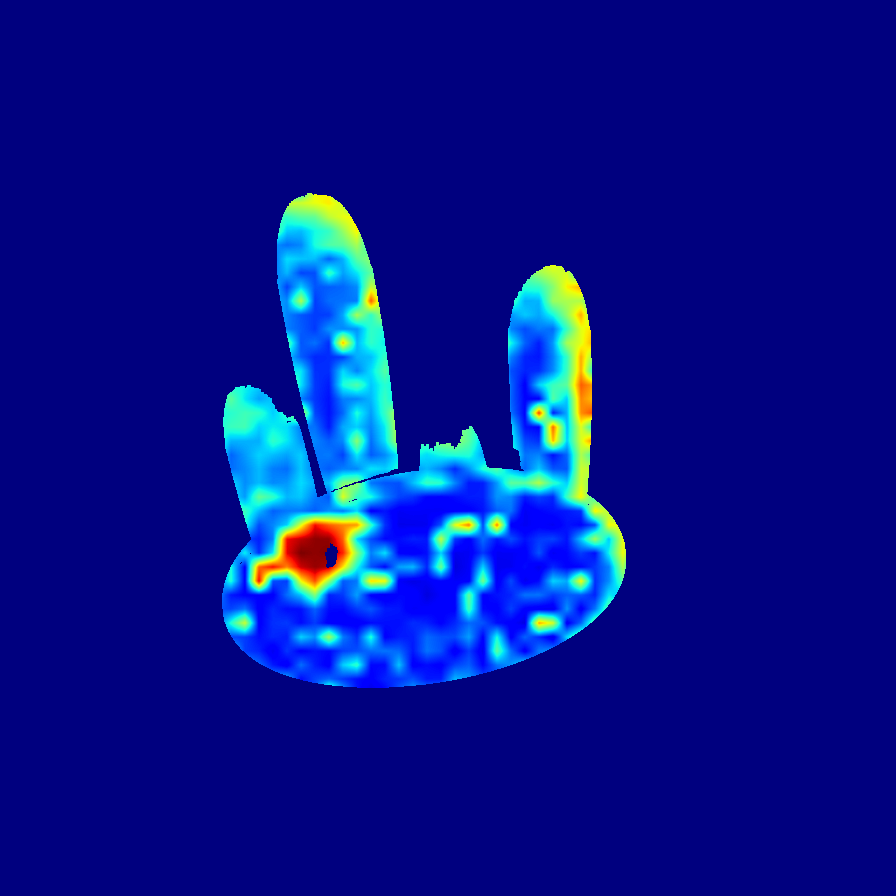} & 
            \includegraphics[width=0.08\linewidth,angle=180,origin=c]{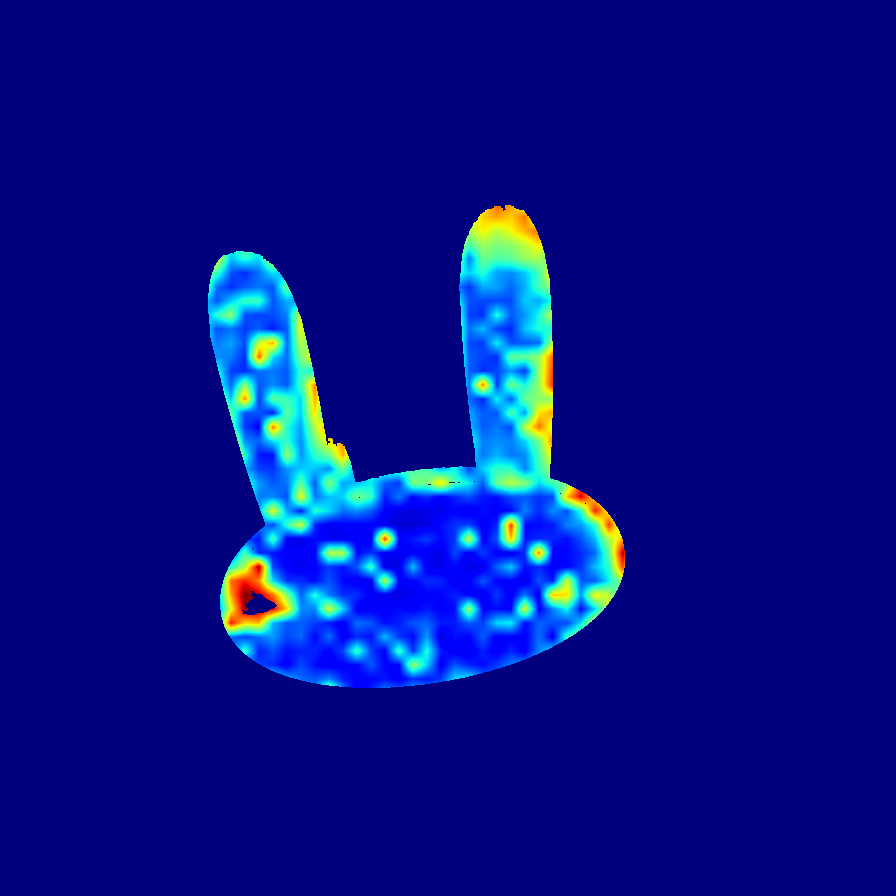} & 
            \includegraphics[width=0.08\linewidth,angle=180,origin=c]{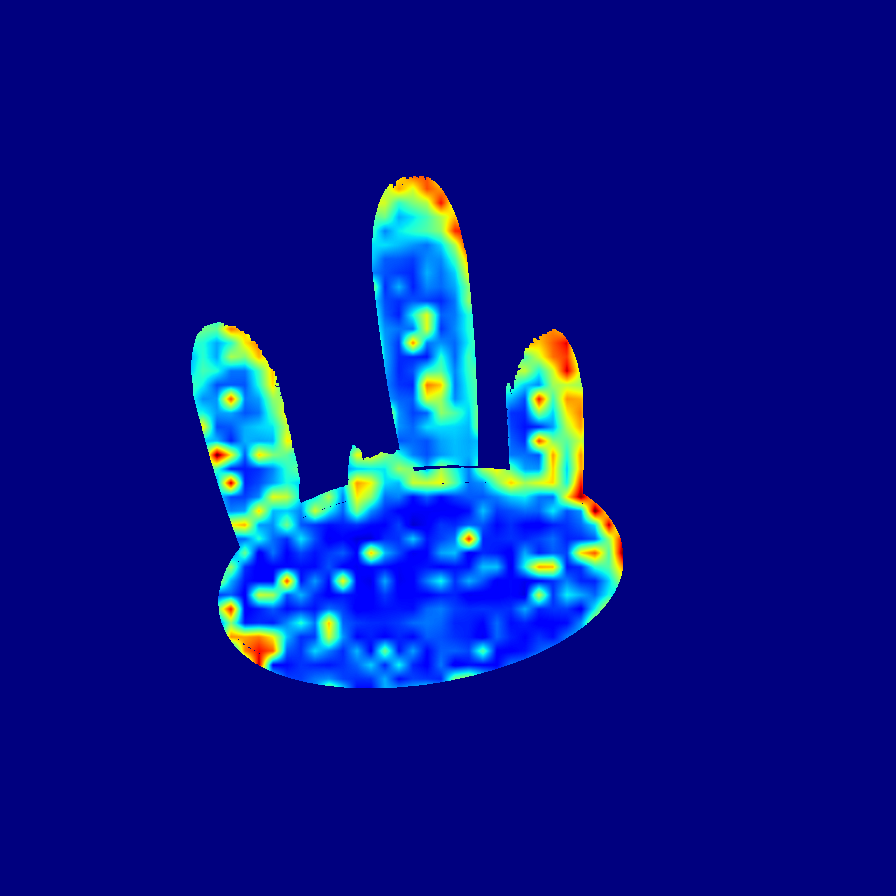} & 
            \includegraphics[width=0.08\linewidth,angle=180,origin=c]{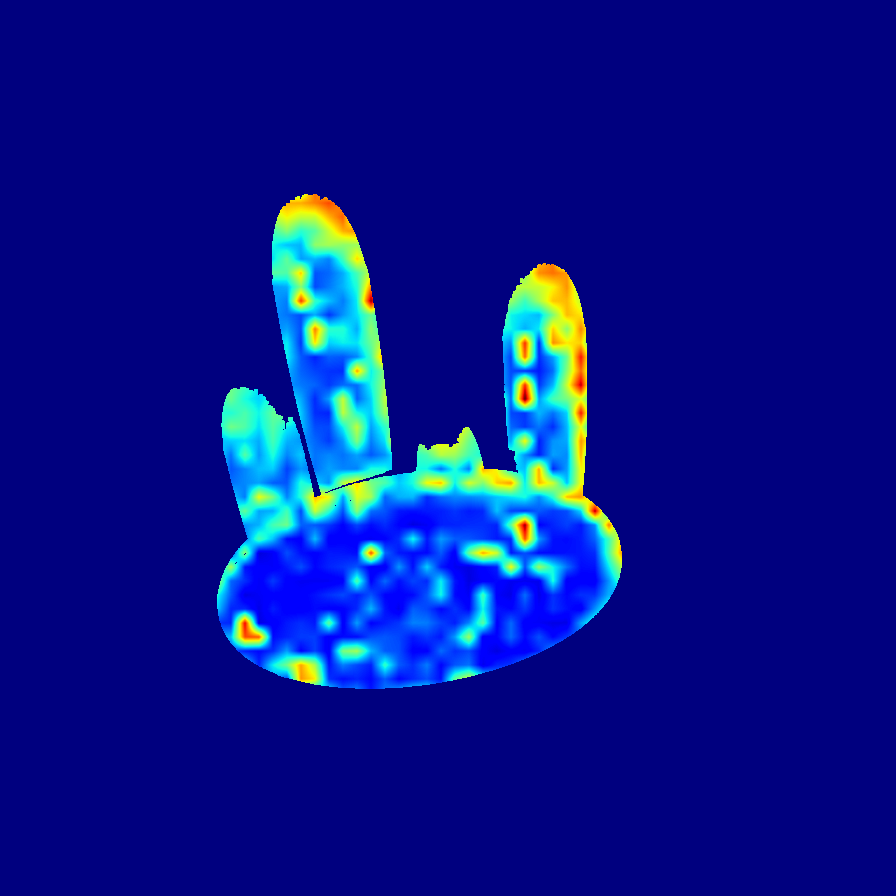} & 
            \includegraphics[width=0.08\linewidth,angle=180,origin=c]{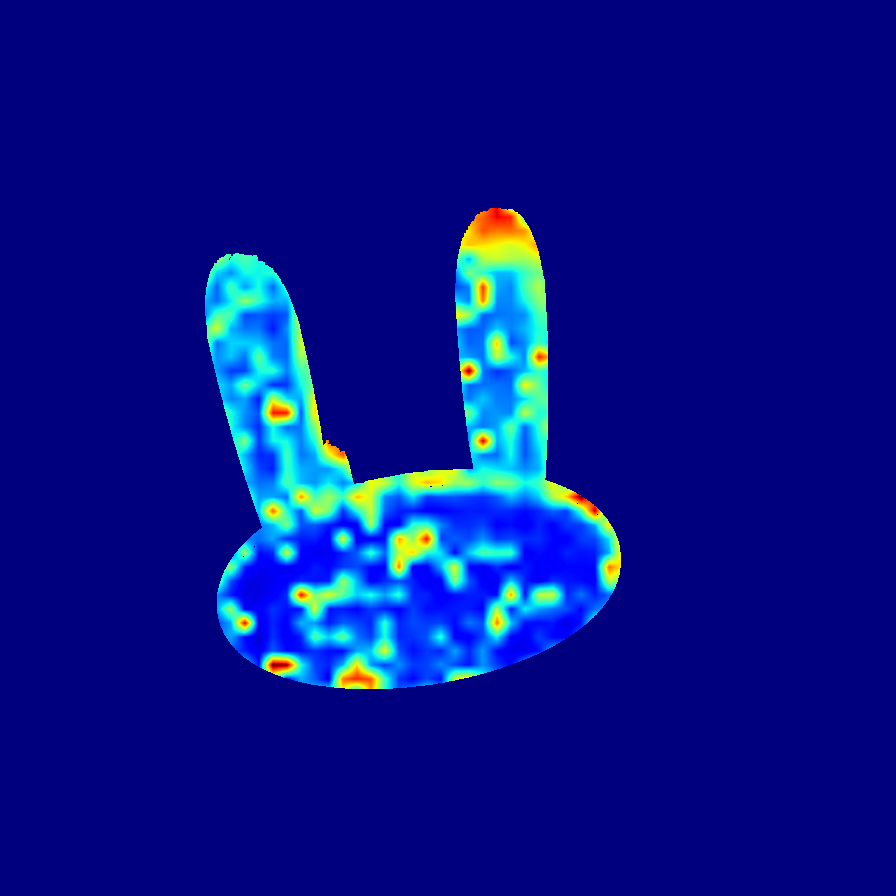} & 
            \includegraphics[width=0.08\linewidth,angle=180,origin=c]{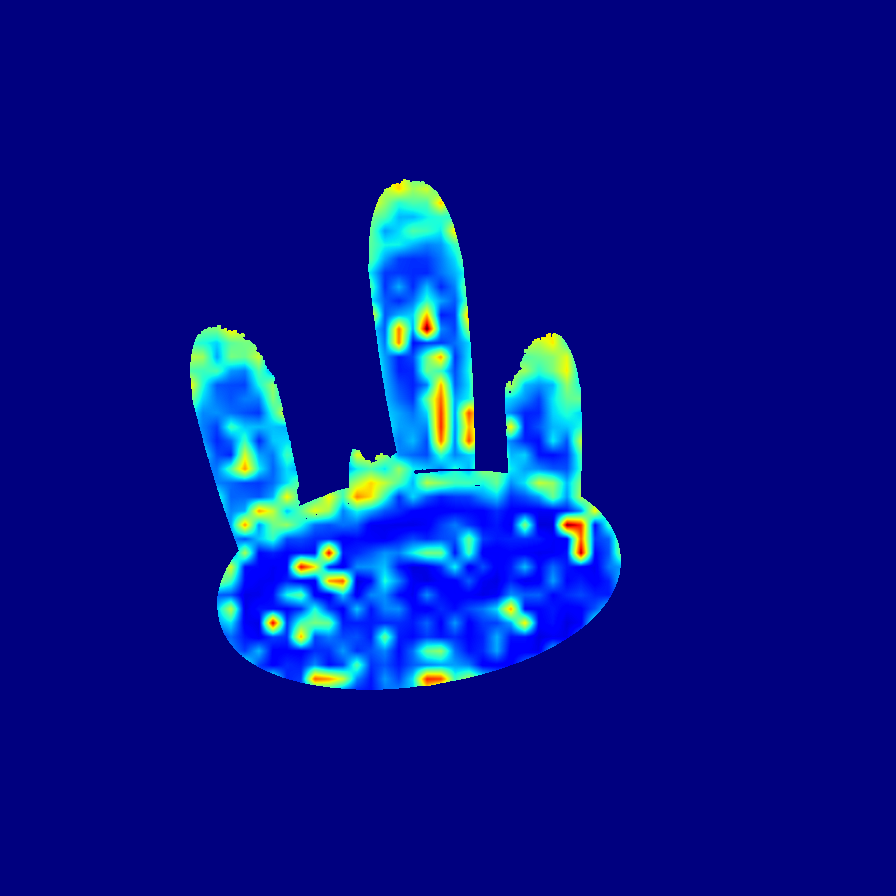} & 
            \includegraphics[width=0.08\linewidth,angle=180,origin=c]{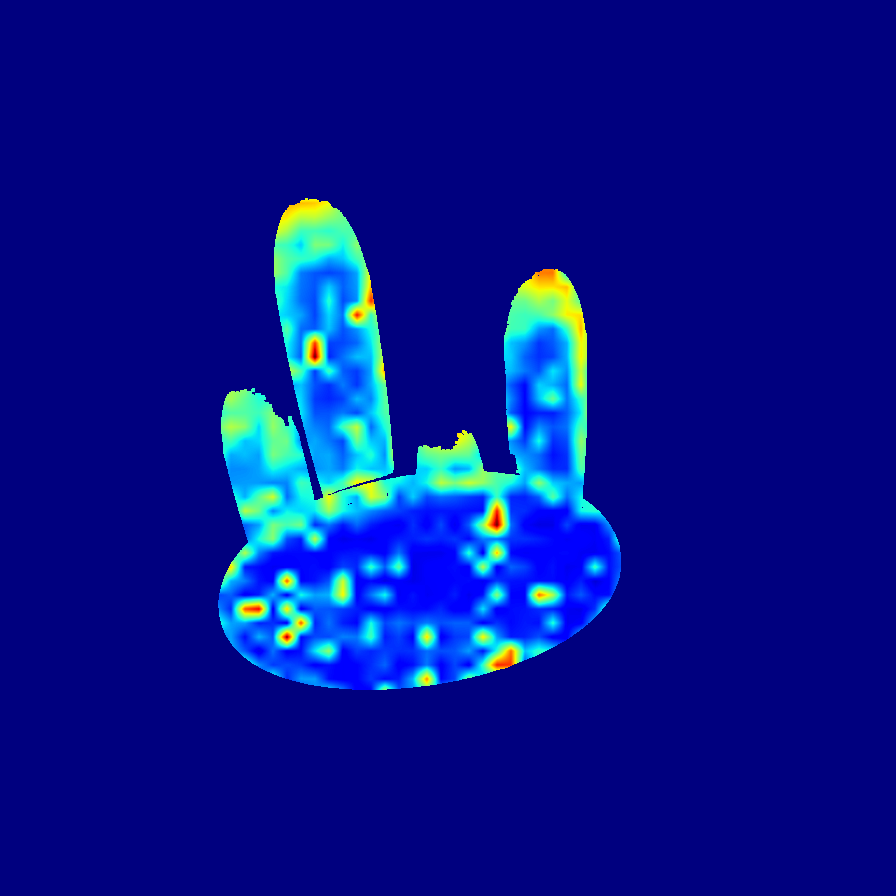} & 
            \includegraphics[width=0.08\linewidth,angle=180,origin=c]{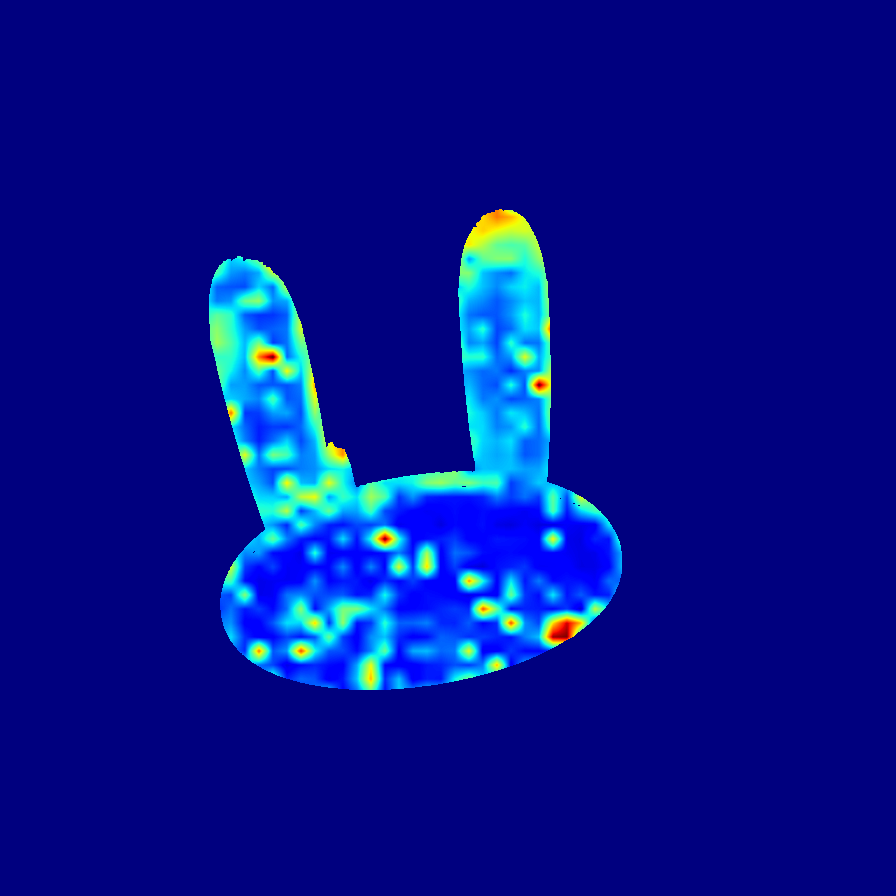} &
            \includegraphics[width=0.08\linewidth,angle=180,origin=c]{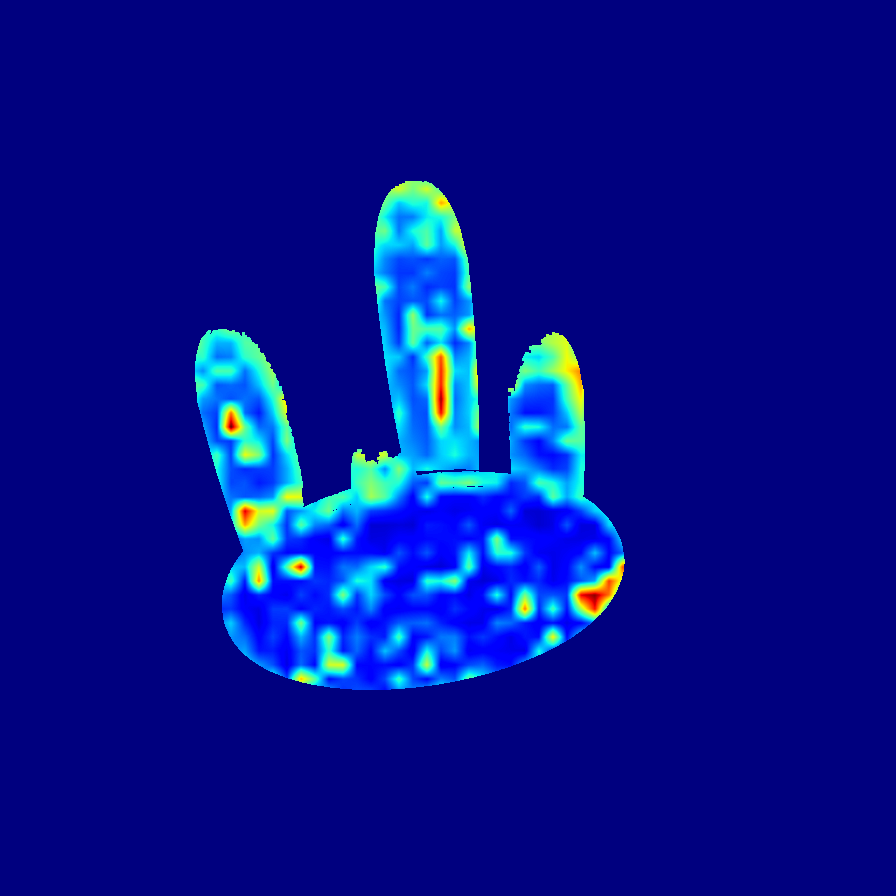} \\

            & \rotatebox{90}{\quad Ens.} &
            \includegraphics[width=0.08\linewidth,angle=180,origin=c]{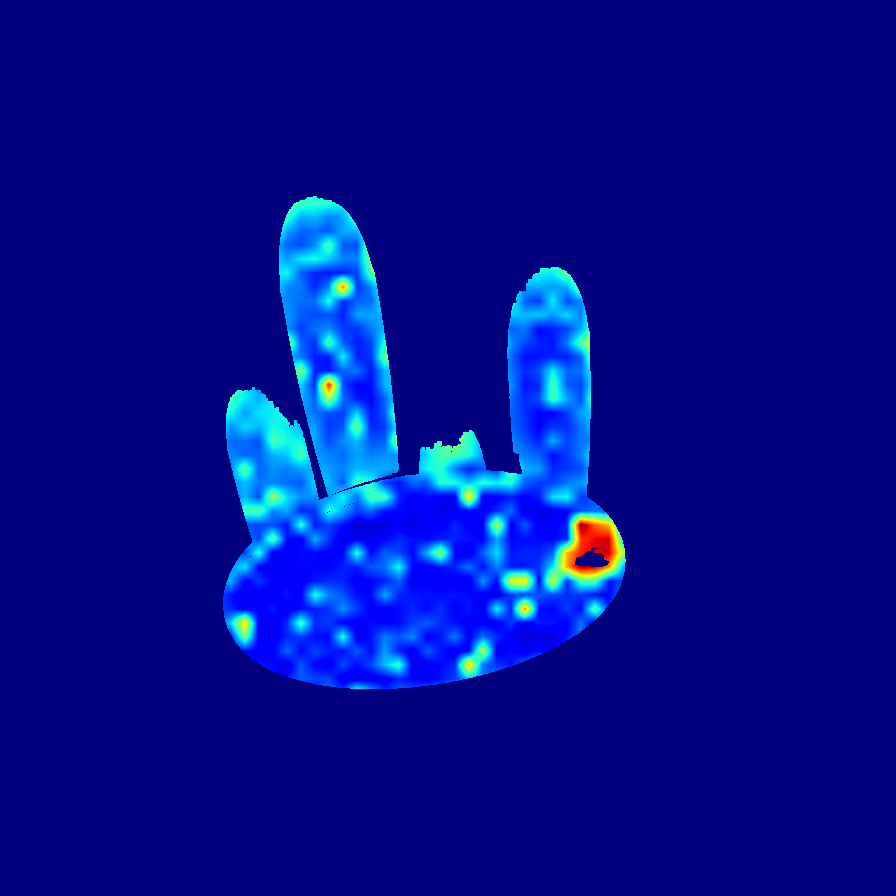} & 
            \includegraphics[width=0.08\linewidth,angle=180,origin=c]{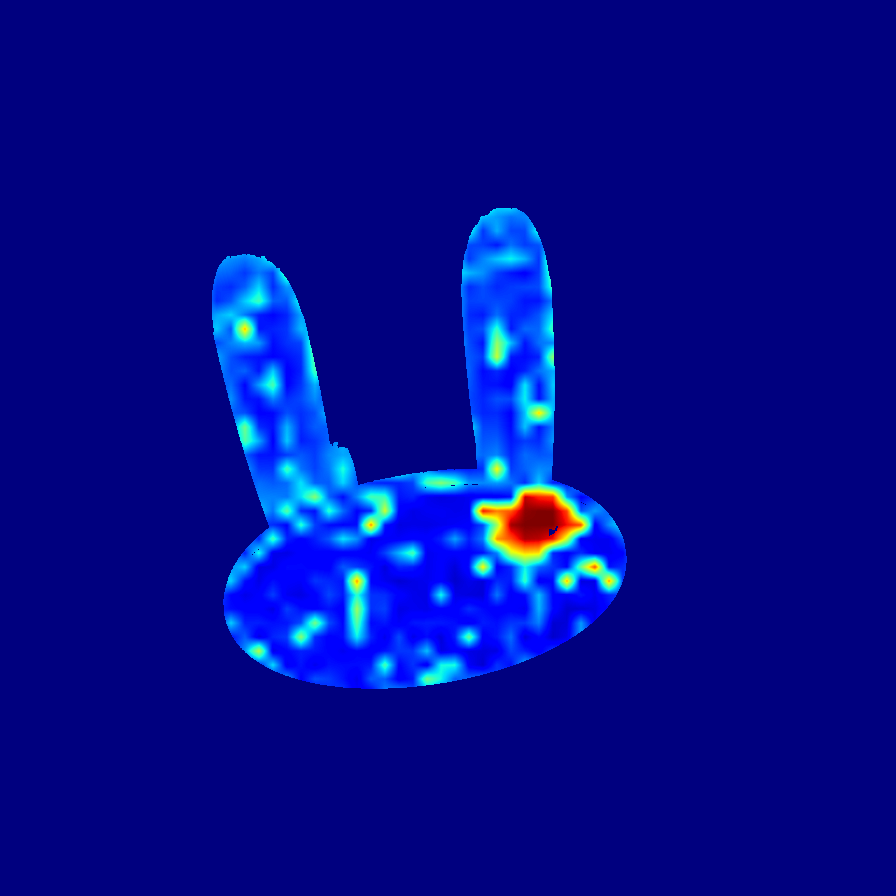} & 
            \includegraphics[width=0.08\linewidth,angle=180,origin=c]{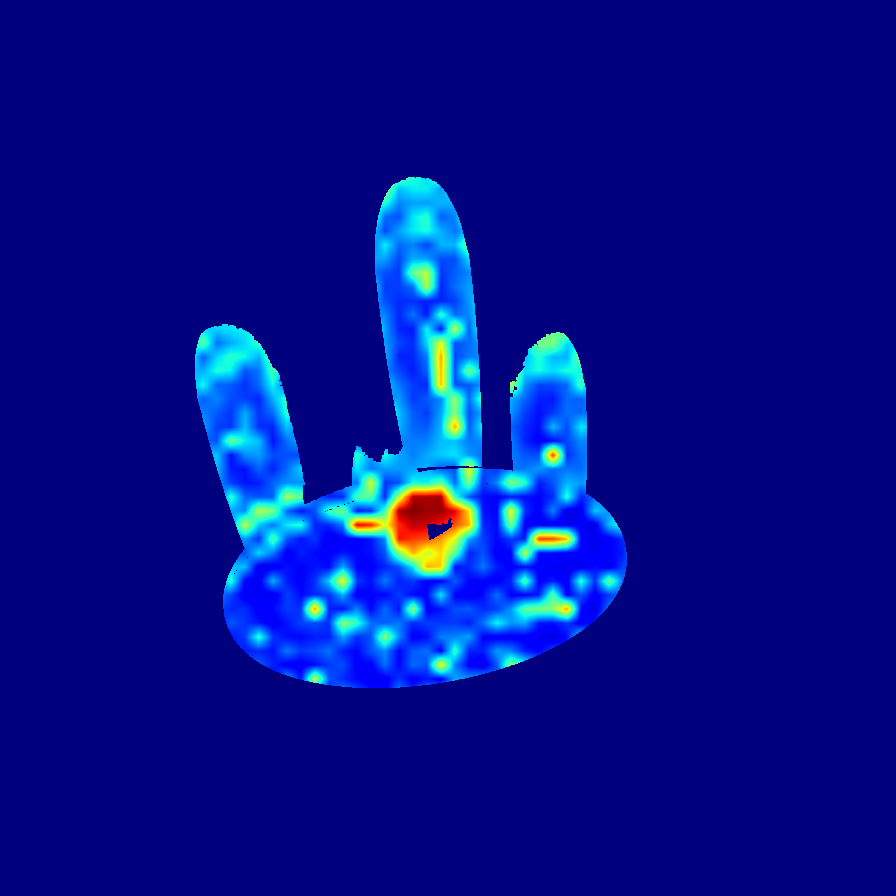} & 
            \includegraphics[width=0.08\linewidth,angle=180,origin=c]{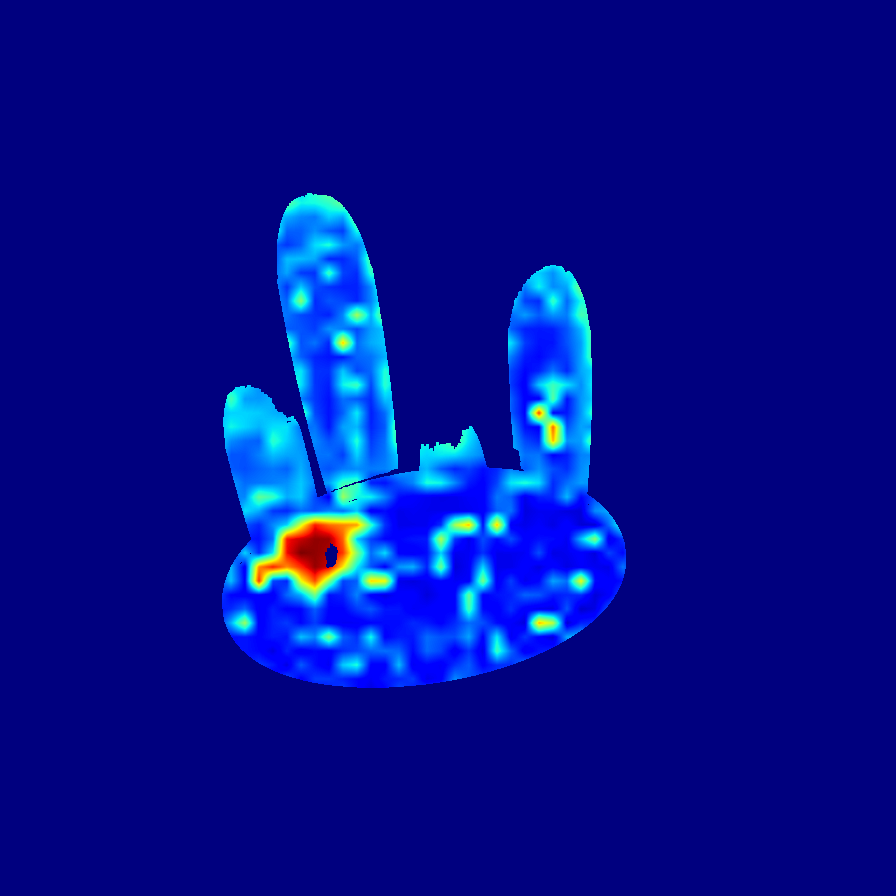} & 
            \includegraphics[width=0.08\linewidth,angle=180,origin=c]{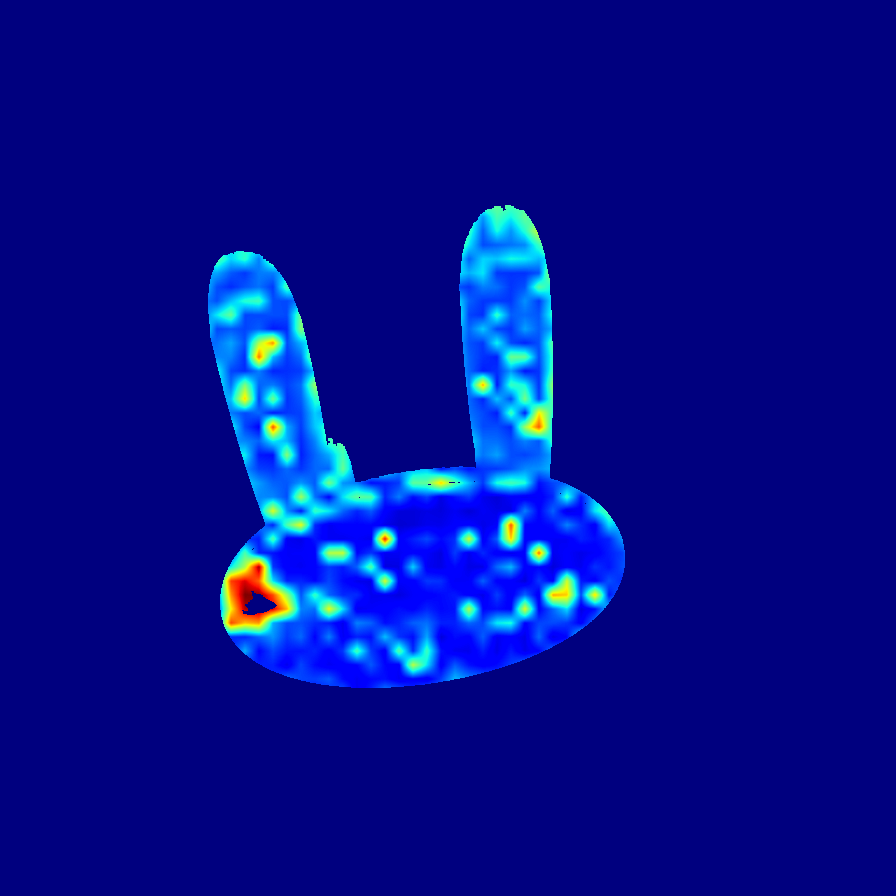} & 
            \includegraphics[width=0.08\linewidth,angle=180,origin=c]{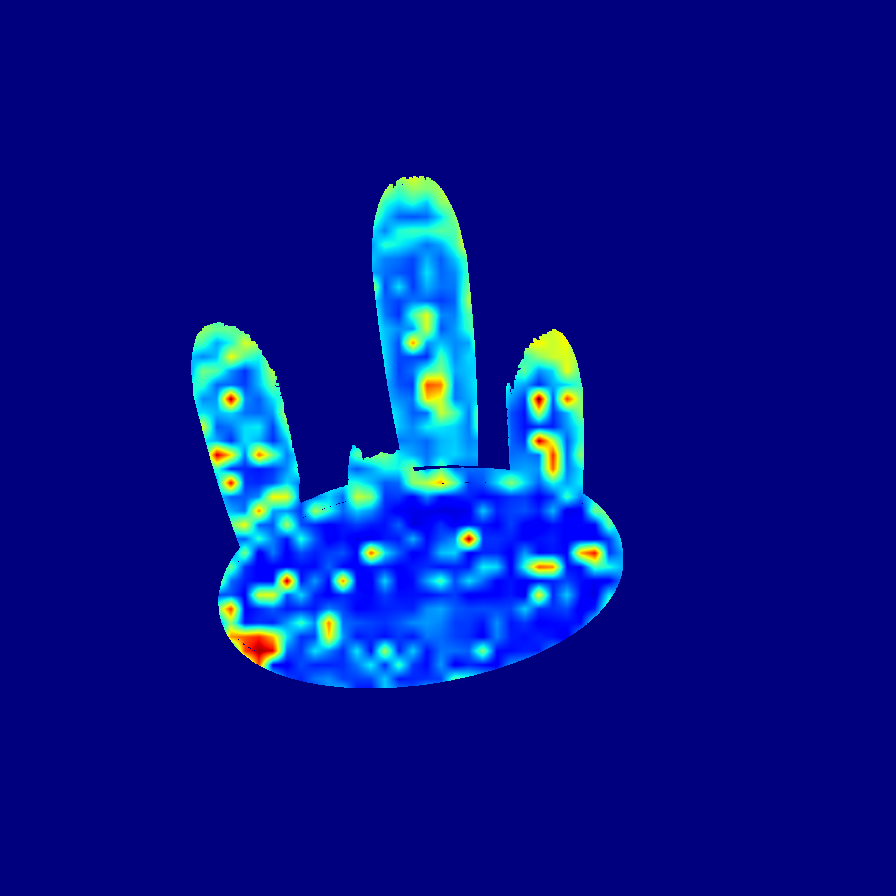} & 
            \includegraphics[width=0.08\linewidth,angle=180,origin=c]{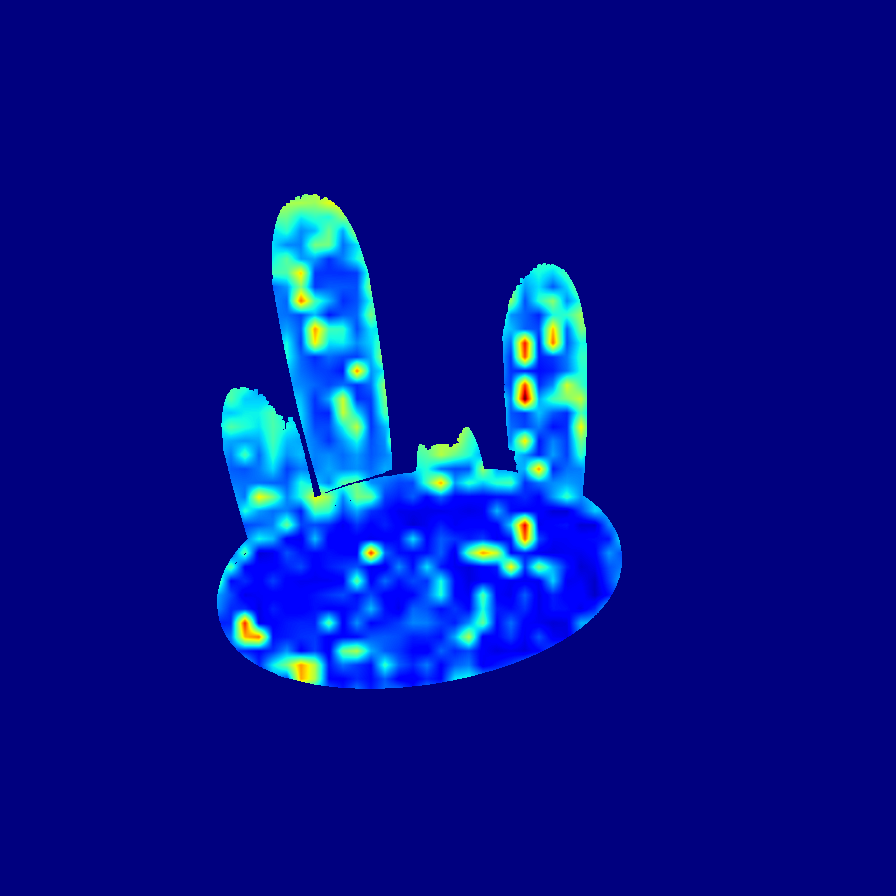} & 
            \includegraphics[width=0.08\linewidth,angle=180,origin=c]{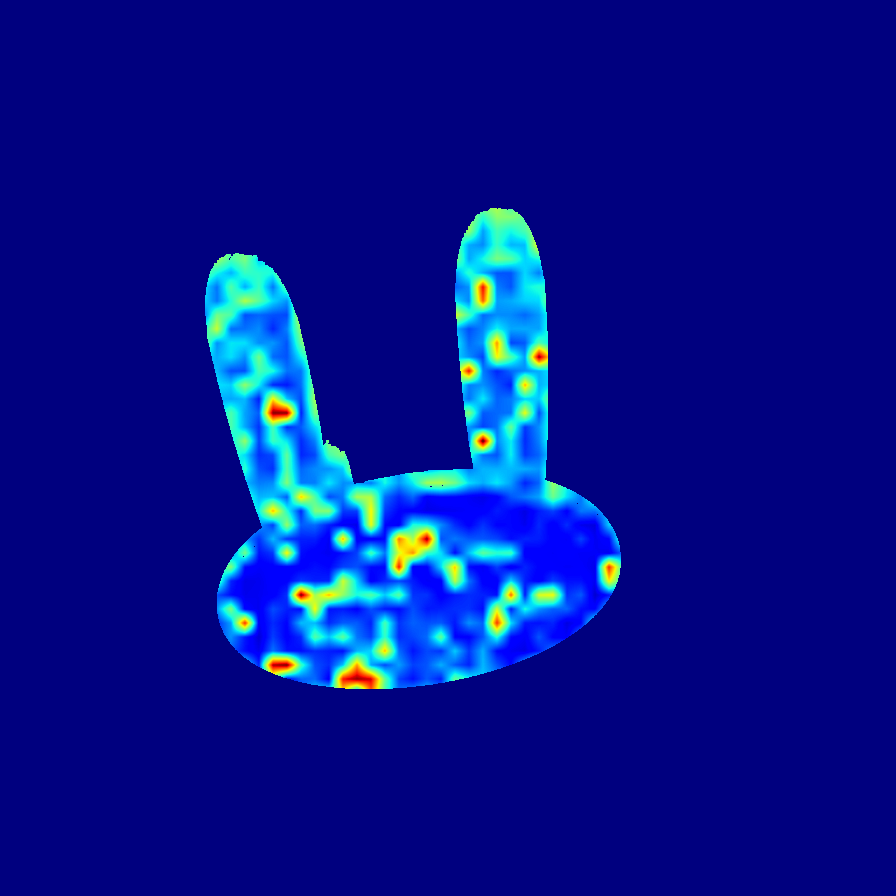} & 
            \includegraphics[width=0.08\linewidth,angle=180,origin=c]{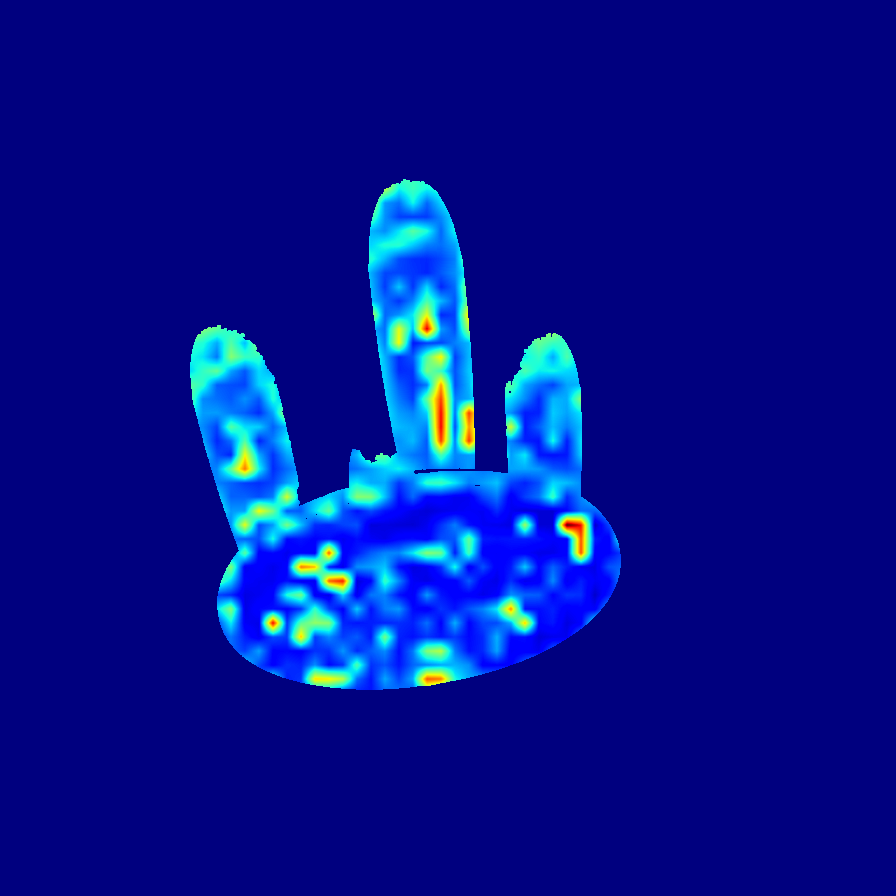} & 
            \includegraphics[width=0.08\linewidth,angle=180,origin=c]{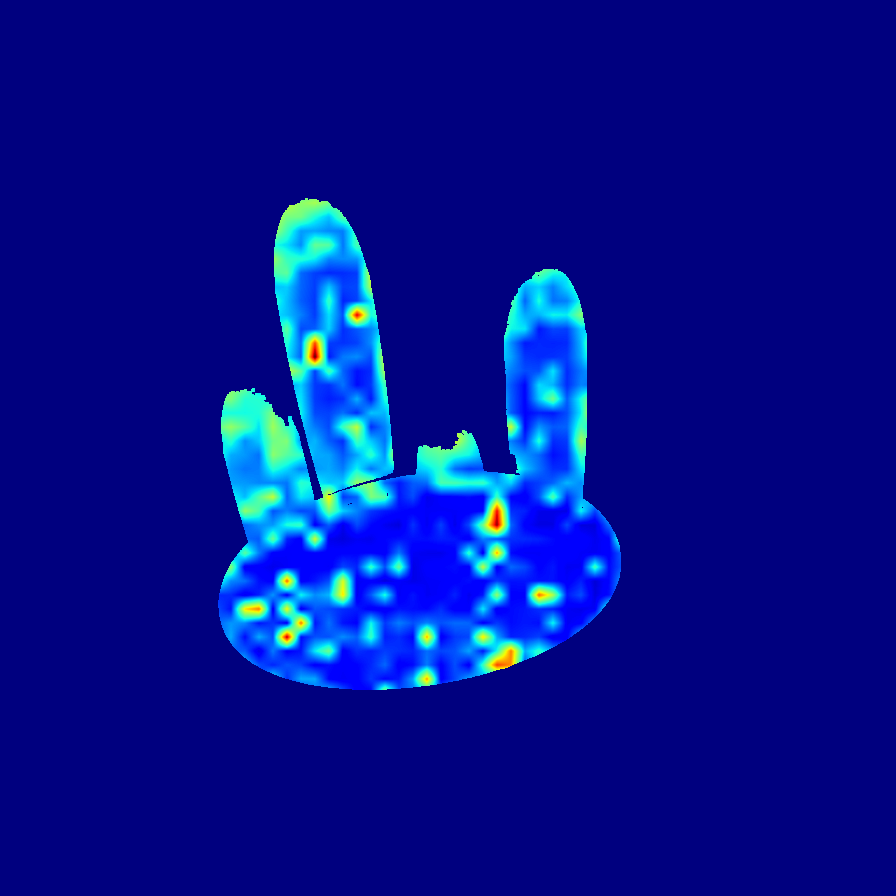} & 
            \includegraphics[width=0.08\linewidth,angle=180,origin=c]{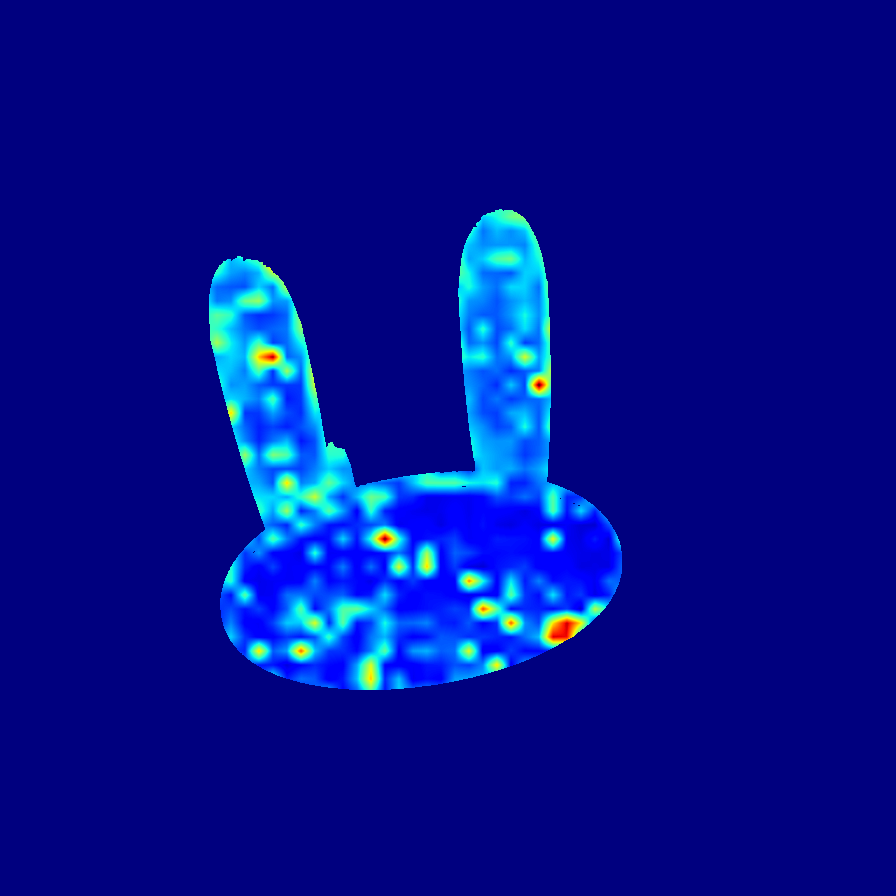} & 
            \includegraphics[width=0.08\linewidth,angle=180,origin=c]{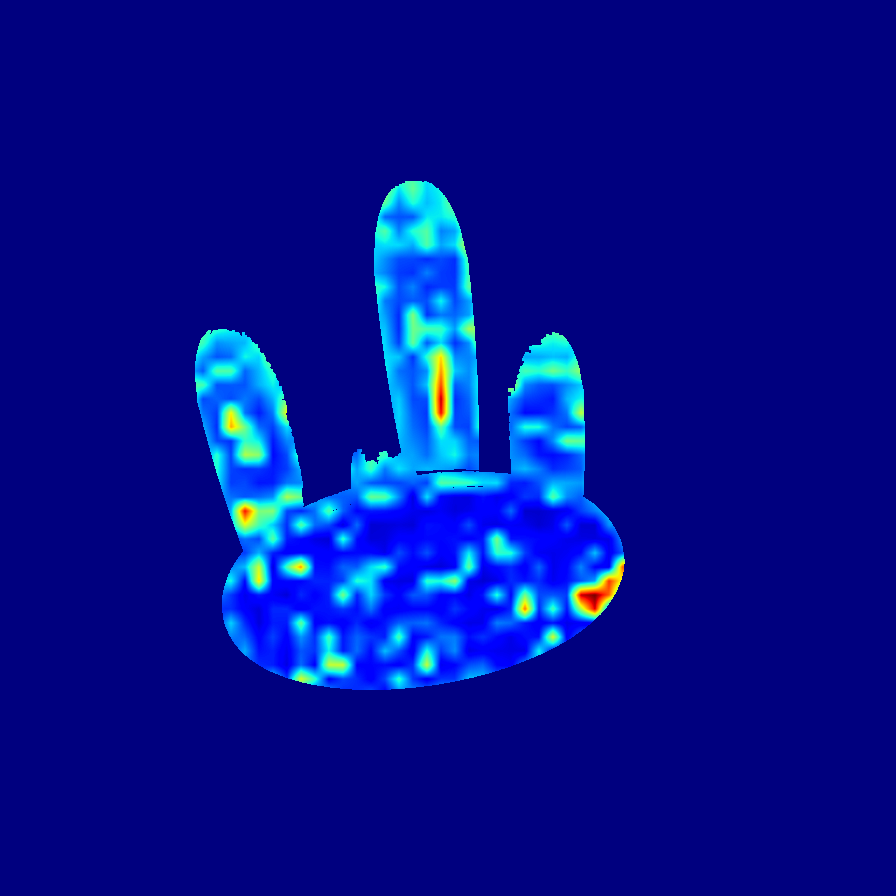} \\
            
        \end{tabular}
    }
    \caption{
        \textbf{Depth-to-Image Cross-Views.}
        }
    \label{fig:all_views_depth2image}
\end{figure*}

        \paragraph{Cross-View Maps}
            We report in~\cref{fig:all_views_image2depth} and~\cref{fig:all_views_depth2image} all the $N \times N$ cross-view visualisations from a test sample of the \emph{Plastic Stool} class from SiM3D.

\end{document}